%% file: main.tex
\documentclass[sigconf, pbalance]{acmart}

\usepackage{booktabs} %

\usepackage[ruled]{algorithm2e} %

\SetAlFnt{\small}
\SetAlCapFnt{\small}
\SetAlCapNameFnt{\small}
\SetAlCapHSkip{0pt}

\acmJournal{TOG}

\usepackage{cleveref}
\usepackage{algorithmic}
\usepackage[export]{adjustbox}
\usepackage{pgfplots}
\usepackage{overpic}
\usepackage{colortbl}

\input{macros.tex}

\newcommand\blfootnote[1]{%
  \begingroup
  \renewcommand\thefootnote{}\footnote{#1}%
  \addtocounter{footnote}{-1}%
  \endgroup
}

\copyrightyear{2024}
\acmYear{2024}
\setcopyright{rightsretained}
\acmConference[SIGGRAPH Conference Papers '24]{Special Interest Group on Computer Graphics and Interactive Techniques Conference Conference Papers '24}{July 27-August 1, 2024}{Denver, CO, USA}
\acmBooktitle{Special Interest Group on Computer Graphics and Interactive Techniques Conference Conference Papers '24 (SIGGRAPH Conference Papers '24), July 27-August 1, 2024, Denver, CO, USA}
\acmDOI{10.1145/3641519.3657430}
\acmISBN{979-8-4007-0525-0/24/07}

\begin{document}
\title{The Chosen One: Consistent Characters in Text-to-Image Diffusion Models}

\author{Omri Avrahami}
\orcid{0000-0002-7628-7525}
\affiliation{
 \institution{The Hebrew University of Jerusalem \\ Google Research}
 \city{Jerusalem}
 \country{Israel}}
\email{omri.avrahami@mail.huji.ac.il}
\authornote{Performed this work while working at Google}

\author{Amir Hertz}
\orcid{0000-0003-3037-3556}
\affiliation{
 \institution{Google Research}
 \city{Tel Aviv}
 \country{Israel}}
\email{hertzamir@gmail.com}

\author{Yael Vinker}
\orcid{0000-0003-4402-7267}
\affiliation{
 \institution{Tel Aviv University \\ Google Research}
 \city{Tel Aviv}
 \country{Israel}}
\email{yaelvi116@gmail.com}
\authornotemark[1]

\author{Moab Arar}
\orcid{0000-0001-8423-3538}
\affiliation{
 \institution{Tel Aviv University \\ Google Research}
 \city{Tel Aviv}
 \country{Israel}}
\email{moab.arar@gmail.com}
\authornotemark[1]

\author{Shlomi Fruchter}
\orcid{0009-0001-1797-2143}
\affiliation{
 \institution{Google Research}
 \city{Tel Aviv}
 \country{Israel}}
\email{shlomi.fruchter@gmail.com}

\author{Ohad Fried}
\orcid{0000-0001-7109-4006}
\affiliation{
 \institution{Reichman University}
 \city{Herzliya}
 \country{Israel}}
\email{ofried@runi.ac.il}

\author{Daniel Cohen-Or}
\orcid{0000-0001-6777-7445}
\affiliation{
 \institution{Tel Aviv University \\ Google Research}
 \city{Tel Aviv}
 \country{Israel}}
\email{cohenor@gmail.com}
\authornotemark[1]

\author{Dani Lischinski}
\orcid{0000-0002-6191-0361}
\affiliation{
 \institution{The Hebrew University of Jerusalem \\ Google Research}
 \city{Jerusalem}
 \country{Israel}}
\email{danix@mail.huji.ac.il}
\authornotemark[1]

\renewcommand\shortauthors{Avrahami et al.}

\input{sections/abstract.tex}



\input{figures/teaser/fig.tex}

\maketitle

\blfootnote{Project page is available at: \textcolor{red}{\href{https://omriavrahami.com/the-chosen-one/}{https://omriavrahami.com/the-chosen-one/}}}

\input{sections/introduction.tex}
\input{sections/related_work.tex}
\input{sections/method.tex}
\input{sections/experiments.tex}
\input{sections/limitations_and_conclusions.tex}

\input{figures/user_control/fig.tex}

\clearpage
\bibliographystyle{ACM-Reference-Format}
\bibliography{egbib}

\appendix
\begin{acks}
    We thank Yael Pitch, Matan Cohen, Neal Wadhwa and Yaron Brodsky for their valuable help and feedback.
\end{acks}
\input{sections/appendices/additional_experiments.tex}
\input{sections/appendices/implmentation_details.tex}
\input{sections/appendices/societal_impact.tex}

\end{document}

%% file: macros.tex
\newcommand{\ignorethis}[1]{}

\newcommand{\etal       }     {{et~al.}}

\newcommand{\eg         }     {{e.g.}}

\newcommand{\ie         }     {{i.e.}}
\newcommand{\etc         }     {{etc.}}
\newcommand{\naive      }     {{na\"{\i}ve}}
\newcommand{\Naive      }     {{Na\"{\i}ve}}

\newcommand{\Reals      }     {{\textrm{I\kern-0.18em R}}}

\newcommand{\change     } [1] {\mbox{{\footnotesize $\Delta$} \kern-3pt}#1}

\definecolor{darkred}{rgb}{0.7,0.1,0.1}
\definecolor{darkgreen}{rgb}{0.1,0.6,0.1}
\definecolor{cyan}{rgb}{0.7,0.0,0.7}
\definecolor{otherblue}{rgb}{0.1,0.4,0.8}
\definecolor{maroon}{rgb}{0.76,.13,.28}
\definecolor{burntorange}{rgb}{0.81,.33,0}

\newif\ifdraft
\drafttrue

\ifdraft
  \newcommand{\todo}[1]{{\color{cyan}[\textbf{TODO:} #1]}}
  \newcommand{\omri}[1]{{\color{burntorange}[\textbf{Omri:} #1]}}
  \newcommand{\yael}[1]{{\color{darkred}[\textbf{Yael:} #1]}}
  \newcommand{\ohad}[1]{{\color{magenta}[\textbf{Ohad:} #1]}}
  
  \newcommand{\danix}[1]{{\color{blue}[\textbf{Danix:} #1]}}

  \newcommand{\dcc}[1]{{\color{red}#1}}

\else
  \newcommand{\omri}[1]{}
   \newcommand {\dcc}[1]{}
  \newcommand{\yael}[1]{}
  \newcommand{\ohad}[1]{}
  \newcommand{\dcor}[1]{}
  \newcommand{\danix}[1]{}
  \newcommand {\note}[1]{}
  \newcommand {\todo}[1]{}

\fi

\newcommand\todosilent[1]{}

\providecommand{\keywords}[1]
{
  \textbf{\textit{Keywords---}} #1
}

\DeclareMathOperator*{\argmin}{argmin}

\newcommand{\expnumber}[2]{{#1}\mathrm{e}{#2}}

\newlength{\ww}

\definecolor{colora}{RGB}{242,78,112}
\newcommand {\colora}[1]{{\color{colora}{#1}}}
\definecolor{colorb}{RGB}{255,176,1}
\newcommand {\colorb}[1]{{\color{colorb}{#1}}}
\definecolor{colorc}{RGB}{41,160,177}
\newcommand {\colorc}[1]{{\color{colorc}{#1}}}
\definecolor{colord}{RGB}{55, 146, 55}
\newcommand {\colord}[1]{{\color{colord}{#1}}}

\newcommand{\model}{M}
\newcommand{\crep}{{\Theta}}
\newcommand{\optrep}{\crep(p)}
\newcommand{\dminc}{d_\textit{min-c}}
\newcommand{\dtgtc}{d_\textit{size-c}}
\newcommand{\dconv}{d_\textit{conv}}
\newcommand{\diter}{d_\textit{iter}}
\newcommand{\ccand}{c_\textit{cohesive}}
\newcommand{\ccen}{c_\textit{cen}}
\newcommand{\Lrec}{\mathcal{L}_\textit{rec}}
\newcommand{\DALLE}{{DALL$\cdot$E}}

%% file: sections/abstract.tex
\begin{abstract}
     Recent advances in text-to-image generation models have unlocked vast potential for visual creativity. However, the users that use these models struggle with the generation of consistent characters, a crucial aspect for numerous real-world applications such as story visualization, game development, asset design, advertising, and more. Current methods typically rely on multiple pre-existing images of the target character or involve labor-intensive manual processes. In this work, we propose a fully automated solution for consistent character generation, with the sole input being a text prompt. We introduce an iterative procedure that, at each stage, identifies a coherent set of images sharing a similar identity and extracts a more consistent identity from this set. Our quantitative analysis demonstrates that our method strikes a better balance between prompt alignment and identity consistency compared to the baseline methods, and these findings are reinforced by a user study. To conclude, we showcase several practical applications of our approach.
\end{abstract}

%% file: figures/teaser/fig.tex
\begin{teaserfigure}
    \centering
    \includegraphics[width=\linewidth]{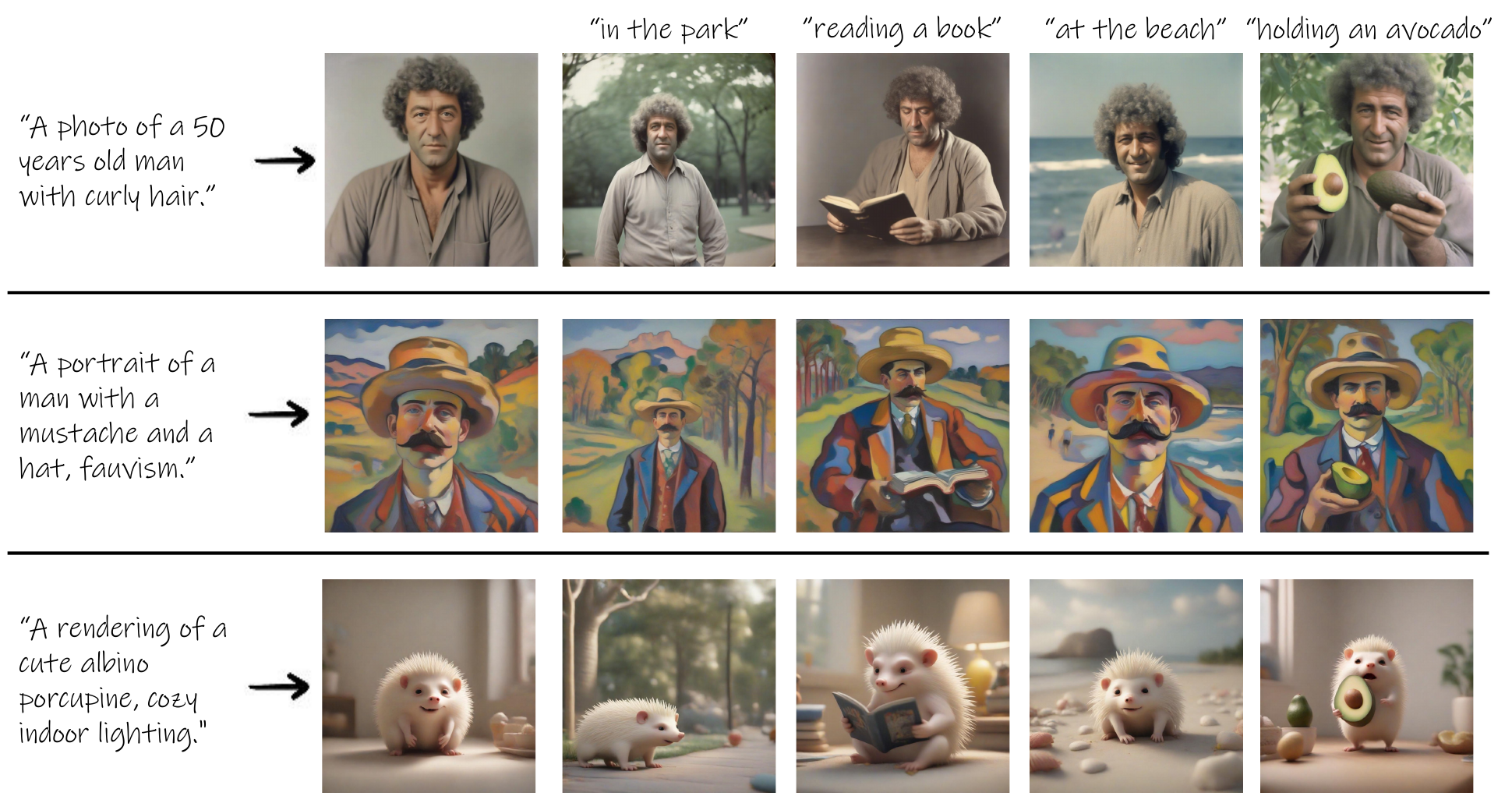} 
    \caption{\textbf{The Chosen One:} Given a text prompt describing a character, our method distills a representation that enables consistent depiction of \emph{the same character} in novel contexts.}
    \label{fig:teaser}
\end{teaserfigure}

%% file: sections/introduction.tex
\section{Introduction}
\label{sec:introduction}

\input{figures/motivation/fig.tex}

The ability to maintain consistency of generated visual content across various contexts, as shown in \Cref{fig:teaser}, plays a central role in numerous creative endeavors. These include illustrating a book, crafting a brand, creating comics, developing presentations, designing webpages, and more. Such consistency serves as the foundation for establishing brand identity, facilitating storytelling, enhancing communication, and nurturing emotional engagement.

Despite the increasingly impressive abilities of text-to-image generative models, the users that use these models struggle with such consistent generation, a shortcoming that we aim to rectify in this work. 
Specifically, we introduce the task of \emph{consistent character generation}, where given an input text prompt describing a character, we derive a representation that enables generating consistent depictions of the same character in novel contexts. Although we refer to characters throughout this paper, our work is in fact applicable to visual subjects in general.

Consider, for example, an illustrator working on a Plasticine cat character. As demonstrated in \Cref{fig:motivation}, providing a state-of-the-art text-to-image model with a prompt describing the character, results in a variety of outcomes, which may lack consistency (top row). In contrast, in this work we show how to distill a consistent representation of the cat (2nd row), which can then be used to depict the \emph{same character} in a multitude of different contexts.

The widespread popularity of text-to-image generative models \cite{Podell2023SDXLIL,Saharia2022PhotorealisticTD,Rombach2021HighResolutionIS,ramesh2022hierarchical}, combined with the need for consistent character generation, has already 
spawned a variety of ad hoc solutions. These include, for example, using celebrity names in prompts~\cite{celebrities_trick} for creating consistent humans, or using image variations~\cite{ramesh2022hierarchical} and filtering them manually by similarity \cite{consistent_generation_tricks}. In contrast to these ad hoc, manually intensive solutions, we propose a fully-automatic principled approach to consistent character generation.

The academic works most closely related to our setting are ones dealing with personalization~\cite{Gal2022AnII,Ruiz2022DreamBoothFT} and story generation~\cite{Rahman2022MakeAStoryVM, Jeong2023ZeroshotGO, Gong2023TaleCrafterIS}. Some of these methods derive a representation for a given character from \emph{several} user-provided images~\cite{Gal2022AnII,Ruiz2022DreamBoothFT,Gong2023TaleCrafterIS}. Others cannot generalize to novel characters that are not in the training data~\cite{Rahman2022MakeAStoryVM}, or rely on textual inversion of an existing depiction of a human face~\cite{Jeong2023ZeroshotGO}.

In this work, we argue that in many applications the goal is to generate \emph{some} consistent character, rather than visually matching a specific appearance. Thus, we address a new setting, where we aim to automatically distill a consistent representation of a character that is only required to comply with a single natural language description. Our method does not require \emph{any} images of the target character as input; thus, it enables creating a \emph{novel} consistent character that does not necessarily resemble any existing visual depiction.

Our fully-automated solution to the task of consistent character generation is based on the assumption that a sufficiently large set of generated images, for a certain prompt, will contain groups of images with shared characteristics. 
Given such a cluster, one can extract a representation that captures the ``common ground'' among its images.
Repeating the process with this representation, we can increase the consistency among the generated images, while still remaining faithful to the original input prompt.

We start by generating a gallery of images based on the provided text prompt, and embed them in a Euclidean space using a pre-trained feature extractor. Next, we cluster these embeddings, and choose the most \emph{cohesive} cluster to serve as the input for a personalization method that attempts to extract a consistent identity. We then use the resulting model to generate the next gallery of images, which should exhibit more consistency, while still depicting the input prompt. This process is repeated iteratively until convergence.

We evaluate our method quantitatively and qualitatively against several baselines, as well as conducting a user study. Finally, we present several applications of our method.

In summary, our contributions are: (1) we formalize the task of consistent character generation, (2) propose a novel solution to this task, and (3) we evaluate our method quantitatively and qualitatively, in addition to a user study, to demonstrate its effectiveness.

%% file: figures/motivation/fig.tex
\begin{figure}[t]
    \centering

    \includegraphics[width=1\linewidth]{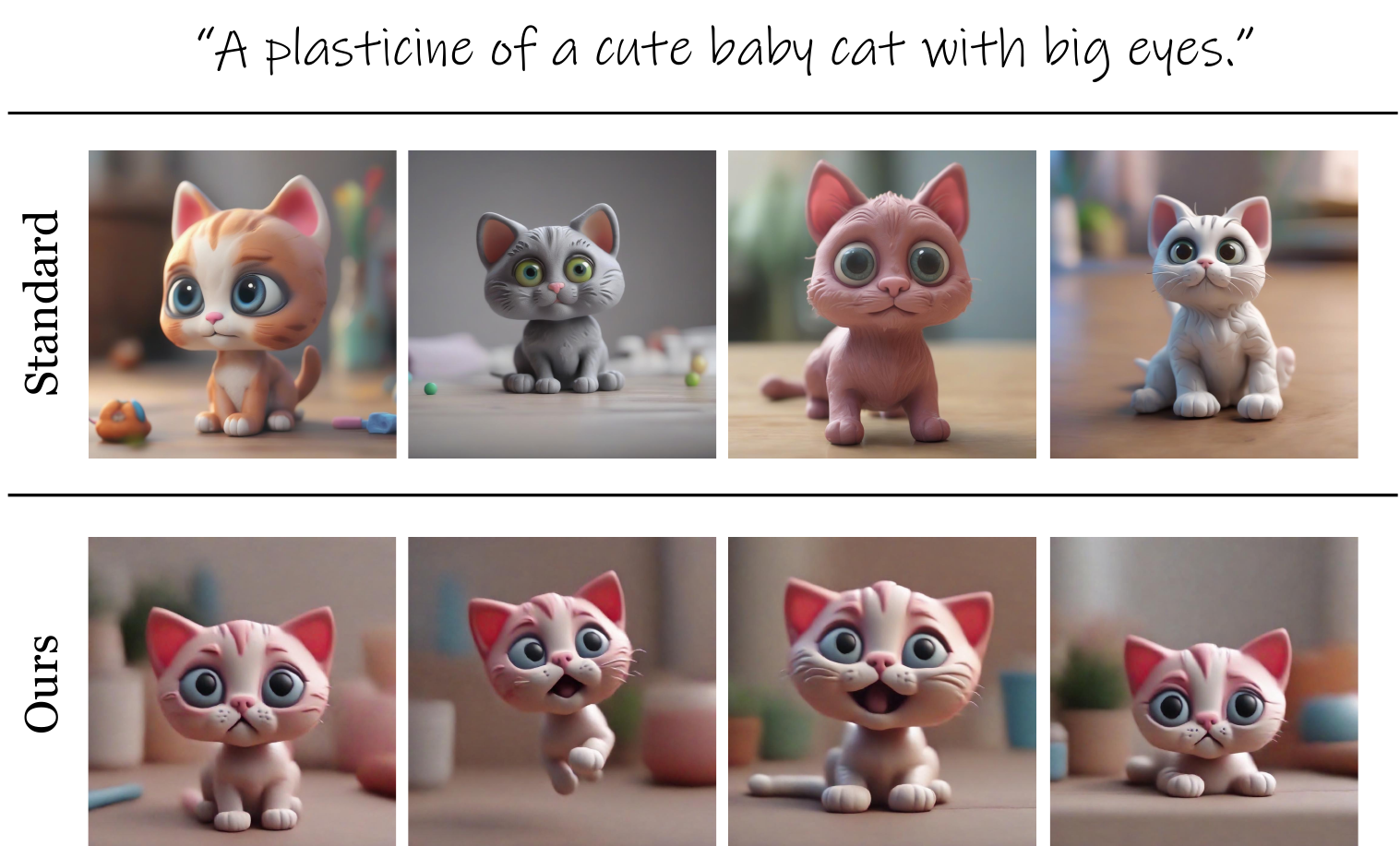}
    
    \caption{\textbf{Identity consistency.} Given the prompt \textit{``a Plasticine of a cute baby cat with big eyes''}, a standard text-to-image diffusion model produces different cats (all corresponding to the input text), whereas our method produces the \emph{same cat}.}
    \label{fig:motivation}
\end{figure}

%% file: sections/related_work.tex
\section{Related Work}
\label{sec:related_work}

\paragraph{Text-to-image generation.}
Text conditioned image generative models (T2I)~\cite{ramesh2022hierarchical,Rombach2021HighResolutionIS, yu2022scaling} show unprecedented capabilities of generating high quality images from mere natural language text descriptions. They are quickly becoming a fundamental tool for any creative vision task. In particular, text-to-image diffusion models~\cite{sohl2015deep,song2019generative,ho2020denoising,song2020denoising, nichol2021glide, Balaji2022eDiffITD} are employed for guided image synthesis~\cite{hertz2022prompt,voynov2022sketch, spatext_2023_CVPR, zhang2023controlnet,mou2023t2i, Chefer2023AttendandExciteAS, Ge2023ExpressiveTG, Couairon2023ZeroshotSL} and image editing tasks~\cite{meng2021sdedit, blended_2022_CVPR, avrahami2023blendedlatent, mokady2022null, pnpDiffusion2022, hertz2023delta, Kawar2022ImagicTR, cao2023masactrl, Patashnik2023LocalizingOS, bar2022text2live, sheynin2022knn}. Using image editing methods, one can edit an image of a given character, and change its pose, \etc, however, these methods cannot ensure consistency of the character in novel contexts, as our problem dictates.

In addition, diffusion models were used in other tasks \cite{Po2023StateOT, Zhang2023TexttoimageDM}, such as: video editing \cite{liu2023video, qi2023fatezero, Molad2023DreamixVD, geyer2023tokenflow, Yang2023RerenderAV, Liu2023VideoP2PVE}, 3D synthesis \cite{poole2022dreamfusion, metzer2023latent, Hllein2023Text2RoomET, Fridman2023SceneScapeTC}, editing \cite{Benaim2022VolumetricDF, Gordon2023BlendedNeRFZO, Zhuang2023DreamEditorT3, Sella2023VoxETV} and texturing \cite{Richardson2023TEXTureTT}, typography generation \cite{Iluz2023WordAsImageFS}, motion generation \cite{Raab2023SingleMD, Tevet2022HumanMD}, and solving inverse problems~\cite{Horwitz2022ConffusionCI}.

\paragraph{Text-to-image personalization.}
Text-conditioned models cannot generate an image of a specific object or character. To overcome this limitation, a line of works utilizes \emph{several} images of the same instance to encapsulate new priors in the generative model. Existing solutions range from optimization of text tokens~\cite{Gal2022AnII, Voynov2023PET, Vinker2023ConceptDF} to fine-tuning the parameters of the entire model~\cite{Ruiz2022DreamBoothFT, Avrahami2023BreakASceneEM}, where in the middle, recent works suggest fine-tuning a small subset of parameters~\cite{lora,lora_diffusion, Kumari2022MultiConceptCO, Han2023SVDiffCP, Tewel2023KeyLockedRO, Alaluf2023ANS, Chen2023AnyDoorZO}. Models trained in this manner can generate consistent images of the same subject. However, they typically require a \emph{collection} of images depicting the subject, which naturally narrows their ability to generate any imaginary character. Moreover, when training on a single input image~\cite{Avrahami2023BreakASceneEM}, these methods tend to overfit and produce similar images with minimal diversity during inference. 

Unlike previous works, our method does not require an input image; instead, it can generate consistent and diverse images of the same character based only on a text description. Additional works are aimed to bypass the personalization training by introducing a dedicated personalization encoder \cite{gal2023encoder, Wei2023ELITEEV, Chen2023SubjectdrivenTG, Jia2023TamingEF, Shi2023InstantBoothPT,Li2023BLIPDiffusionPS, Ye2023IPAdapterTC, arar2023domain, Valevski2023Face0IC}. Given an image and a prompt, these works can produce images with a character similar to the input. However, as shown in \Cref{sec:comparisons}, they lack consistency when generating multiple images from the same input. Concurrently, ConceptLab~\cite{richardson2023conceptlab} is able to generate new members of a broad \emph{category} (\eg, a new pet); in contrast, we seek a consistent \emph{instance} of a character described by the input text prompt. Another line of works, focuses on learning styles \cite{Sohn2023StyleDropTG, Ahn2023DreamStylerPB} from a reference image. On the other hand, our work focuses on generating novel consistent characters rather than styles.

\paragraph{Story visualization.}
Consistent character generation is well studied in the field of story visualization. Early GAN works \cite{li2018storygan,szHucs2022modular} employ a story discriminator for the image-text alignment. Recent works, such as StoryDALL-E \cite{maharana2022storydall} and Make-A-Story \cite{Rahman2022MakeAStoryVM} utilize pre-trained T2I models for the image generation, while an adapter model is trained to embed story captions and previous images into the T2I model.
However, those methods cannot generalize to novel characters, as they are trained over specific datasets.
More closely related, Jeong \etal~\cite{Jeong2023ZeroshotGO} generate consistent storybooks by combining textual inversion with a face-swapping mechanism; therefore, their work relies on images of existing human-like characters.
TaleCrafter \cite{Gong2023TaleCrafterIS} presents a comprehensive pipeline for storybook visualization. However, their consistent character module is based on an existing personalization method that requires fine-tuning on \emph{several} images of the same character.

\paragraph{Manual methods.}
Other attempts for achieving consistent character generation using a generative model rely on ad hoc and manually-intensive tricks such as using text tokens of a celebrity, or a combination of celebrities \cite{celebrities_trick} in order to create a consistent human; however, the generated characters resemble the original celebrities, and this approach does not generalize to other character types (\eg, animals). Users have also proposed to ensure consistency by manually crafting very long and elaborate text prompts \cite{consistent_generation_tricks}, or by using image variations~\cite{ramesh2022hierarchical} and filtering them manually by similarity \cite{consistent_generation_tricks}. Other users suggested generating a full design sheet of a character, then manually filter the best results and use them for further generation~\cite{design_sheet_trick}. All these methods are manual, labor-intensive, and ad hoc for specific domains (\eg, humans). In contrast, our method is fully automated and domain-agnostic.

%% file: sections/method.tex
\section{Method}
\label{sec:method}

\input{figures/method/fig.tex}

\input{algorithms/consistent_generation.tex}

As stated earlier, our goal in this work is to enable generation of consistent images of a character (or another kind of visual subject) based on a textual description. We achieve this by iteratively customizing a pre-trained text-to-image model, using sets of images generated by the model itself as training data. Intuitively, we refine the representation of the target character by repeatedly funneling the model's output into a consistent identity.
Once the process has converged, the resulting model can be used to generate consistent images of the target character in novel contexts.
In this section, we describe our method in detail.

Formally, we are given a text-to-image model $\model_\crep$, parameterized by $\crep$, and a text prompt $p$ that describes a target character. The parameters $\crep$ consist of a set of model weights $\theta$ and a set of custom text embeddings $\tau$. We seek a representation $\optrep$, s.t., the parameterized model $\model_{\optrep}$ is able to generate consistent images of the character described by $p$ in novel contexts.

Our approach, described in \Cref{alg:consistent_generation} and depicted in \Cref{fig:method}, is based on the premise that a sufficiently large set of images generated by $\model$ for the same text prompt, but with different seeds, will reflect the non-uniform density of the manifold of generated images. Specifically, we expect to find some groups of images with shared characteristics. The ``common ground'' among the images in one of these groups can be used to refine the representation $\optrep$ so as to better capture and fit the target. We therefore propose to iteratively cluster the generated images, and use the most cohesive cluster to refine $\optrep$. This process is repeated, with the refined representation $\optrep$, until convergence.
Below, we describe the clustering and the representation refinement components of our method in detail.

\subsection{Identity Clustering}
\label{sec:identity_clustering}

\input{figures/embedding_visualization/fig.tex}

We start each iteration by using $\model_\crep$, parameterized with the current representation $\crep$, to generate a collection of $N$ images, each corresponding to a different random seed. Each image is embedded in a high-dimensional semantic embedding space, using a feature extractor $F$, to form a set of embeddings $S = \bigcup_N F(\model_\crep(p))$. In our experiments, we use DINOv2 \cite{Oquab2023DINOv2LR} as the feature extractor $F$.

Next, we use the K-MEANS++~\cite{Arthur2007kmeansTA} algorithm to cluster the embeddings of the generated images according to cosine similarity in the embedding space. We filter the resulting collection of clusters $C$ by removing all clusters whose size is below a pre-defined threshold $\dminc$, as it was shown \cite{Avrahami2023BreakASceneEM} that personalization algorithms are prone to overfitting on small datasets. Among the remaining clusters, we choose the most \emph{cohesive} one to serve as the input for the identity extraction stage (see \Cref{fig:embedding_visualization}). We define the cohesion of a cluster $c$ as the average distance between the members of $c$ and its centroid $\ccen$:
\begin{equation}
    \label{eq:cohesion}
    \textit{cohesion}(c) = \frac{1}{|c|} \sum_{e \in c} \Vert e - \ccen \Vert^2.
\end{equation}

In \Cref{fig:embedding_visualization} we show a visualization of the DINOv2 embedding space, where the high-dimensional embeddings $S$ are projected into 2D using t-SNE~\cite{Hinton2002StochasticNE} and colored according to their K-MEANS++~\cite{Arthur2007kmeansTA} clusters. Some of the embeddings are clustered together more tightly than others, and the black cluster is chosen as the most cohesive one.

\subsection{Identity Extraction}
\label{sec:identity_extraction}

Depending on the diversity of the image set generated in the current iteration, the most cohesive cluster $\ccand$ may still exhibit an inconsistent identity, as can be seen in \Cref{fig:method}. The representation $\crep$ is therefore not yet ready for consistent generation, and we further refine it by training on the images in $\ccand$ to extract a more consistent identity. This refinement is performed using text-to-image personalization methods \cite{Gal2022AnII,Ruiz2022DreamBoothFT}, which aim to extract a character from a given set of several images that already depict a \emph{consistent identity}. 
While we apply them to a set of images which are not completely consistent, the fact that these images are chosen based on their semantic similarity to each other, enables these methods to nevertheless distill a common identity from them. This way, our method can overcome the inconsistencies that may emerge due to the feature extractor $F$ or the clustering algorithm.

We base our solution on a pre-trained Stable Diffusion XL (SDXL) \cite{Podell2023SDXLIL} model, which utilizes two text encoders: CLIP \cite{Radford2021LearningTV} and OpenCLIP \cite{OpenCLIP}. We perform textual inversion \cite{Gal2022AnII} to add a new pair of textual tokens $\tau$, one for each of the two text encoders. However, we found that this parameter space is not expressive enough, as demonstrated in \Cref{sec:ablation_study}, hence we also update the model weights $\theta$ via a low-rank adaptation (LoRA) \cite{lora,lora_diffusion} of the self- and cross-attention layers of the model.

We use the standard denoising loss:
\begin{equation}
    \Lrec = \mathbb{E}_{x \sim \ccand, z \sim E(x), \epsilon \sim \mathcal{N}(0, 1), t }\Big[ \Vert \epsilon - \epsilon_{\optrep}(z_{t},t) \Vert_{2}^{2}\Big],
    \label{eq:masked_LDM_loss}
\end{equation}
where $\ccand$ is the chosen cluster, $E(x)$ is the VAE encoder of SDXL, $\epsilon$ is the sample's noise and $t$ is the time step, $z_t$ is the latent $z$ noised to time step $t$. We optimize $\Lrec$ over $\crep = (\theta,\tau)$, the union of the LoRA weights and the newly-added textual tokens.

\subsection{Convergence}
\label{sec:iterative_convergence}

As explained earlier (\Cref{alg:consistent_generation} and \Cref{fig:method}), the above process is performed iteratively.
Note that the representation $\crep$ extracted in each iteration is the one used to generate the set of $N$ images for the next iteration. The generated images are thus funneled into a consistent identity.

Rather than using a fixed number of iterations, we apply a convergence criterion that enables early stopping. After each iteration, we calculate the average pairwise Euclidean distance between all $N$ embeddings of the newly-generated images, and stop when this distance is smaller than a pre-defined threshold $\dconv$.

Finally, it should be noticed that our method is non-deterministic, \ie, when running our method multiple times, on the same input prompt $p$, different consistent characters will be generated. This is aligned with the one-to-many nature of our task. For more details and examples, please refer to the supplementary material.

%% file: figures/method/fig.tex
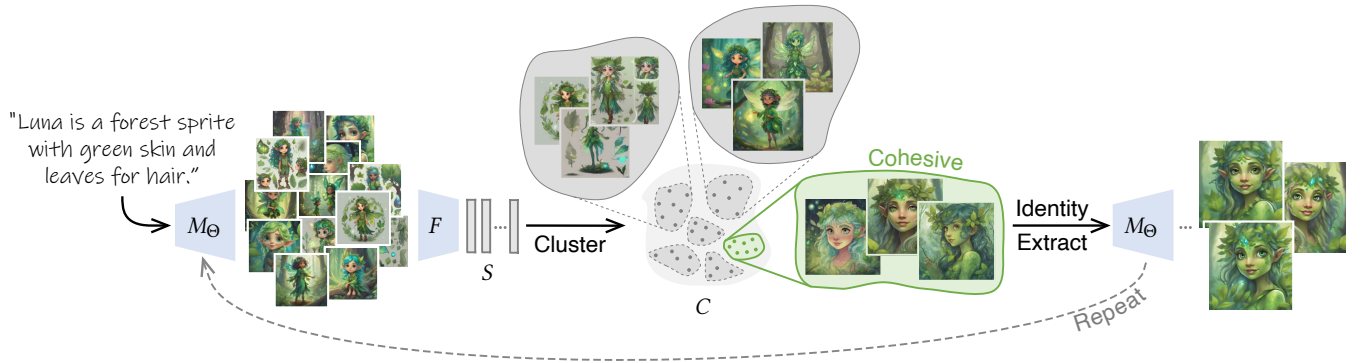
\begin{figure*}[t]
    \centering
    \begin{overpic}[width=1\linewidth]{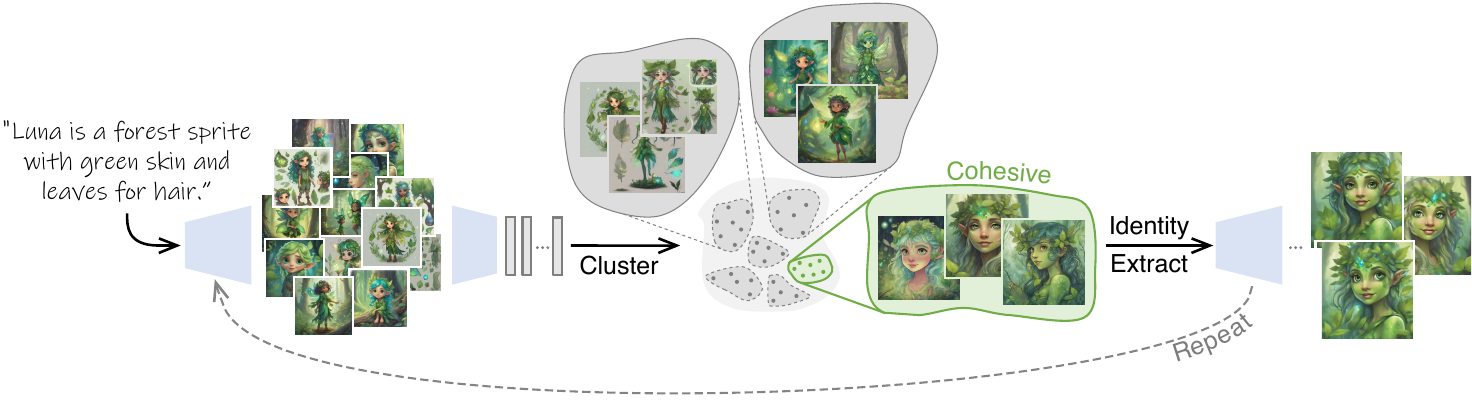}
        \put(13.5,10){$\model_\crep$}
        \put(31.5,10){$F$}
        \put(35.5,6.5){$S$}
        \put(51.5,4){$C$}
        \put(83.5,10){$\model_\crep$}
    \end{overpic}
    \caption{\textbf{Method overview.} Given an input text prompt, we start by generating numerous images using the text-to-image model $\model_\crep$, which are embedded into a semantic feature space using the feature extractor $F$. Next, these embeddings are clustered and the most cohesive group is chosen, since it contains images with shared characteristics. The ``common ground'' among the images in this set is used to refine the representation $\crep$ to better capture and fit the target. These steps are iterated until convergence to a consistent identity.}
    \label{fig:method}
\end{figure*}

%% file: algorithms/consistent_generation.tex
\begin{algorithm}[t]
    \caption{Consistent Character Generation}
    \label{alg:consistent_generation}
    \begin{algorithmic}
        \STATE \textbf{Input:} Text-to-image diffusion model $\model$, parameterized by $\crep = (\theta,\tau)$, where $\theta$ are the LoRA weights and $\tau$ is a set of custom text embeddings, target prompt $p$, feature extractor $F$.
        \STATE \textbf{Hyper-parameters:} number of generated images per step $N$, minimum cluster size $\dminc$, target cluster size $\dtgtc$, convergence criterion $\dconv$, maximum number of iterations $\diter$
        \STATE \textbf{Output:} a consistent representation $\optrep$
        \vspace{0.5em}
        \REPEAT
            \STATE $S = \bigcup_{N} F(\model_\crep (p))$
            \STATE $C = \text{K-MEANS++}(S, k=\lfloor N/\dtgtc \rfloor)$
            \STATE $C = \left\{ c \in C | \dminc < |c|\right\}$ \COMMENT{filter small clusters}
            \STATE $\ccand = \argmin \limits_{c \in C} \frac{1}{|c|} \sum_{e \in c} \Vert e - \ccen \Vert^2$
            \STATE $\Theta = \argmin \limits_{(\theta, \tau)} \Lrec$ over $\ccand$
        \UNTIL $\dconv \ge \frac{1}{{|S|}^2} \sum_{s_1, s_2 \in S} \Vert s_1 - s_2 \Vert^2$
        \RETURN $\crep$
    \end{algorithmic}
\end{algorithm}

%% file: figures/embedding_visualization/fig.tex
\begin{figure}[t]
    \centering
    \includegraphics[width=0.96\linewidth]{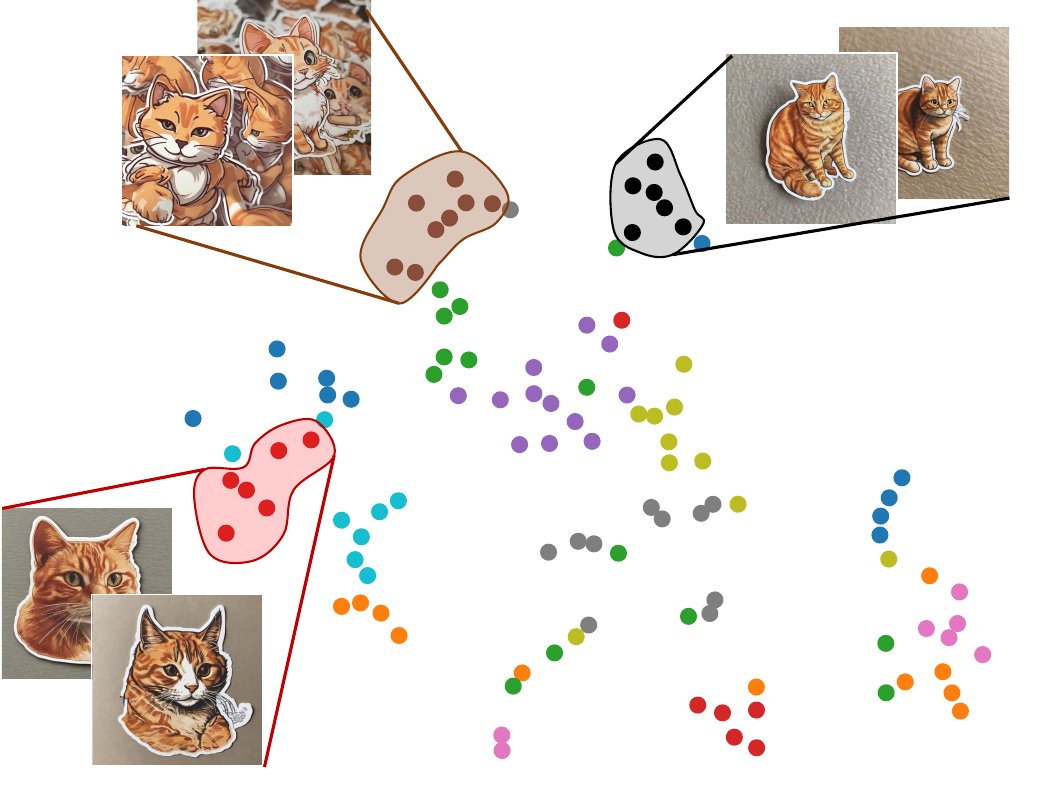}
    \caption{\textbf{Embedding visualization.} Given generated images for the text prompt \textit{``a sticker of a ginger cat''}, we project the set $S$ of their high-dimensional embeddings into 2D using t-SNE~\cite{Hinton2002StochasticNE} and indicate different K-MEANS++~\cite{Arthur2007kmeansTA} clusters using different colors. 
    Representative images are shown for three of the clusters. It may be seen that images in each cluster share the same characteristics: black cluster --- full body cats, red cluster --- cat heads, brown cluster --- images with multiple cats. According to our cohesion measure~\eqref{eq:cohesion},
    the black cluster is the most cohesive, and therefore, chosen for identity extraction (or refinement).}
    \label{fig:embedding_visualization}
\end{figure}

%% file: sections/experiments.tex
\section{Experiments}
\label{sec:experimens}

In \Cref{sec:comparisons} we compare our method against several baselines, both qualitatively and quantitatively. Next, in \Cref{sec:user_study} we describe the user study we conducted and present its results. The results of an ablation study are reported in \Cref{sec:ablation_study}. Finally, in \Cref{sec:applications} we demonstrate several applications of our method.

\subsection{Qualitative and Quantitative Comparison}
\label{sec:comparisons}

\input{figures/qualitative_comparison/fig.tex}
\input{figures/quantitative_comparison/fig.tex}

We compared our method against the most related personalization techniques \cite{Gal2022AnII,lora_diffusion,ELITE-implmentation,Li2023BLIPDiffusionPS,Ye2023IPAdapterTC}. In each experiment, each of these techniques is used to extract a character from a single image,  generated by SDXL \cite{Podell2023SDXLIL} from an input prompt $p$. The same prompt $p$ is also provided as input to our method. 
Textual Inversion (TI) \cite{Gal2022AnII} optimizes a textual token using several images of the same concept, and we converted it to support SDXL by learning \emph{two} text tokens, one for each of its text encoders, as we did in our method. In addition, we used LoRA DreamBooth~\cite{lora_diffusion} (LoRA DB), which we found less prone to overfitting than standard DB. Furthermore, we compared against all available image encoder techniques that encode a single image into the textual space of the diffusion model for later generation in novel contexts: BLIP-Diffusion~\cite{Li2023BLIPDiffusionPS}, ELITE~\cite{ELITE-implmentation}, and IP-adapter~\cite{Ye2023IPAdapterTC}. For all the baselines, we used the same prompt $p$ to generate a single image, and used it to extract the identity via optimization (TI and LoRA DB) or encoding (ELITE, BLIP-diffusion and IP-adapter).

In \Cref{fig:qualitative_comparison} we qualitatively compare our method against the above baselines. While TI \cite{Gal2022AnII}, BLIP-diffusion \cite{Li2023BLIPDiffusionPS} and IP-adapter \cite{Ye2023IPAdapterTC} are able to follow the specified prompt, they fail to produce a consistent character. LoRA DB \cite{lora_diffusion} succeeds in consistent generation, but it does not always respond to the prompt. Furthermore, the resulting character is generated in the same fixed pose. ELITE \cite{Wei2023ELITEEV} struggles with prompt following and the generated characters tend to be deformed. In comparison, our method is able to follow the prompt and maintain consistency, while generating appealing characters in different poses and viewing angles.

In order to automatically evaluate our method and the baselines quantitatively, we instructed ChatGPT~\cite{chatgpt} to generate prompts for characters of different types (\eg, animals, creatures, objects, \etc) in different styles (\eg, stickers, animations, photorealistic images, \etc). Each of these prompts was then used to extract a consistent character by our method and by each of the baselines. Next, we generated these characters in a predefined collection of novel contexts. For a visual comparison, please refer to the supplementary material.

We employ two standard evaluation metrics: prompt similarity and identity consistency, which are commonly used in the personalization literature \cite{Gal2022AnII,Ruiz2022DreamBoothFT,Avrahami2023BreakASceneEM}. Prompt similarity measures the correspondence between the generated images and the input text prompt. We use the standard CLIP \cite{Radford2021LearningTV} similarity, \ie, the normalized cosine similarity between the CLIP image embedding of the generated images and the CLIP text embedding of the source prompts. For measuring identity consistency, we calculate the pairwise similarity between the CLIP image embeddings of generated images of the same concept across different contexts (\ie, when using different text prompts for the same character).

As can be seen in \Cref{fig:quantitative_comparison_and_user_study} (left), there is an inherent trade-off between prompt similarity and identity consistency: LoRA DB and ELITE exhibit high identity consistency, while sacrificing prompt similarity. TI and BLIP-diffusion achieve high prompt similarity but low identity consistency. Our method achieves better identity consistency than IP-adapter, which is significant from the user's perspective, as supported by our user study.

\subsection{User Study}
\label{sec:user_study}
We conducted a user study to evaluate our method, using the Amazon Mechanical Turk (AMT) platform \cite{amt}. We used the same generated prompts and samples that were used in \Cref{sec:comparisons} and asked the evaluators to rate the prompt similarity and identity consistency of each result on a Likert scale of 1--5. For ranking the prompt similarity, the evaluators were presented with the target text prompt and the result of our method and the baselines on the same page, and were asked to rate each of the images. For identity consistency, for each of the generated concepts, we compared our method and the baselines by randomly choosing pairs of generated images with different target prompts, and the evaluators were asked to rate on a scale of 1--5 whether the images contain the same main character. Again, all the pairs of the same character for the different baselines were shown on the same page.

As can be seen in \Cref{fig:quantitative_comparison_and_user_study} (right), our method again exhibits a good balance between identity consistency and prompt similarity, with a wider gap separating it from the baselines. For more details and statistical significance analysis, read the supplementary material.

\subsection{Ablation Study}
\label{sec:ablation_study}
We conducted an ablation study for the following cases: (1) \emph{Without clustering} --- we omit the clustering step described in \Cref{sec:identity_clustering}, and instead simply generate 5 images according to the input prompt. (2) \emph{Without LoRA} --- we reduce the optimizable representation $\crep$ in the identity extraction stage, as described in \Cref{sec:identity_extraction}, to consist of only the newly-added text tokens without the additional LoRA weights. (3) \emph{With re-initialization} --- instead of using the latest representation $\crep$ in each of the optimization iterations, as described in \Cref{sec:iterative_convergence}, we re-initialize it in each iteration. (4) \emph{Single iteration} --- rather than iterating until convergence (\Cref{sec:iterative_convergence}), we stop after a single iteration.

As can be seen in \Cref{fig:quantitative_comparison_and_user_study} (left), all of the above key components are crucial for achieving a consistent identity in the final result: (1) removing the clustering harms the identity extraction stage because the training set is too diverse, (2) reducing the representation causes underfitting, as the model does not have enough parameters to properly capture the identity, (3) re-initializing the representation in each iteration, or (4) performing a single iteration, does not allow the model to converge into a single identity.

For a visual comparison of the ablation study, as well as comparison of alternative feature extractors (DINOv1 \cite{Caron2021EmergingPI} and CLIP \cite{Radford2021LearningTV}), please refer to the supplementary material.

\subsection{Applications}
\label{sec:applications}

\input{figures/applications/fig.tex}

As demonstrated in \Cref{fig:applications}, our method can be used for various down-stream tasks, such as (a) Illustrating a story by breaking it into a different scenes and using the same consistent character for all of them. (b) Local text-driven image editing by integrating Blended Latent Diffusion~\cite{avrahami2023blendedlatent, blended_2022_CVPR} --- a consistent character can be injected into a specified location of a provided background image, in a novel pose specified by a text prompt. (c) Generating a consistent character with an additional pose control using  ControlNet~\cite{zhang2023controlnet}. For more details, please refer to the supplementary material. In addition, as demonstrated in \Cref{fig:user_control} instead of choosing the most cohesive cluster automatically, as explained in \Cref{sec:identity_clustering}, a user can manually select one of the clusters according to their preferences, to affect the final result.

%% file: figures/qualitative_comparison/fig.tex
\begin{figure*}[t]
    \centering
    \setlength{\tabcolsep}{3.5pt}
    \renewcommand{\arraystretch}{0.4}
    \setlength{\ww}{0.295\columnwidth}
    \begin{tabular}{ccccccc}
        &
        \textbf{TI} &
        \textbf{LoRA DB} &
        \textbf{ELITE} &
        \textbf{BLIP-diff} &
        \textbf{IP-Adapter} &
        \textbf{Ours}
        \\

        &
        \cite{Gal2022AnII} &
        \cite{lora_diffusion} &
        \cite{Wei2023ELITEEV} &
        \cite{Li2023BLIPDiffusionPS} &
        \cite{Ye2023IPAdapterTC} &
        \\
        
        \rotatebox[origin=c]{90}{\phantom{a}}
        \rotatebox[origin=c]{90}{\phantom{a}}
        \rotatebox[origin=c]{90}{\textit{``indoors''}} &
        {\includegraphics[valign=c, width=\ww]{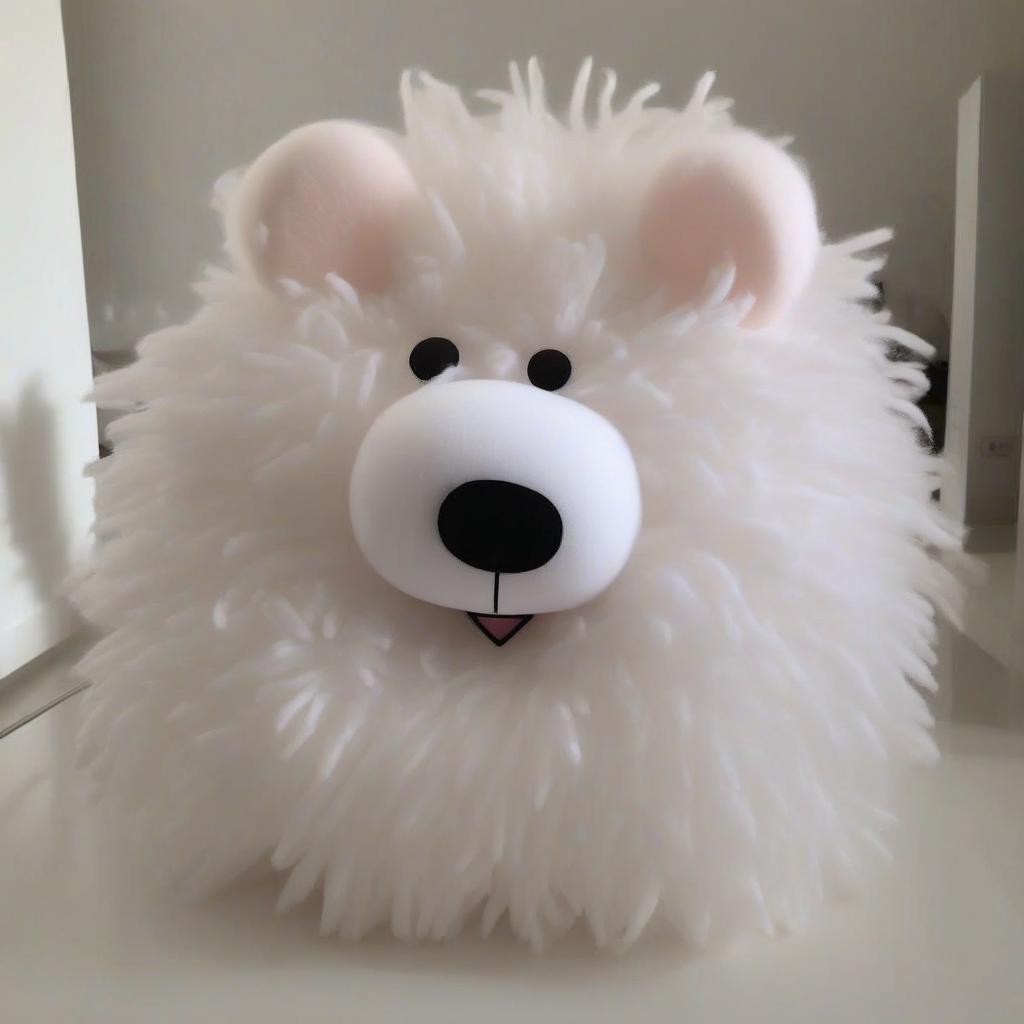}} &
        {\includegraphics[valign=c, width=\ww]{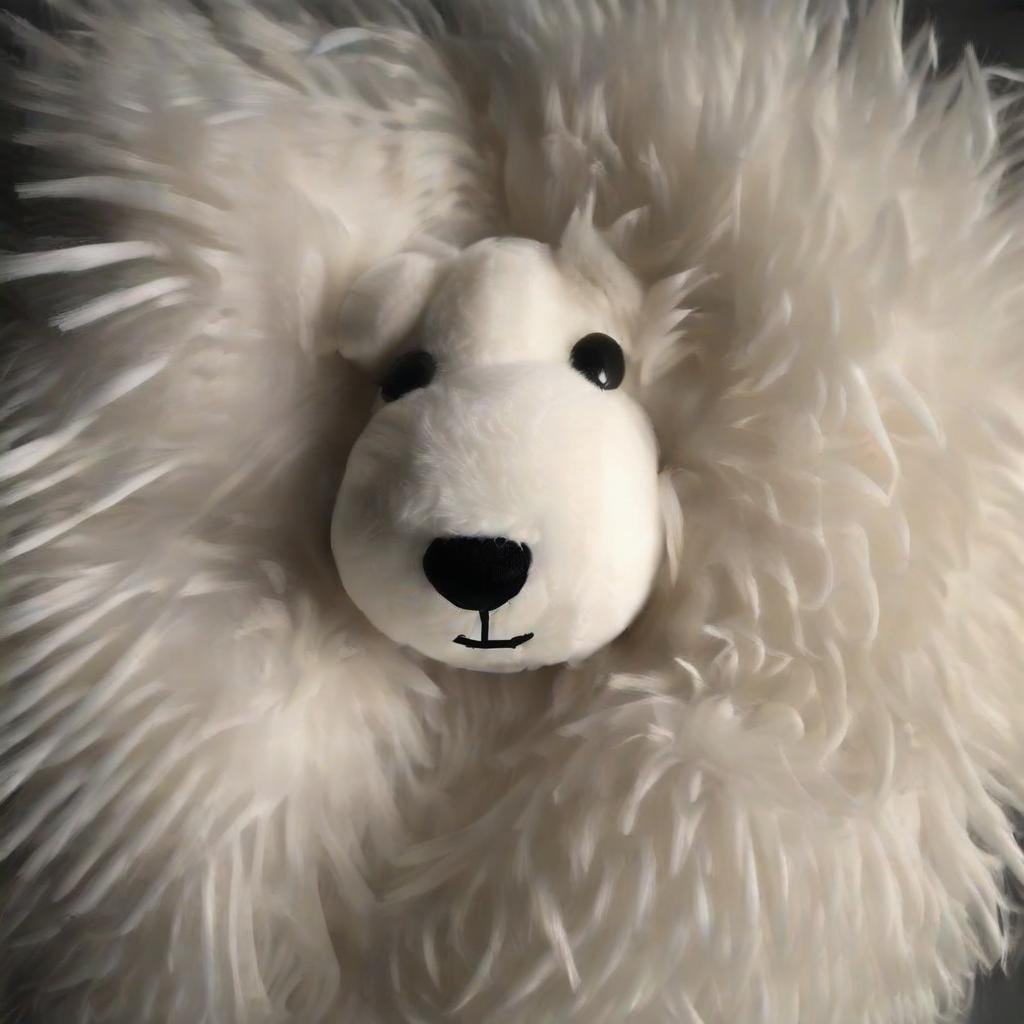}} &
        {\includegraphics[valign=c, width=\ww]{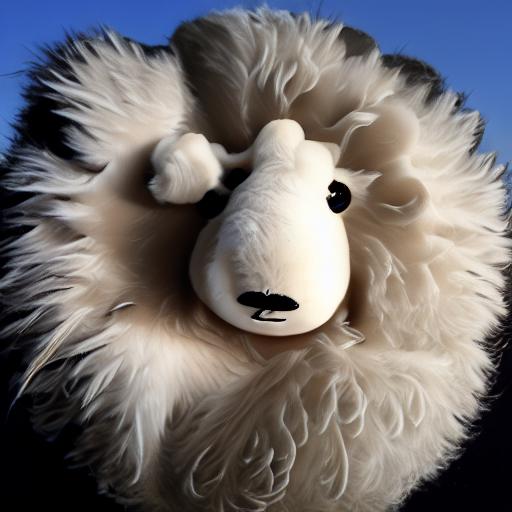}} &
        {\includegraphics[valign=c, width=\ww]{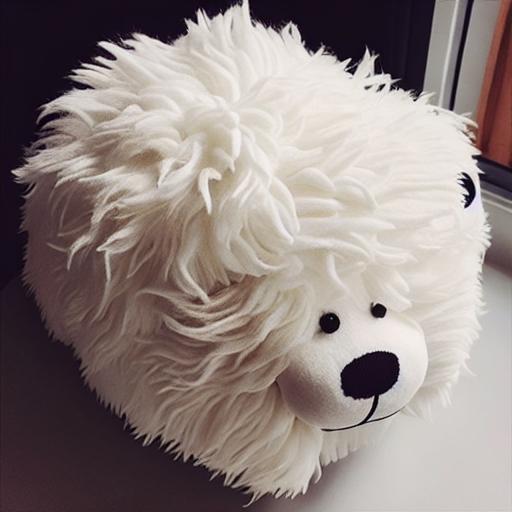}} &
        {\includegraphics[valign=c, width=\ww]{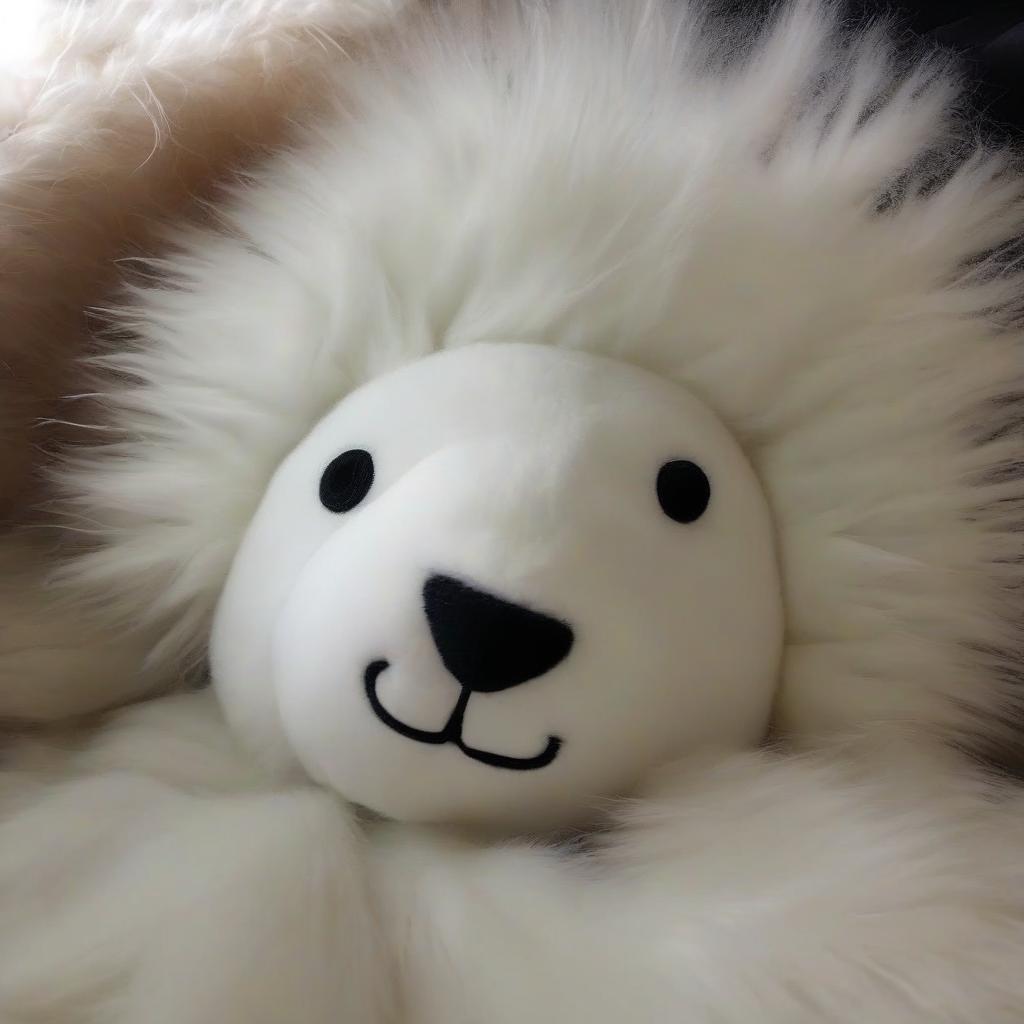}} &
        {\includegraphics[valign=c, width=\ww]{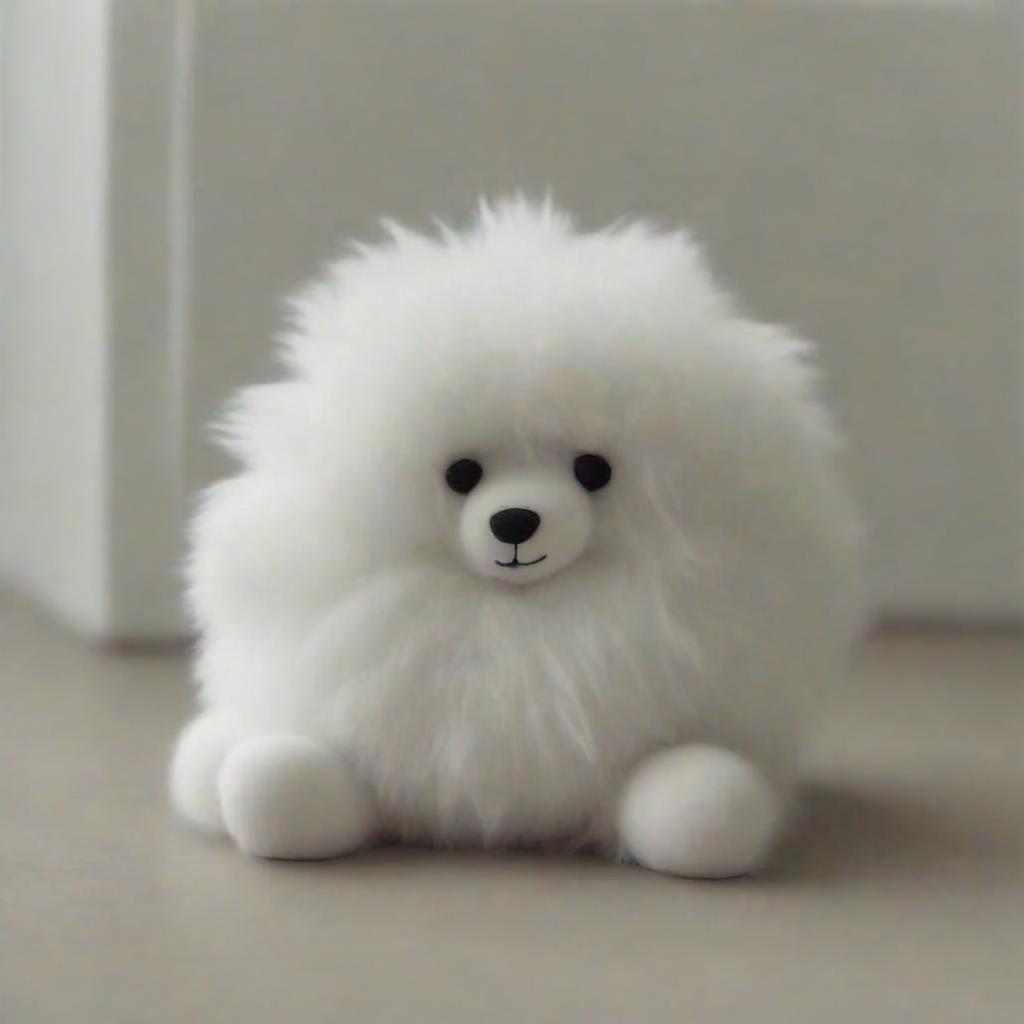}}
        \\
        \\

        \rotatebox[origin=c]{90}{\phantom{a}}
        \rotatebox[origin=c]{90}{\phantom{a}}
        \rotatebox[origin=c]{90}{\textit{``in the park''}} &
        {\includegraphics[valign=c, width=\ww]{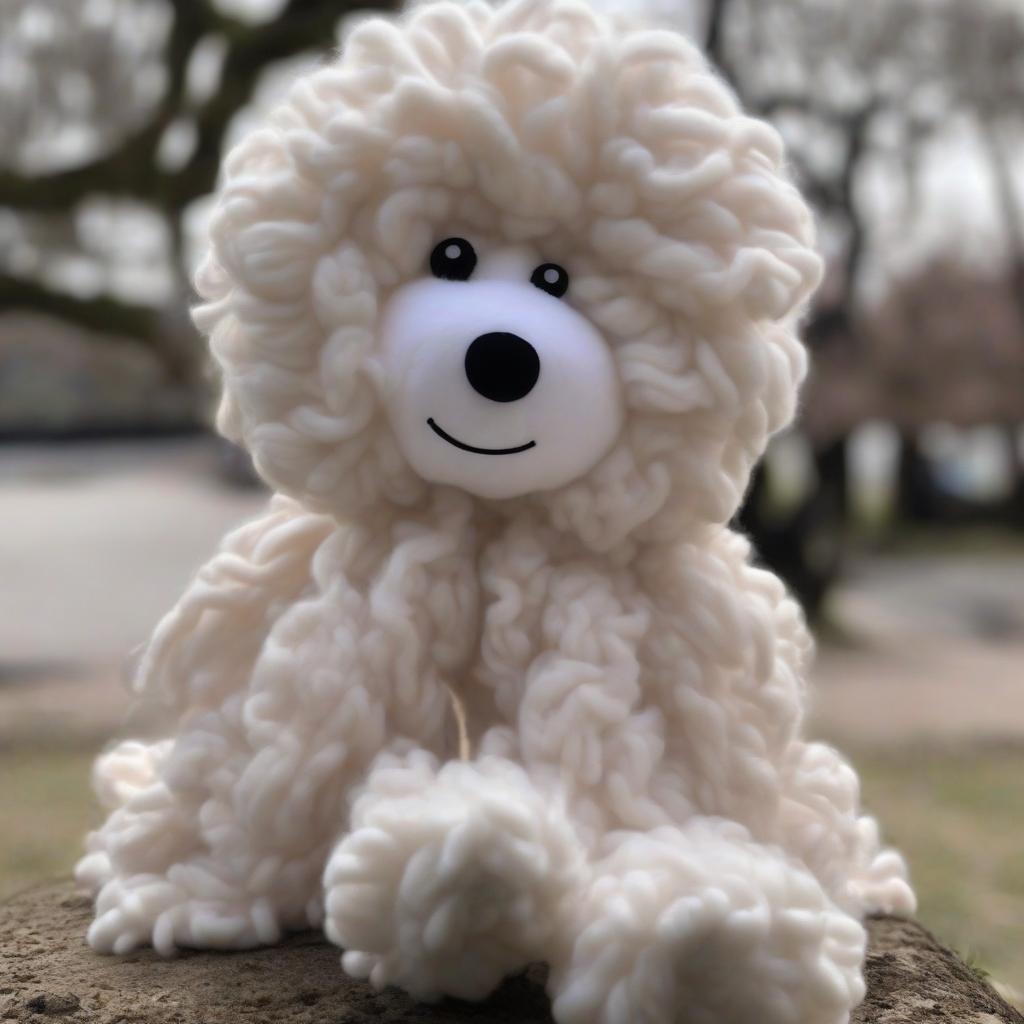}} &
        {\includegraphics[valign=c, width=\ww]{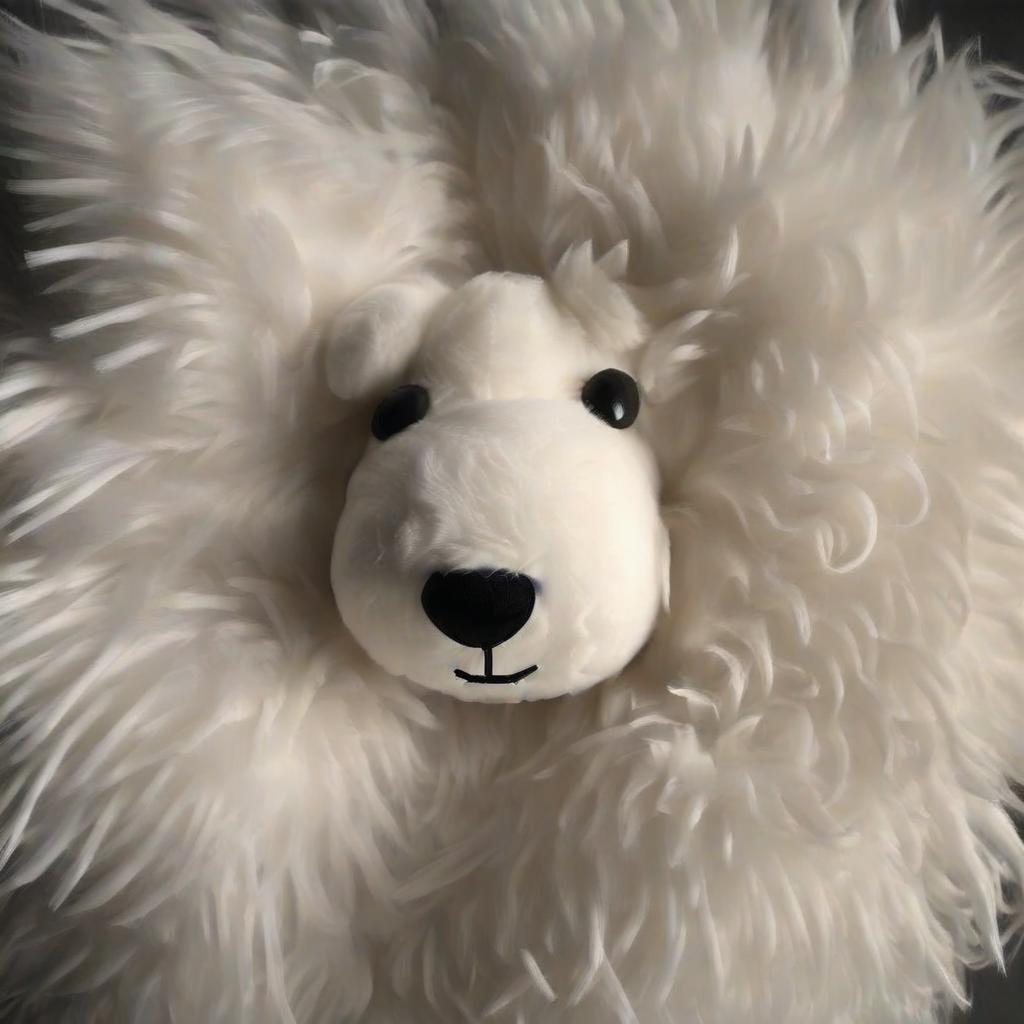}} &
        {\includegraphics[valign=c, width=\ww]{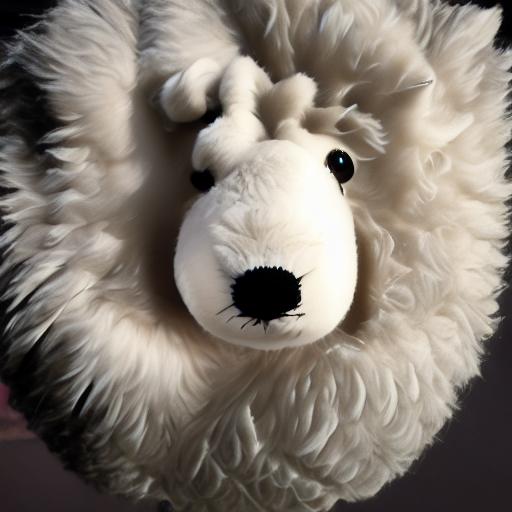}} &
        {\includegraphics[valign=c, width=\ww]{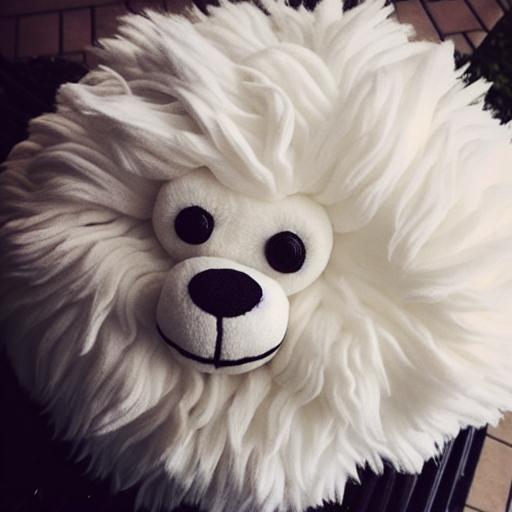}} &
        {\includegraphics[valign=c, width=\ww]{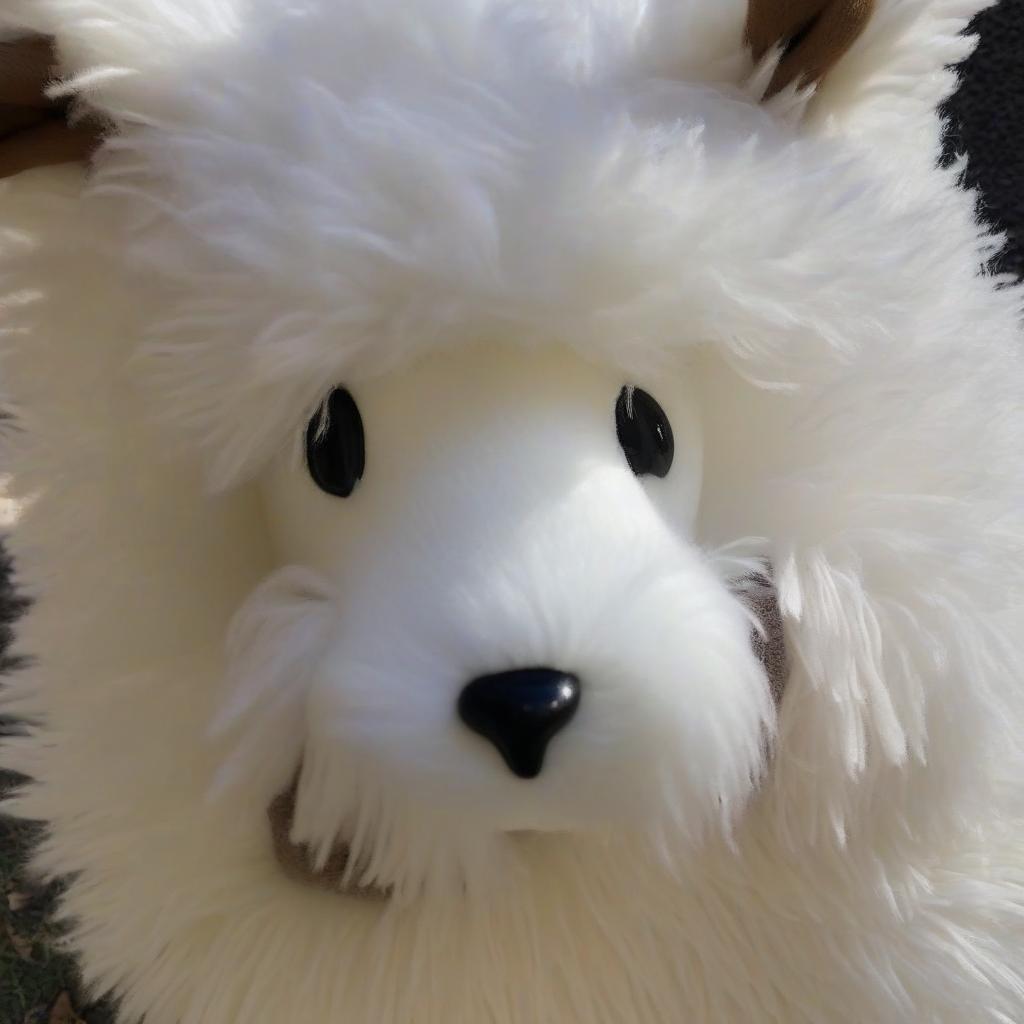}} &
        {\includegraphics[valign=c, width=\ww]{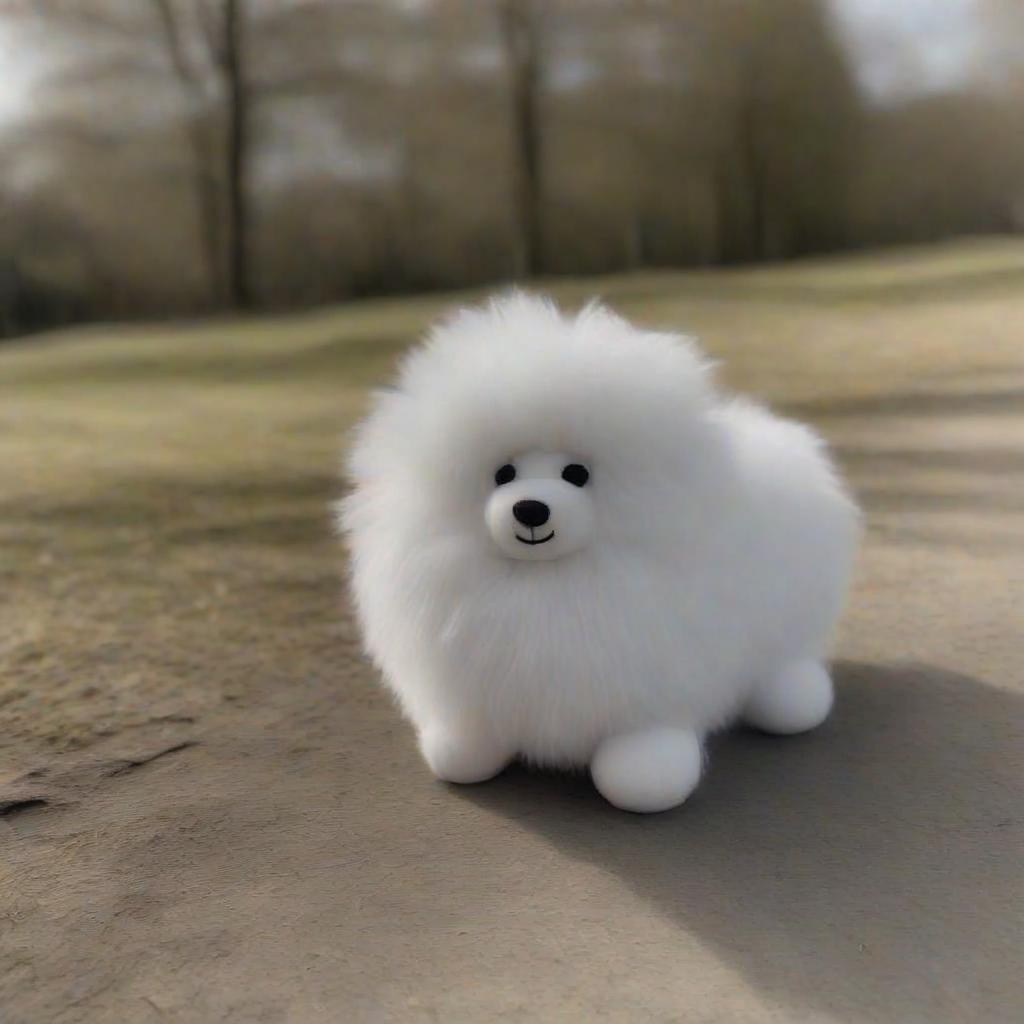}}
        \\
        \\
        
        \multicolumn{7}{c}{\textit{``a photo of a white fluffy toy''}}
        \\
        \\
        \midrule

        \\
        \rotatebox[origin=c]{90}{\phantom{a}}
        \rotatebox[origin=c]{90}{\textit{``wearing a red}}
        \rotatebox[origin=c]{90}{\textit{hat in the street''}} &
        {\includegraphics[valign=c, width=\ww]{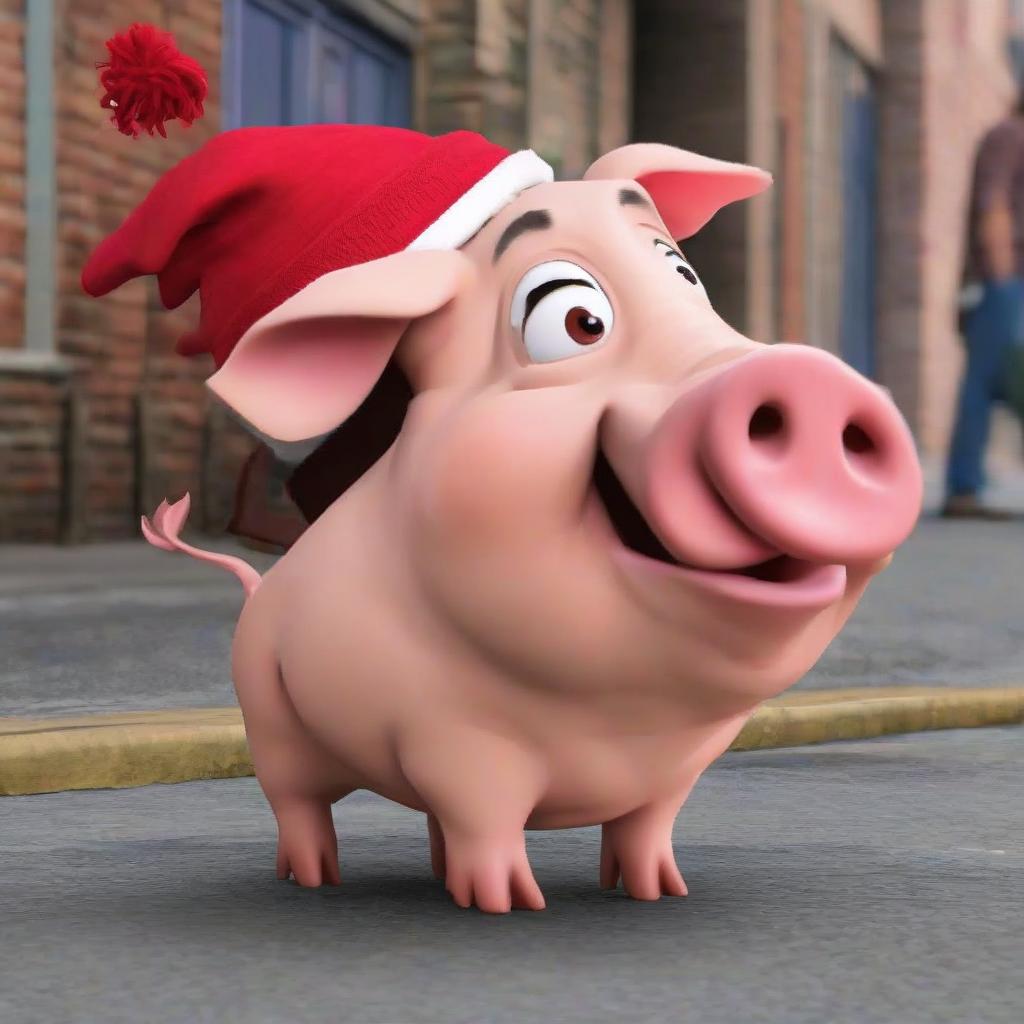}} &
        {\includegraphics[valign=c, width=\ww]{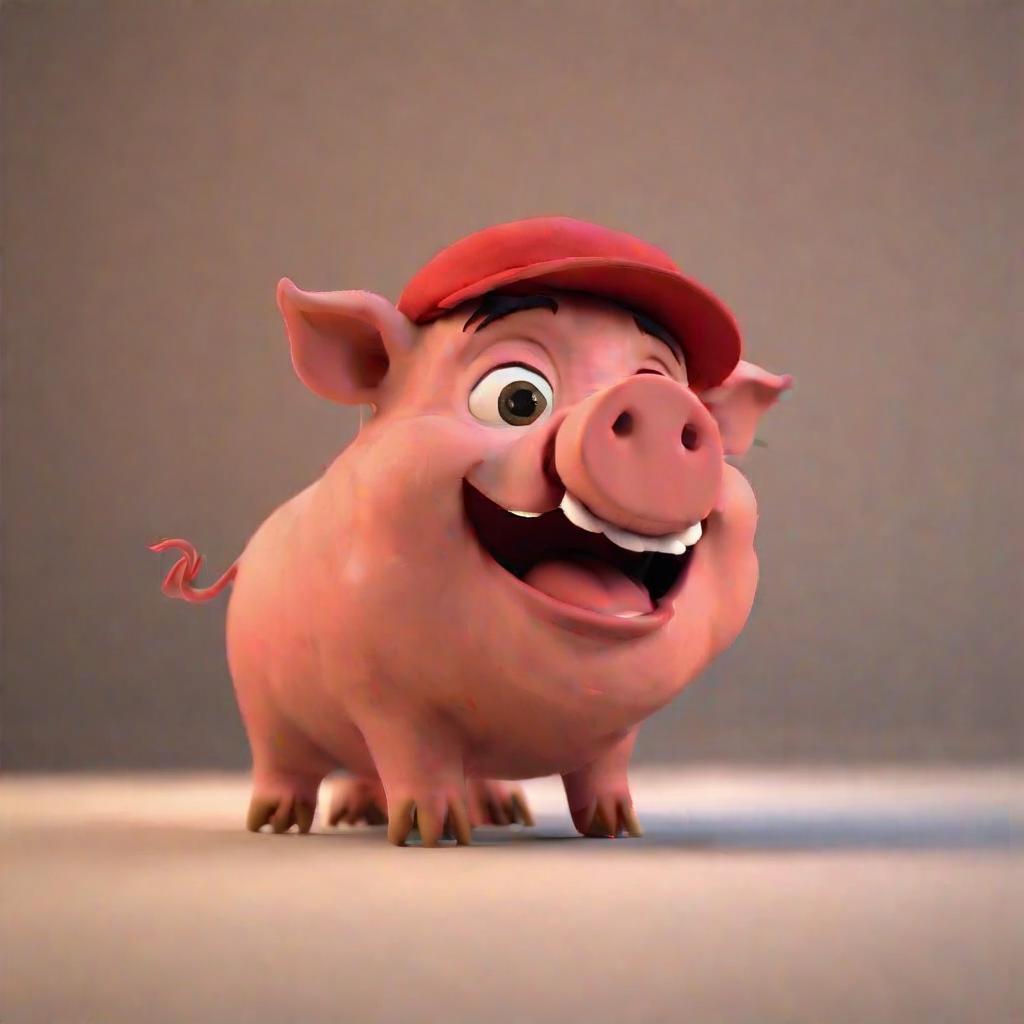}} &
        {\includegraphics[valign=c, width=\ww]{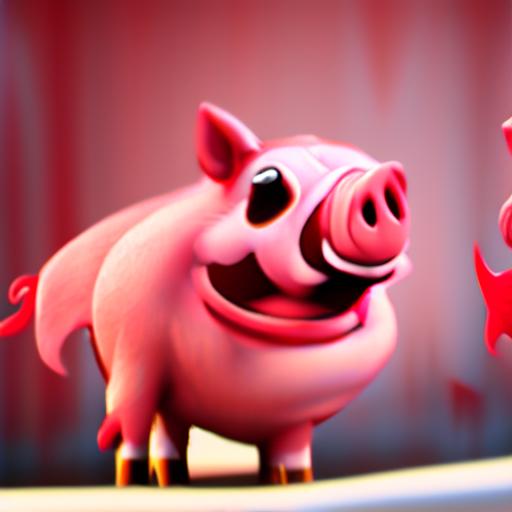}} &
        {\includegraphics[valign=c, width=\ww]{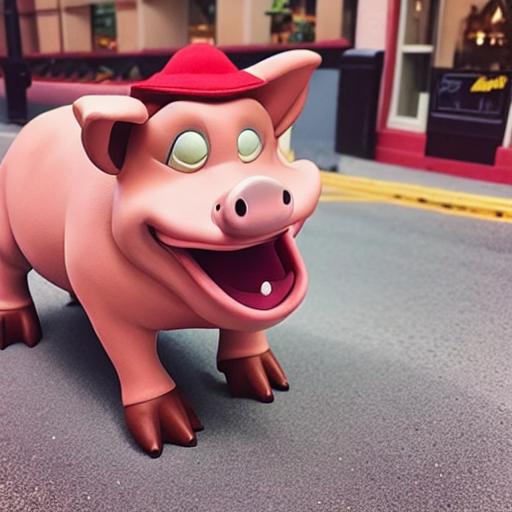}} &
        {\includegraphics[valign=c, width=\ww]{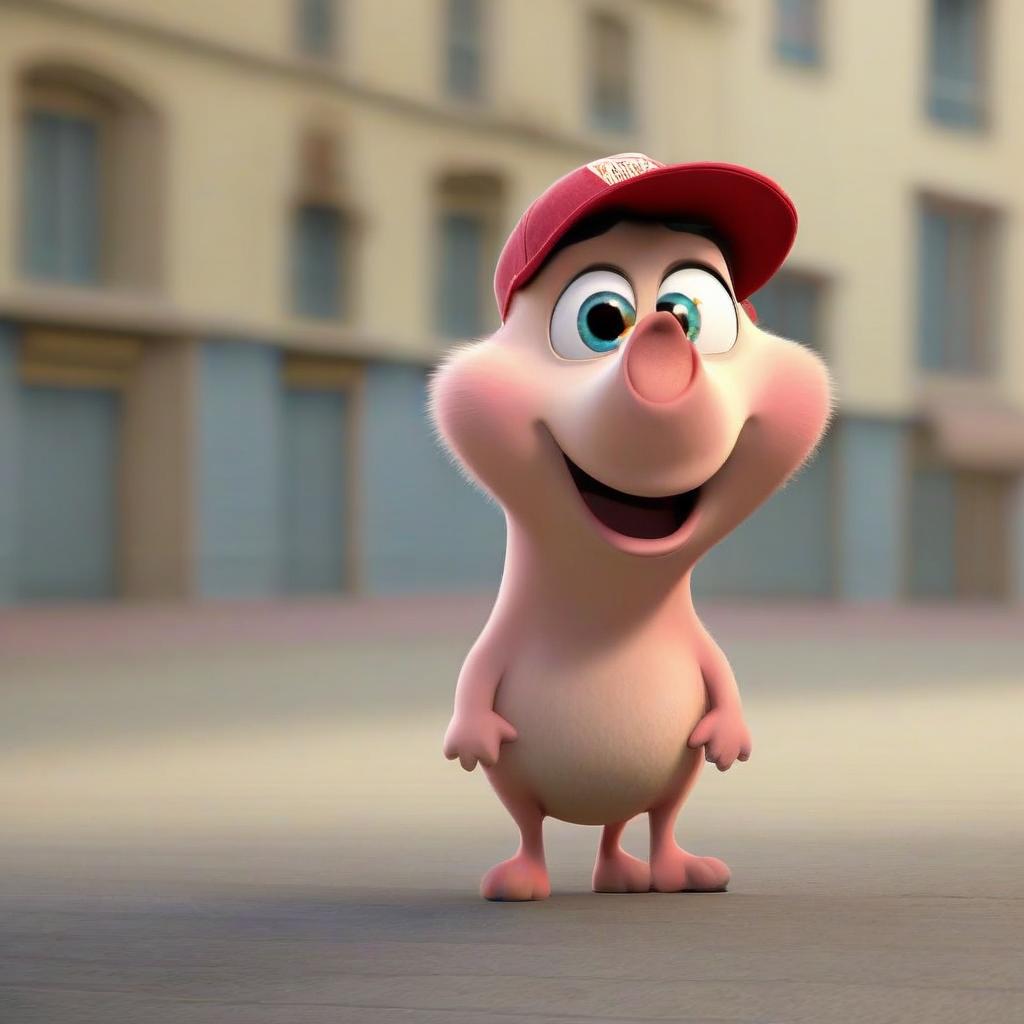}} &
        {\includegraphics[valign=c, width=\ww]{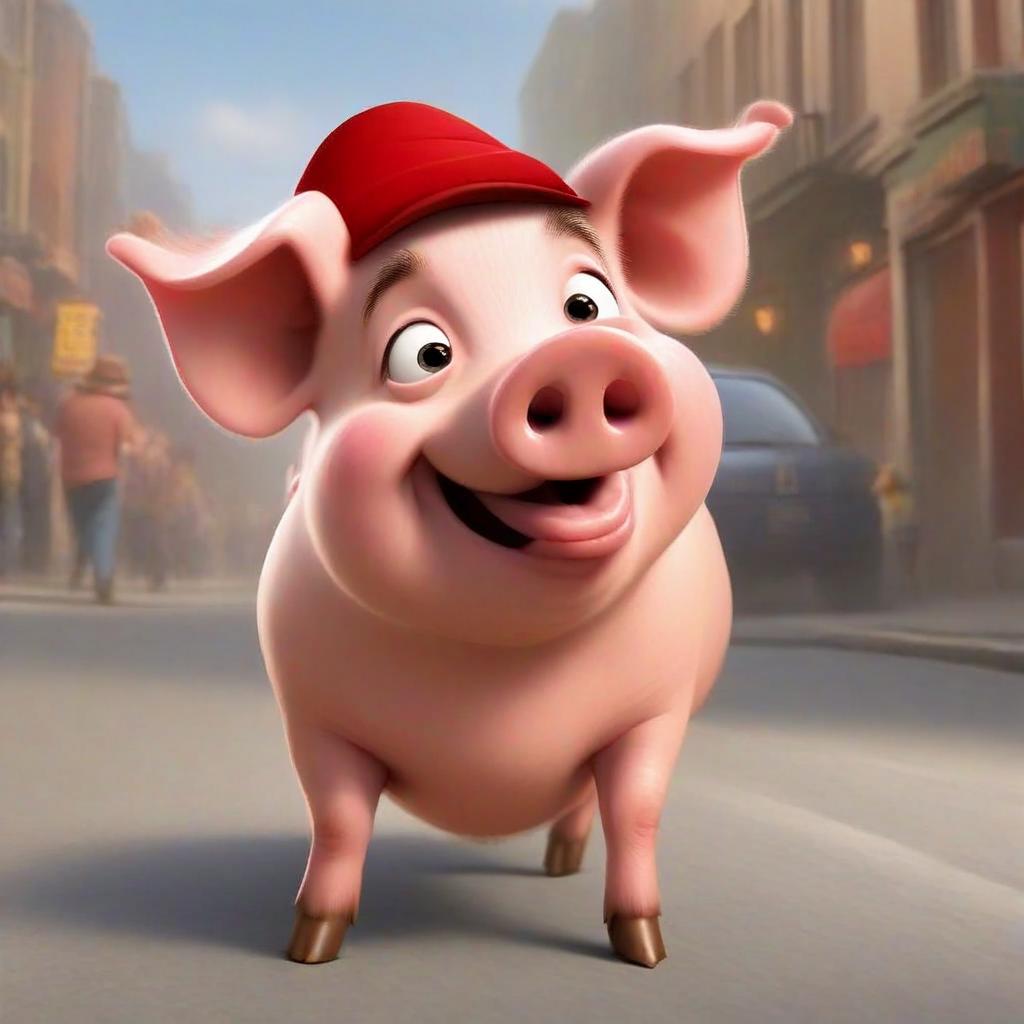}}
        \\
        \\

        \rotatebox[origin=c]{90}{\phantom{a}}
        \rotatebox[origin=c]{90}{\textit{``jumping near}}
        \rotatebox[origin=c]{90}{\textit{the river''}} &
        {\includegraphics[valign=c, width=\ww]{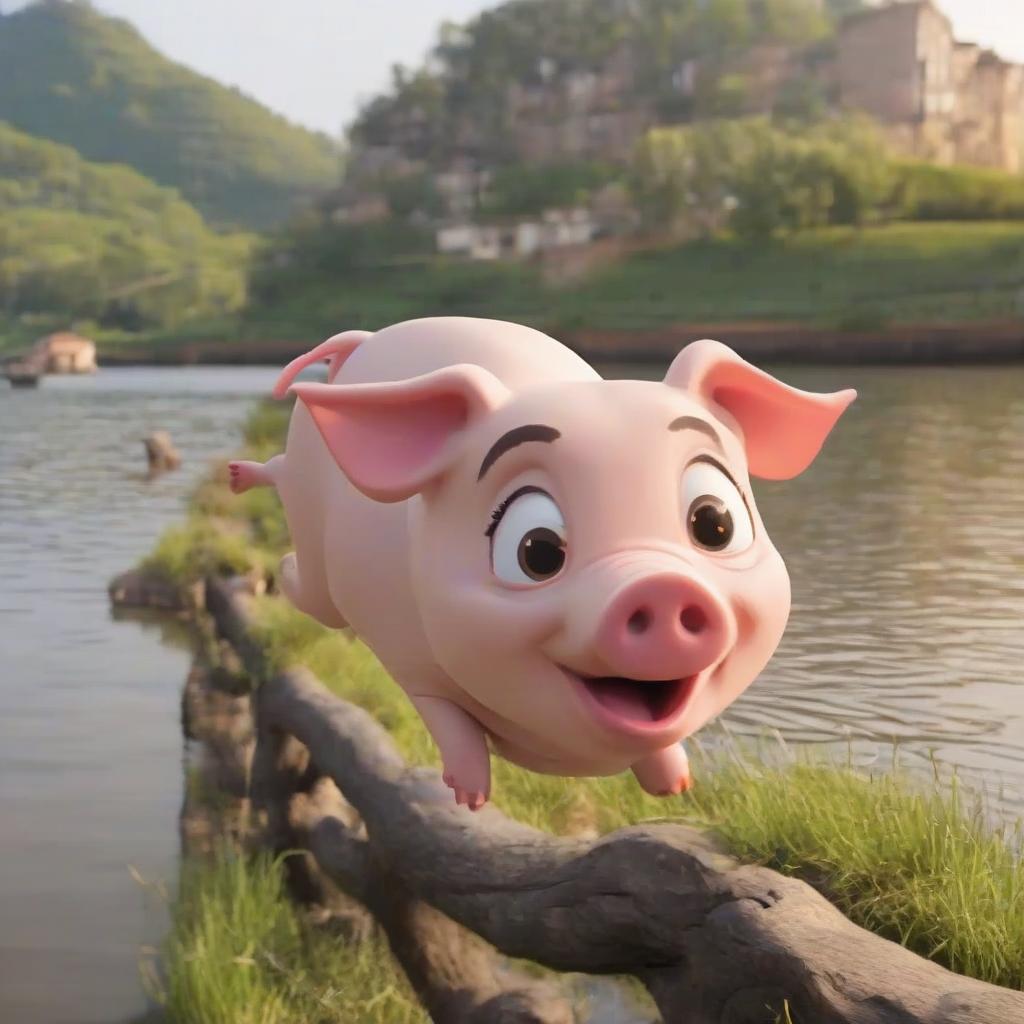}} &
        {\includegraphics[valign=c, width=\ww]{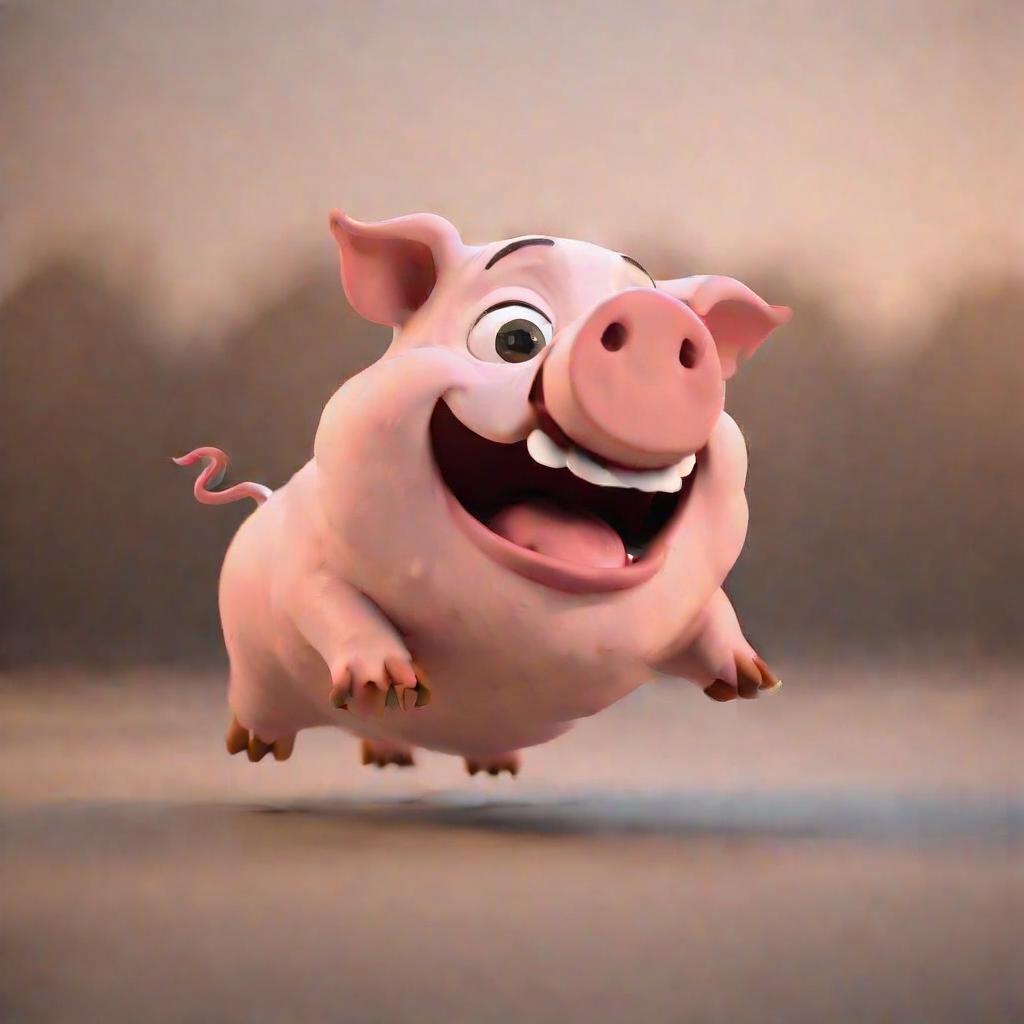}} &
        {\includegraphics[valign=c, width=\ww]{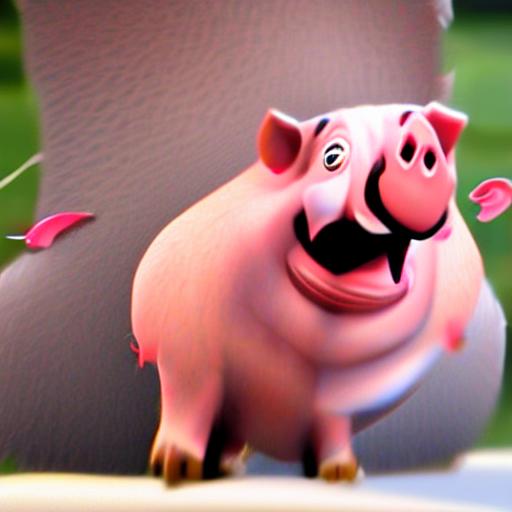}} &
        {\includegraphics[valign=c, width=\ww]{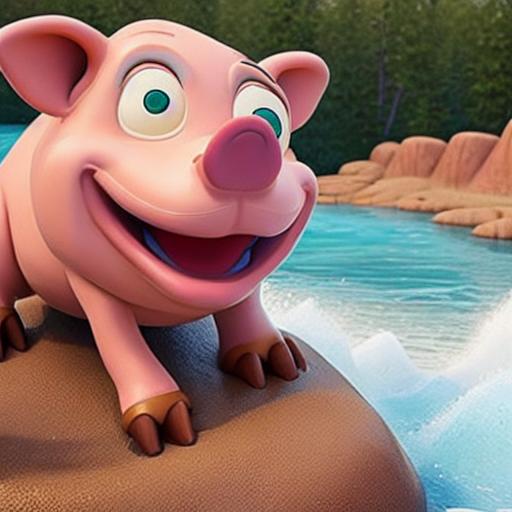}} &
        {\includegraphics[valign=c, width=\ww]{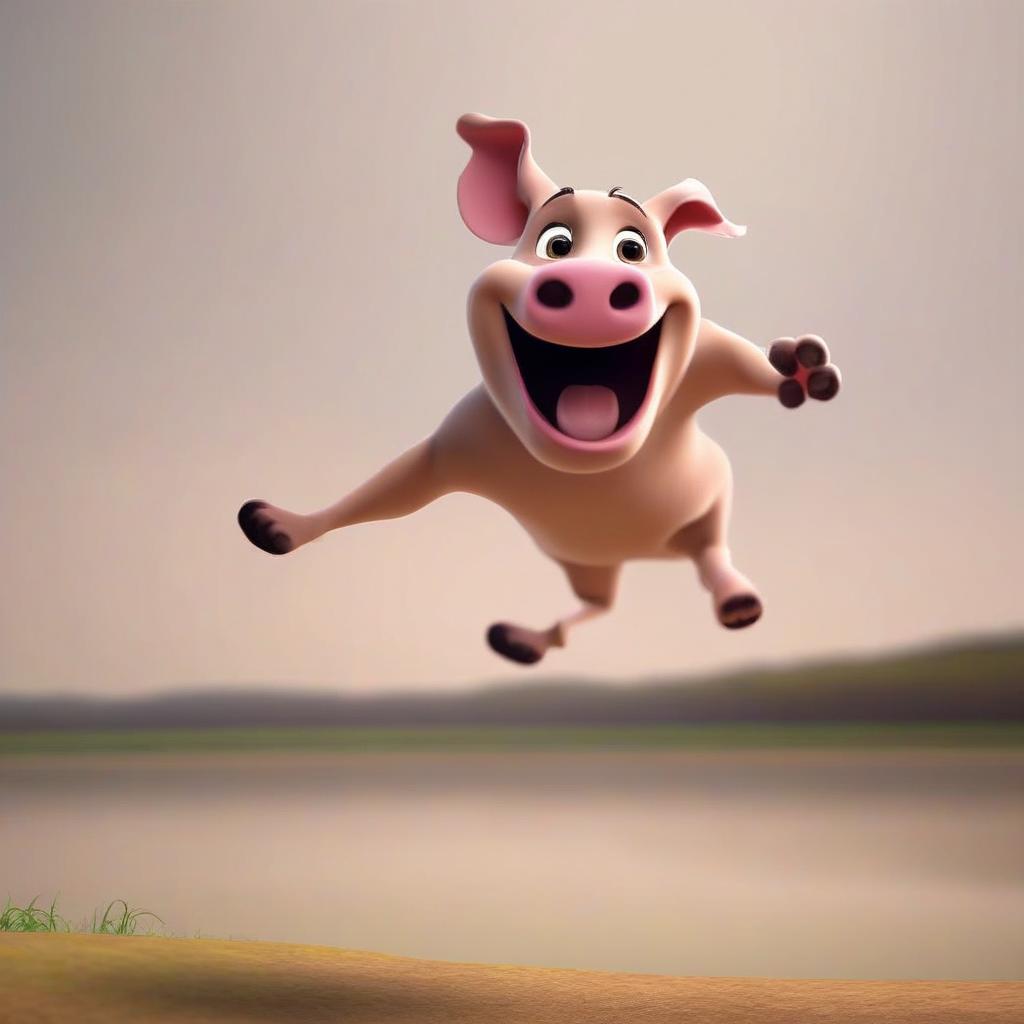}} &
        {\includegraphics[valign=c, width=\ww]{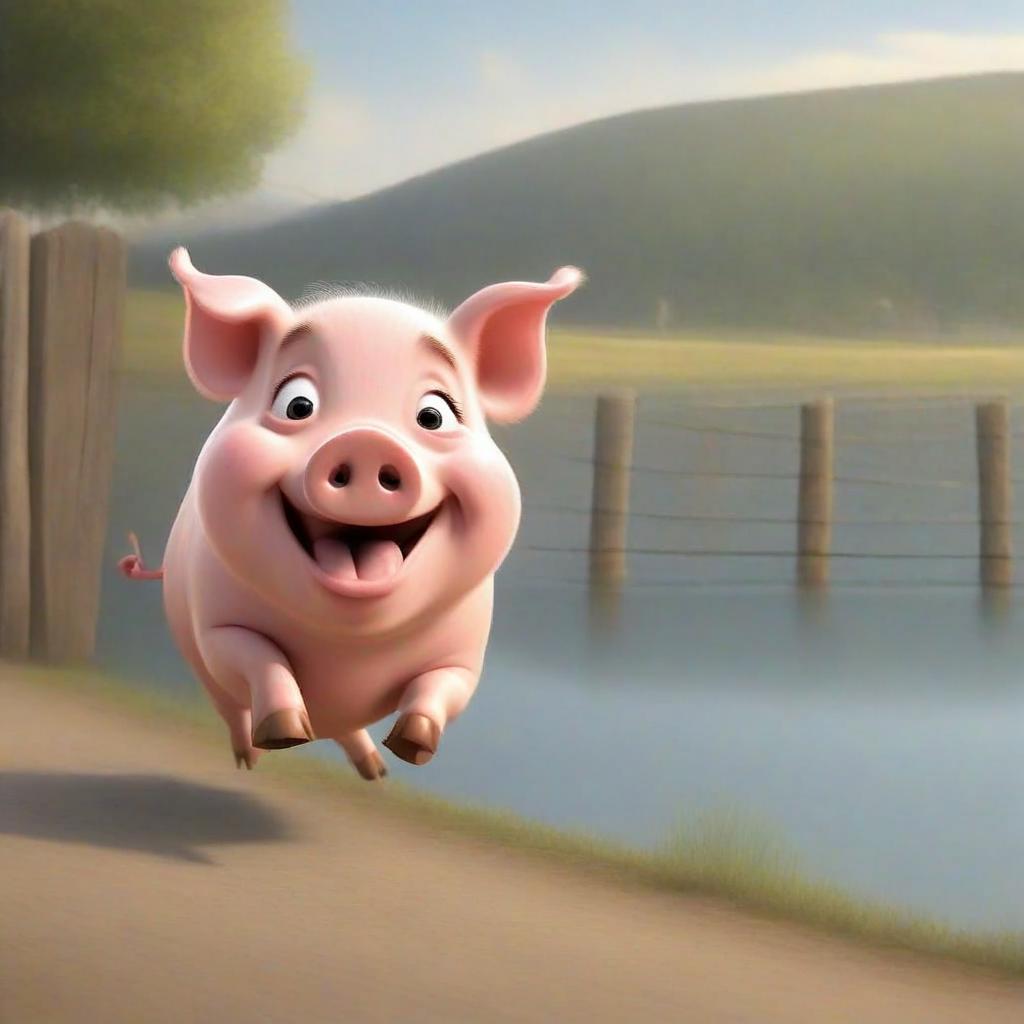}}
        \\
        \\

        \multicolumn{7}{c}{\textit{``a 3D animation of a happy pig''}}
        \\
        \\
        \midrule

        \\
        \rotatebox[origin=c]{90}{\textit{``near the}}
        \rotatebox[origin=c]{90}{\textit{Golden Gate}}
        \rotatebox[origin=c]{90}{\textit{Bridge''}} &
        {\includegraphics[valign=c, width=\ww]{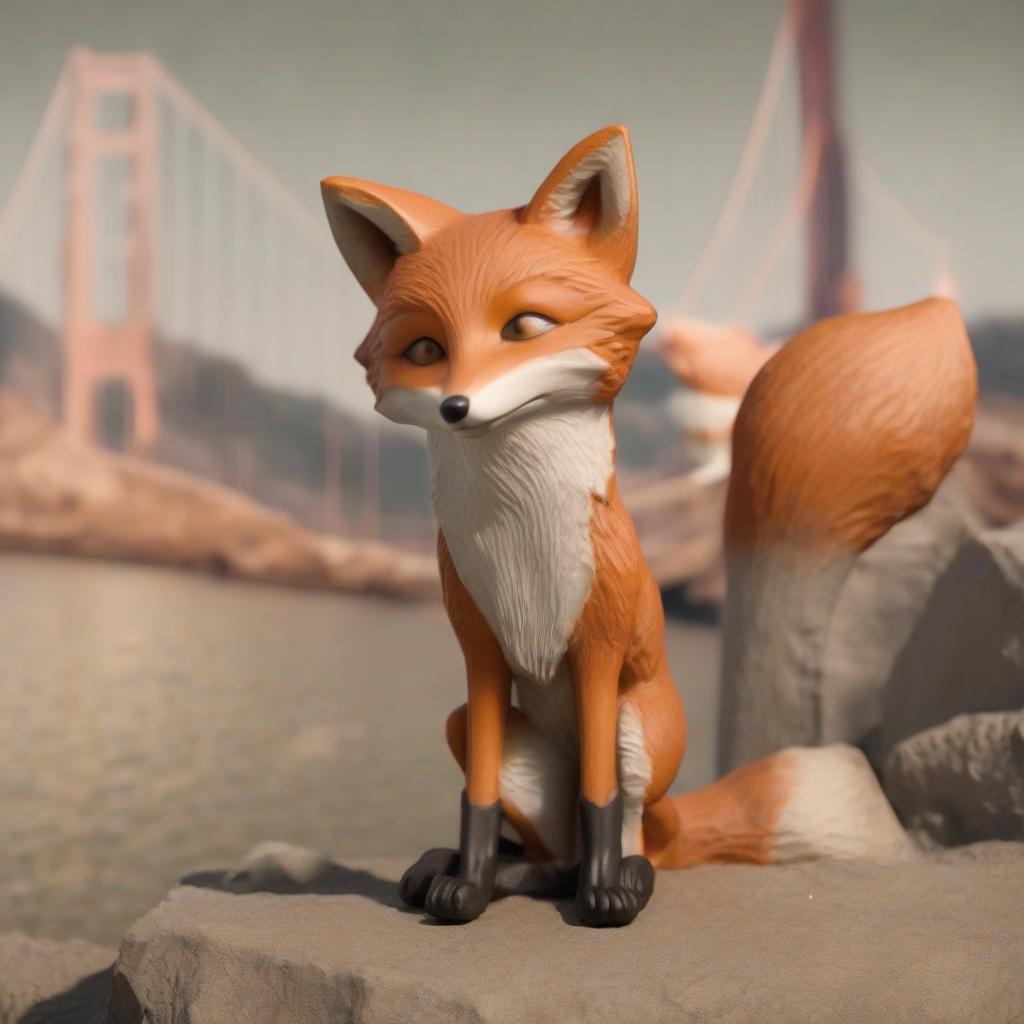}} &
        {\includegraphics[valign=c, width=\ww]{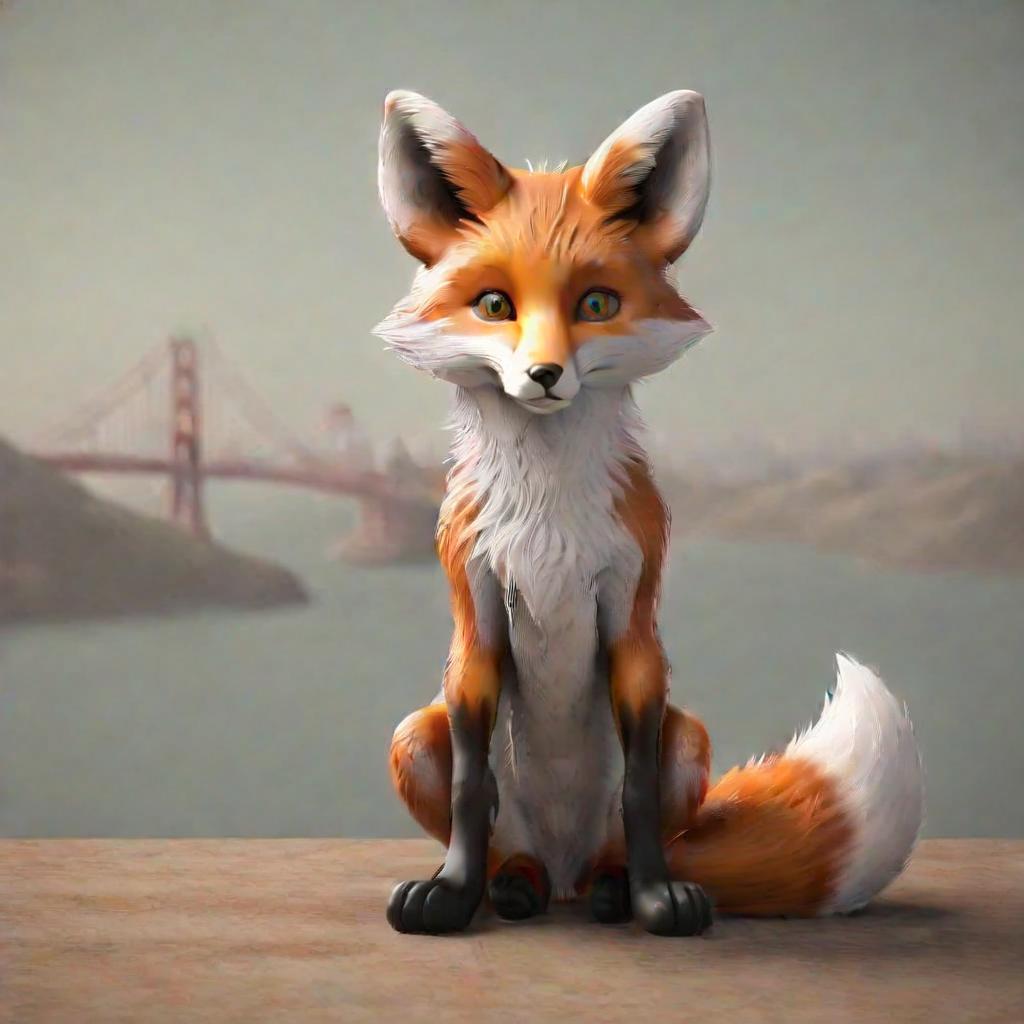}} &
        {\includegraphics[valign=c, width=\ww]{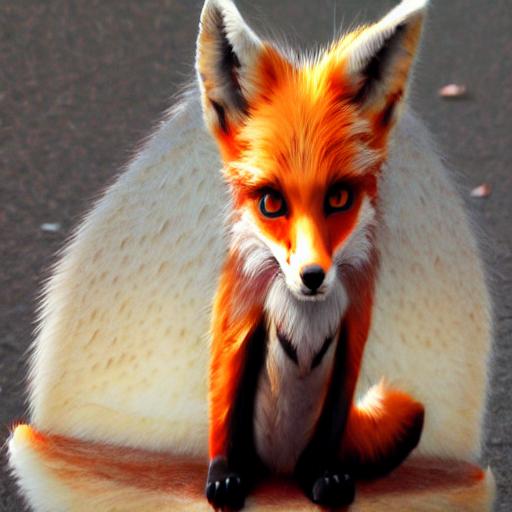}} &
        {\includegraphics[valign=c, width=\ww]{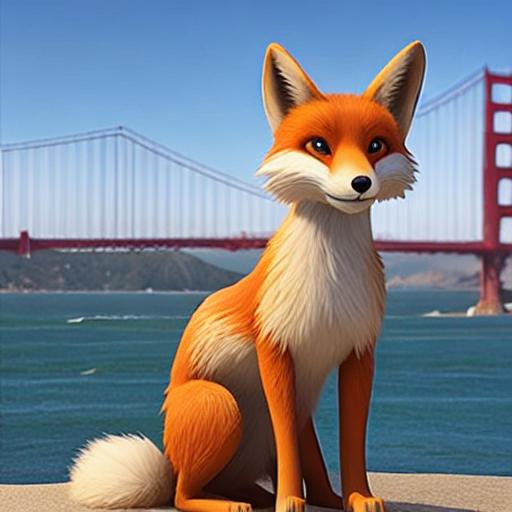}} &
        {\includegraphics[valign=c, width=\ww]{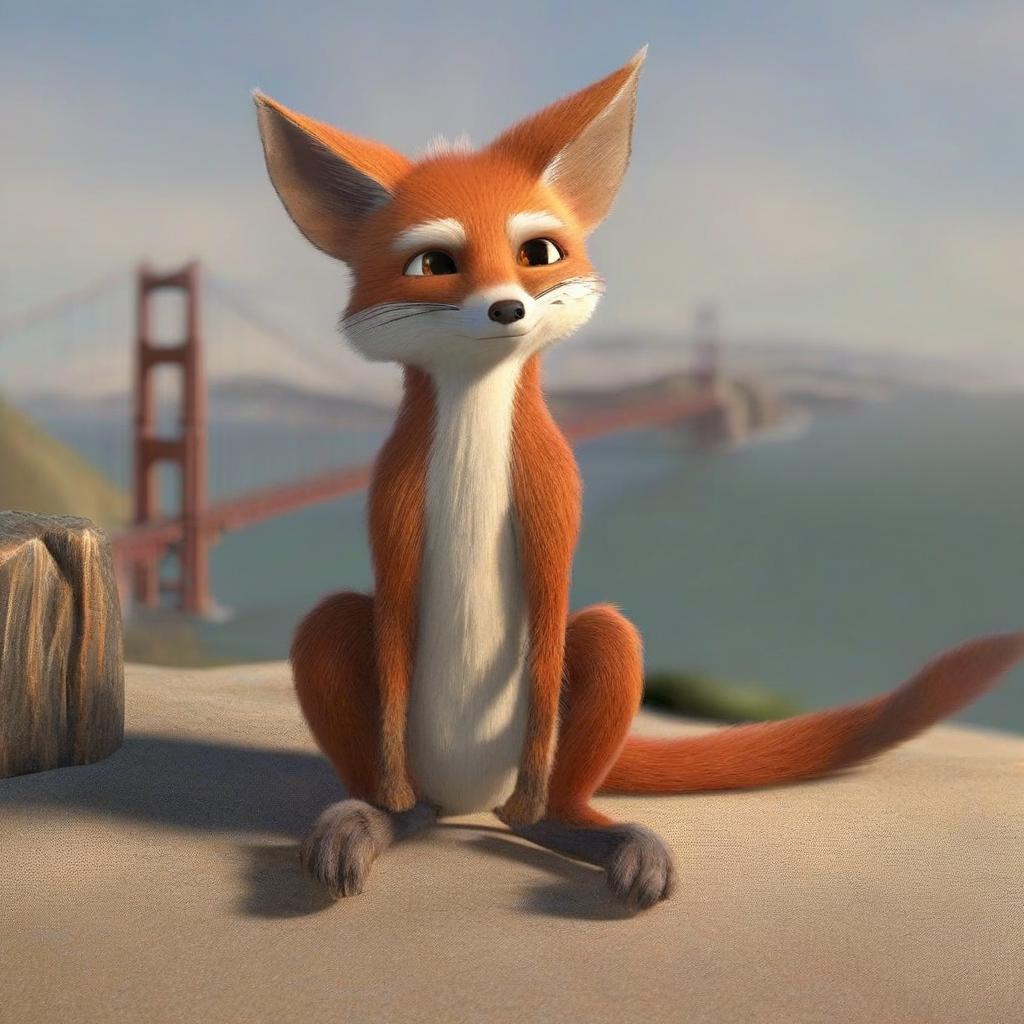}} &
        {\includegraphics[valign=c, width=\ww]{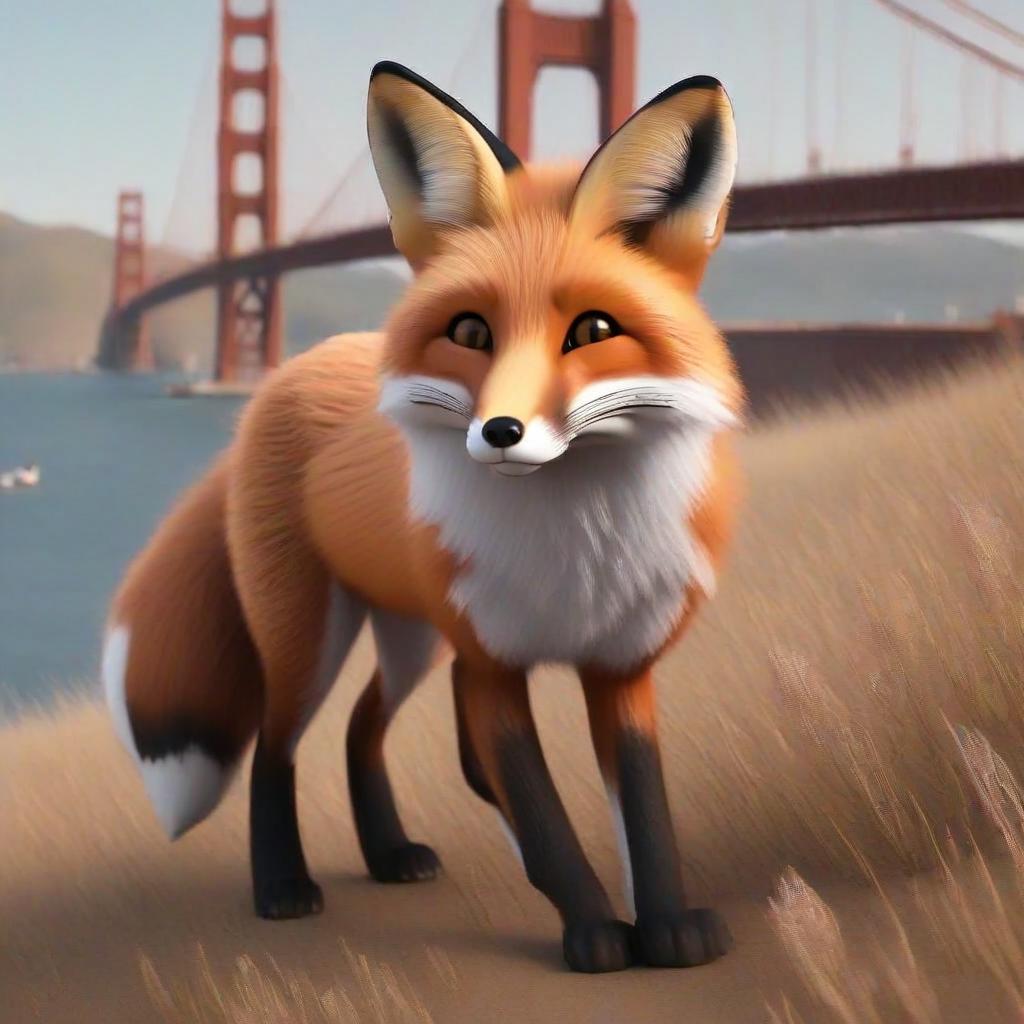}}
        \\
        \\

        \rotatebox[origin=c]{90}{\phantom{a}}
        \rotatebox[origin=c]{90}{\phantom{a}}
        \rotatebox[origin=c]{90}{\textit{``in the snow''}} &
        {\includegraphics[valign=c, width=\ww]{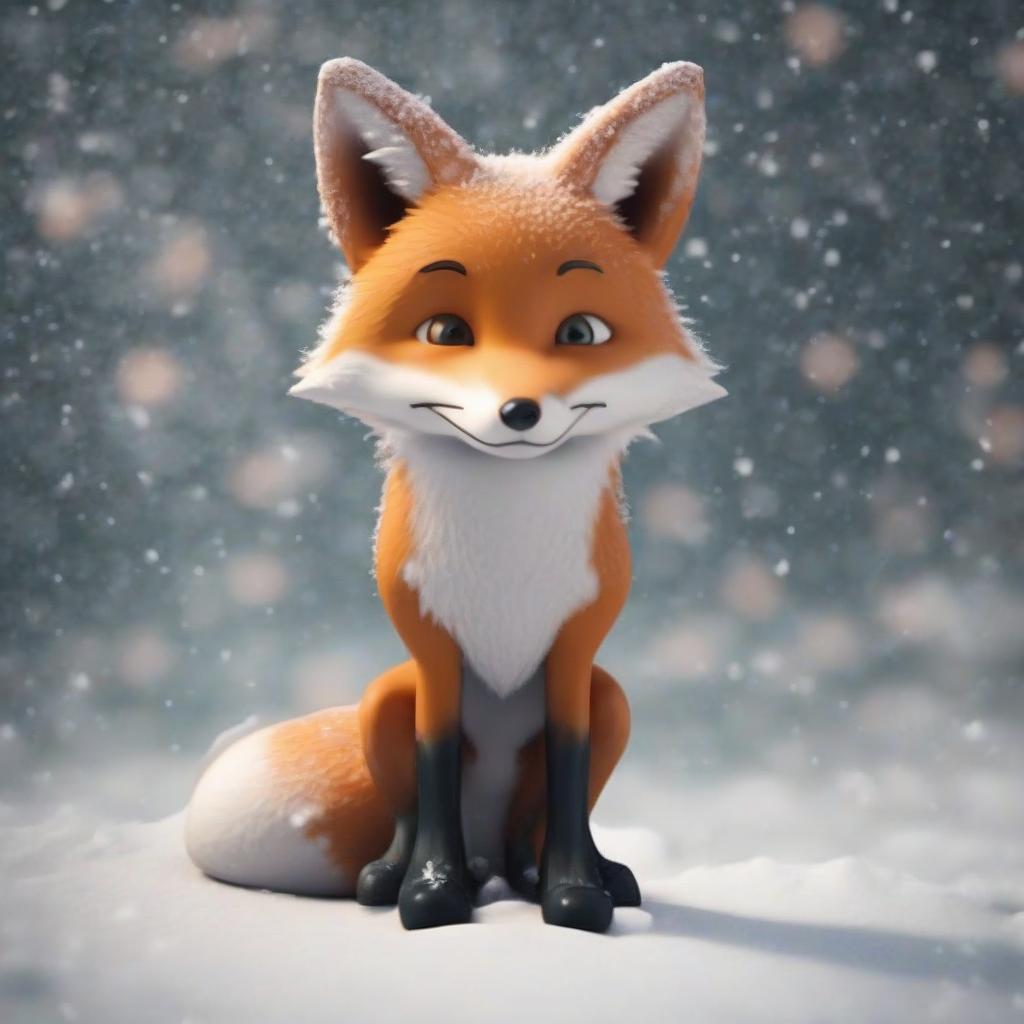}} &
        {\includegraphics[valign=c, width=\ww]{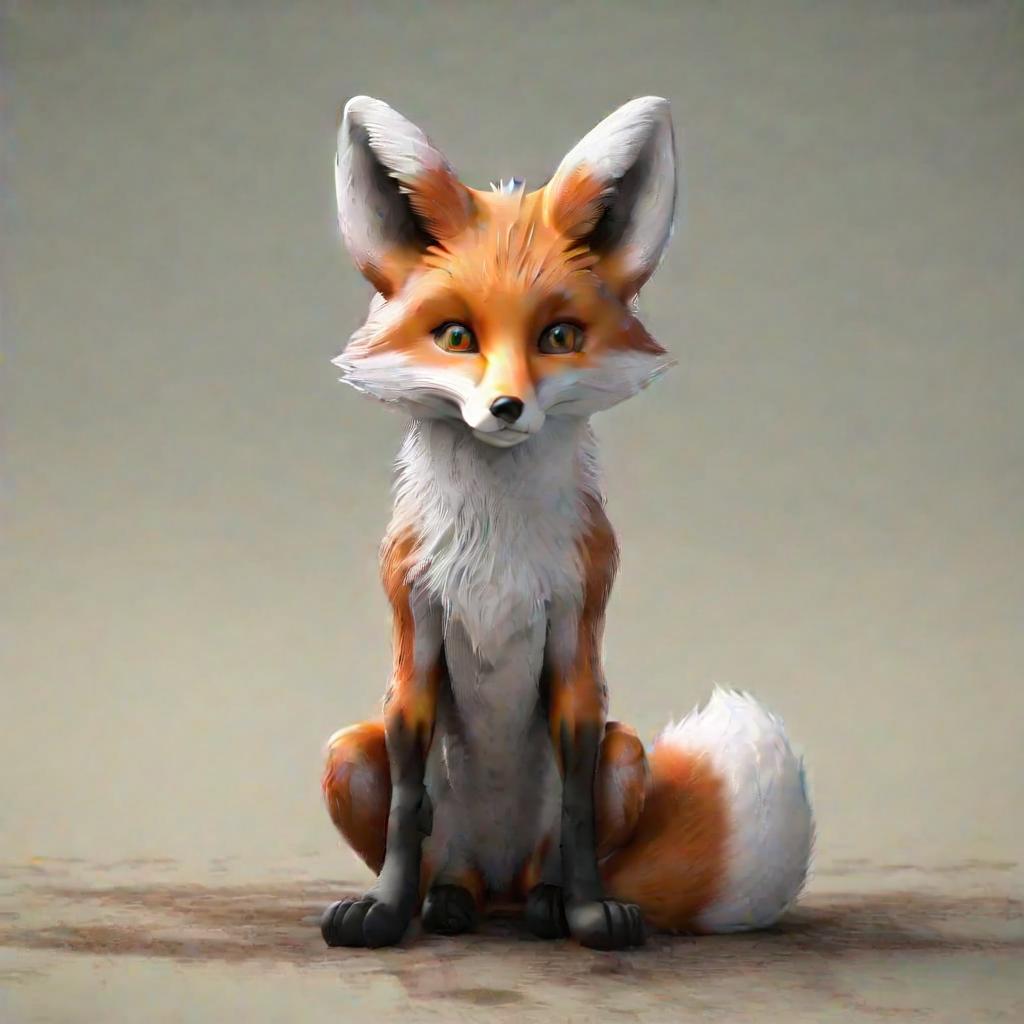}} &
        {\includegraphics[valign=c, width=\ww]{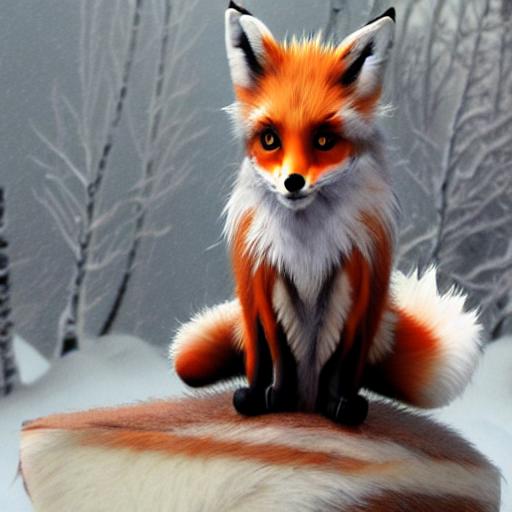}} &
        {\includegraphics[valign=c, width=\ww]{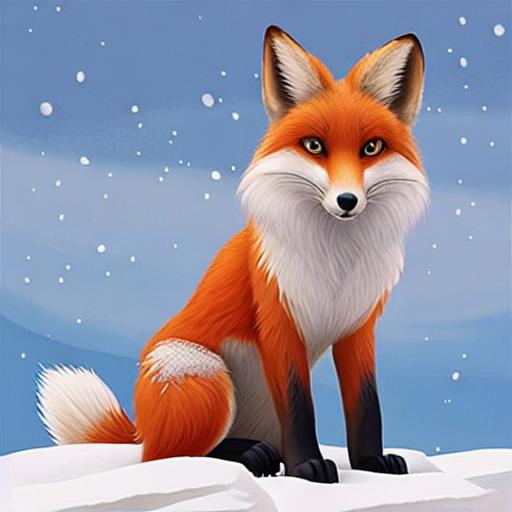}} &
        {\includegraphics[valign=c, width=\ww]{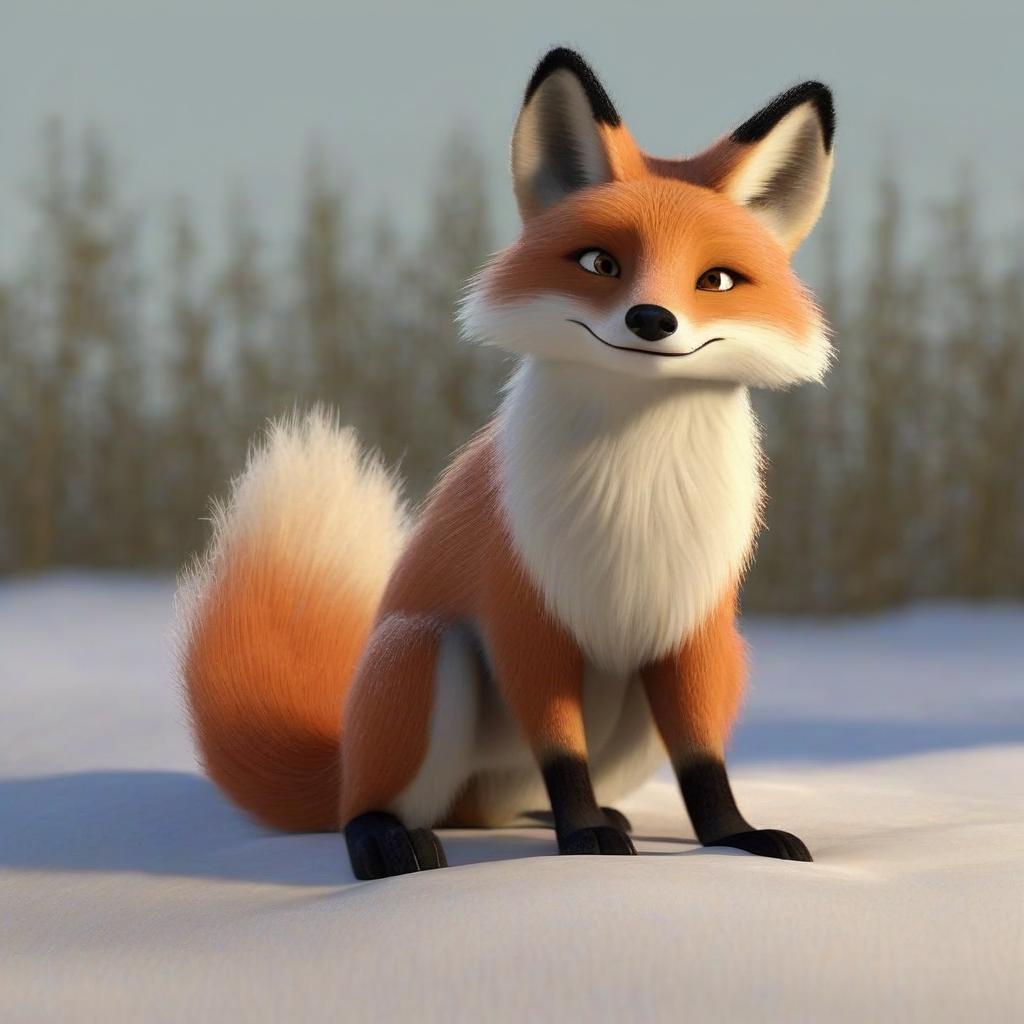}} &
        {\includegraphics[valign=c, width=\ww]{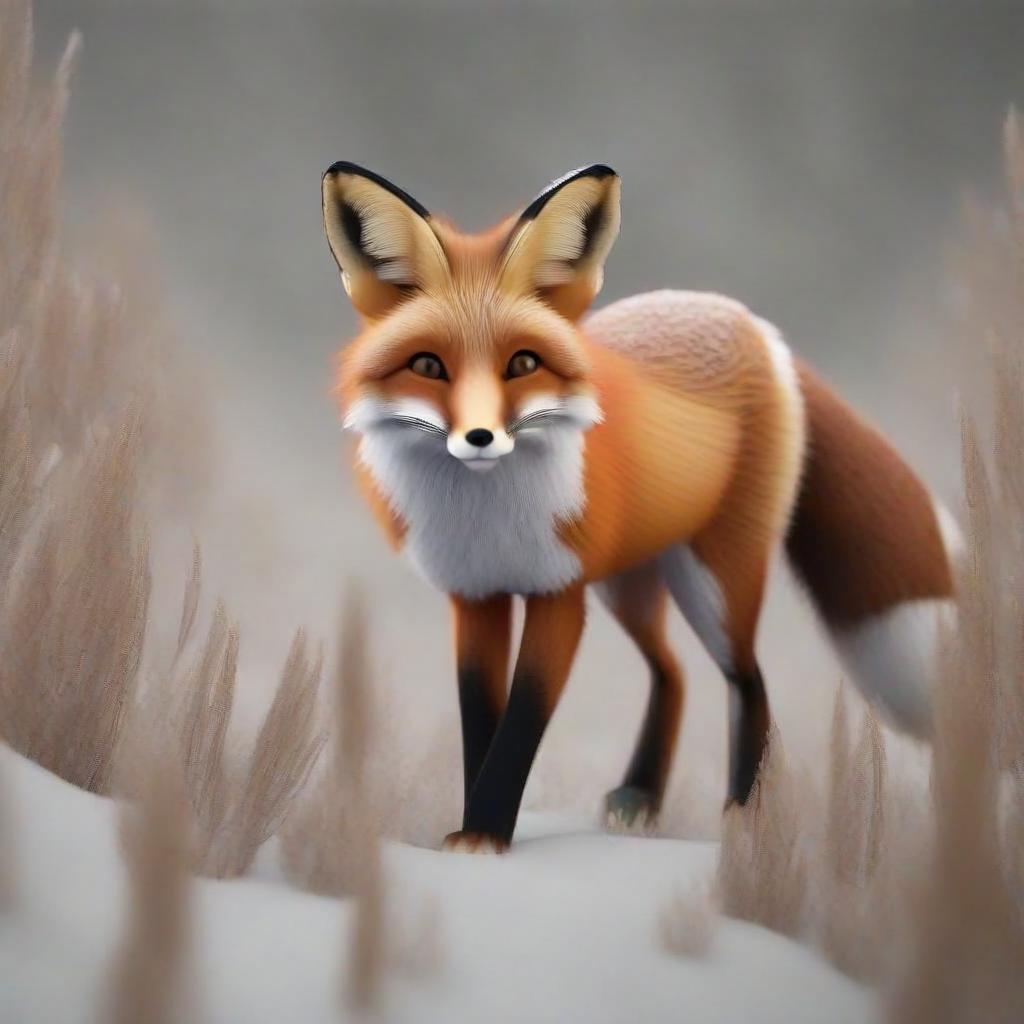}}
        \\
        \\

        \multicolumn{7}{c}{\textit{``a rendering of a fox, full body''}}

    \end{tabular}
    
    \caption{\textbf{Qualitative comparison.} We compare our method against several baselines: TI \cite{Gal2022AnII}, BLIP-diffusion \cite{Li2023BLIPDiffusionPS} and IP-adapter \cite{Ye2023IPAdapterTC} are able to follow the target prompts, but do not preserve a consistent identity. LoRA DB \cite{lora_diffusion} is able to maintain consistency, but it does not always follow the prompt. Furthermore, the character is generated in the same fixed pose. ELITE \cite{Wei2023ELITEEV} struggles with prompt following and also tends to generate deformed characters. On the other hand, our method is able to follow the prompt and maintain consistent identities, while generating the characters in different poses and viewing angles.}
    \label{fig:qualitative_comparison}
\end{figure*}

%% file: figures/quantitative_comparison/fig.tex
\begin{figure*}[t]
    \begin{tabular}{c @{\hspace{4\tabcolsep}} c}
        \begin{tikzpicture} [thick,scale=0.9, every node/.style={scale=1}]

            \def\MarkSize{.75em}
            \protected\def\ToWest#1{
              \llap{#1\kern\MarkSize}\phantom{#1}
            }
            \protected\def\ToSouth#1{
              \sbox0{#1}
              \smash{
                \rlap{
                  \kern-.5\dimexpr\wd0 + \MarkSize\relax
                  \lower\dimexpr.575em+\ht0\relax\copy0
                }
              }
              \hphantom{#1}
            }

            \begin{axis}[
                xlabel={Automatic prompt similarity ($\rightarrow$)}, 
                ylabel={Automatic identity consistency ($\rightarrow$)},
                compat=newest,
                xmax=0.211,
                ymax=0.9,
                width=9.5cm,
            ]
                \addplot[
                    scatter/classes={a={blue}, b={red}, c={green}, o={orange}},
                    scatter,
                    mark=*, 
                    only marks, 
                    scatter src=explicit symbolic,
                    nodes near coords*={\Label},
                    visualization depends on={value \thisrow{label} \as \Label}
                ] table [meta=class] {
                    x y class label
                    0.1788745292 0.7757437728 a \scriptsize{TI}
                    0.153539295 0.8775047056 a \scriptsize{LoRA DB}
                    0.1525347698 0.8537772401 a \scriptsize{ELITE}
                    0.1832544055 0.7527157749 a \ToSouth{\scriptsize{BLIP-diffusion}}
                    0.1695447199 0.8319113668 a \ToSouth{\scriptsize{IP-adapter}}
                    0.2042037527 0.7093161861 b \scriptsize{Ours single iter.}
                    0.1985133551 0.6938554896 b \ToSouth{\scriptsize{Ours w/o LoRA}}
                    0.1884804174 0.7574920302 b \scriptsize{Ours w reinit.}
                    0.1951256738811935 0.7333991974592209 b \ToSouth{\scriptsize{Ours w/o clustering}}
                    0.1657731262 0.8447209117 c \scriptsize{Ours}
                };
            \end{axis}
        \end{tikzpicture} 
        &

        \begin{tikzpicture} [thick,scale=0.9, every node/.style={scale=1}]
            \begin{axis}[
                xlabel={User prompt similarity ranking ($\rightarrow$)}, 
                ylabel={User identity consistency ($\rightarrow$)},
                compat=newest,
                ymax=3.8,
                width=9.5cm,
            ]
                \addplot[
                    scatter/classes={a={blue}, b={red}, c={green}},
                    scatter,
                    mark=*, 
                    only marks, 
                    scatter src=explicit symbolic,
                    nodes near coords*={\Label},
                    visualization depends on={value \thisrow{label} \as \Label}
                ] table [meta=class] {
                    x y class label
                    3.317934783 3.171195652 a \scriptsize{TI}
                    3.030797101 3.673913043 a \scriptsize{LoRA DB}
                    2.874094203 3.229166667 a \scriptsize{ELITE}
                    3.359601449 2.767210145 a \scriptsize{BLIP}
                    3.259057971 2.991847826 a \scriptsize{IP-Adapter}
                    3.302717391 3.487318841 c \scriptsize{Ours}
                };

            \end{axis}
        \end{tikzpicture}
    \end{tabular}

    \caption{\textbf{Quantitative Comparison and User Study.} (Left) We compared our method quantitatively with various baselines in terms of identity consistency and prompt similarity, as explained in \Cref{sec:comparisons}. LoRA DB and ELITE maintain high identity consistency, while sacrificing prompt similarity. TI and BLIP-diffusion achieve high prompt similarity but low identity consistency. We also ablated some components of our method: removing the clustering stage, reducing the optimizable representation, re-initializing the representation in each iteration and performing only a single iteration. All of the ablated cases resulted in a significant degradation of consistency. (Right) The user study rankings also demonstrate that our method is balancing between identity consistency and prompt similarity.}
    \label{fig:quantitative_comparison_and_user_study}
\end{figure*}
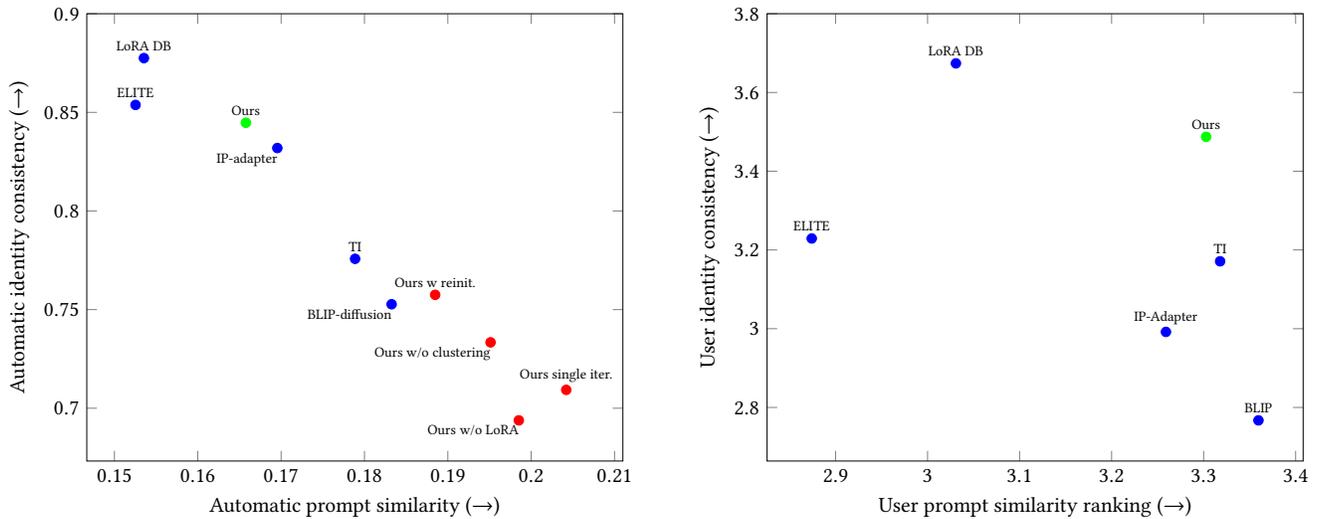

%% file: figures/applications/fig.tex
\begin{figure}[t]
    \centering
    \setlength{\tabcolsep}{1pt}
    \renewcommand{\arraystretch}{0.65}
    \setlength{\ww}{0.224\columnwidth}
    \begin{tabular}{ccccc}

        &
        \multicolumn{4}{c}{\scriptsize{\textit{``This is a story about Jasper, \colora{a cute mink with a brown jacket and red pants.}}}}
        \\
        &
        \multicolumn{4}{c}{\scriptsize{\textit{Jasper started his day by \colorb{jogging on the beach}, and afterwards, \colorc{he enjoyed a}}}}
        \\
        &
        \multicolumn{4}{c}{\scriptsize{\textit{\colorc{coffee meetup with a friend in the heart of New York City}. As the day drew}}}
        \\
        &
        \multicolumn{4}{c}{\scriptsize{\textit{to a close, \colord{he settled into his cozy apartment to review a paper}.''}}}
        \\

        \rotatebox[origin=c]{90}{\scriptsize{(a) Story}} 
        \rotatebox[origin=c]{90}{\scriptsize{illustration}}
        &
        {\includegraphics[valign=c, width=\ww]{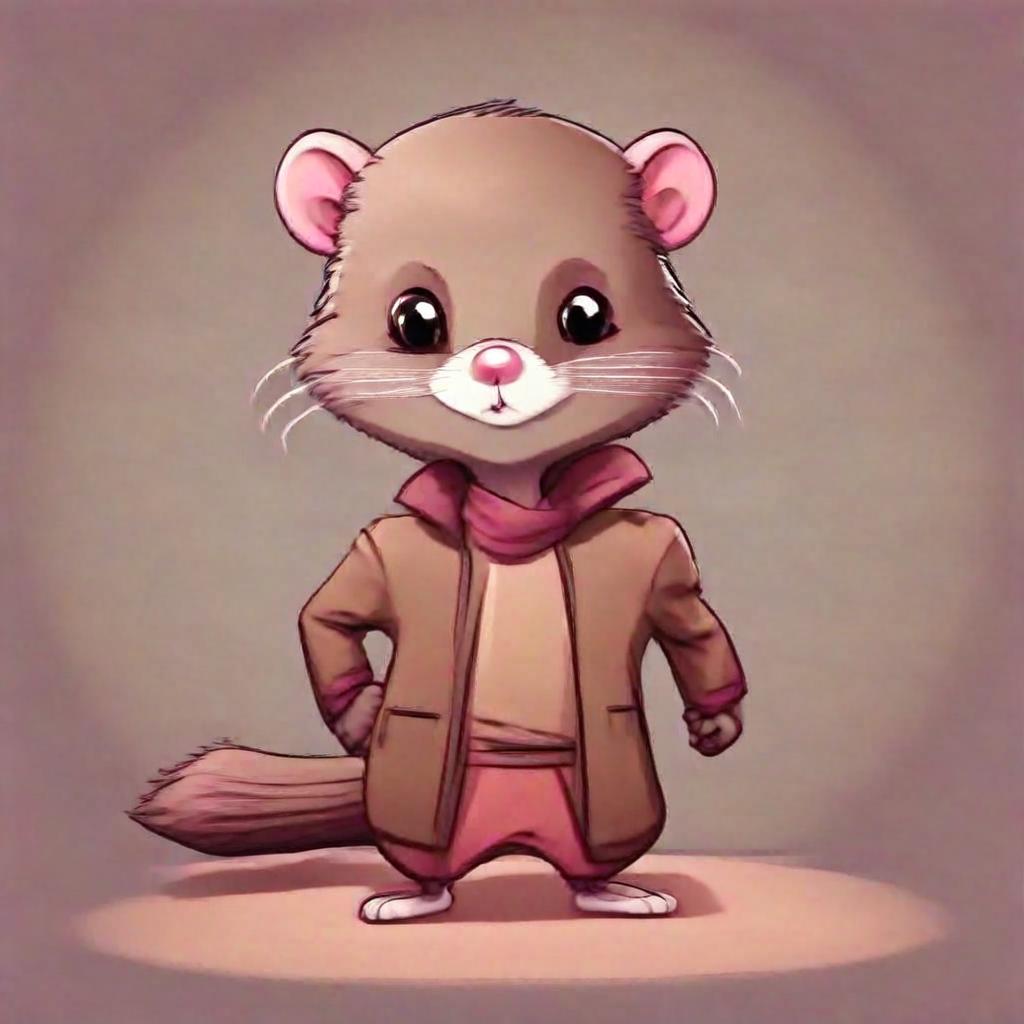}} &
        {\includegraphics[valign=c, width=\ww]{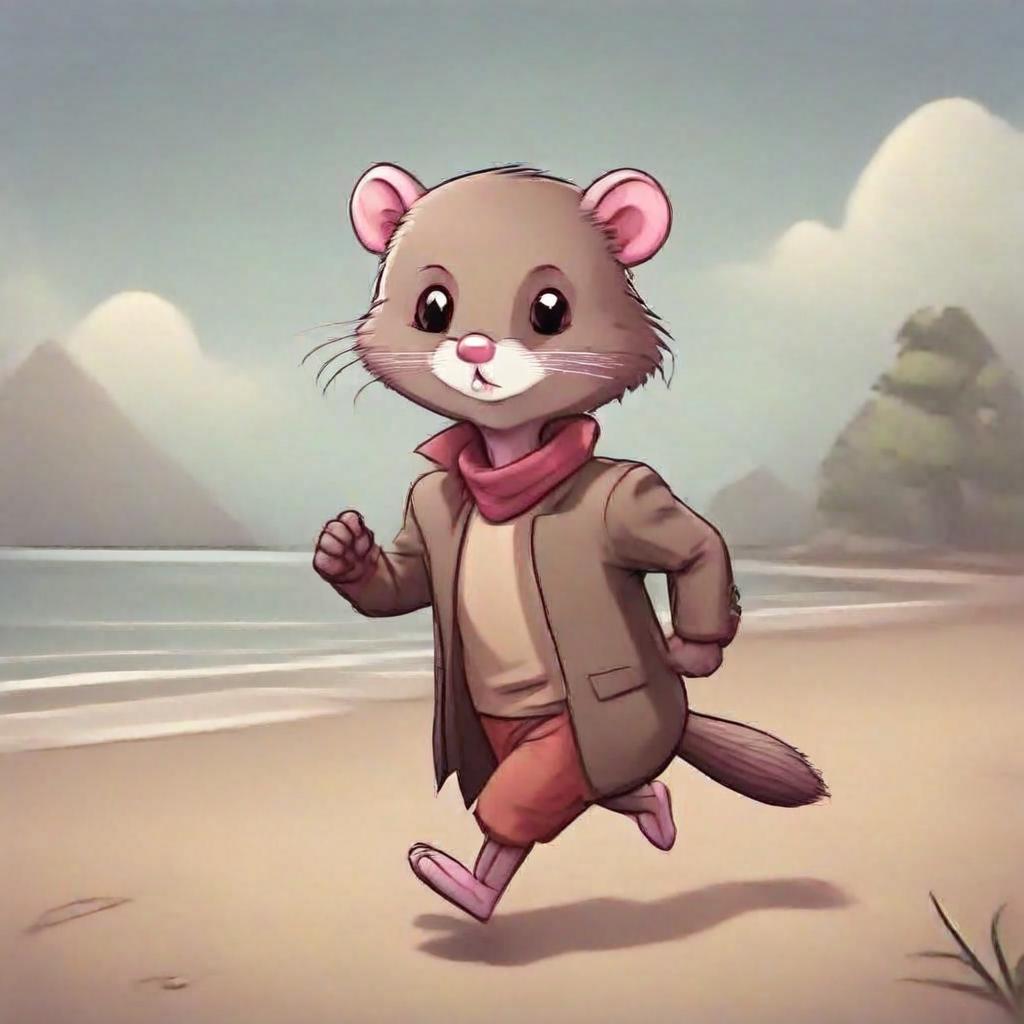}} &
        {\includegraphics[valign=c, width=\ww]{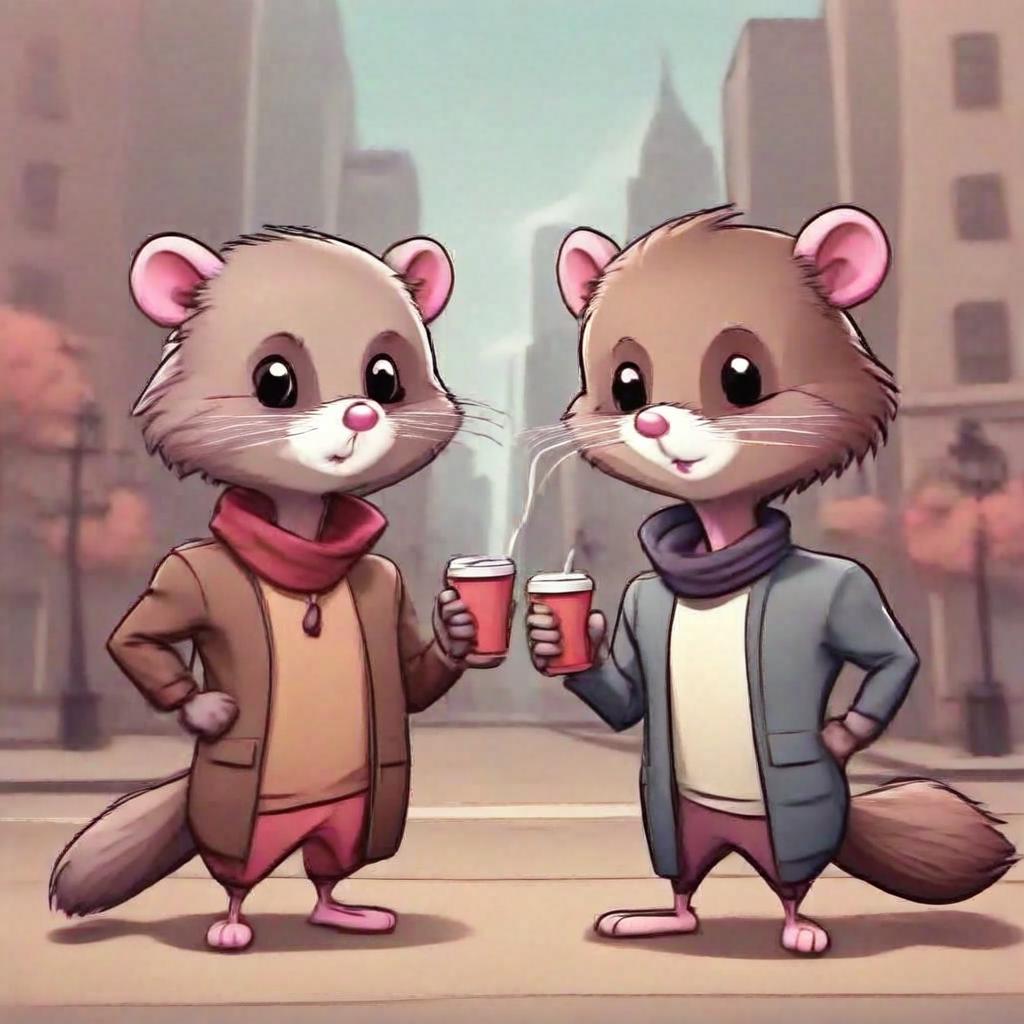}} &
        {\includegraphics[valign=c, width=\ww]{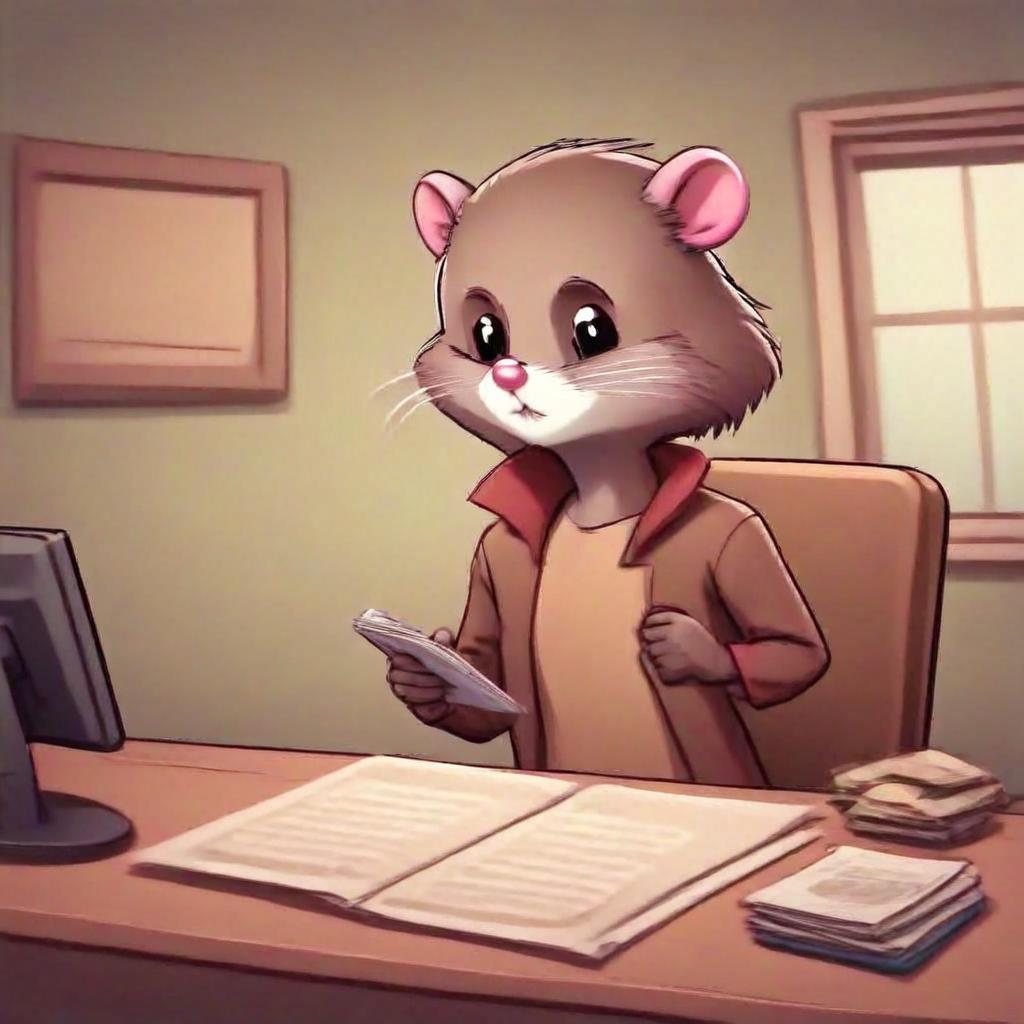}}
        \\

        &
        \scriptsize{\colora{Scene 1}} &
        \scriptsize{\colorb{Scene 2}} &
        \scriptsize{\colorc{Scene 3}} &
        \scriptsize{\colord{Scene 4}}
        \vspace{0.1cm}
        \\

        \multicolumn{5}{c}{\scriptsize{\textit{``a Plasticine of a cute baby cat with big eyes''}}}
        \\

        \rotatebox[origin=c]{90}{\scriptsize{(b) Local}} 
        \rotatebox[origin=c]{90}{\scriptsize{image editing}}
        &
        {\includegraphics[valign=c, width=\ww]{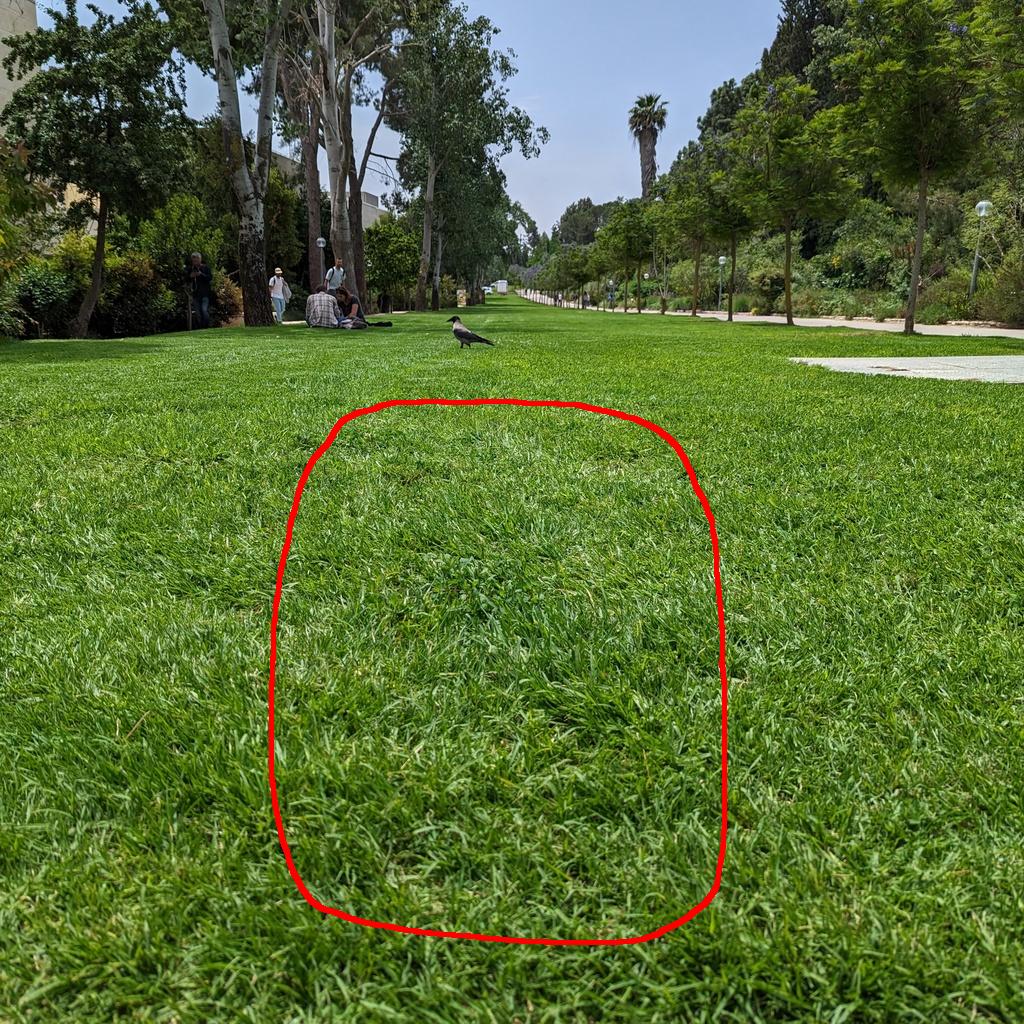}} &
        {\includegraphics[valign=c, width=\ww]{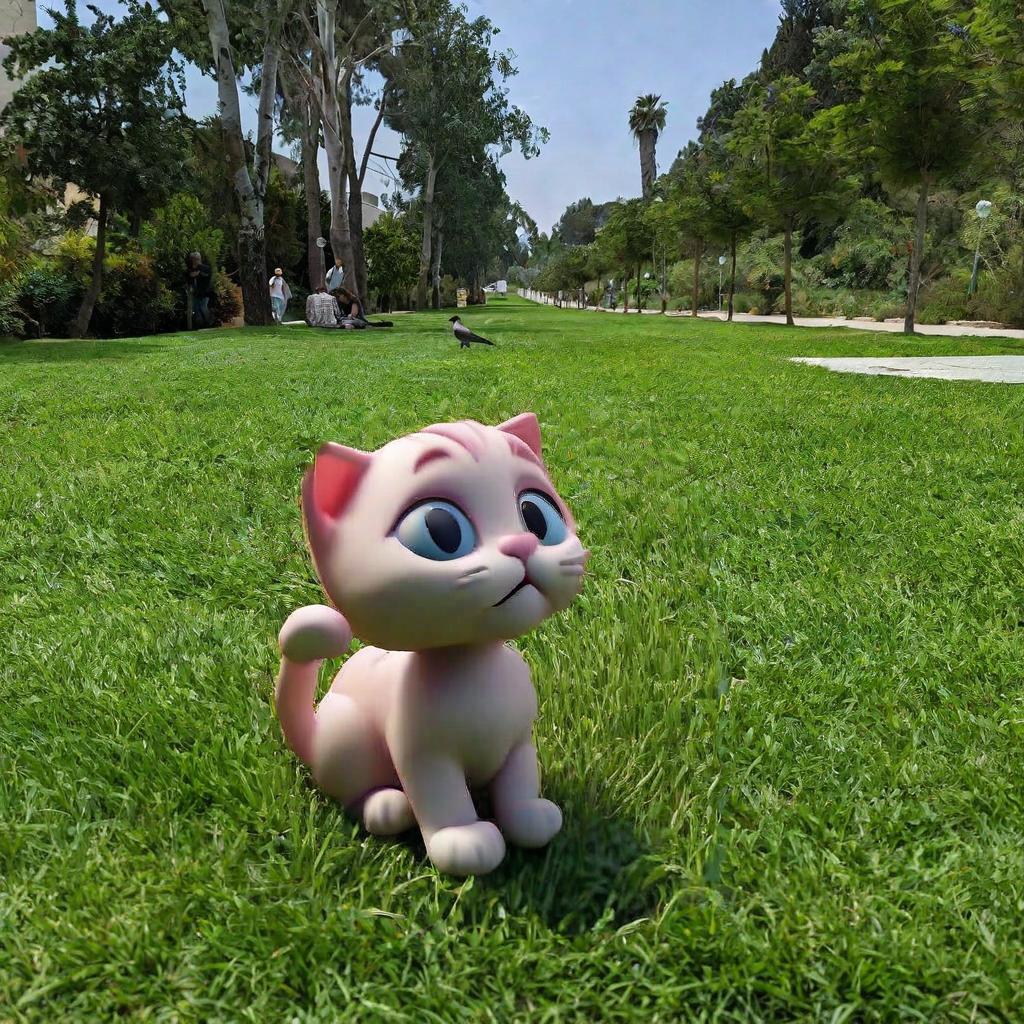}} &
        {\includegraphics[valign=c, width=\ww]{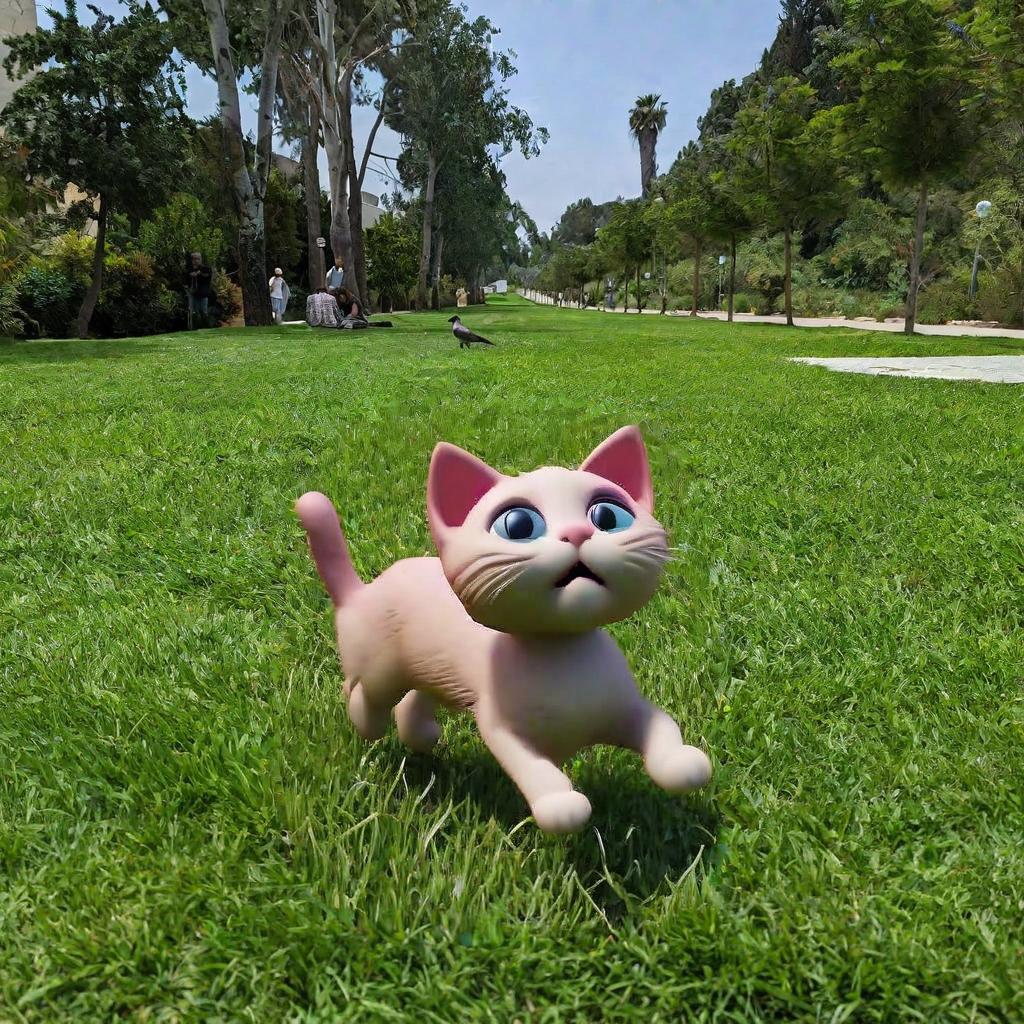}} &
        {\includegraphics[valign=c, width=\ww]{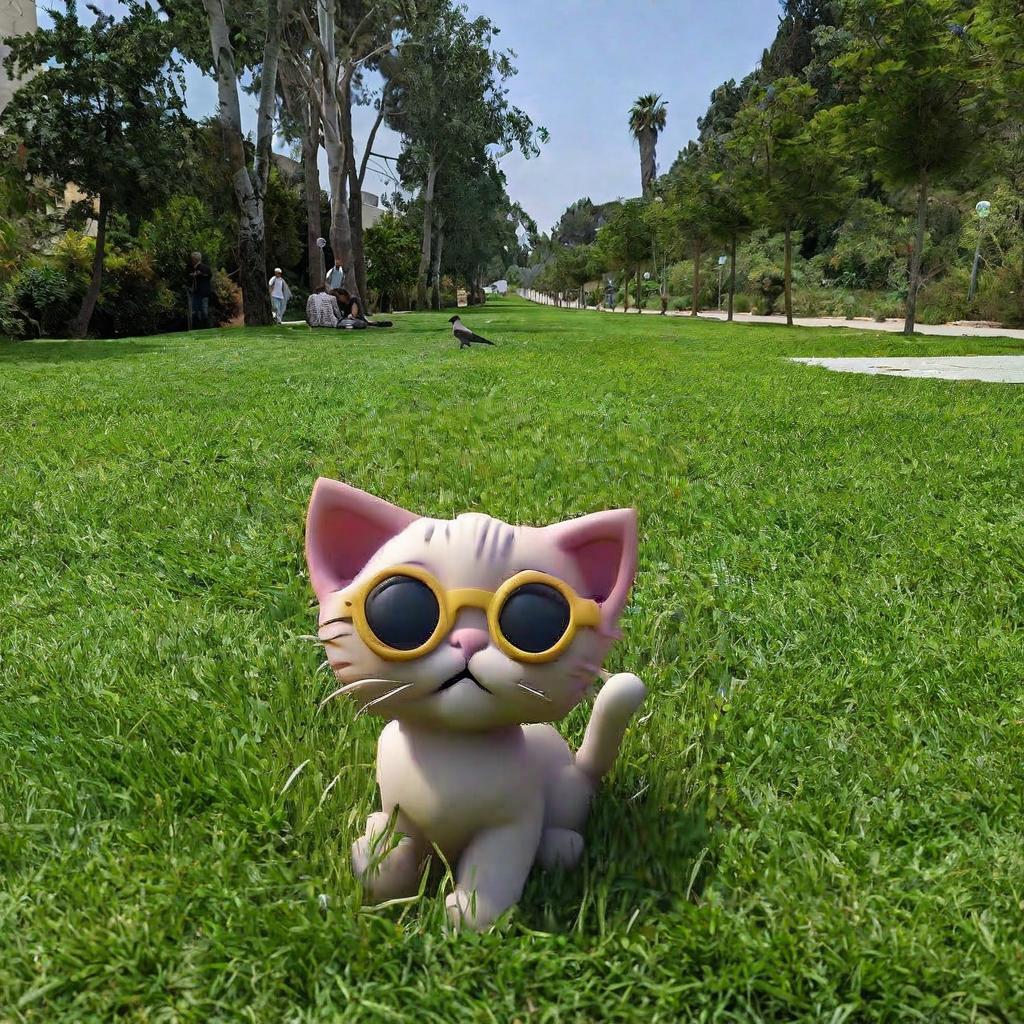}}
        \\

        &
        \scriptsize{Image + mask} &
        \scriptsize{\textit{``sitting''}} &
        \scriptsize{\textit{`` jumping''}} &
        \scriptsize{\textit{``wearing}}
        \\

        &&&&
        \scriptsize{\textit{sunglasses''}}
        \vspace{0.1cm}
        \\

        \multicolumn{5}{c}{\scriptsize{\textit{``a photo of a ginger woman with long hair''}}}
        \\

        \rotatebox[origin=c]{90}{\scriptsize{(c) Additional}} 
        \rotatebox[origin=c]{90}{\scriptsize{pose control}}
        &
        {\includegraphics[valign=c, width=\ww]{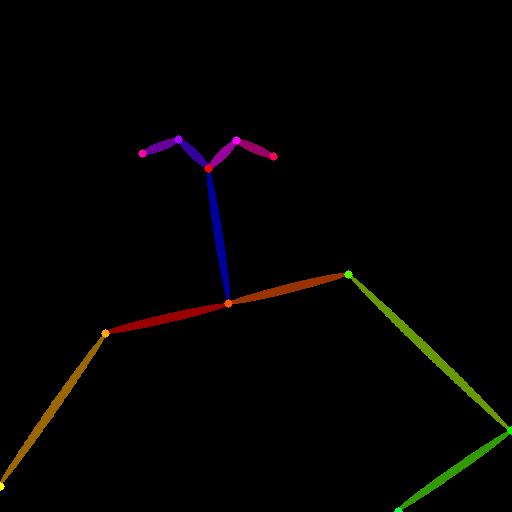}} &
        {\includegraphics[valign=c, width=\ww]{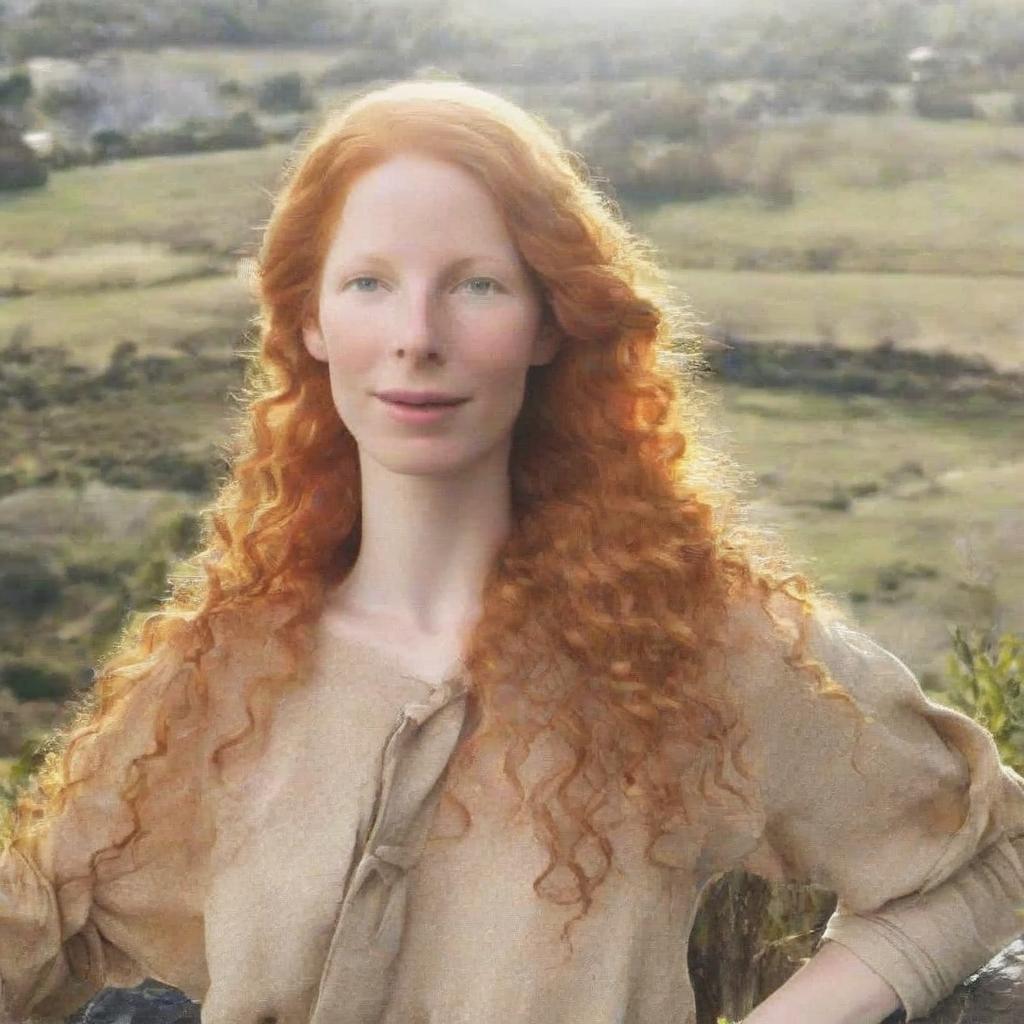}} &
        {\includegraphics[valign=c, width=\ww]{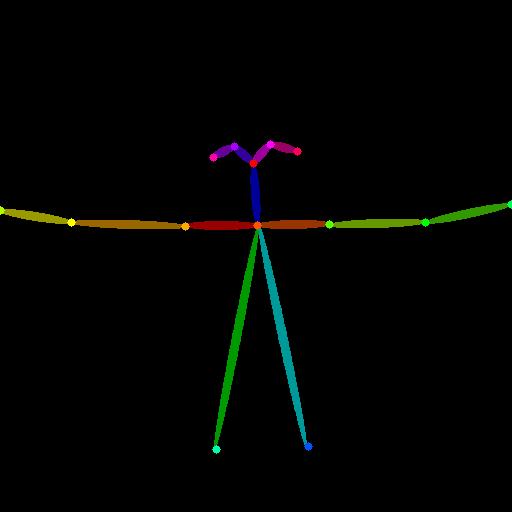}} &
        {\includegraphics[valign=c, width=\ww]{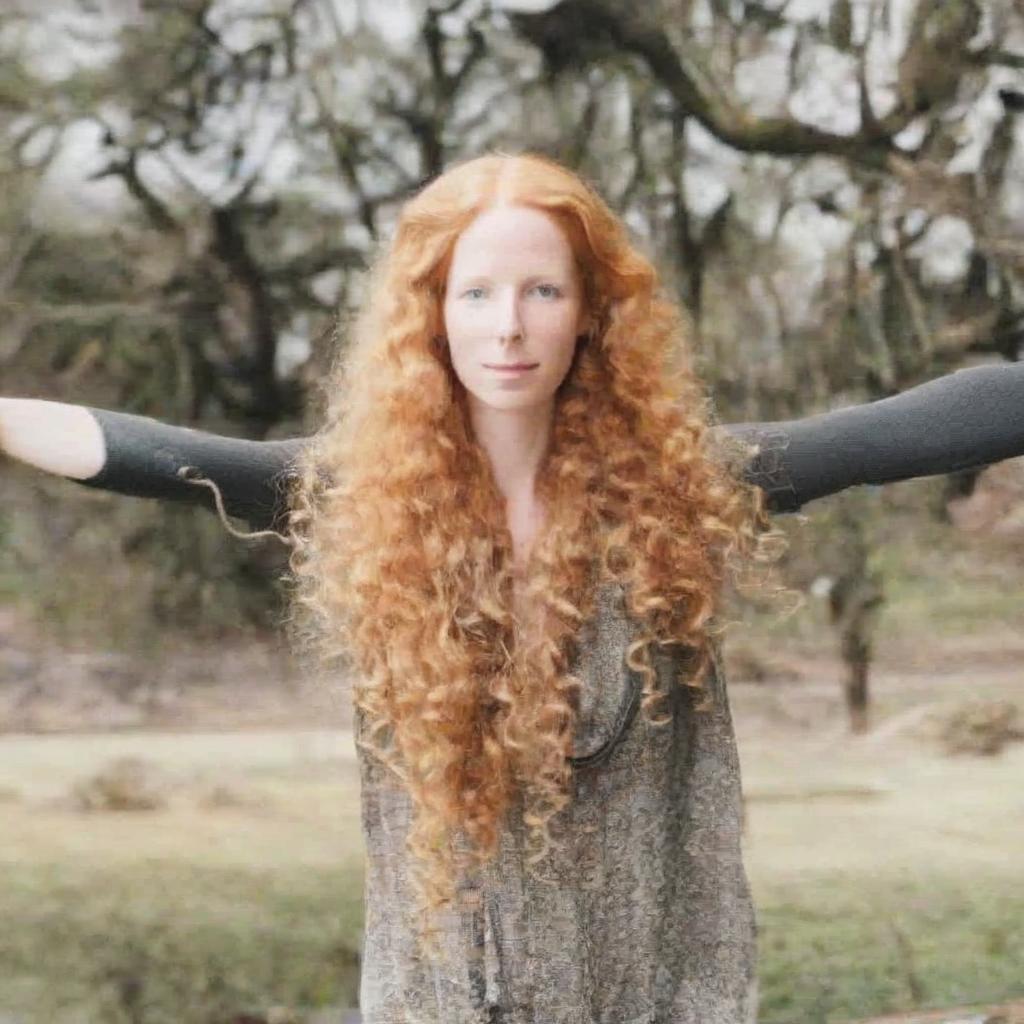}}
        \\

        &
        \scriptsize{Input Pose 1} &
        \scriptsize{Result 1} &
        \scriptsize{Input Pose 2} &
        \scriptsize{Result 2}
        \\

    \end{tabular}
    
    \caption{\textbf{Applications.} Our method can be used for various applications: (a) Illustrating a full story with the same consistent character. (b) Local text-driven image editing via integration with Blended Latent Diffusion~\cite{avrahami2023blendedlatent,blended_2022_CVPR}. (c) Generating a consistent character with an additional pose control via integration with ControlNet~\cite{zhang2023controlnet}.}
    \label{fig:applications}
\end{figure}

%% file: sections/limitations_and_conclusions.tex
\section{Limitations and Conclusions}
\label{sec:limitations_and_conclusions}

\input{figures/limitations/fig.tex}

We found our method to suffer from the following limitations: (a) Inconsistent identity --- in some cases, our method is not able to converge to a fully consistent identity (without overfitting). As demonstrated in \Cref{fig:limitations}(a), when trying to generate a portrait of a robot, our method generated robots with slightly different colors and shapes (\eg, different arms). This may result from a prompt that is too general, for which identity clustering (\Cref{sec:identity_clustering}) is not able to find a sufficiently cohesive set. (b) Inconsistent supporting characters/elements --- although our method is able to find a consistent identity for the character described by the input prompt, the identities of other characters, related to the input character (\eg, their pet), might be inconsistent. For example, in \Cref{fig:limitations}(b) the input prompt $p$ to our method described only the girl, and when asked to generate the girl with her cat, different cats were generated. In addition, our framework does not support finding multiple concepts concurrently \cite{Avrahami2023BreakASceneEM}. (c) Spurious attributes --- we found that in some cases, our method binds additional attributes, which are not part of the input text prompt, with the final identity of the character. For example, in \Cref{fig:limitations}(c), the input text prompt was \textit{``a sticker of a ginger cat''}, however, our method added green leaves to the generated sticker, even though it was not asked to do so. This stems from the stochastic nature of the text-to-image model --- the model added these leaves in some of the stickers generated during the identity clustering stage (\Cref{sec:identity_clustering}), and the stickers containing the leaves happened to form the most cohesive set $\ccand$. One way to mitigate it is to let the user choose one of the most cohesive clusters according to their preferences, instead of selecting it automatically. (d) Significant computational cost --- each iteration of our method involves generating a large number of images, and learning the identity of the most cohesive cluster. It takes about 20 minutes to converge into a consistent identity. Reducing the computational costs is an appealing direction for further research. (e) Simplistic characters --- we found that our method tends to generate simplistic scences (single and mostly centered objects), which may be caused by the ``averaging'' effect of the identity extraction stage, as explained in \Cref{sec:identity_extraction}.

In conclusion, in this paper we offered the first fully-automated solution to the problem of consistent character generation. We hope that our work will pave the way for future advancements, as we believe this technology of consistent character generation may have a disruptive effect on numerous sectors, including education, storytelling, entertainment, fashion, brand design, advertising, and more.

%% file: figures/limitations/fig.tex
\begin{figure}[t]
    \vspace{0.8cm}  %
    \centering
    \setlength{\tabcolsep}{1pt}
    \renewcommand{\arraystretch}{0.7}
    \setlength{\ww}{0.224\columnwidth}
    \begin{tabular}{ccccc}

        \rotatebox[origin=c]{90}{\scriptsize{(a) Inconsistent}} 
        \rotatebox[origin=c]{90}{\scriptsize{identity}} 
        &
        {\includegraphics[valign=c, width=\ww]{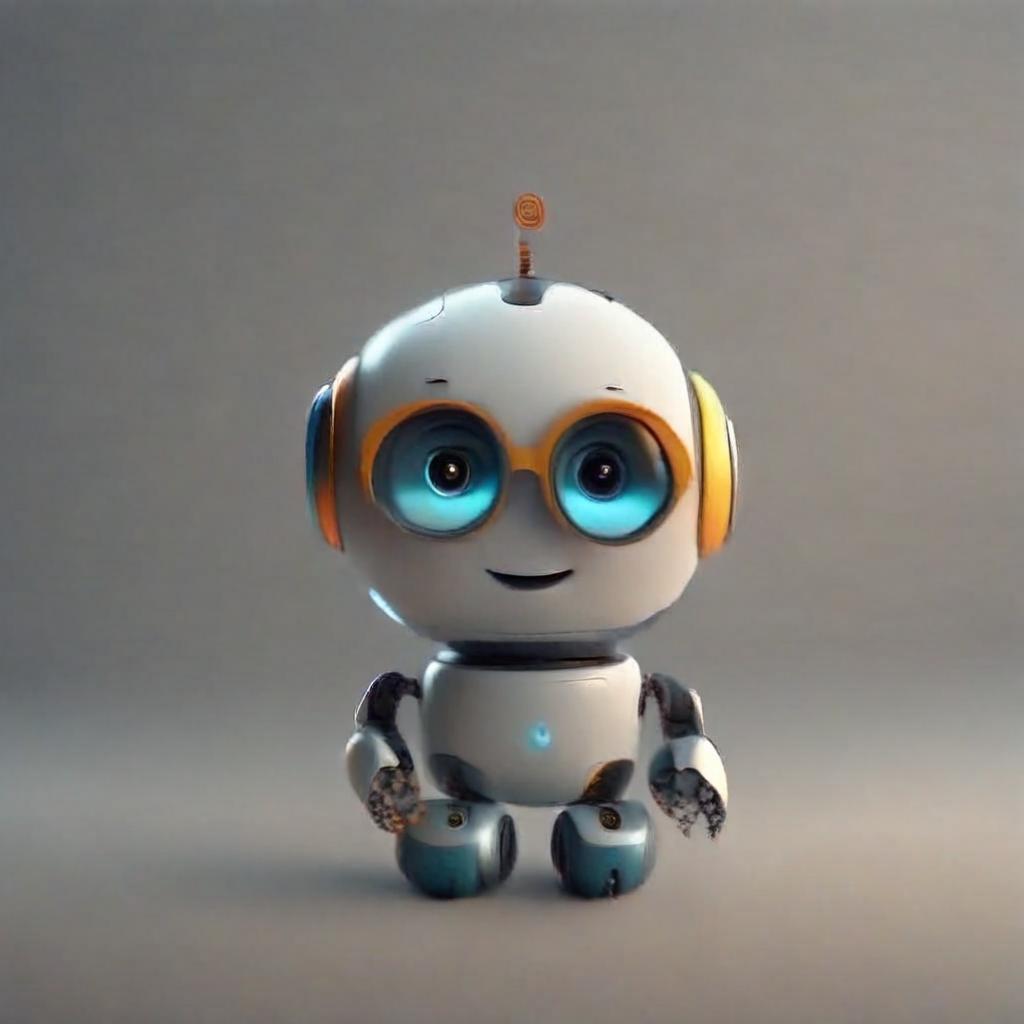}} &
        {\includegraphics[valign=c, width=\ww]{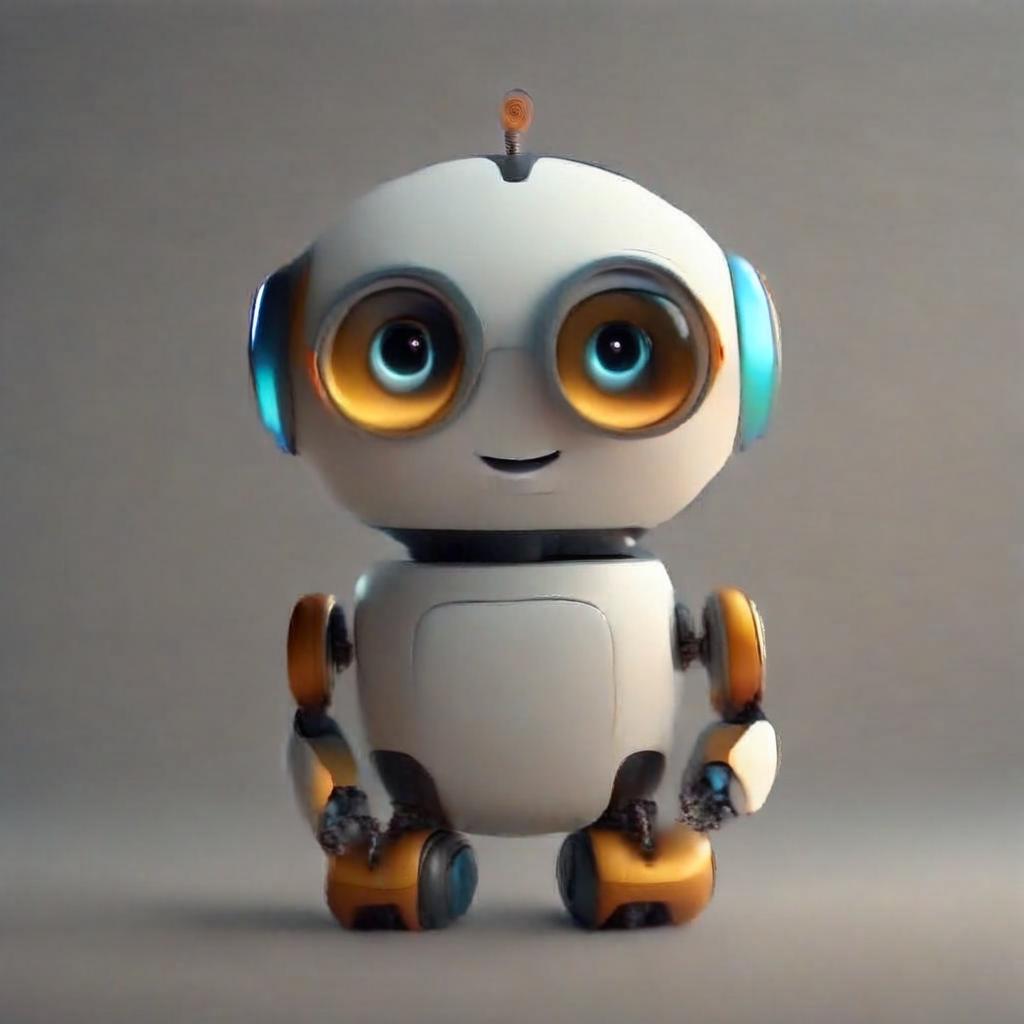}} &
        {\includegraphics[valign=c, width=\ww]{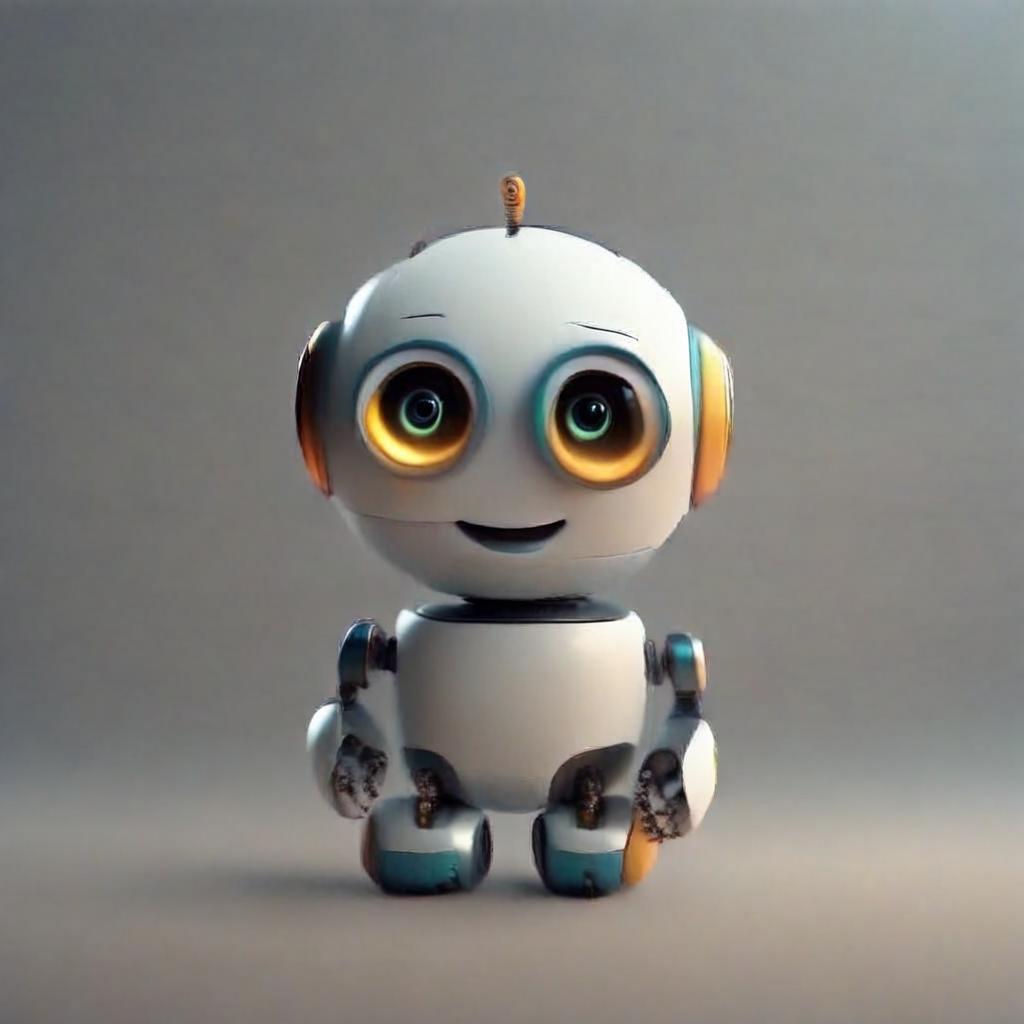}} &
        {\includegraphics[valign=c, width=\ww]{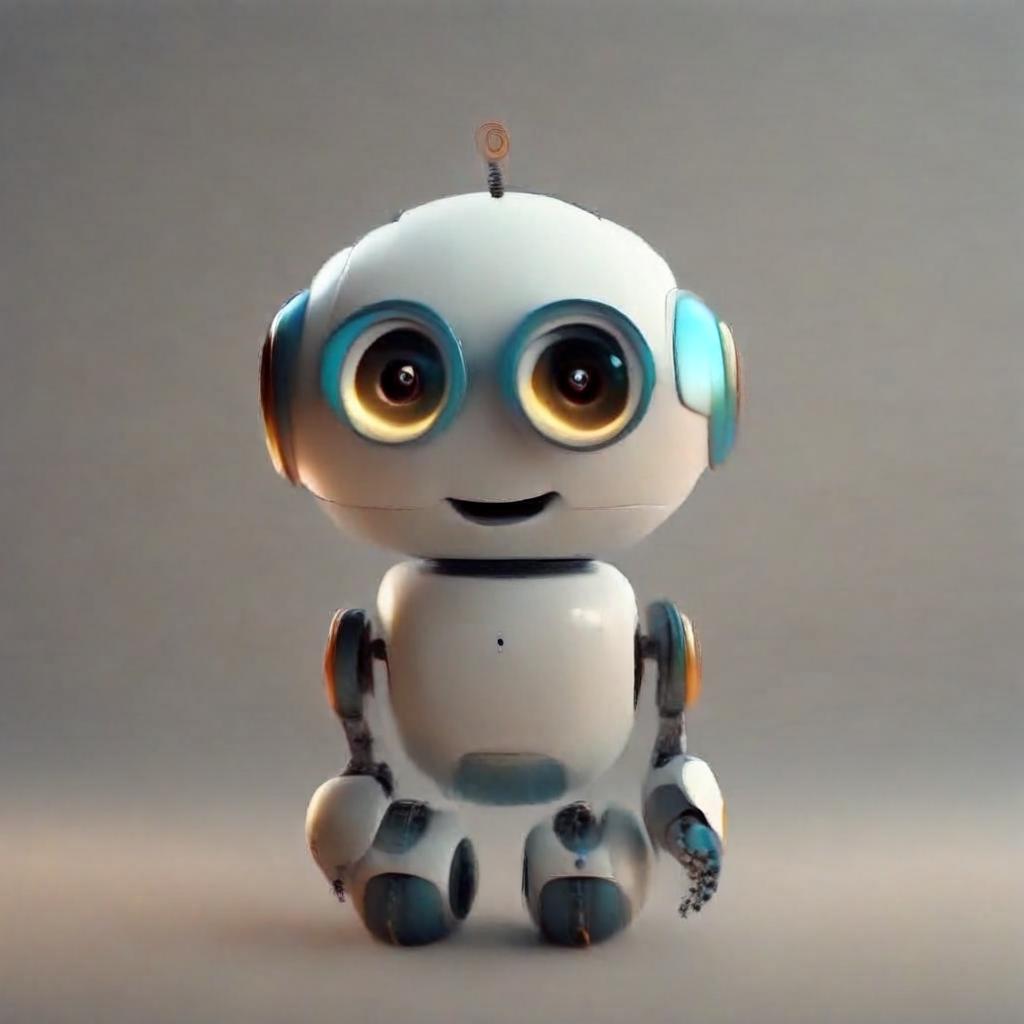}}
        \\
        
        &
        \multicolumn{4}{c}{\scriptsize{\textit{``a portrait of a round robot with glasses ...''}}}
        \vspace{0.1cm}
        \\

        \rotatebox[origin=c]{90}{\scriptsize{(b) Inconsistent}} 
        \rotatebox[origin=c]{90}{\scriptsize{supporting elements}} 
        &
        {\includegraphics[valign=c, width=\ww]{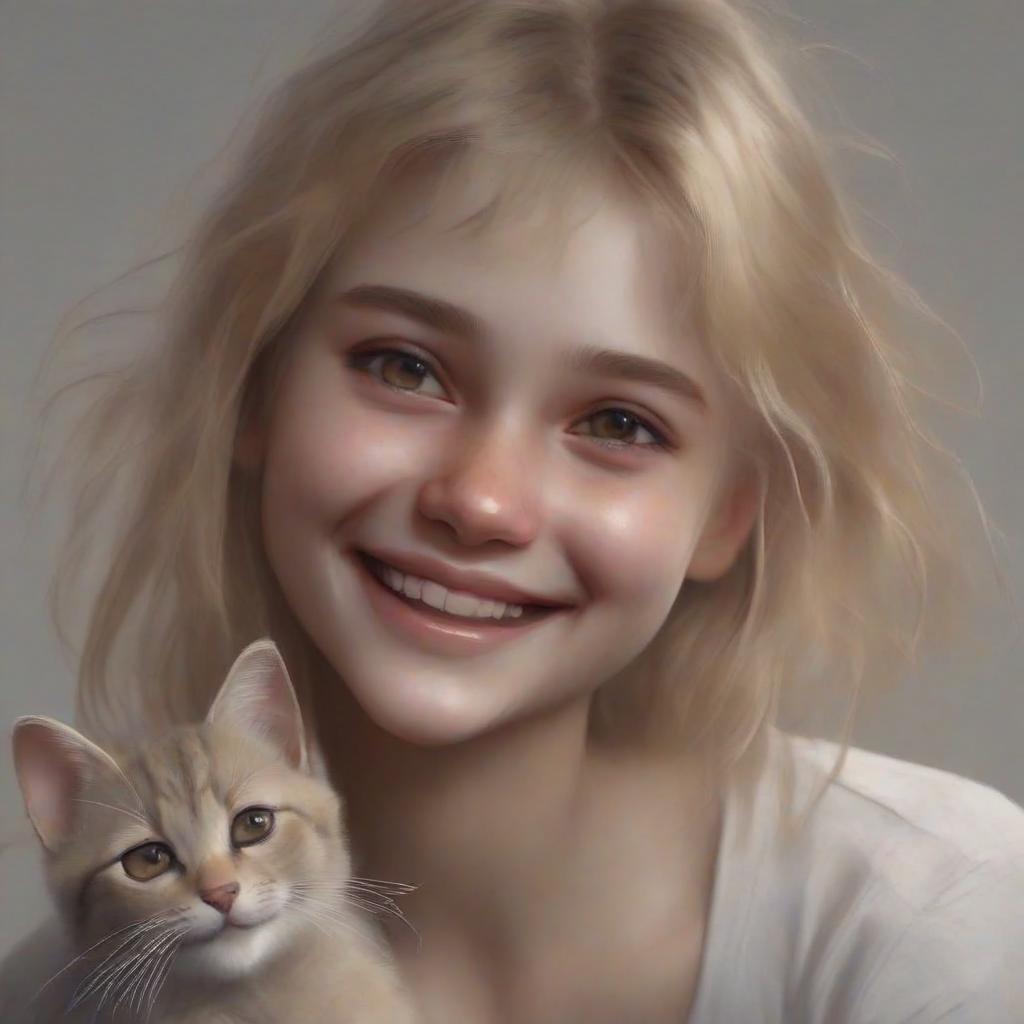}} &
        {\includegraphics[valign=c, width=\ww]{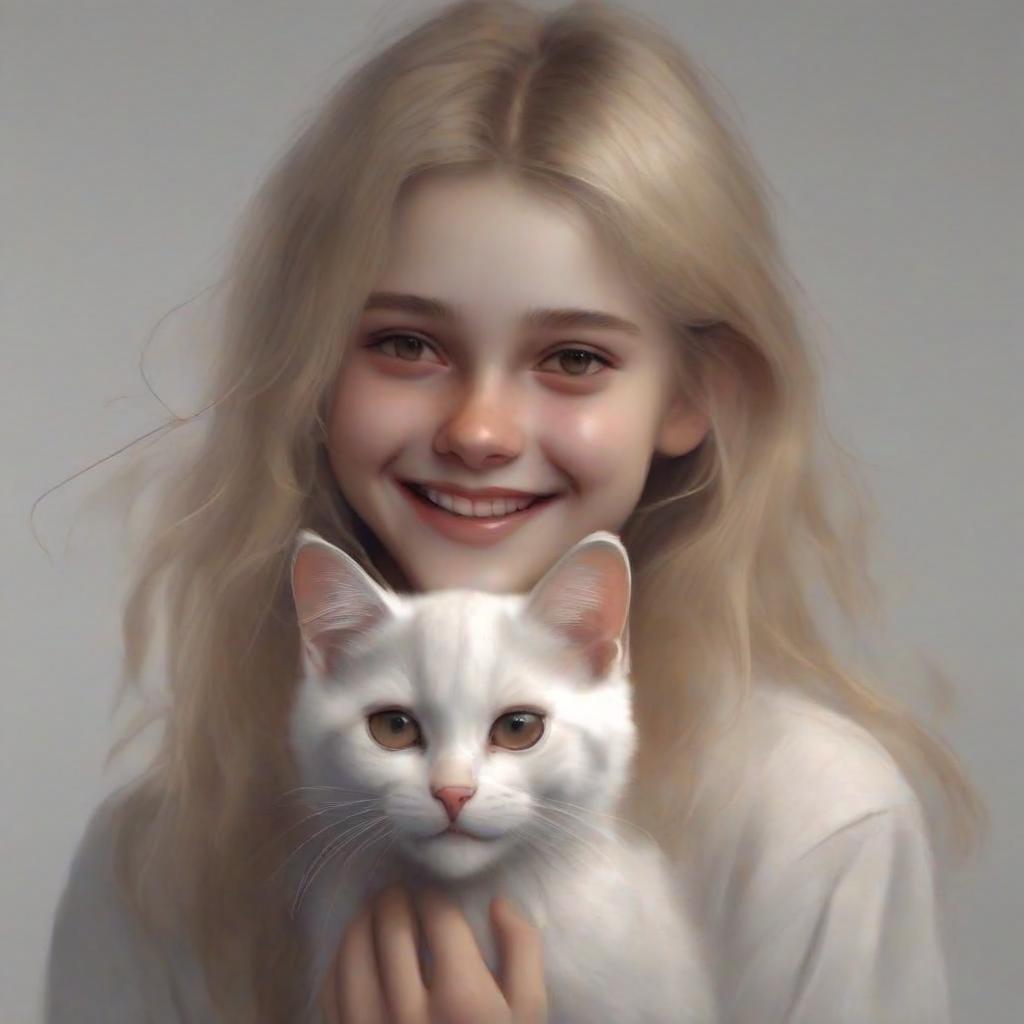}} &
        {\includegraphics[valign=c, width=\ww]{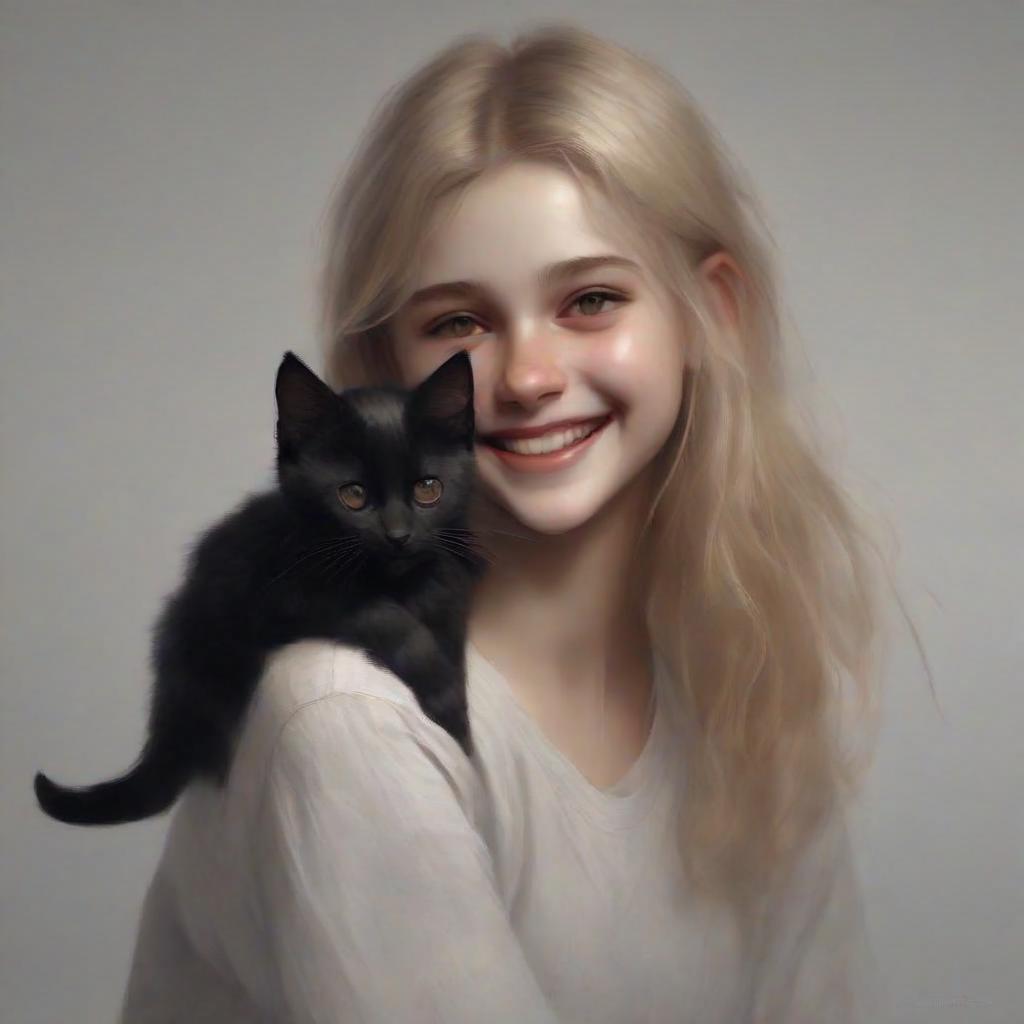}} &
        {\includegraphics[valign=c, width=\ww]{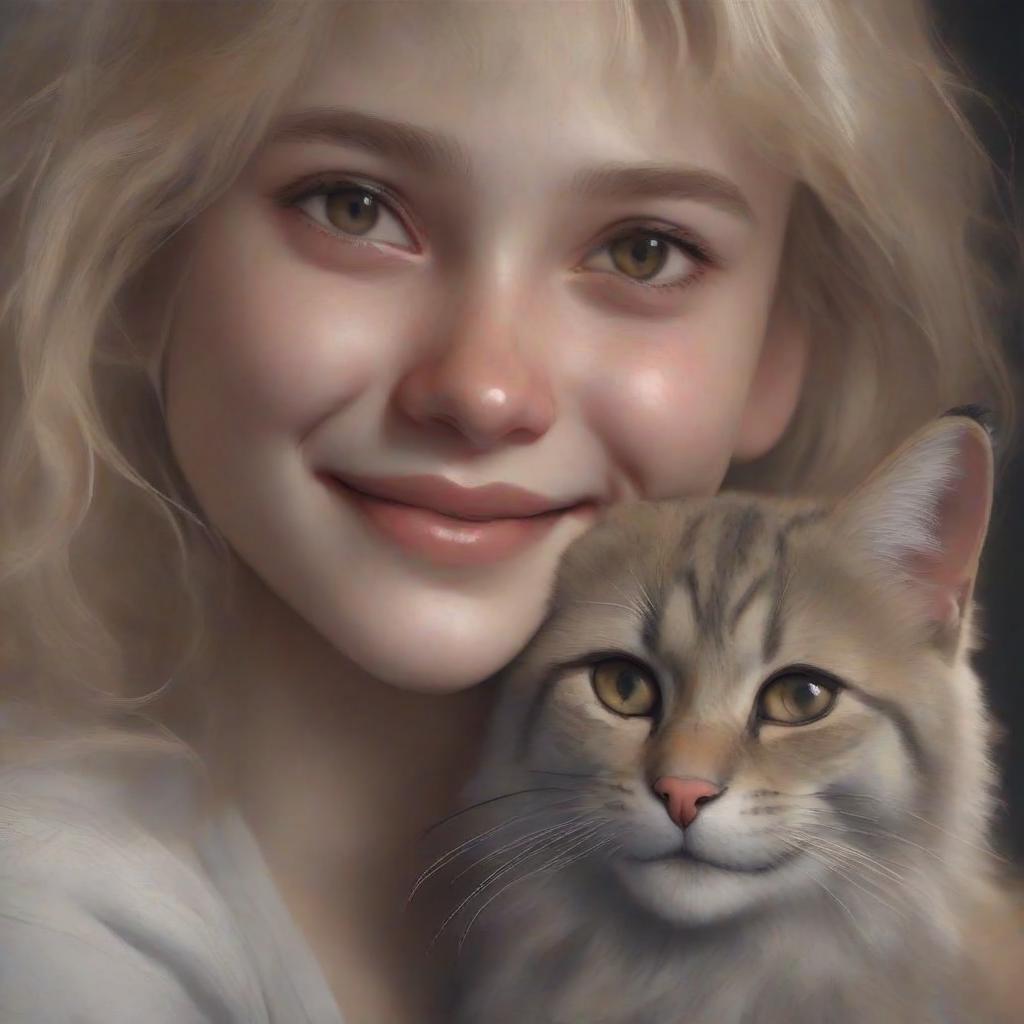}}
        \\

        &
        \multicolumn{4}{c}{\scriptsize{\textit{``a hyper-realistic digital painting of a happy girl, brown eyes...''}}}
        \\
        &
        \multicolumn{4}{c}{\scriptsize{+ \textit{``with her cat''}}}
        \vspace{0.1cm}
        \\

        \rotatebox[origin=c]{90}{\scriptsize{(c) Spurious}}
        \rotatebox[origin=c]{90}{\scriptsize{attributes}}
        &
        {\includegraphics[valign=c, width=\ww]{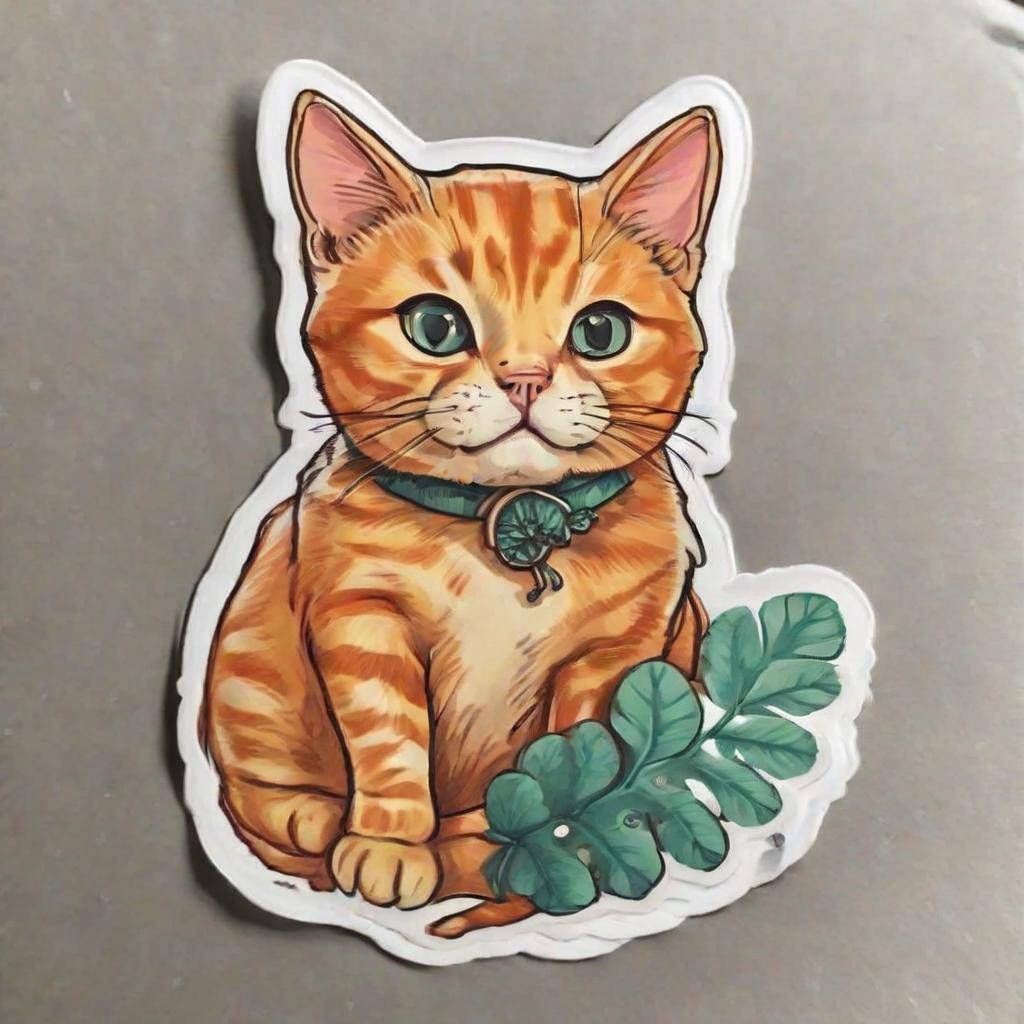}} &
        {\includegraphics[valign=c, width=\ww]{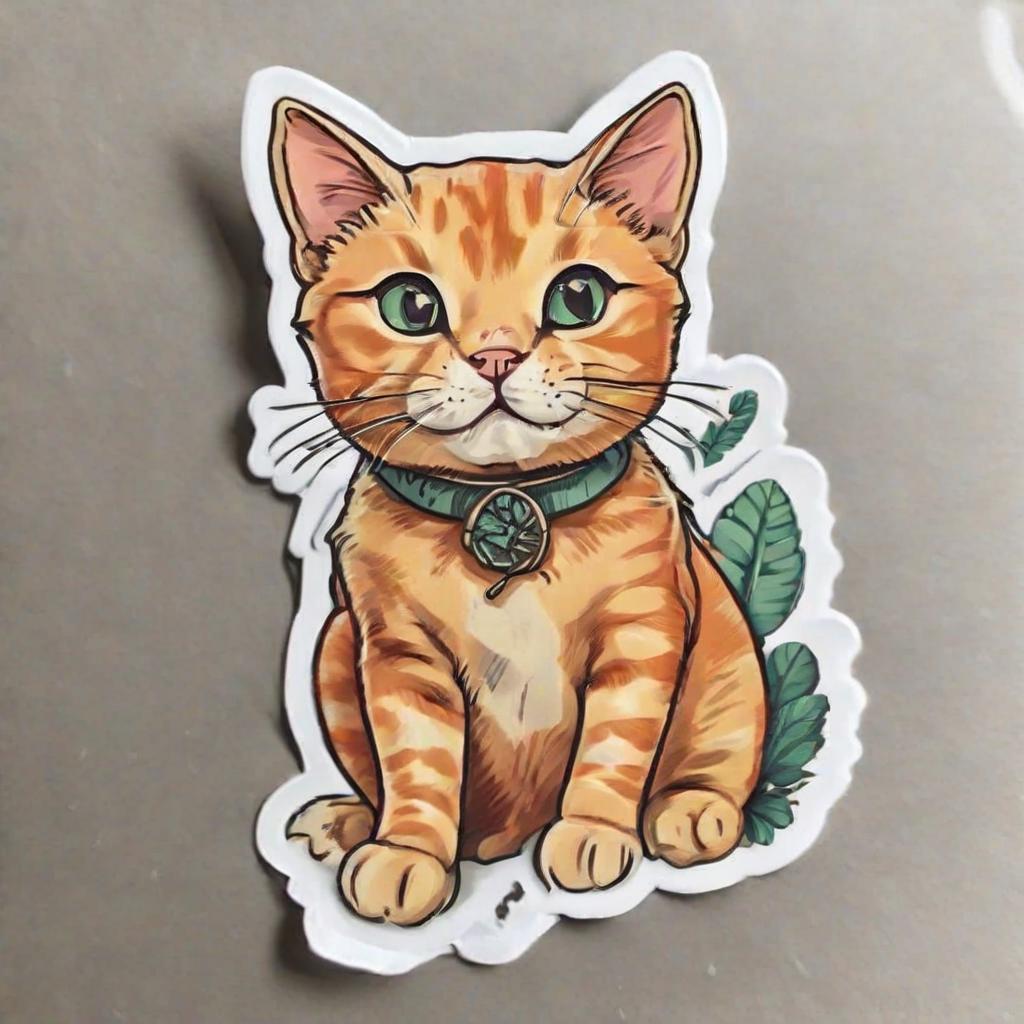}} &
        {\includegraphics[valign=c, width=\ww]{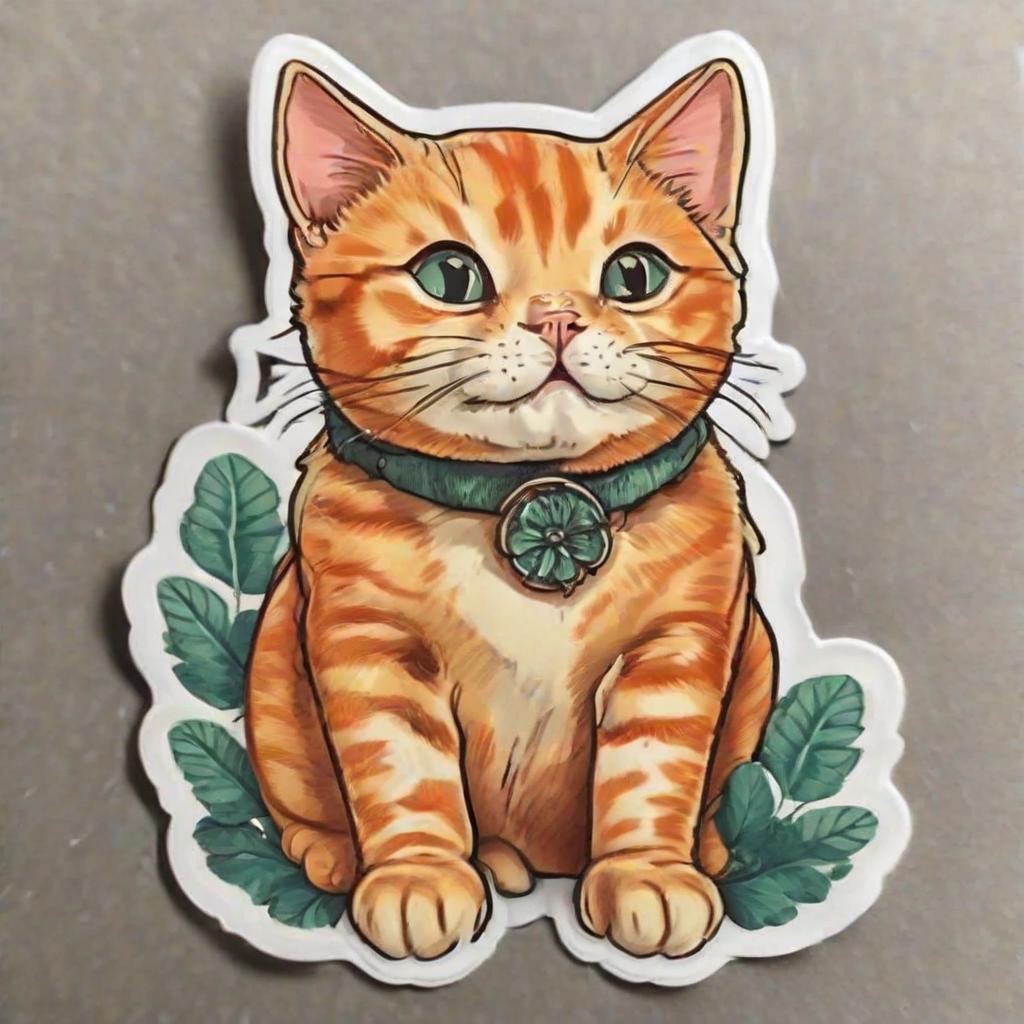}} &
        {\includegraphics[valign=c, width=\ww]{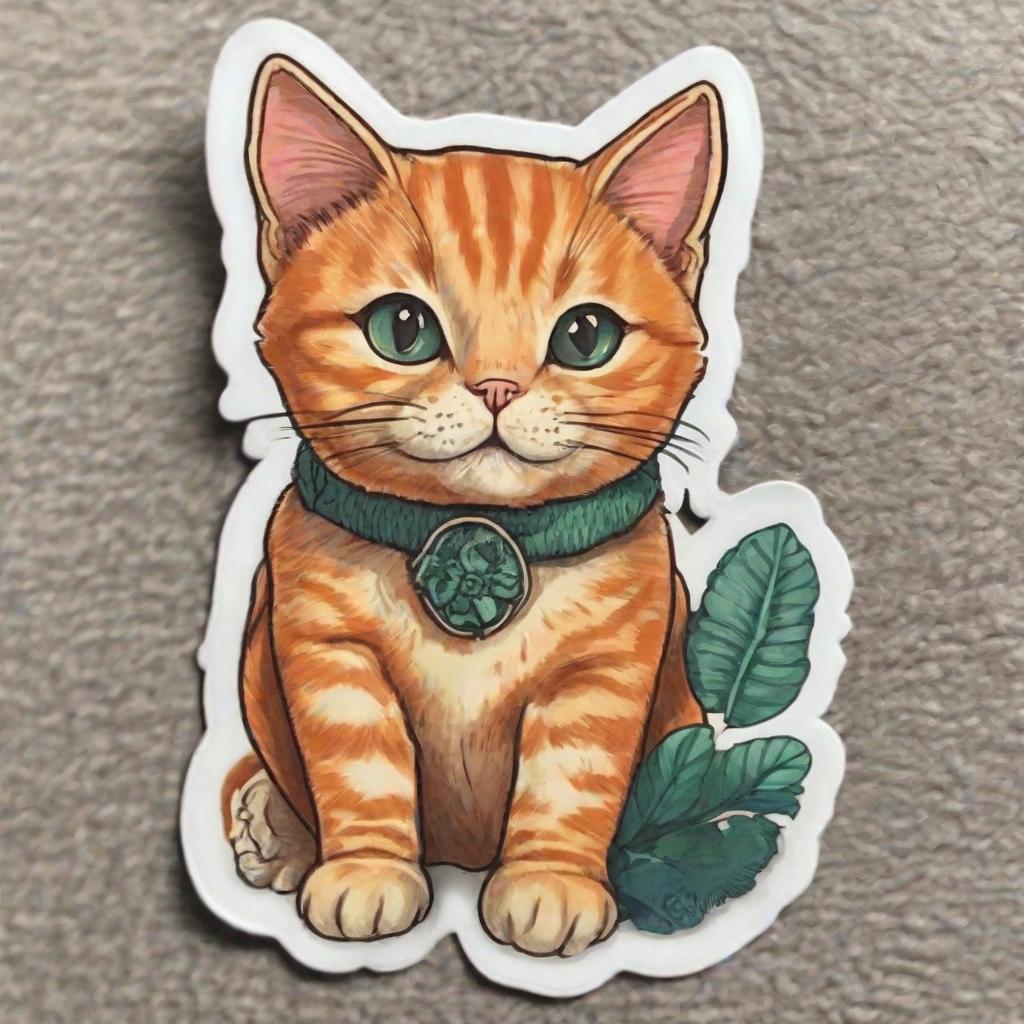}}
        \\

        &
        \multicolumn{4}{c}{\scriptsize{\textit{``a sticker of a ginger cat''}}}

    \end{tabular}
    
    \caption{\textbf{Limitations.} Our method suffers from the following limitations: (a) in some cases, our method is not able to converge to a fully consistent identity --- notice slight color and arm shape changes. (b) Our method is not able to associate a consistent identity to a supporting character that may appear with the main extracted character, for example our method generates different cats for the same girl. (c) Our method sometimes adds spurious attributes to the character, that were not present in the text prompt. For example, it learns to associate green leaves with the cat sticker.}
    \label{fig:limitations}
\end{figure}

%% file: figures/user_control/fig.tex
\begin{figure*}[t]
    \centering
    \includegraphics[width=1\linewidth]{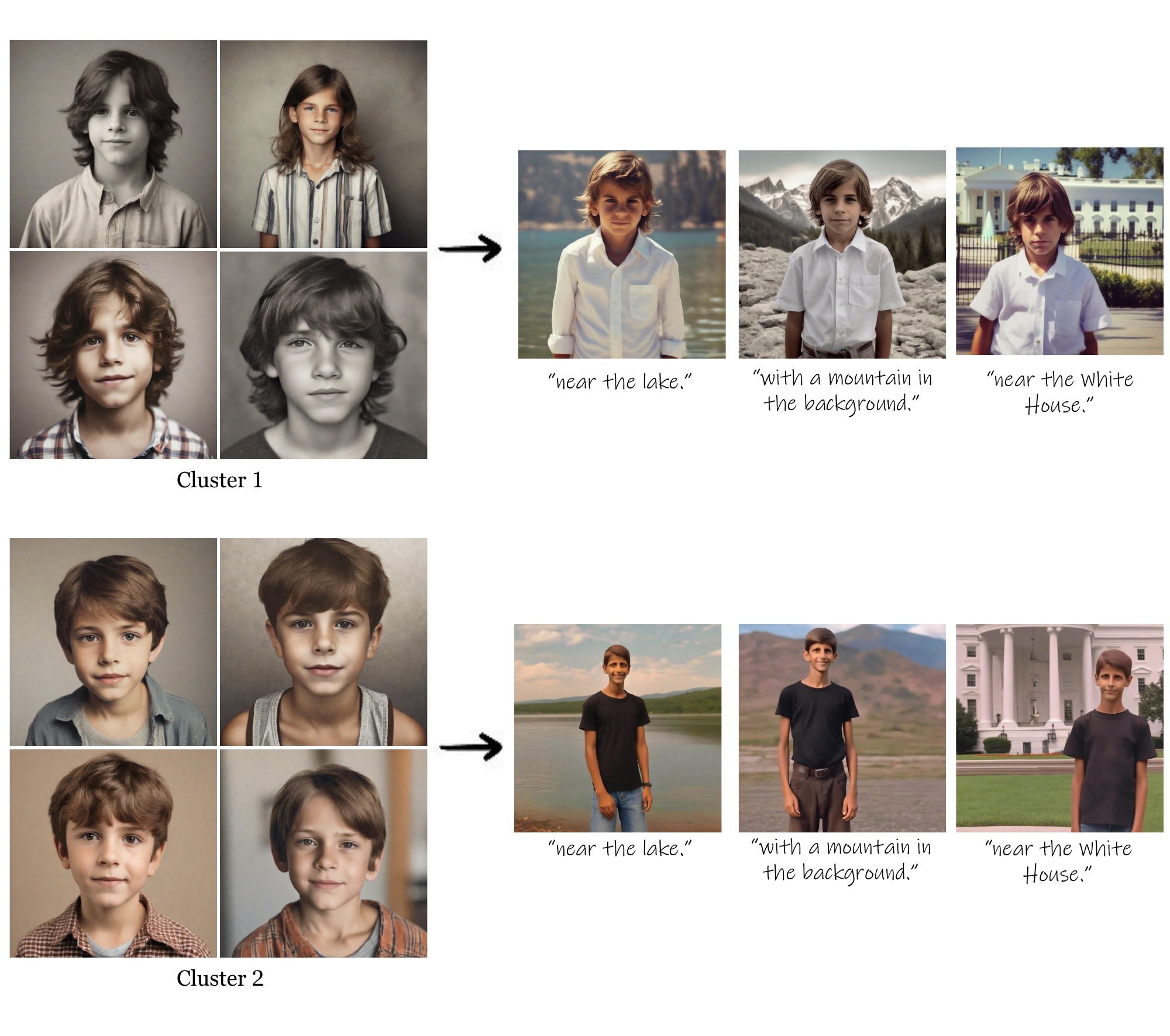}
    \caption{\textbf{User control.} Instead of choosing the most cohesive cluster automatically, as explained in \Cref{sec:identity_clustering}, a user can manually select one of the clusters according to their preferences, to affect the final result. For example, given the text prompt \textit{``a photo of a boy with brown hair''}, the user can control the hairstyle of the generated character by choosing the appropriate cluster.}
    \label{fig:user_control}
\end{figure*}

%% file: sections/appendices/additional_experiments.tex
\section{Additional Experiments}
\label{sec:additional_experiments}

Below, we provide additional experiments that were omitted from the main paper. In \Cref{sec:addtional_comparisons_and_results} we provide additional comparisons and results of our method, and demonstrate its non-deterministic nature in \Cref{sec:nondeterminism}. In \Cref{sec:naive_baselines} we compare our method against two \naive{} baselines. \Cref{sec:additional_feature_extractors} presents the results of our method using different feature extractors. Lastly, in \Cref{sec:dataset_non_memorization} we provide results that reduce the concerns of dataset memorization by our method.

\subsection{Additional Comparisons and Results}
\label{sec:addtional_comparisons_and_results}

\input{figures/automatic_qualitative_comparison/fig_baselines.tex}
\input{figures/qualitative_comparison/fig_additional.tex}
\input{figures/automatic_qualitative_comparison/fig_ablations.tex}
\input{figures/general_objects/fig.tex}
\input{figures/teaser/fig_extended.tex}
\input{figures/life_story/fig.tex}

In \Cref{fig:automatic_qualitative_baselines_comparison} we provide a qualitative comparison on the automatically generated prompts, and in \Cref{fig:additional_qualitative_comparison_additional} we provide an additional qualitative comparison.

Concurrently to our work, the \DALLE{} 3 model \cite{BetkerImprovingIG} was commercially released as part of the paid ChatGPT Plus \cite{chatgpt} subscription, enabling generating images in a conversational setting. We tried, using a conversation, to create a consistent character of a Plasticine cat, as demonstrated in \Cref{fig:dalle3_comparison}. As can be seen, the generated characters share only some of the characteristics (\eg, big eyes) but not all of them (\eg, colors, textures and shapes).

In \Cref{fig:automatic_qualitative_ablations_comparison} we provide a qualitative comparison of the ablated cases. In addition, as demonstrated in \Cref{fig:general_objects}, our approach is applicable to consistent generation of a wide range of subjects, without the requirement for them to necessarily depict human characters or creatures. \Cref{fig:extended_teaser} shows additional results of our method, demonstrating a variety of character styles. Lastly, in \Cref{fig:life_story} we demonstrate the ability of creating a fully consistent ``life story" of a character using our method.

\subsection{Non-determinism of Our Method}
\label{sec:nondeterminism}

\input{figures/nondeterminism/fig_person.tex}
\input{figures/nondeterminism/fig_cat.tex}

In \Cref{fig:nondeterminism_person,fig:nondeterminism_cat} we demonstrate the non-deterministic nature of our method. Using the same text prompt, we run our method multiple times with different initial seeds, thereby generating a different set of images for the identity clustering stage (\Cref{sec:identity_clustering}).
Consequently, the most cohesive cluster $\ccand$ is different in each run, yielding different consistent identities. This behavior of our method is aligned with the one-to-many nature of our task --- a single text prompt may correspond to many identities.

\subsection{\Naive{} Baselines}
\label{sec:naive_baselines}
\input{figures/automatic_qualitative_comparison/fig_naive_baselines.tex}
\input{figures/quantitative_comparison/naive_baselines_comparison.tex}

\input{figures/dalle3/fig_dalle3.tex}

As explained in \Cref{sec:comparisons},
we compared our method against a version of TI~\cite{Gal2022AnII} and LoRA DB~\cite{lora_diffusion} that were trained on a single image (with a single identity). Instead, we could generate a small set of five images for the given prompt (that are not guaranteed to be of the same identity), and use this small dataset for TI and LoRA DB baselines, referred to as \emph{TI multi} and \emph{LoRA DB multi}, respectively. As can be seen in \Cref{fig:qualitative_naive_baselines,fig:naive_baselines_comparison}, these baselines fail to achieve satisfactory identity consistency.

\subsection{Additional Feature Extractors}
\label{sec:additional_feature_extractors}

\input{figures/quantitative_comparison/feature_extractors_comparison.tex}
\input{figures/automatic_qualitative_comparison/fig_feature_extractors.tex}

Instead of using DINOv2 \cite{Oquab2023DINOv2LR} features for the identity clustering stage (\Cref{sec:identity_clustering}),
we also experimented with two alternative feature extractors: DINOv1~\cite{Caron2021EmergingPI} and CLIP~\cite{Radford2021LearningTV} image encoder. We quantitatively evaluate our method with each of these feature extractors in terms of identity consistency and prompt similarity, as explained in
\Cref{sec:comparisons}.
As can be seen in \Cref{fig:feature_extractors_comparison}, DINOv1 produces higher identity consistency, while sacrificing prompt similarity, whereas CLIP achieves higher prompt similarity at the expense of identity consistency. Qualitatively, as demonstrated in \Cref{fig:automatic_qualitative_feature_extractor_comparison}, we found the DINOv1 extractor to perform similarly to DINOv2, whereas CLIP produces results with a slightly lower identity consistency.

\subsection{Additional Clustering Visualization}
\label{sec:additional_clustering_visualization}

In \Cref{fig:clustering_visualization} we provide a visualization of the clustering algorithm described in
\Cref{sec:identity_clustering}.
As can be seen, given the input text prompt \textit{``a purple astronaut, digital art, smooth, sharp focus, vector art''}, in the first iteration (top three rows), our algorithm divides the generated image set into three clusters: (1) focusing on the astronaut's head, (2) an astronaut with no face, and (3) a full body astronaut. In the second iteration (bottom three rows), all the clusters share the same identity, that was extracted in the first iteration, as described in
\Cref{sec:identity_extraction},
and our algorithm divides them into clusters by their pose.

\input{figures/clustering_visualization/fig.tex}

\subsection{Dataset Non-Memorization}
\label{sec:dataset_non_memorization}

\input{figures/dataset_non_memorization/fig.tex}

Our method is able to produce consistent characters, which raises the question of whether these characters already exist in the training data of the generative model. 
We employed SDXL~\cite{Podell2023SDXLIL} as our text-to-image model, whose training dataset is, unfortunately, undisclosed in the paper~\cite{Podell2023SDXLIL}. Consequently, we relied on the most likely overlapping dataset, LAION-5B~\cite{Schuhmann2022LAION5BAO}, which was also utilized by Stable Diffusion V2.

To probe for dataset memorization, we found the top 5 nearest neighbors in the dataset in terms of CLIP~\cite{Radford2021LearningTV} image similarity, for a few representative characters from our paper, using an open-source solution \cite{clip_retrival}. As demonstrated in \Cref{fig:dataset_non_memorization}, our method does not simply memorize images from the LAION-5B dataset.

\subsection{Stable Diffusion 2 Results}
\label{sec:sd2_results}

\input{figures/sd2_results/fig.tex}

We experimented with a version of our method that uses the Stable Diffusion 2~\cite{Rombach2021HighResolutionIS} model. The implementation is the same as explained in \Cref{sec:method_implementation_details}, with the following changes: 
(1) The set of custom text embeddings $\tau$ in the character representation $\crep$ (as explained in
Section 2 in the main paper
), contains only one text embedding. (2) We used a higher learning rate of 5e-4. The rest of the implementation details are the same. More specifically, we used Stable Diffusion v2.1 implementation from Diffusers \cite{von-platen-etal-2022-diffusers} library.

As can be seen in \Cref{fig:sd2_results}, when using the Stable Diffusion 2 backbone, our method can extract a consistent character, however, as expected, the results are of a lower quality than when using the SDXL ~\cite{Podell2023SDXLIL} backbone that we use in the rest of this paper.

%% file: figures/automatic_qualitative_comparison/fig_baselines.tex
\begin{figure*}[t]
    \centering
    \setlength{\tabcolsep}{3.5pt}
    \renewcommand{\arraystretch}{0.2}
    \setlength{\ww}{0.3\columnwidth}
    \begin{tabular}{ccccccc}
        &
        \textbf{TI} &
        \textbf{LoRA DB} &
        \textbf{ELITE} &
        \textbf{BLIP-diff} &
        \textbf{IP-Adapter} &
        \textbf{Ours}
        \\

        &
        \cite{Gal2022AnII} &
        \cite{lora_diffusion} &
        \cite{Wei2023ELITEEV} &
        \cite{Li2023BLIPDiffusionPS} &
        \cite{Ye2023IPAdapterTC} &
        \\

        \rotatebox[origin=c]{90}{\textit{``drinking}}
        \rotatebox[origin=c]{90}{\textit{a beer''}} &
        {\includegraphics[valign=c, width=\ww]{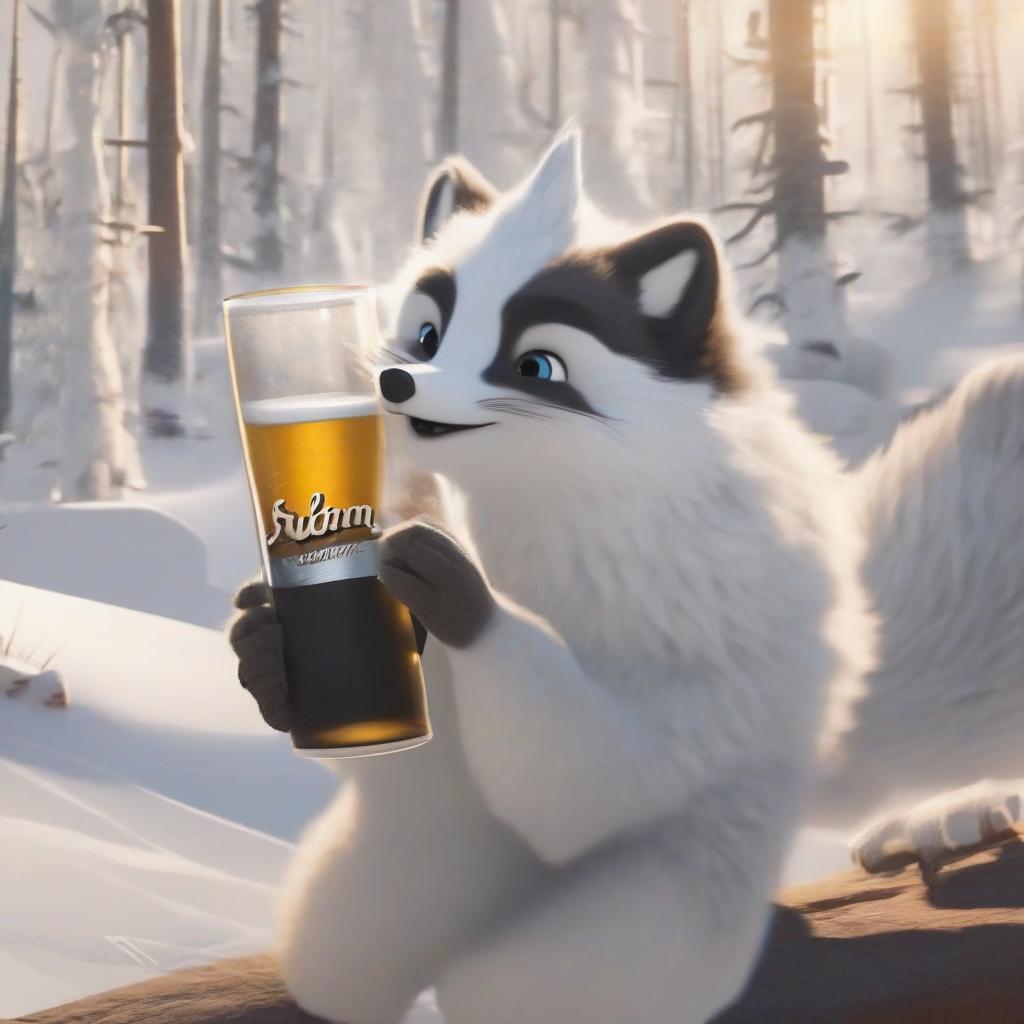}} &
        {\includegraphics[valign=c, width=\ww]{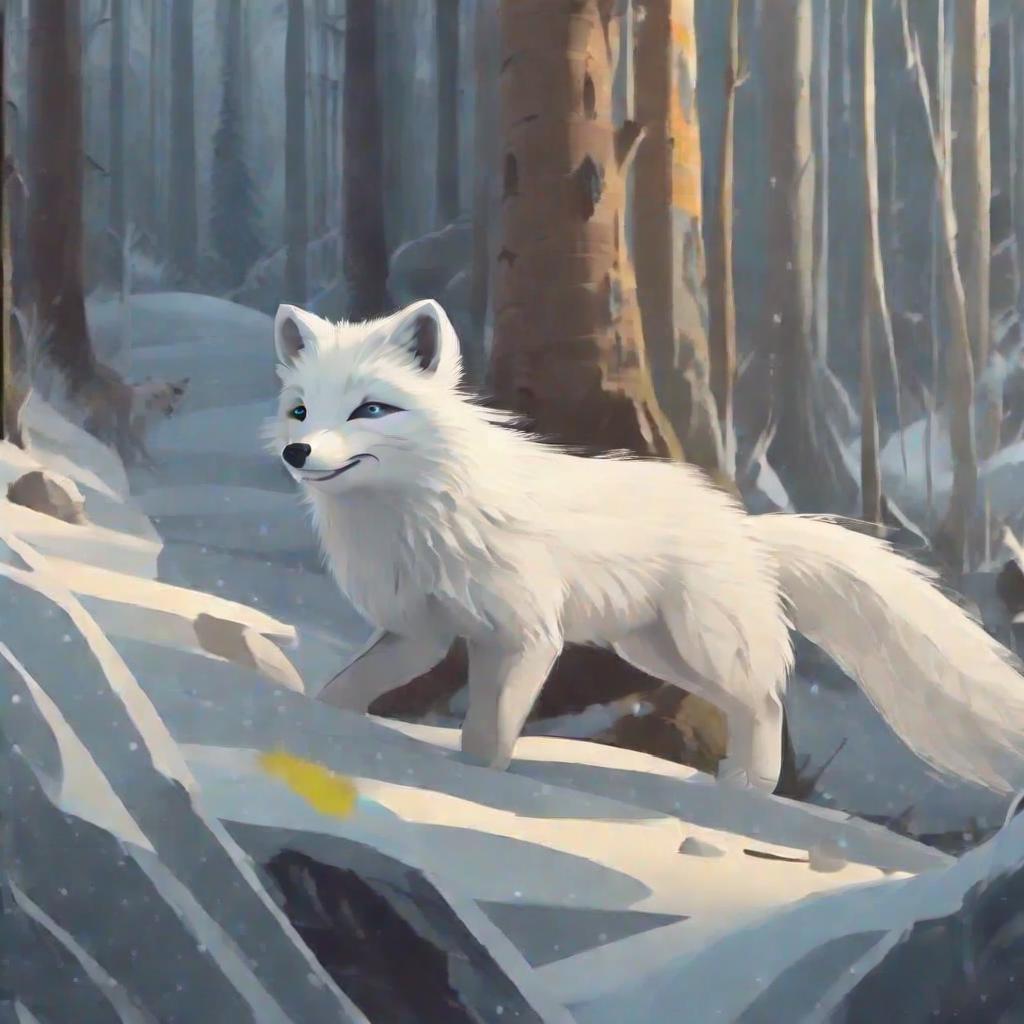}} &
        {\includegraphics[valign=c, width=\ww]{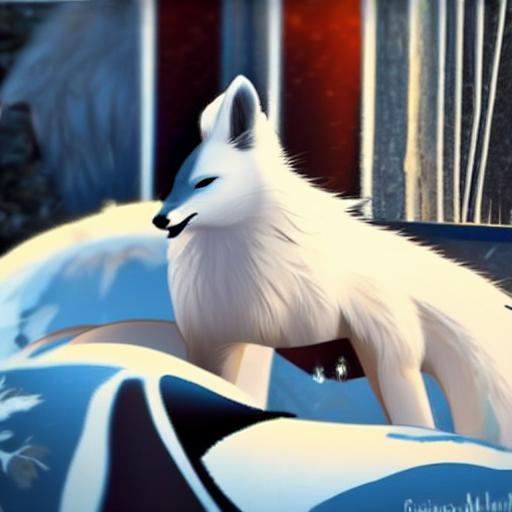}} &
        {\includegraphics[valign=c, width=\ww]{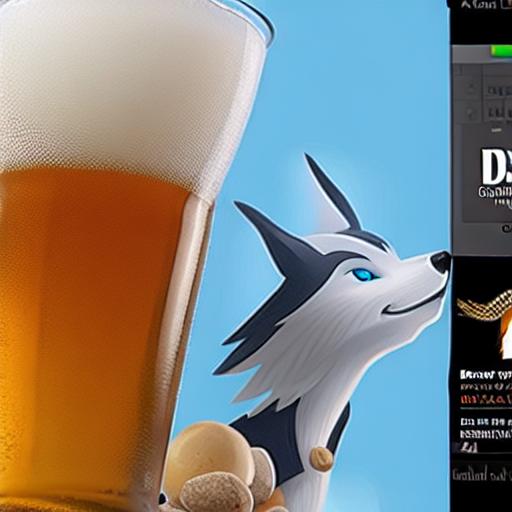}} &
        {\includegraphics[valign=c, width=\ww]{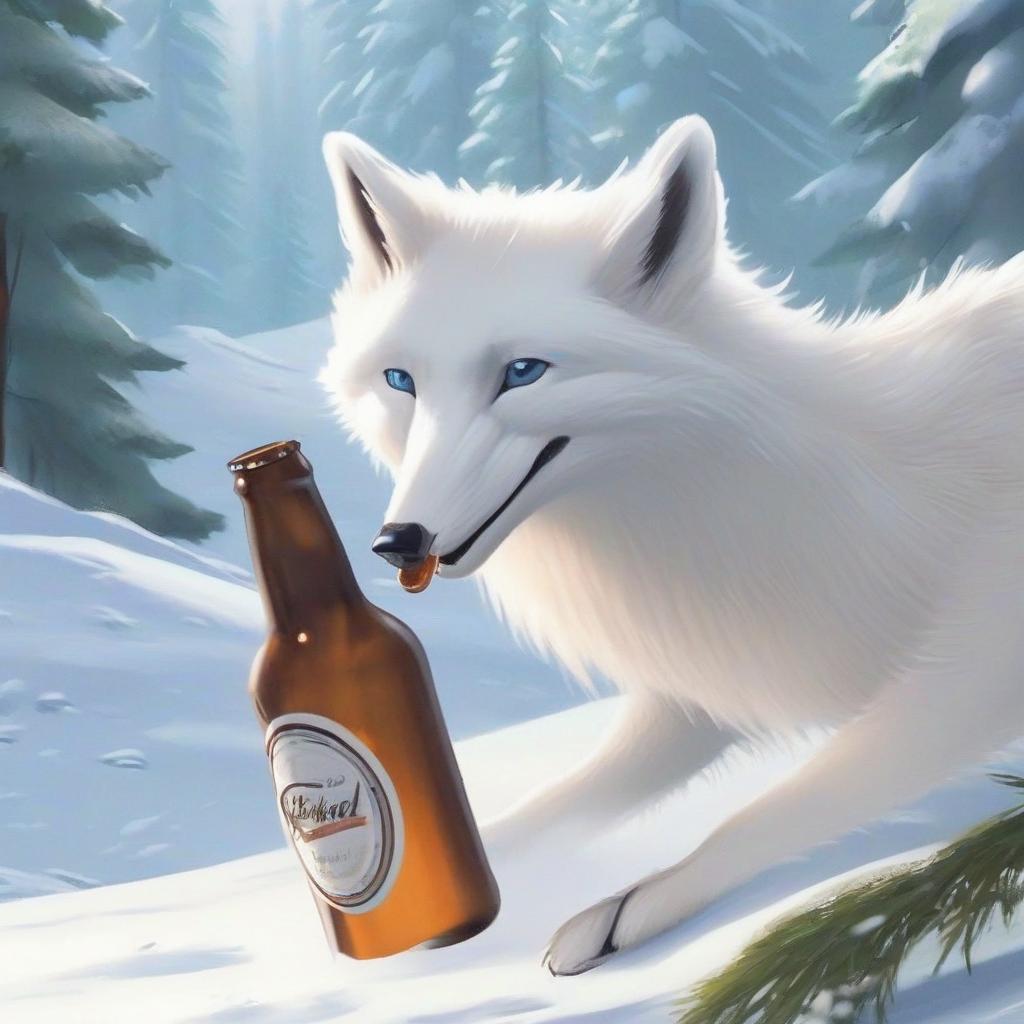}} &
        {\includegraphics[valign=c, width=\ww]{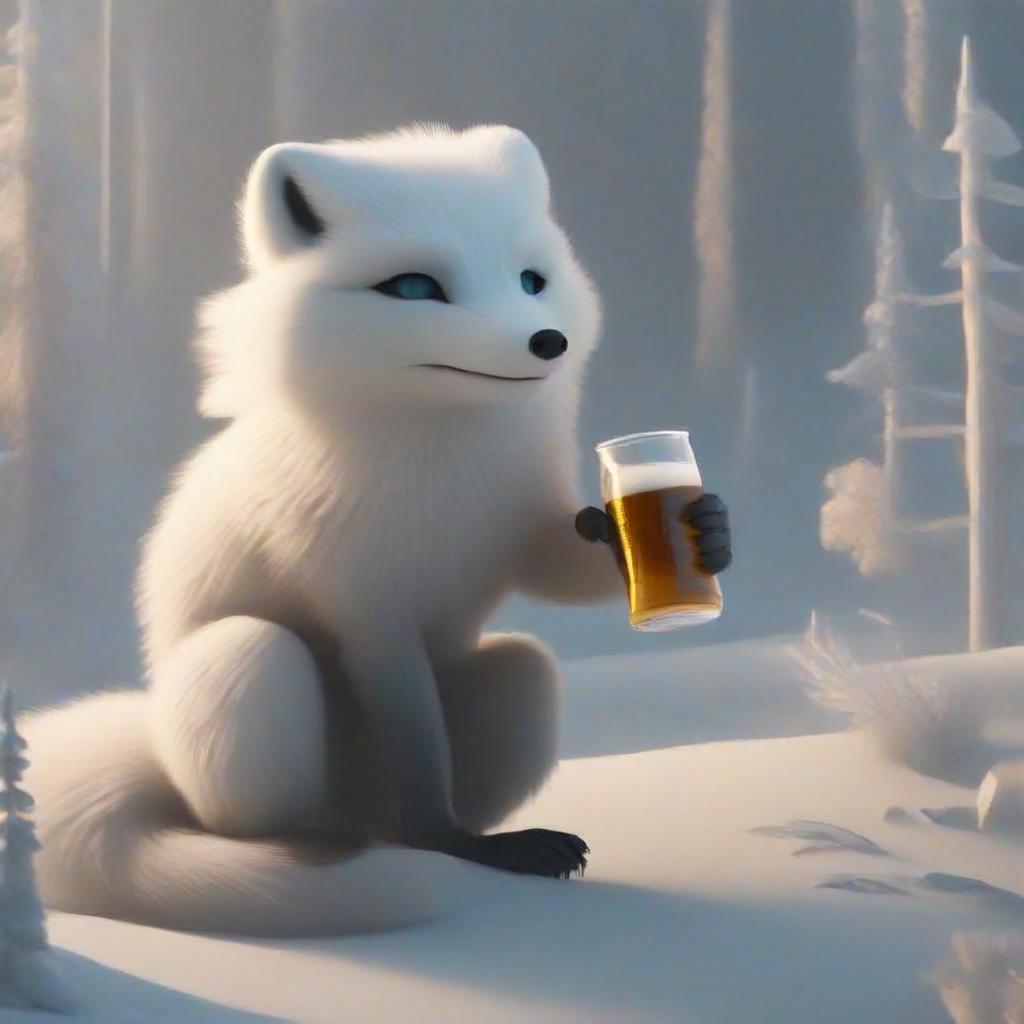}}
        \\
        \\

        \rotatebox[origin=c]{90}{\textit{``with a city in}}
        \rotatebox[origin=c]{90}{\textit{the background''}} &
        {\includegraphics[valign=c, width=\ww]{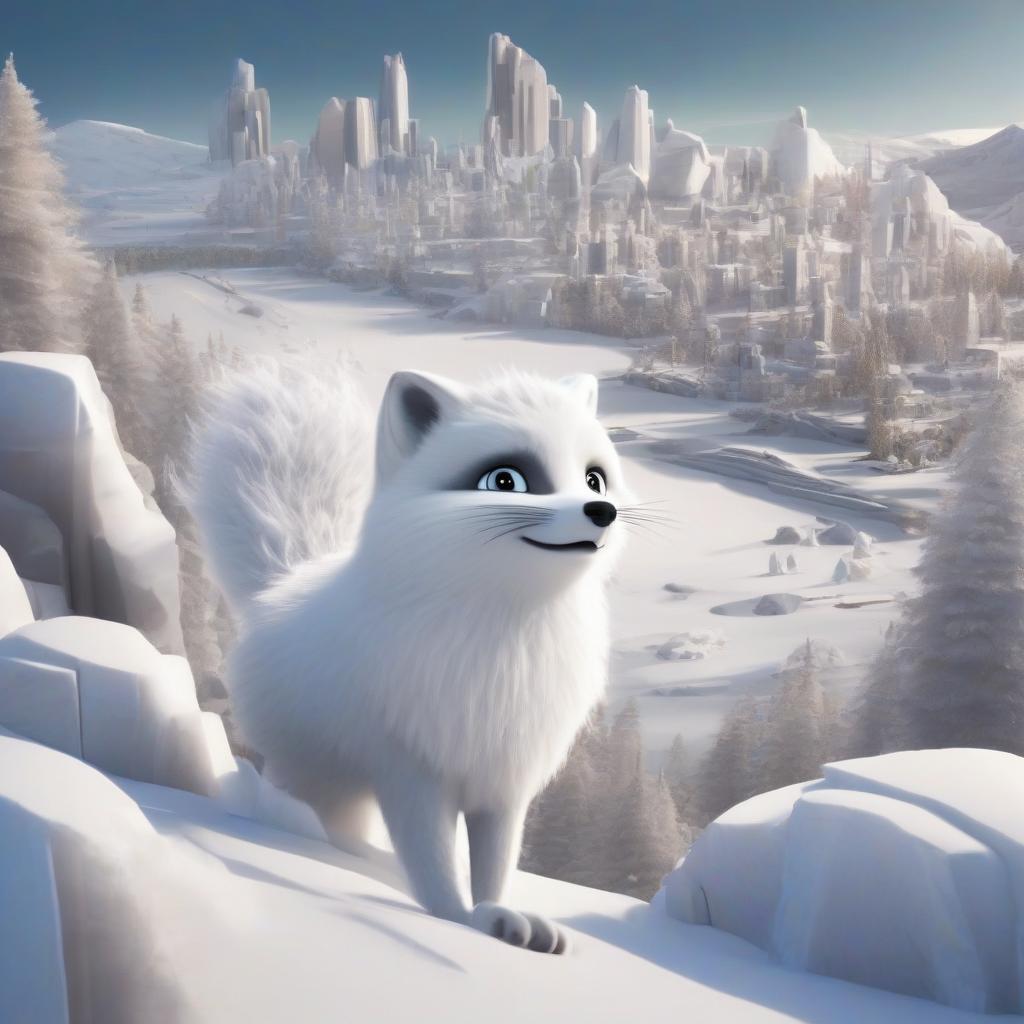}} &
        {\includegraphics[valign=c, width=\ww]{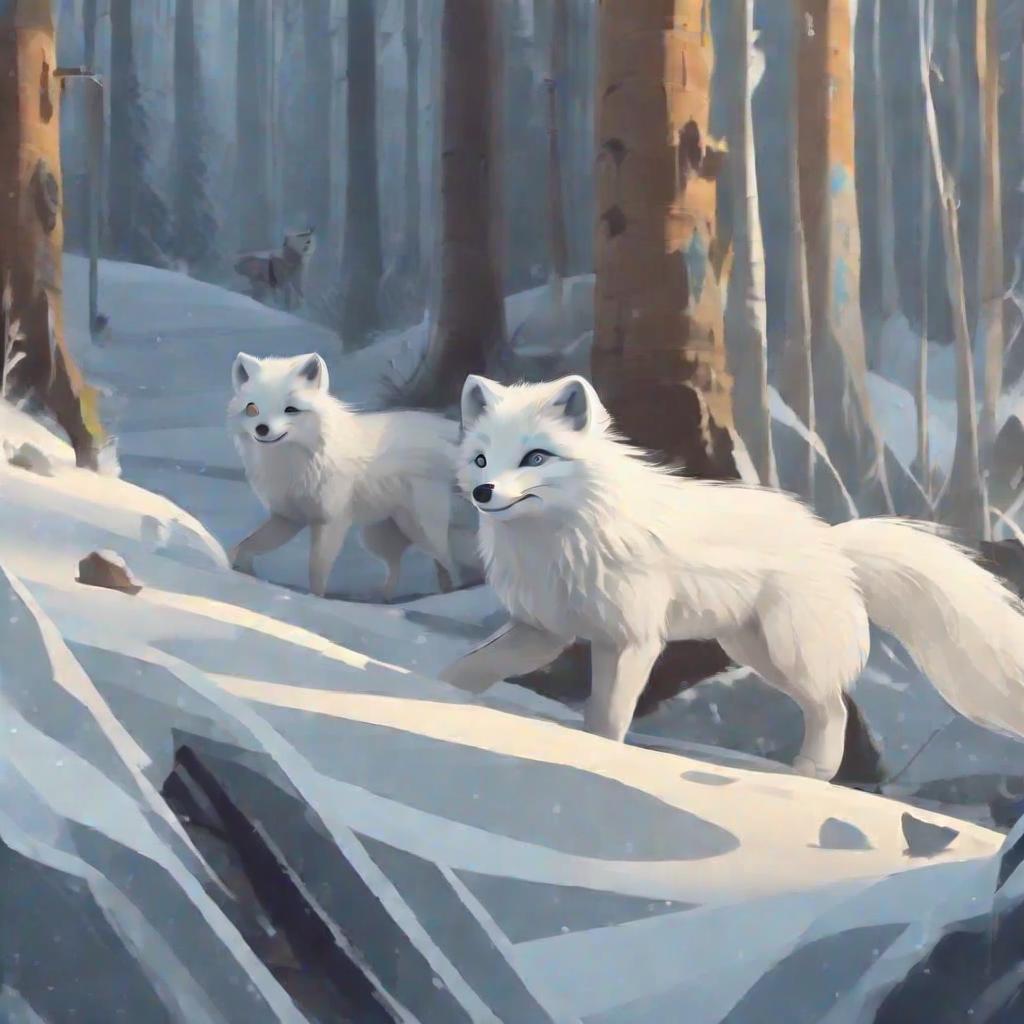}} &
        {\includegraphics[valign=c, width=\ww]{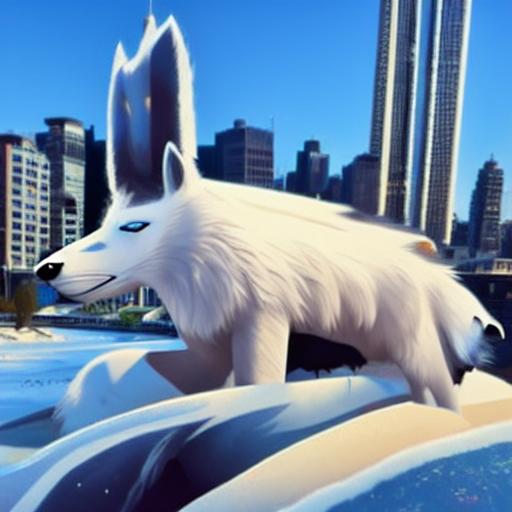}} &
        {\includegraphics[valign=c, width=\ww]{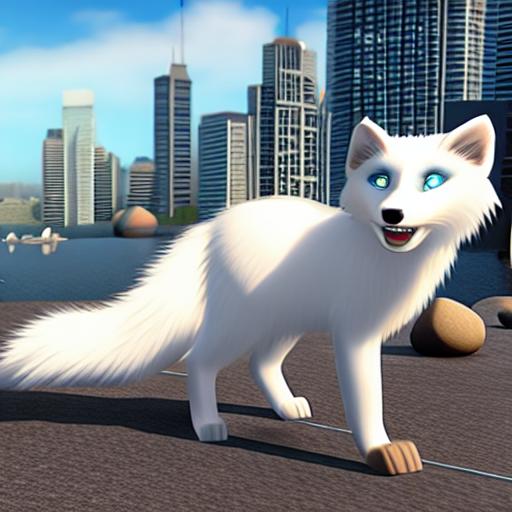}} &
        {\includegraphics[valign=c, width=\ww]{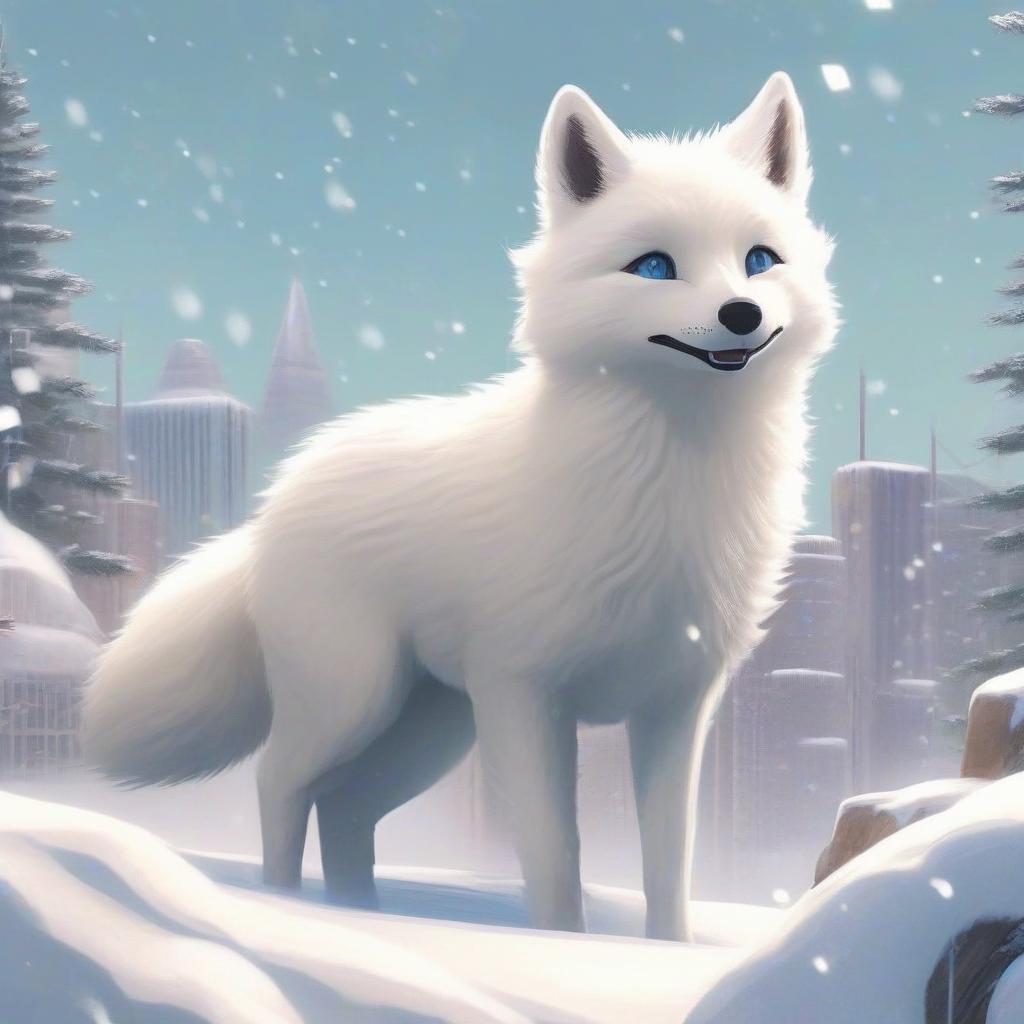}} &
        {\includegraphics[valign=c, width=\ww]{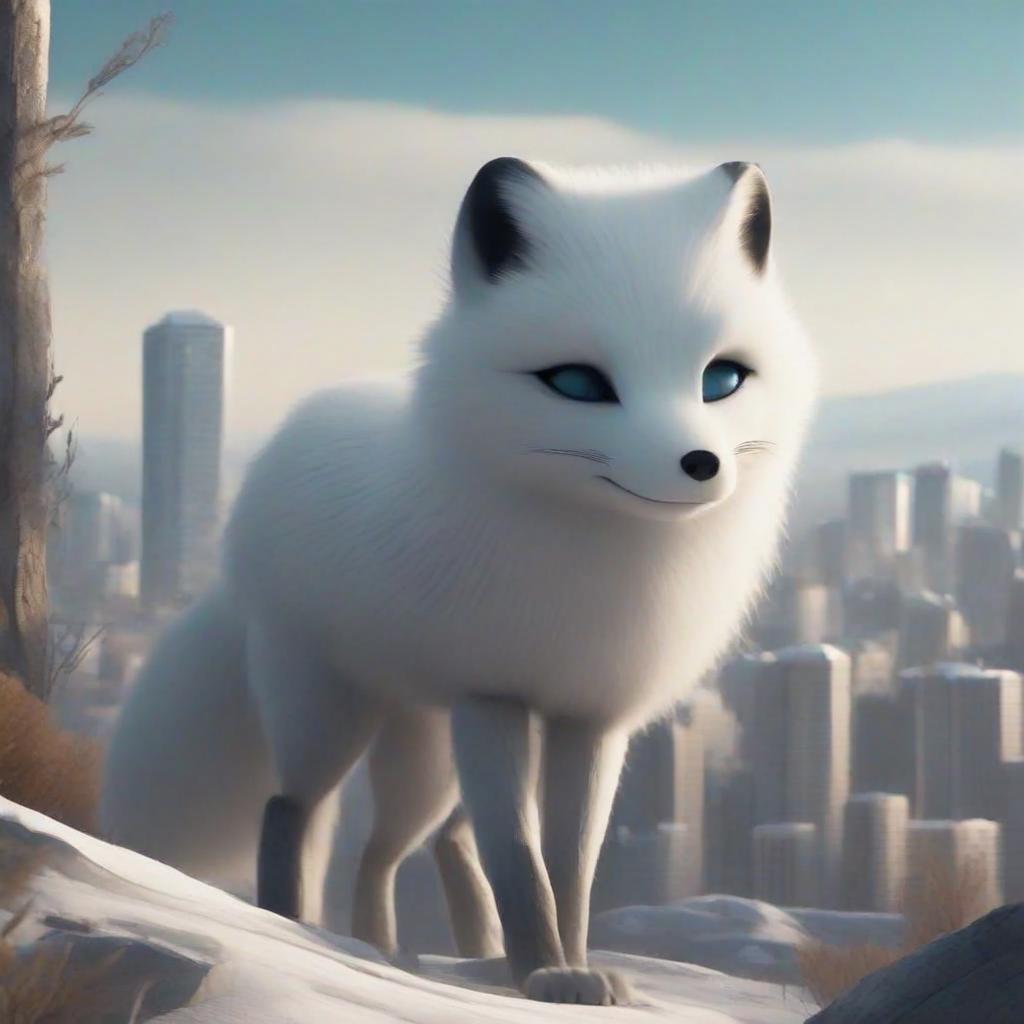}}
        \\
        \\

        \\
        \\
        \multicolumn{7}{c}{\textit{``a 2D animation of captivating Arctic fox with fluffy fur, bright eyes, and}}
        \\
        \multicolumn{7}{c}{\textit{nimble movements, bringing the magic of the icy wilderness to animated life''}}
        \\
        \\
        \\
        \midrule

        \\
        \rotatebox[origin=c]{90}{\textit{``eating}}
        \rotatebox[origin=c]{90}{\textit{a burger''}} &
        {\includegraphics[valign=c, width=\ww]{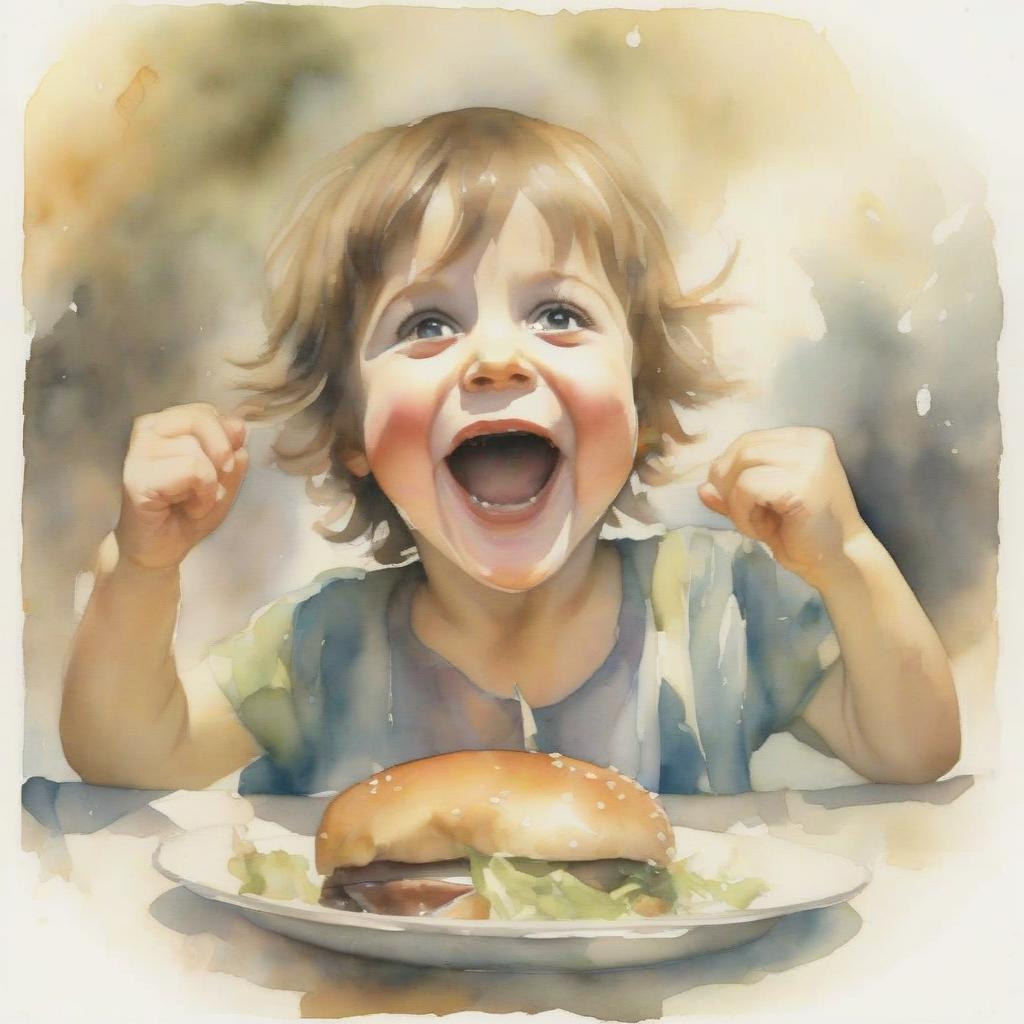}} &
        {\includegraphics[valign=c, width=\ww]{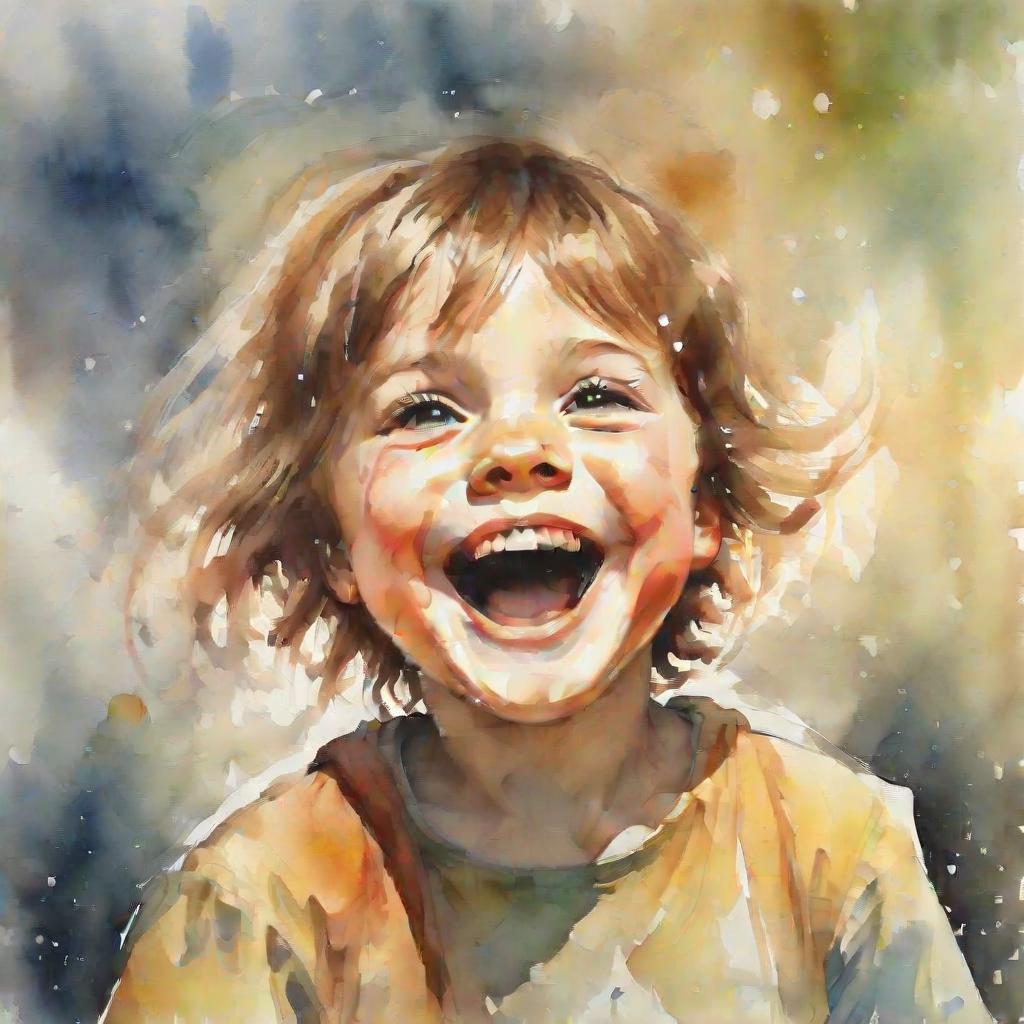}} &
        {\includegraphics[valign=c, width=\ww]{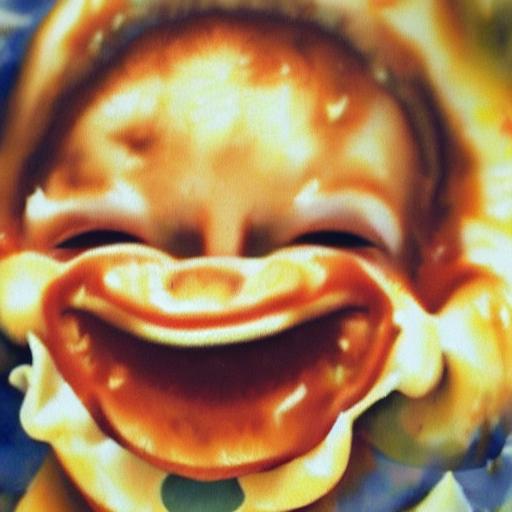}} &
        {\includegraphics[valign=c, width=\ww]{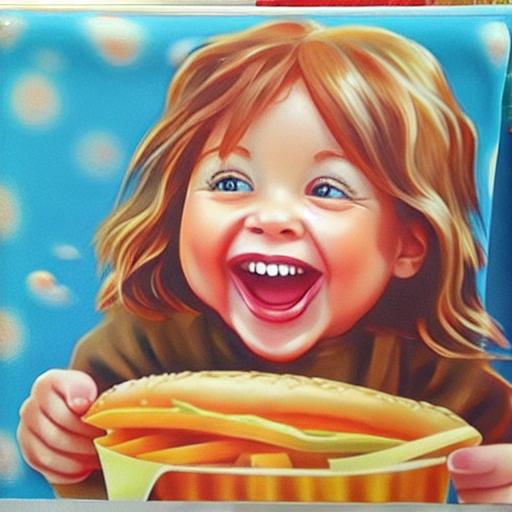}} &
        {\includegraphics[valign=c, width=\ww]{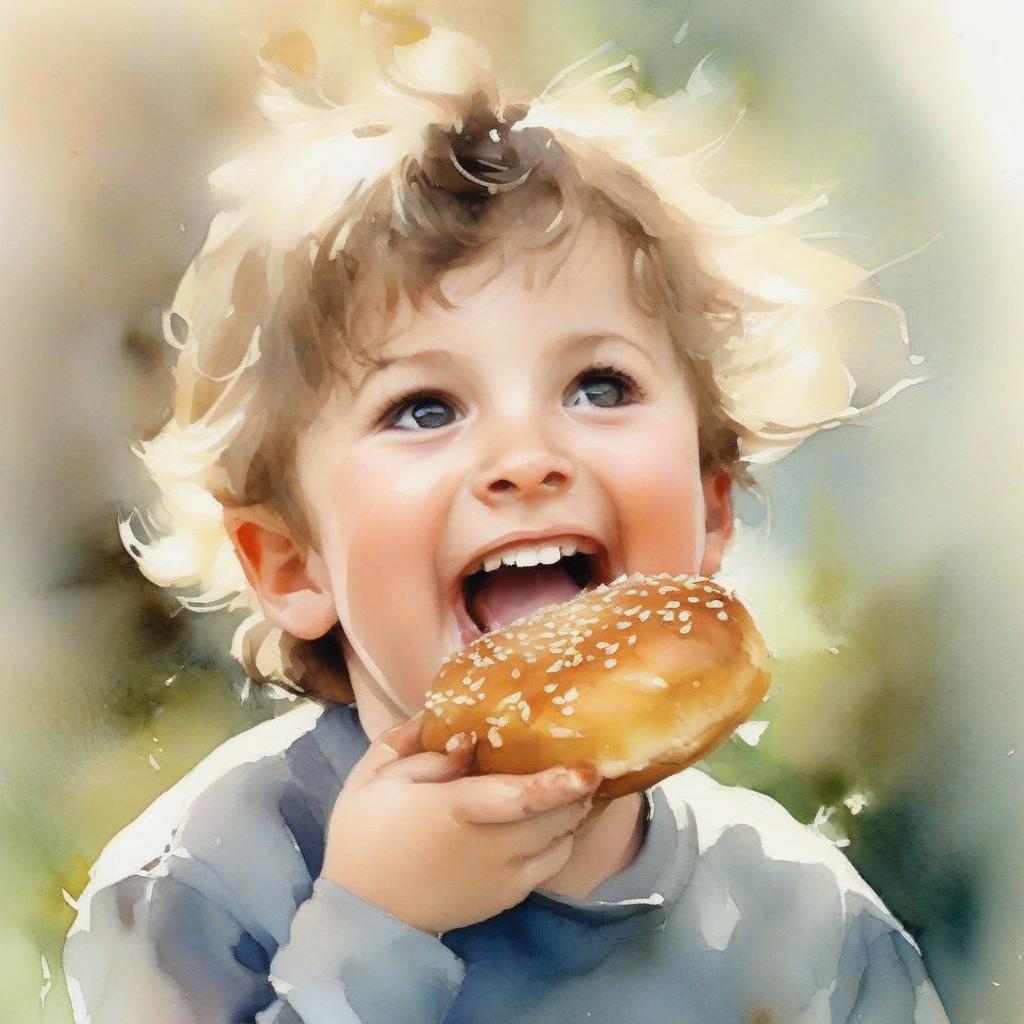}} &
        {\includegraphics[valign=c, width=\ww]{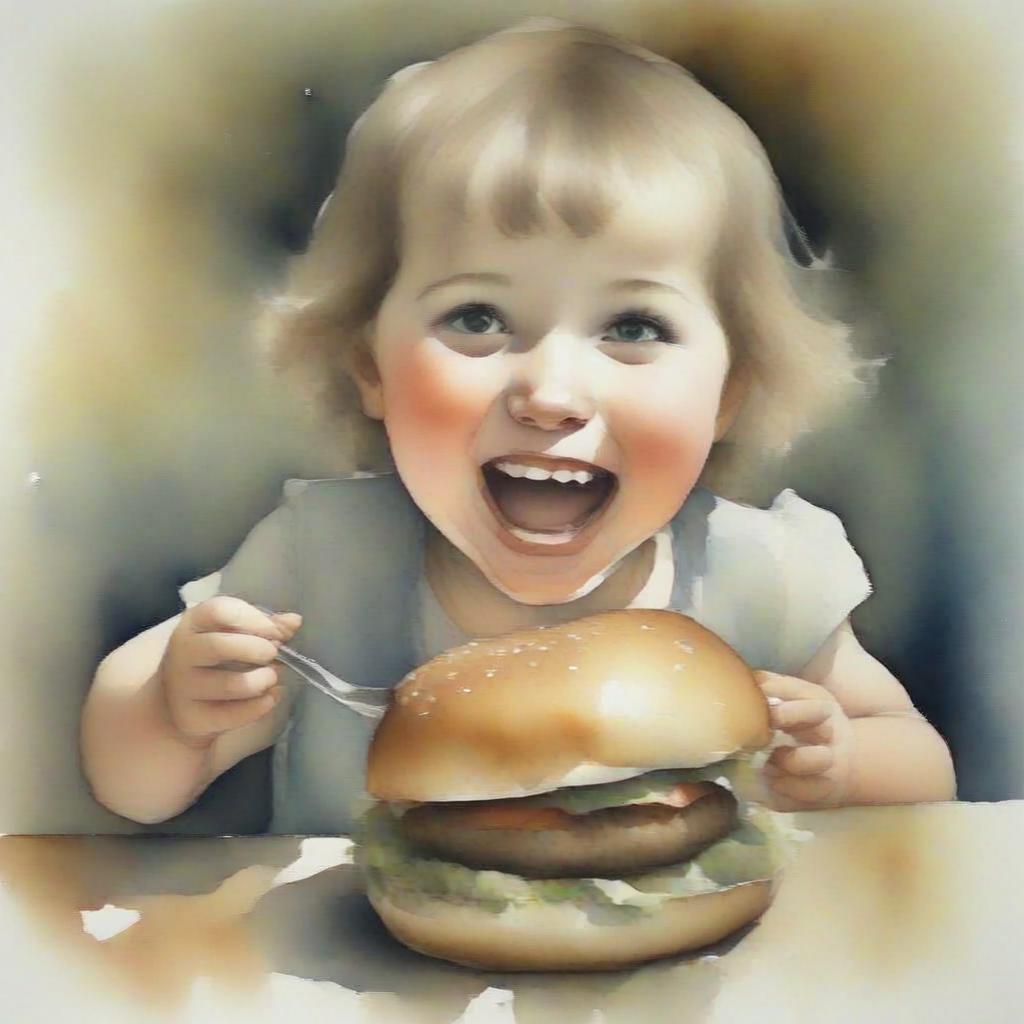}}
        \\
        \\

        \rotatebox[origin=c]{90}{\textit{``wearing a}}
        \rotatebox[origin=c]{90}{\textit{blue hat''}} &
        {\includegraphics[valign=c, width=\ww]{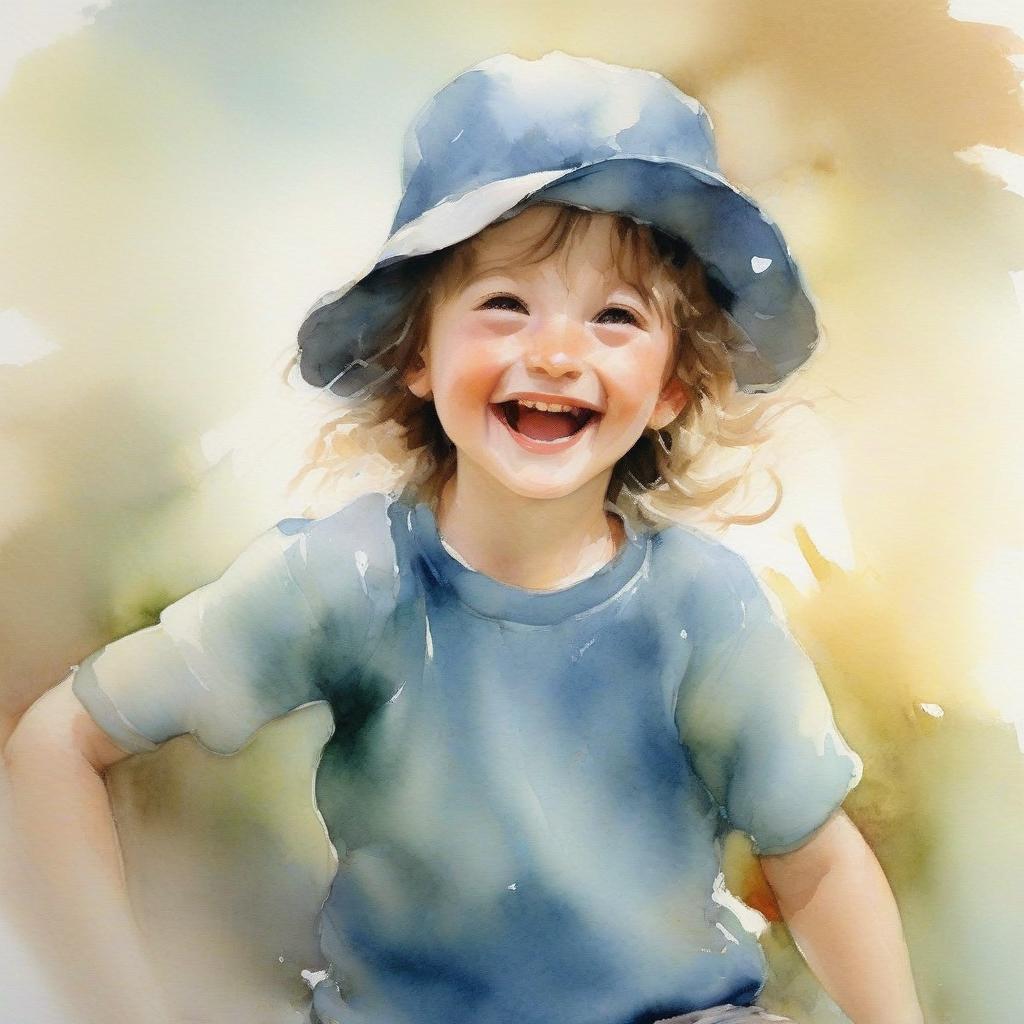}} &
        {\includegraphics[valign=c, width=\ww]{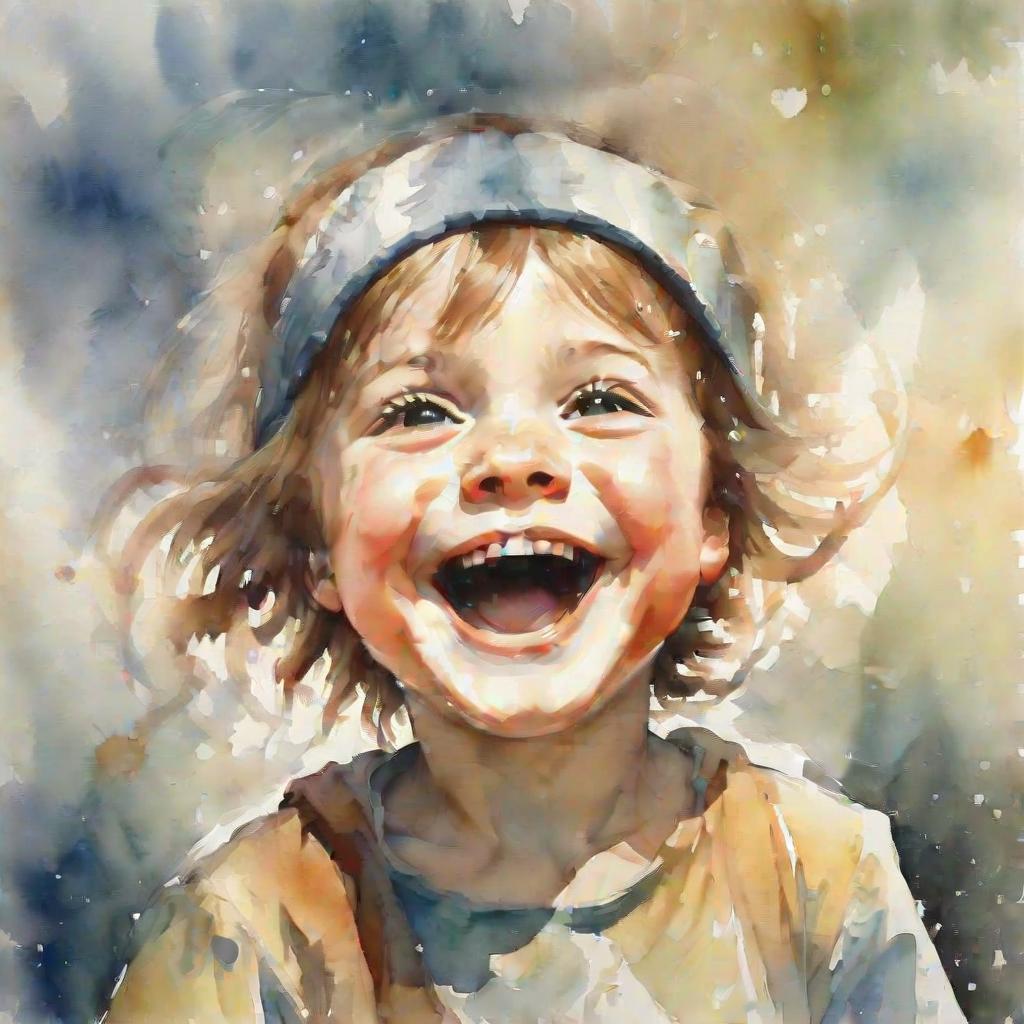}} &
        {\includegraphics[valign=c, width=\ww]{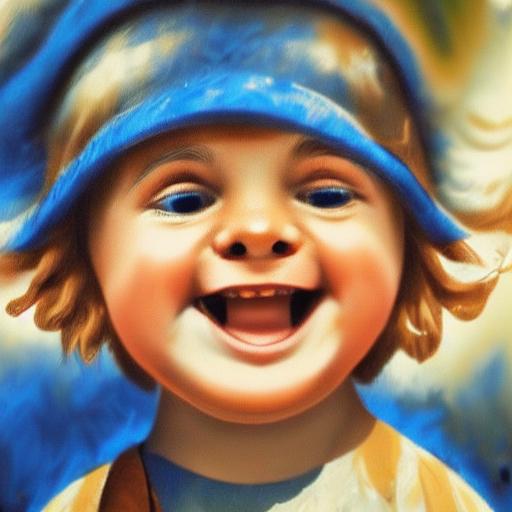}} &
        {\includegraphics[valign=c, width=\ww]{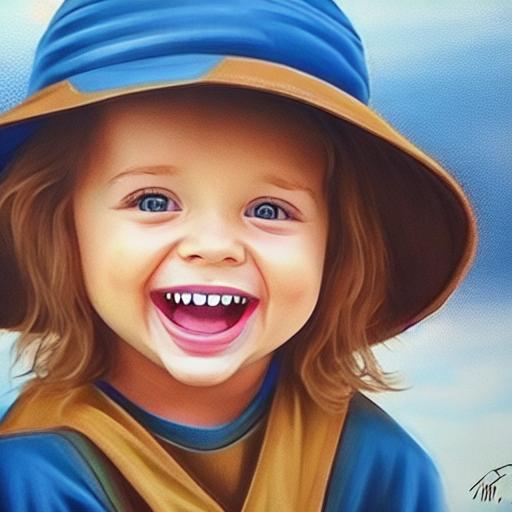}} &
        {\includegraphics[valign=c, width=\ww]{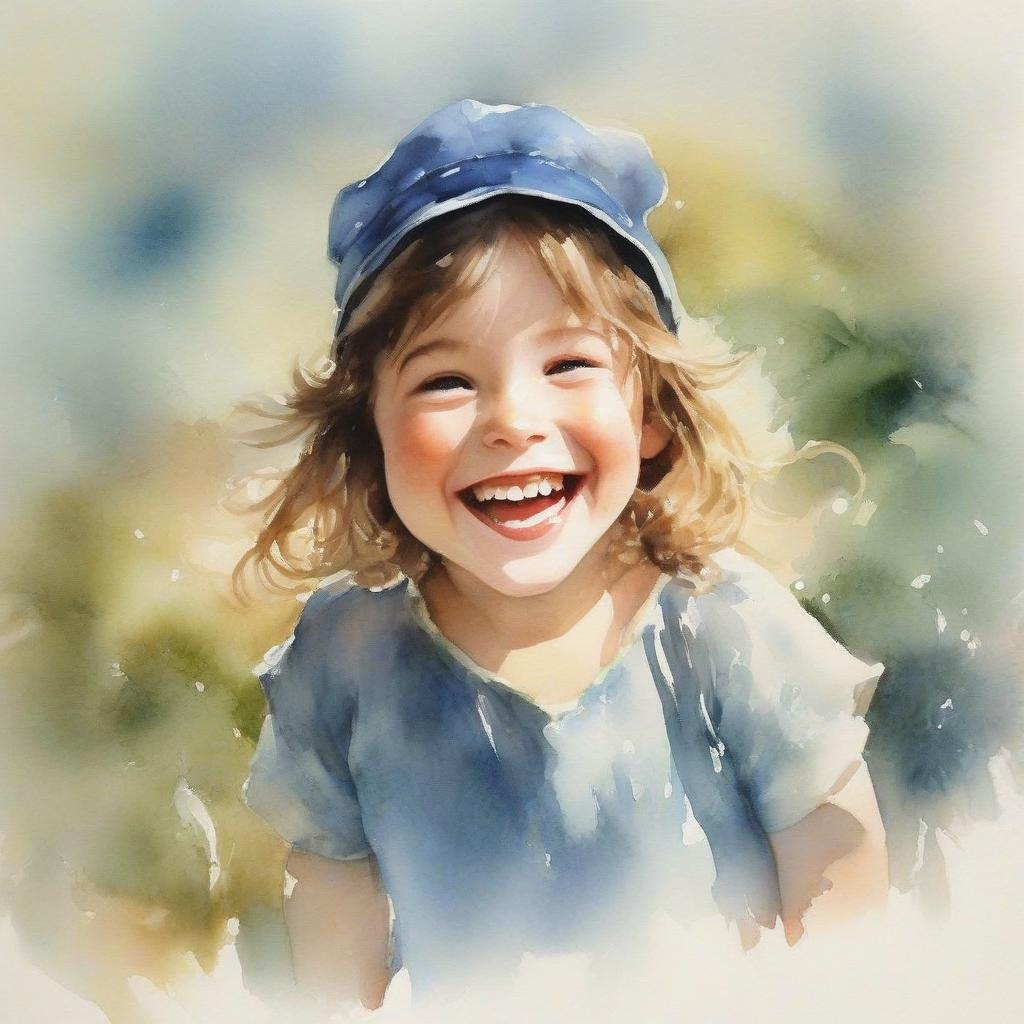}} &
        {\includegraphics[valign=c, width=\ww]{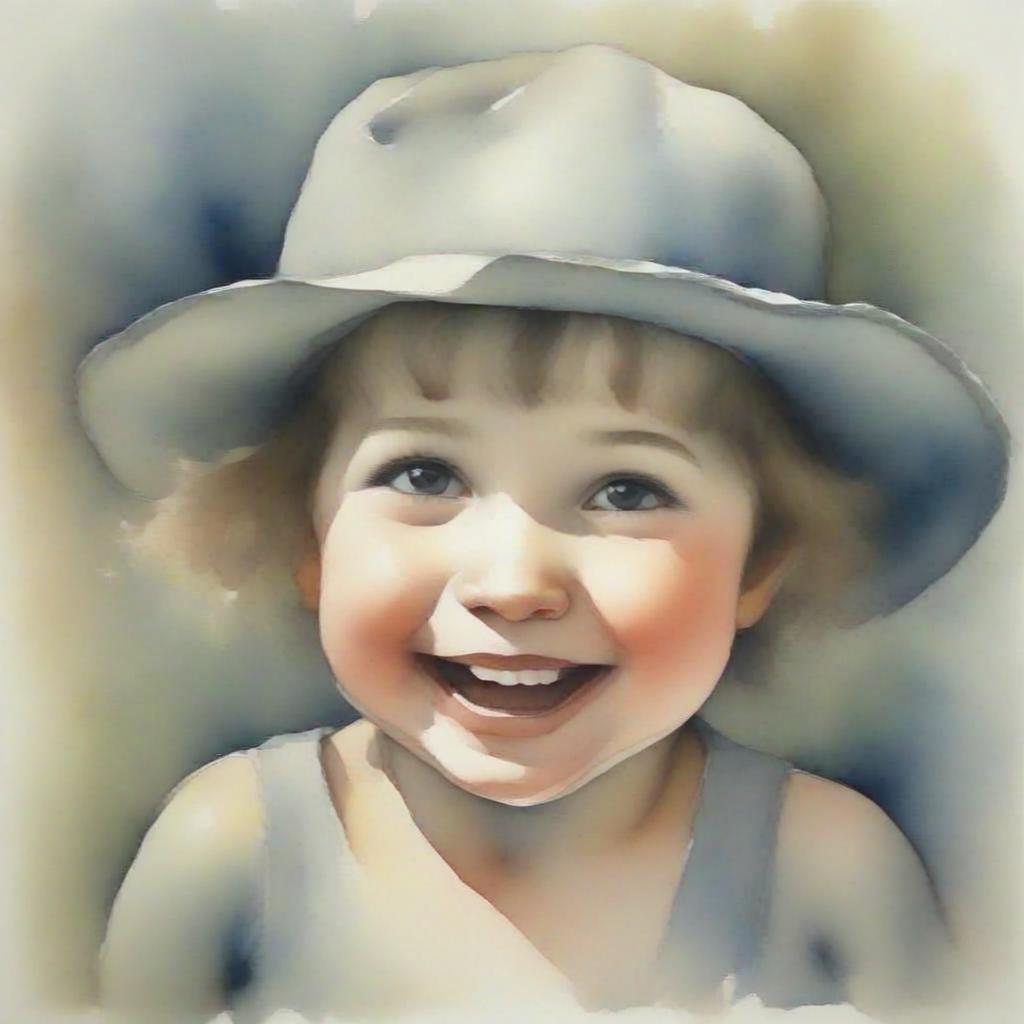}}
        \\
        \\

        \\
        \\
        \multicolumn{7}{c}{\textit{``a watercolor portrayal of a joyful child, radiating innocence and wonder with}}
        \\
        \multicolumn{7}{c}{\textit{rosy cheeks and a genuine, wide-eyed smile''}}
        \\
        \\
        \midrule

        \\
        \rotatebox[origin=c]{90}{\textit{``near the Statue}}
        \rotatebox[origin=c]{90}{\textit{of Liberty''}} &
        {\includegraphics[valign=c, width=\ww]{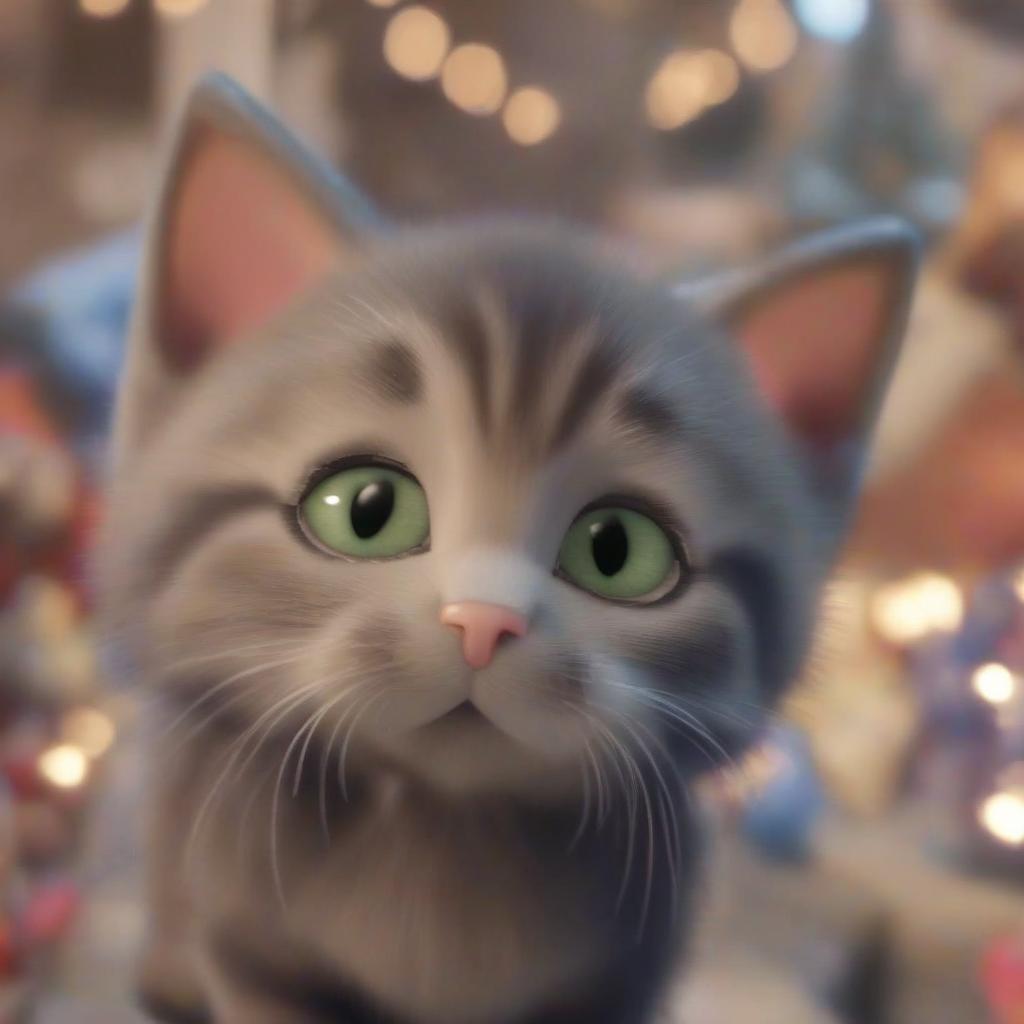}} &
        {\includegraphics[valign=c, width=\ww]{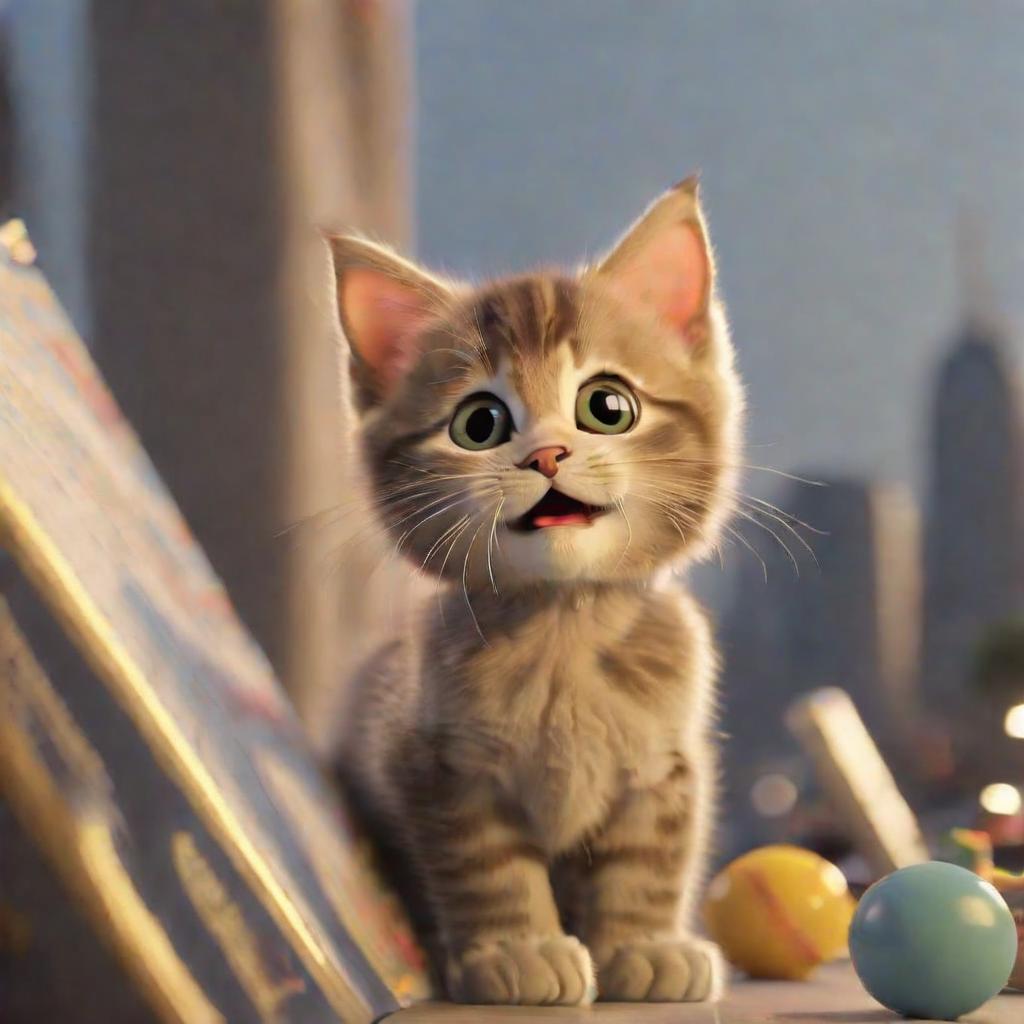}} &
        {\includegraphics[valign=c, width=\ww]{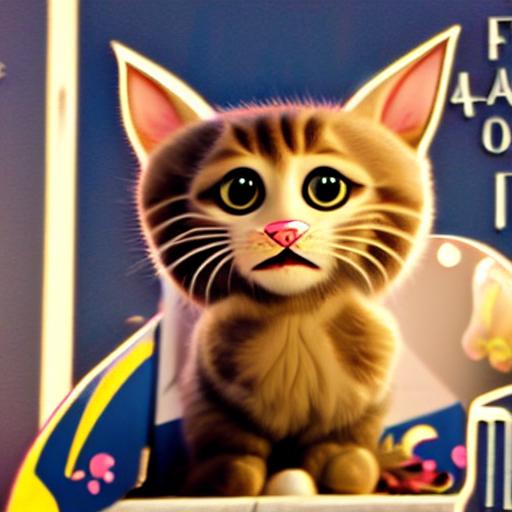}} &
        {\includegraphics[valign=c, width=\ww]{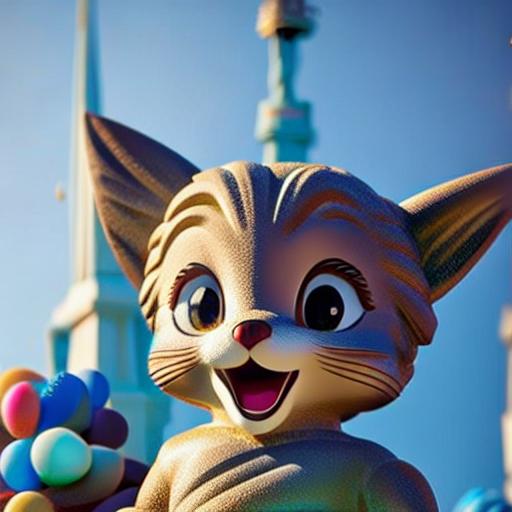}} &
        {\includegraphics[valign=c, width=\ww]{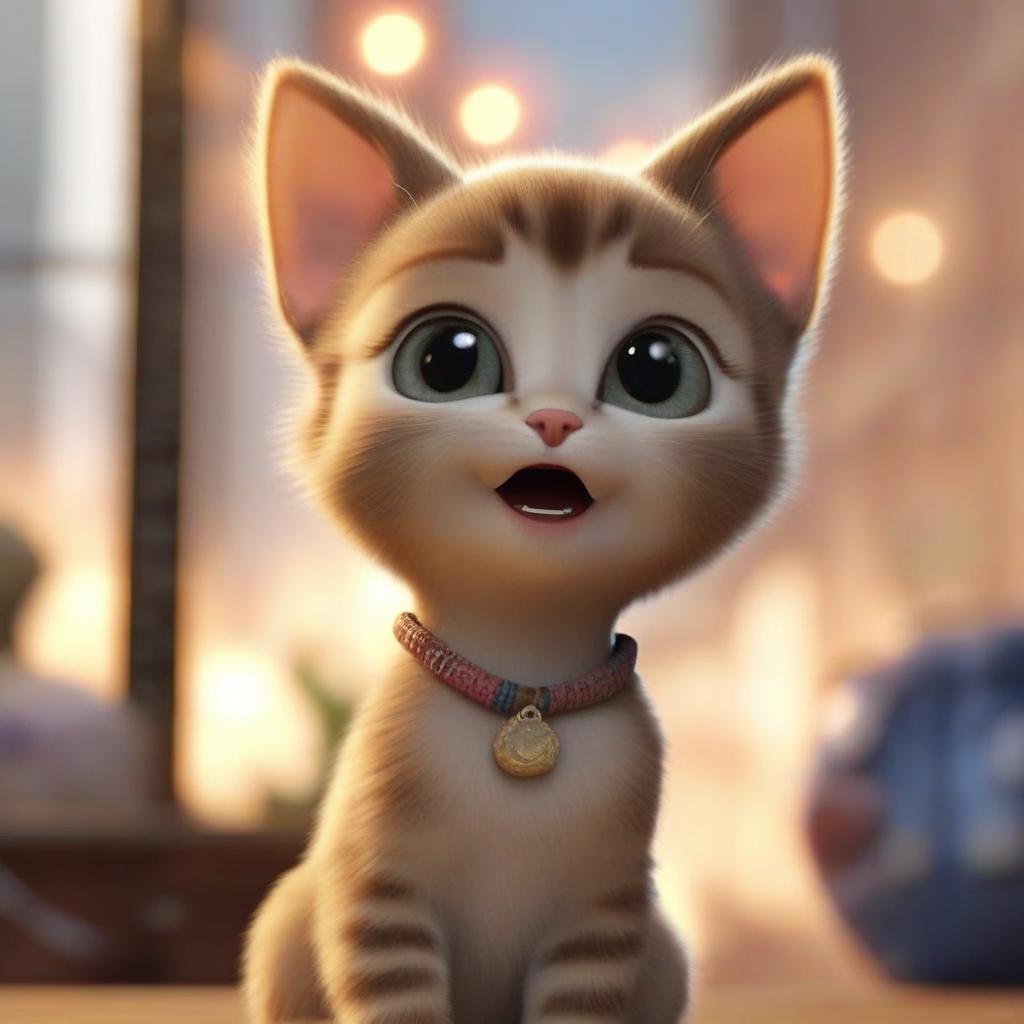}} &
        {\includegraphics[valign=c, width=\ww]{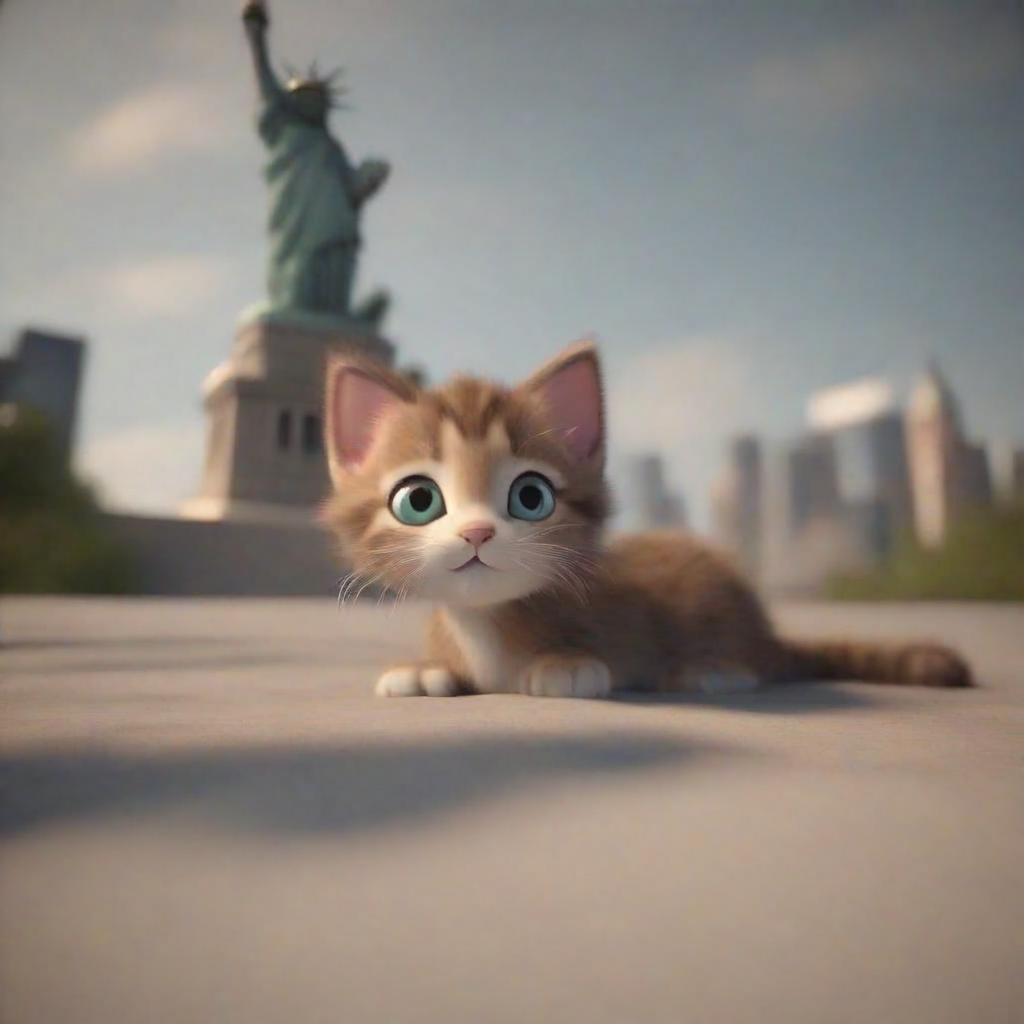}}
        \\
        \\

        \rotatebox[origin=c]{90}{\textit{``as a}}
        \rotatebox[origin=c]{90}{\textit{police officer''}} &
        {\includegraphics[valign=c, width=\ww]{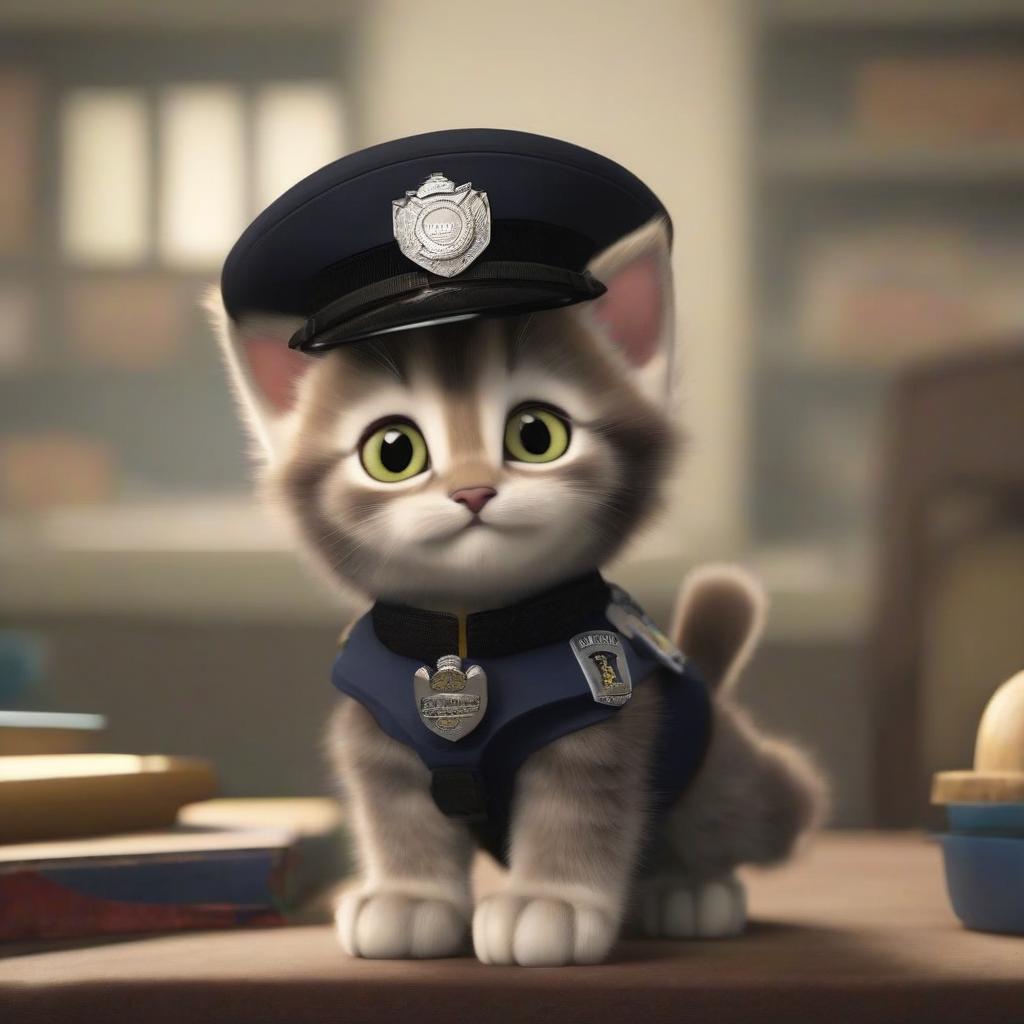}} &
        {\includegraphics[valign=c, width=\ww]{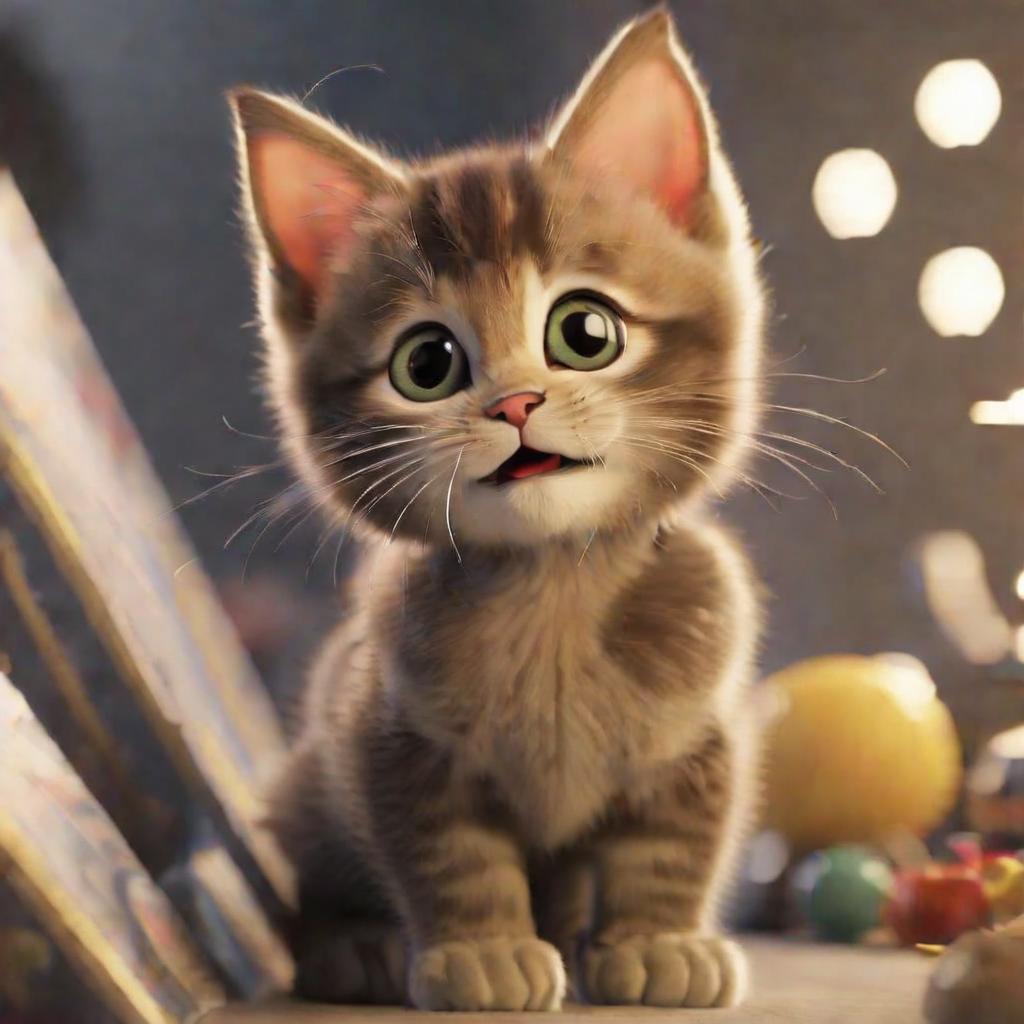}} &
        {\includegraphics[valign=c, width=\ww]{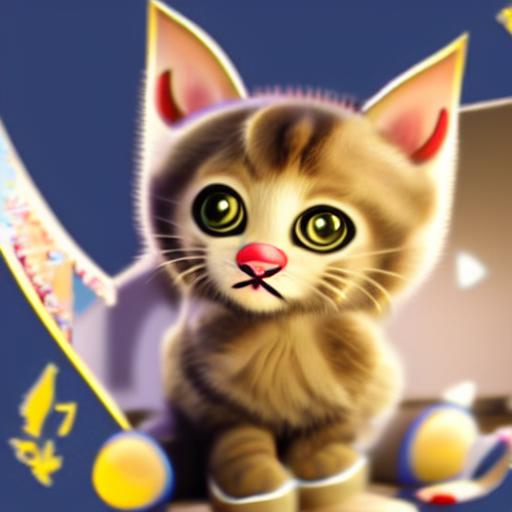}} &
        {\includegraphics[valign=c, width=\ww]{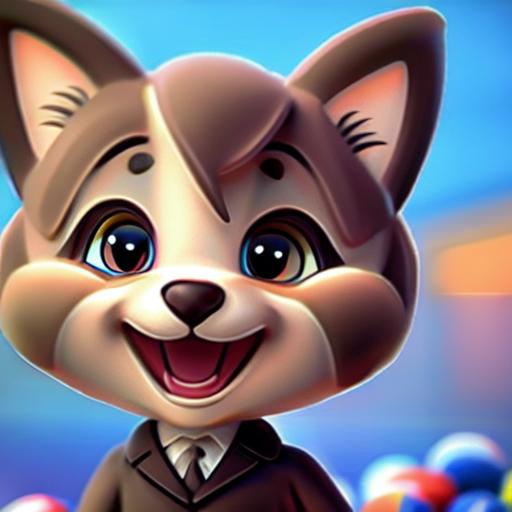}} &
        {\includegraphics[valign=c, width=\ww]{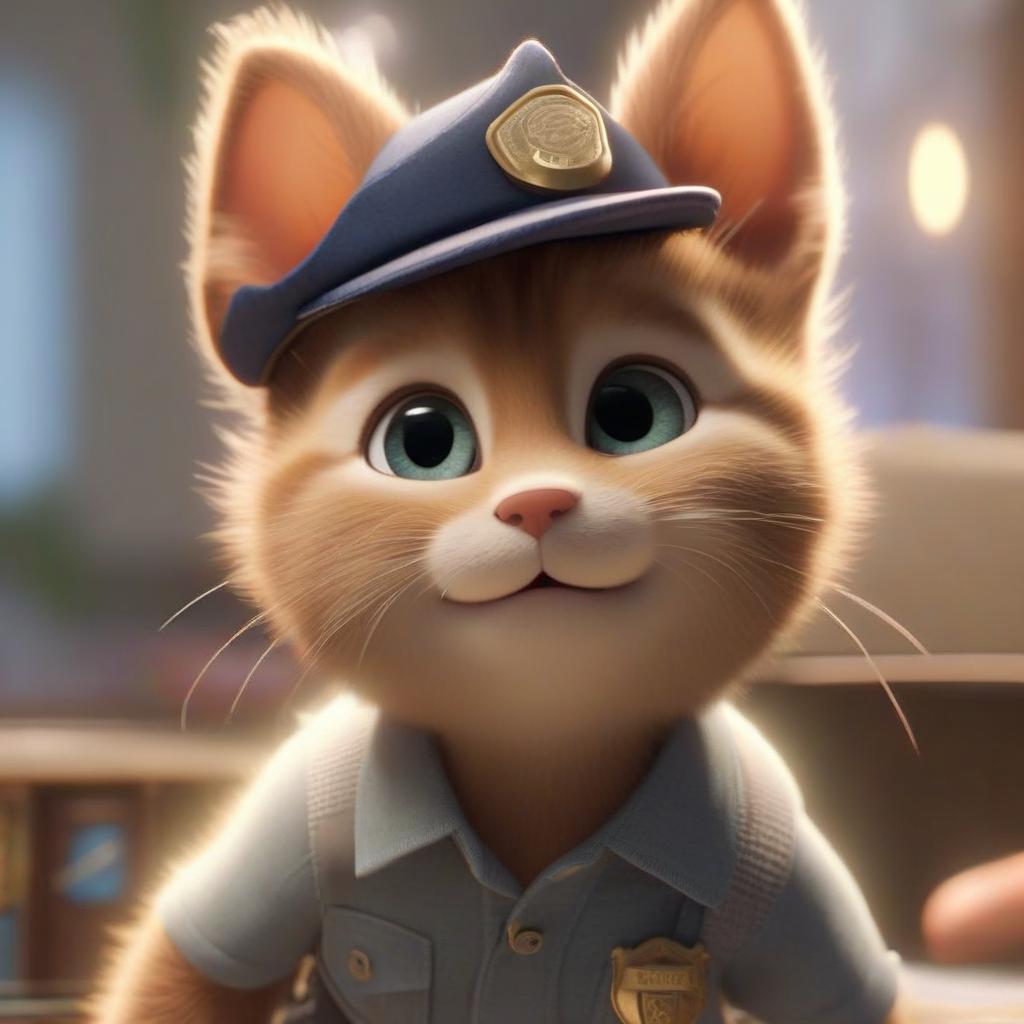}} &
        {\includegraphics[valign=c, width=\ww]{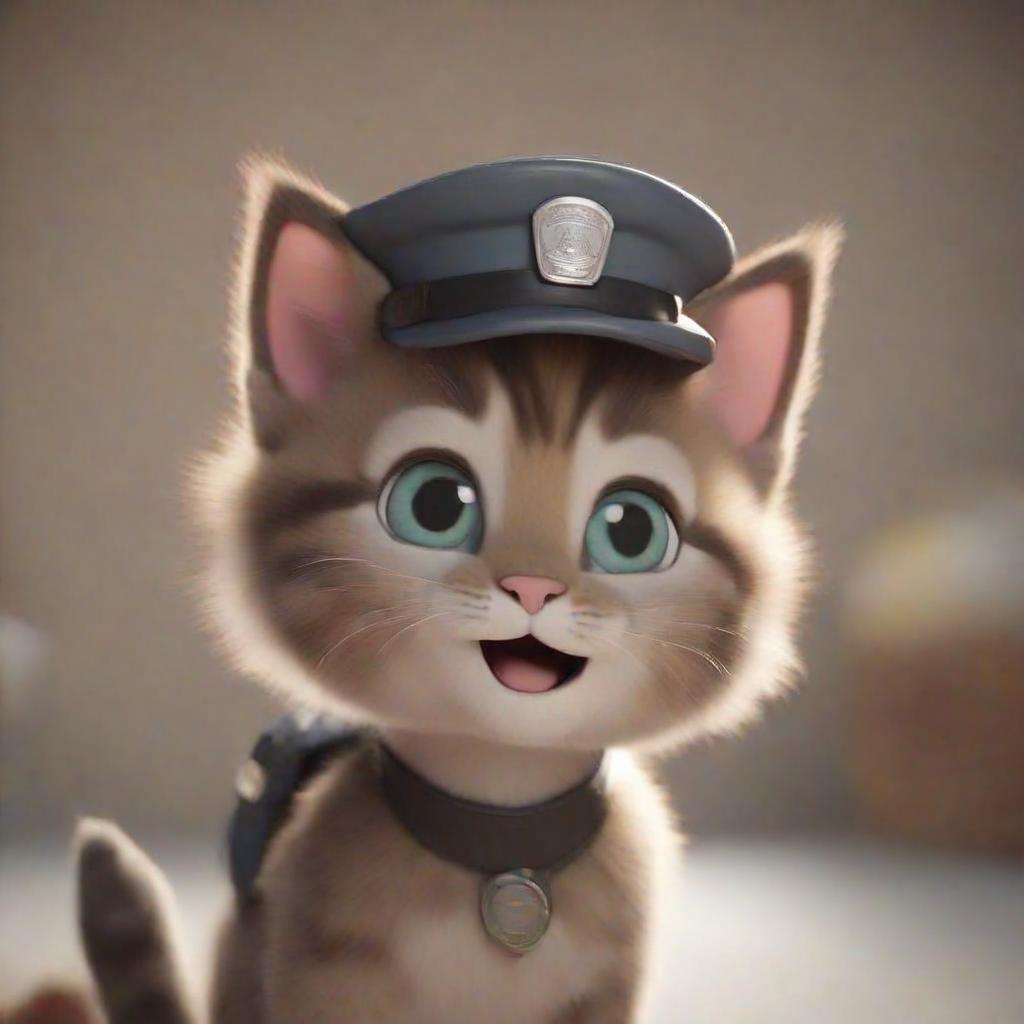}}
        \\
        \\

        \\
        \\
        \multicolumn{7}{c}{\textit{``a 3D animation of a playful kitten, with bright eyes and a}}
        \\
        \multicolumn{7}{c}{\textit{ mischievous expression, embodying youthful curiosity and joy''}}

    \end{tabular}
    
    \caption{\textbf{Qualitative comparison to baselines on the automatically generated prompts.} We compared our method against several baselines: TI \cite{Gal2022AnII}, BLIP-diffusion \cite{Li2023BLIPDiffusionPS} and IP-adapter \cite{Ye2023IPAdapterTC} are able to correspond to the target prompt but fail to produce consistent results. LoRA DB \cite{lora_diffusion} is able to achieve consistency, but it does not always follow to the prompt, in addition, the generate character is being generated in the same fixed pose. ELITE \cite{Wei2023ELITEEV} struggles with following the prompt and also tends to generate deformed characters. Our method is able to follow the prompt, and generate consistent characters in different poses and viewing angles.}
    \label{fig:automatic_qualitative_baselines_comparison}
\end{figure*}

%% file: figures/qualitative_comparison/fig_additional.tex
\begin{figure*}[t]
    \centering
    \setlength{\tabcolsep}{3.5pt}
    \renewcommand{\arraystretch}{0.4}
    \setlength{\ww}{0.29\columnwidth}
    \begin{tabular}{ccccccc}
        &
        \textbf{TI} &
        \textbf{LoRA DB} &
        \textbf{ELITE} &
        \textbf{BLIP-diff} &
        \textbf{IP-Adapter} &
        \textbf{Ours}
        \\

        &
        \cite{Gal2022AnII} &
        \cite{lora_diffusion} &
        \cite{Wei2023ELITEEV} &
        \cite{Li2023BLIPDiffusionPS} &
        \cite{Ye2023IPAdapterTC} &
        \\

        \\
        \rotatebox[origin=c]{90}{\phantom{a}}
        \rotatebox[origin=c]{90}{\textit{``in the}}
        \rotatebox[origin=c]{90}{\textit{desert''}} &
        {\includegraphics[valign=c, width=\ww]{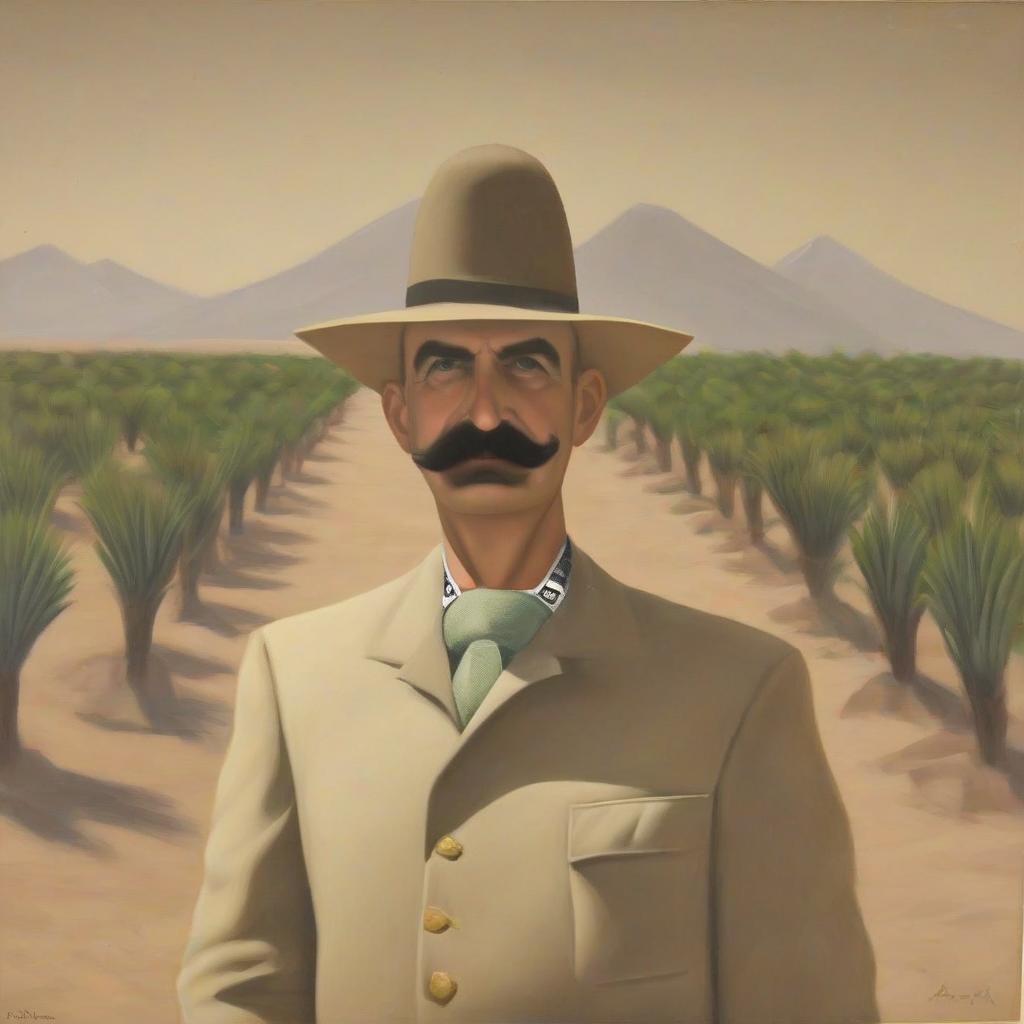}} &
        {\includegraphics[valign=c, width=\ww]{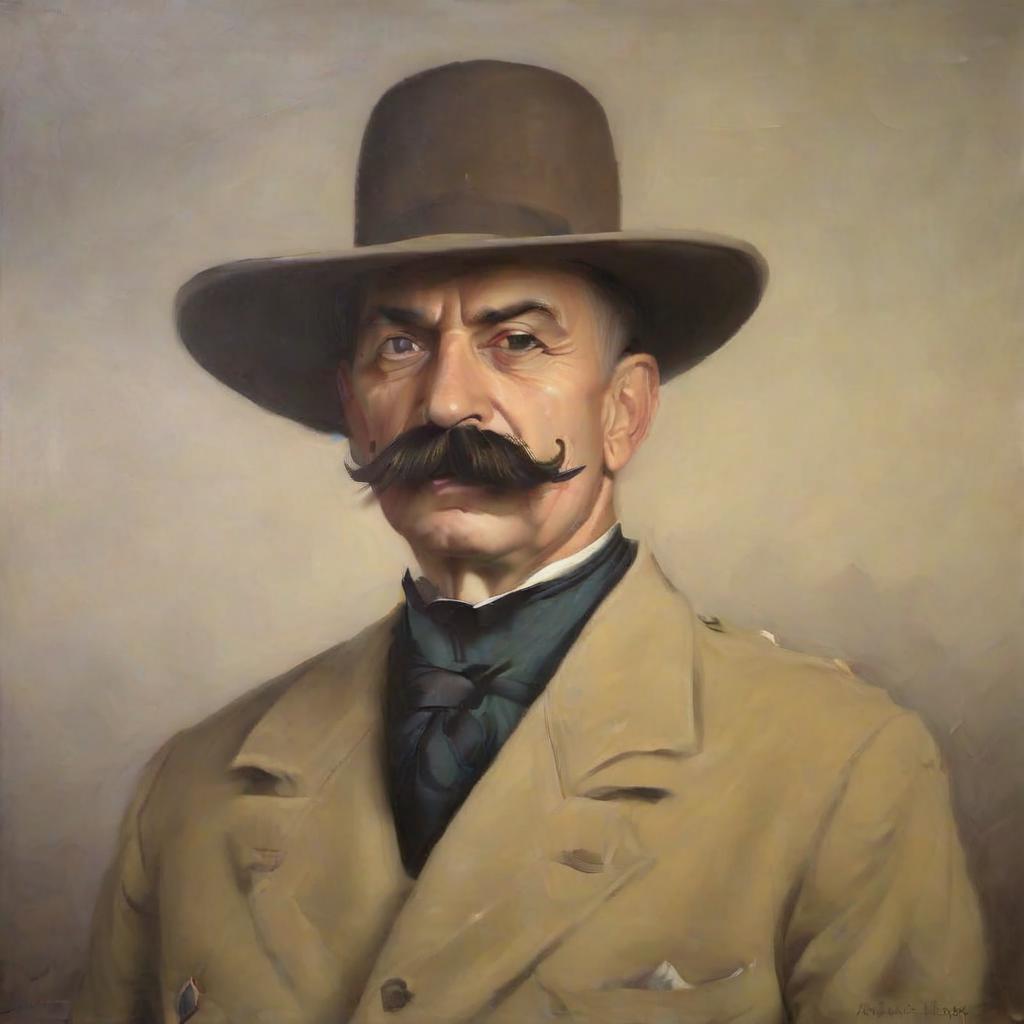}} &
        {\includegraphics[valign=c, width=\ww]{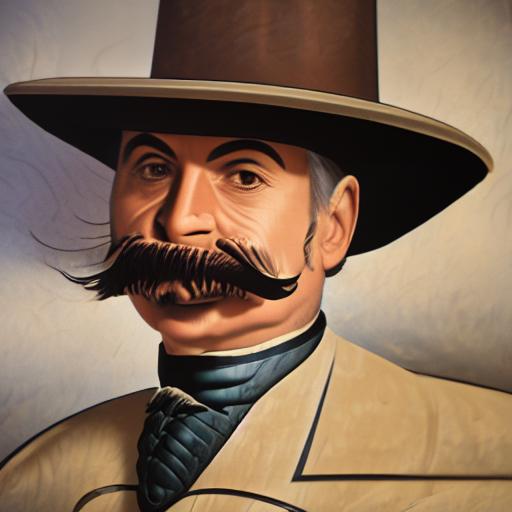}} &
        {\includegraphics[valign=c, width=\ww]{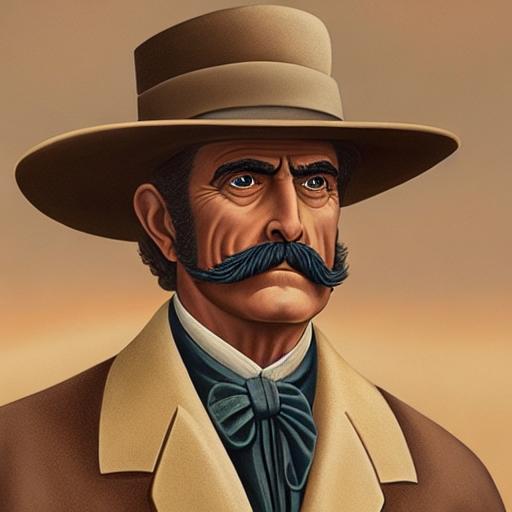}} &
        {\includegraphics[valign=c, width=\ww]{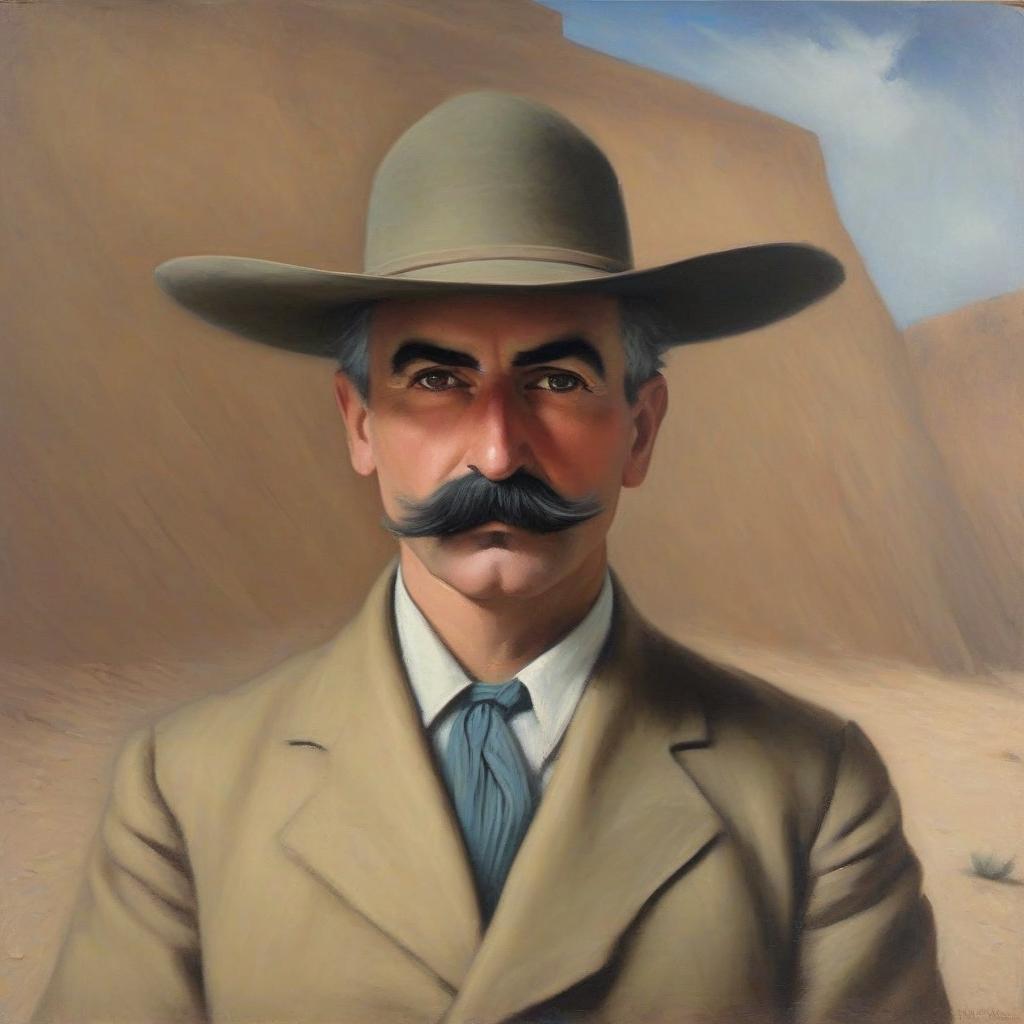}} &
        {\includegraphics[valign=c, width=\ww]{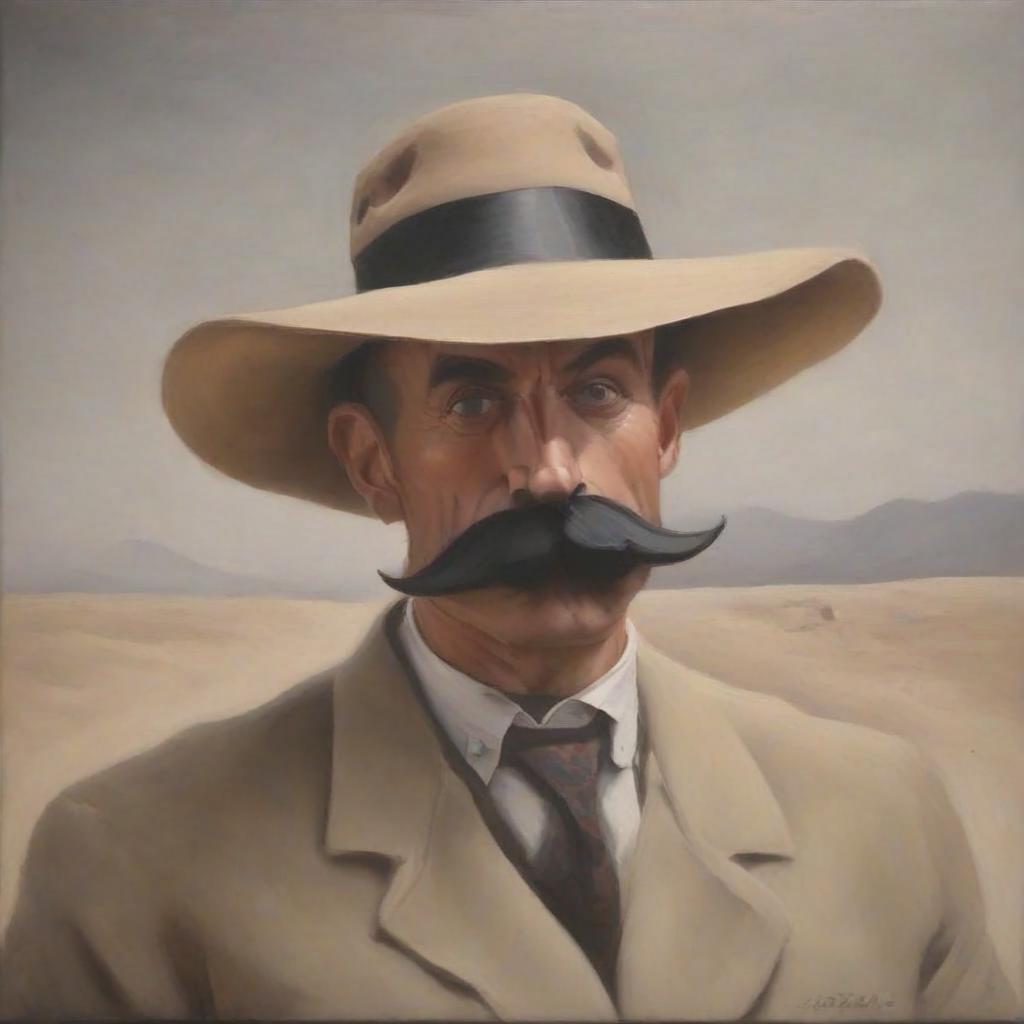}}
        \\
        \\

        \rotatebox[origin=c]{90}{\textit{``taking a}}
        \rotatebox[origin=c]{90}{\textit{picture with}}
        \rotatebox[origin=c]{90}{\textit{his phone''}} &
        {\includegraphics[valign=c, width=\ww]{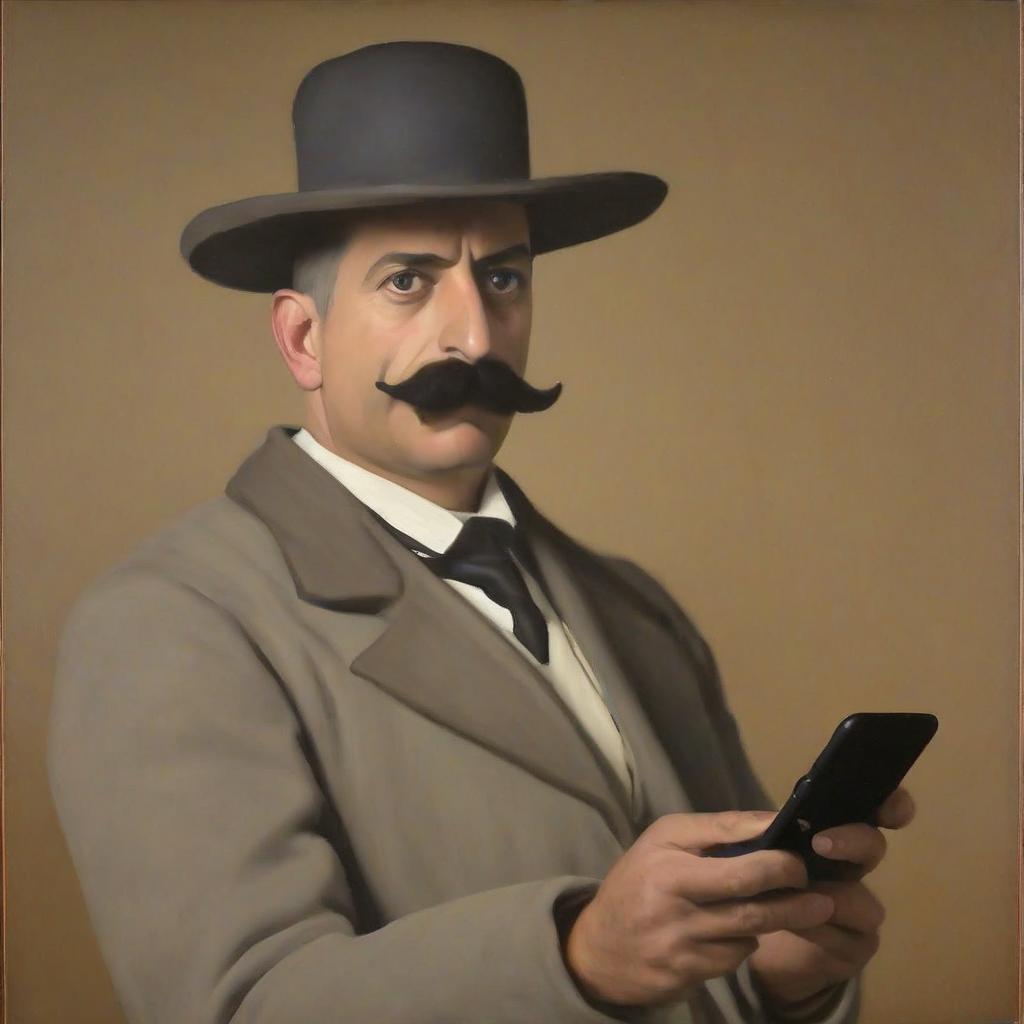}} &
        {\includegraphics[valign=c, width=\ww]{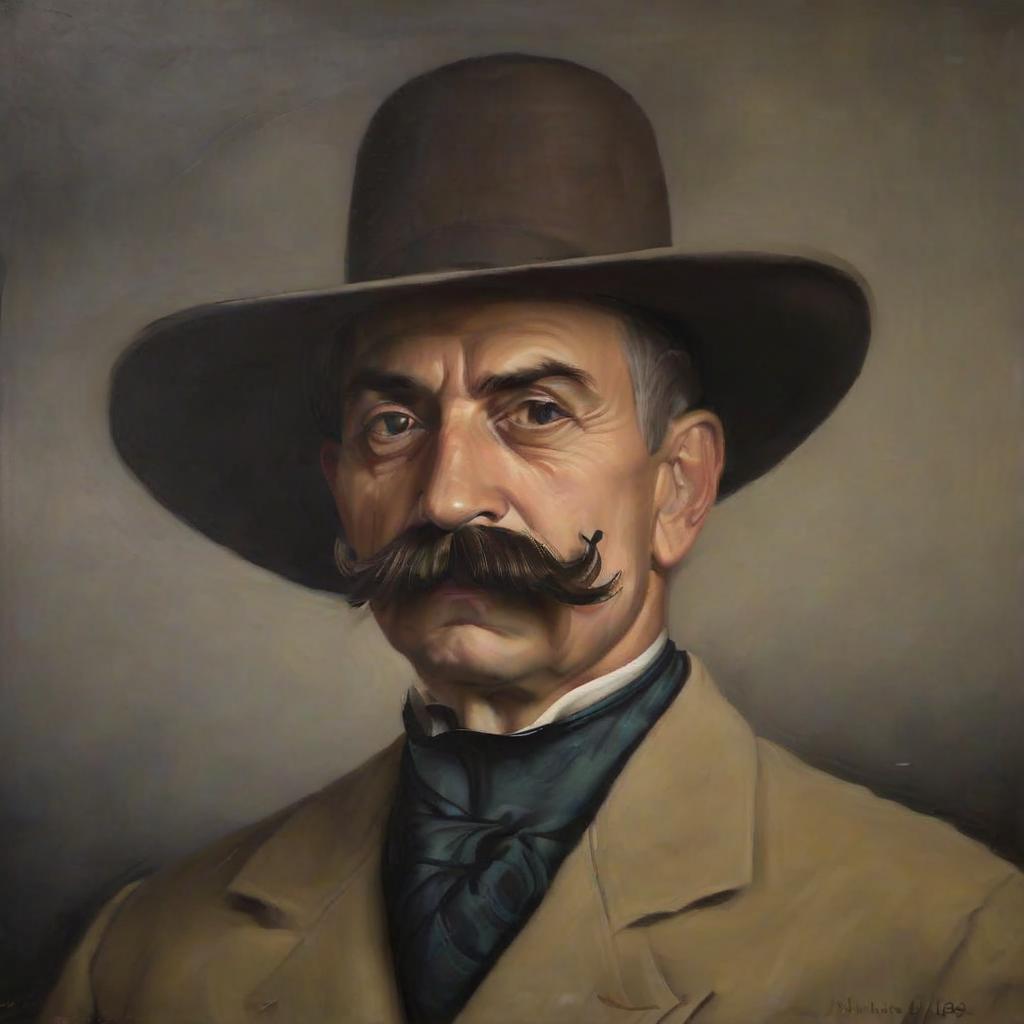}} &
        {\includegraphics[valign=c, width=\ww]{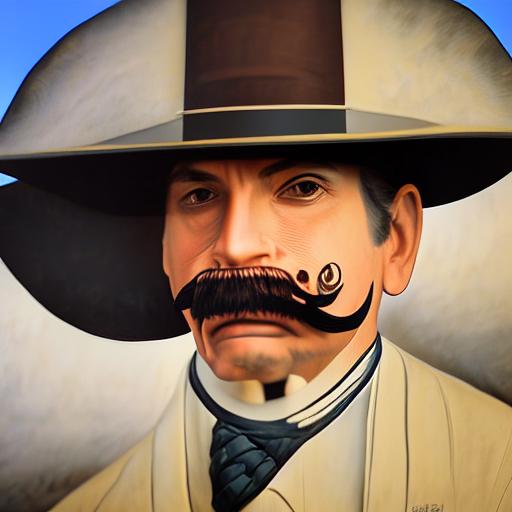}} &
        {\includegraphics[valign=c, width=\ww]{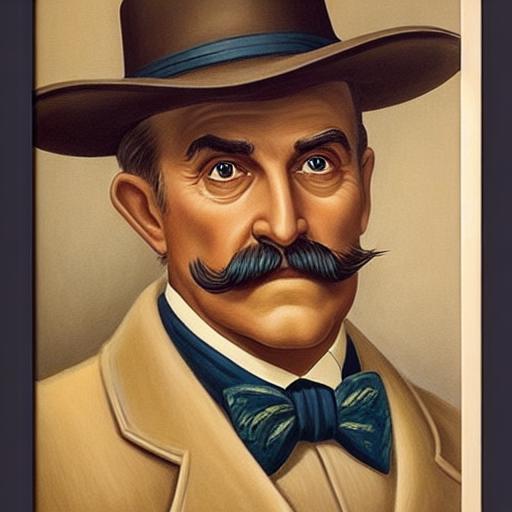}} &
        {\includegraphics[valign=c, width=\ww]{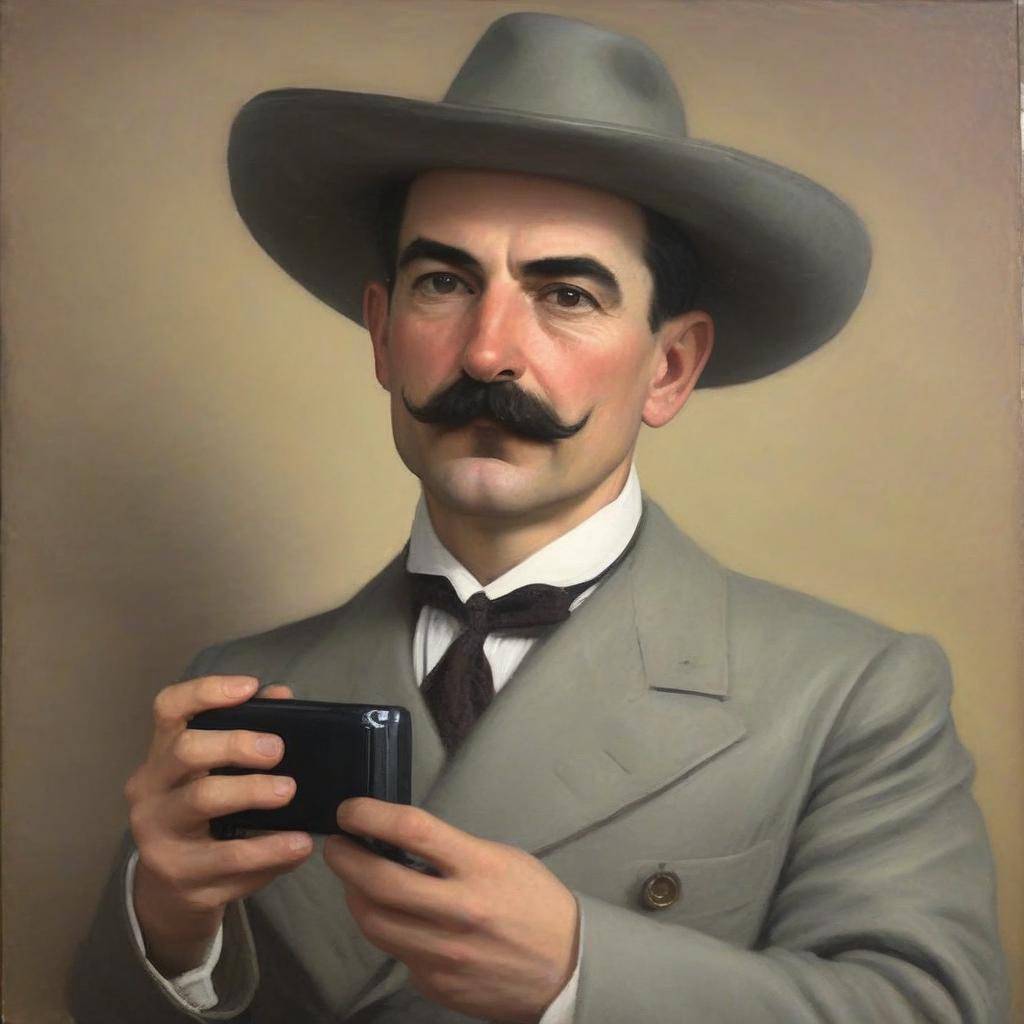}} &
        {\includegraphics[valign=c, width=\ww]{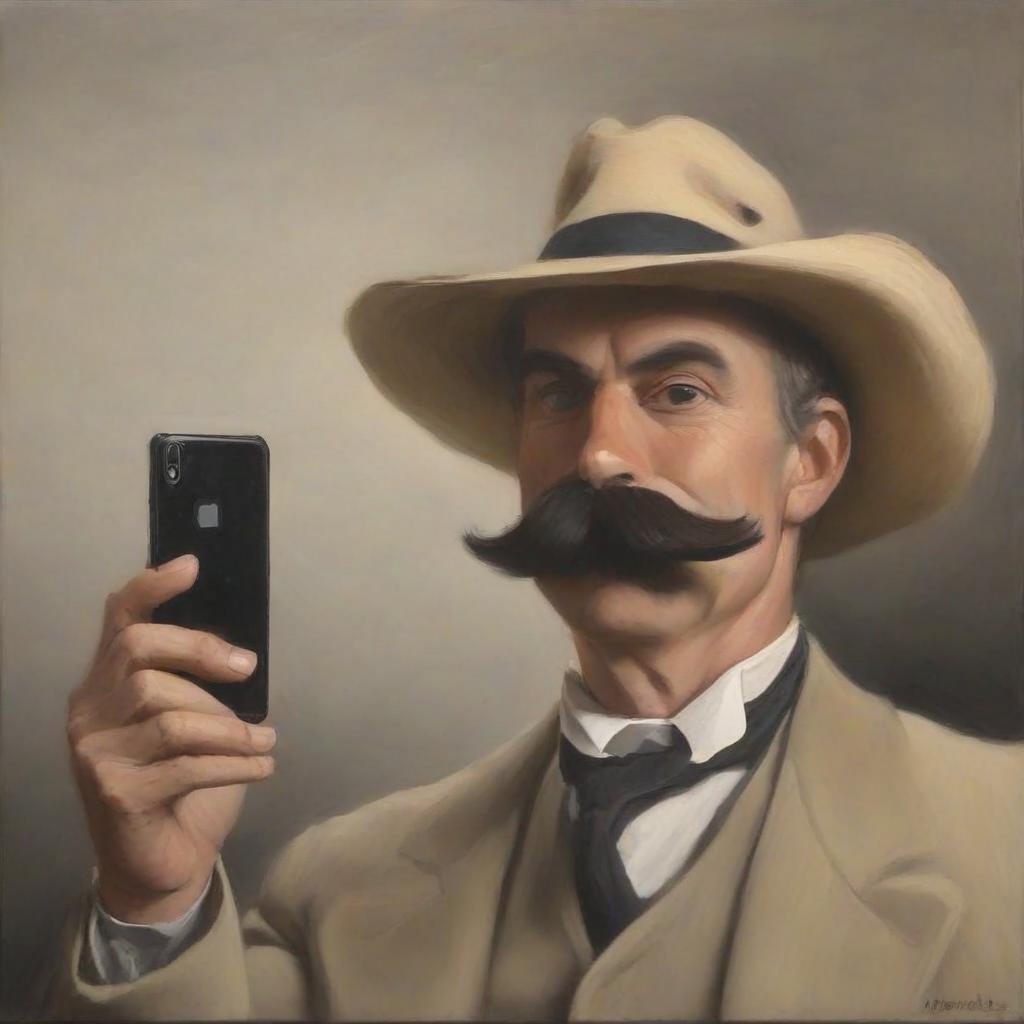}}
        \\
        \\

        \multicolumn{7}{c}{\textit{``an oil painting of a man with a mustache and a hat''}}
        \\
        \\
        \midrule

        \\
        \rotatebox[origin=c]{90}{\phantom{a}}
        \rotatebox[origin=c]{90}{\textit{``working on}}
        \rotatebox[origin=c]{90}{\textit{his laptop''}} &
        {\includegraphics[valign=c, width=\ww]{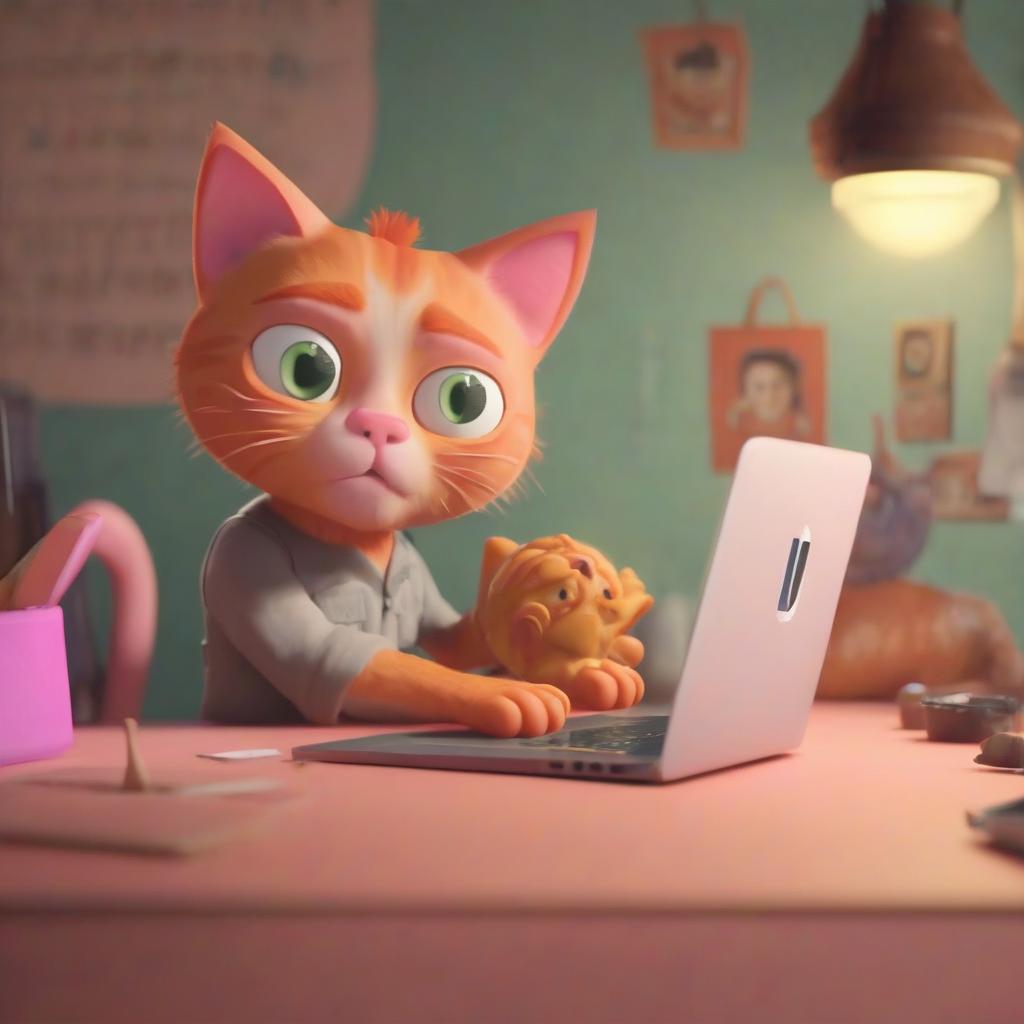}} &
        {\includegraphics[valign=c, width=\ww]{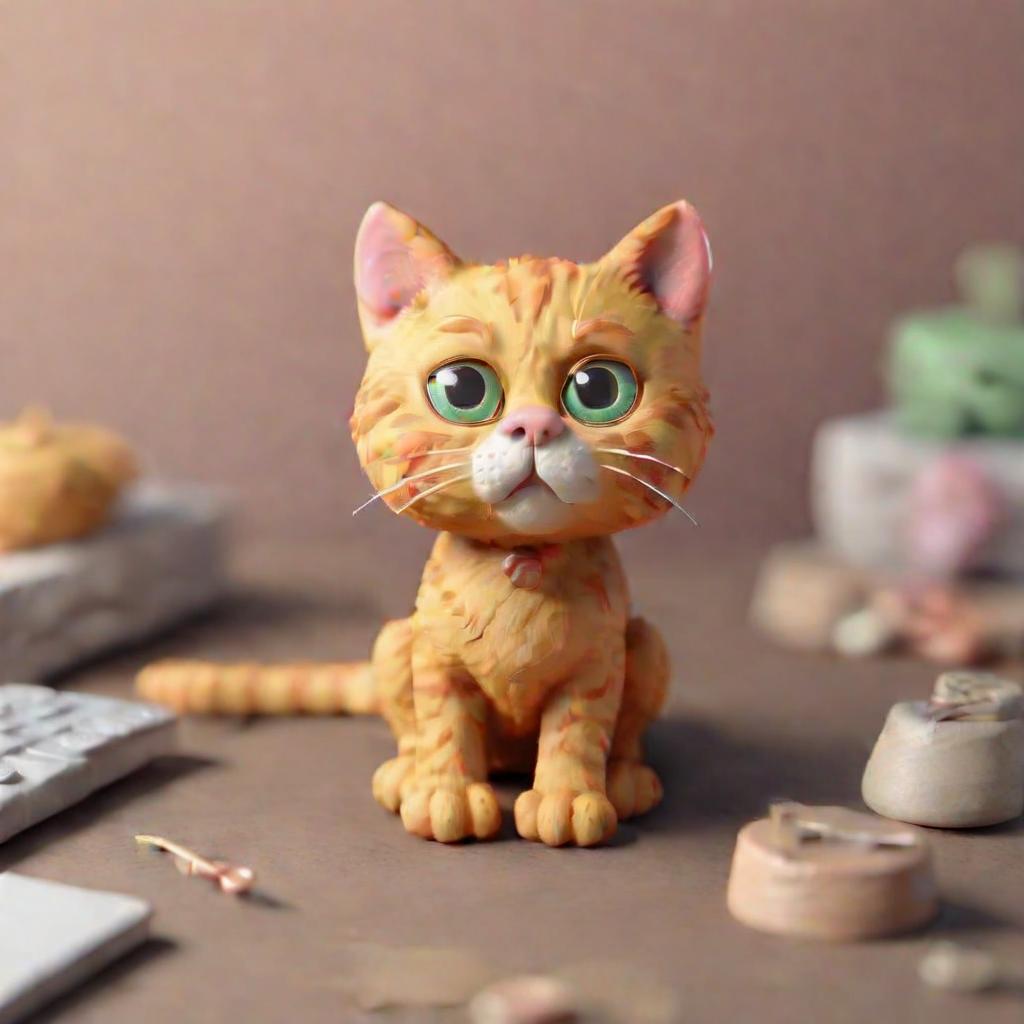}} &
        {\includegraphics[valign=c, width=\ww]{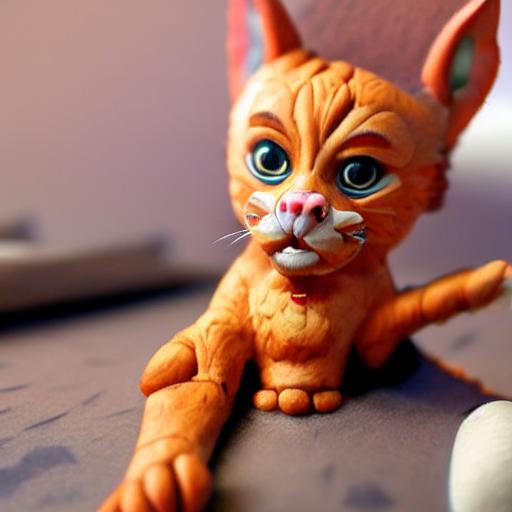}} &
        {\includegraphics[valign=c, width=\ww]{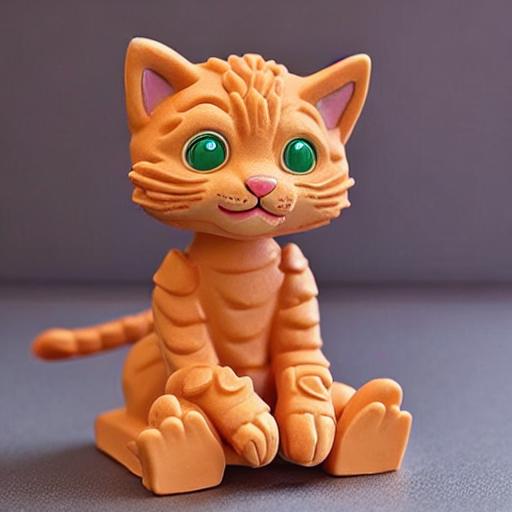}} &
        {\includegraphics[valign=c, width=\ww]{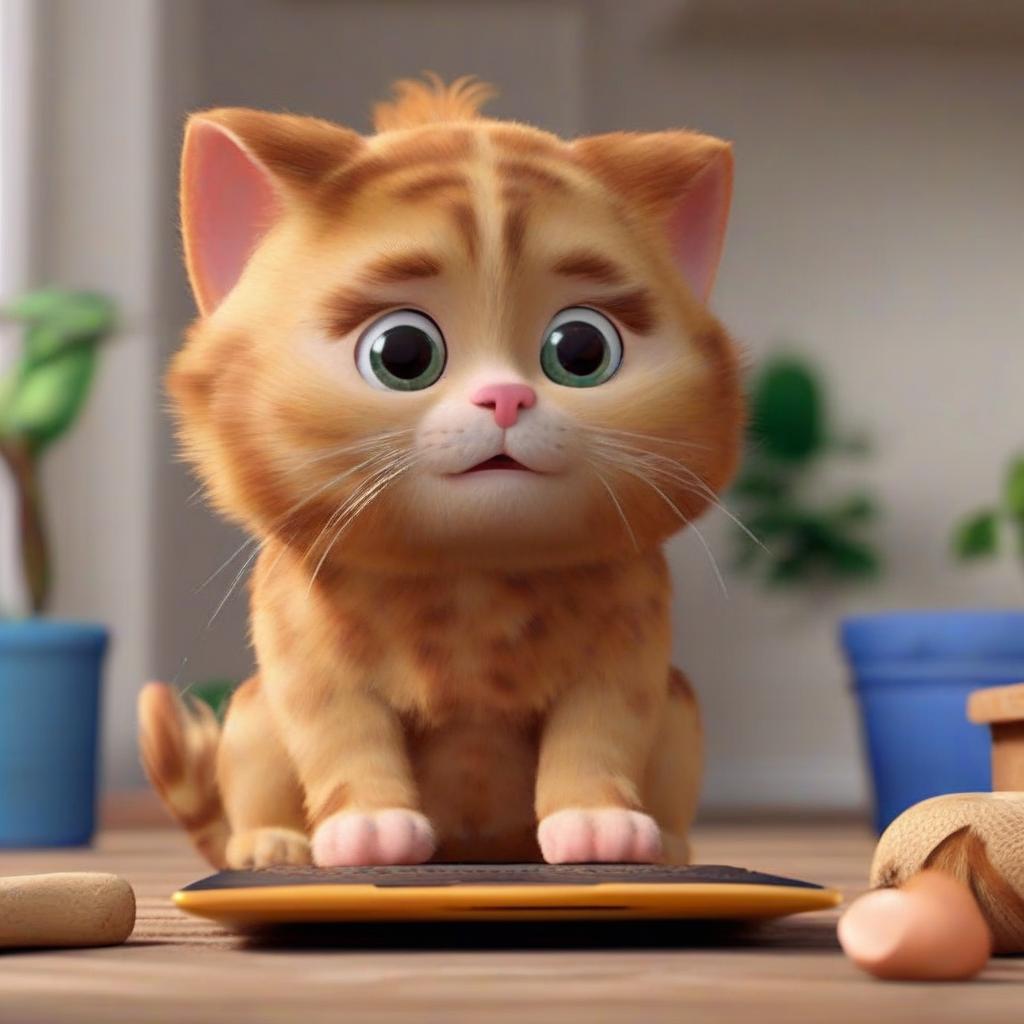}} &
        {\includegraphics[valign=c, width=\ww]{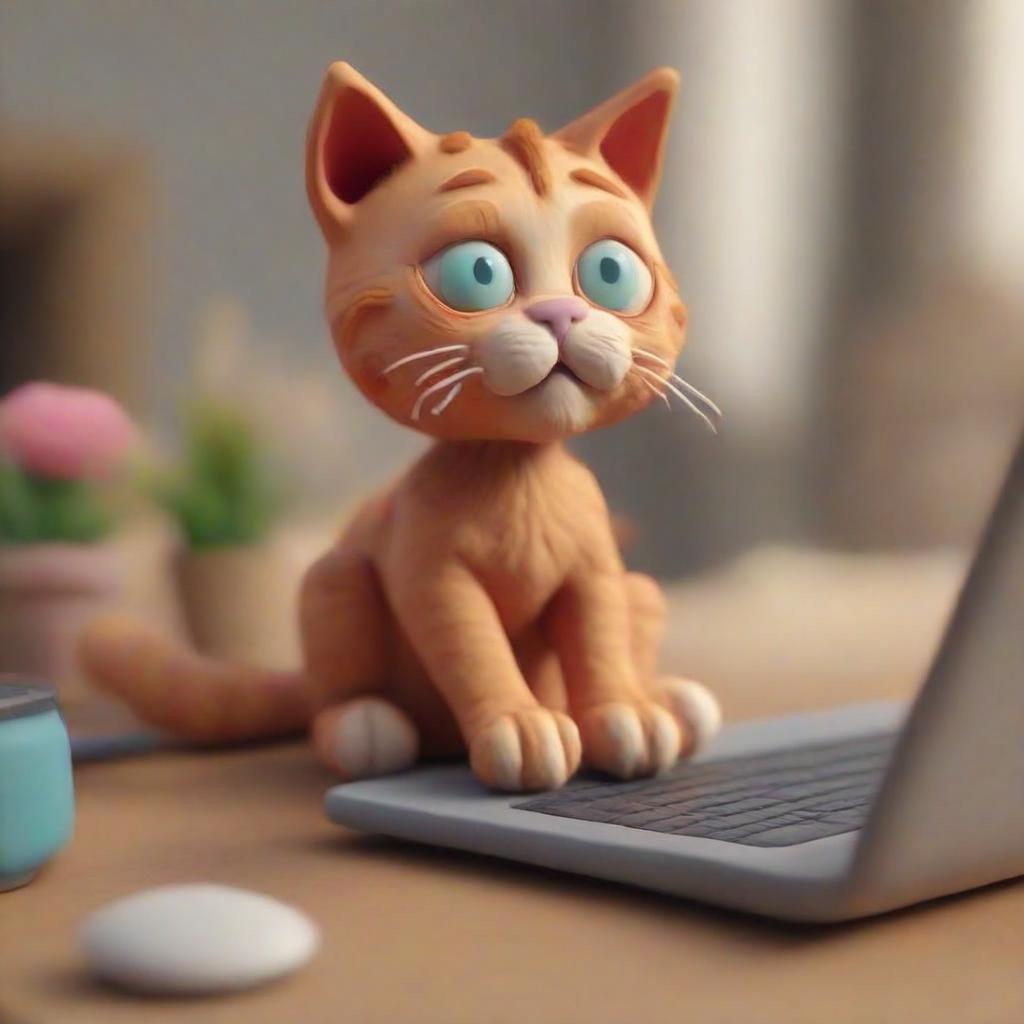}}
        \\
        \\

        \rotatebox[origin=c]{90}{\phantom{a}}
        \rotatebox[origin=c]{90}{\textit{``eating}}
        \rotatebox[origin=c]{90}{\textit{a burger''}} &
        {\includegraphics[valign=c, width=\ww]{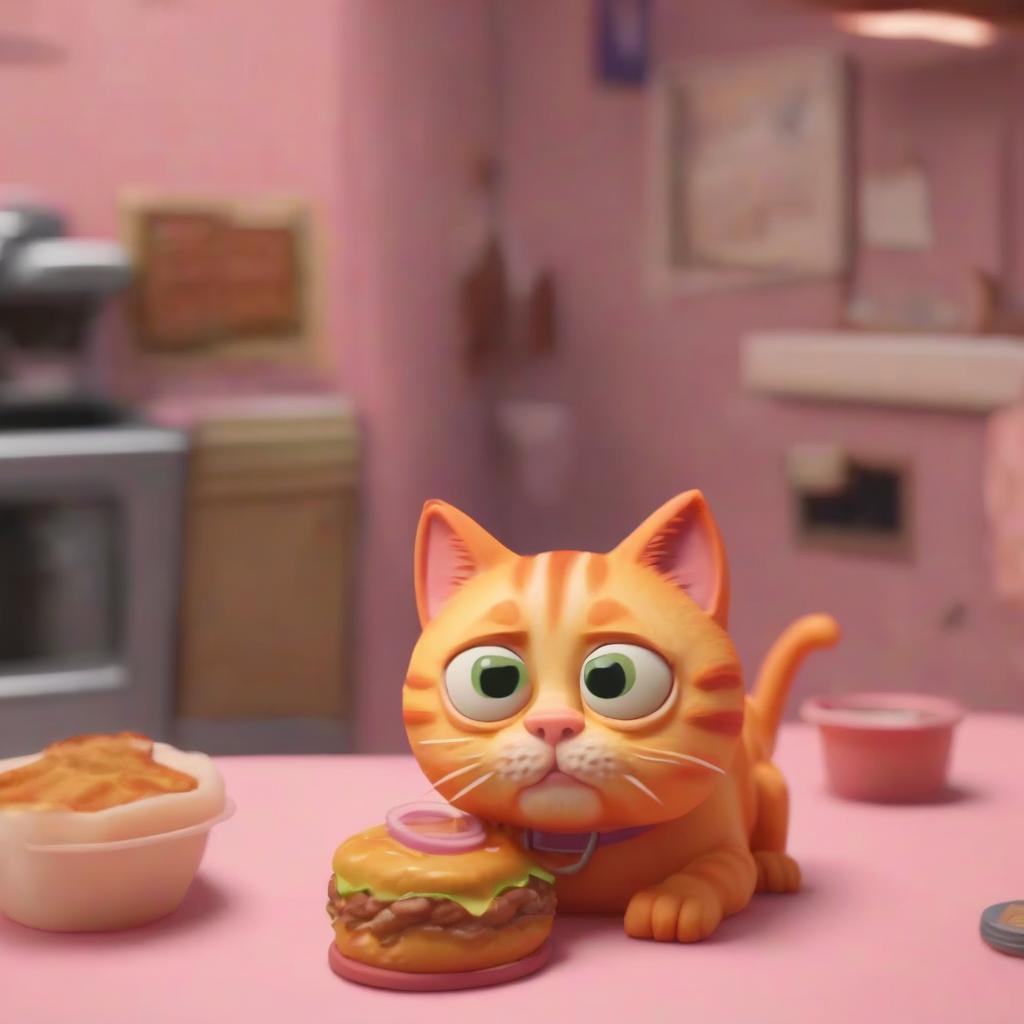}} &
        {\includegraphics[valign=c, width=\ww]{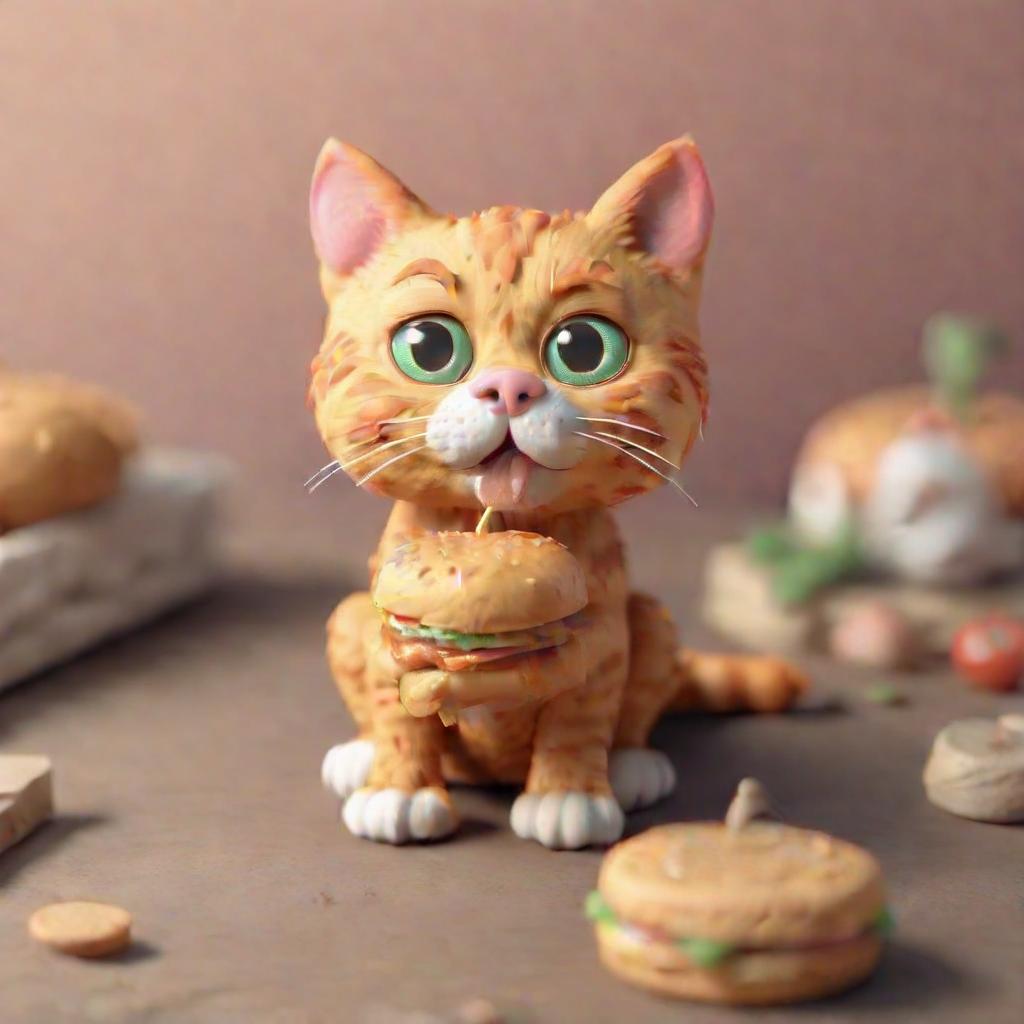}} &
        {\includegraphics[valign=c, width=\ww]{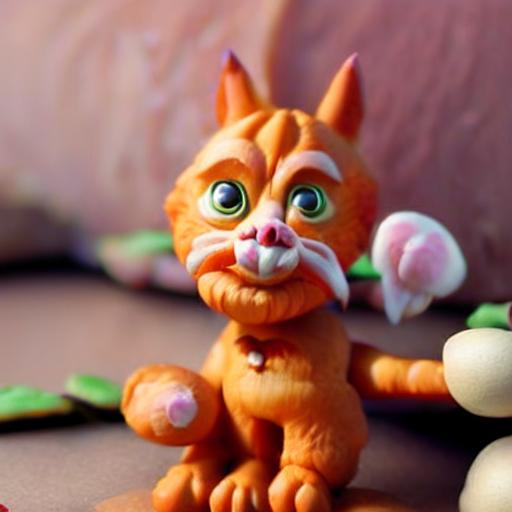}} &
        {\includegraphics[valign=c, width=\ww]{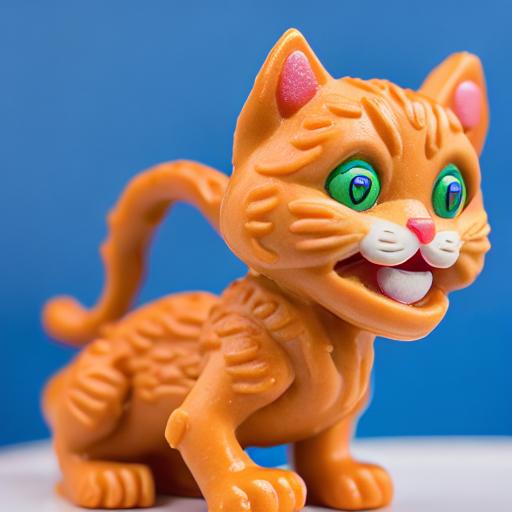}} &
        {\includegraphics[valign=c, width=\ww]{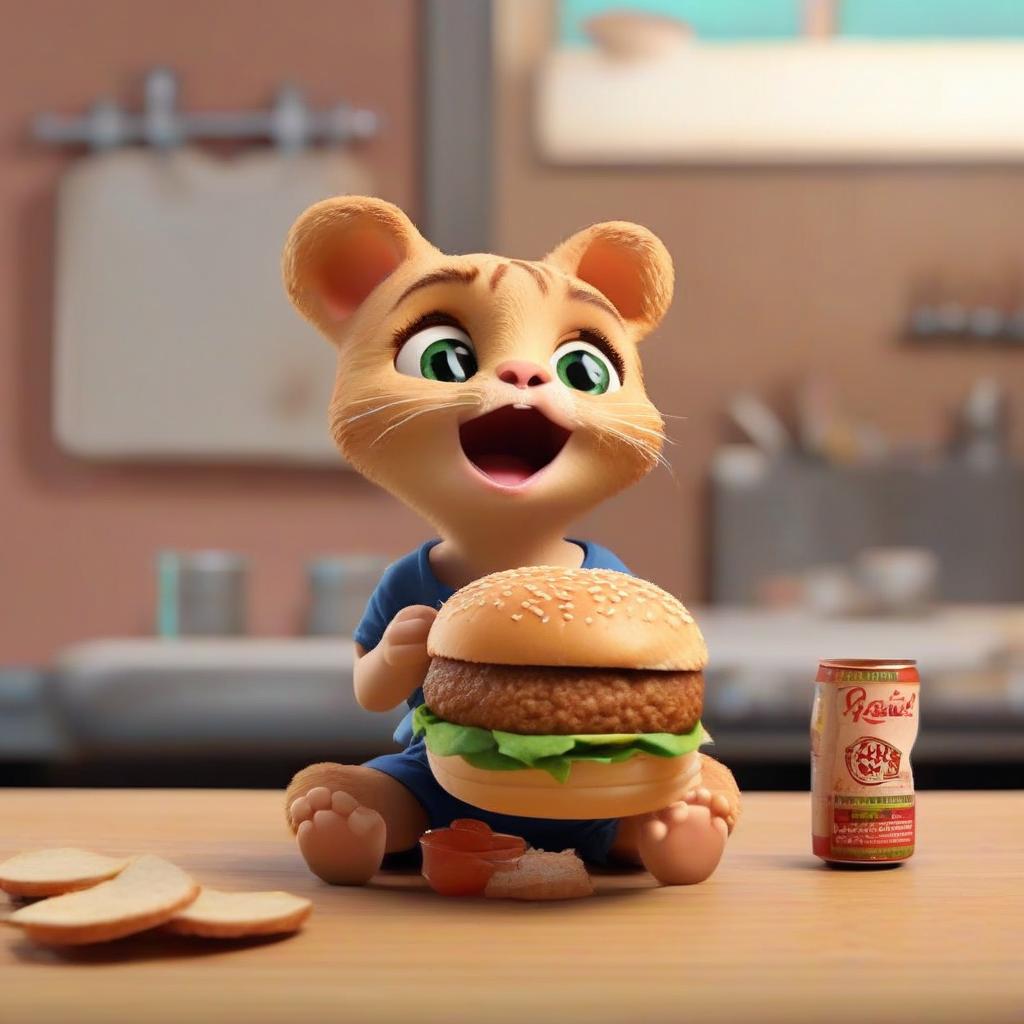}} &
        {\includegraphics[valign=c, width=\ww]{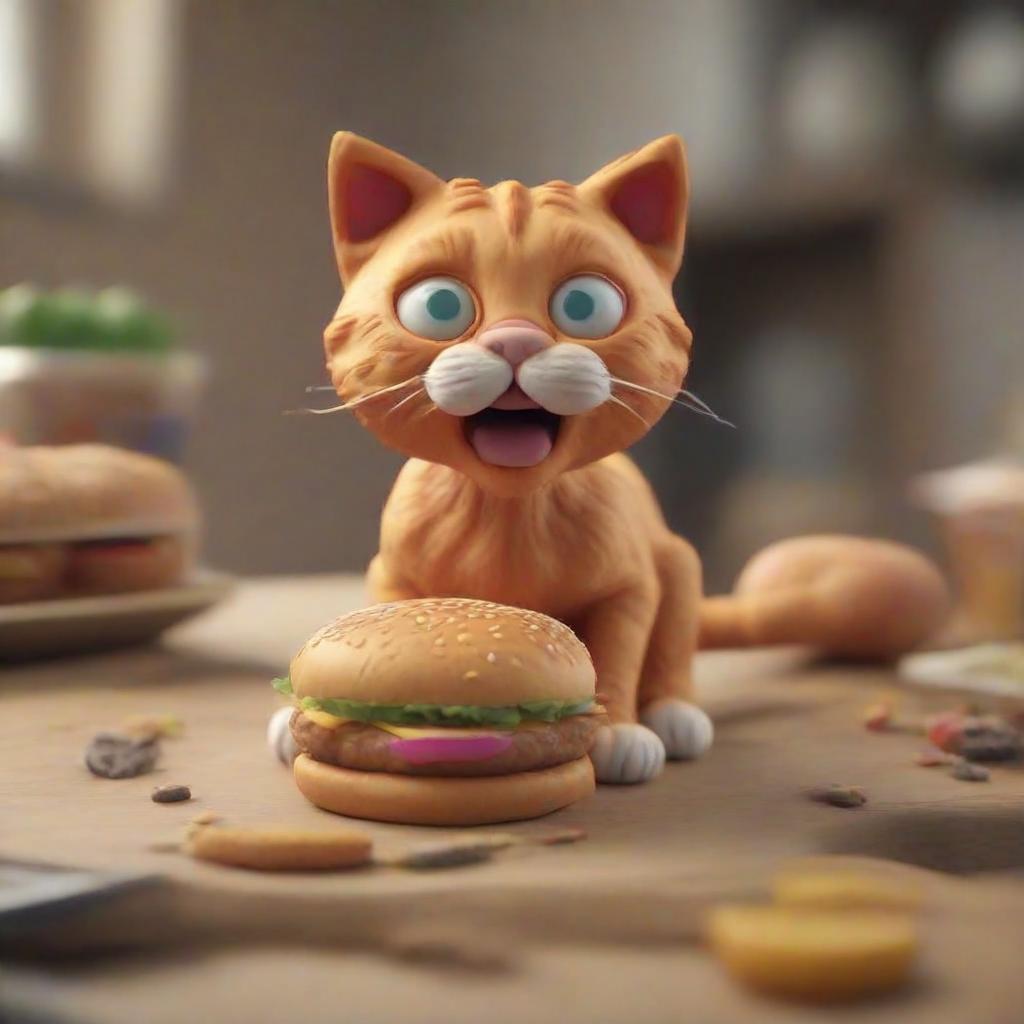}}
        \\
        \\

        \multicolumn{7}{c}{\textit{``a Plasticine of a cute baby cat with big eyes''}}
        \\
        \\
        \midrule

        \\
        \rotatebox[origin=c]{90}{\phantom{a}}
        \rotatebox[origin=c]{90}{\textit{``celebrating in}}
        \rotatebox[origin=c]{90}{\textit{a party''}} &
        {\includegraphics[valign=c, width=\ww]{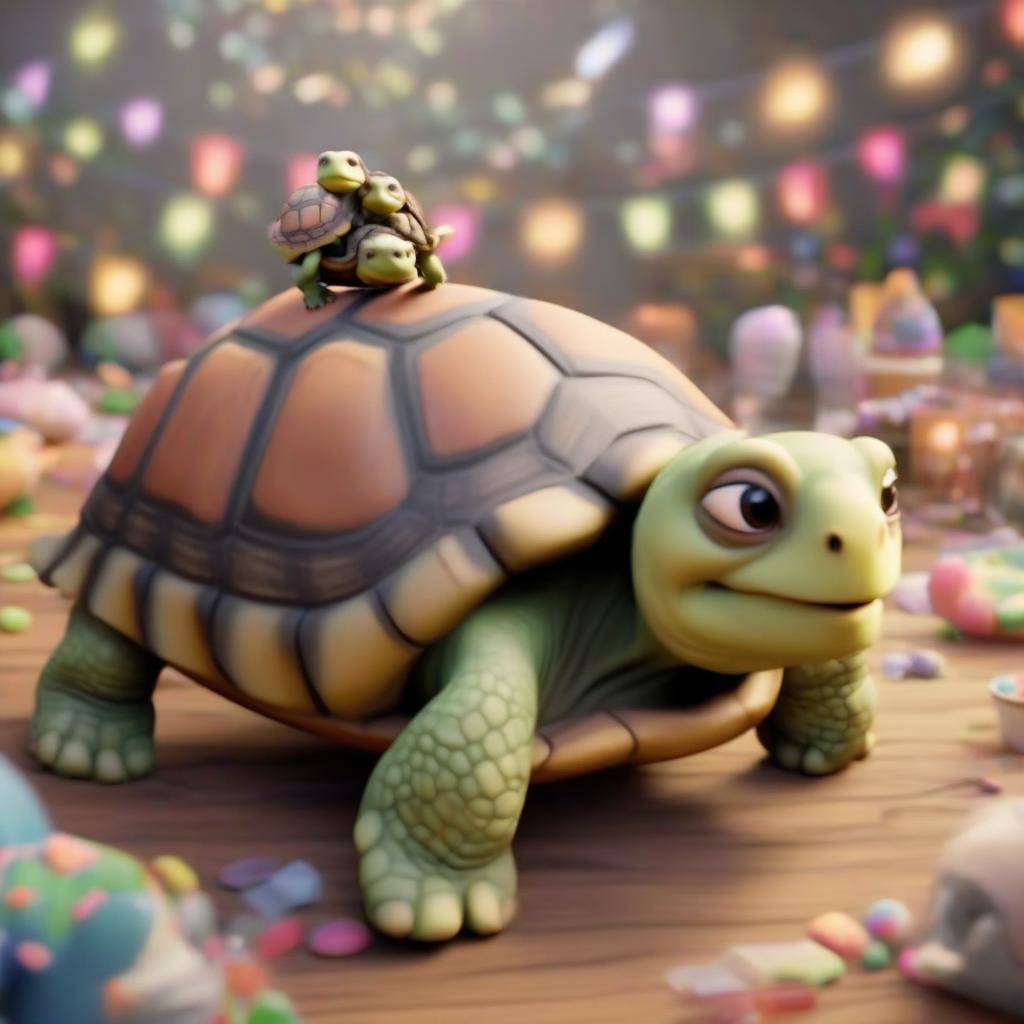}} &
        {\includegraphics[valign=c, width=\ww]{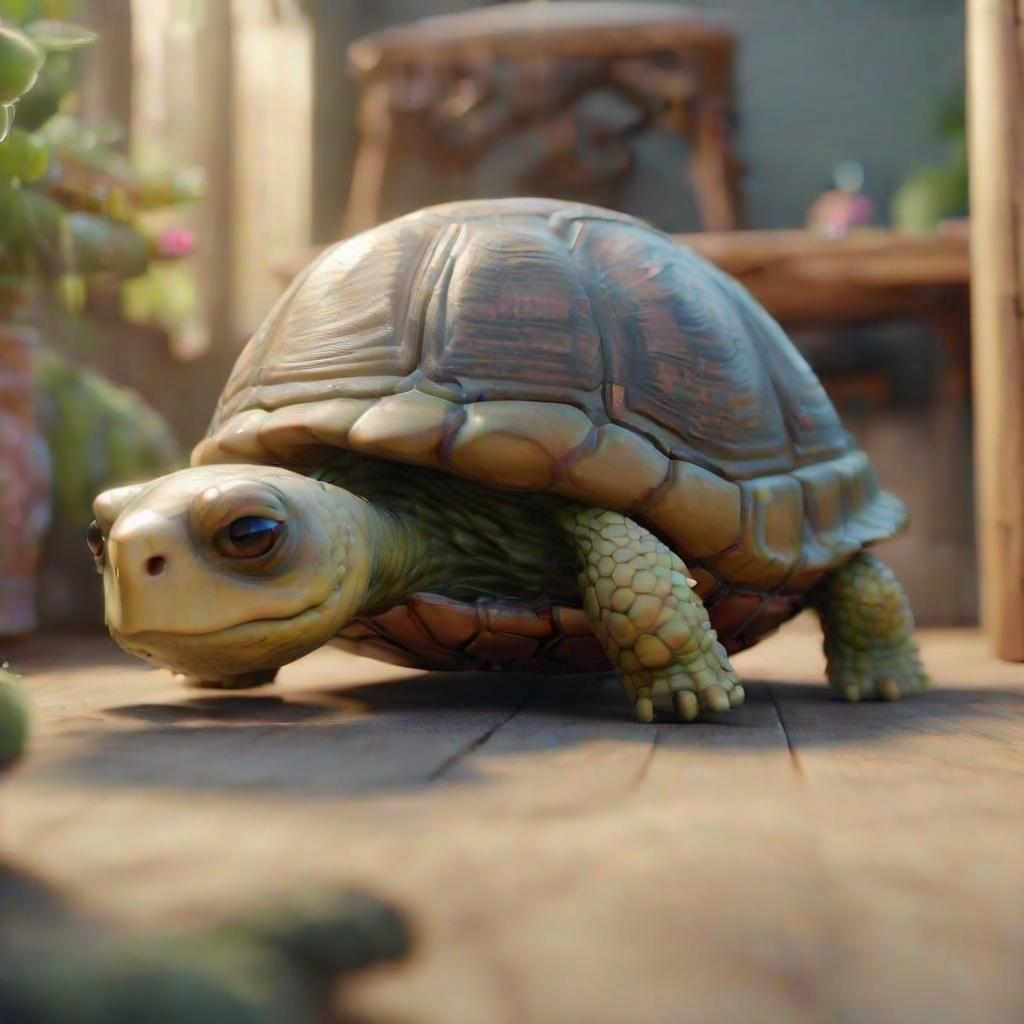}} &
        {\includegraphics[valign=c, width=\ww]{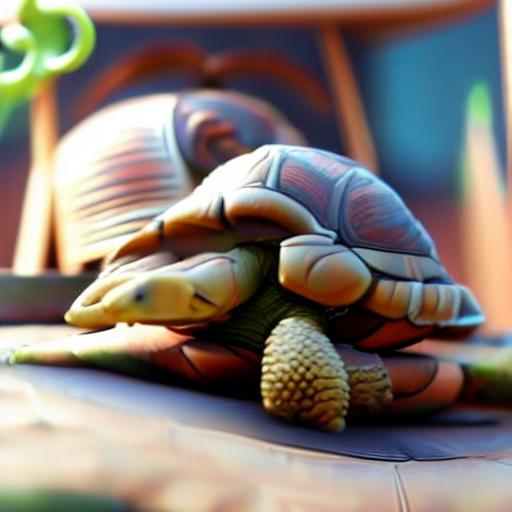}} &
        {\includegraphics[valign=c, width=\ww]{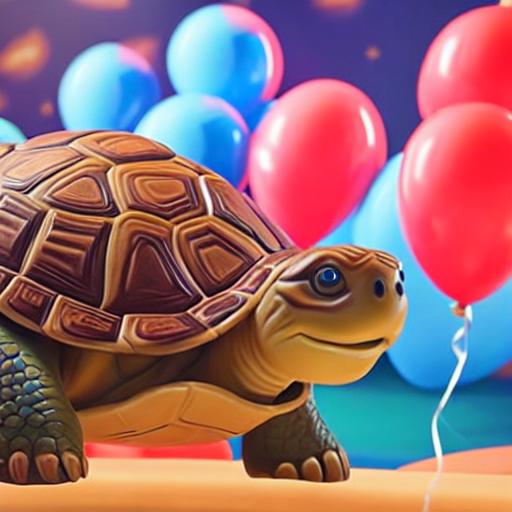}} &
        {\includegraphics[valign=c, width=\ww]{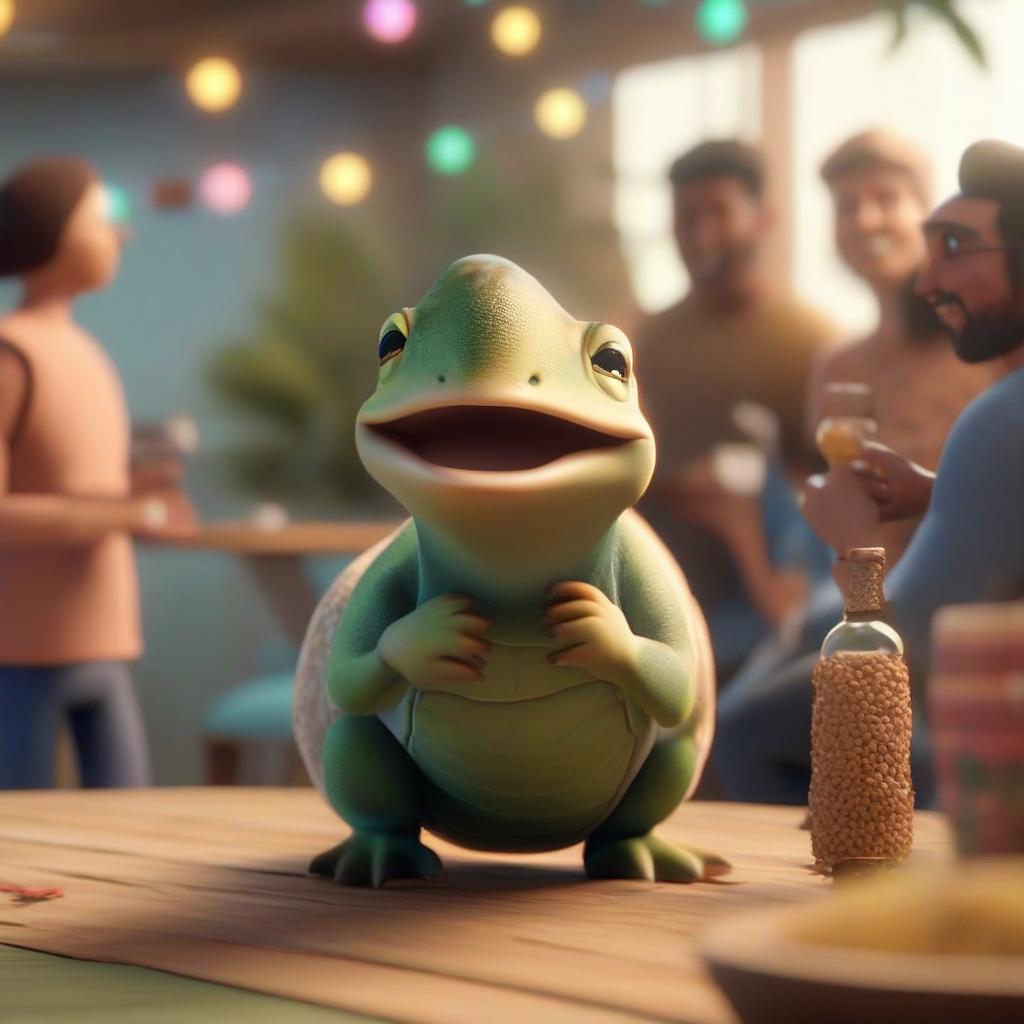}} &
        {\includegraphics[valign=c, width=\ww]{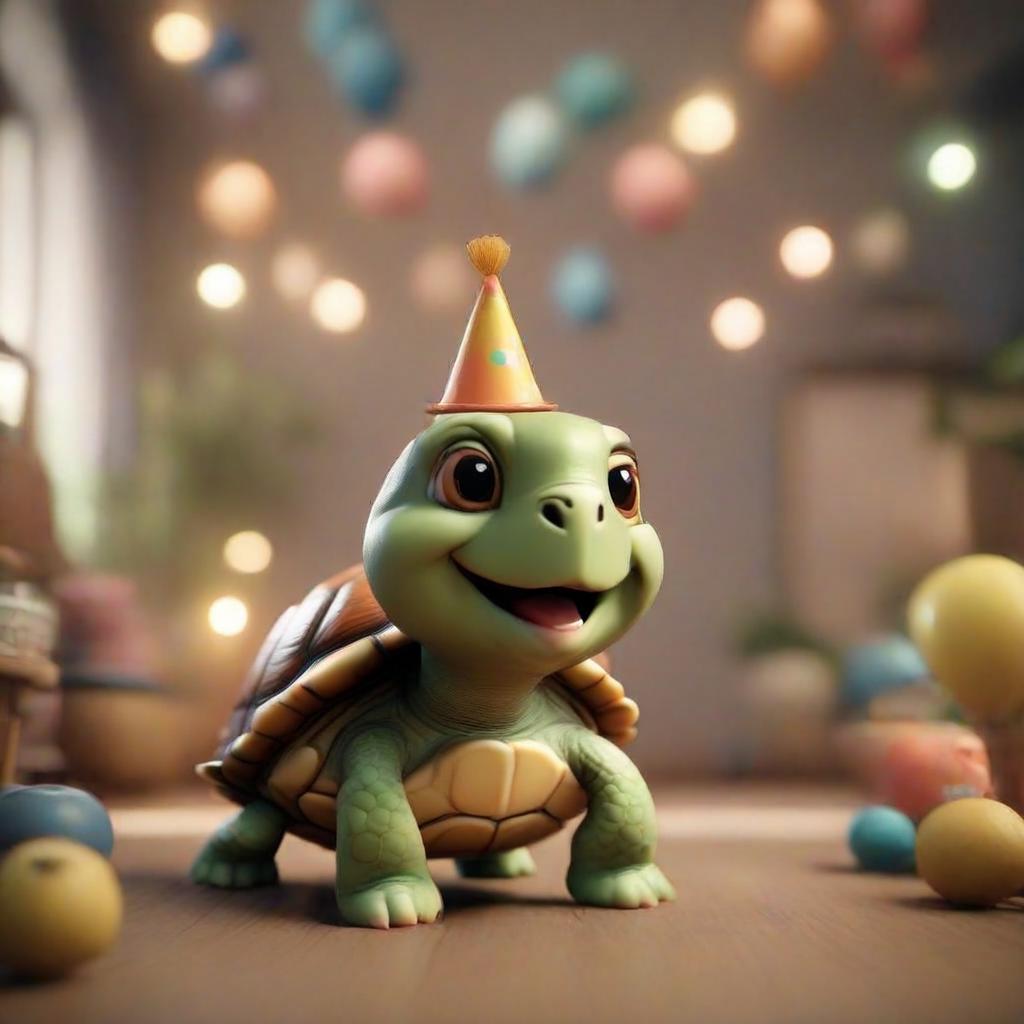}}
        \\
        \\

        \rotatebox[origin=c]{90}{\phantom{a}}
        \rotatebox[origin=c]{90}{\phantom{a}}
        \rotatebox[origin=c]{90}{\textit{``in the forest''}} &
        {\includegraphics[valign=c, width=\ww]{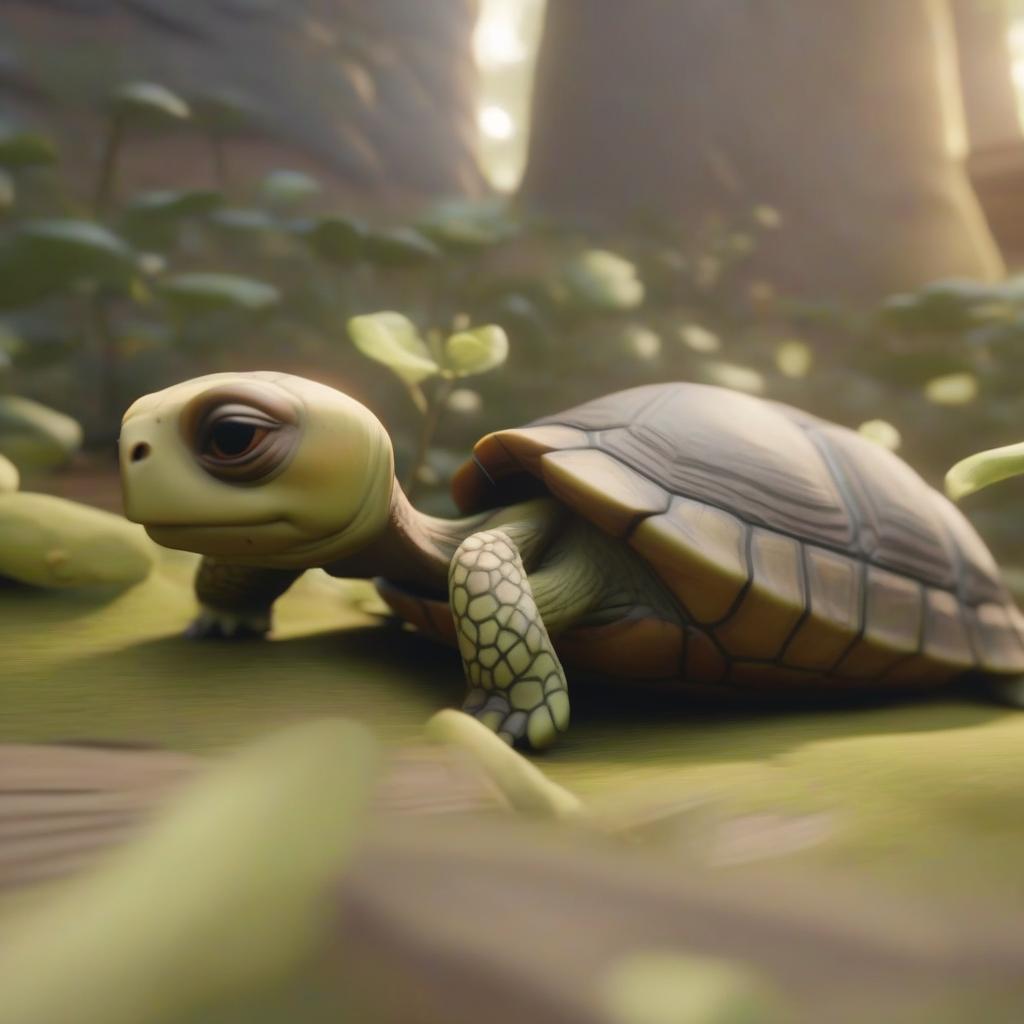}} &
        {\includegraphics[valign=c, width=\ww]{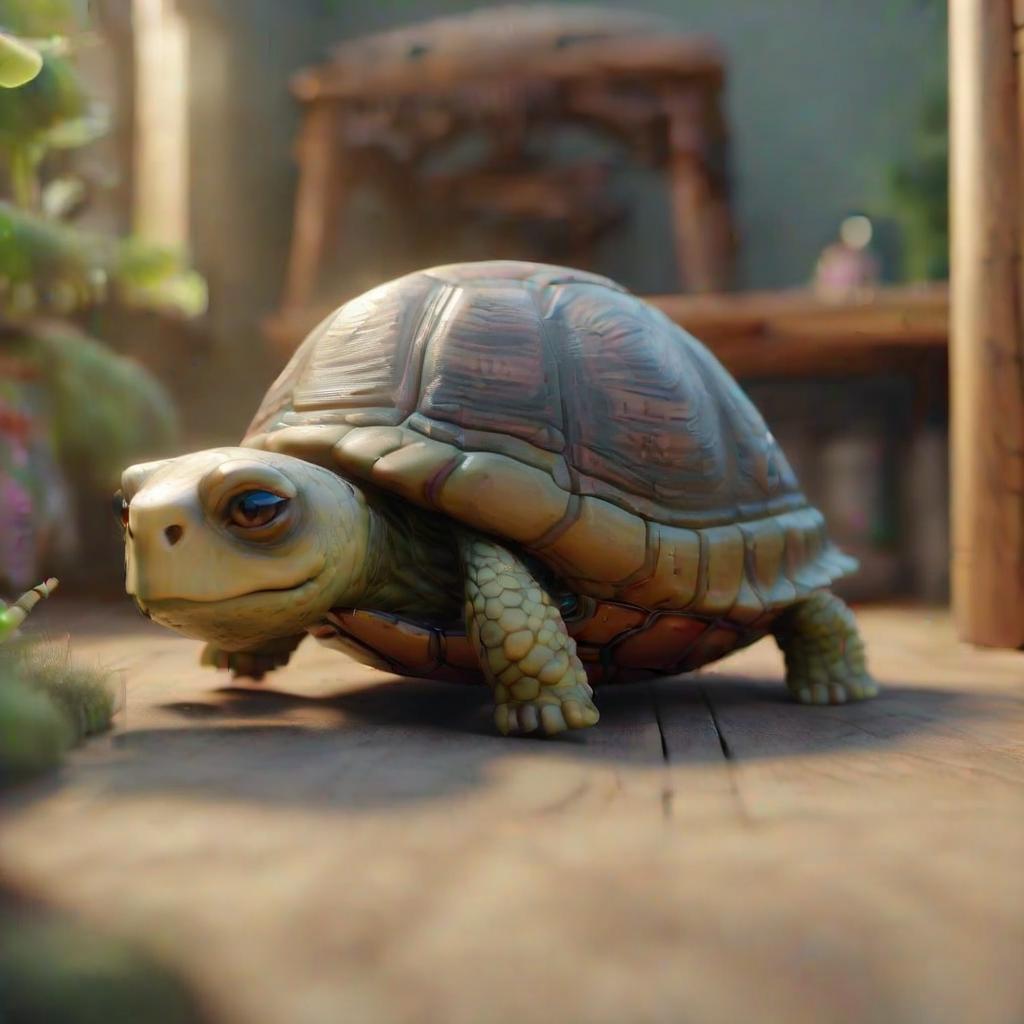}} &
        {\includegraphics[valign=c, width=\ww]{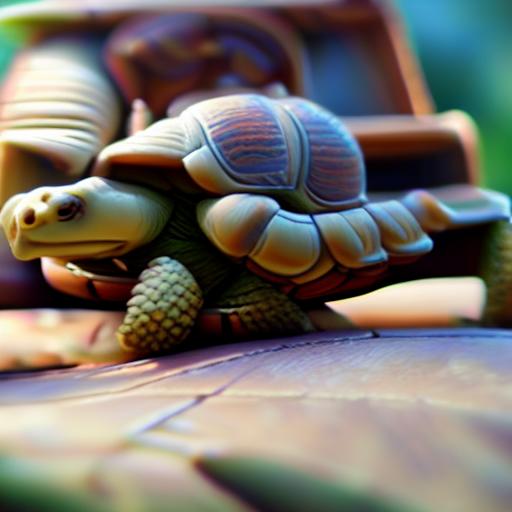}} &
        {\includegraphics[valign=c, width=\ww]{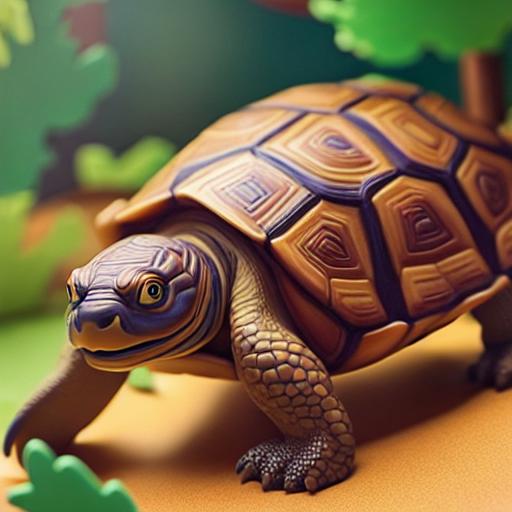}} &
        {\includegraphics[valign=c, width=\ww]{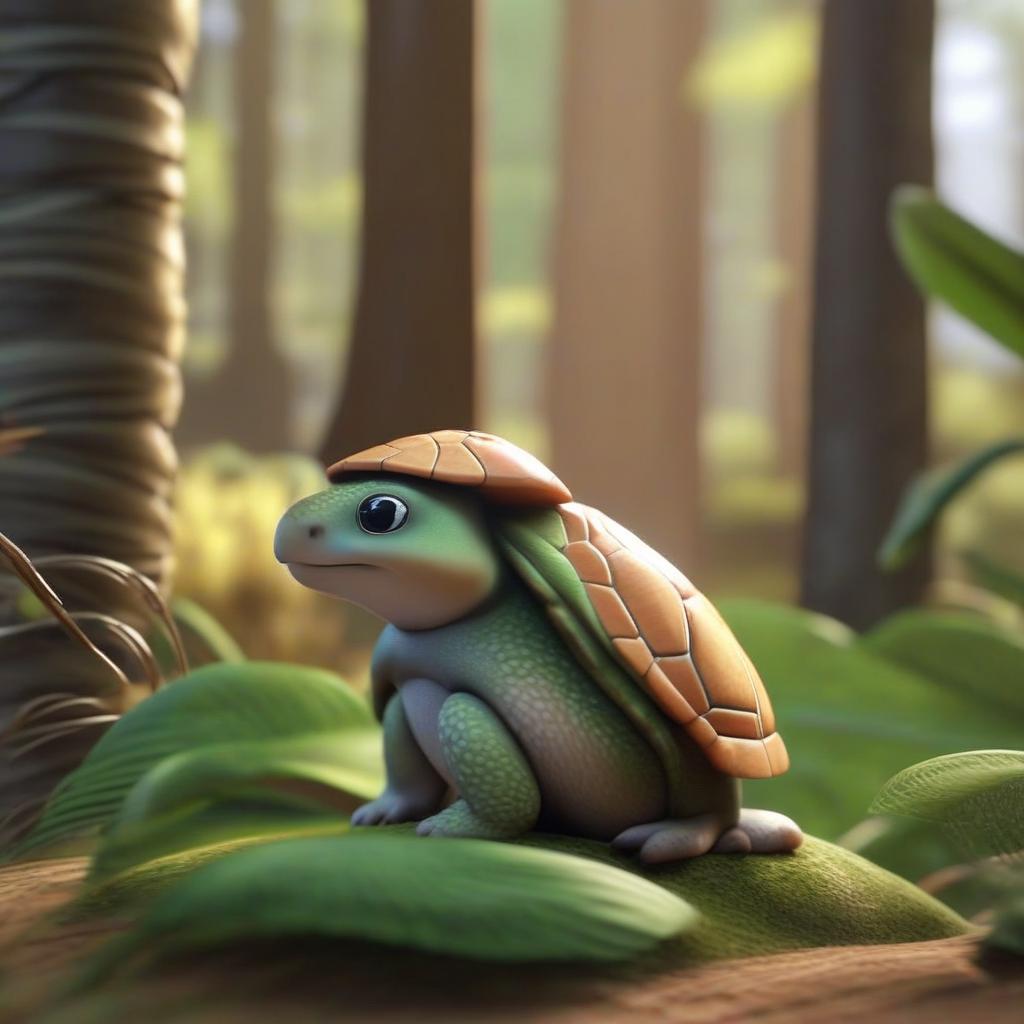}} &
        {\includegraphics[valign=c, width=\ww]{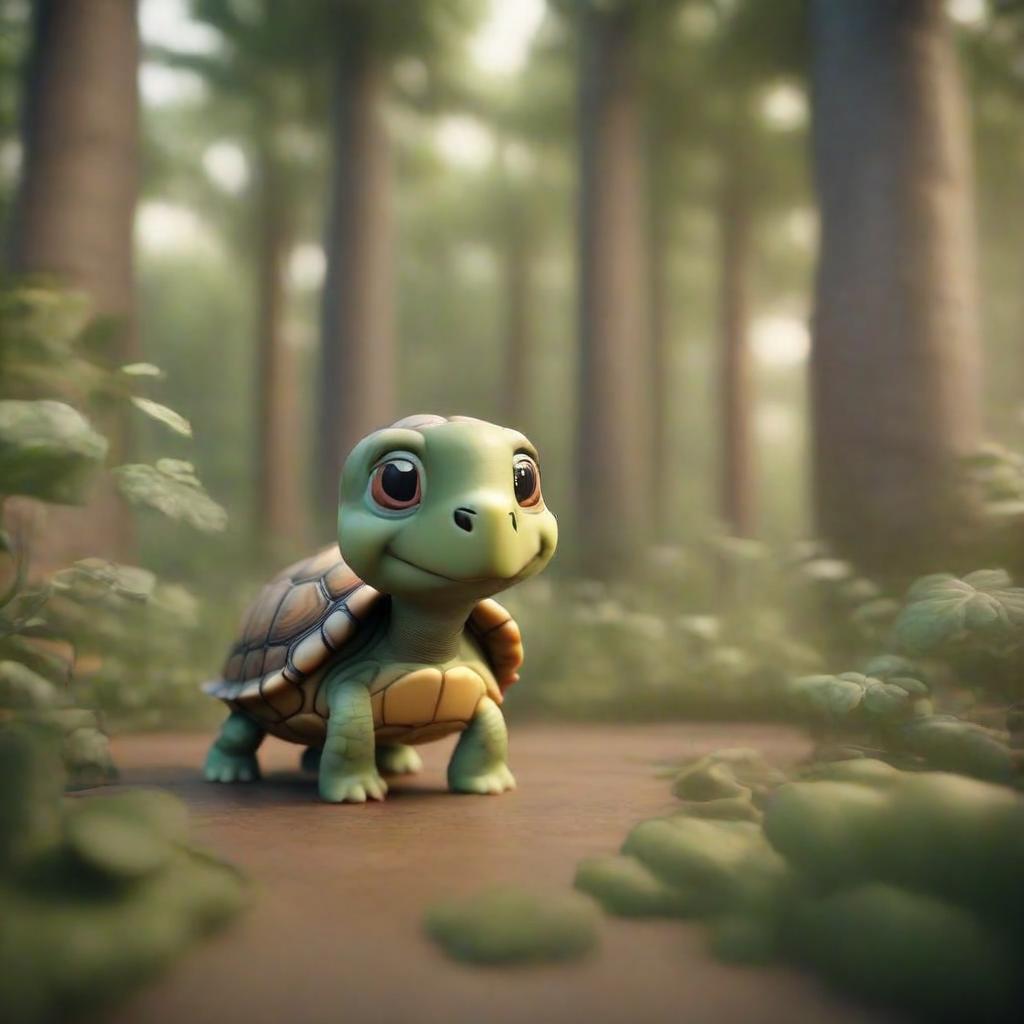}}
        \\
        \\

        \multicolumn{7}{c}{\textit{``a rendering of a cute turtle, cozy lighting ...''}}

    \end{tabular}
    
    \caption{\textbf{Additional qualitative comparisons to baselines.} 
    We compared our method against several baselines: TI \cite{Gal2022AnII}, BLIP-diffusion \cite{Li2023BLIPDiffusionPS} and IP-adapter \cite{Ye2023IPAdapterTC} are able to correspond to the target prompt but fail to produce consistent results. LoRA DB \cite{lora_diffusion} is able to achieve consistency, but it does not always follow to the prompt, in addition, the generate character is being generated in the same fixed pose. ELITE \cite{Wei2023ELITEEV} struggles with following the prompt and also tends to generate deformed characters. On the other hand, our method is able to follow the prompt, and generate consistent characters in different poses and viewing angles.}
    \label{fig:additional_qualitative_comparison_additional}
\end{figure*}

%% file: figures/automatic_qualitative_comparison/fig_ablations.tex
\begin{figure*}[t]
    \centering
    \setlength{\tabcolsep}{2.5pt}
    \renewcommand{\arraystretch}{0.2}
    \setlength{\ww}{0.345\columnwidth}
    \begin{tabular}{cccccc}
        &
        \textbf{Ours single iter.} &
        \textbf{Ours w/o clust.} &
        \textbf{Ours w/o LoRA} &
        \textbf{Ours w reinit.} &
        \textbf{Ours}
        \\

        \rotatebox[origin=c]{90}{\phantom{a}}
        \rotatebox[origin=c]{90}{\textit{``drinking a beer''}} &
        {\includegraphics[valign=c, width=\ww]{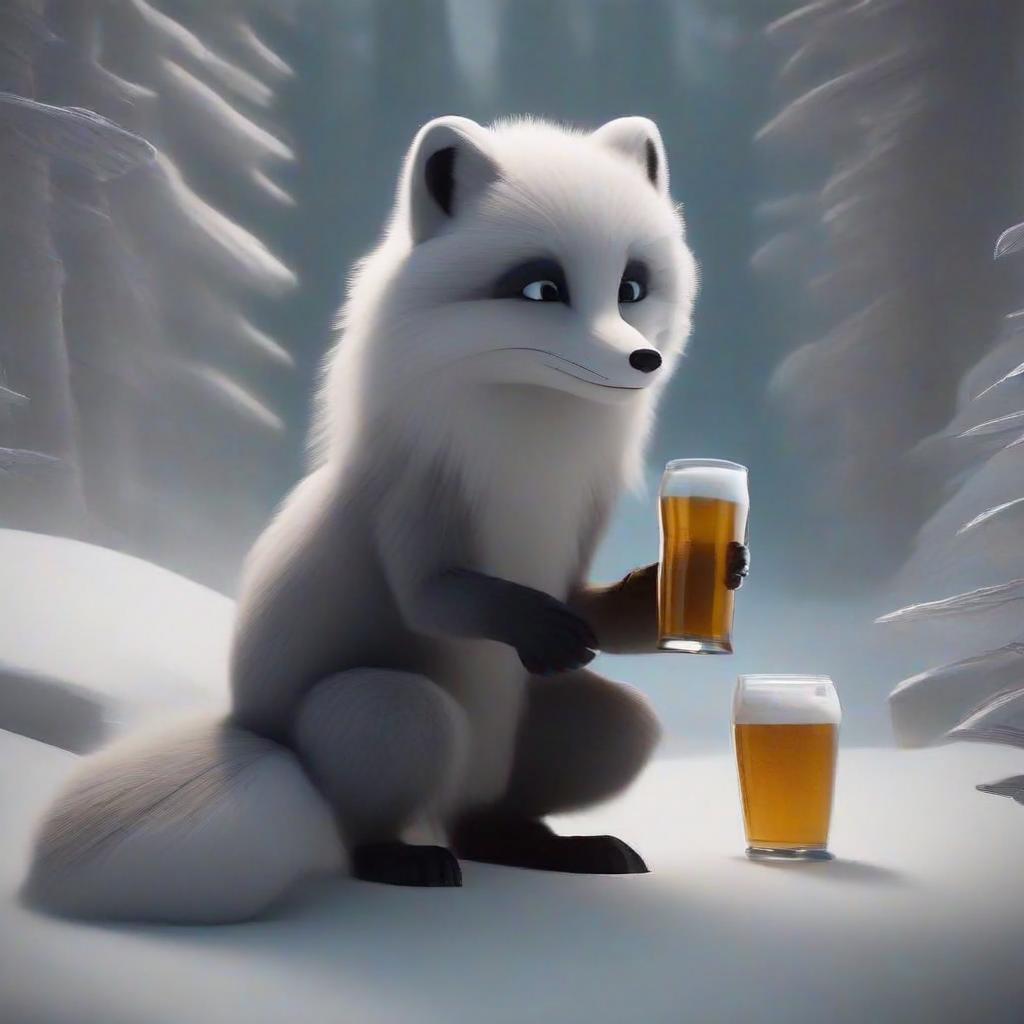}} &
        {\includegraphics[valign=c, width=\ww]{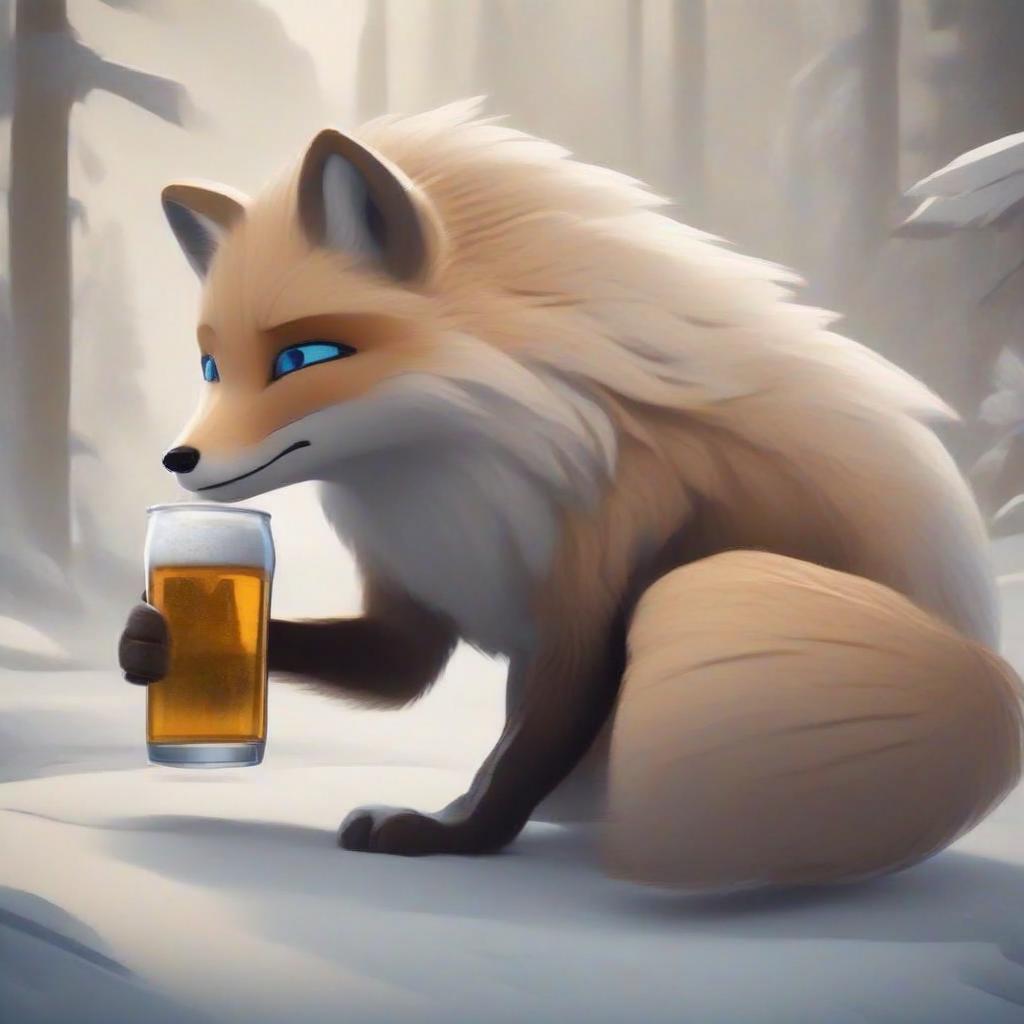}} &
        {\includegraphics[valign=c, width=\ww]{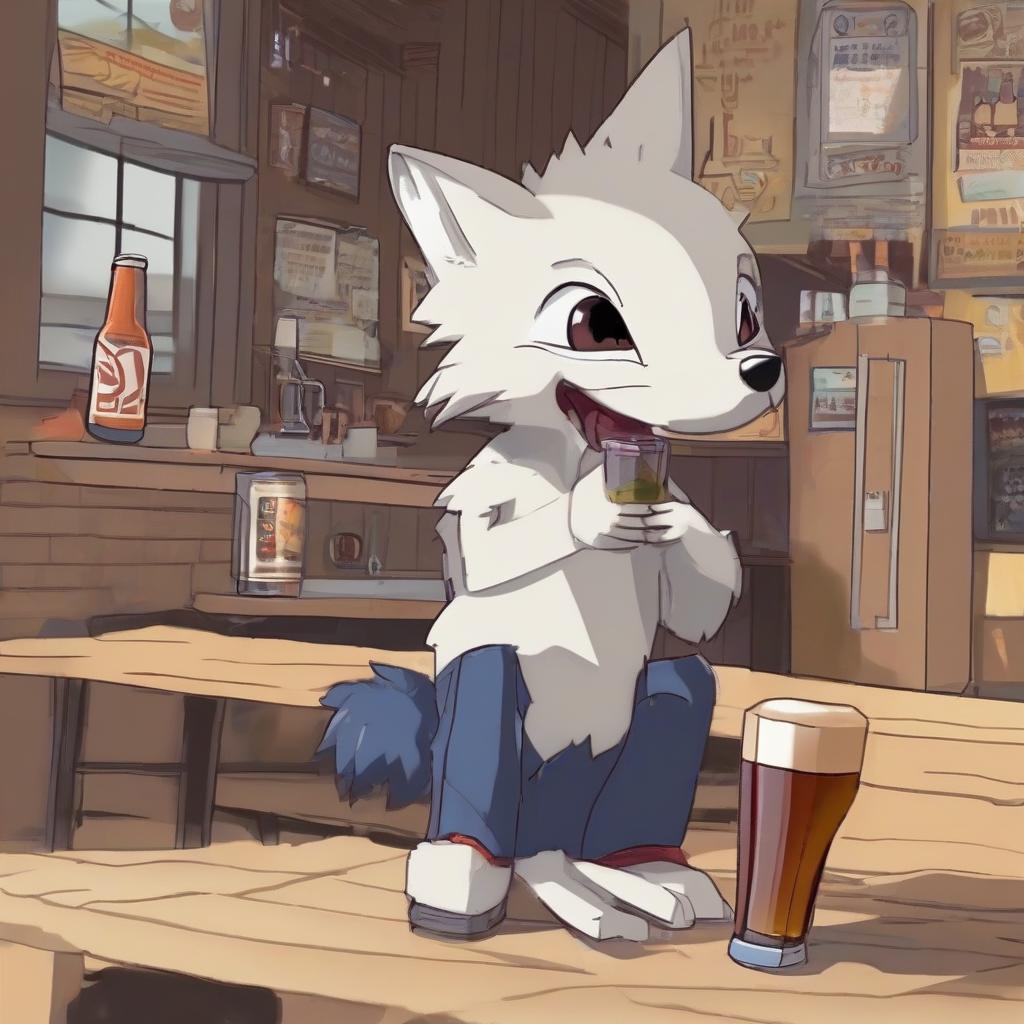}} &
        {\includegraphics[valign=c, width=\ww]{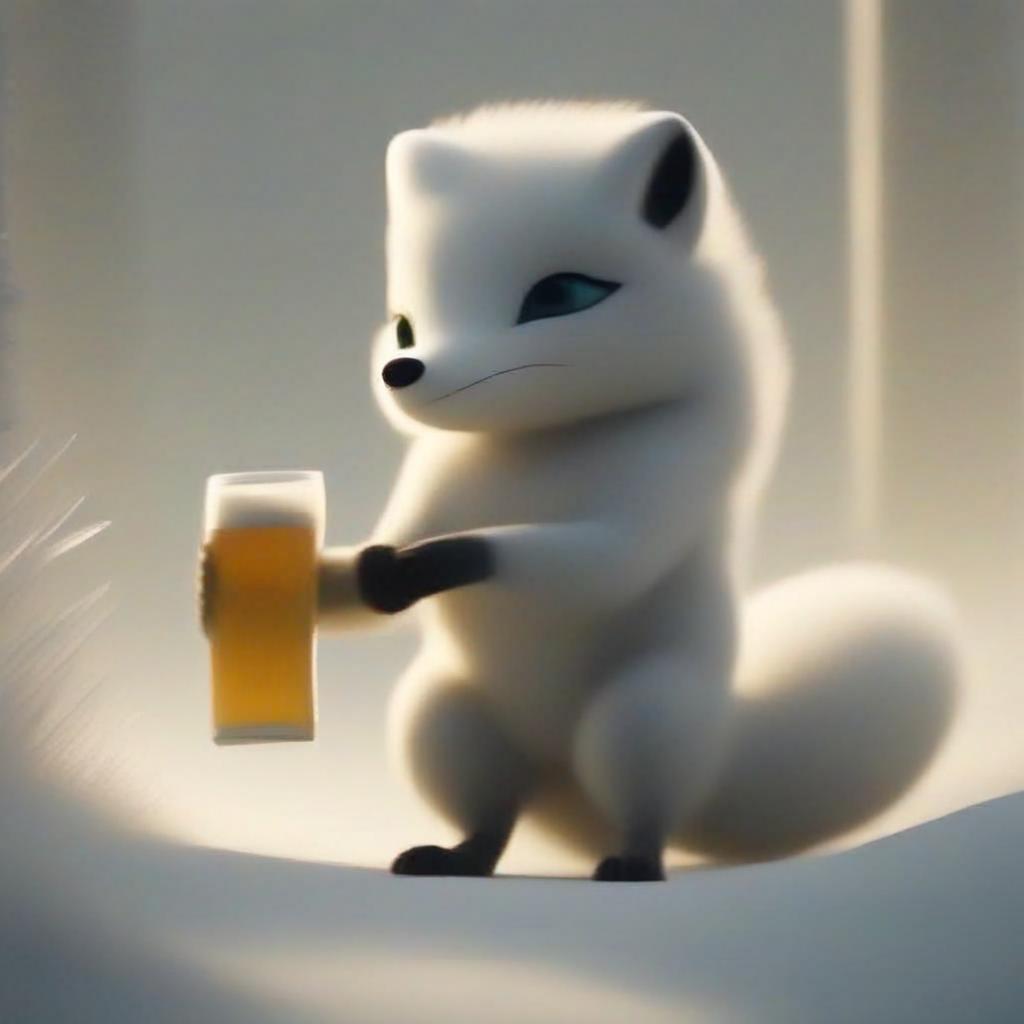}} &
        {\includegraphics[valign=c, width=\ww]{figures/automatic_qualitative_comparison/assets/fox/ours/beer_0.jpg}}
        \\
        \\

        \rotatebox[origin=c]{90}{\textit{``with a city in}}
        \rotatebox[origin=c]{90}{\textit{the background''}} &
        {\includegraphics[valign=c, width=\ww]{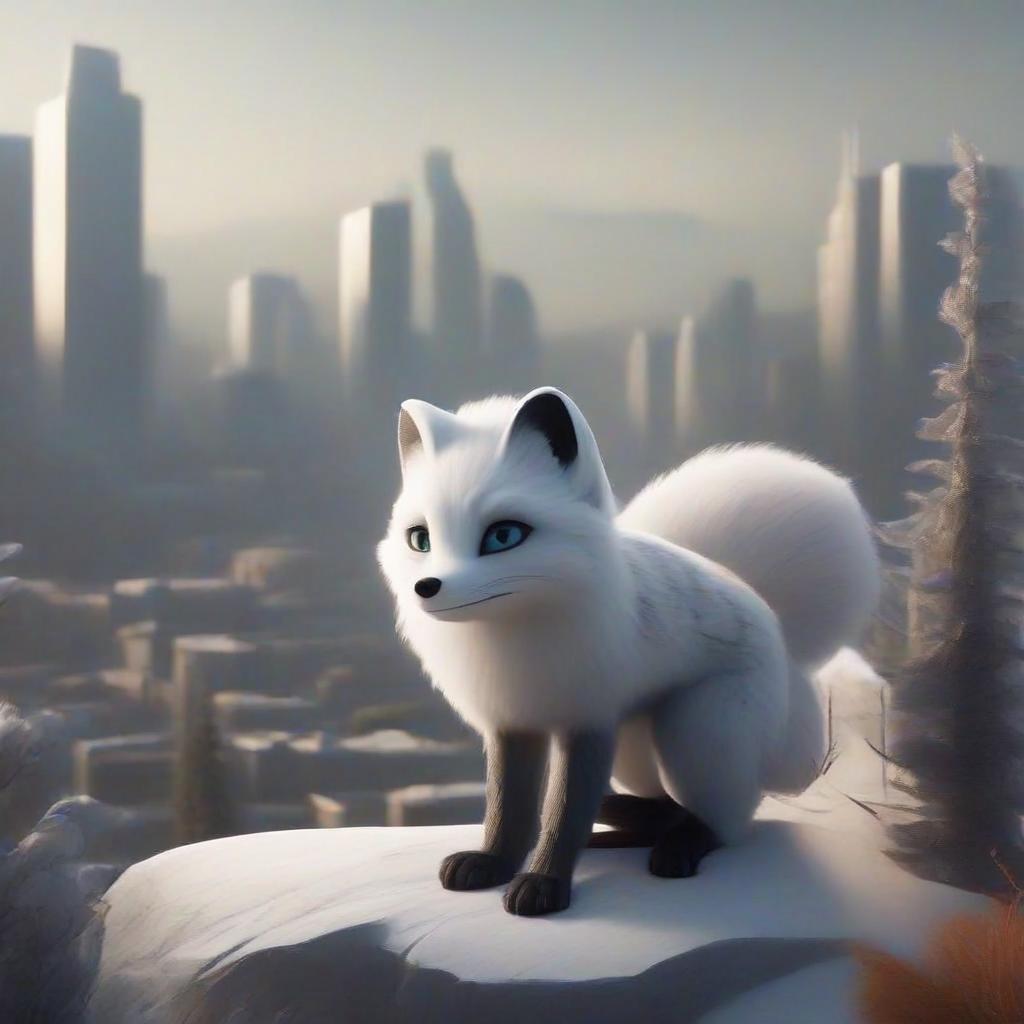}} &
        {\includegraphics[valign=c, width=\ww]{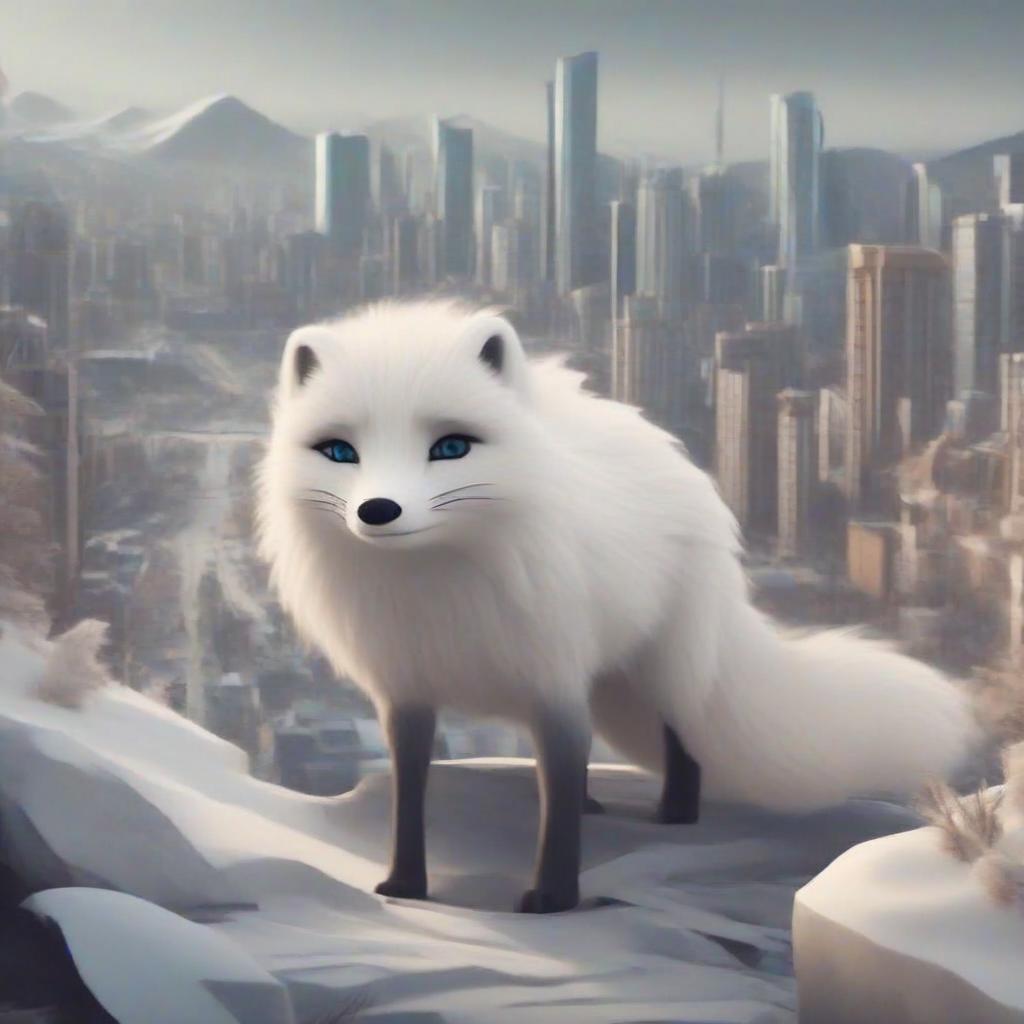}} &
        {\includegraphics[valign=c, width=\ww]{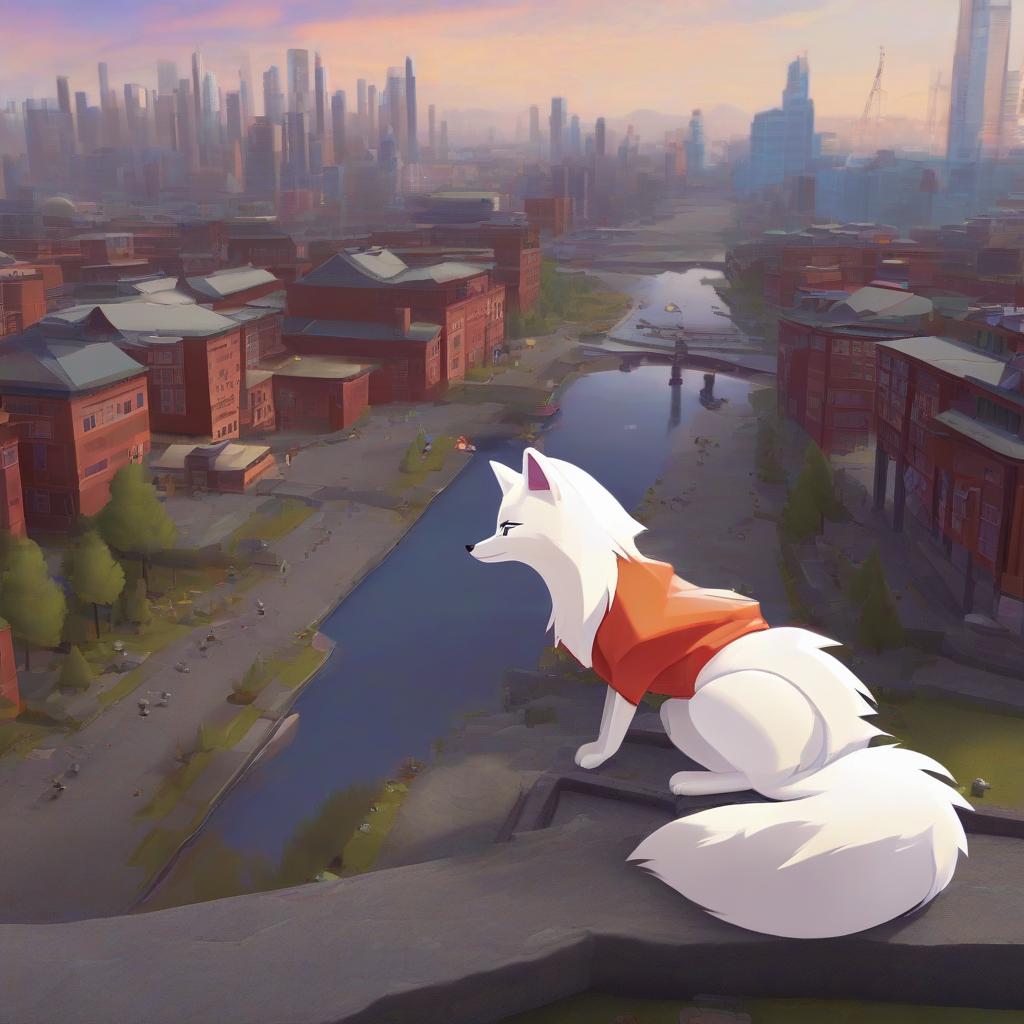}} &
        {\includegraphics[valign=c, width=\ww]{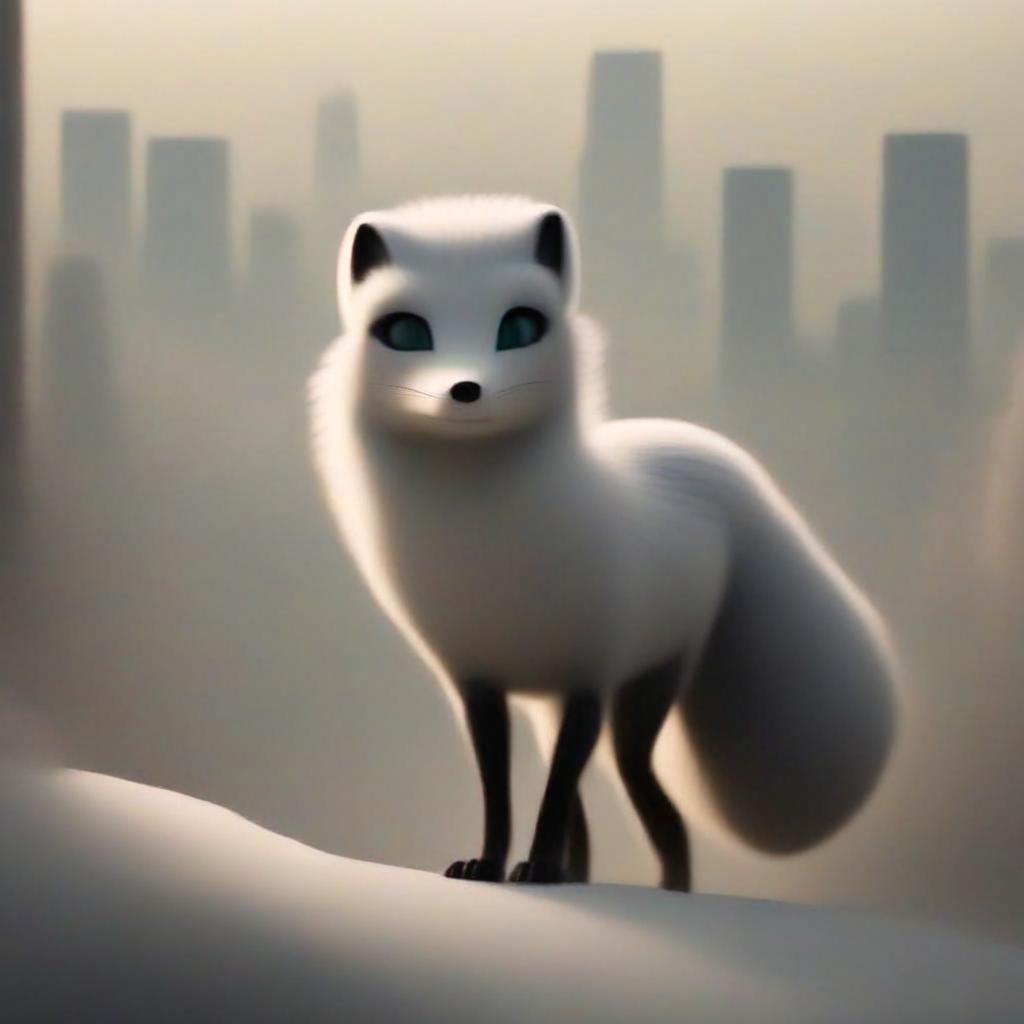}} &
        {\includegraphics[valign=c, width=\ww]{figures/automatic_qualitative_comparison/assets/fox/ours/city_0.jpg}}
        \\
        \\

        \\
        \\
        \multicolumn{6}{c}{\textit{``a 2D animation of captivating Arctic fox with fluffy fur, bright}}
        \\
        \multicolumn{6}{c}{\textit{and nimble movements, bringing the magic of the icy wilderness to animated life''}}
        \\
        \\
        \midrule

        \\
        \rotatebox[origin=c]{90}{\phantom{a}}
        \rotatebox[origin=c]{90}{\textit{``eating a burger''}} &
        {\includegraphics[valign=c, width=\ww]{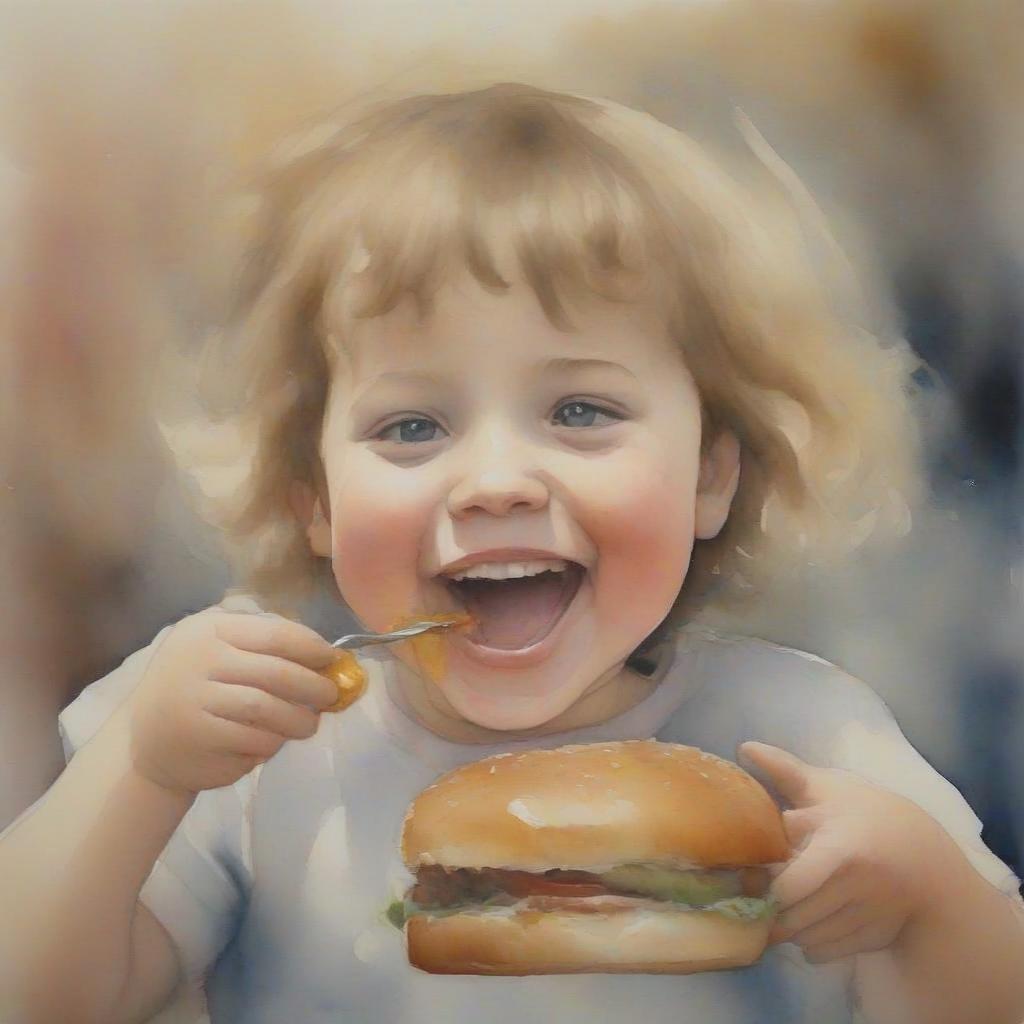}} &
        {\includegraphics[valign=c, width=\ww]{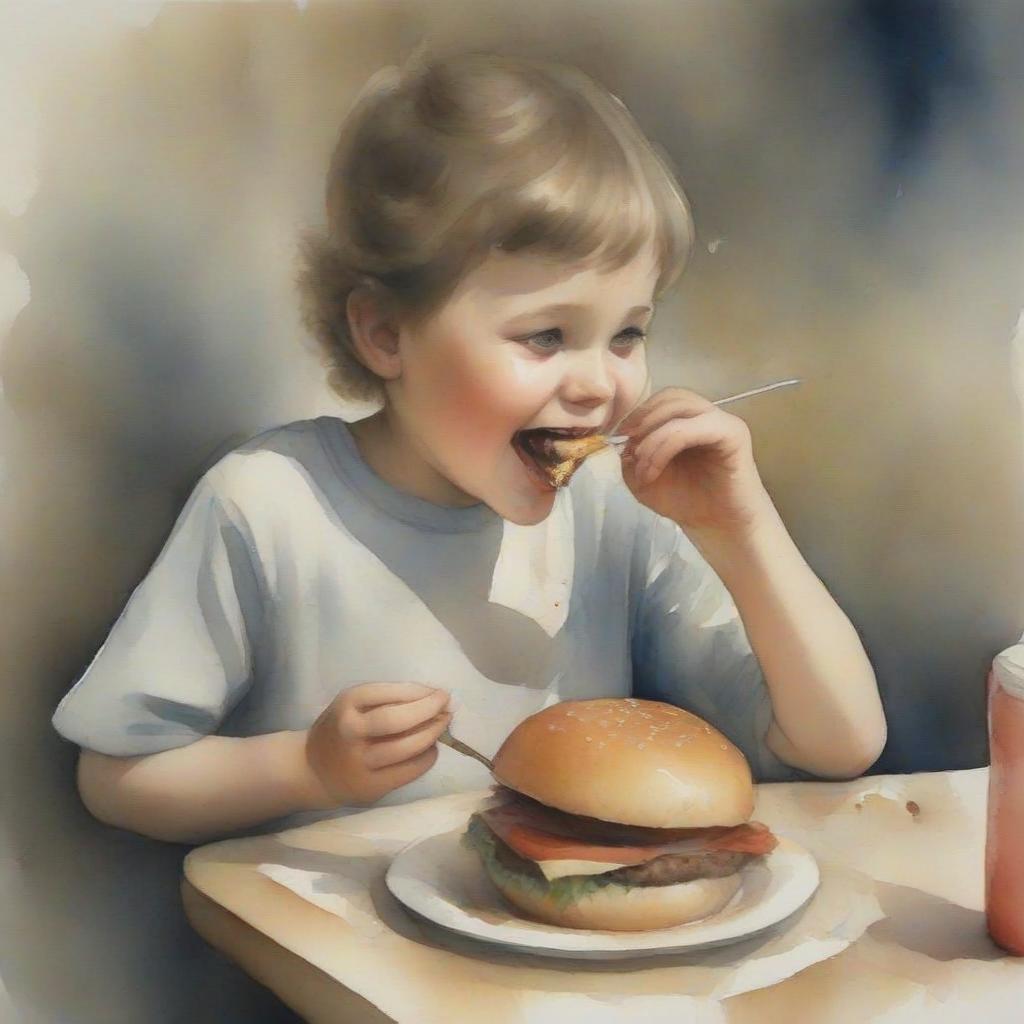}} &
        {\includegraphics[valign=c, width=\ww]{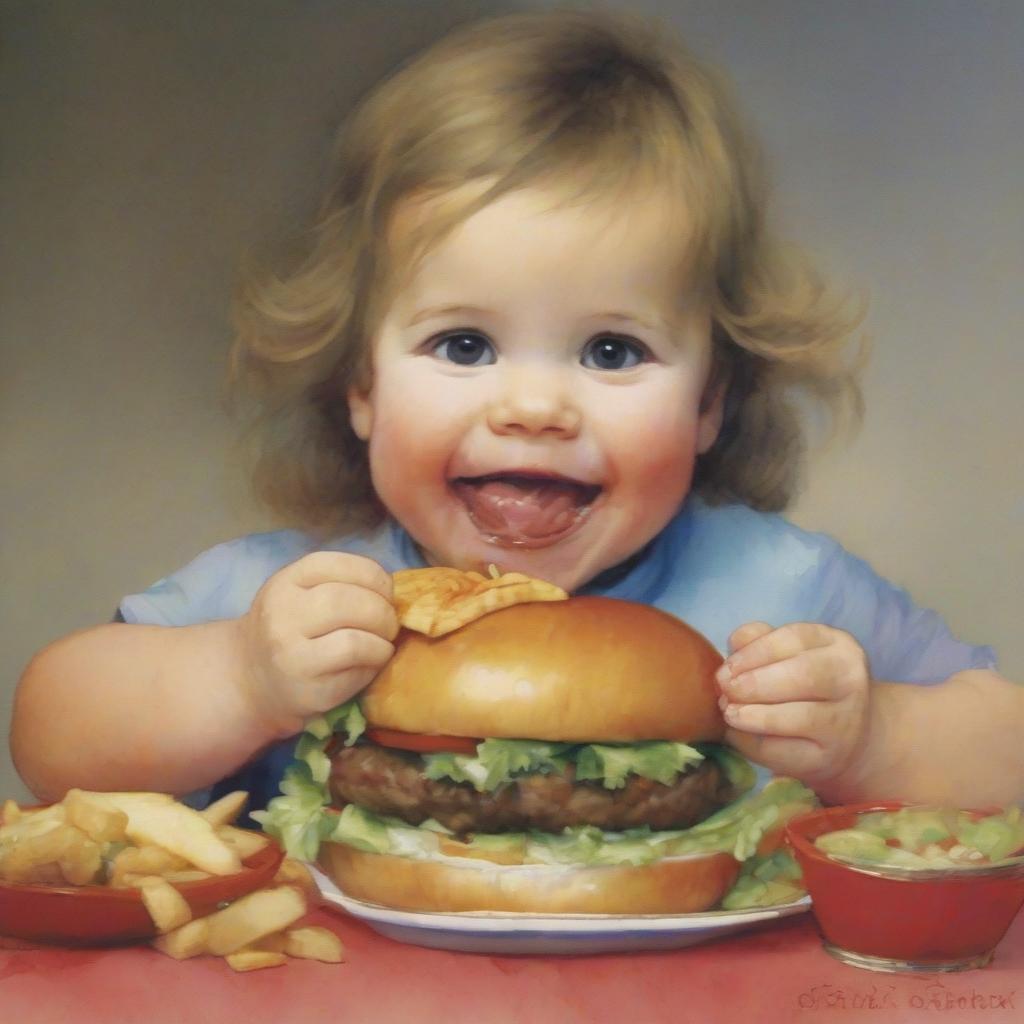}} &
        {\includegraphics[valign=c, width=\ww]{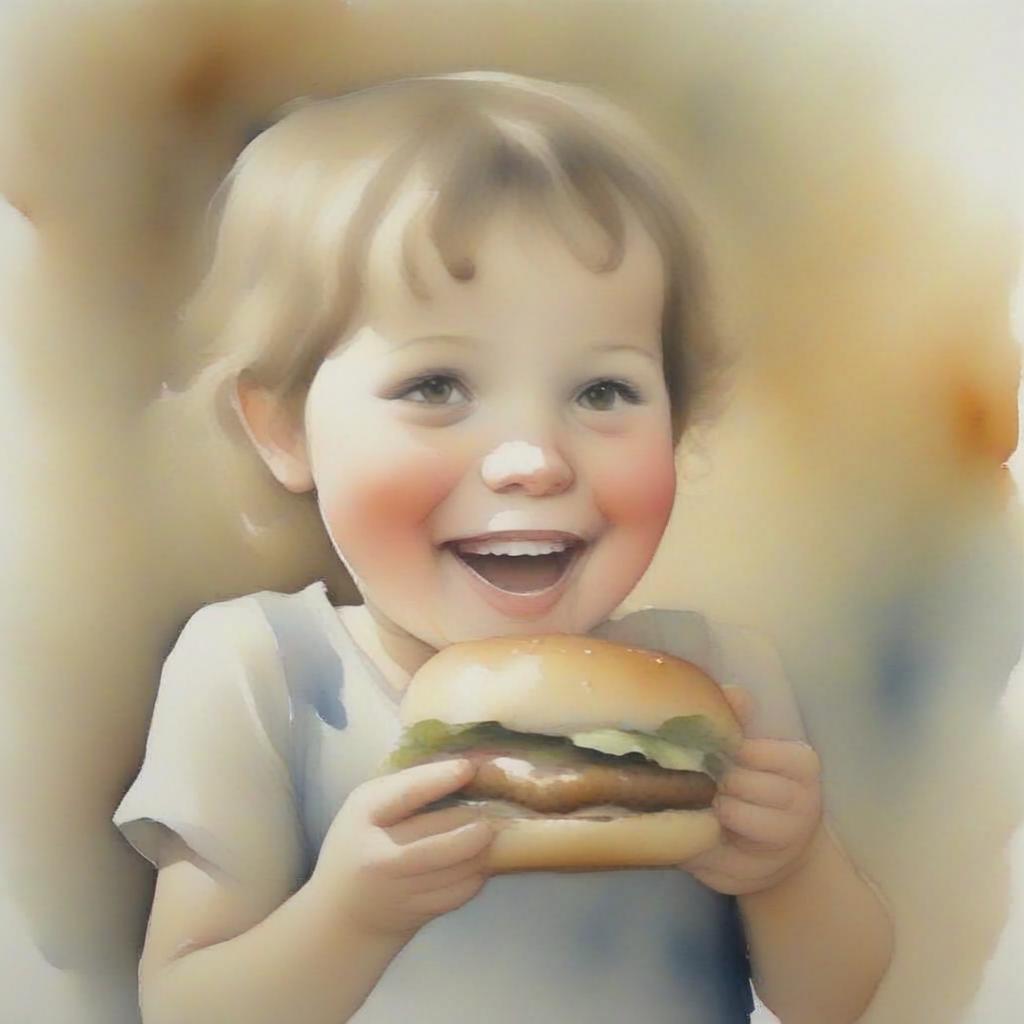}} &
        {\includegraphics[valign=c, width=\ww]{figures/automatic_qualitative_comparison/assets/child/ours/burger.jpg}}
        \\
        \\

        \rotatebox[origin=c]{90}{\textit{``wearing a}}
        \rotatebox[origin=c]{90}{\textit{blue hat''}} &
        {\includegraphics[valign=c, width=\ww]{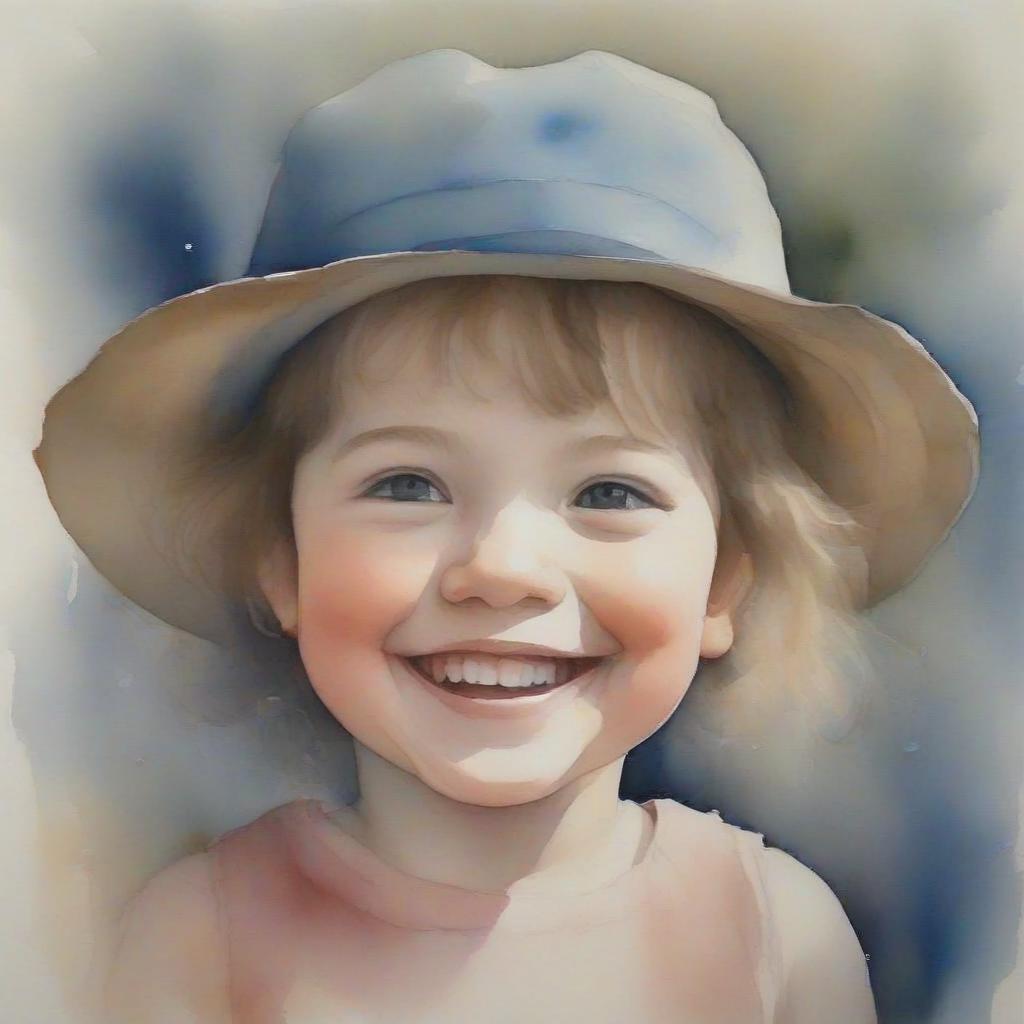}} &
        {\includegraphics[valign=c, width=\ww]{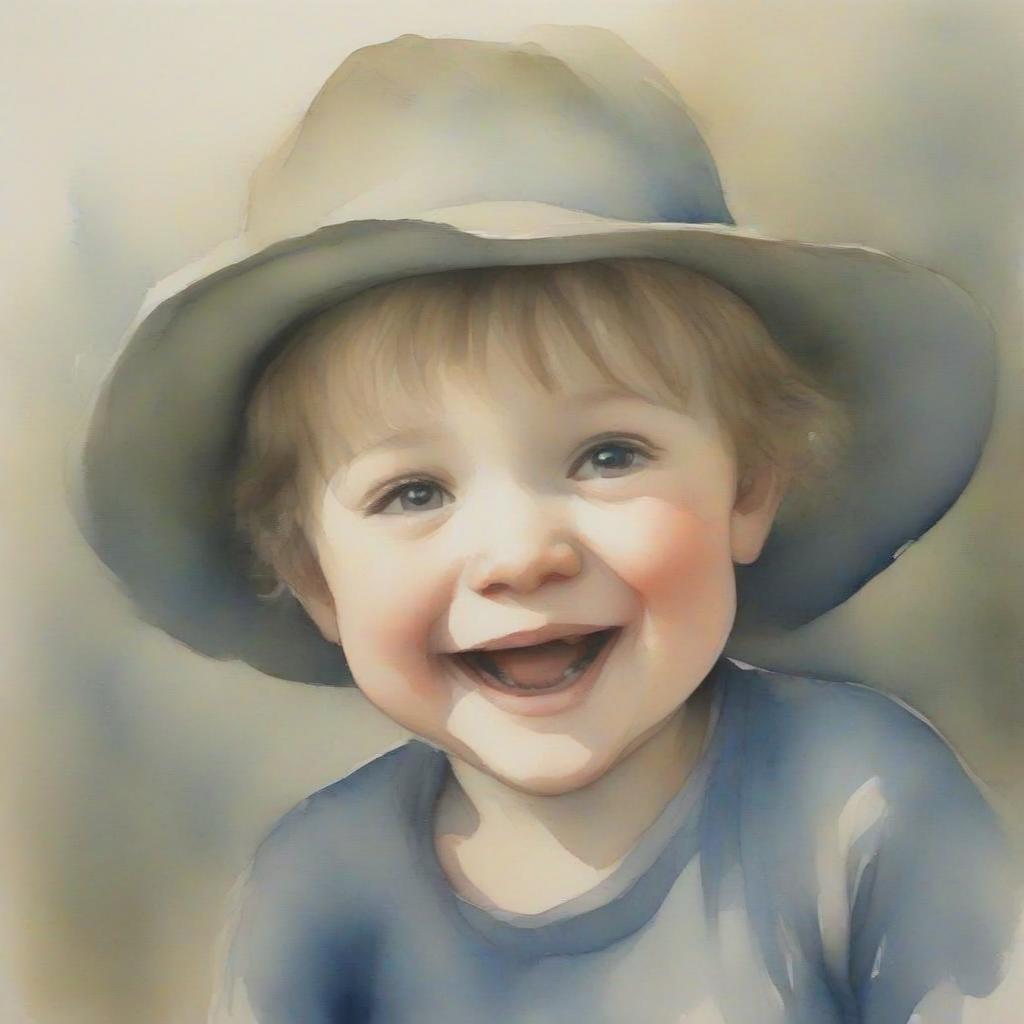}} &
        {\includegraphics[valign=c, width=\ww]{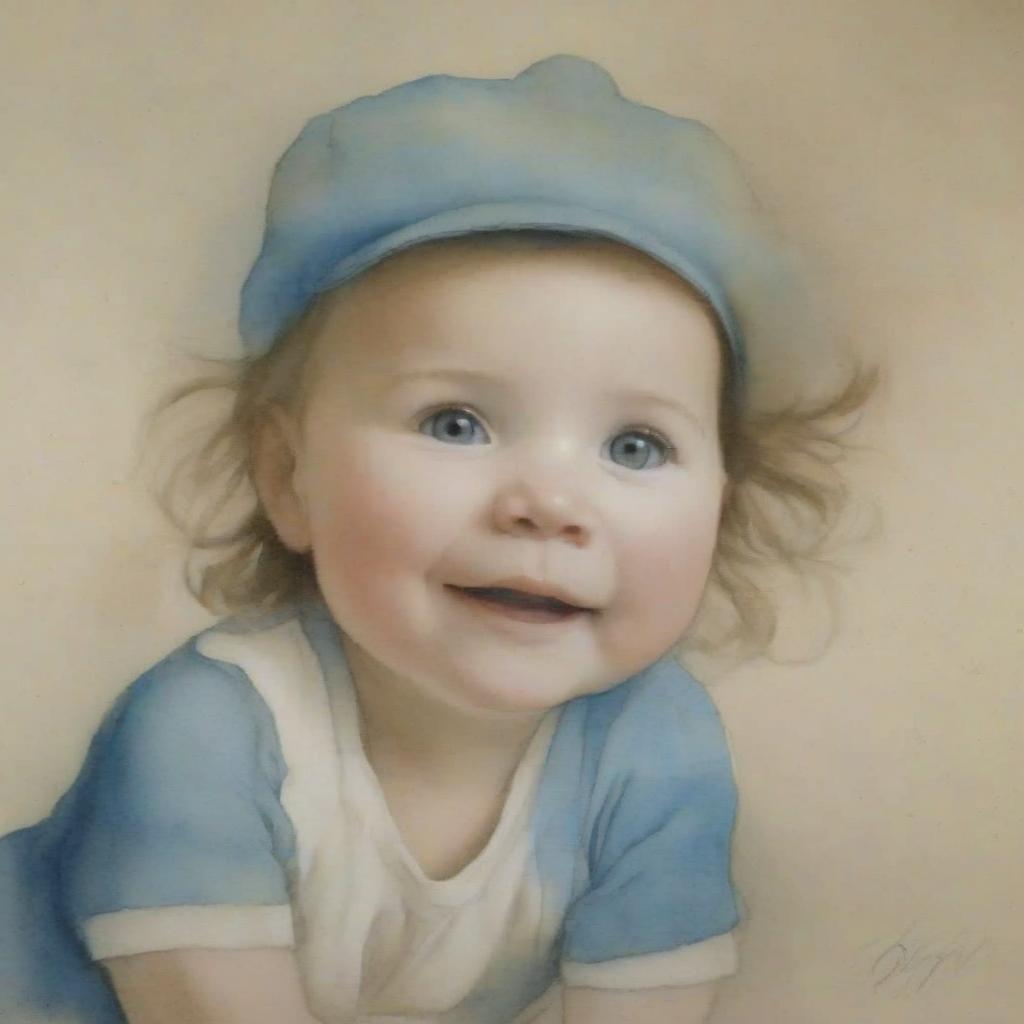}} &
        {\includegraphics[valign=c, width=\ww]{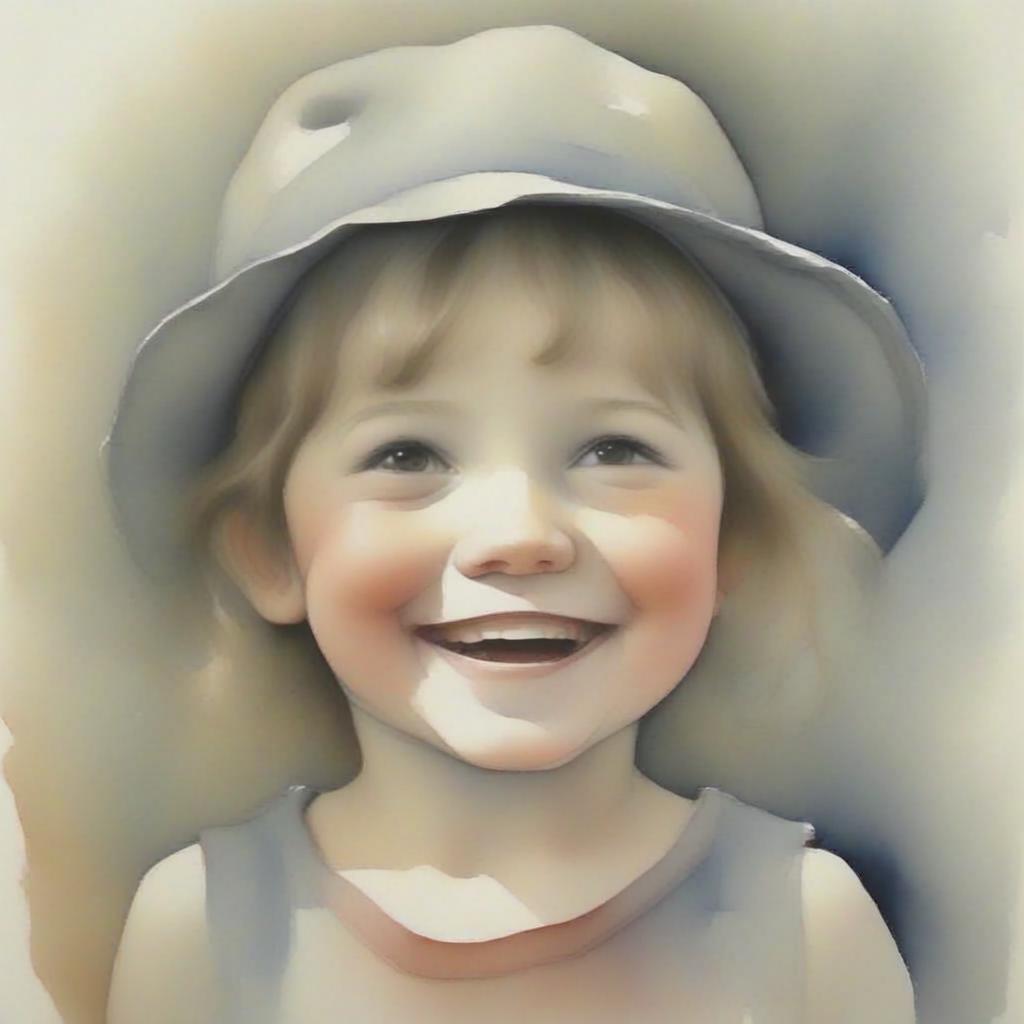}} &
        {\includegraphics[valign=c, width=\ww]{figures/automatic_qualitative_comparison/assets/child/ours/hat.jpg}}
        \\
        \\

        \\
        \\
        \multicolumn{6}{c}{\textit{``a watercolor portrayal of a joyful child, radiating innocence}}
        \\
        \multicolumn{6}{c}{\textit{and wonder with rosy cheeks and a genuine, wide-eyed smile''}}
        \\
        \\
        \midrule

        \\
        \rotatebox[origin=c]{90}{\textit{``near the}}
        \rotatebox[origin=c]{90}{\textit{Statue of Liberty''}} &
        {\includegraphics[valign=c, width=\ww]{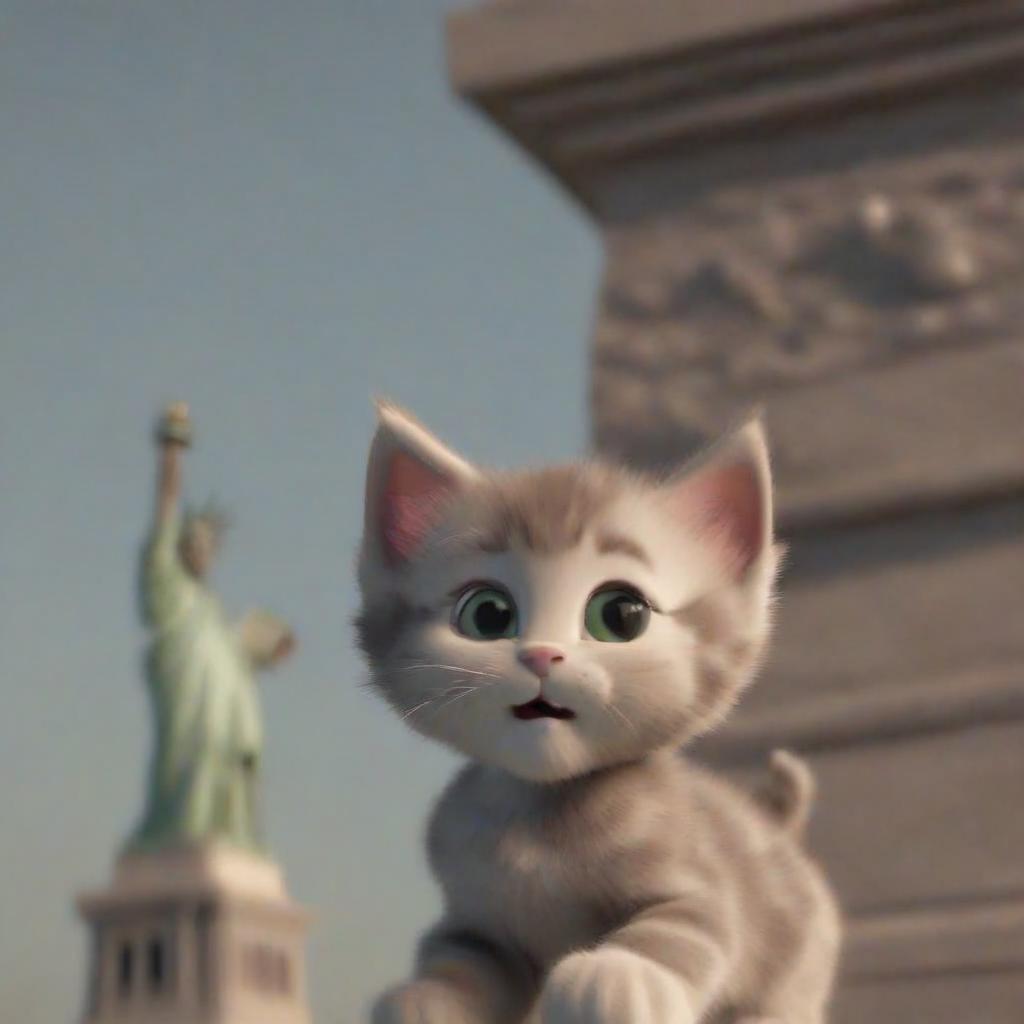}} &
        {\includegraphics[valign=c, width=\ww]{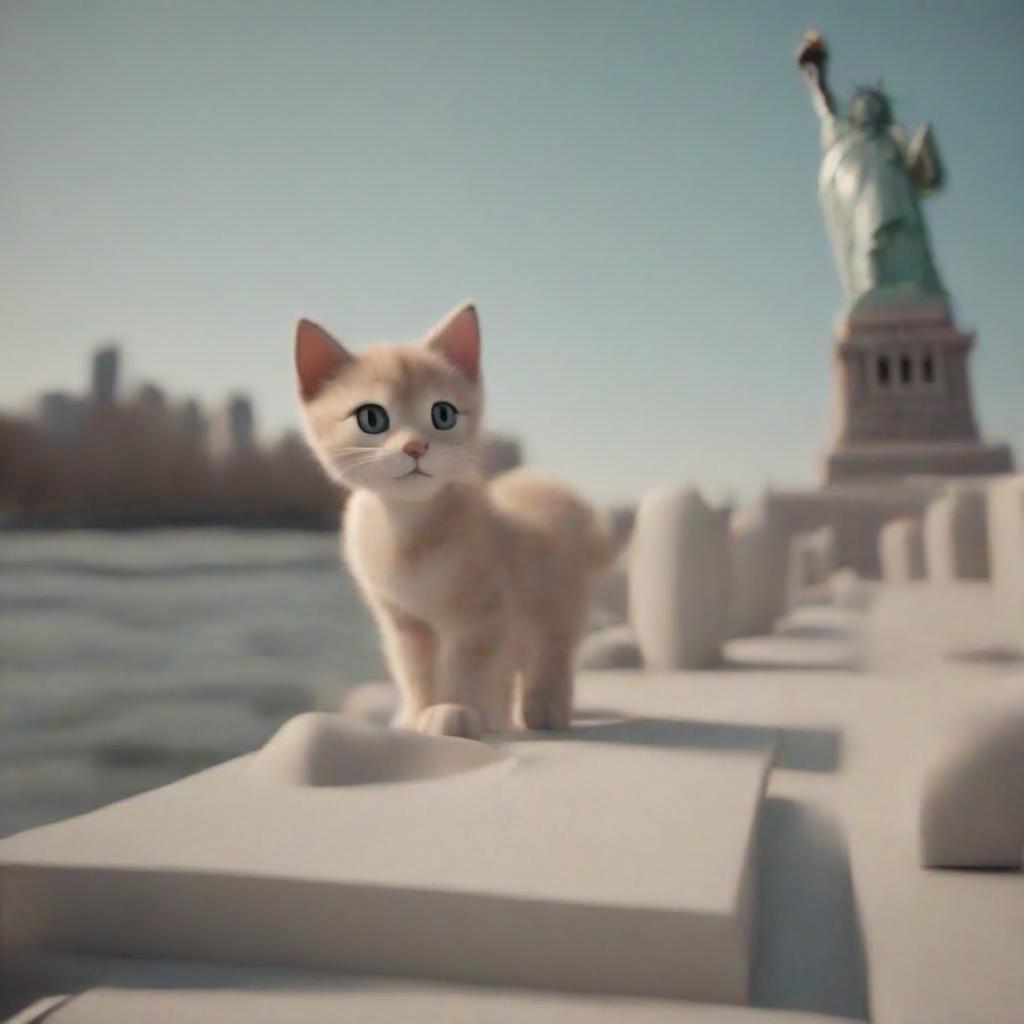}} &
        {\includegraphics[valign=c, width=\ww]{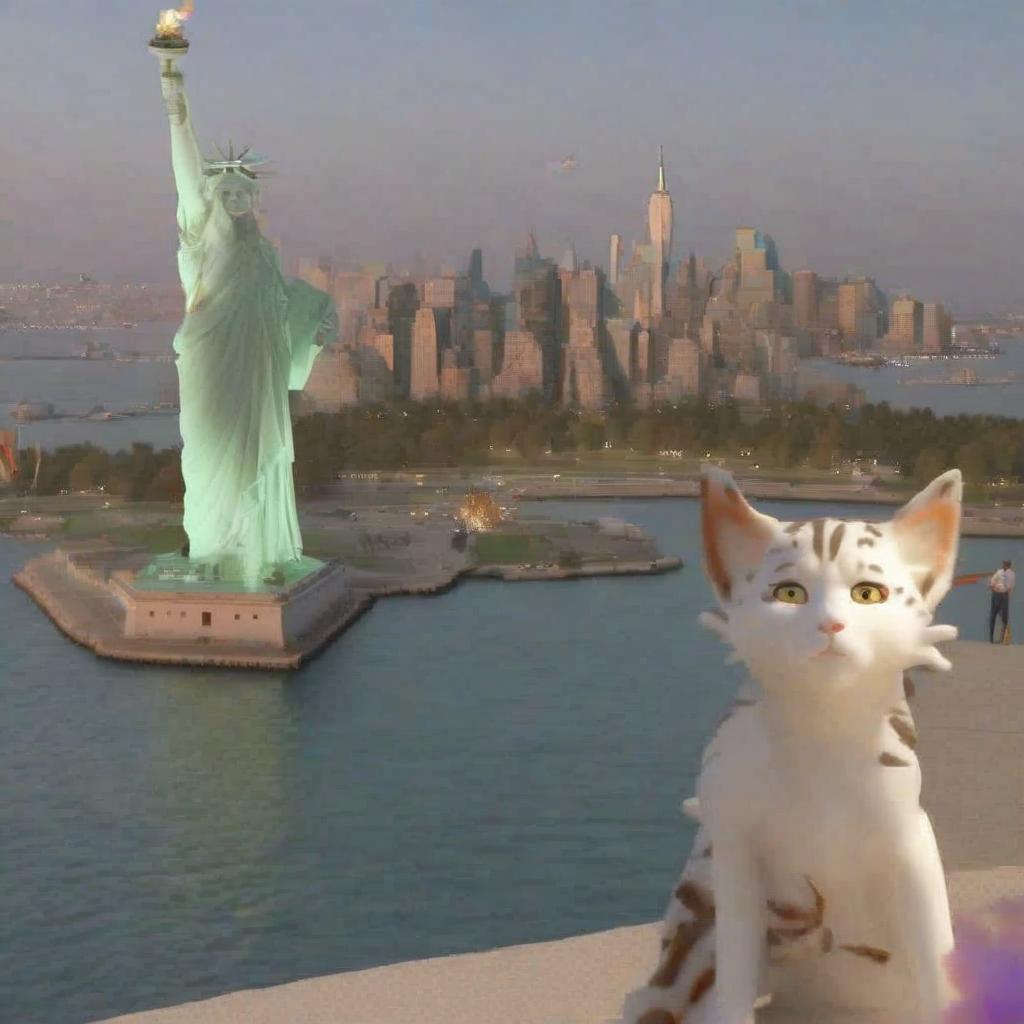}} &
        {\includegraphics[valign=c, width=\ww]{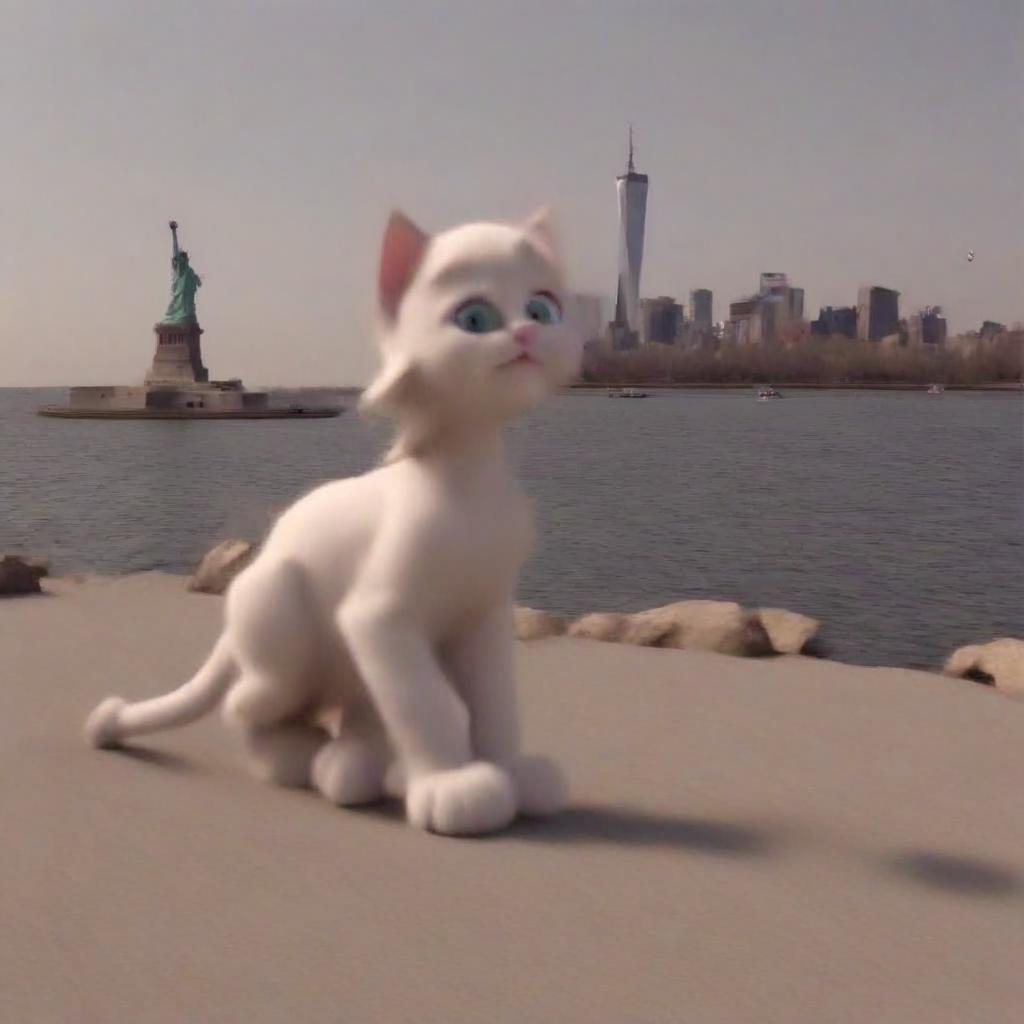}} &
        {\includegraphics[valign=c, width=\ww]{figures/automatic_qualitative_comparison/assets/cat/ours/liberty.jpg}}
        \\
        \\

        \rotatebox[origin=c]{90}{\textit{``as a police}}
        \rotatebox[origin=c]{90}{\textit{officer''}} &
        {\includegraphics[valign=c, width=\ww]{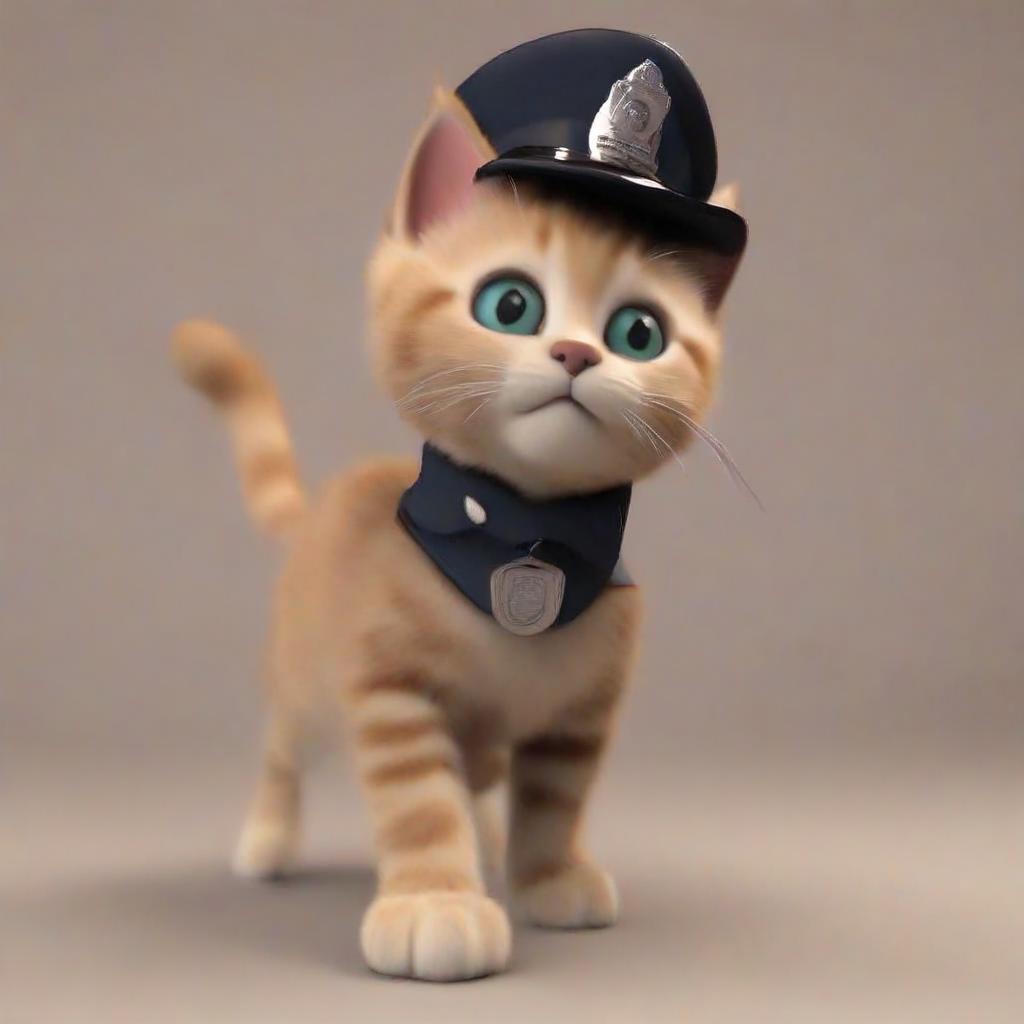}} &
        {\includegraphics[valign=c, width=\ww]{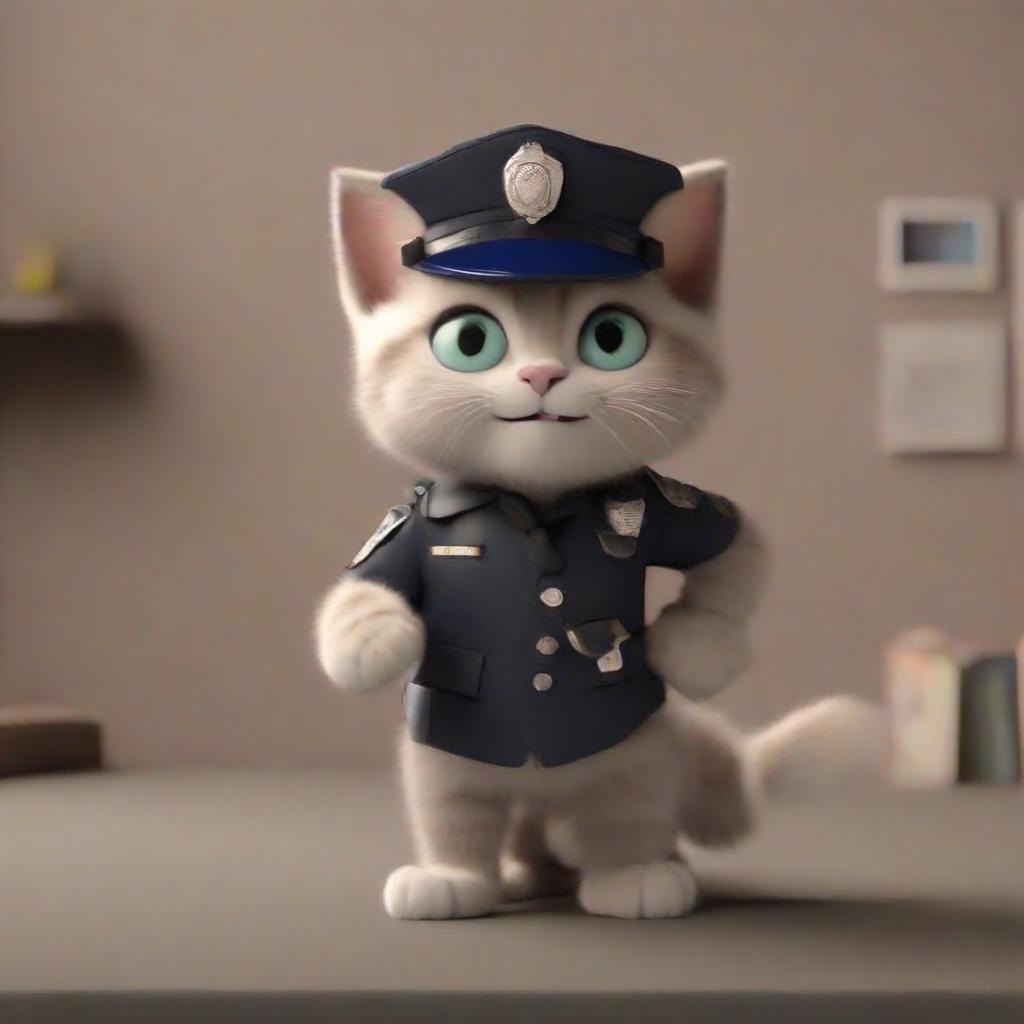}} &
        {\includegraphics[valign=c, width=\ww]{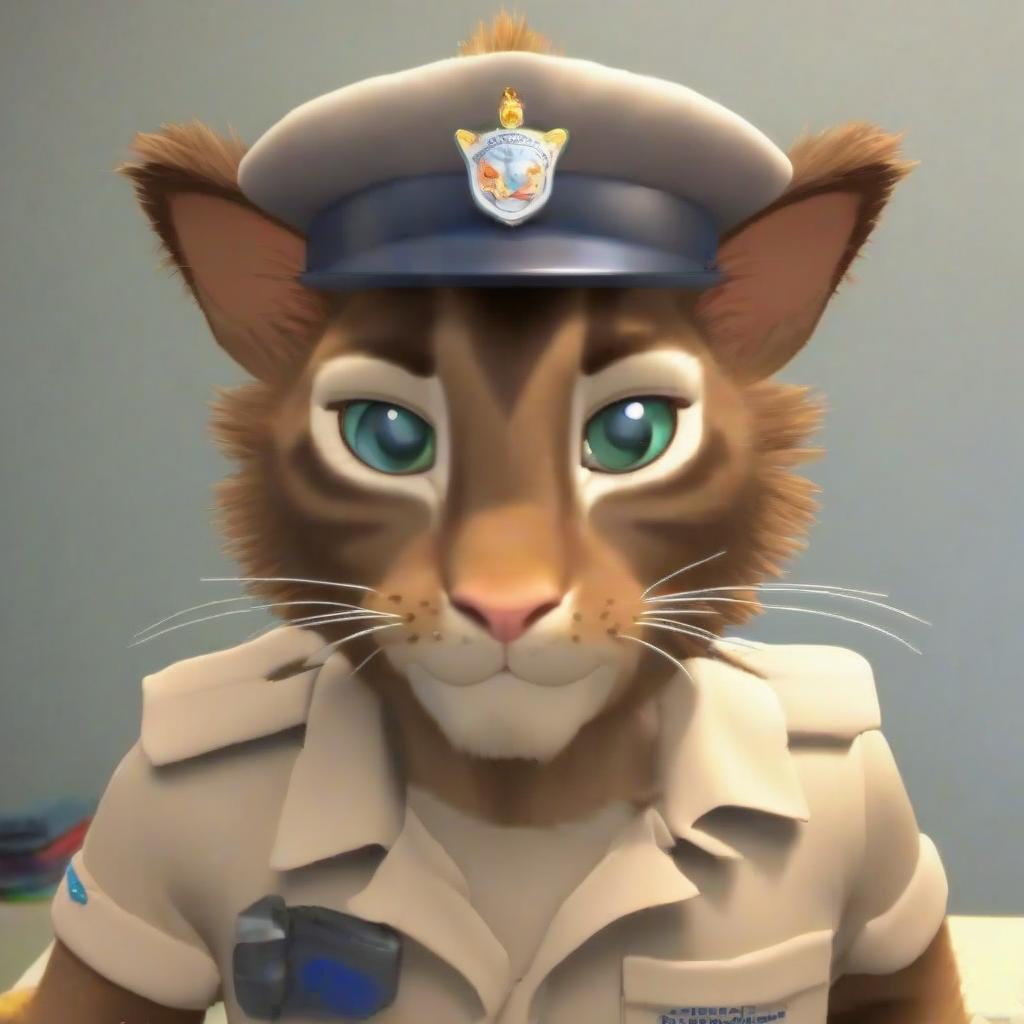}} &
        {\includegraphics[valign=c, width=\ww]{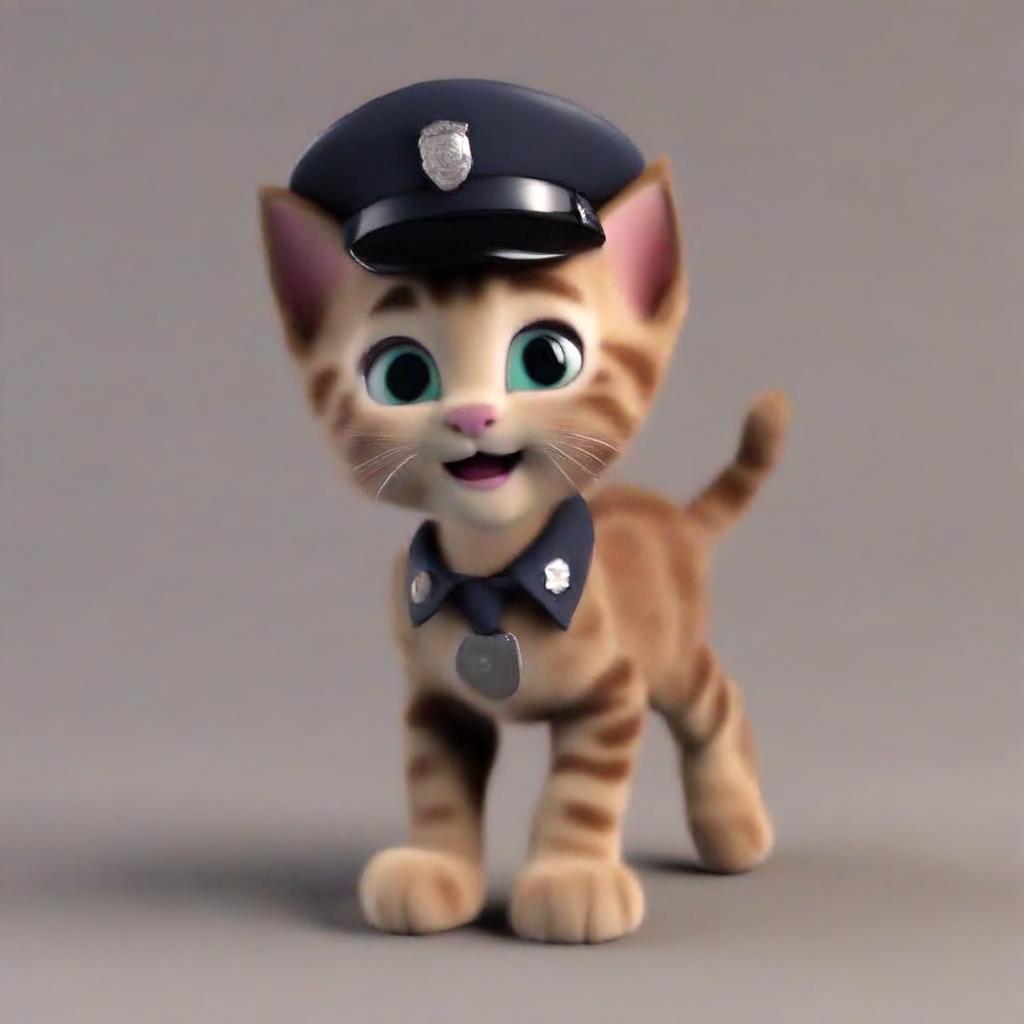}} &
        {\includegraphics[valign=c, width=\ww]{figures/automatic_qualitative_comparison/assets/cat/ours/police.jpg}}
        \\
        \\

        \\
        \\
        \multicolumn{6}{c}{\textit{``a 3D animation of a playful kitten, with bright eyes and}}
        \\
        \multicolumn{6}{c}{\textit{a mischievous expression, embodying youthful curiosity and joy''}}

    \end{tabular}
    
    \caption{\textbf{Qualitative comparison of ablations.} We ablated the following components of our method: using a single iteration, removing the clustering stage, removing the LoRA trainable parameters, using the same initial representation at every iteration. As can be seen, all these ablated cases struggle with preserving the character's consistency.}
    \label{fig:automatic_qualitative_ablations_comparison}
\end{figure*}

%% file: figures/general_objects/fig.tex
\begin{figure*}[t]
    \centering
    \setlength{\tabcolsep}{1.5pt}
    \renewcommand{\arraystretch}{0.7}
    \setlength{\ww}{0.4\columnwidth}
    \begin{tabular}{ccccc}

        \textit{``in the desert''} &
        \textit{``in Times Square''} &
        \textit{``near a lake''} &
        \textit{``near the Eiffel Tower''} &
        \textit{``near the Taj Mahal''}
        \\

        {\includegraphics[valign=c, width=\ww]{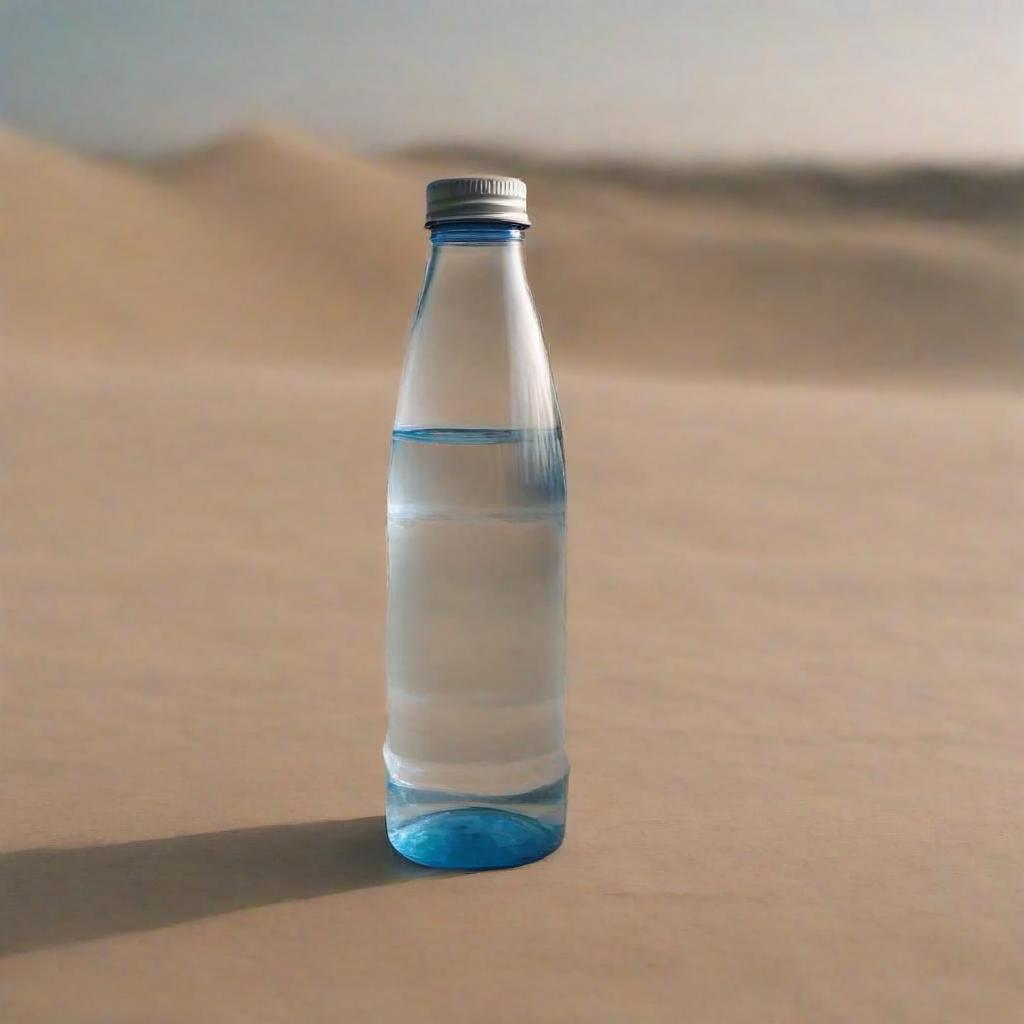}} &
        {\includegraphics[valign=c, width=\ww]{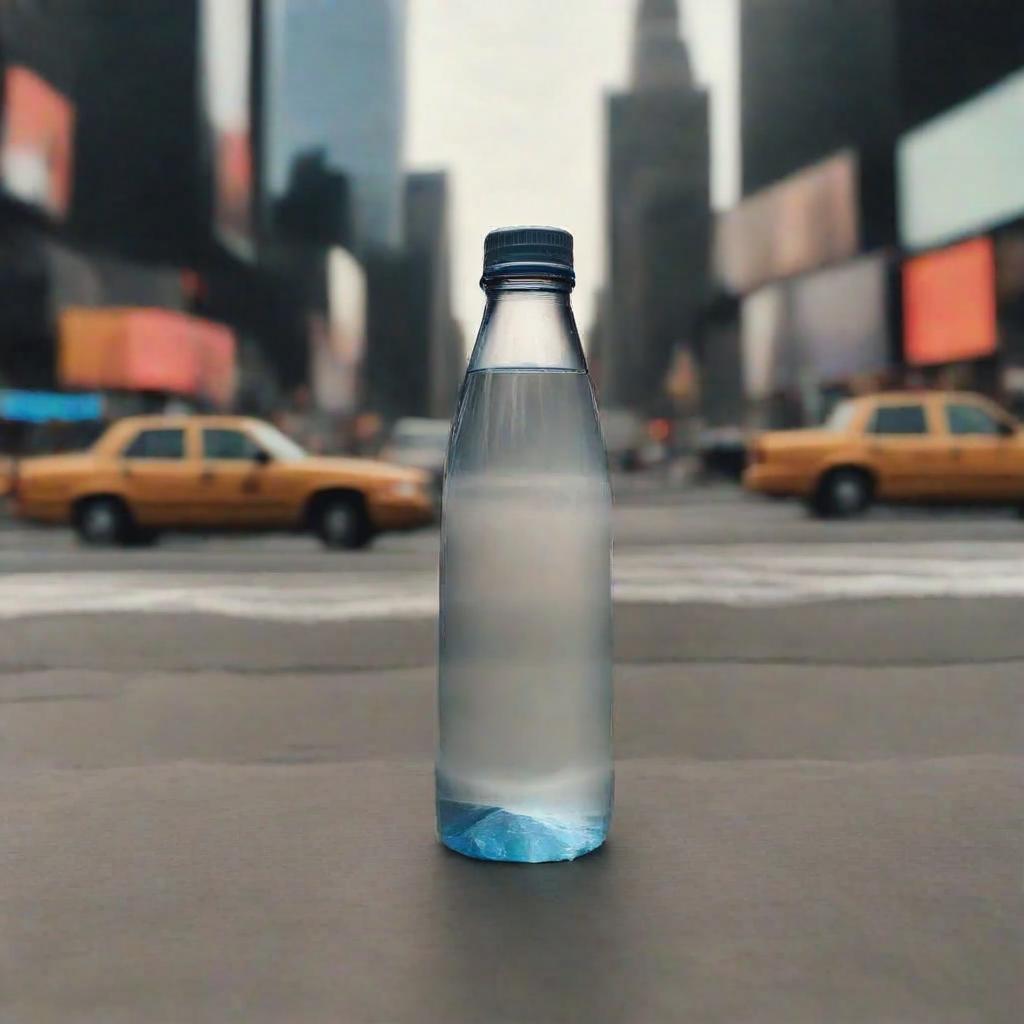}} &
        {\includegraphics[valign=c, width=\ww]{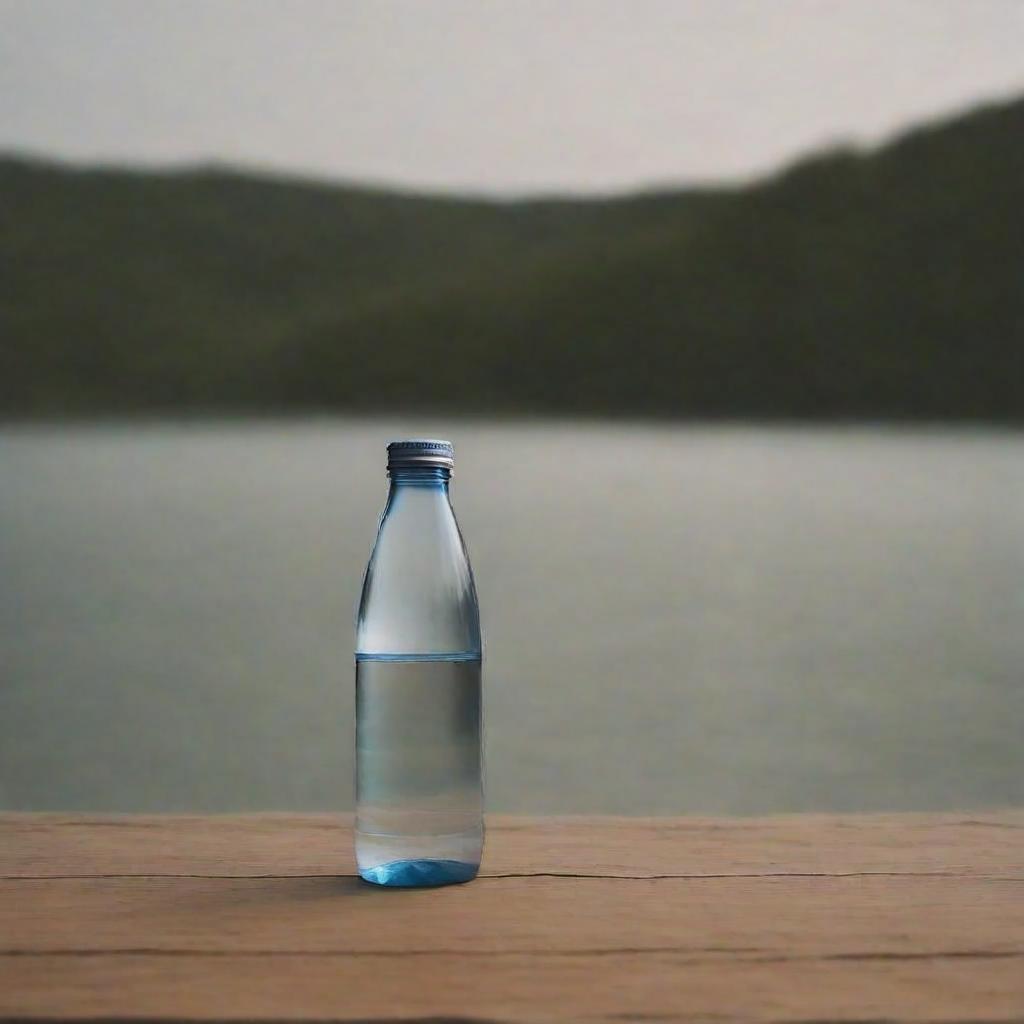}} &
        {\includegraphics[valign=c, width=\ww]{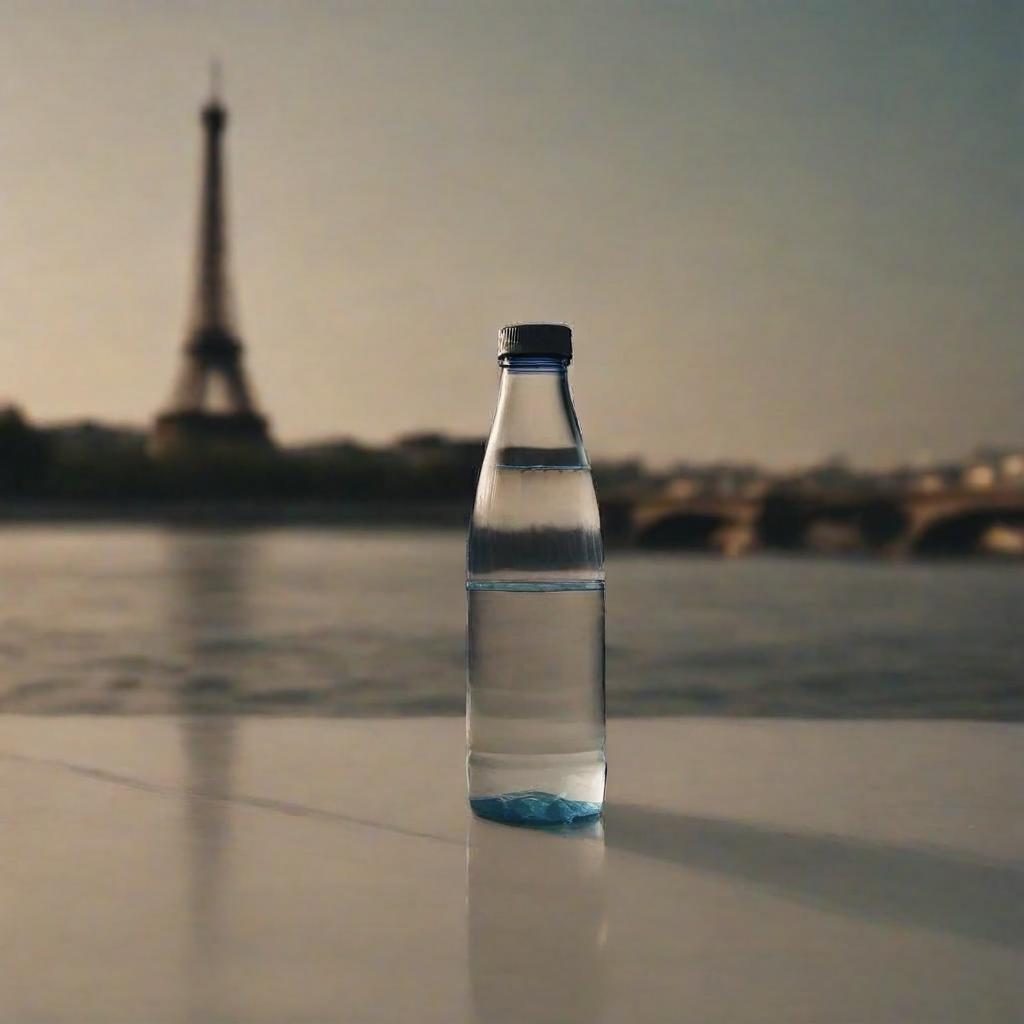}} &
        {\includegraphics[valign=c, width=\ww]{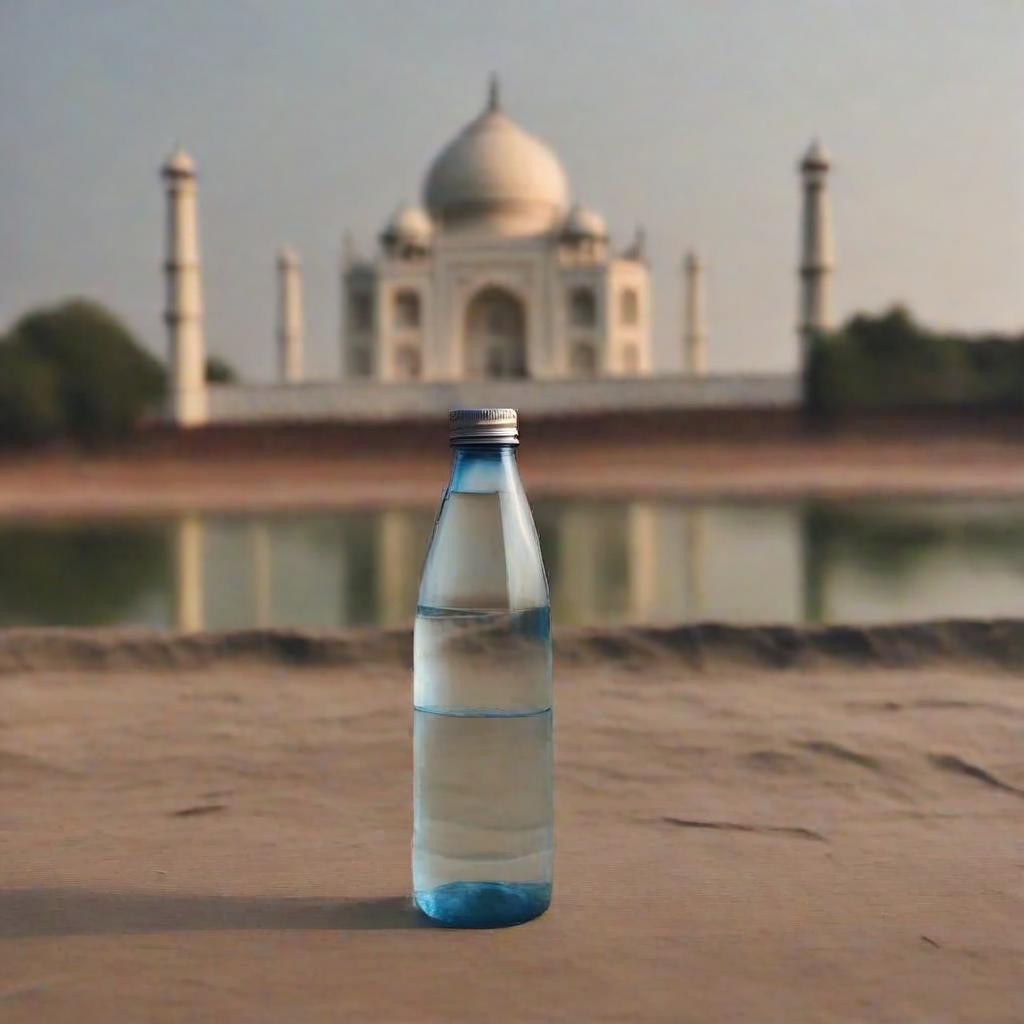}}
        \\
        \\
        
        \multicolumn{5}{c}{{\textit{``a photo of a bottle of water''}}}
        \\
        \\

        {\includegraphics[valign=c, width=\ww]{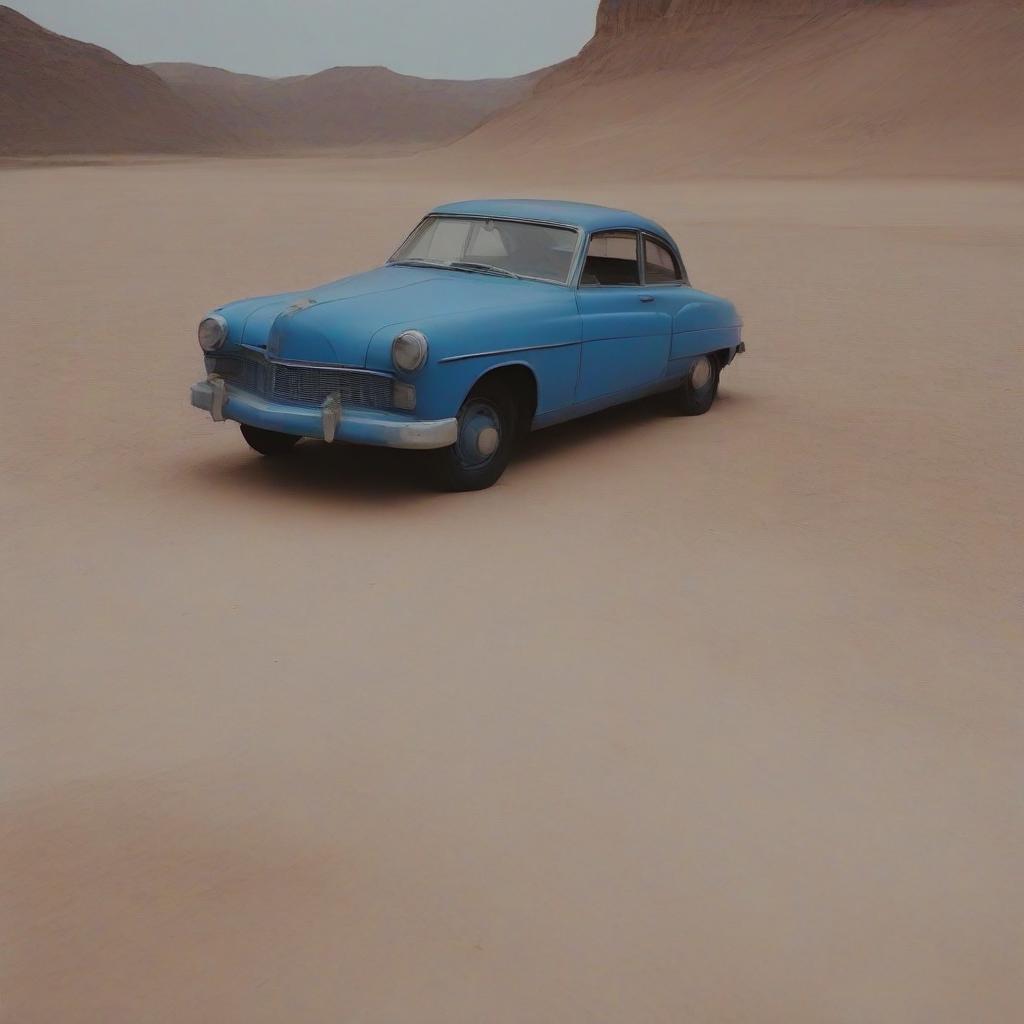}} &
        {\includegraphics[valign=c, width=\ww]{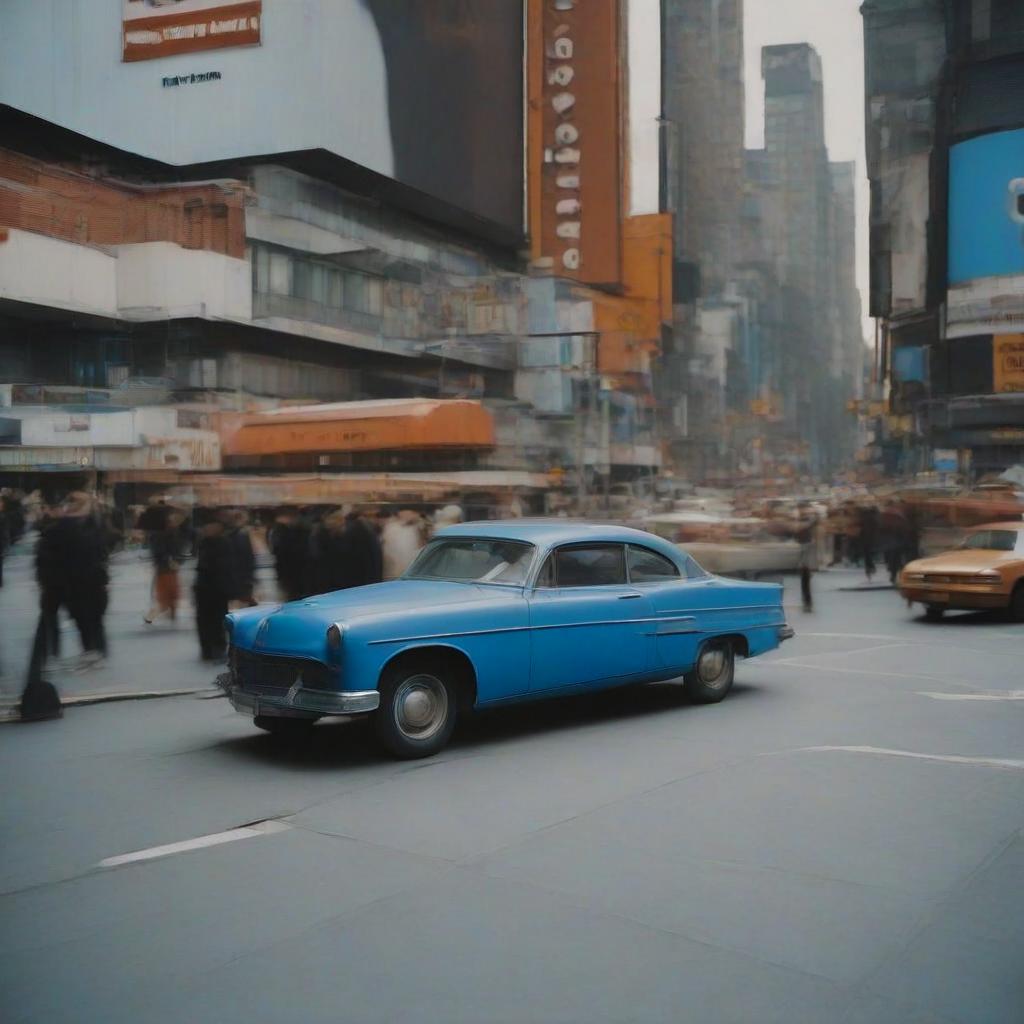}} &
        {\includegraphics[valign=c, width=\ww]{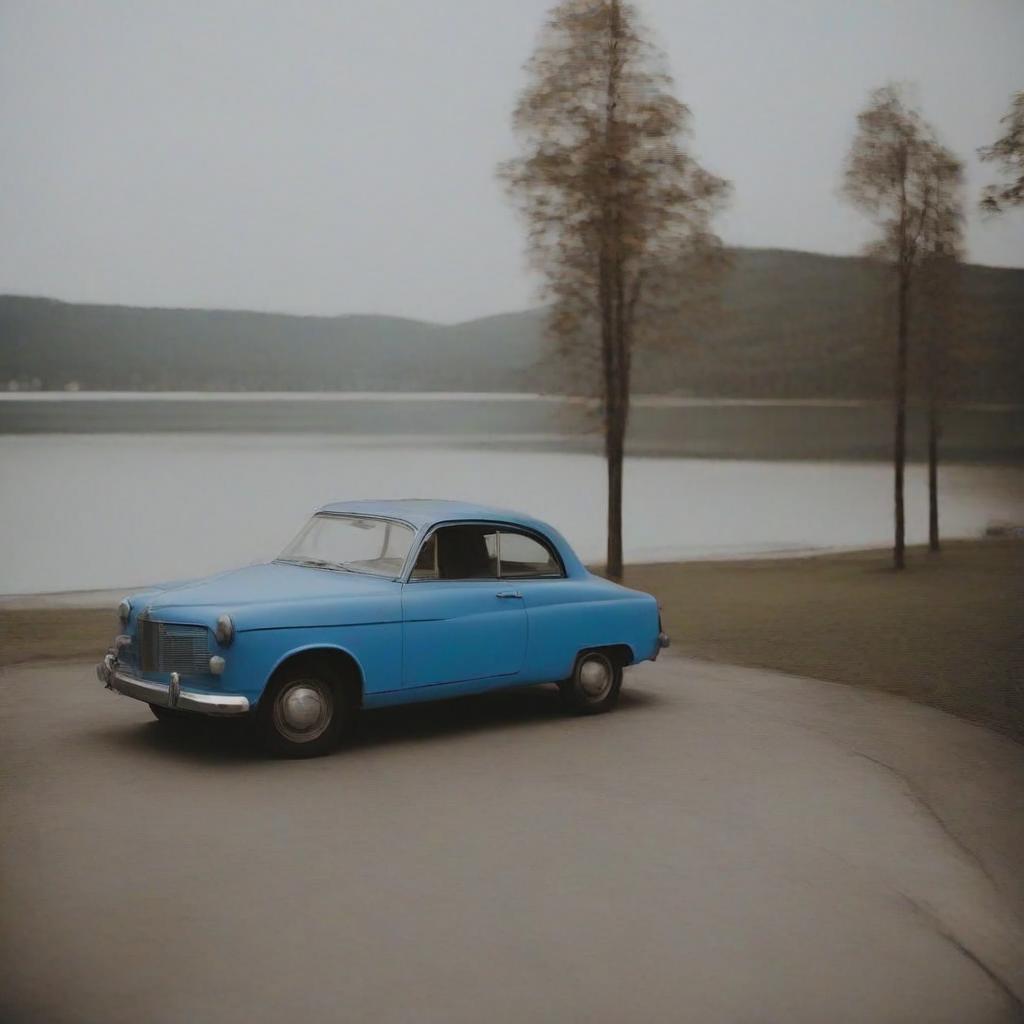}} &
        {\includegraphics[valign=c, width=\ww]{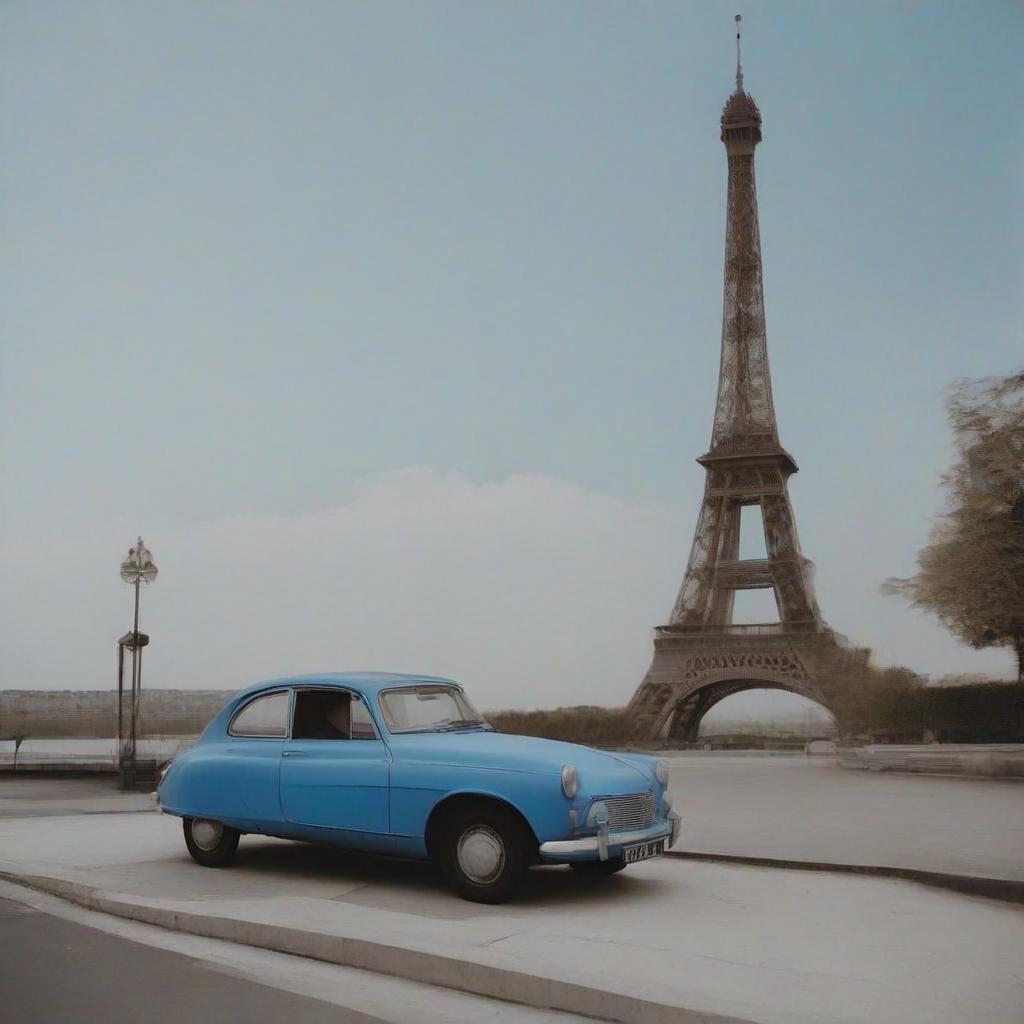}} &
        {\includegraphics[valign=c, width=\ww]{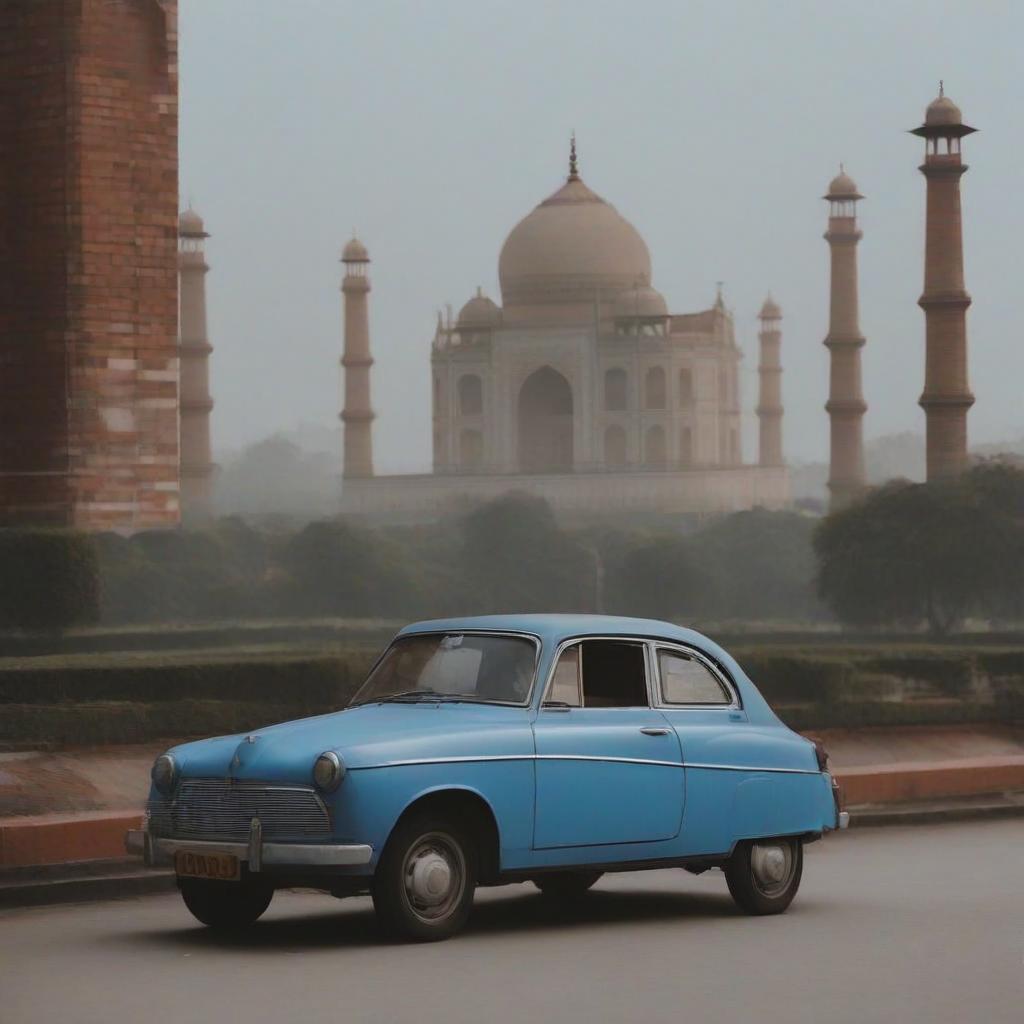}}
        \\
        \\
        
        \multicolumn{5}{c}{{\textit{``a photo of a blue car''}}}
        \\
        \\

        {\includegraphics[valign=c, width=\ww]{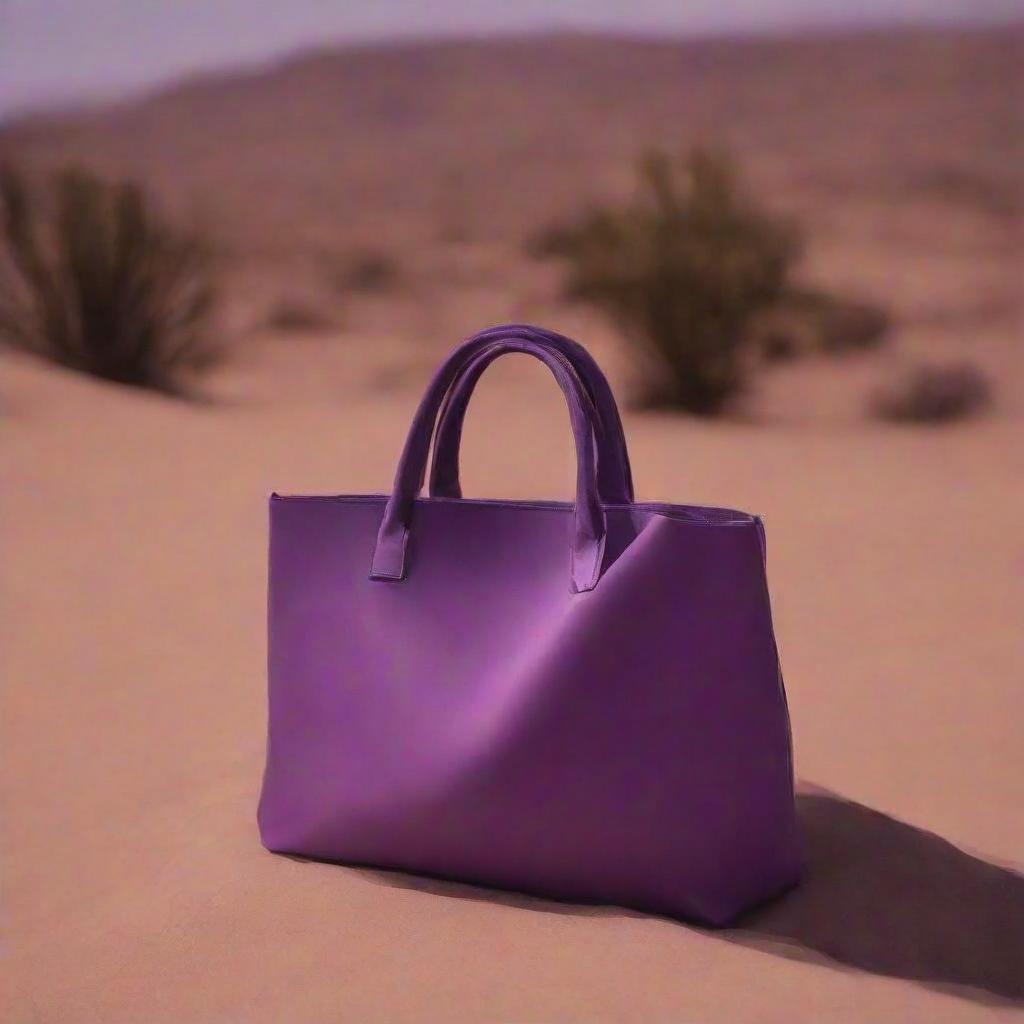}} &
        {\includegraphics[valign=c, width=\ww]{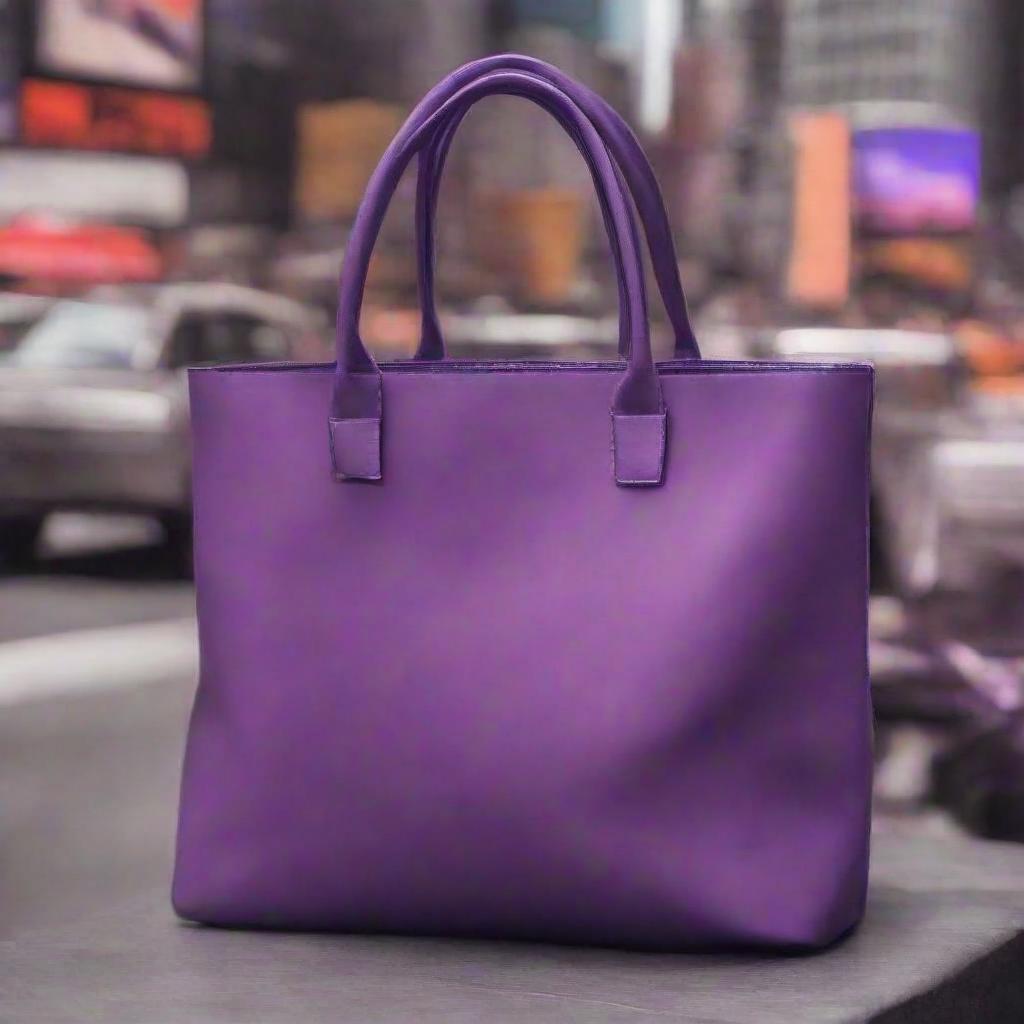}} &
        {\includegraphics[valign=c, width=\ww]{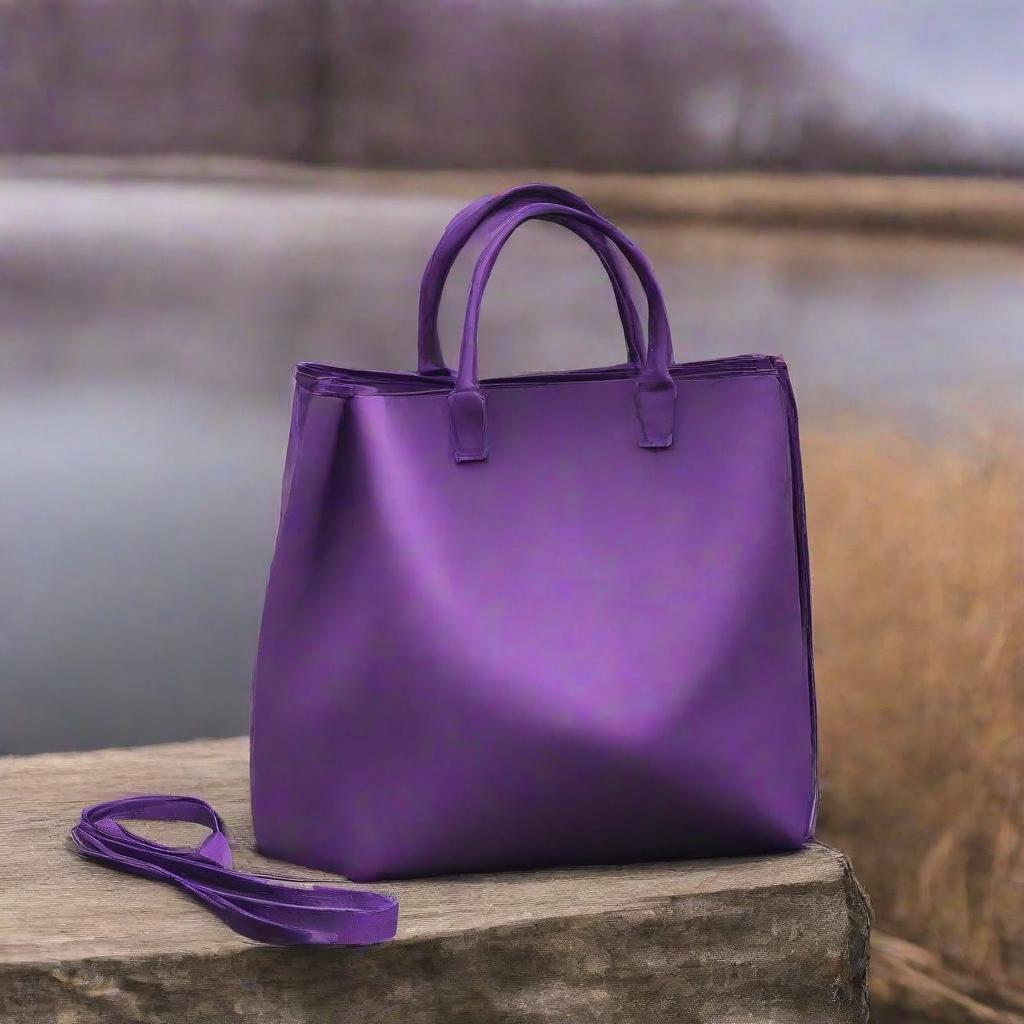}} &
        {\includegraphics[valign=c, width=\ww]{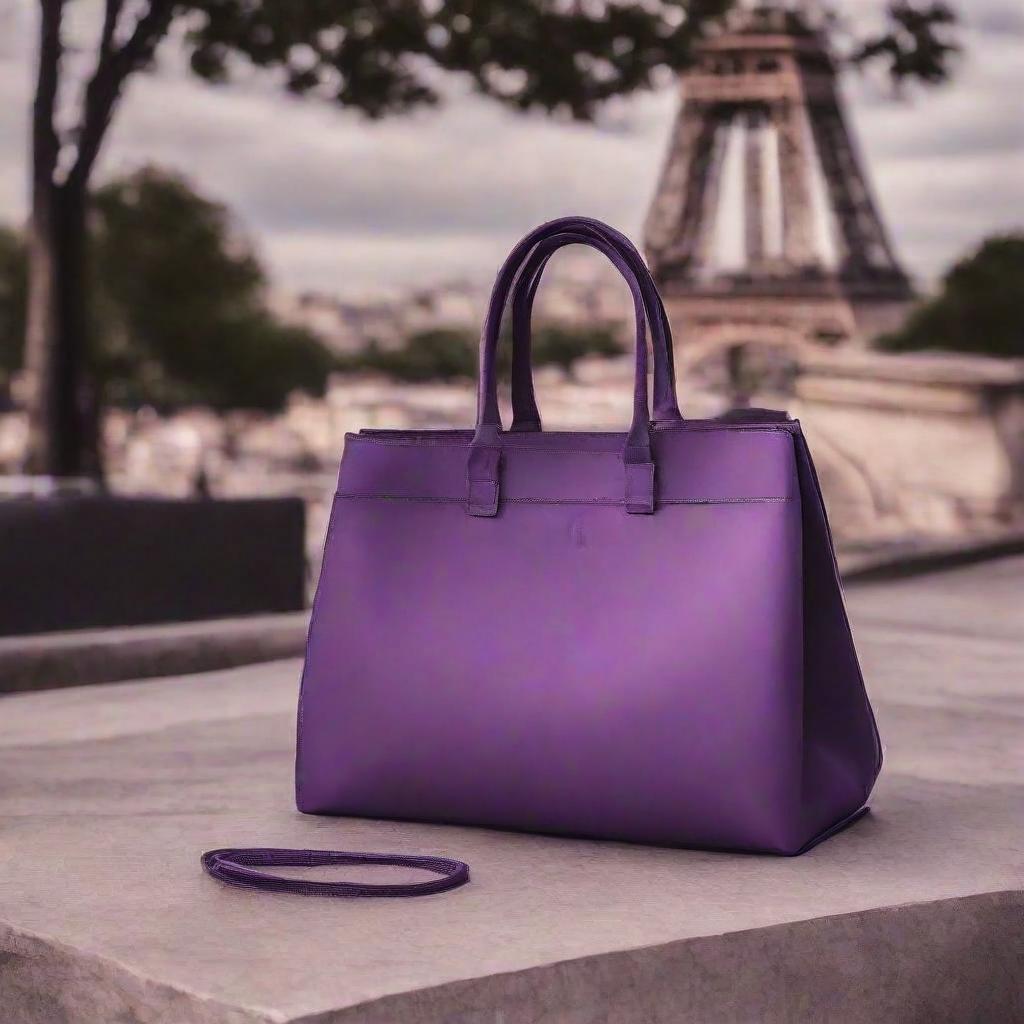}} &
        {\includegraphics[valign=c, width=\ww]{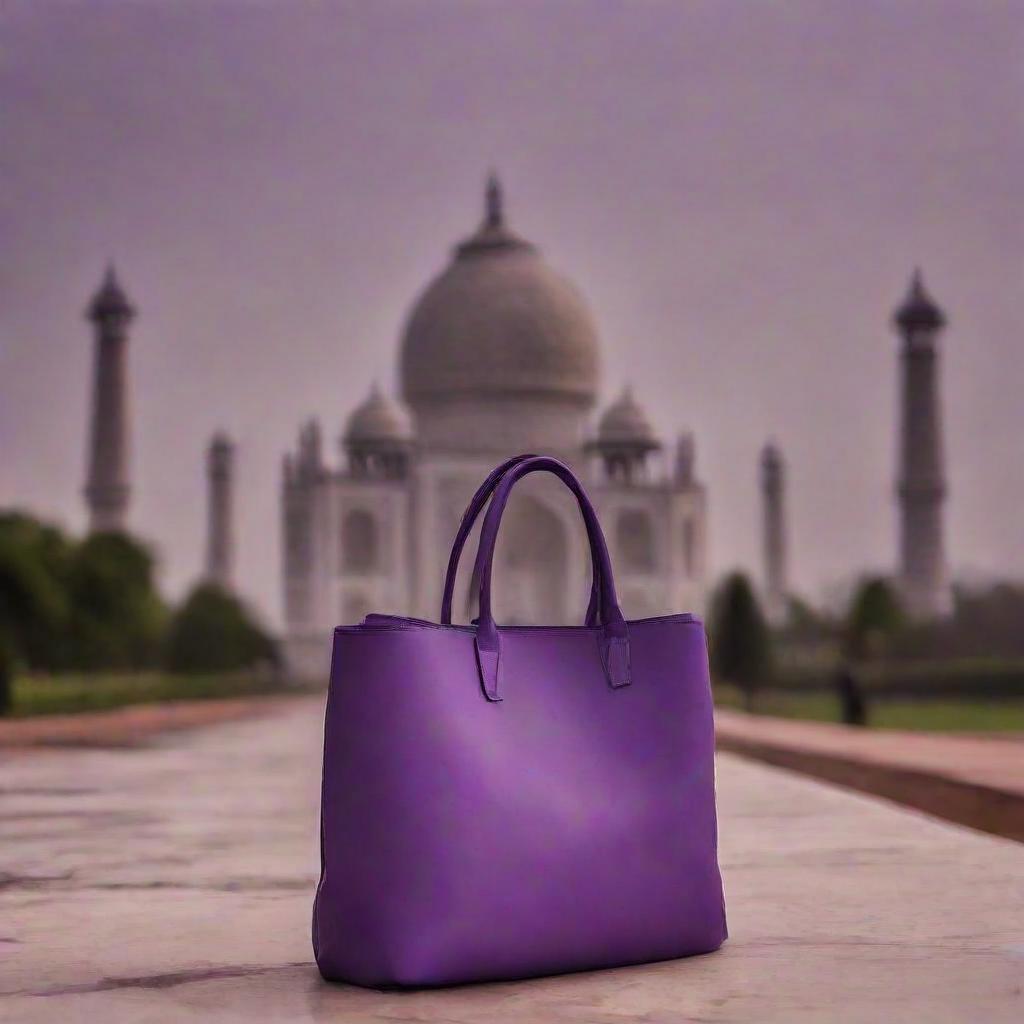}}
        \\
        \\
        
        \multicolumn{5}{c}{{\textit{``a photo of a purple bag''}}}
        \\
        \\

        {\includegraphics[valign=c, width=\ww]{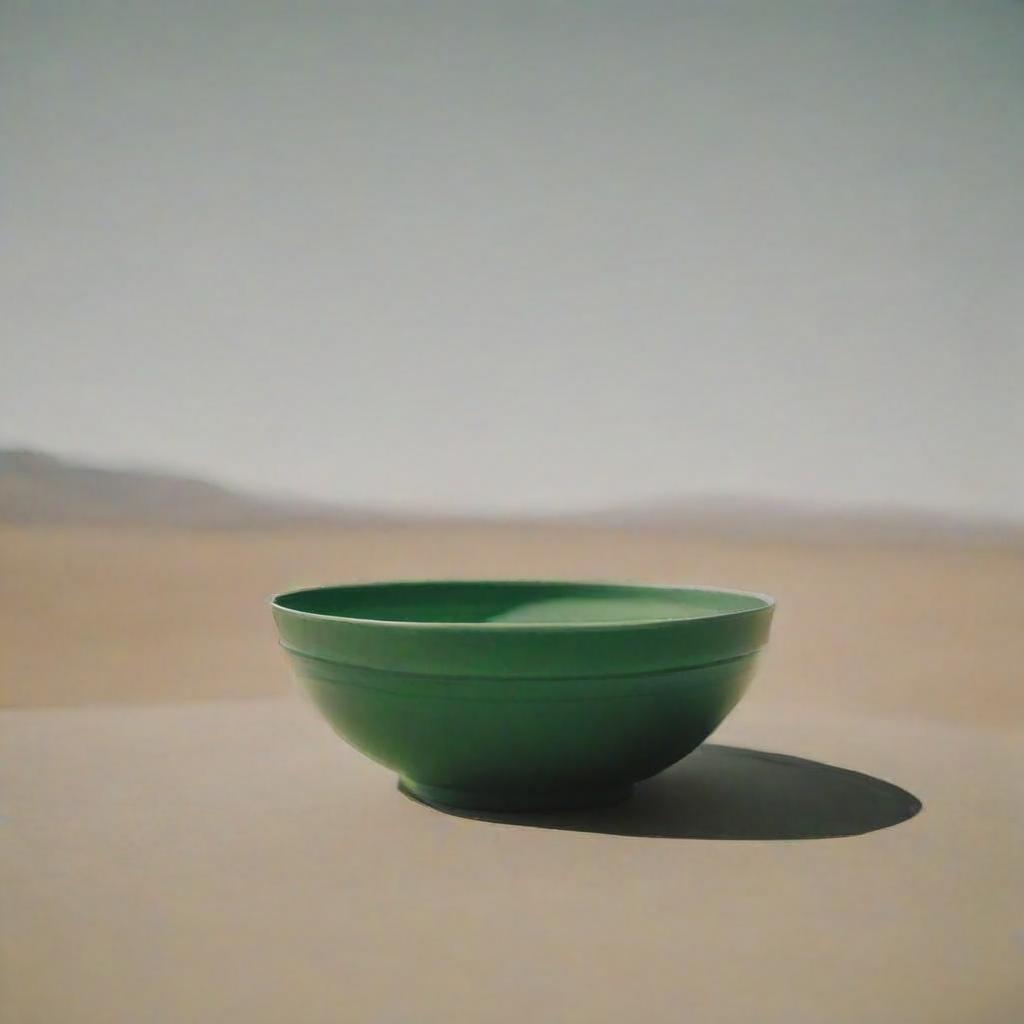}} &
        {\includegraphics[valign=c, width=\ww]{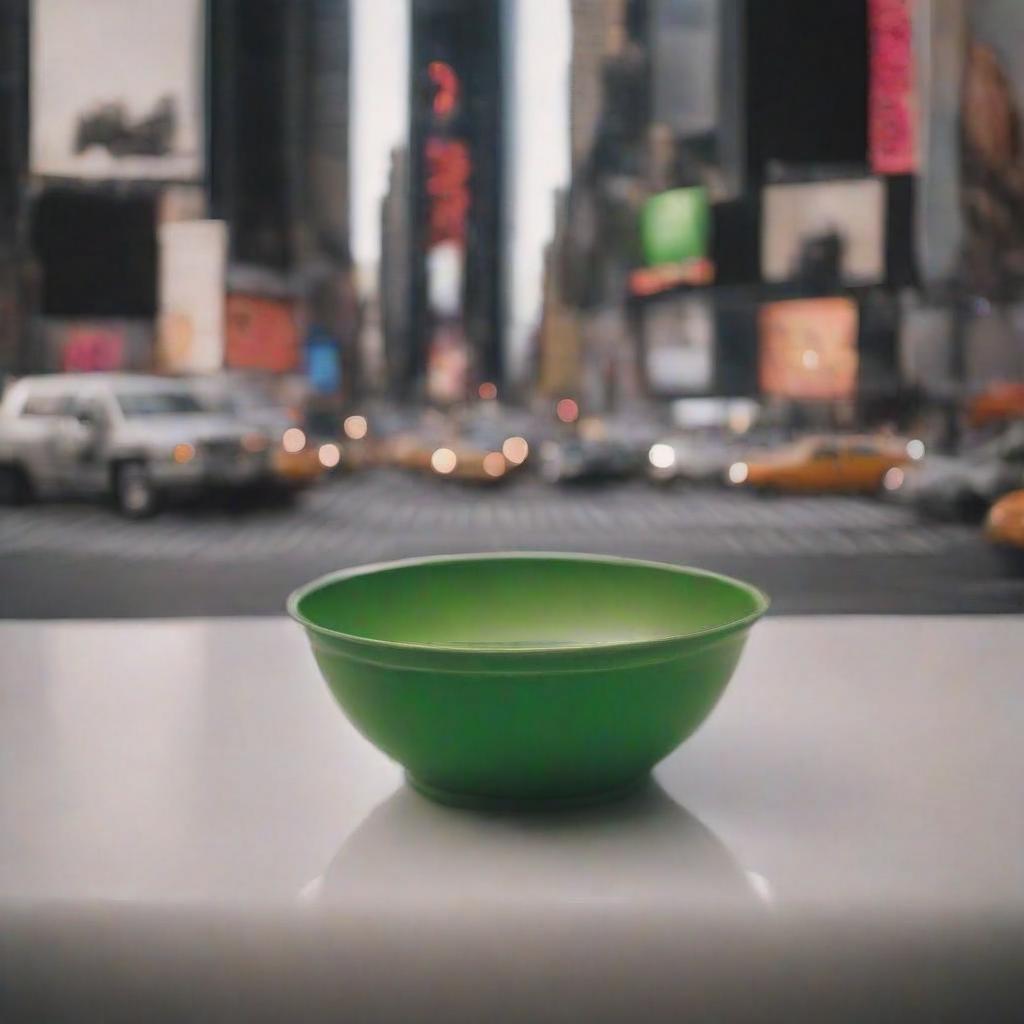}} &
        {\includegraphics[valign=c, width=\ww]{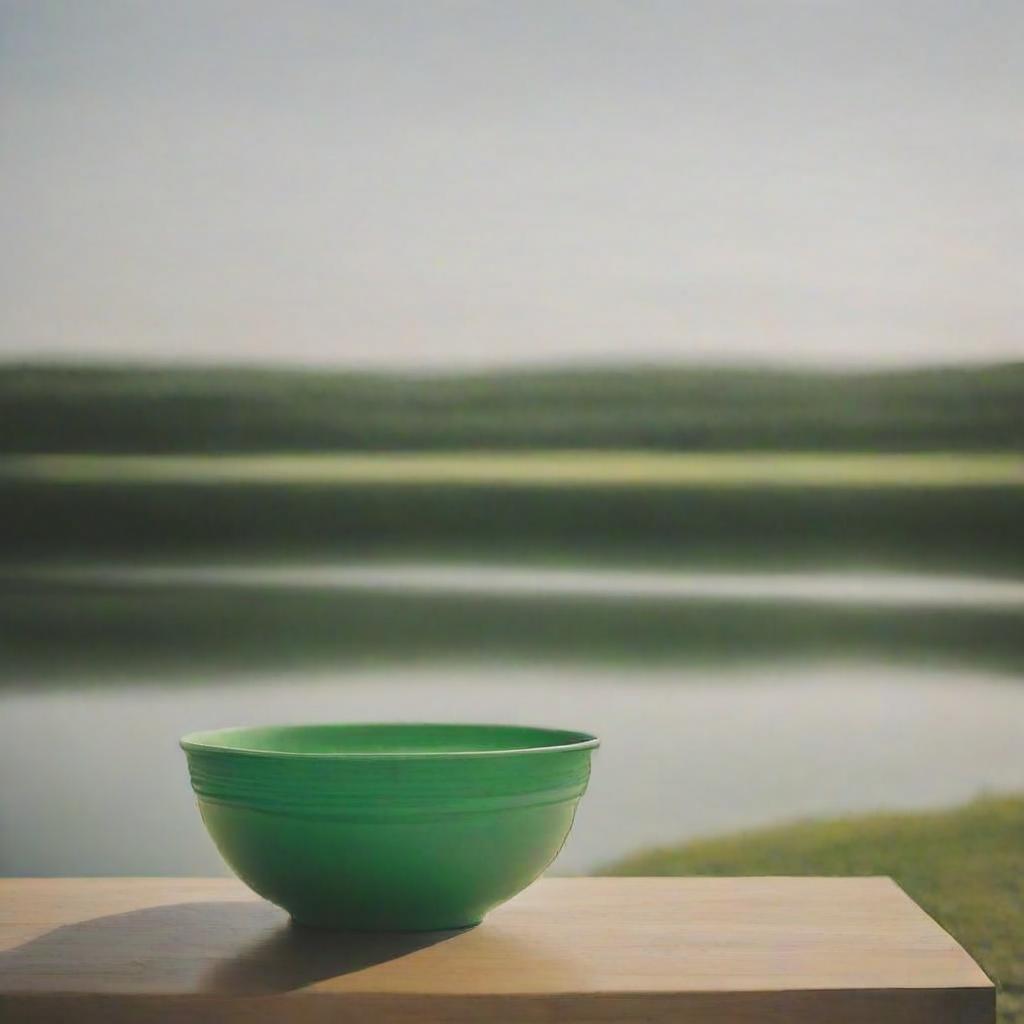}} &
        {\includegraphics[valign=c, width=\ww]{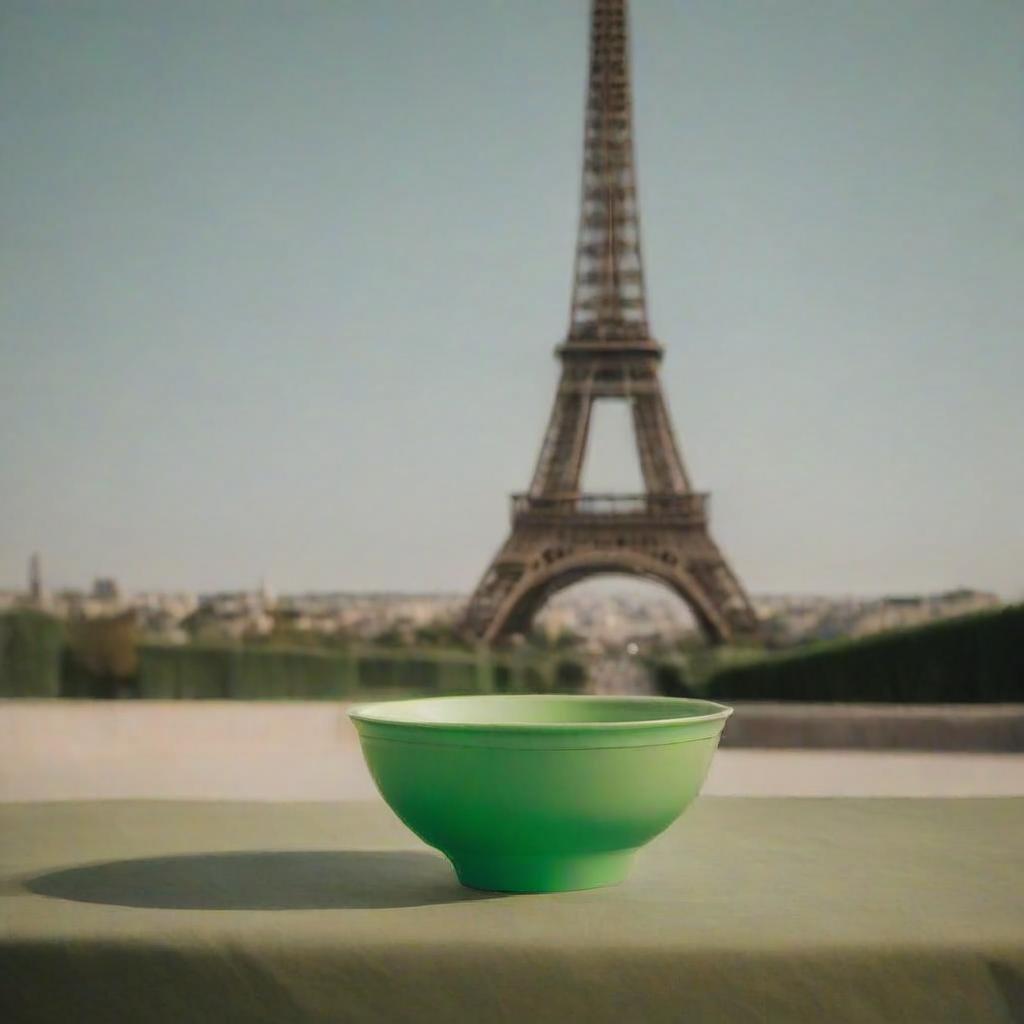}} &
        {\includegraphics[valign=c, width=\ww]{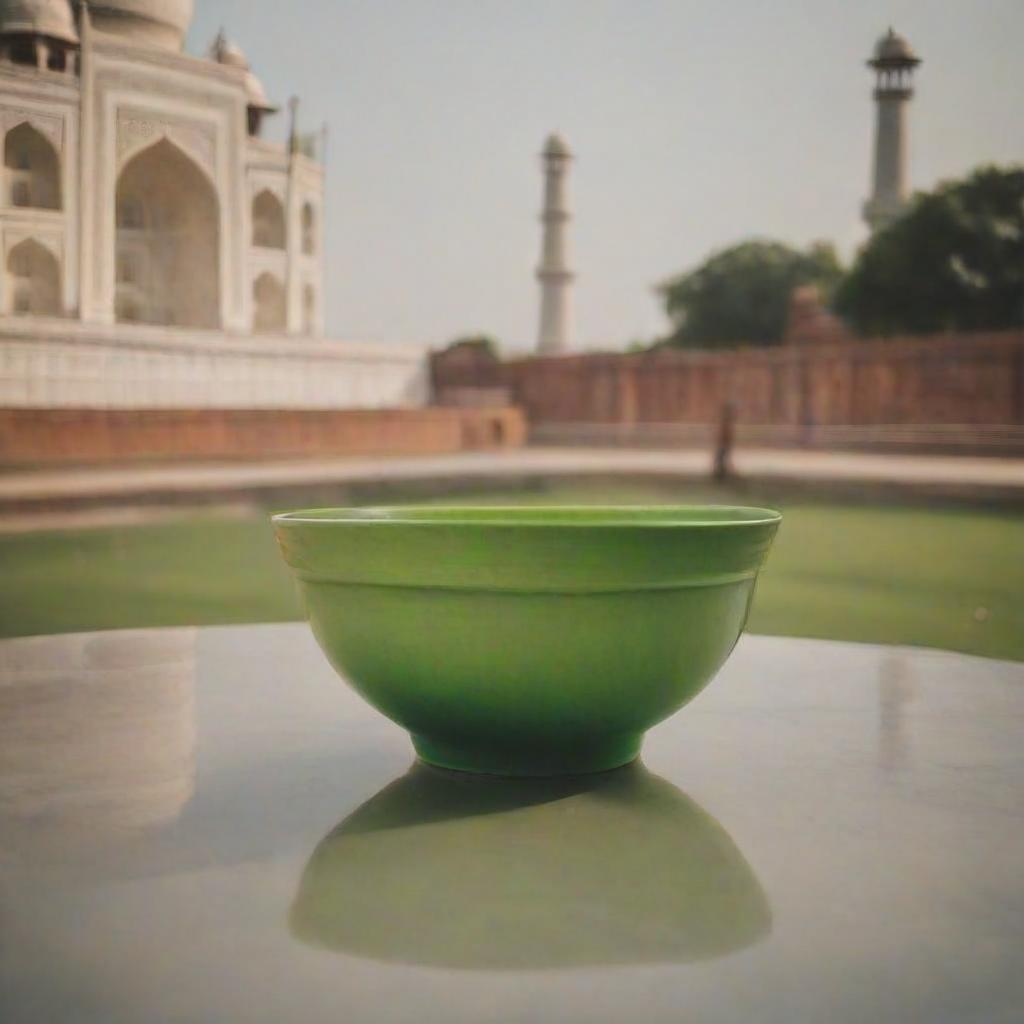}}
        \\
        \\
        
        \multicolumn{5}{c}{{\textit{``a photo of a green bowl''}}}

    \end{tabular}
    
    \caption{\textbf{Consistent generation of non-character objects.} Our approach is applicable to a wide range of objects, without the requirement for them to depict human characters or creatures.}
    \label{fig:general_objects}
\end{figure*}

%% file: figures/teaser/fig_extended.tex
\begin{figure*}[t]
    \centering
    \setlength{\tabcolsep}{0.5pt}
    \renewcommand{\arraystretch}{1.0}
    \setlength{\ww}{0.4\columnwidth}
    \begin{tabular}{ccccc}
        &&&&
        \textit{``holding an}
        \\

        &
        \textit{``in the park''} &
        \textit{``reading a book''} &
        \textit{``at the beach''} &
        \textit{avocado''}
        \\

        {\includegraphics[valign=c, width=\ww]{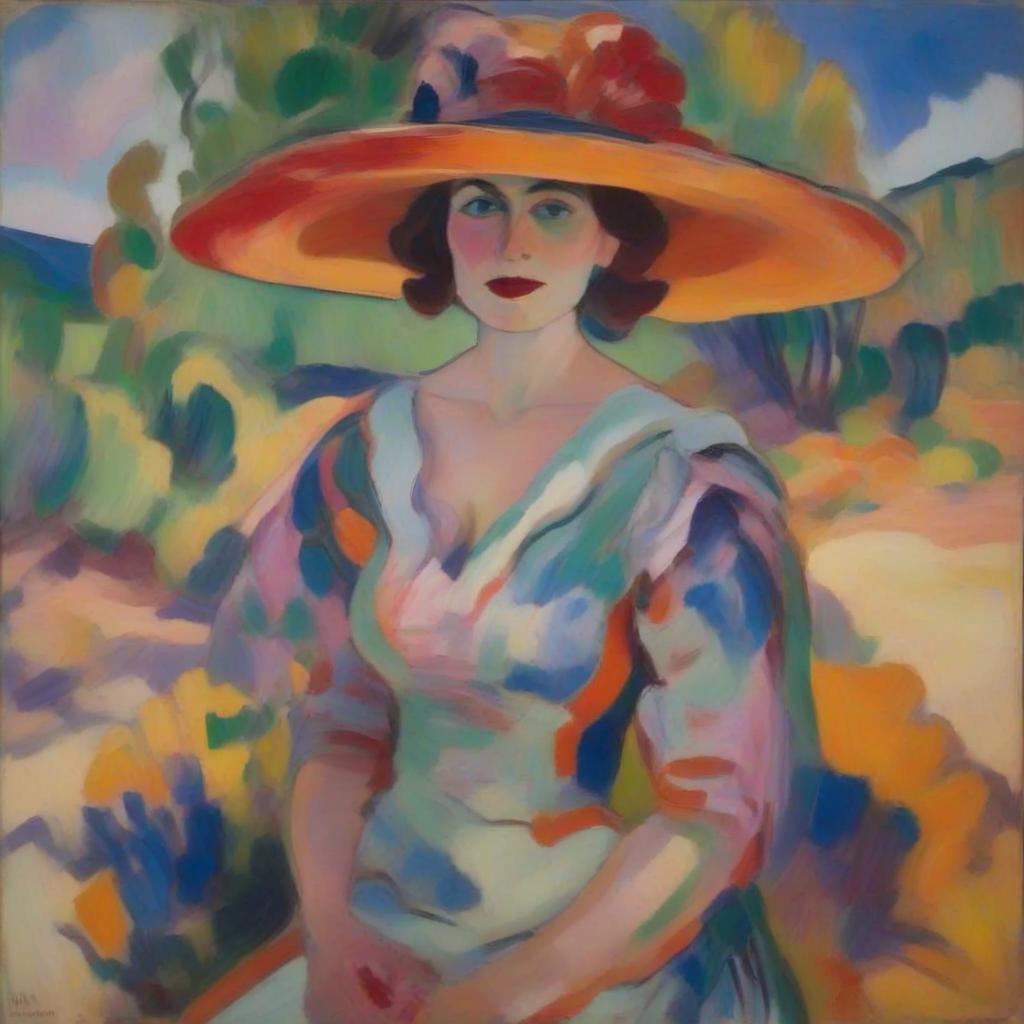}} &
        {\includegraphics[valign=c, width=\ww]{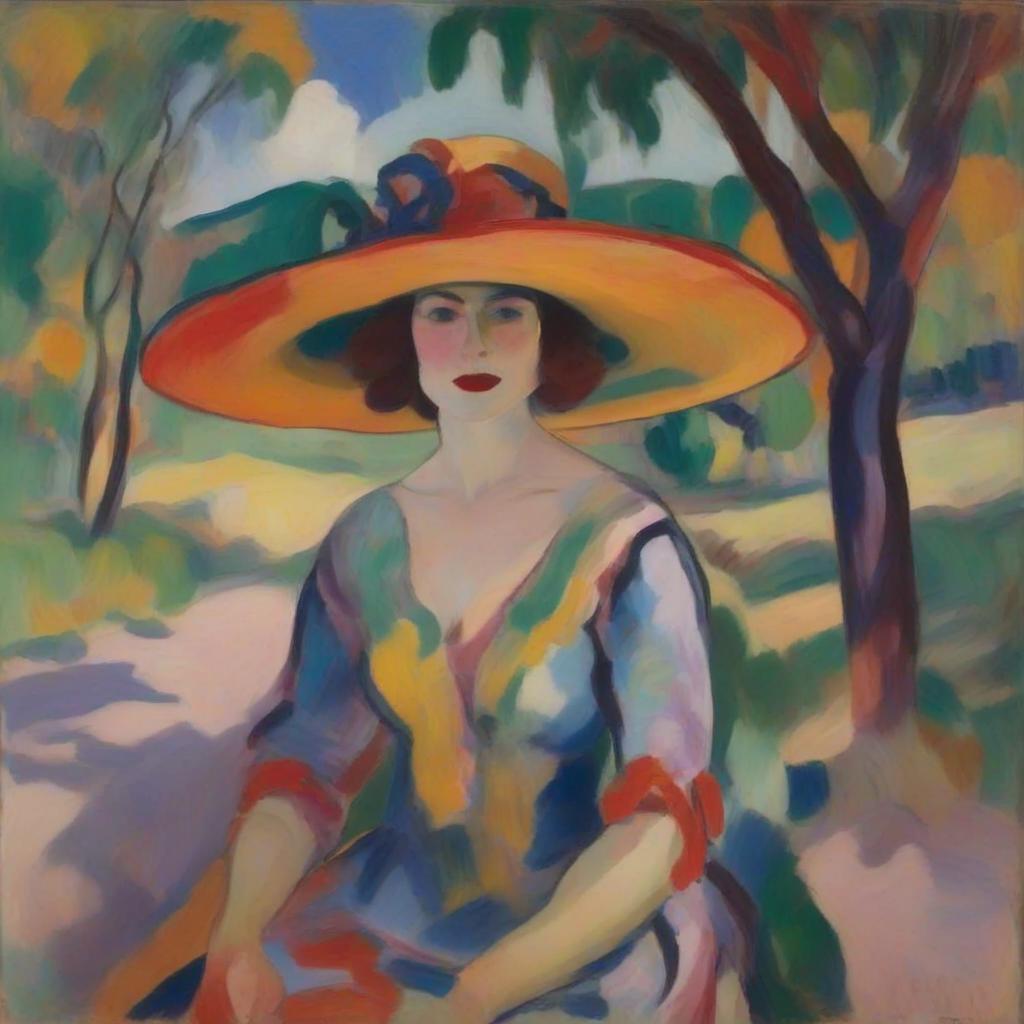}} &
        {\includegraphics[valign=c, width=\ww]{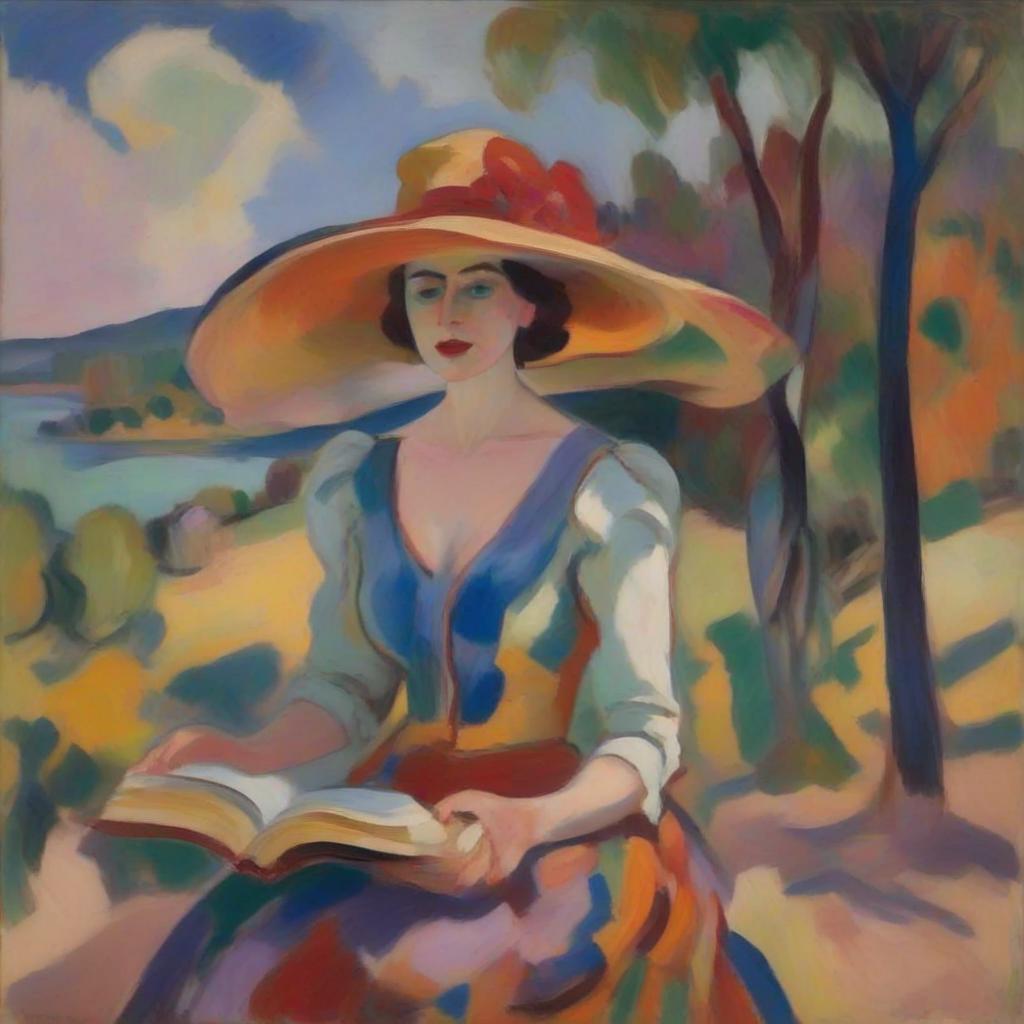}} &
        {\includegraphics[valign=c, width=\ww]{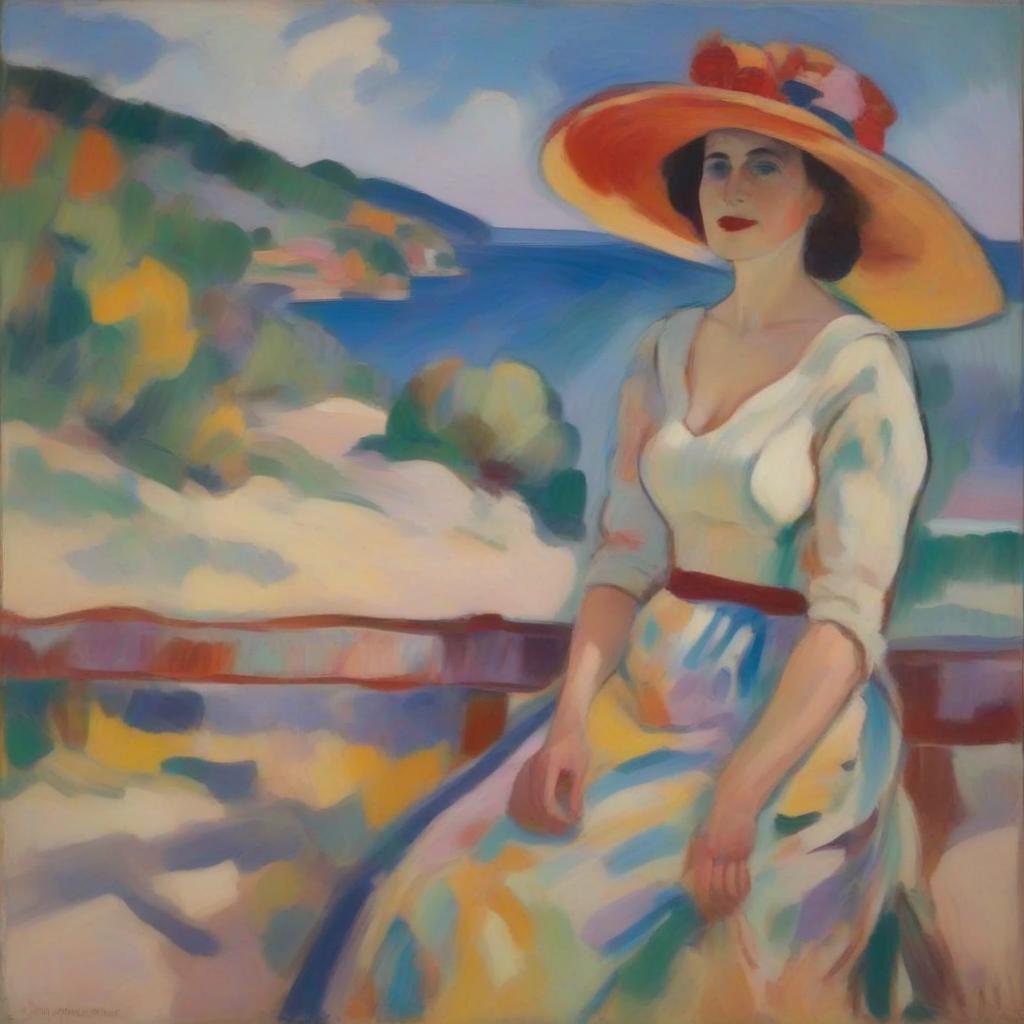}} &
        {\includegraphics[valign=c, width=\ww]{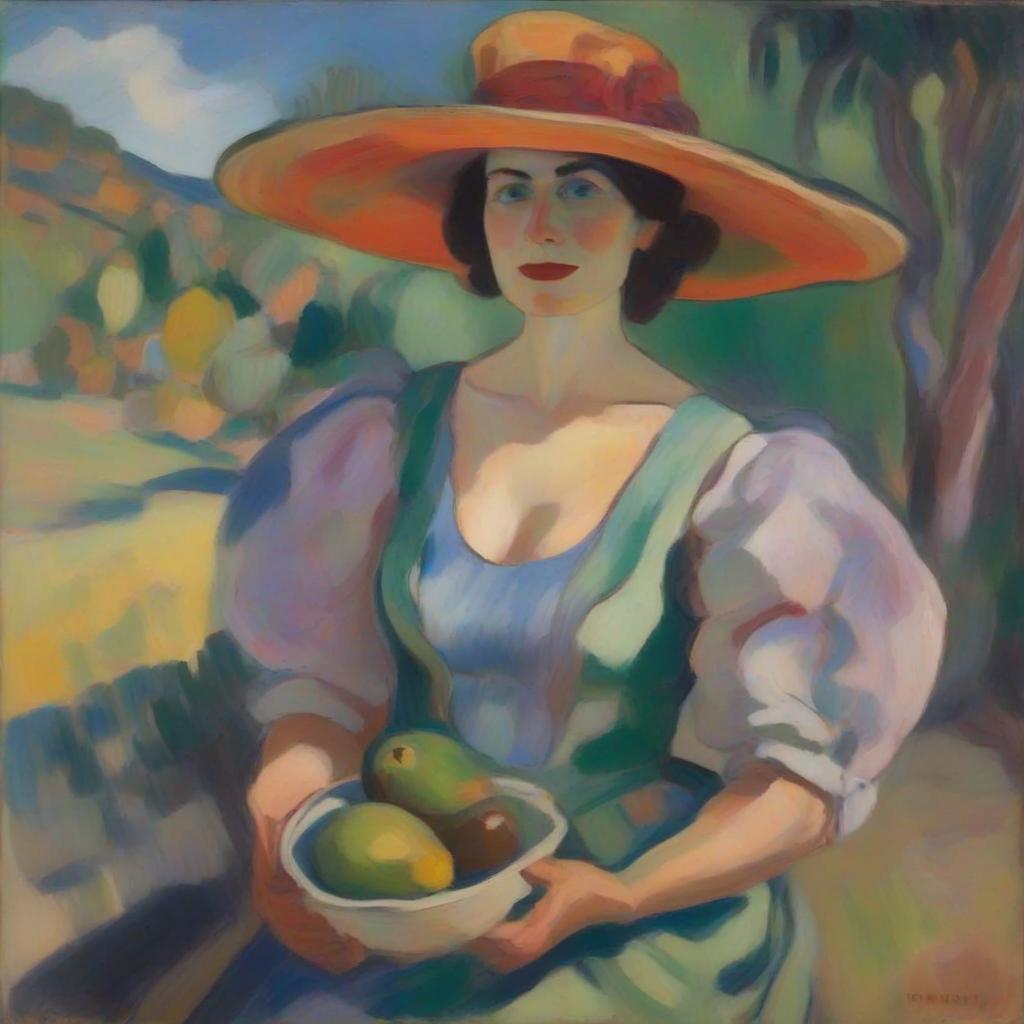}}
        \\

        \multicolumn{5}{c}{\textit{``a portrait of a woman with a large hat in a scenic environment, fauvism''}}
        \\
        \\

        {\includegraphics[valign=c, width=\ww]{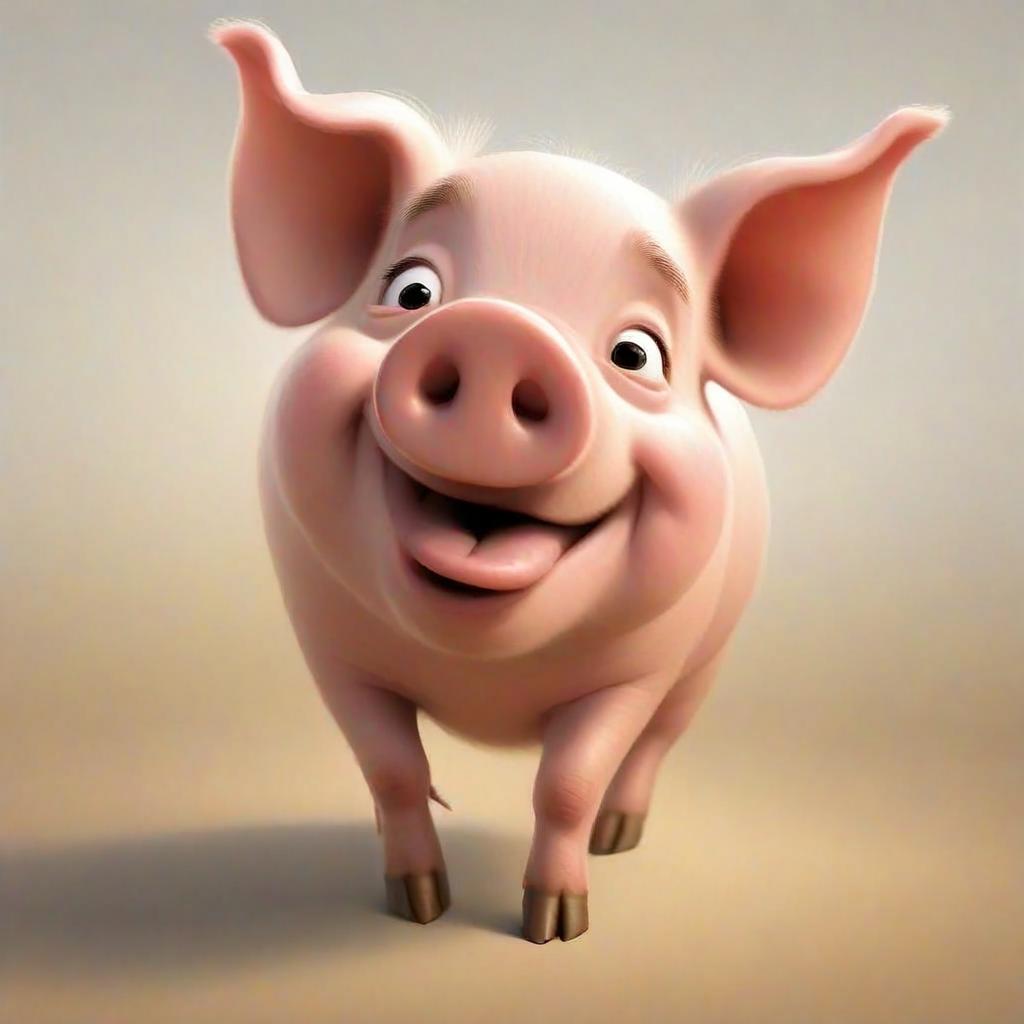}} &
        {\includegraphics[valign=c, width=\ww]{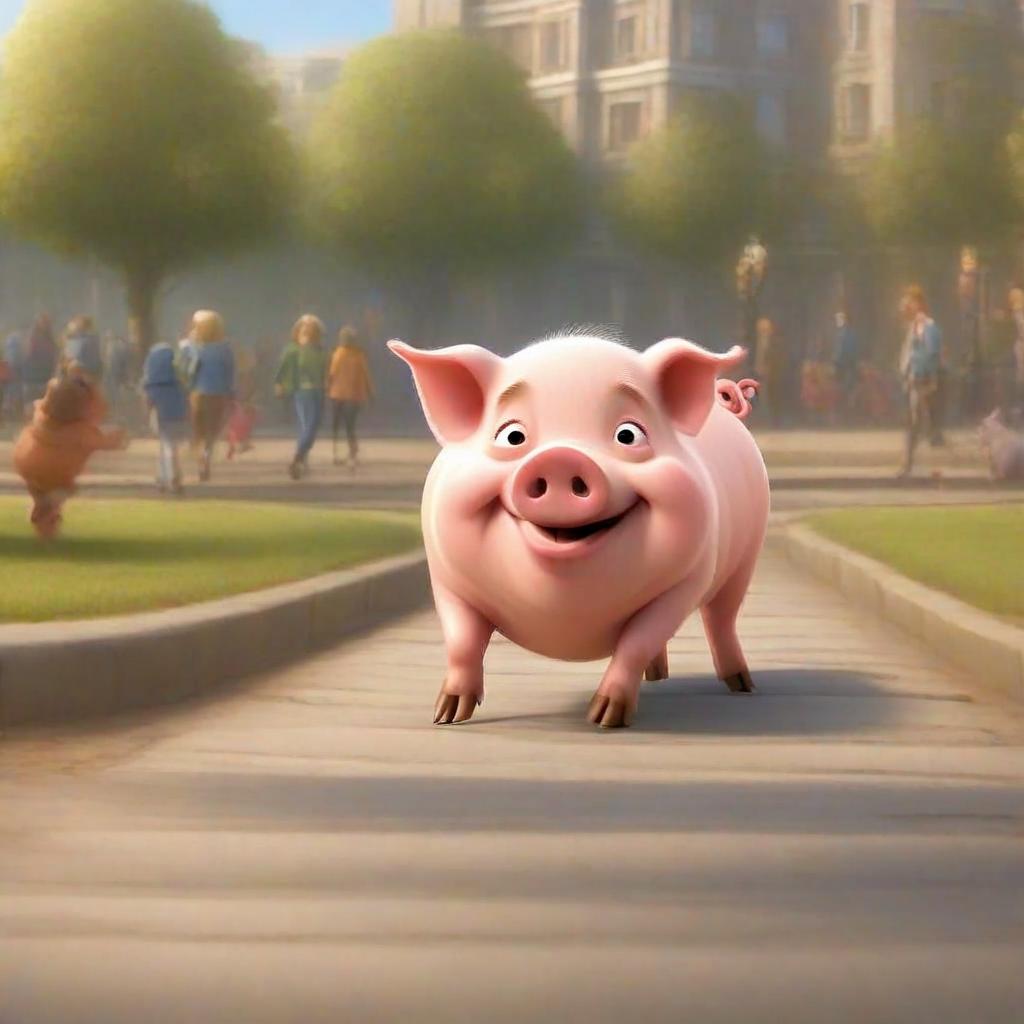}} &
        {\includegraphics[valign=c, width=\ww]{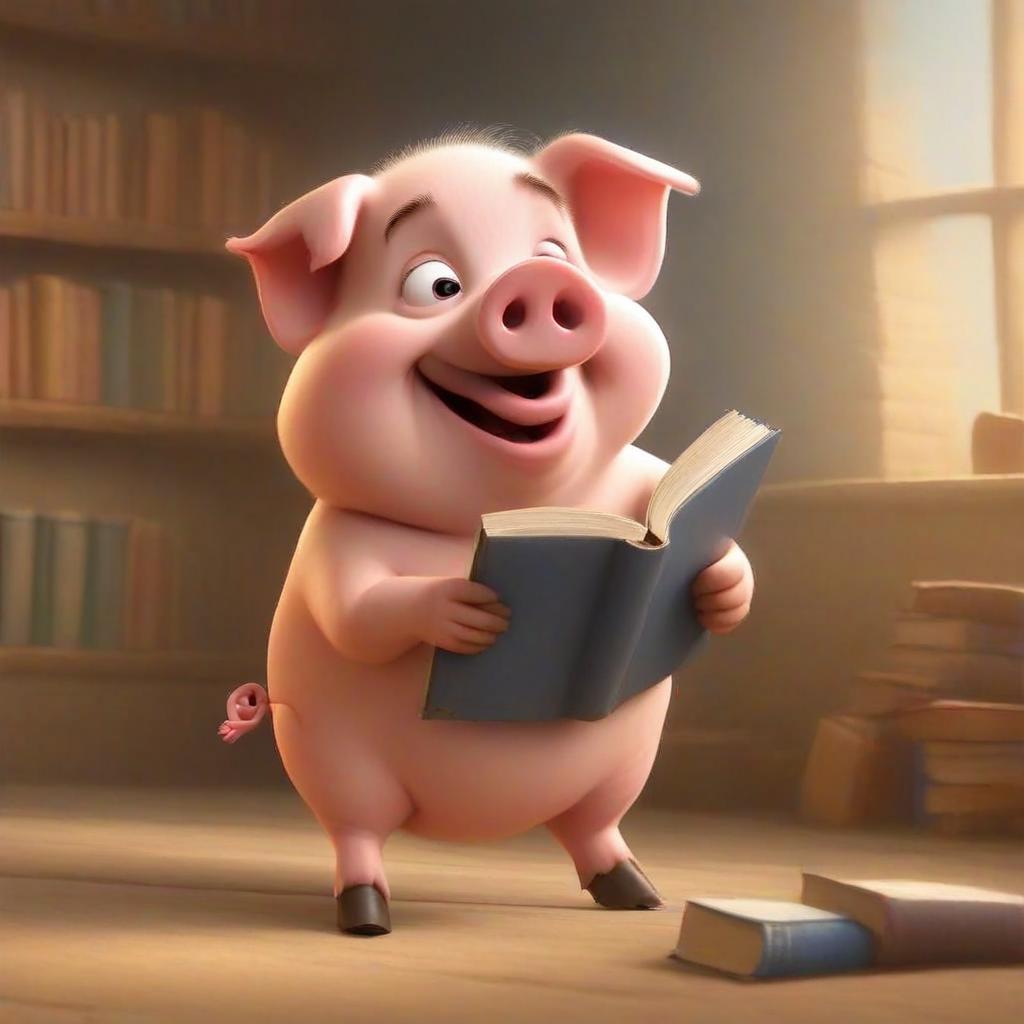}} &
        {\includegraphics[valign=c, width=\ww]{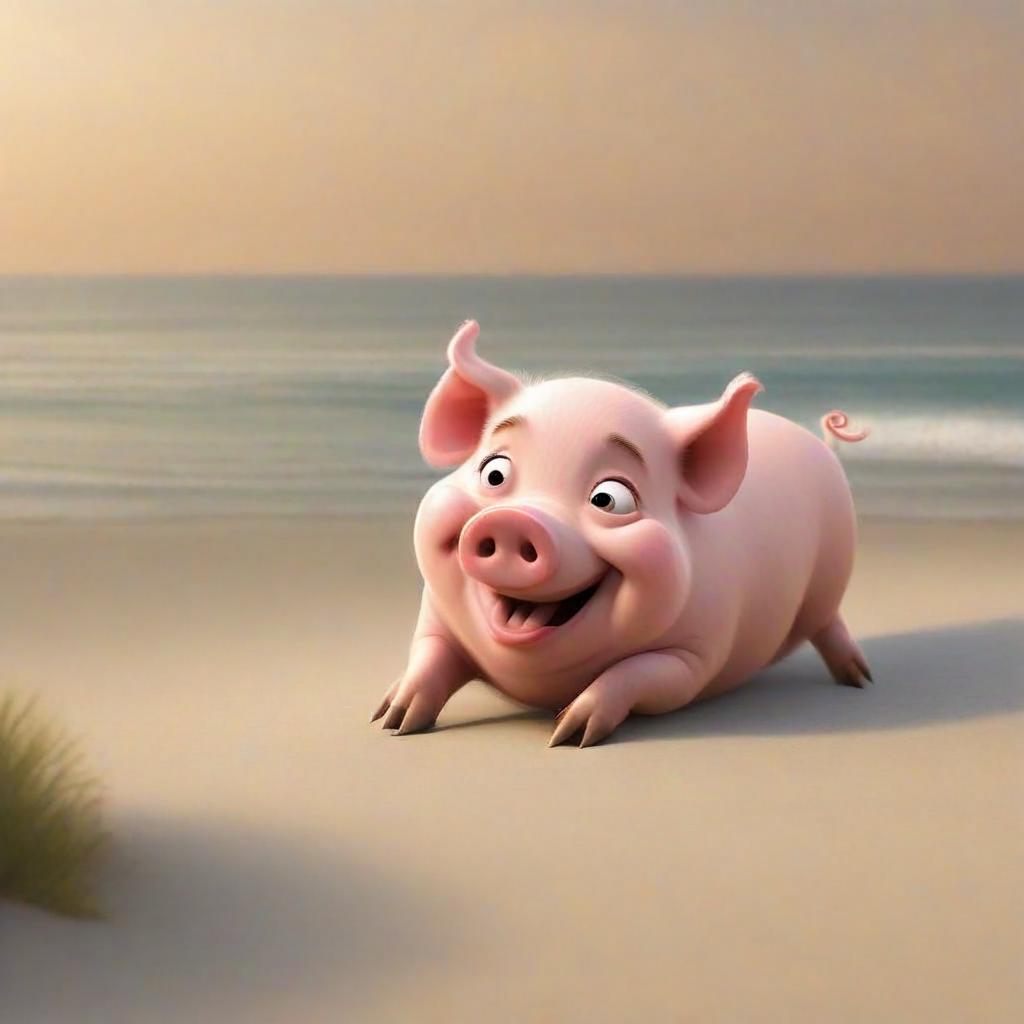}} &
        {\includegraphics[valign=c, width=\ww]{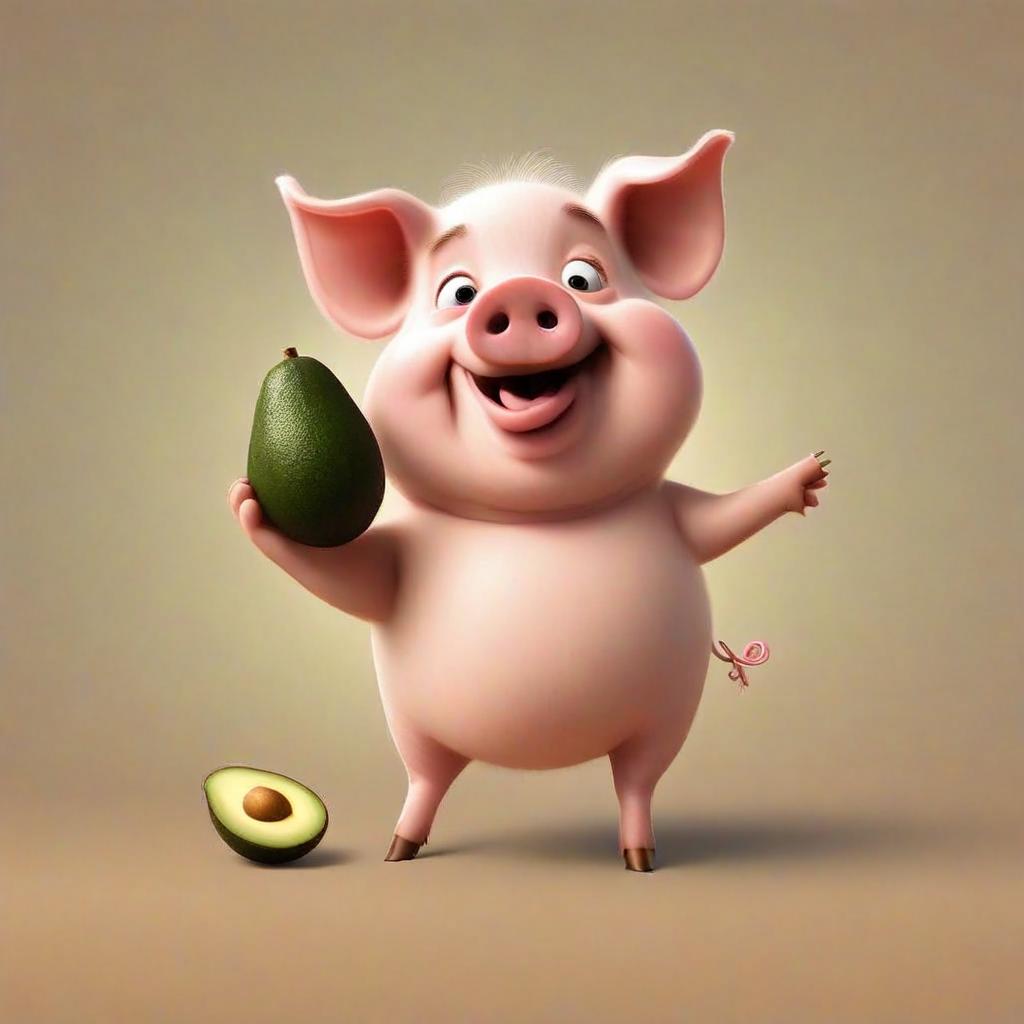}}
        \\

        \multicolumn{5}{c}{\textit{``a 3D animation of a happy pig''}}
        \\
        \\

        {\includegraphics[valign=c, width=\ww]{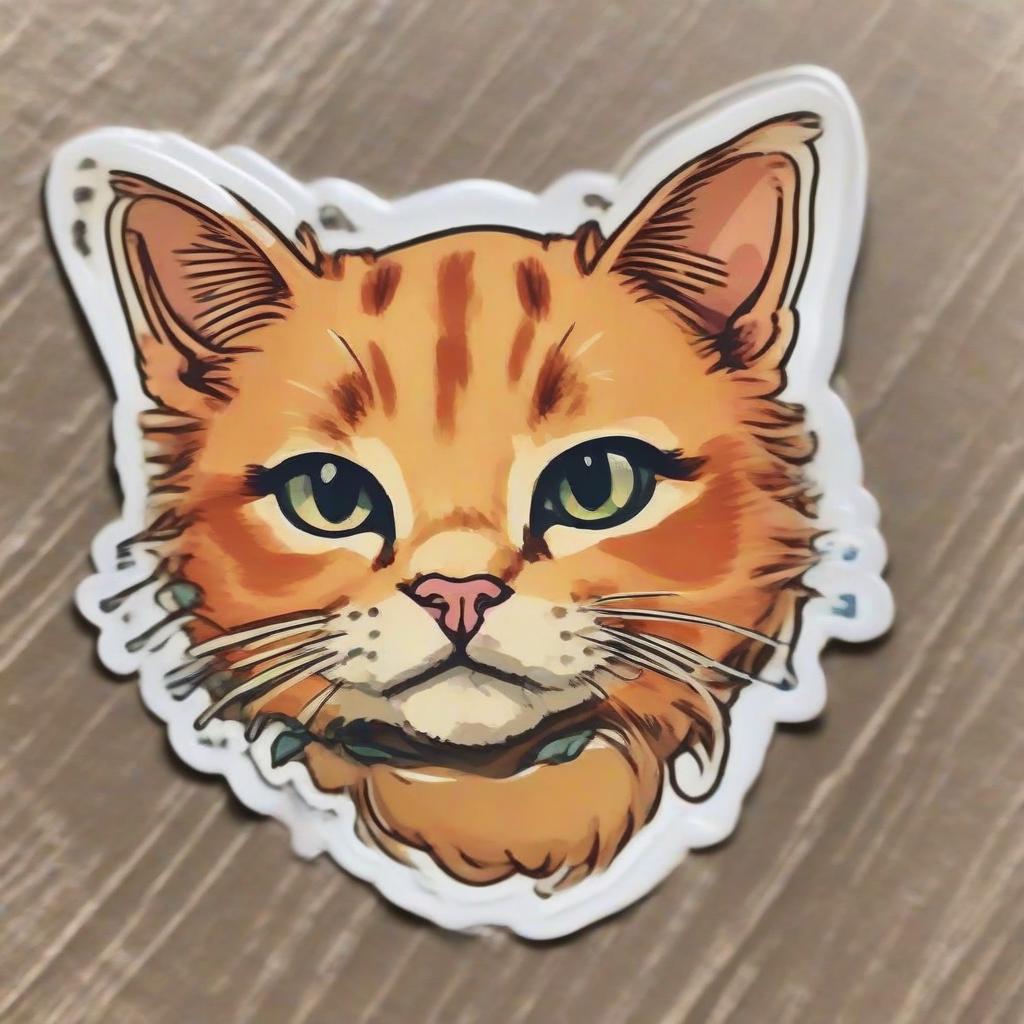}} &
        {\includegraphics[valign=c, width=\ww]{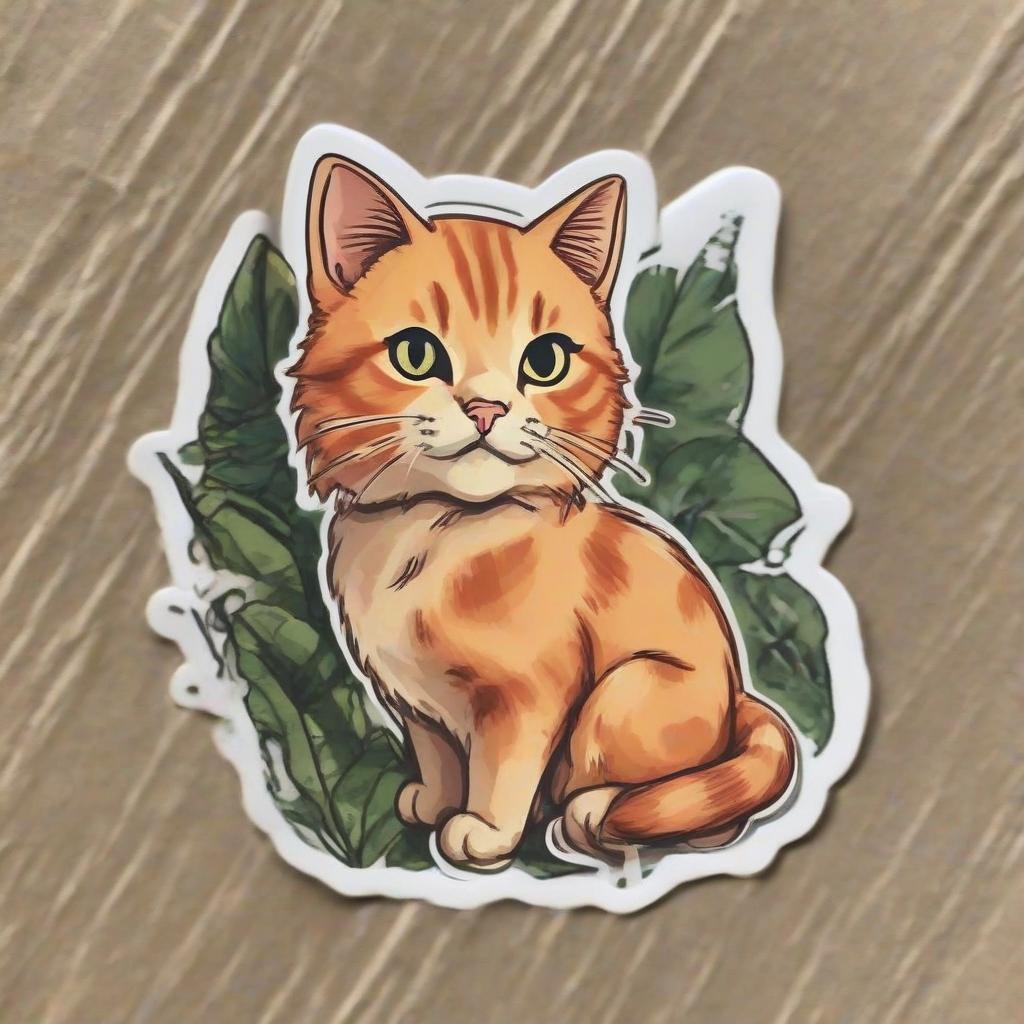}} &
        {\includegraphics[valign=c, width=\ww]{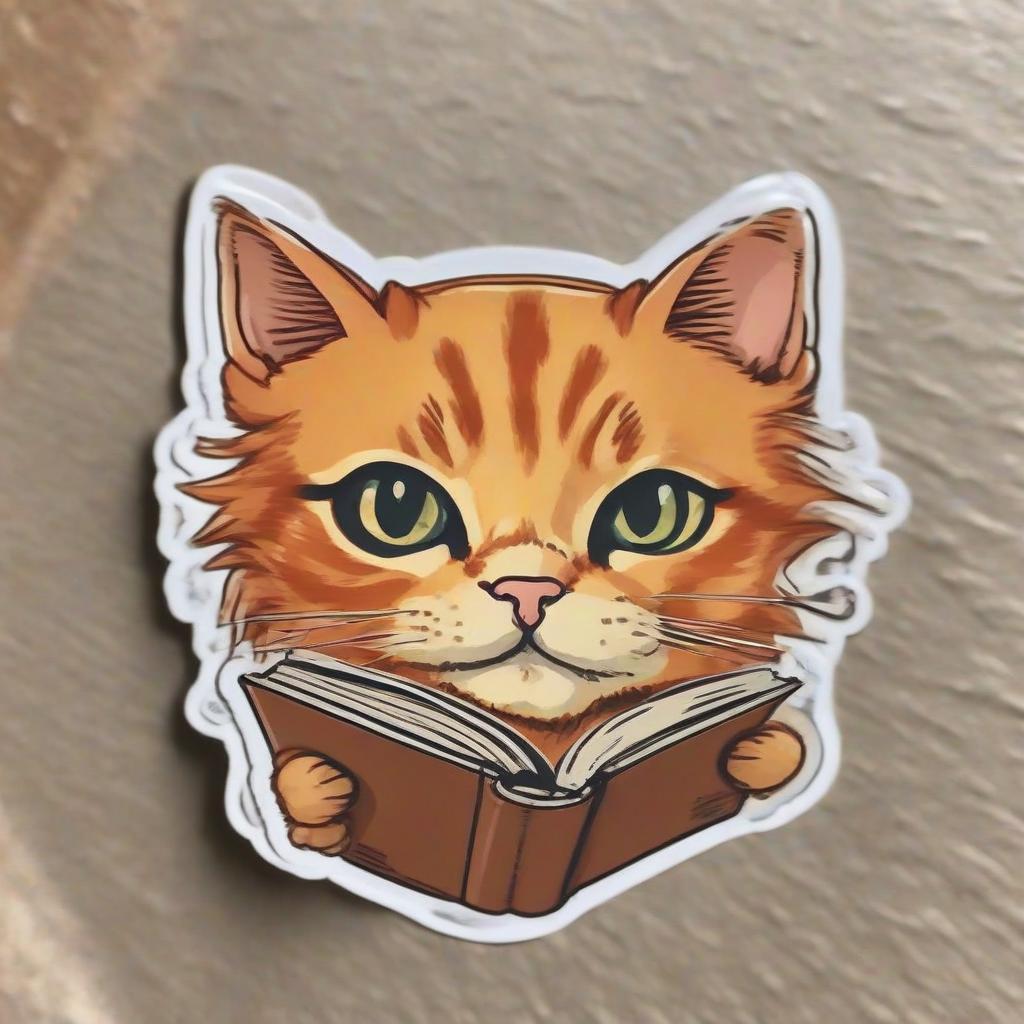}} &
        {\includegraphics[valign=c, width=\ww]{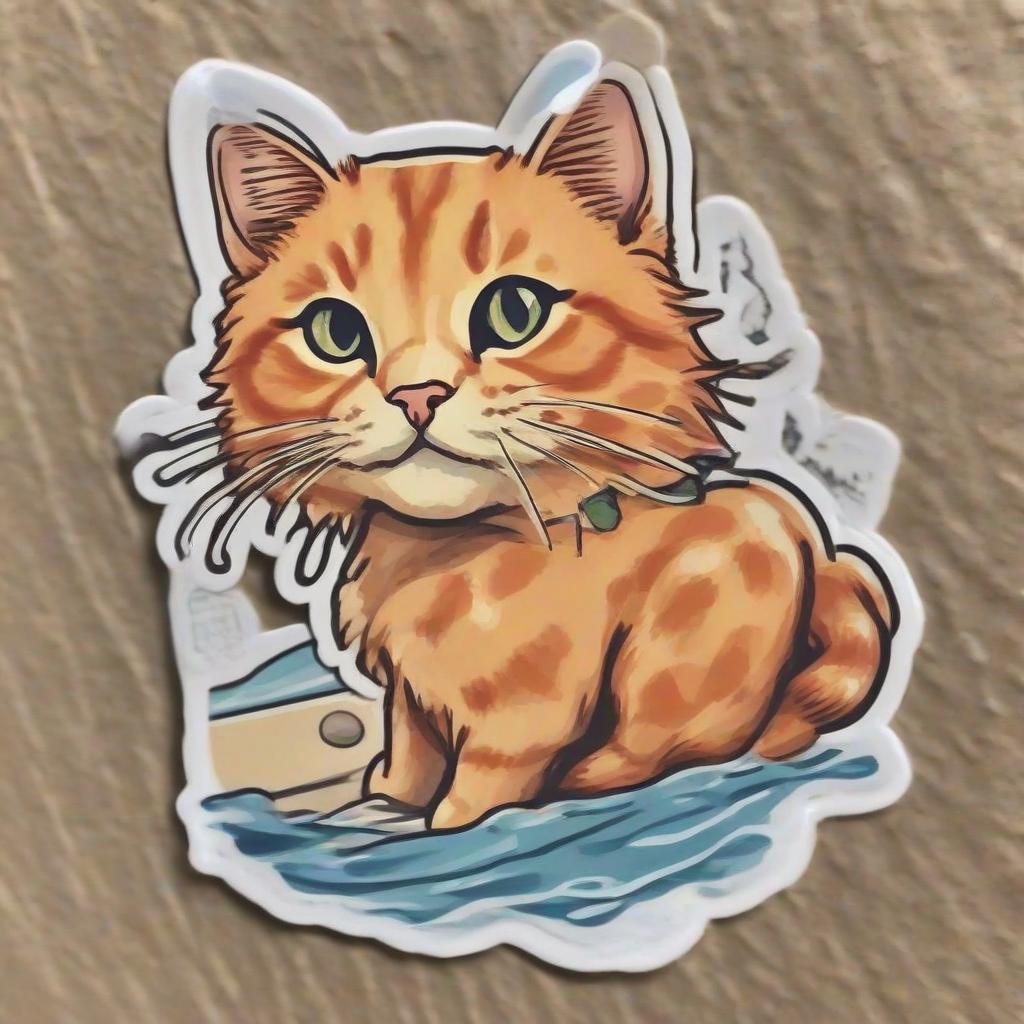}} &
        {\includegraphics[valign=c, width=\ww]{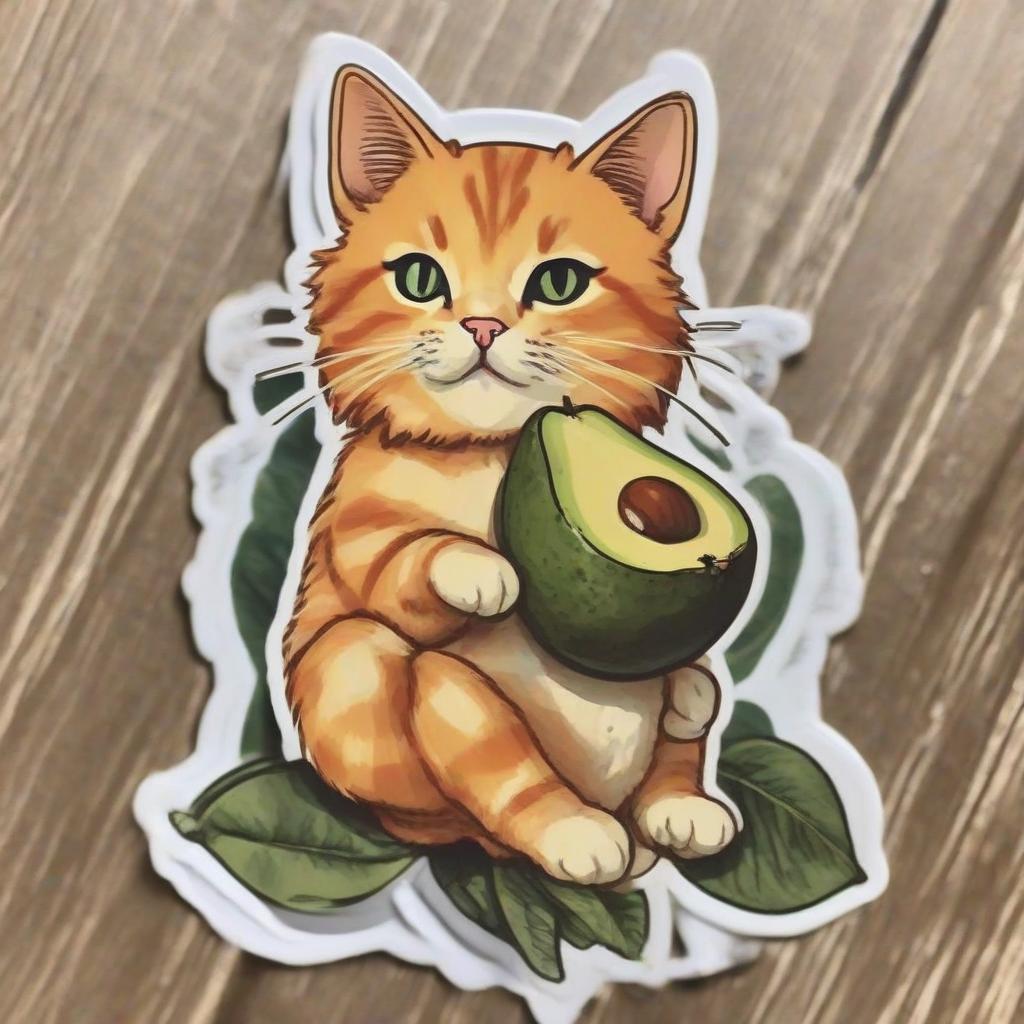}}
        \\

        \multicolumn{5}{c}{\textit{``a sticker of a ginger cat''}}
        \\
        \\

        {\includegraphics[valign=c, width=\ww]{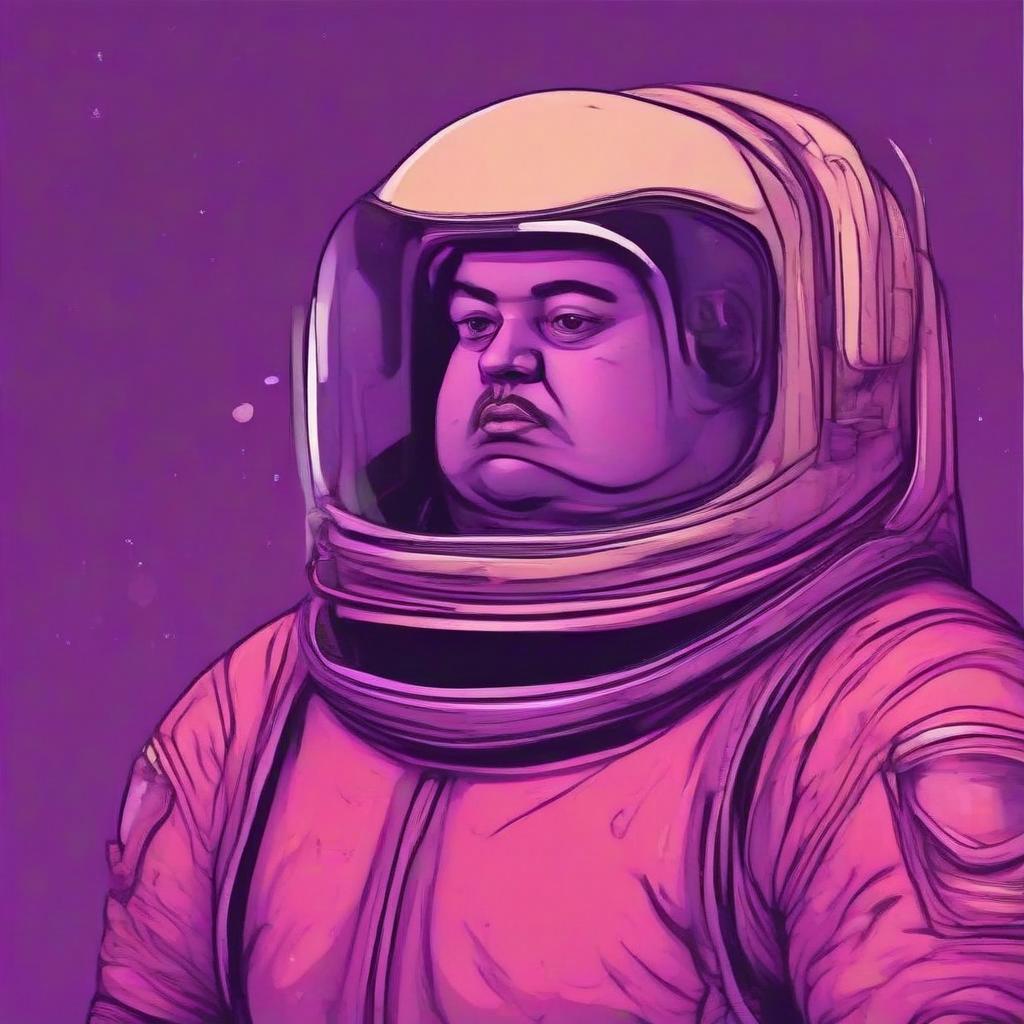}} &
        {\includegraphics[valign=c, width=\ww]{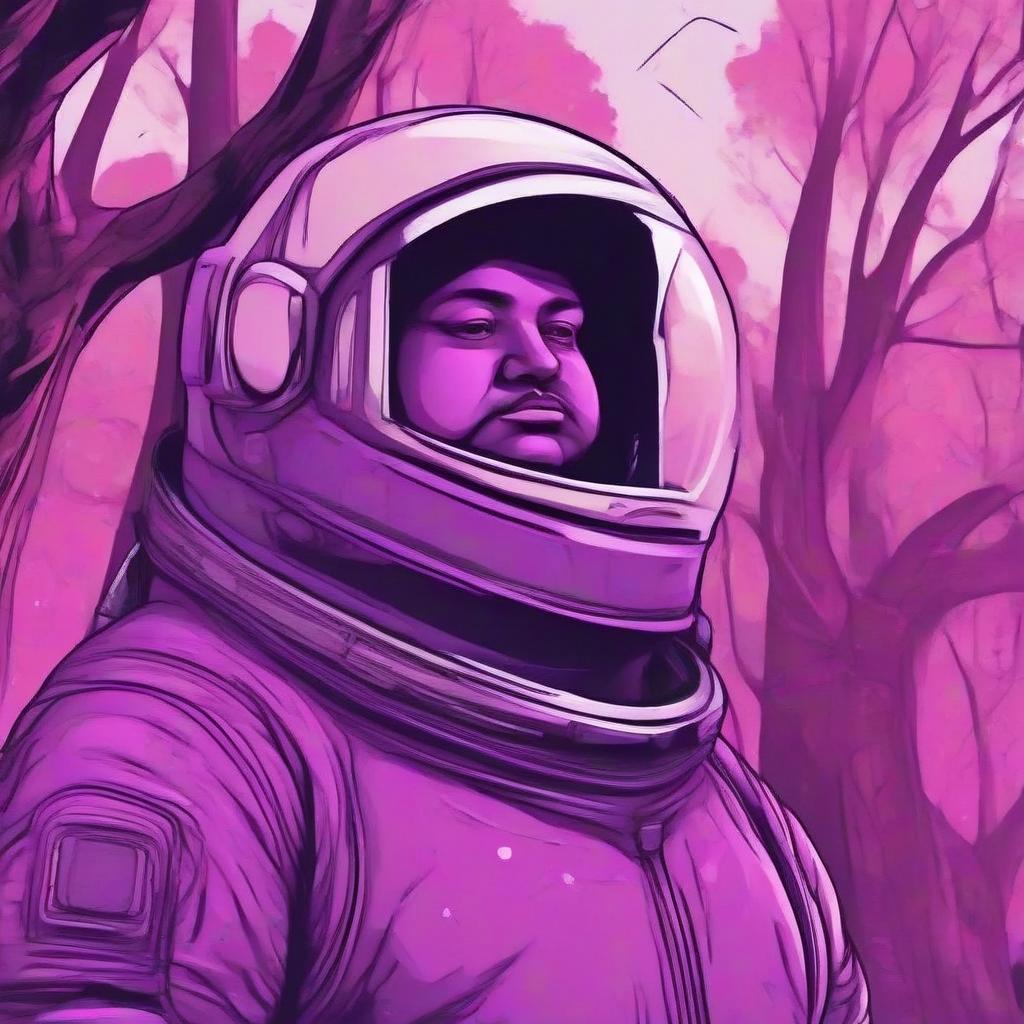}} &
        {\includegraphics[valign=c, width=\ww]{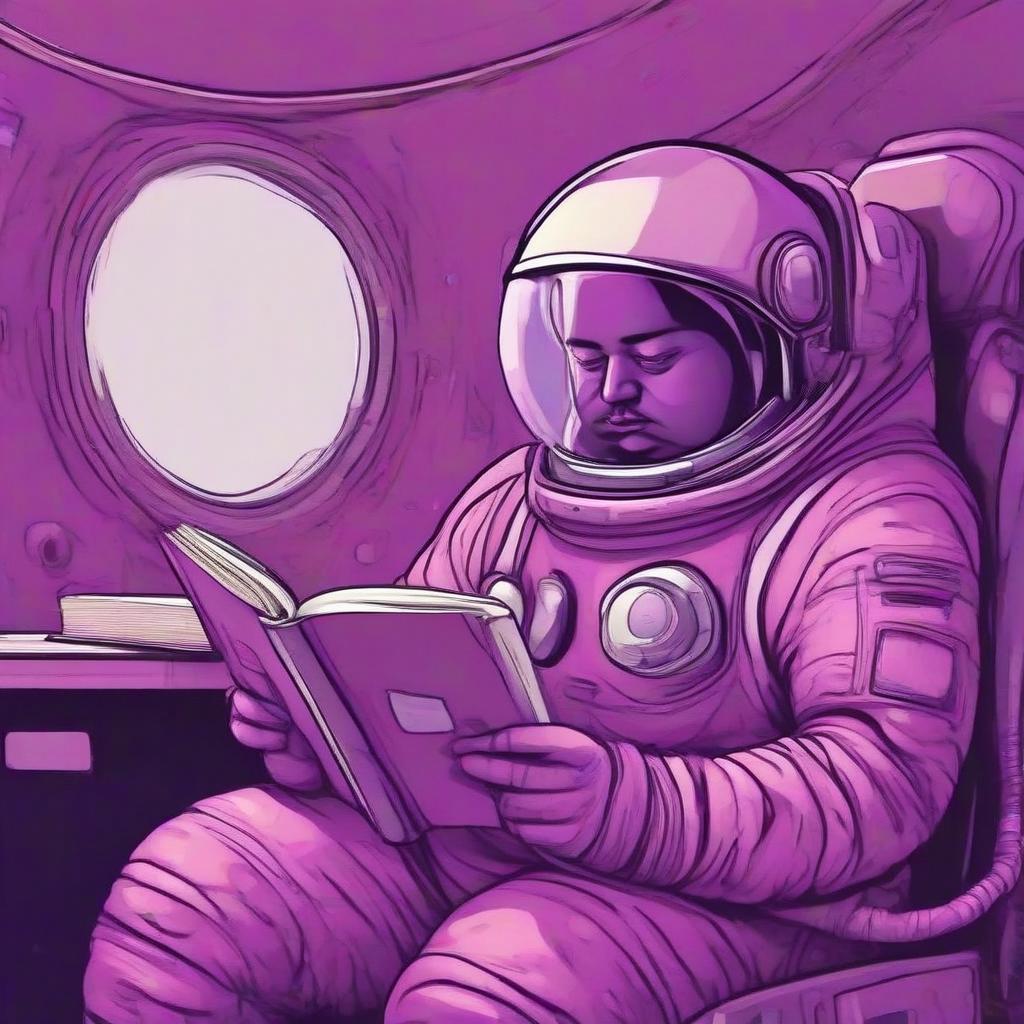}} &
        {\includegraphics[valign=c, width=\ww]{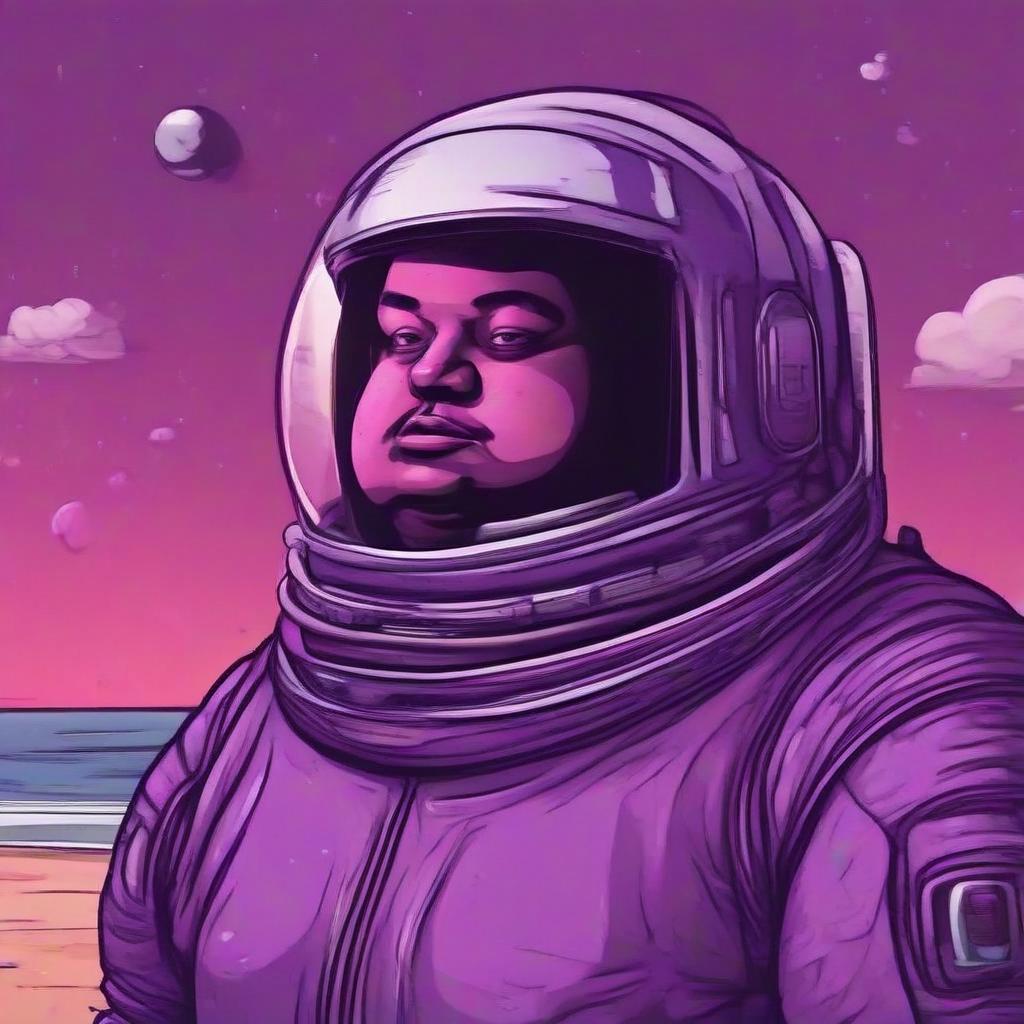}} &
        {\includegraphics[valign=c, width=\ww]{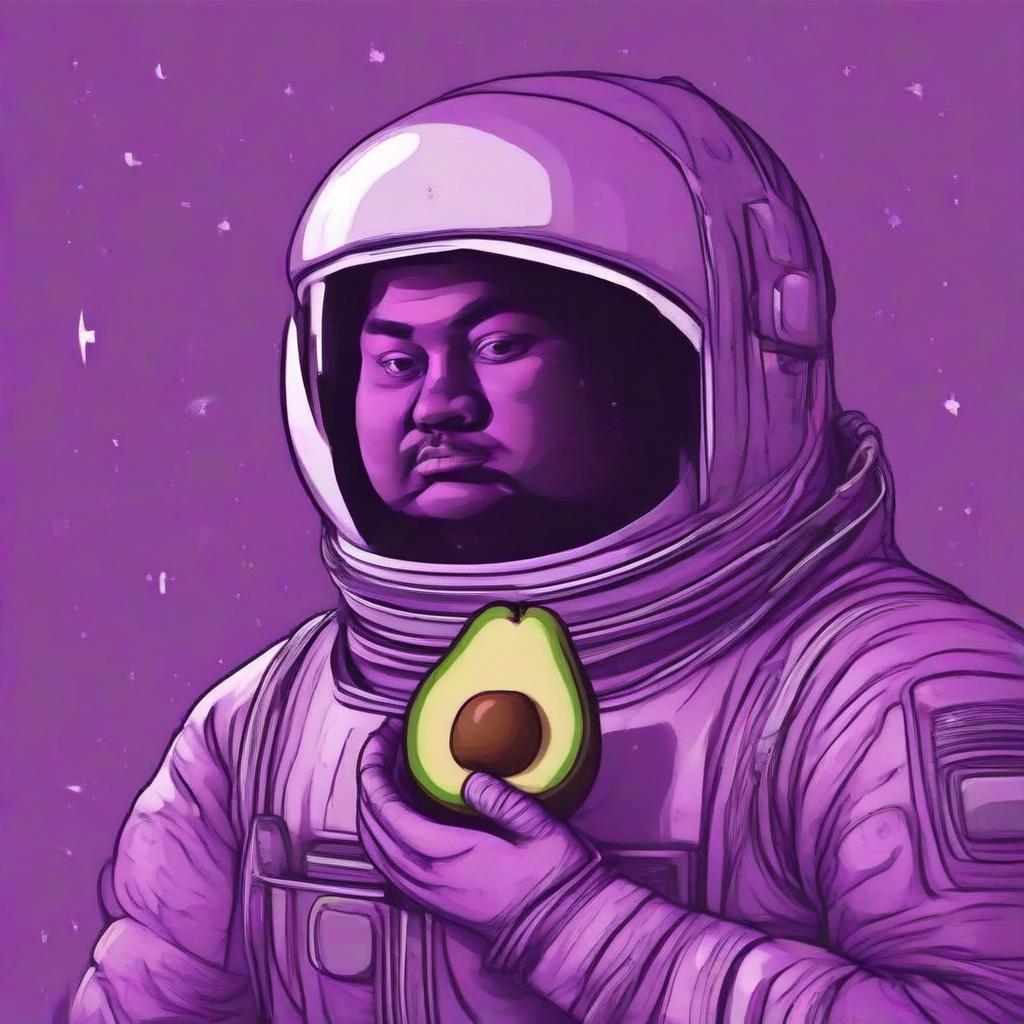}}
        \\

        \multicolumn{5}{c}{\textit{``a purple astronaut, digital art, smooth, sharp focus, vector art''}}

    \end{tabular}
    
    \caption{\textbf{Additional results.} Our method is able to consistently generate different types and styles of characters, \eg, paintings, animations, stickers and vector art.}
    \label{fig:extended_teaser}
\end{figure*}

%% file: figures/life_story/fig.tex
\begin{figure*}[t]
    \centering
    \setlength{\tabcolsep}{1.0pt}
    \renewcommand{\arraystretch}{1.0}
    \setlength{\ww}{0.37\columnwidth}
    \begin{tabular}{ccccc}
        {\includegraphics[valign=c, width=\ww]{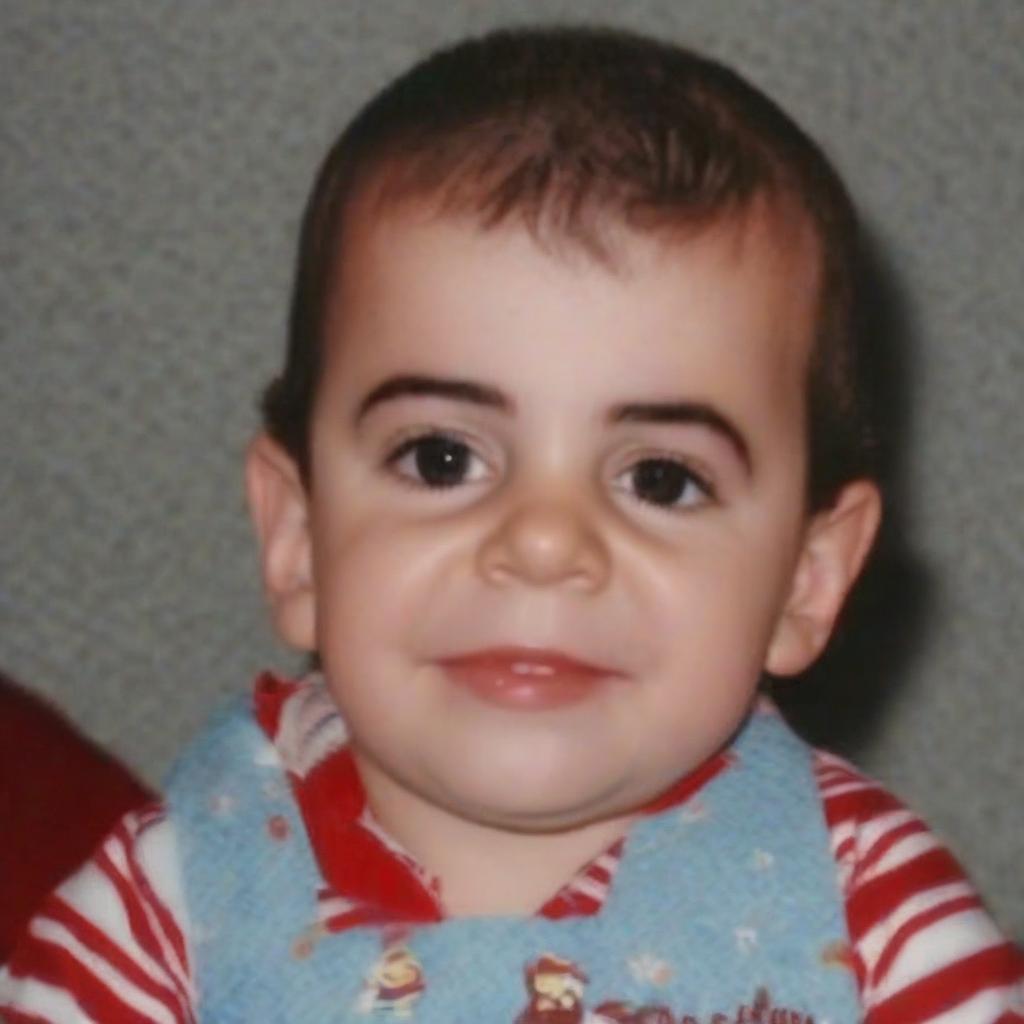}} &
        {\includegraphics[valign=c, width=\ww]{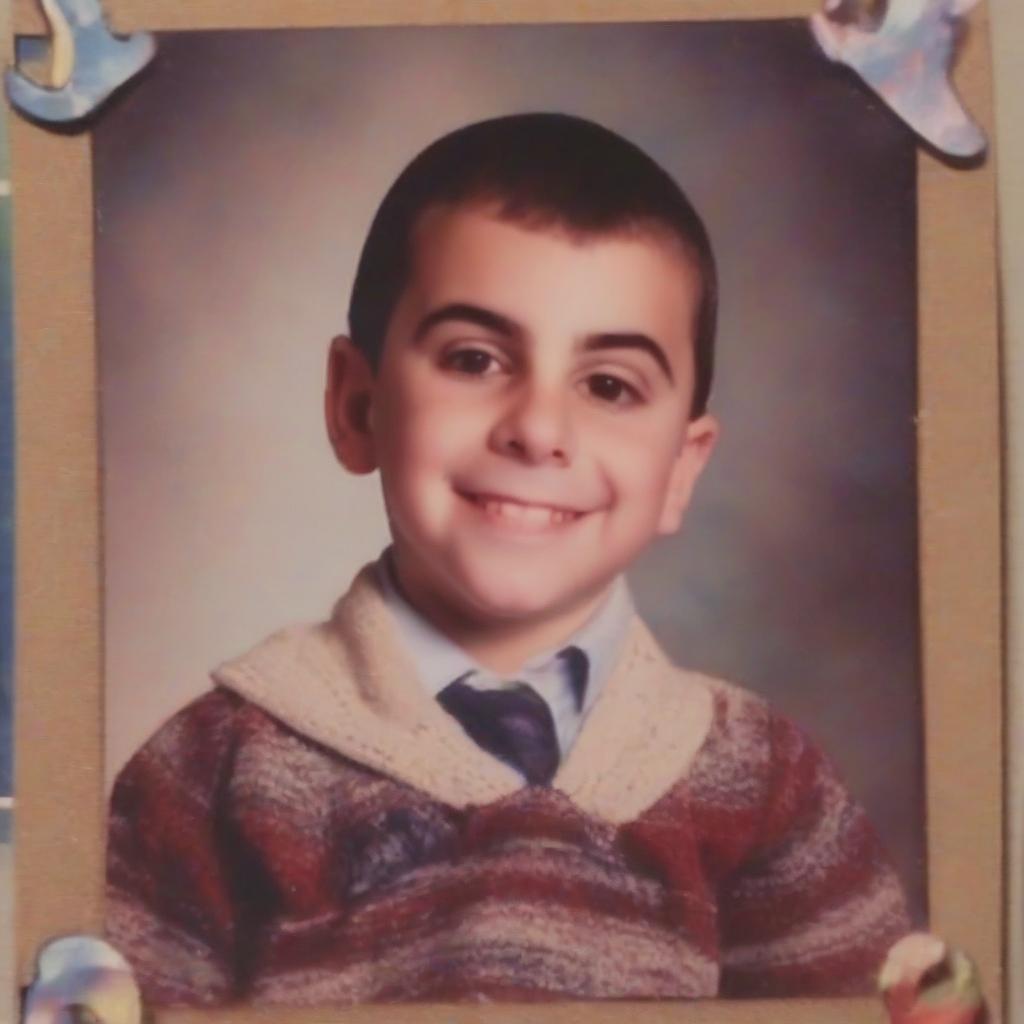}} &
        {\includegraphics[valign=c, width=\ww]{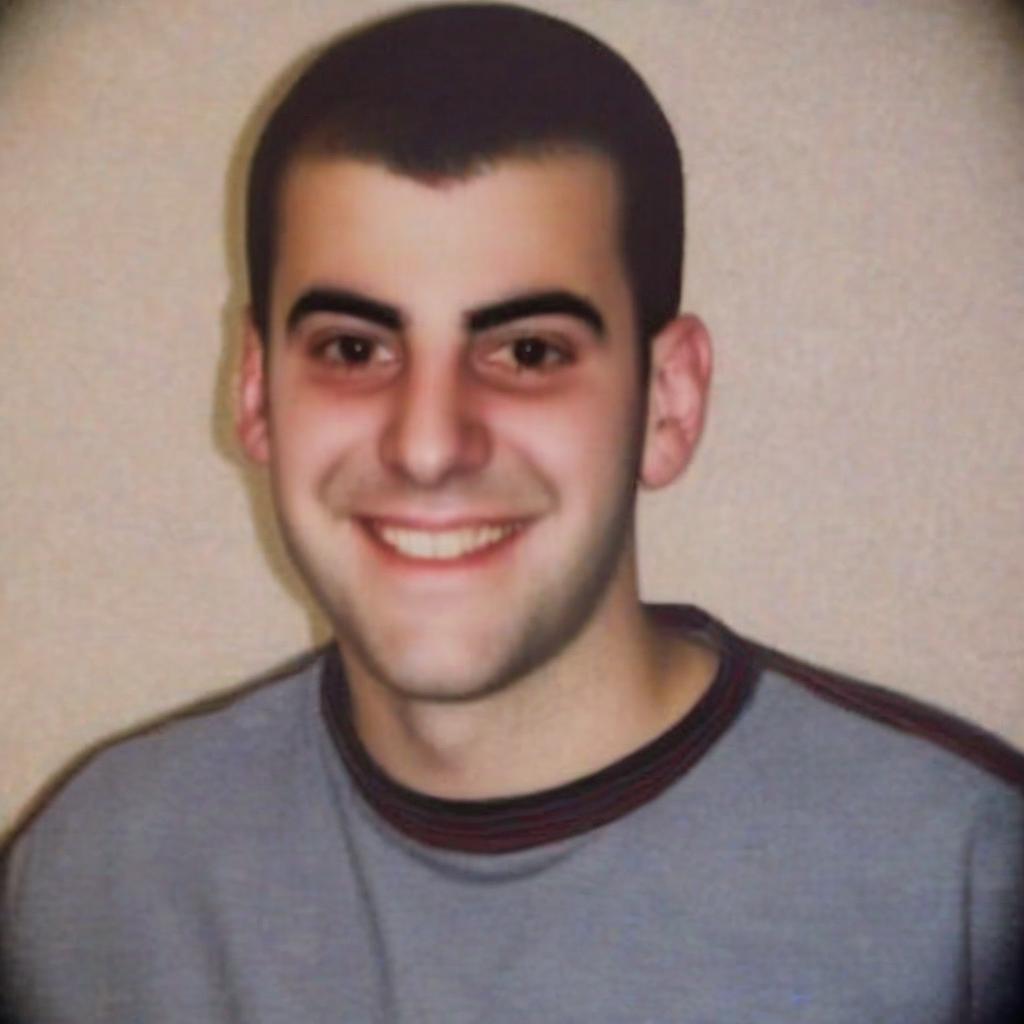}} &
        {\includegraphics[valign=c, width=\ww]{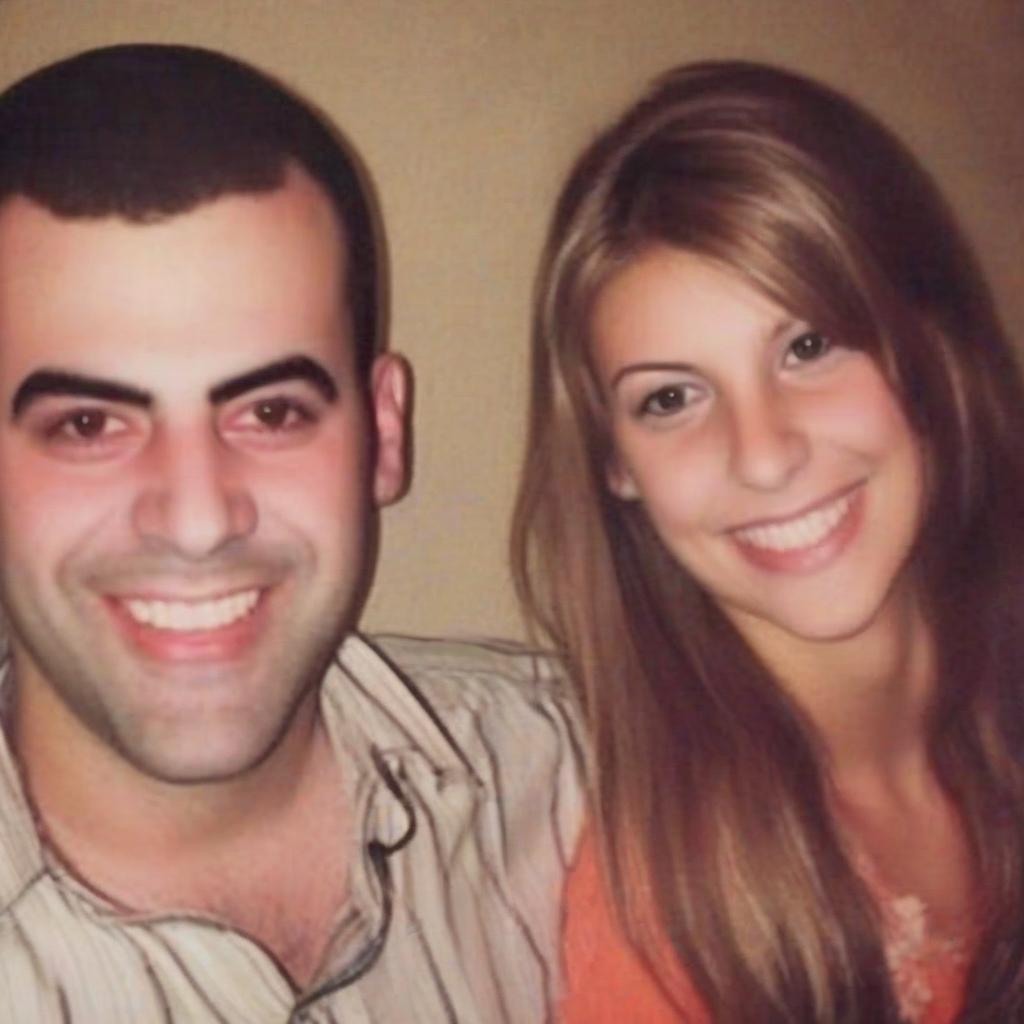}} &
        {\includegraphics[valign=c, width=\ww]{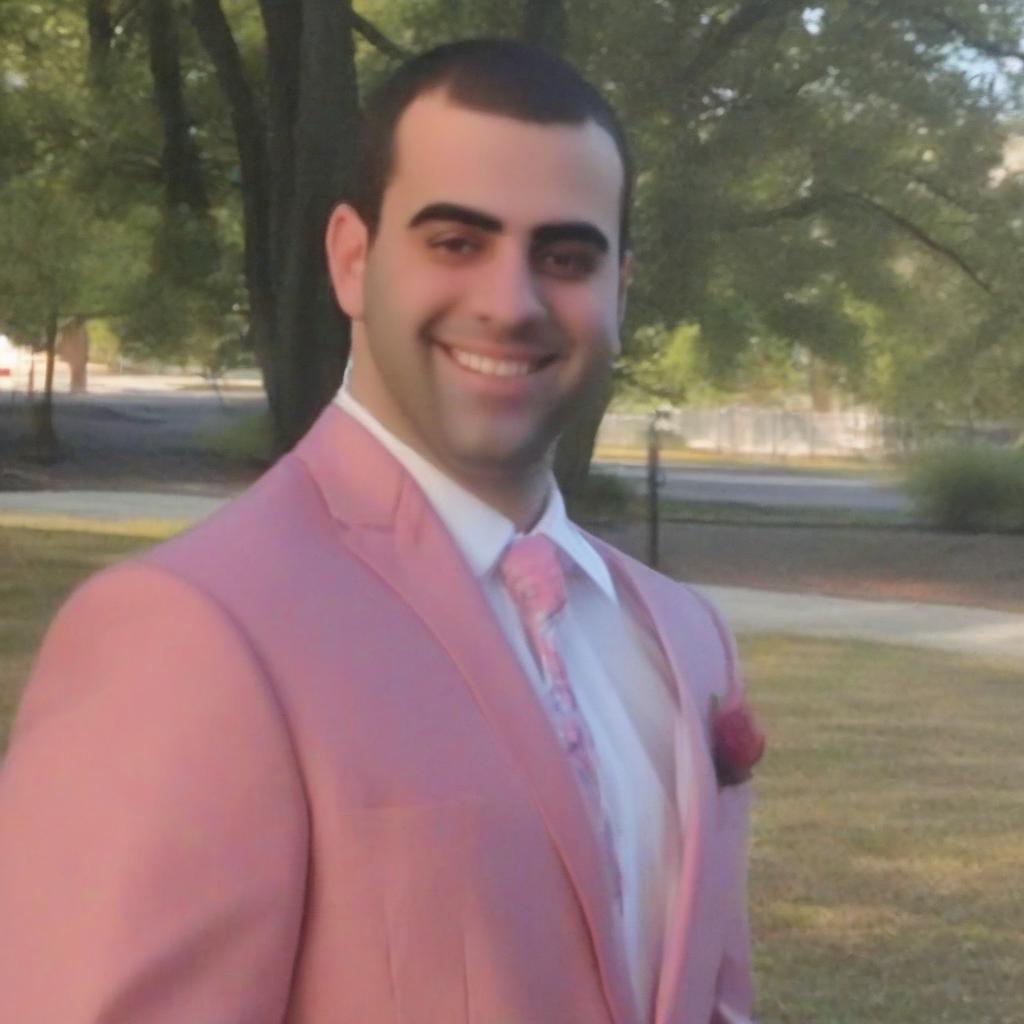}}
        \\

        \textit{``as a baby''} &
        \textit{``as a small child''} &
        \textit{``as a teenager''} &
        \textit{``with his} &
        \textit{``before the prom''}
        \\
        &
        &
        &
        \textit{first girlfriend''}
        \\
        \\

        {\includegraphics[valign=c, width=\ww]{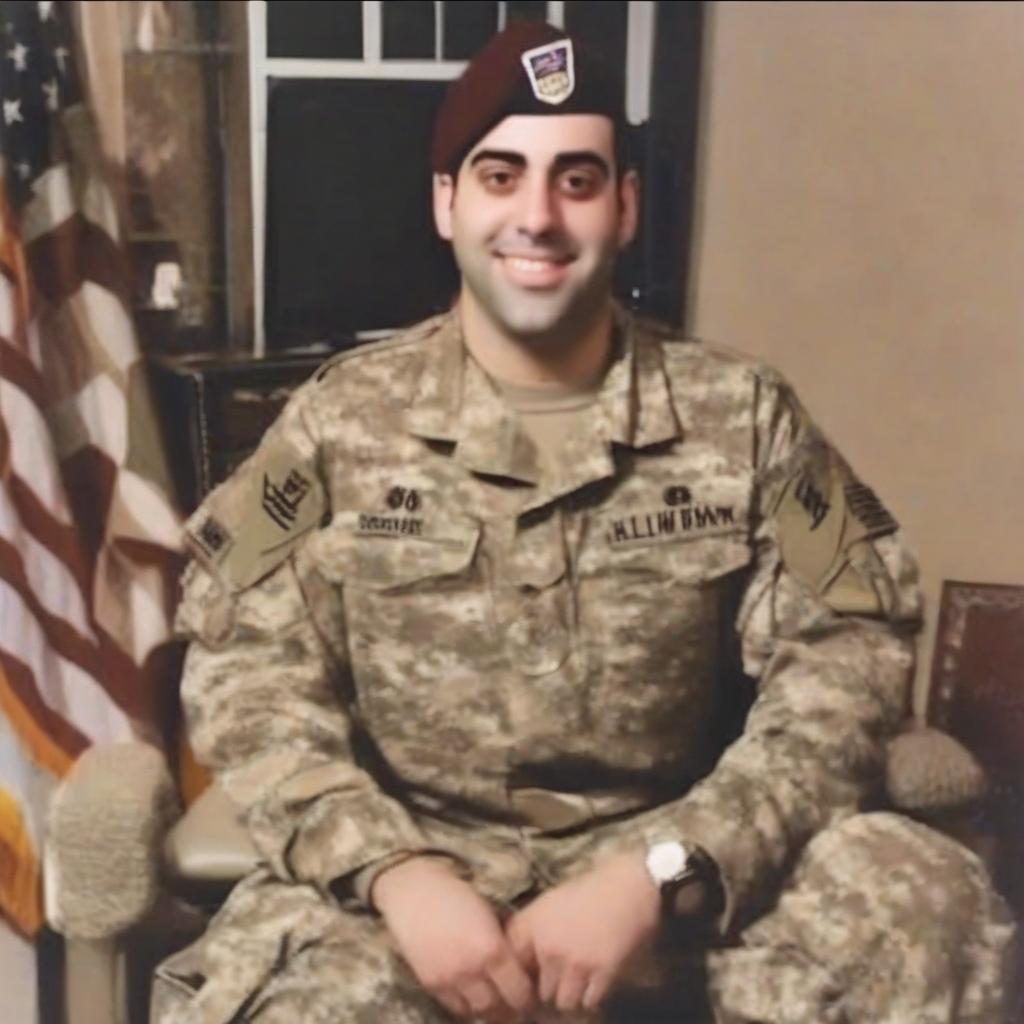}} &
        {\includegraphics[valign=c, width=\ww]{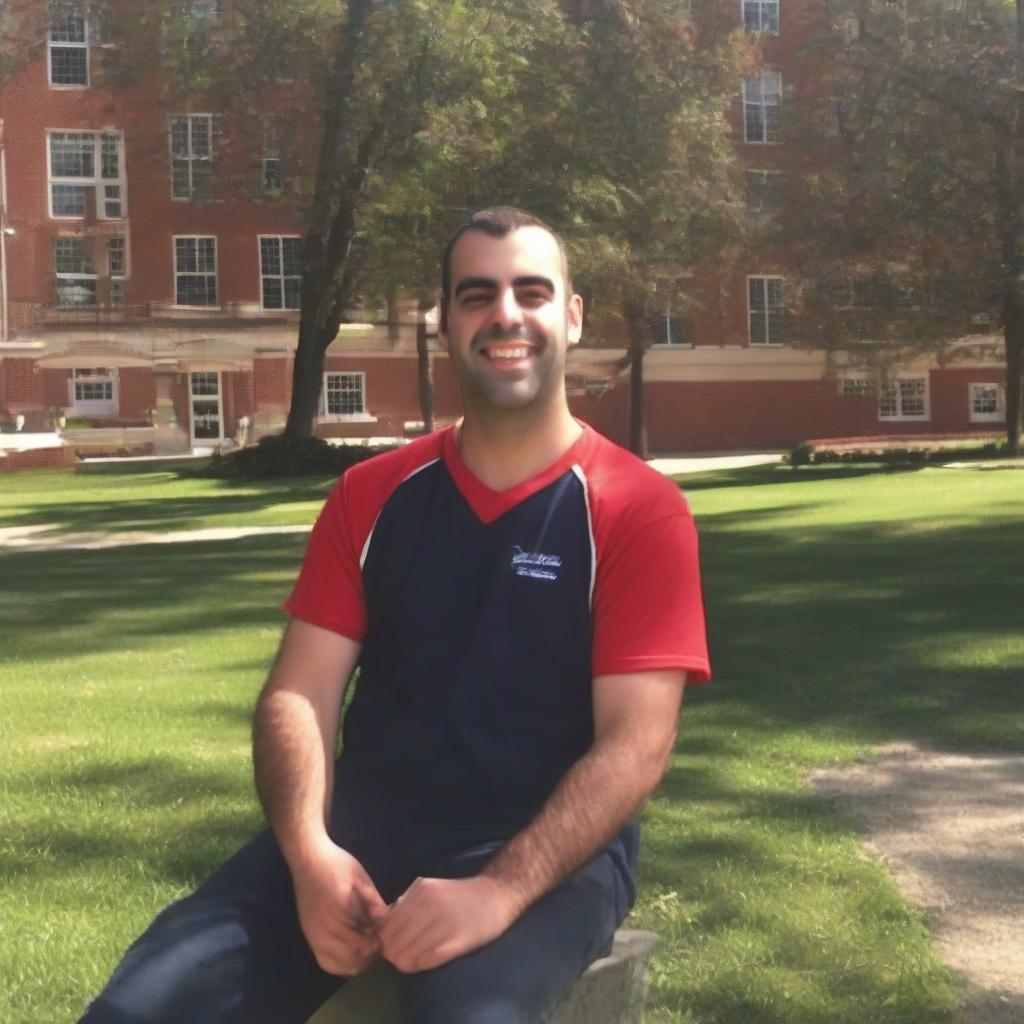}} &
        {\includegraphics[valign=c, width=\ww]{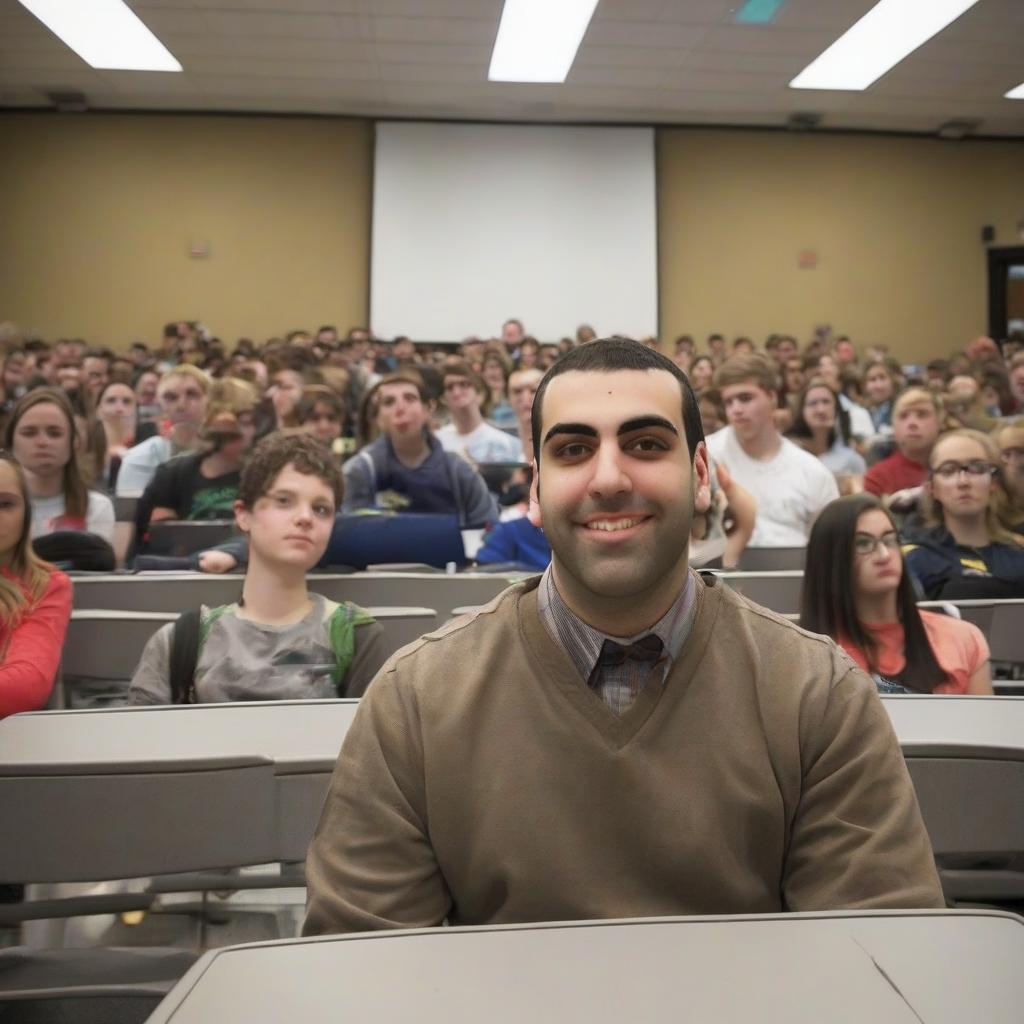}} &
        {\includegraphics[valign=c, width=\ww]{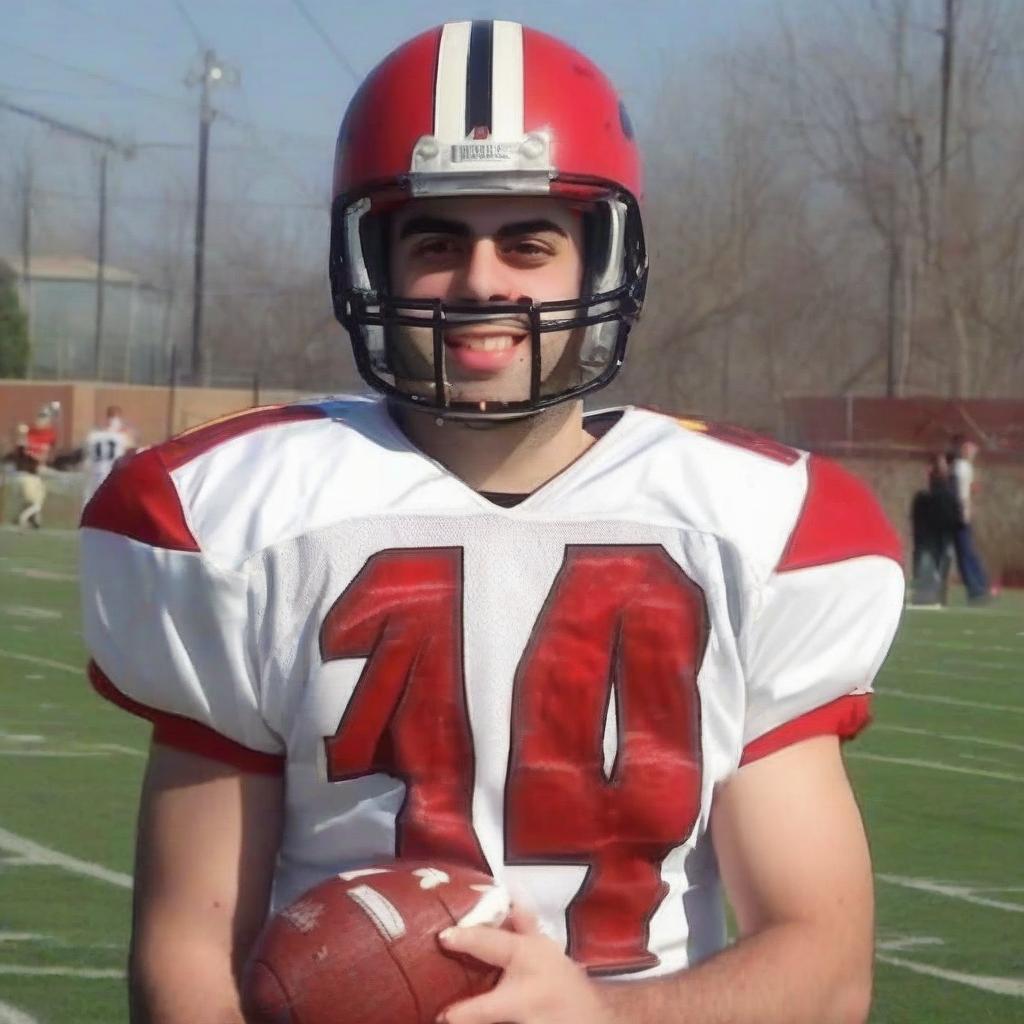}} &
        {\includegraphics[valign=c, width=\ww]{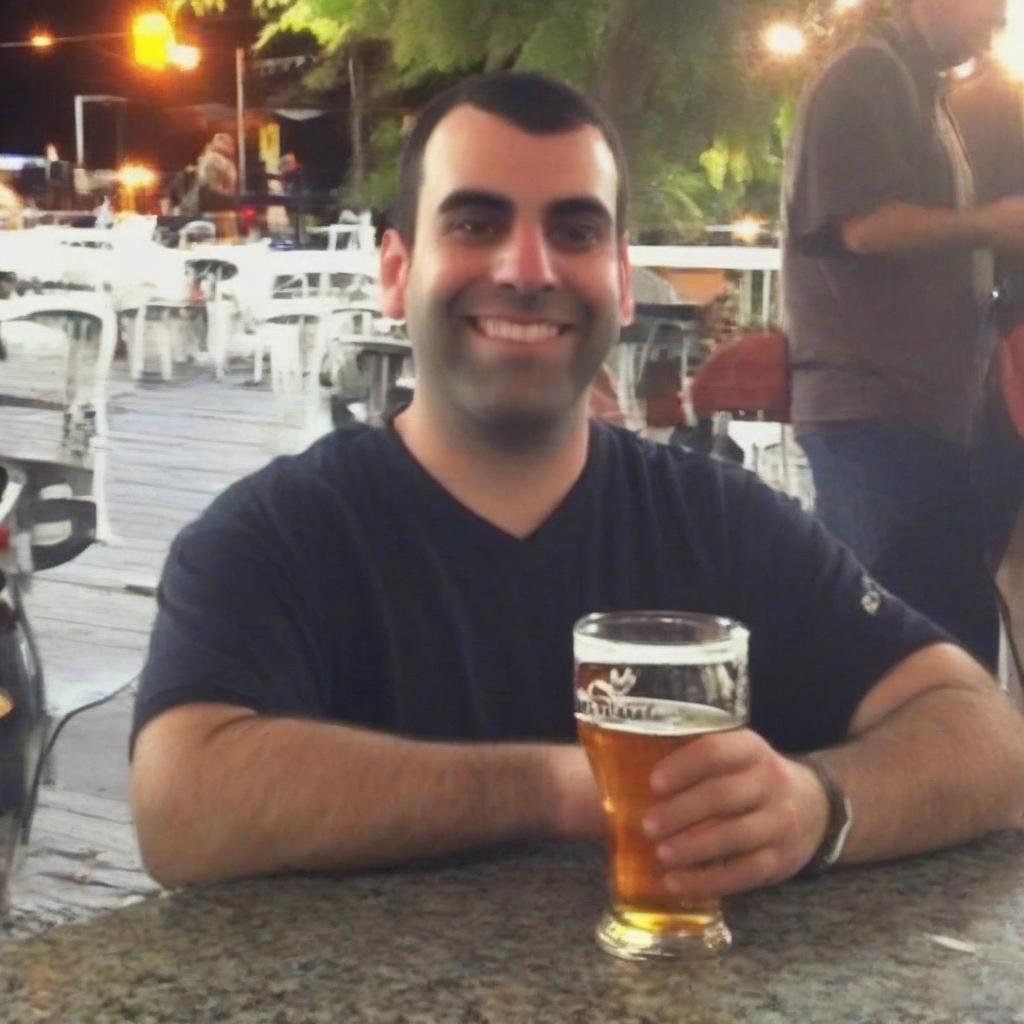}}
        \\

        \textit{``as a soldier''} &
        \textit{``in the} &
        \textit{``sitting in a lecture''} &
        \textit{``playing football''} &
        \textit{``drinking a beer''}
        \\
        &
        \textit{college campus''} &
        \\
        \\

        {\includegraphics[valign=c, width=\ww]{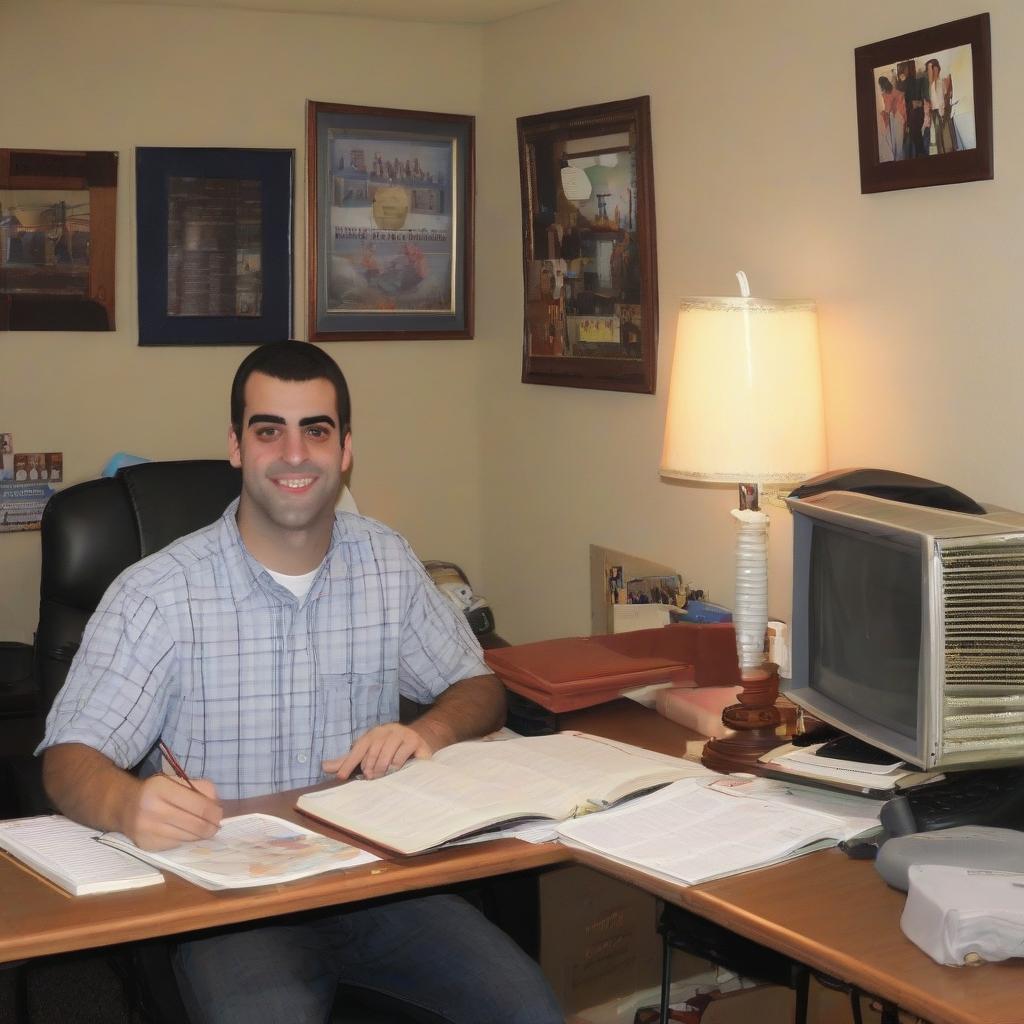}} &
        {\includegraphics[valign=c, width=\ww]{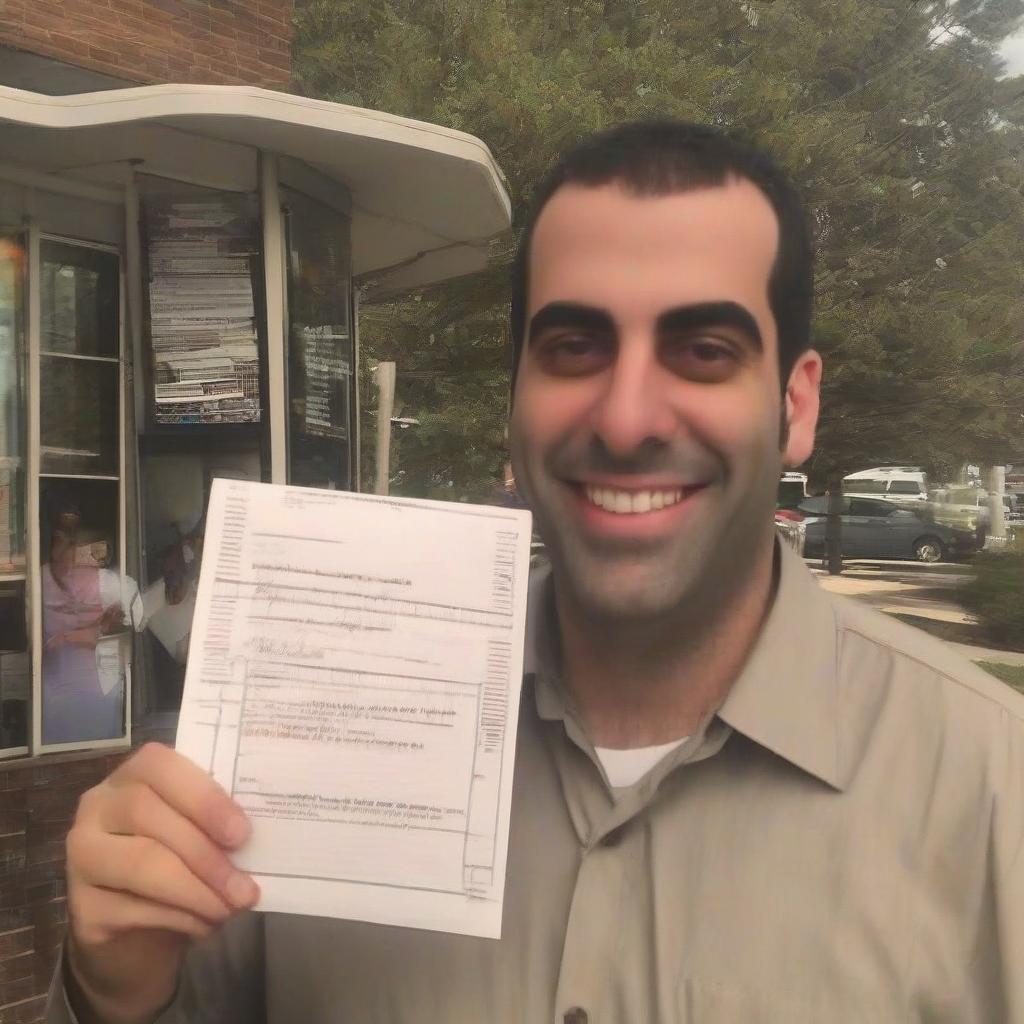}} &
        {\includegraphics[valign=c, width=\ww]{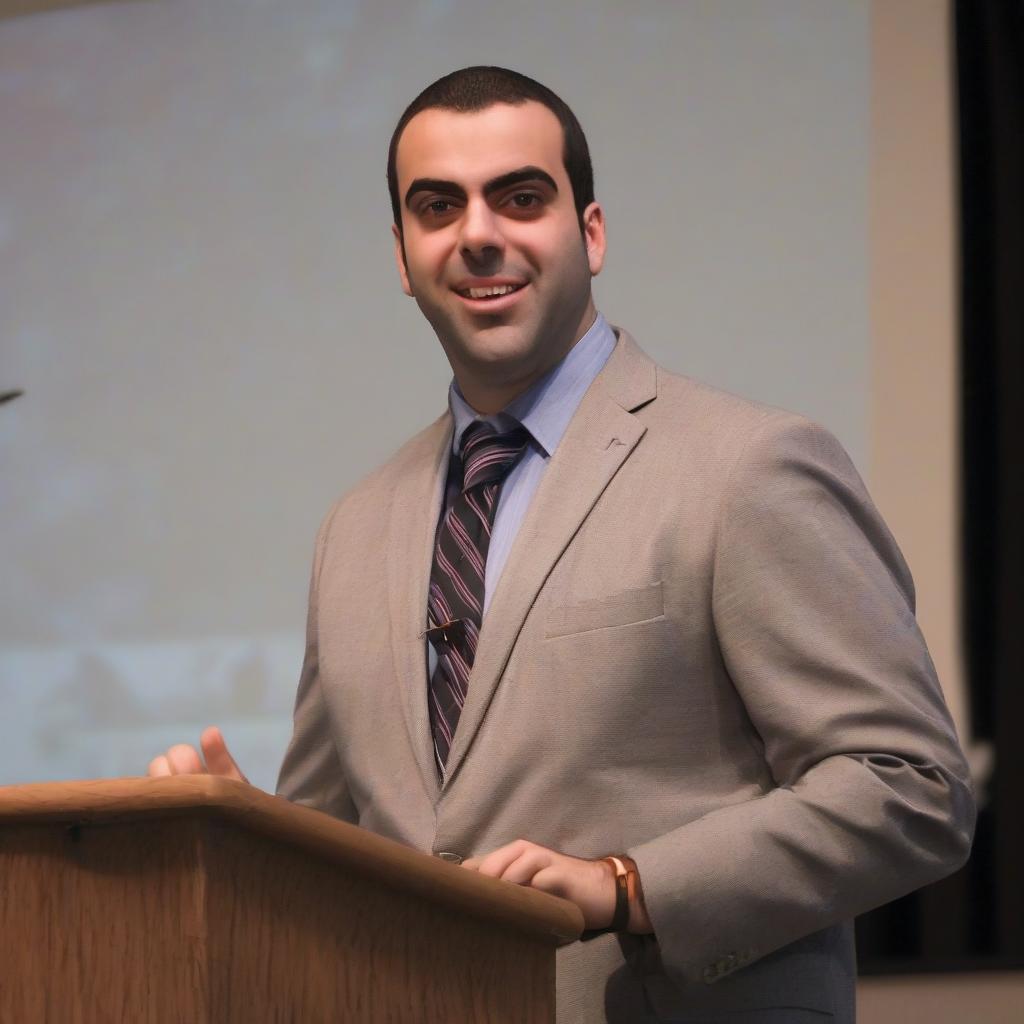}} &
        {\includegraphics[valign=c, width=\ww]{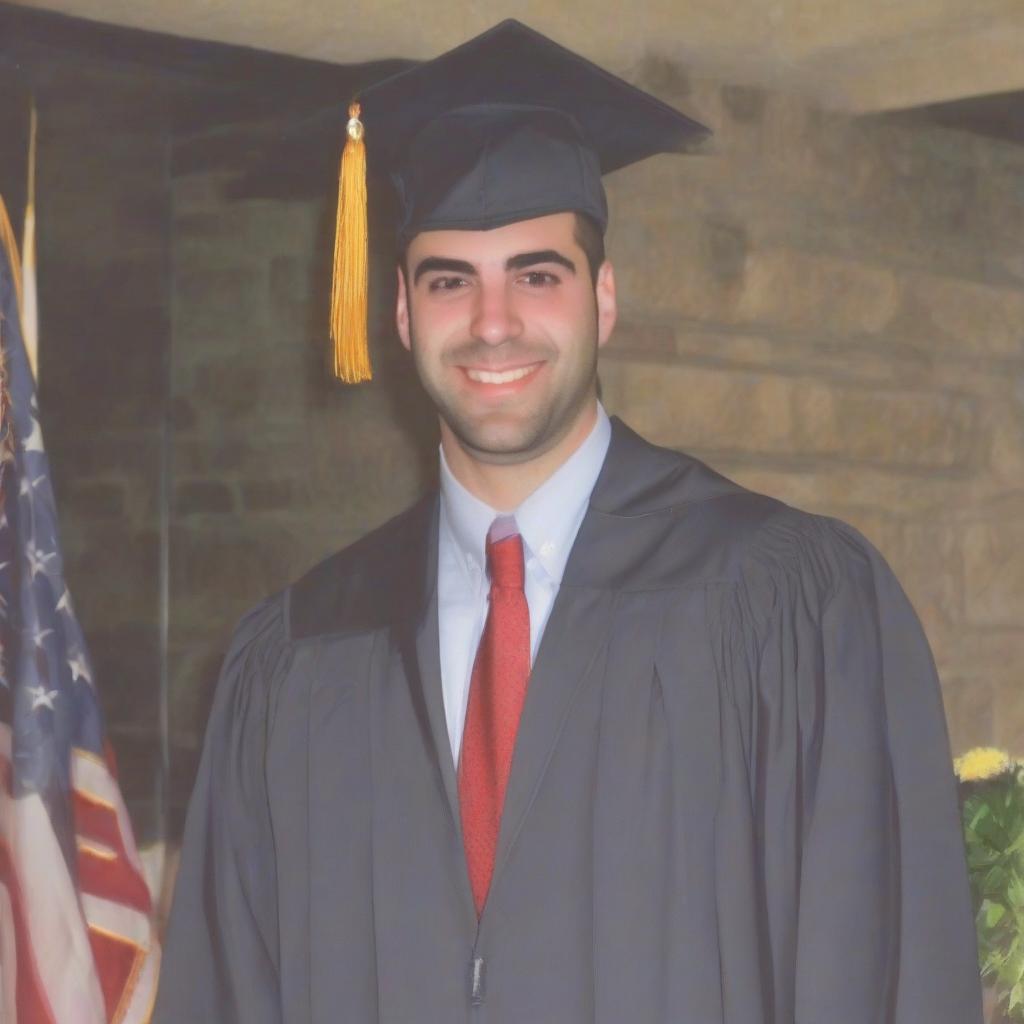}} &
        {\includegraphics[valign=c, width=\ww]{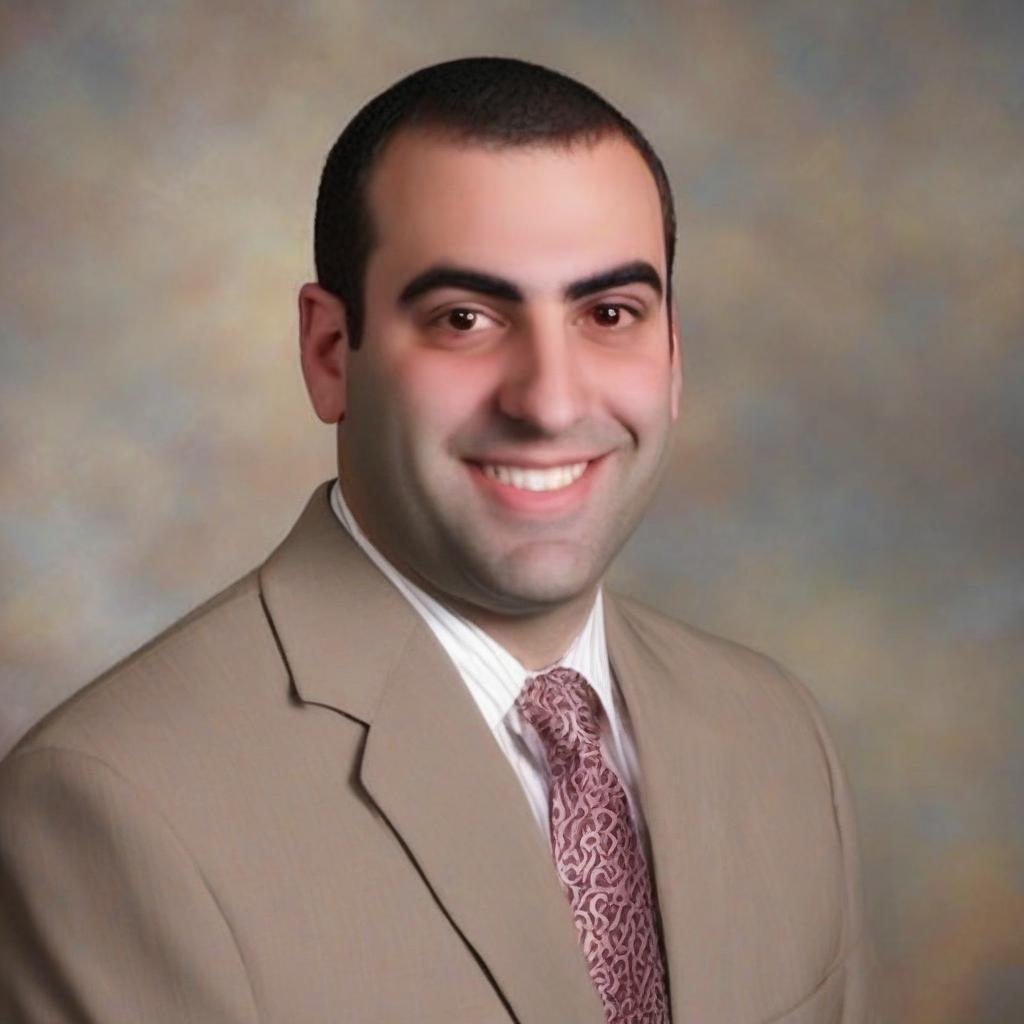}}
        \\

        \textit{``studying in} &
        \textit{``happy with his} &
        \textit{``giving a talk} &
        \textit{``graduating from} &
        \textit{``a profile picture''}
        \\
        \textit{his room''} &
        \textit{accepted paper''} &
        \textit{in a conference''} &
        \textit{college''} &
        \\
        \\

        {\includegraphics[valign=c, width=\ww]{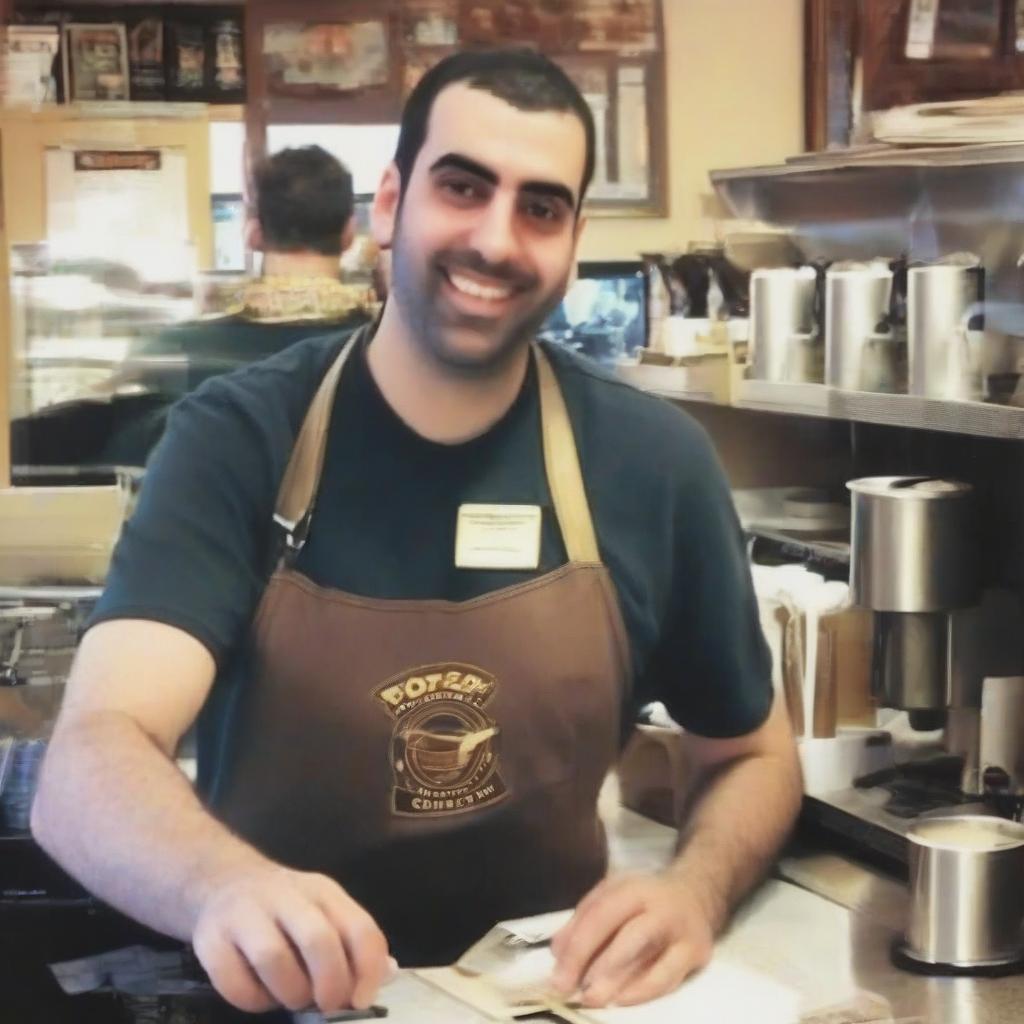}} &
        {\includegraphics[valign=c, width=\ww]{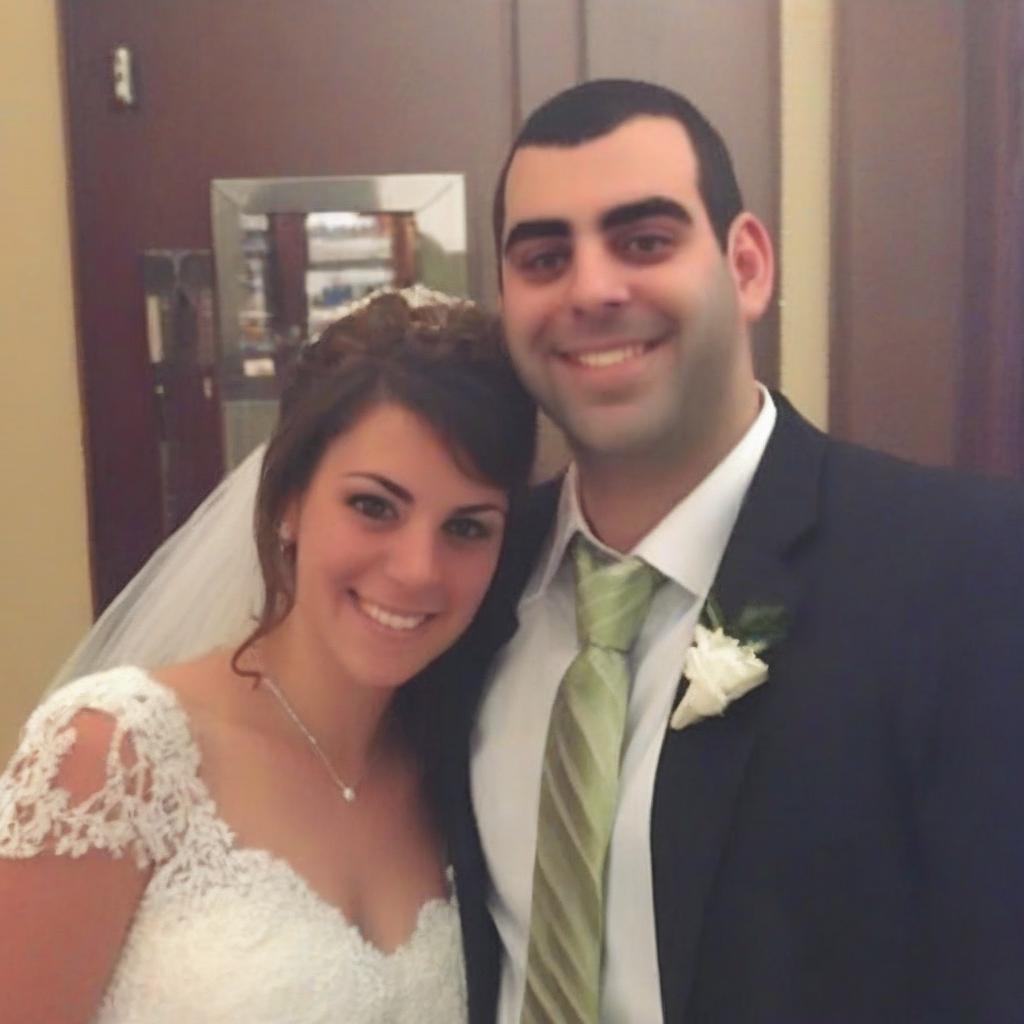}} &
        {\includegraphics[valign=c, width=\ww]{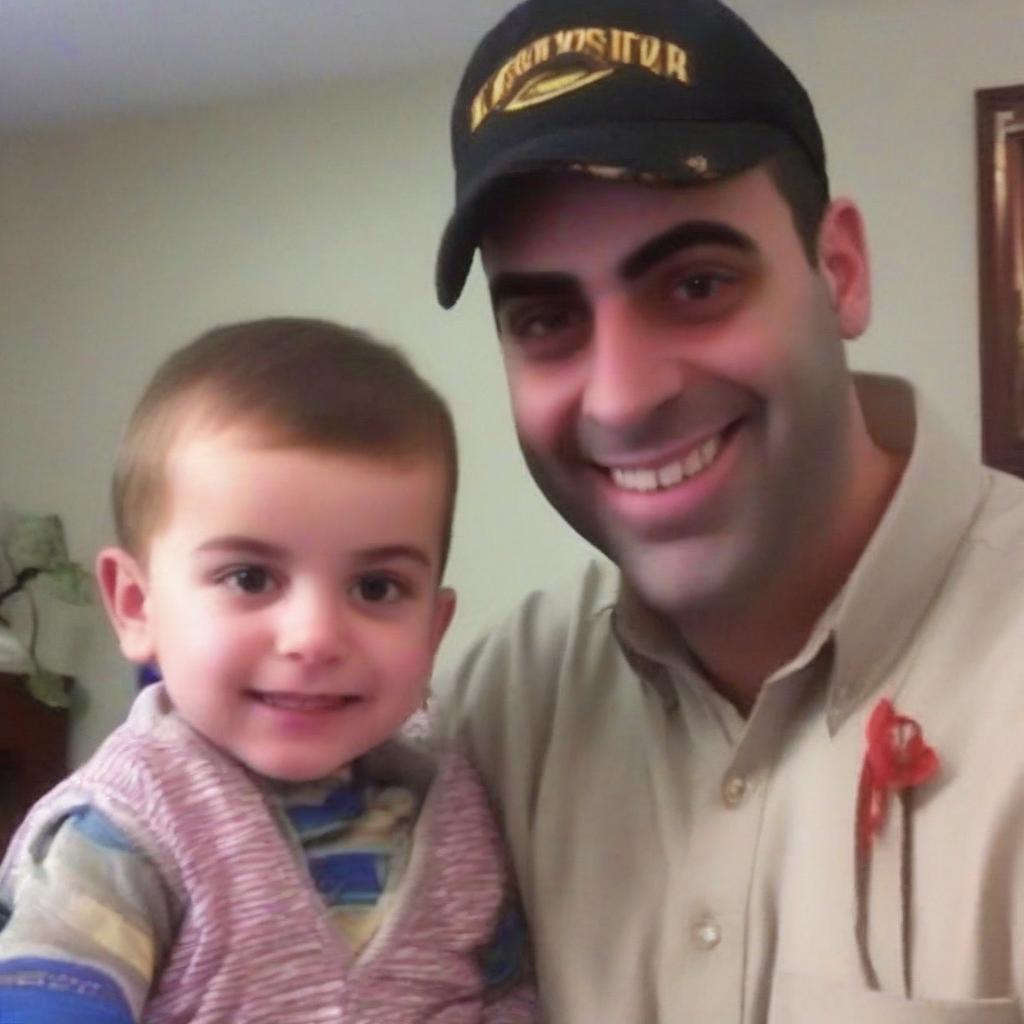}} &
        {\includegraphics[valign=c, width=\ww]{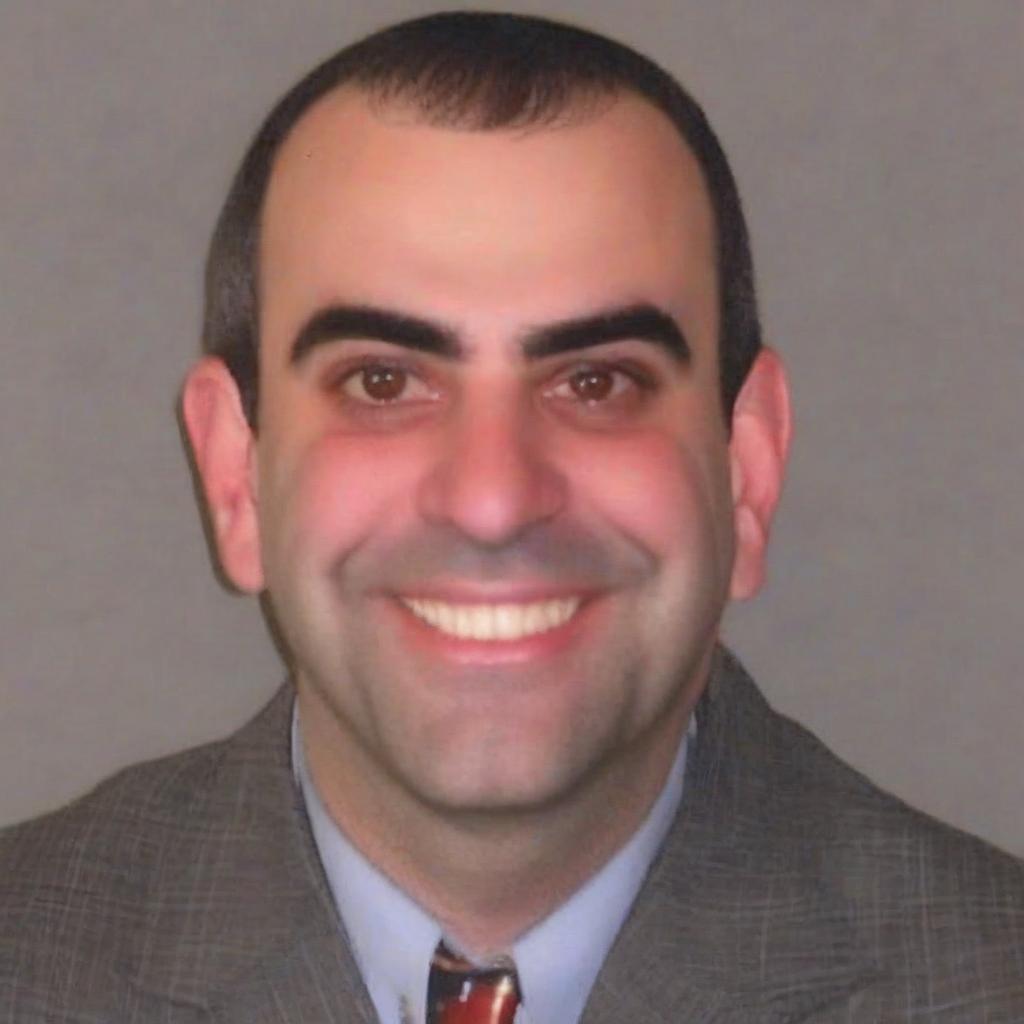}} &
        {\includegraphics[valign=c, width=\ww]{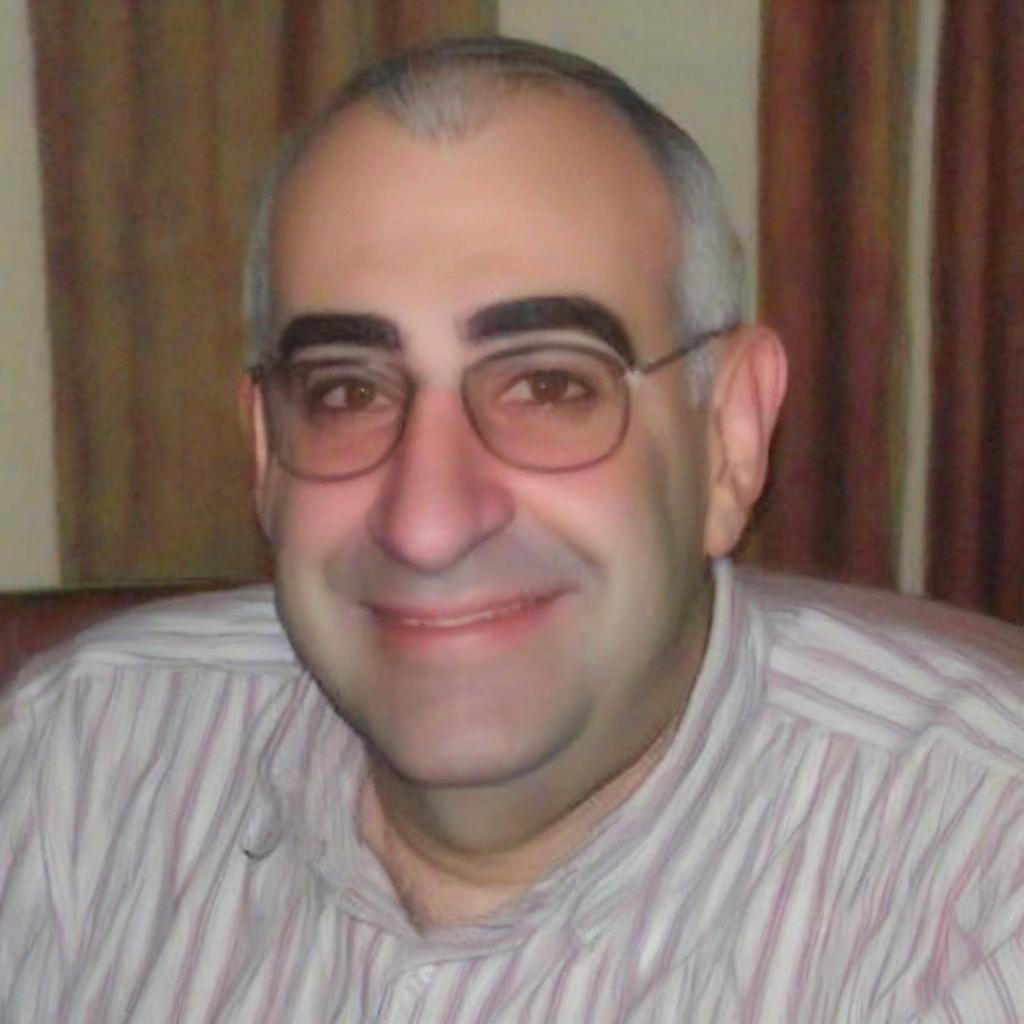}}
        \\

        \textit{``working in a} &
        \textit{``in his wedding''} &
        \textit{``with his} &
        \textit{``as a 50} &
        \textit{``as a 70}
        \\
        \textit{coffee shop''} &
        &
        \textit{small  child''} &
        \textit{years old man''} &
        \textit{years old man''}
        \\
        \\

        {\includegraphics[valign=c, width=\ww]{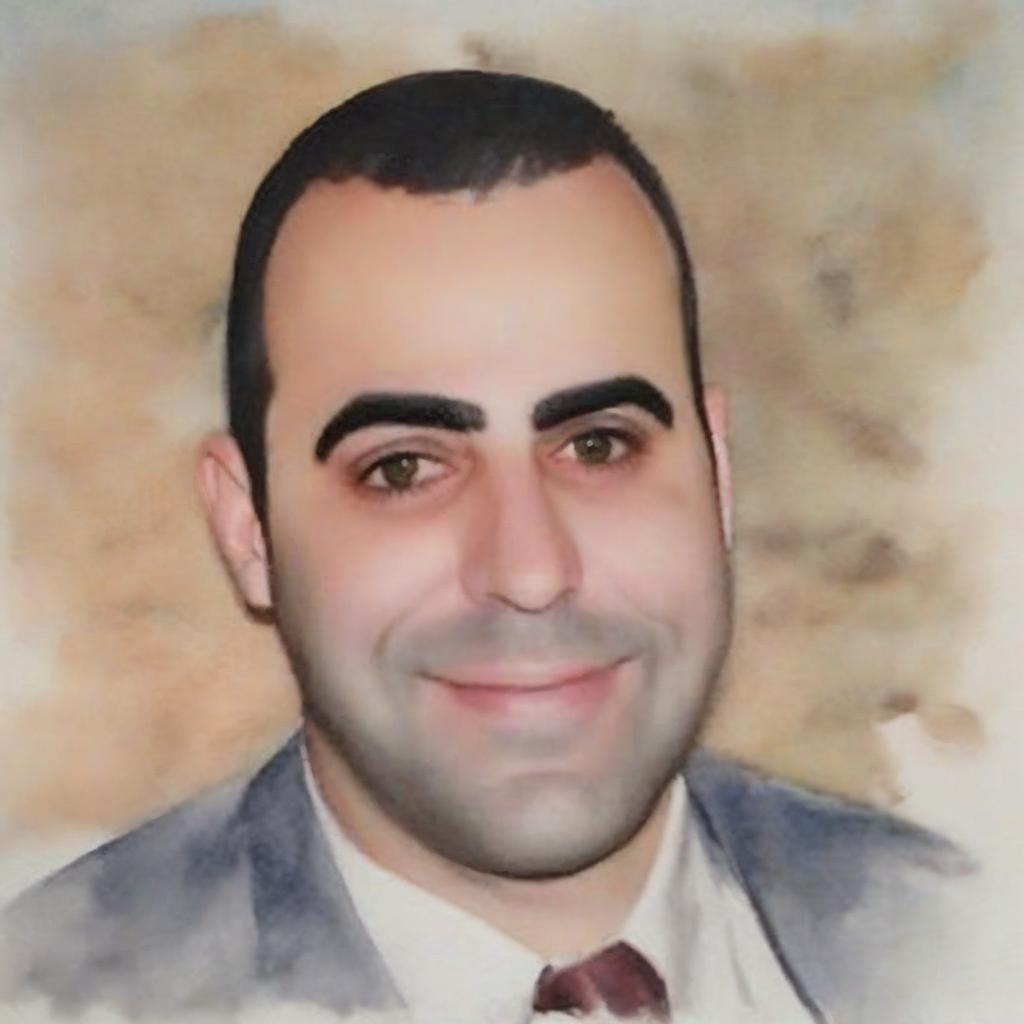}} &
        {\includegraphics[valign=c, width=\ww]{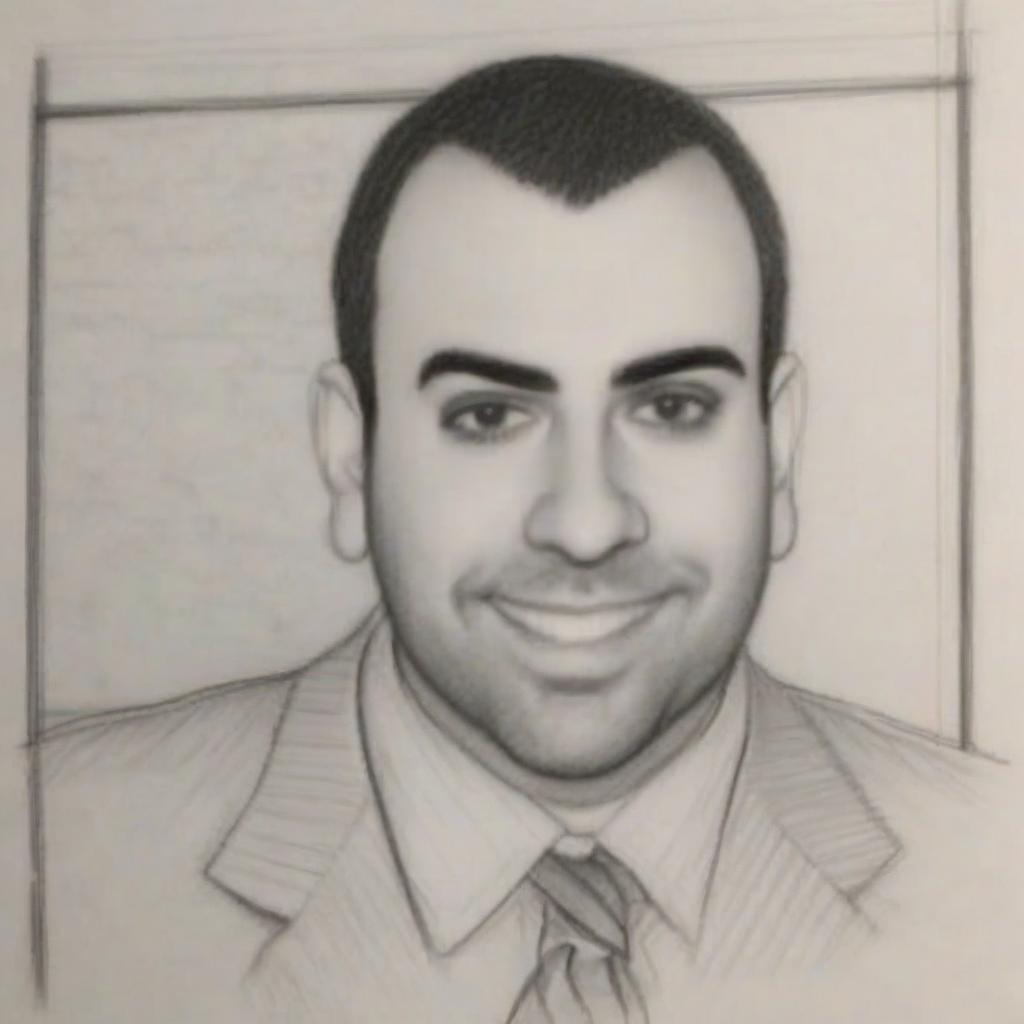}} &
        {\includegraphics[valign=c, width=\ww]{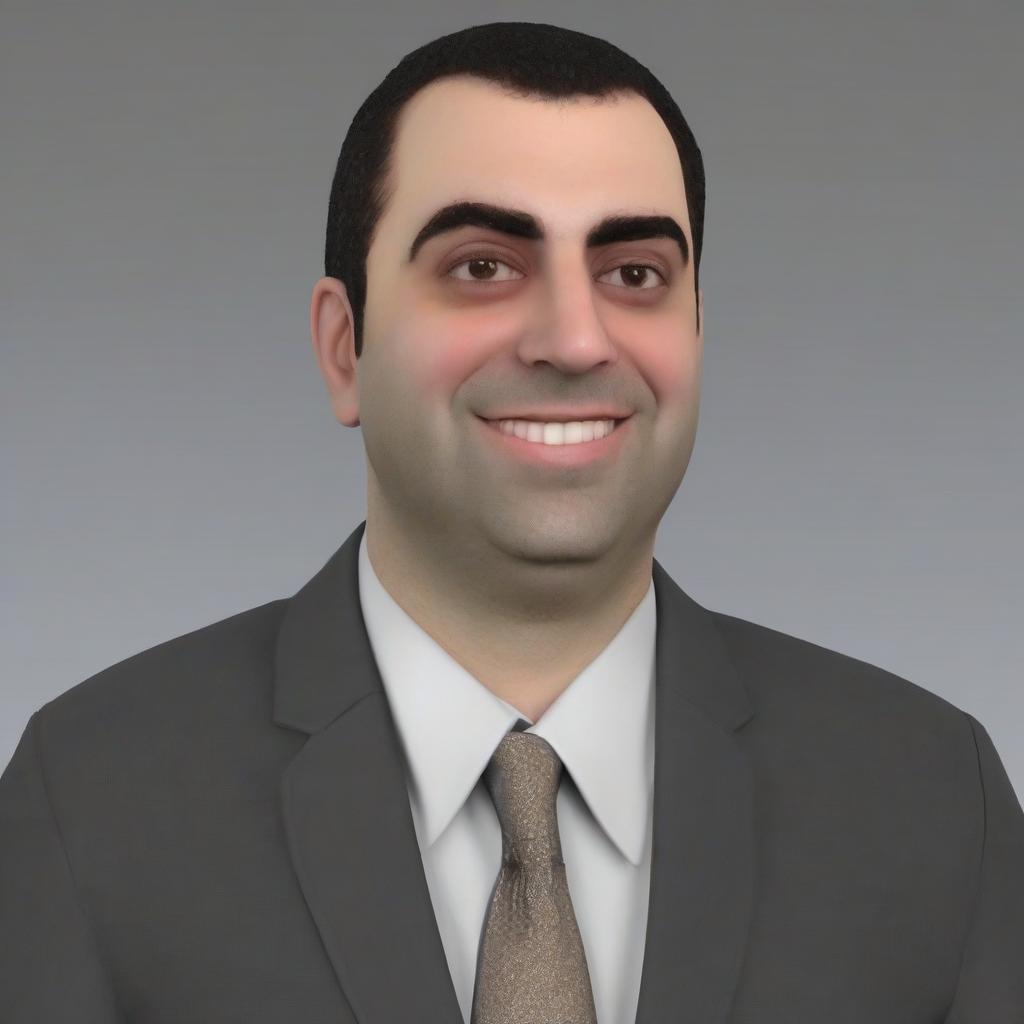}} &
        {\includegraphics[valign=c, width=\ww]{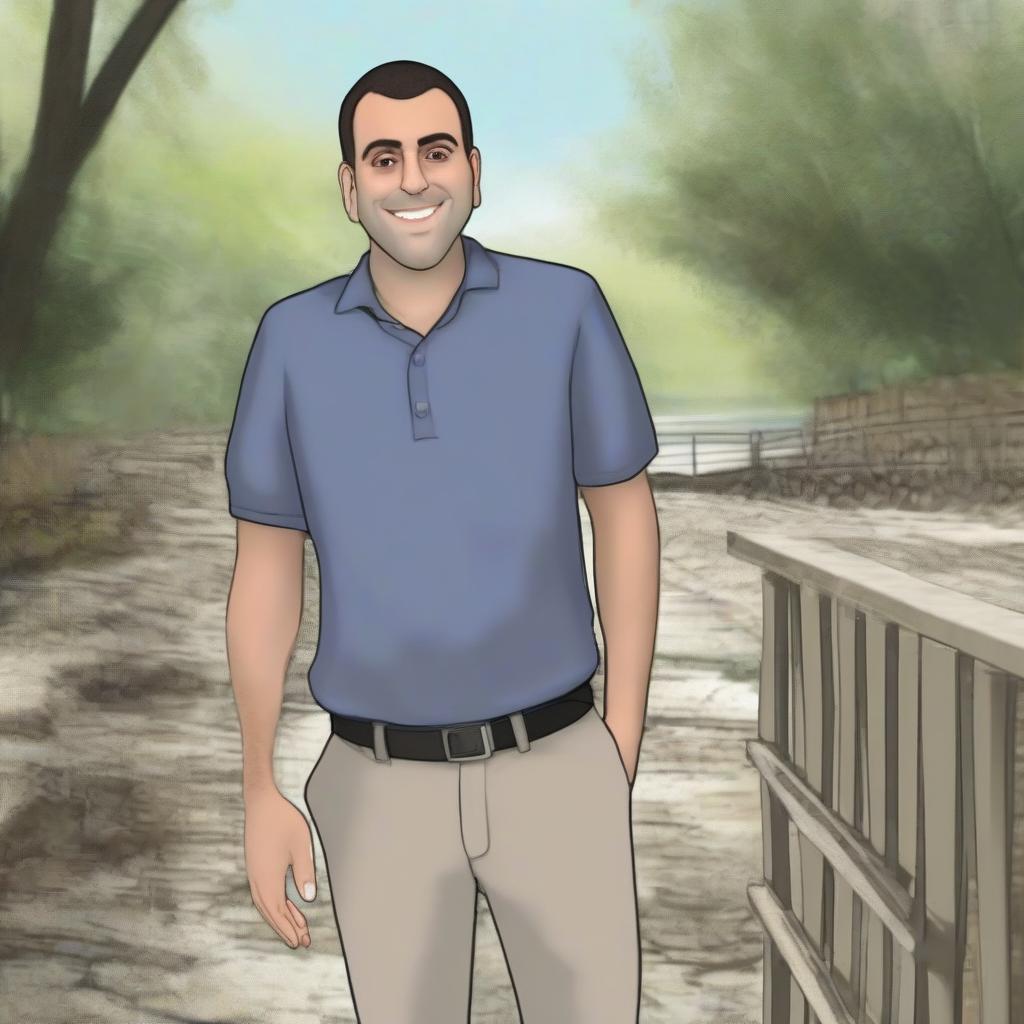}} &
        {\includegraphics[valign=c, width=\ww]{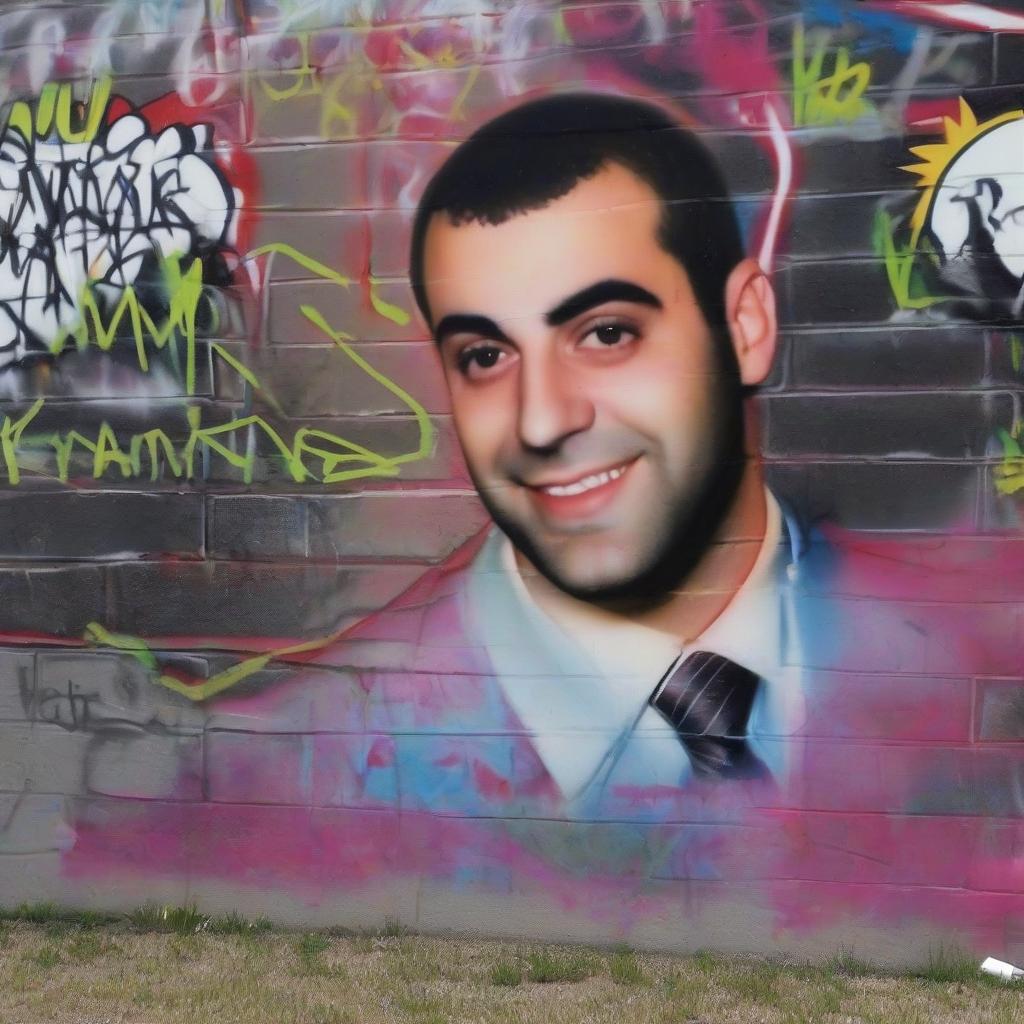}}
        \\

        \textit{``a watercolor} &
        \textit{``a pencil sketch''} &
        \textit{``a rendered avatar''} &
        \textit{``a 2D animation''} &
        \textit{``a graffiti''}
        \\
        \textit{painting''} &

    \end{tabular}
    
    \caption{\textbf{Life story.} Given a text prompt describing a fictional character, \textit{``a photo of a man with short black hair''}, we can generate a consistent life story for that character, demonstrating the applicability of our method for story generation.}
    \label{fig:life_story}
\end{figure*}

%% file: figures/nondeterminism/fig_person.tex
\begin{figure*}[t]
    \centering
    \setlength{\tabcolsep}{0.5pt}
    \renewcommand{\arraystretch}{1.0}
    \setlength{\ww}{0.4\columnwidth}
    \begin{tabular}{ccccc}
        &&&&
        \textit{``holding an}
        \\

        &
        \textit{``in the park''} &
        \textit{``reading a book''} &
        \textit{``at the beach''} &
        \textit{avocado''}
        \\

        {\includegraphics[valign=c, width=\ww]{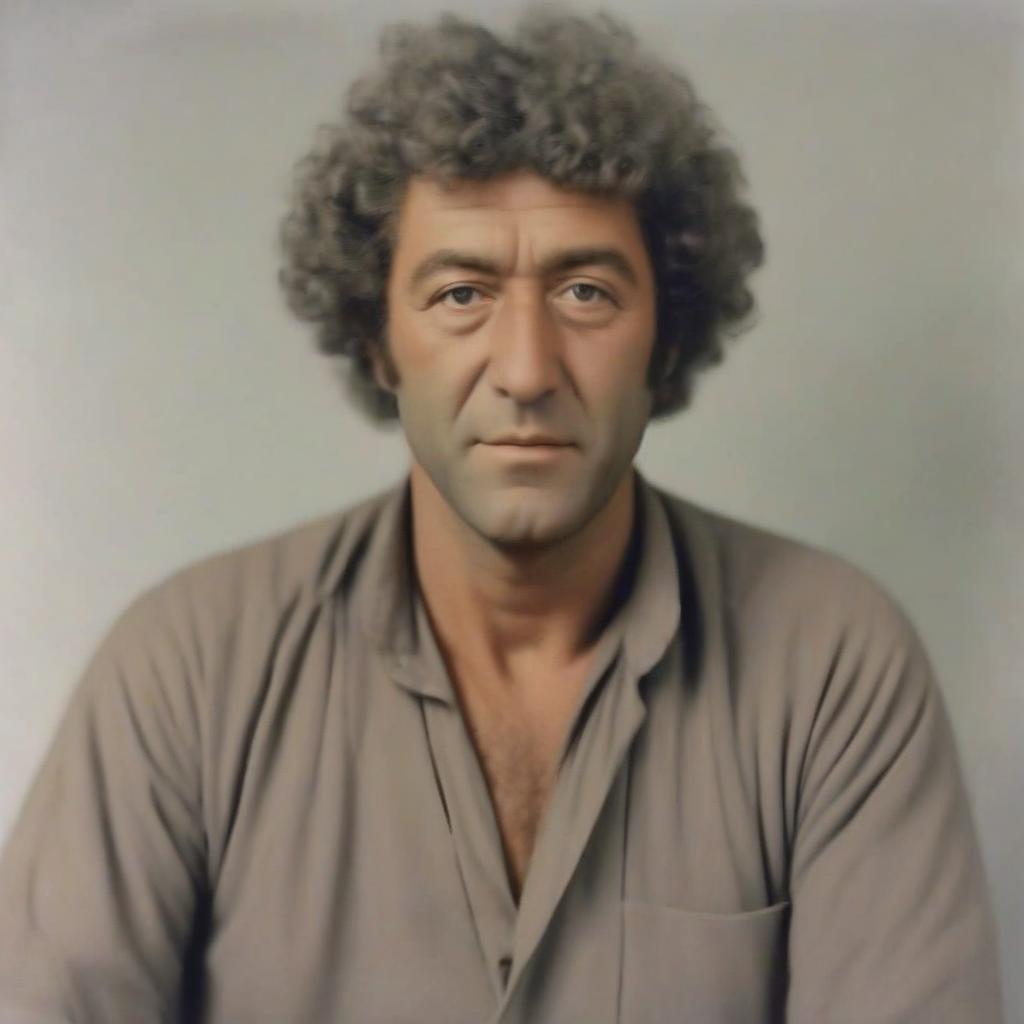}} &
        {\includegraphics[valign=c, width=\ww]{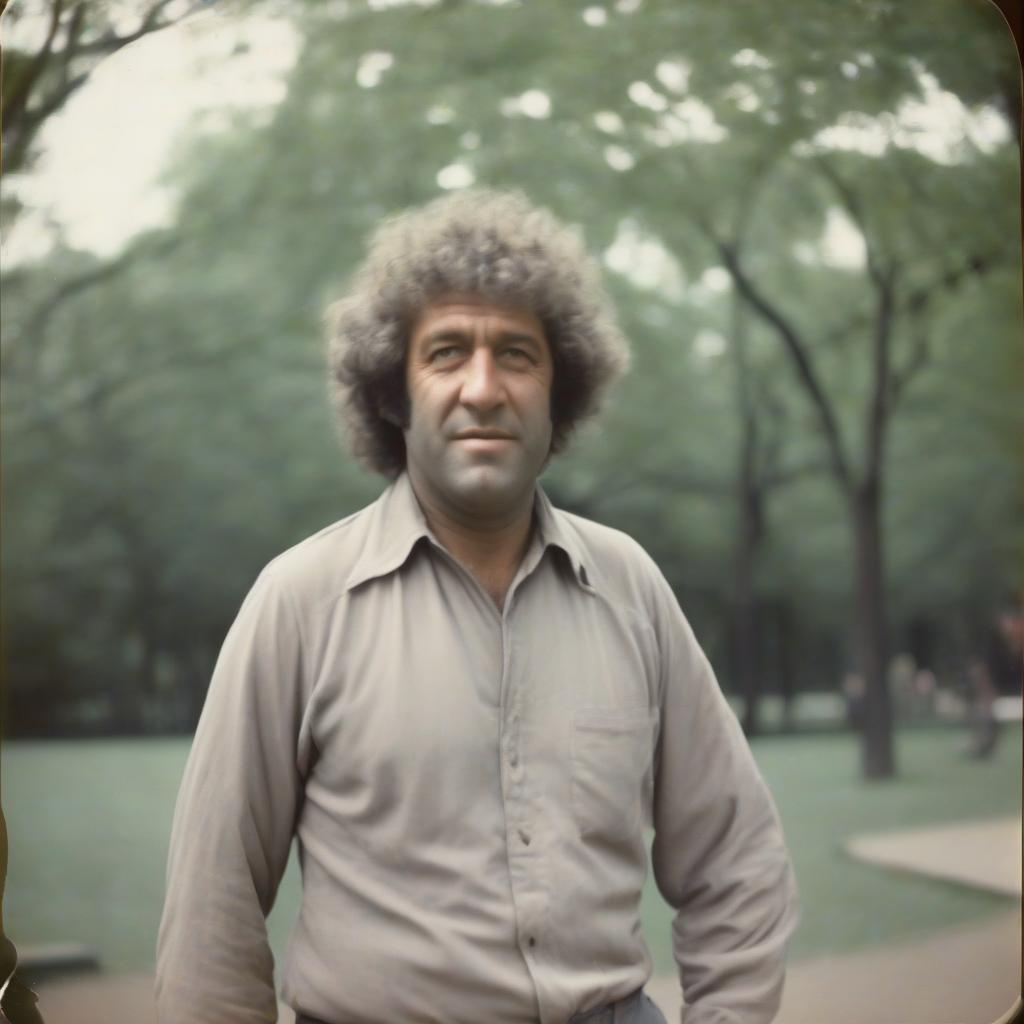}} &
        {\includegraphics[valign=c, width=\ww]{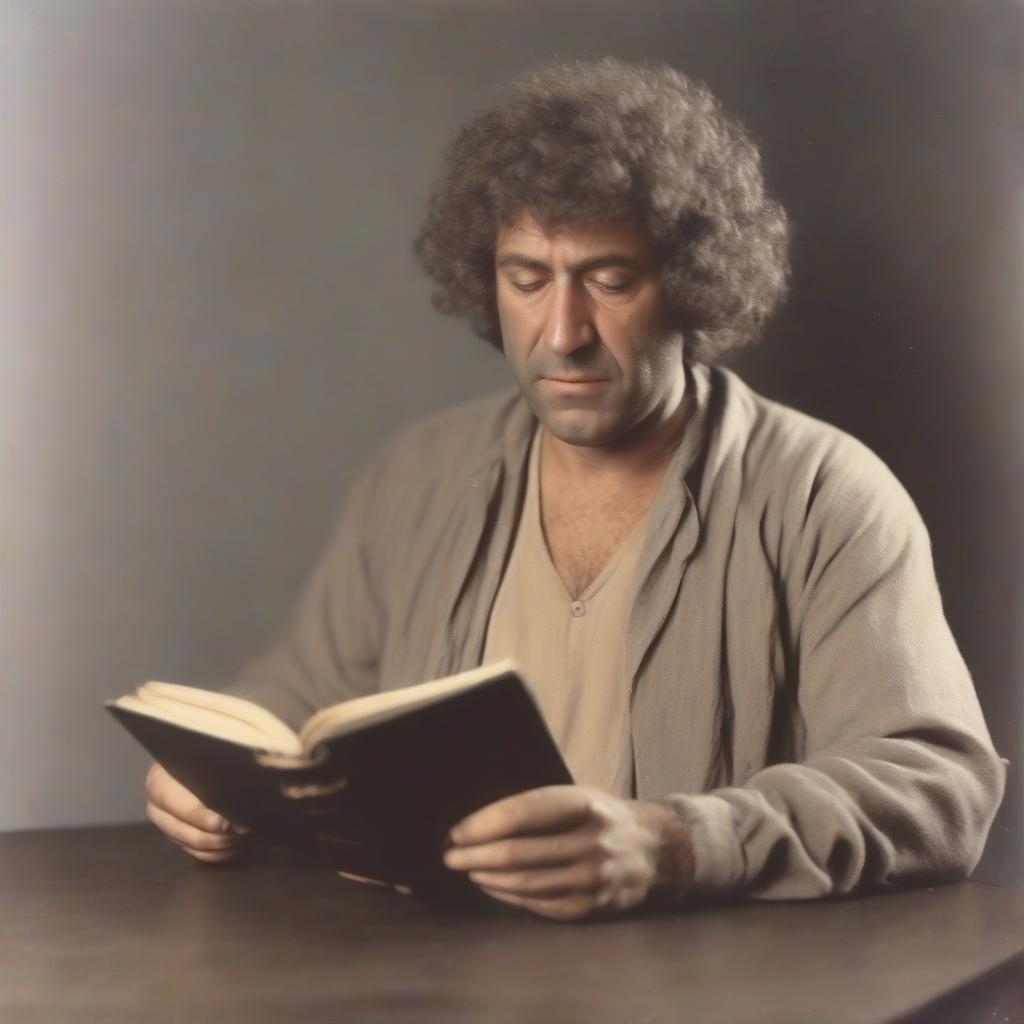}} &
        {\includegraphics[valign=c, width=\ww]{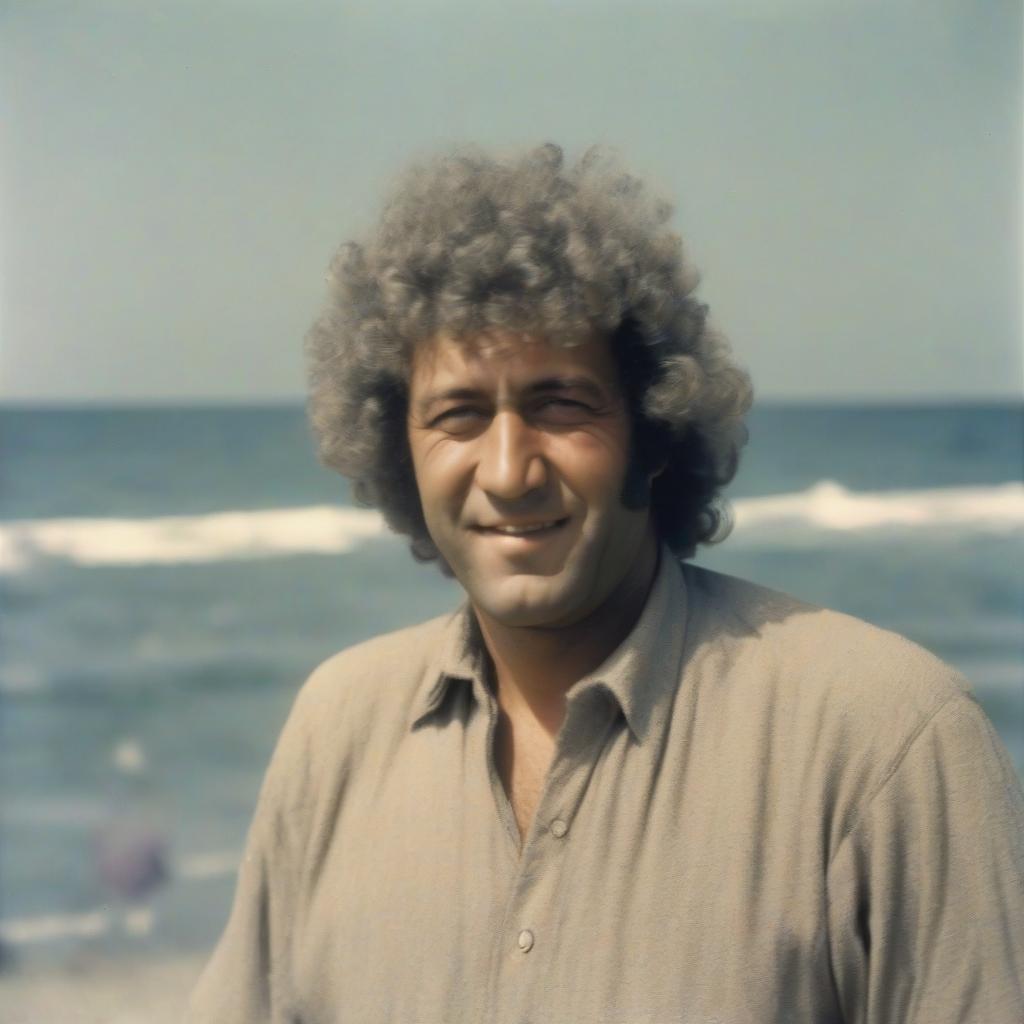}} &
        {\includegraphics[valign=c, width=\ww]{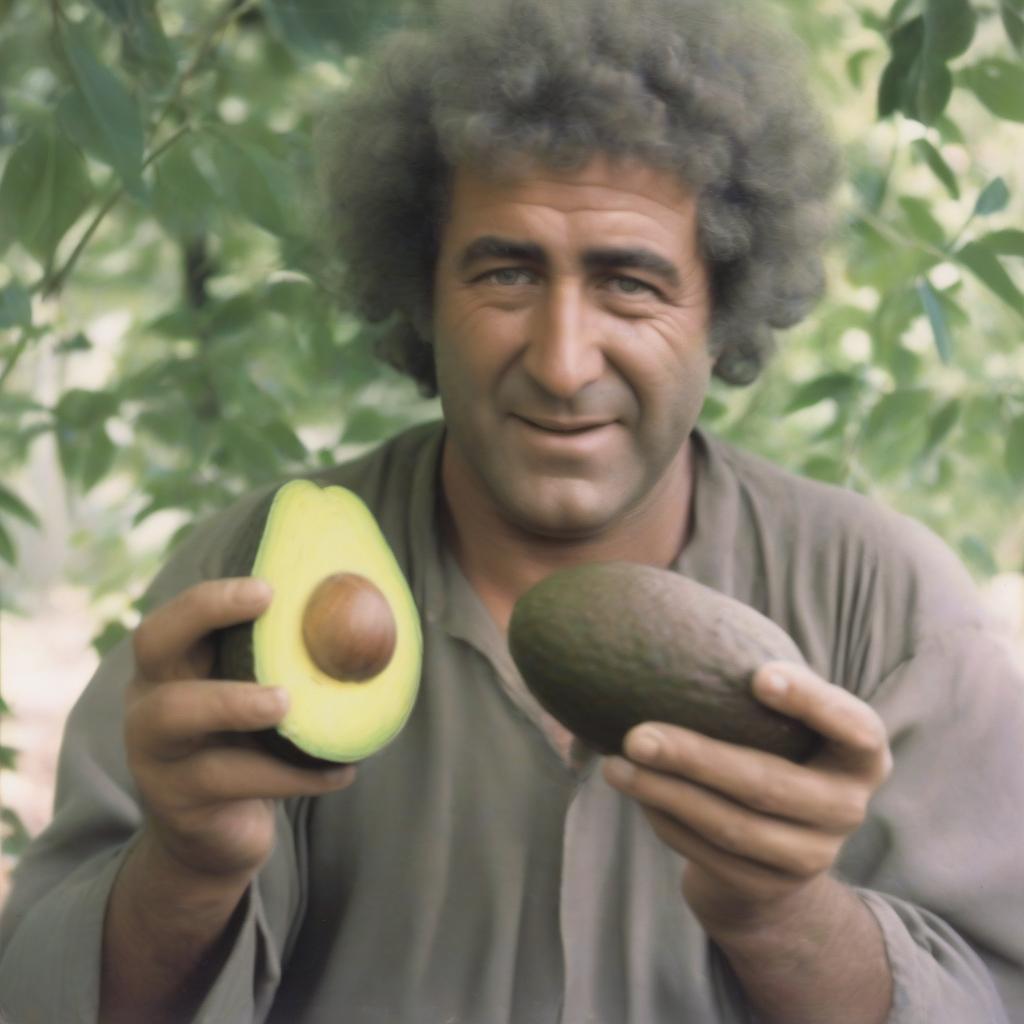}}
        \\
        \\

        {\includegraphics[valign=c, width=\ww]{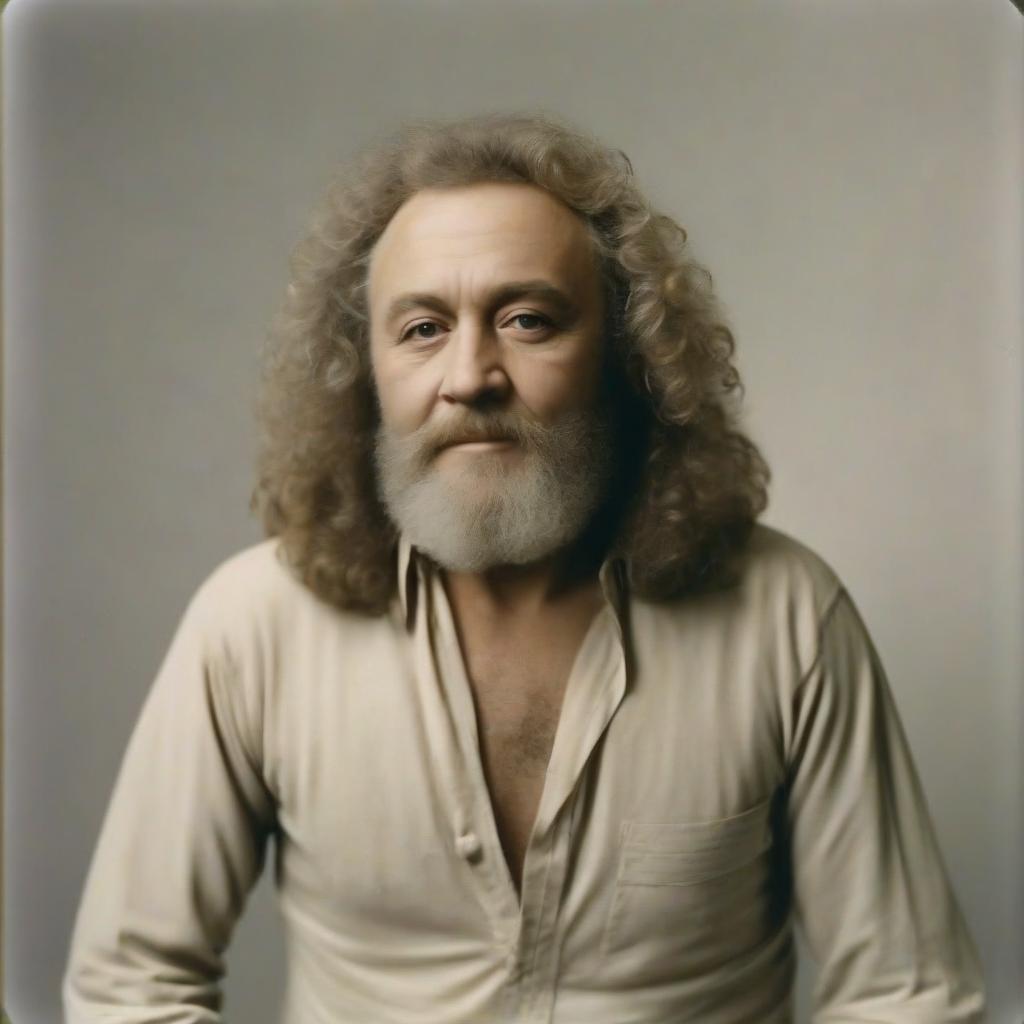}} &
        {\includegraphics[valign=c, width=\ww]{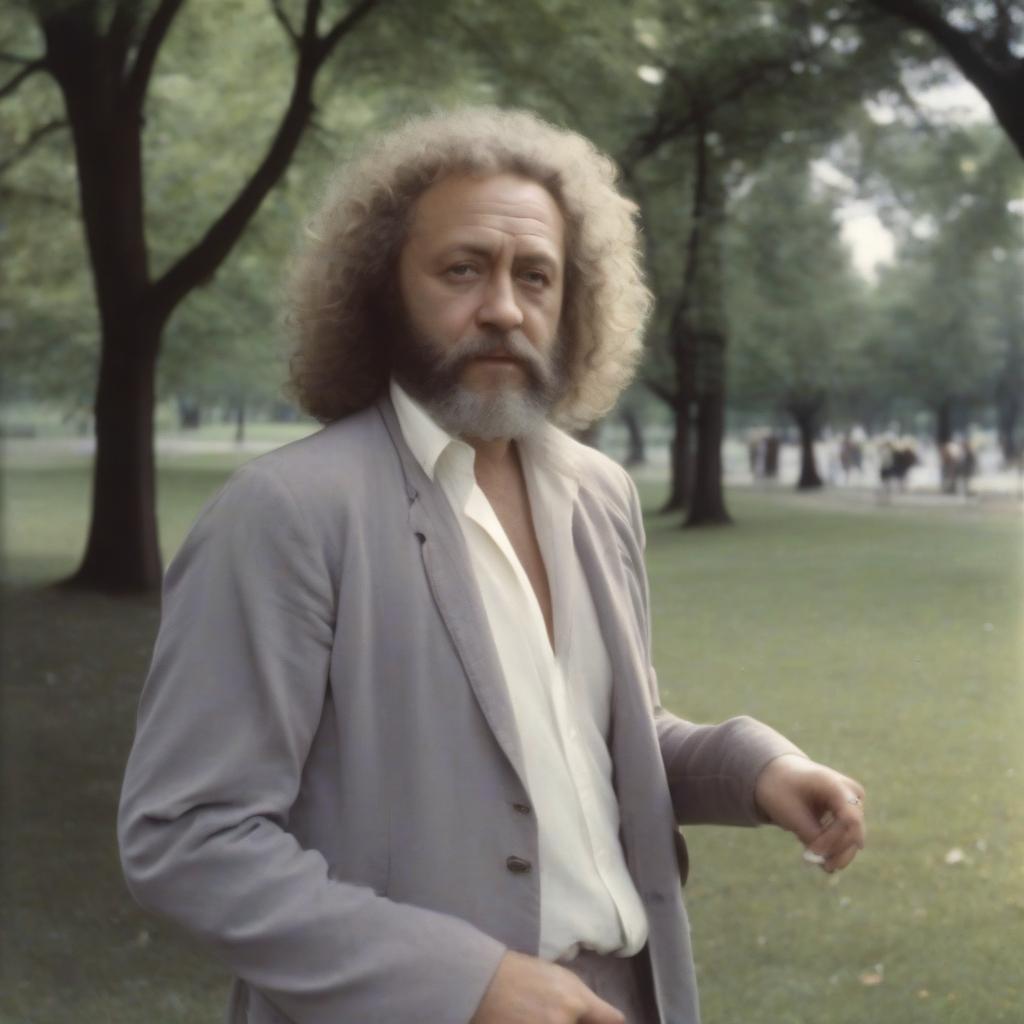}} &
        {\includegraphics[valign=c, width=\ww]{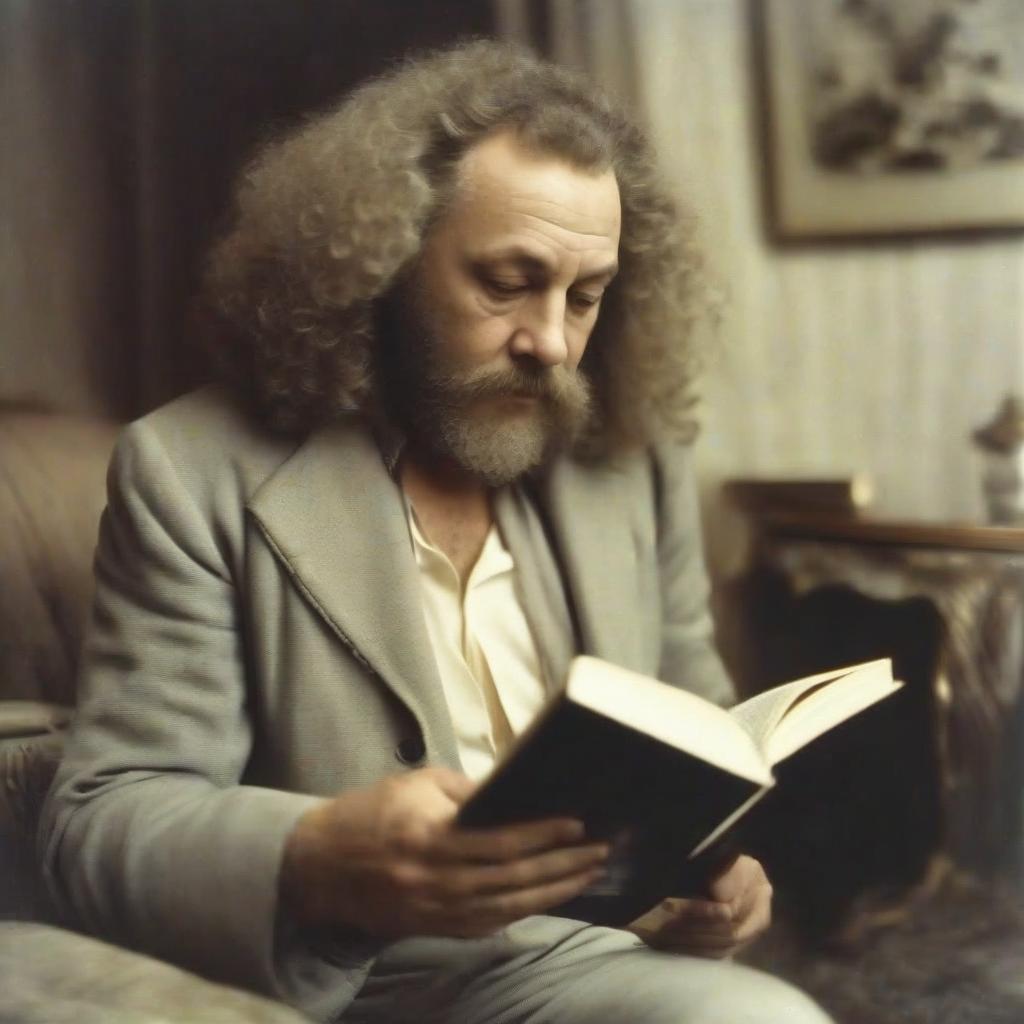}} &
        {\includegraphics[valign=c, width=\ww]{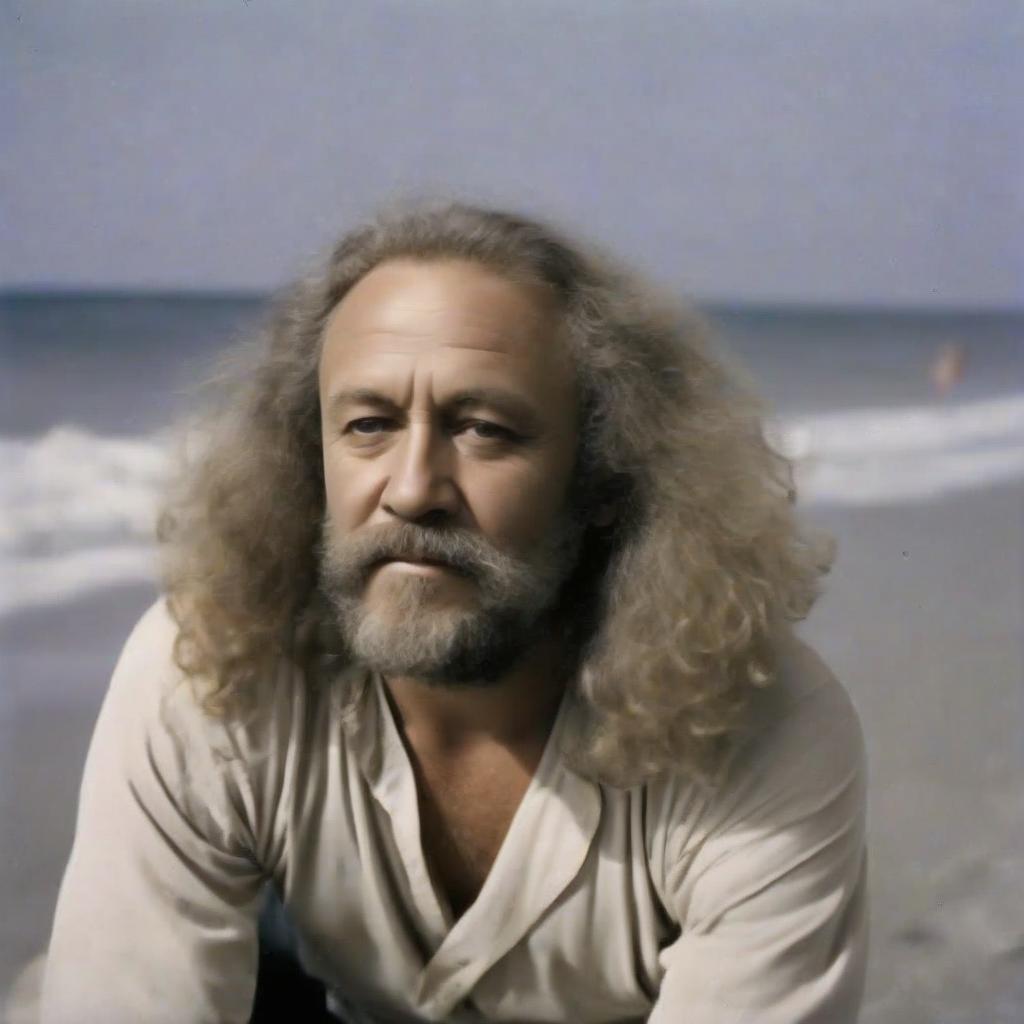}} &
        {\includegraphics[valign=c, width=\ww]{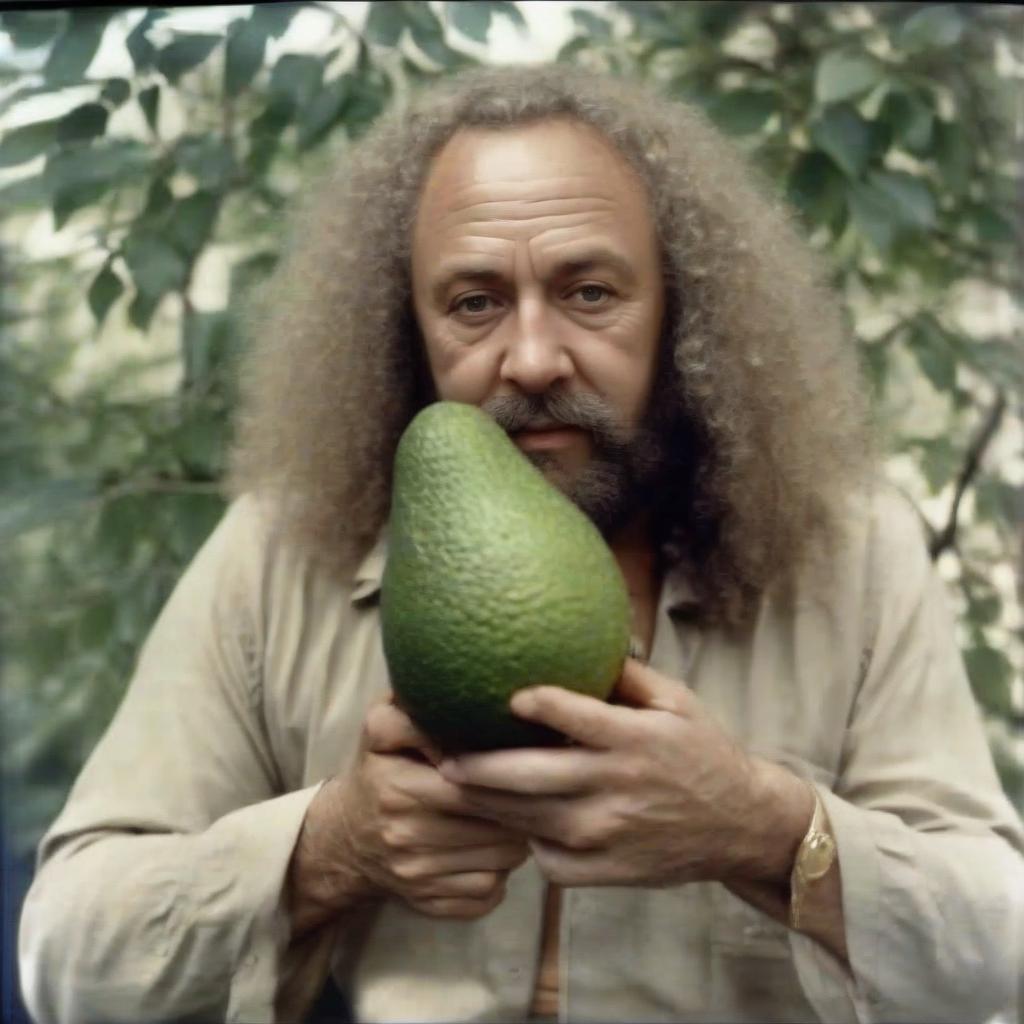}}
        \\
        \\

        {\includegraphics[valign=c, width=\ww]{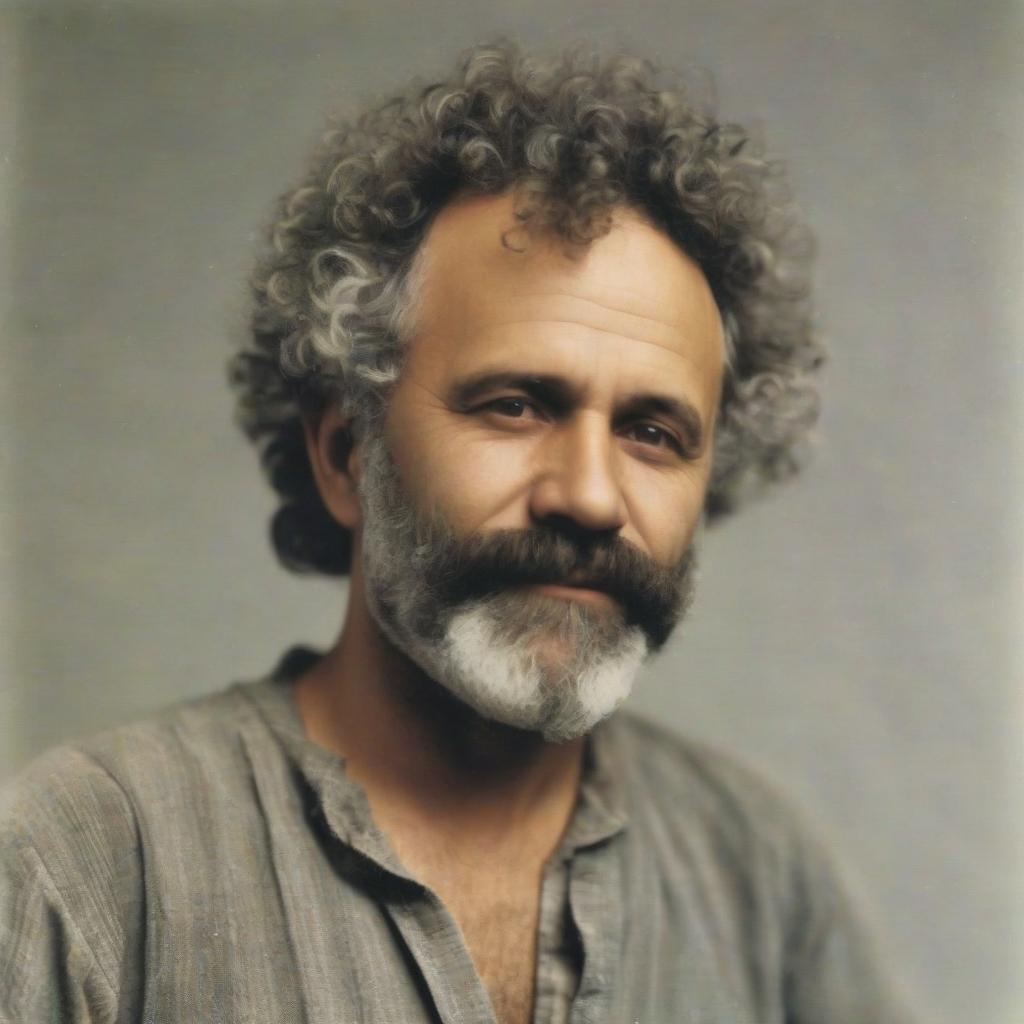}} &
        {\includegraphics[valign=c, width=\ww]{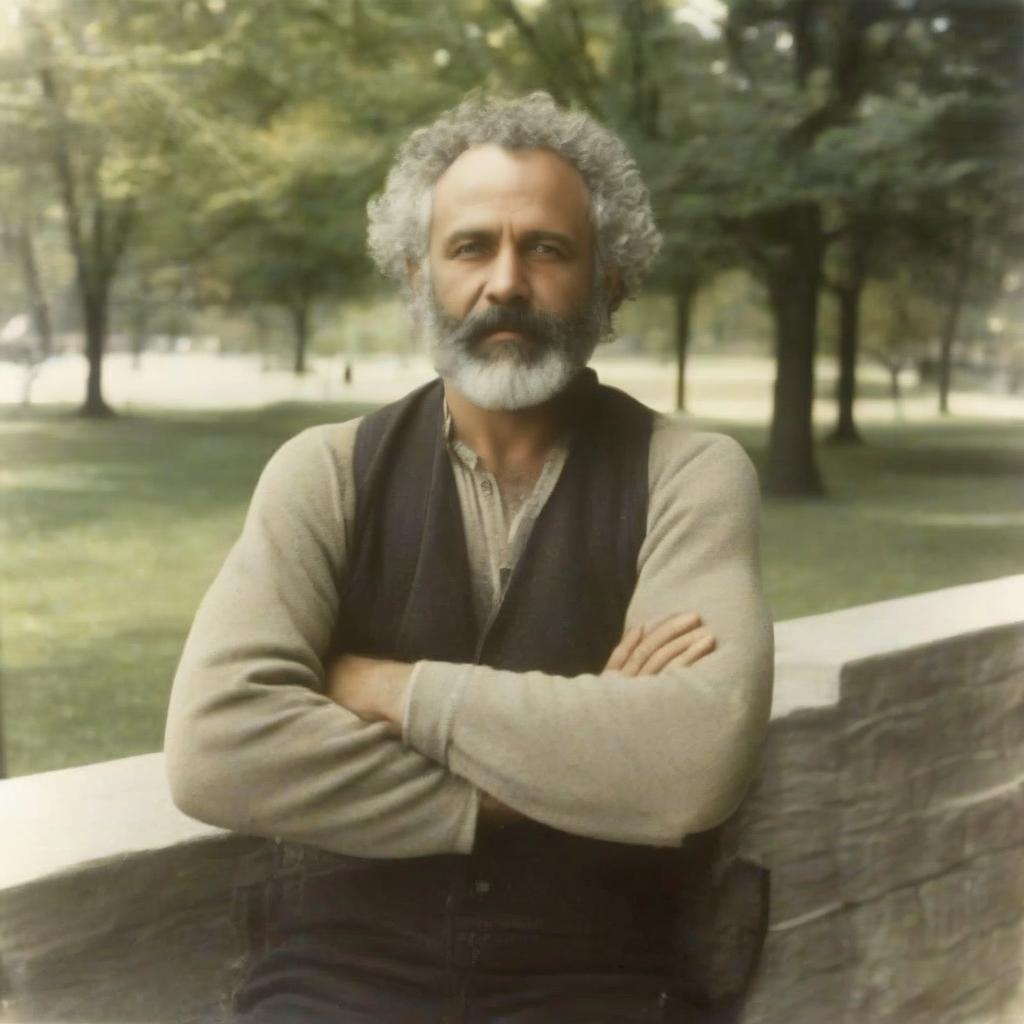}} &
        {\includegraphics[valign=c, width=\ww]{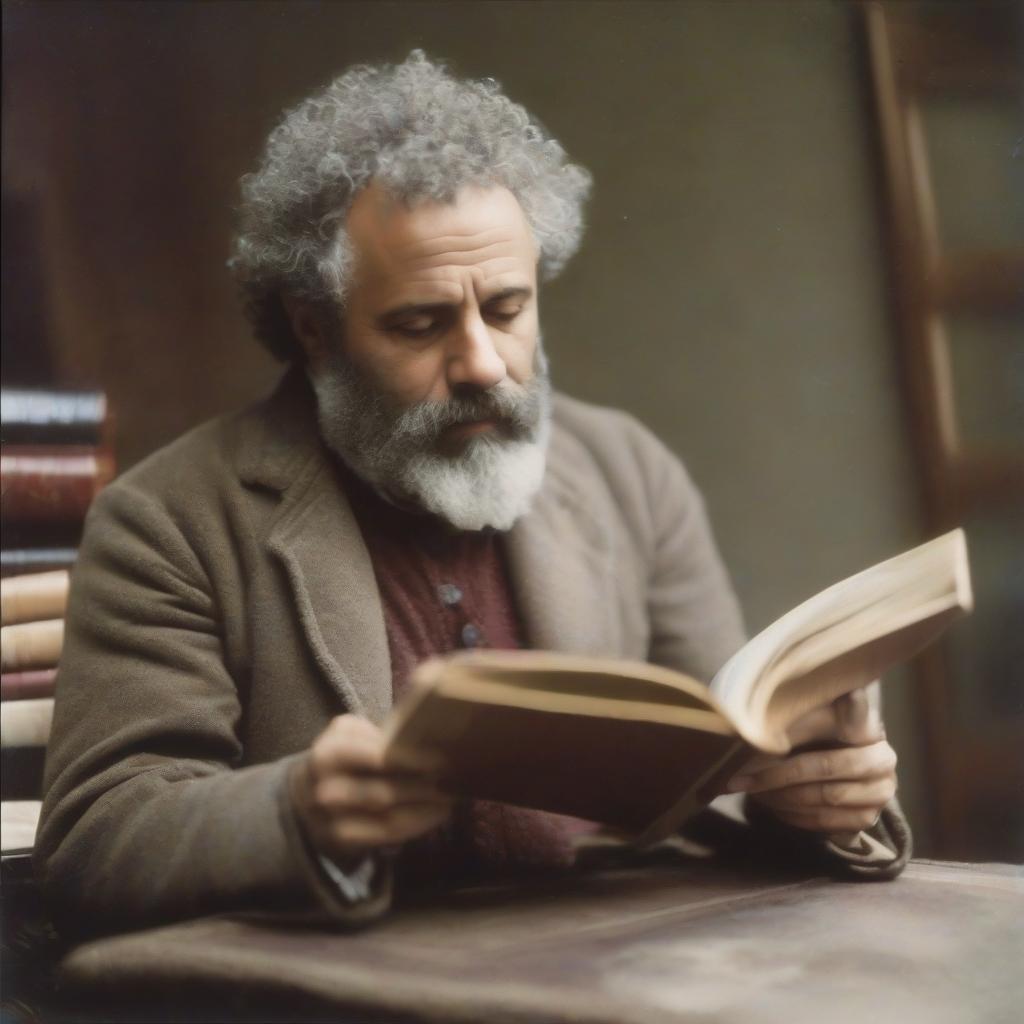}} &
        {\includegraphics[valign=c, width=\ww]{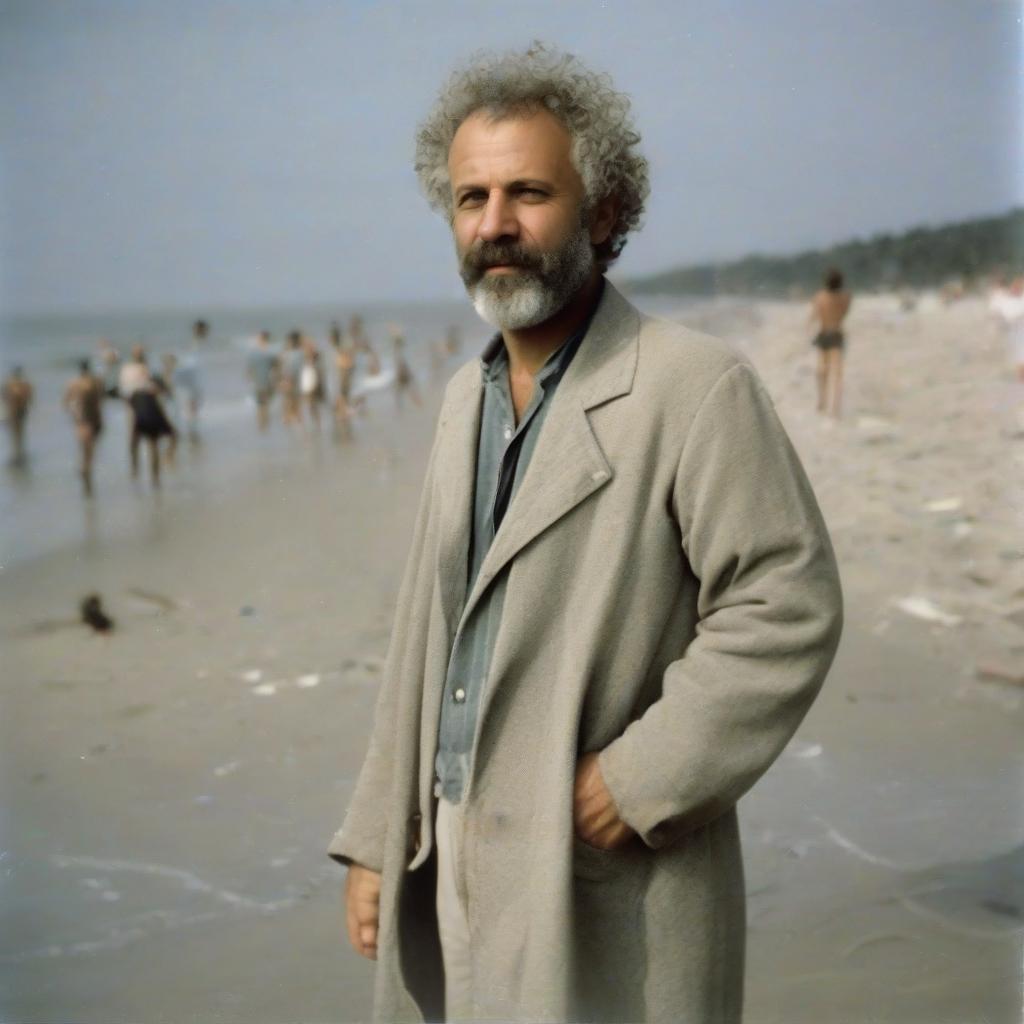}} &
        {\includegraphics[valign=c, width=\ww]{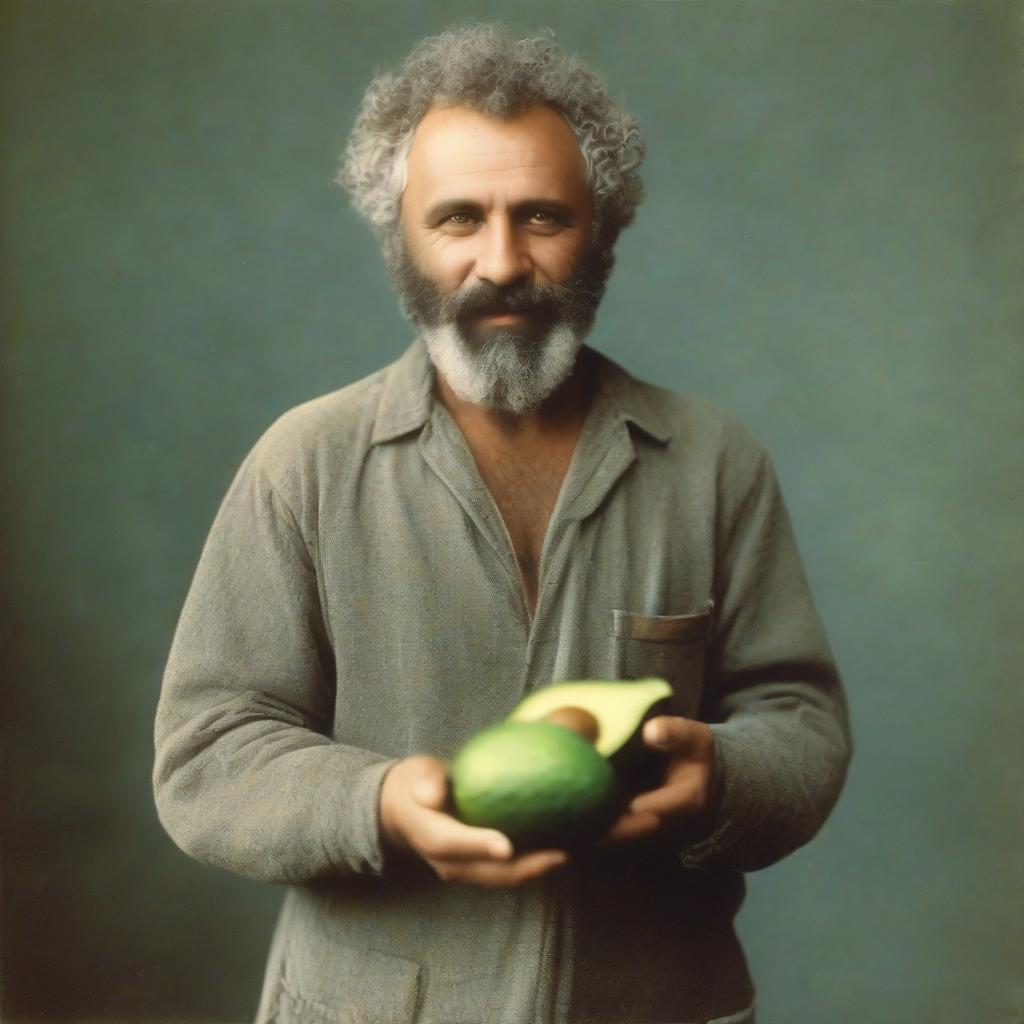}}
        \\
        \\

        {\includegraphics[valign=c, width=\ww]{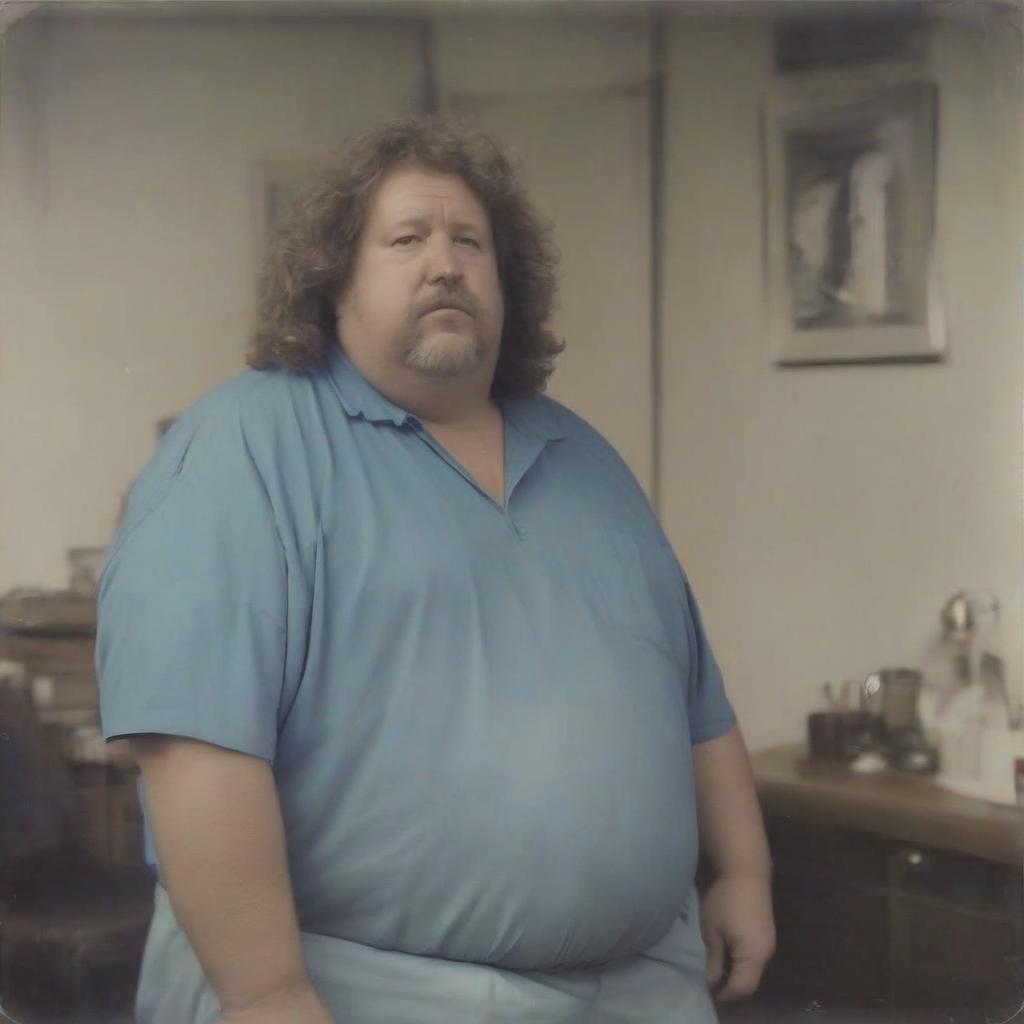}} &
        {\includegraphics[valign=c, width=\ww]{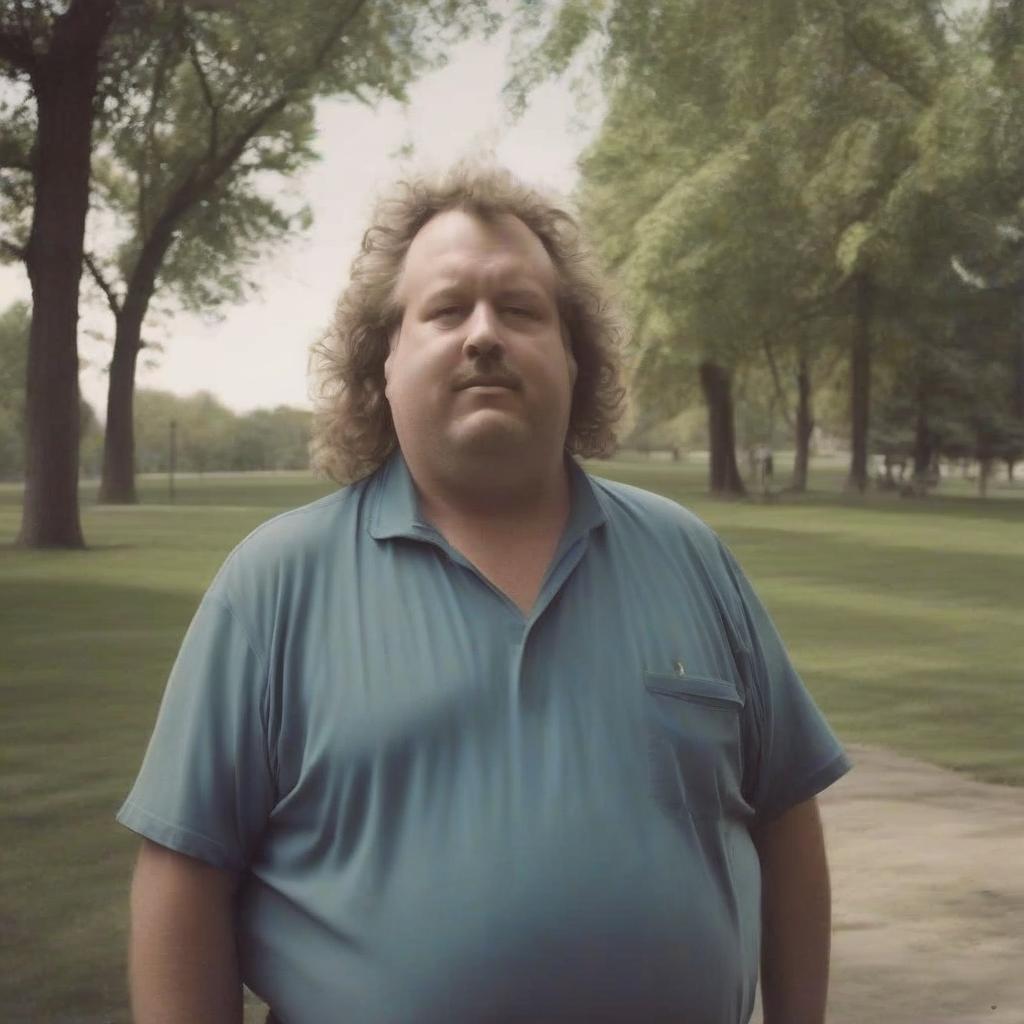}} &
        {\includegraphics[valign=c, width=\ww]{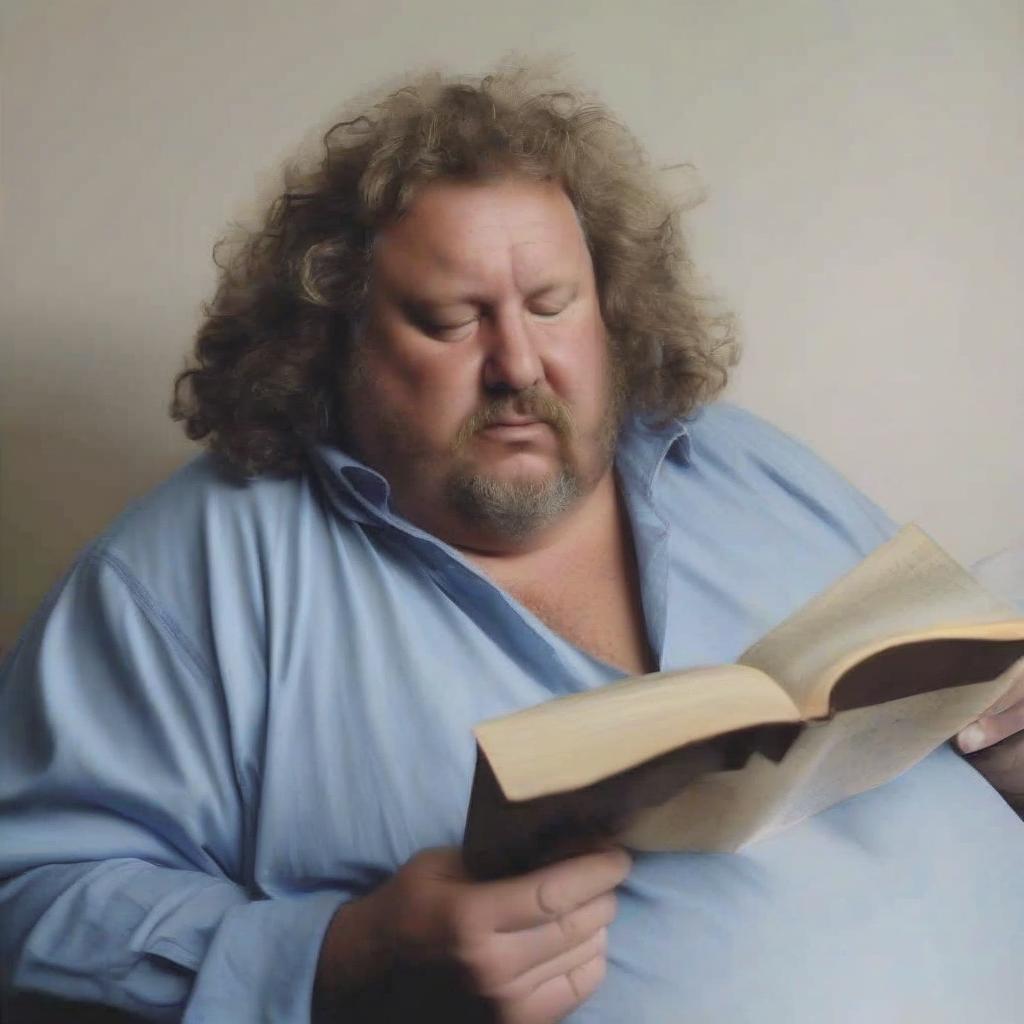}} &
        {\includegraphics[valign=c, width=\ww]{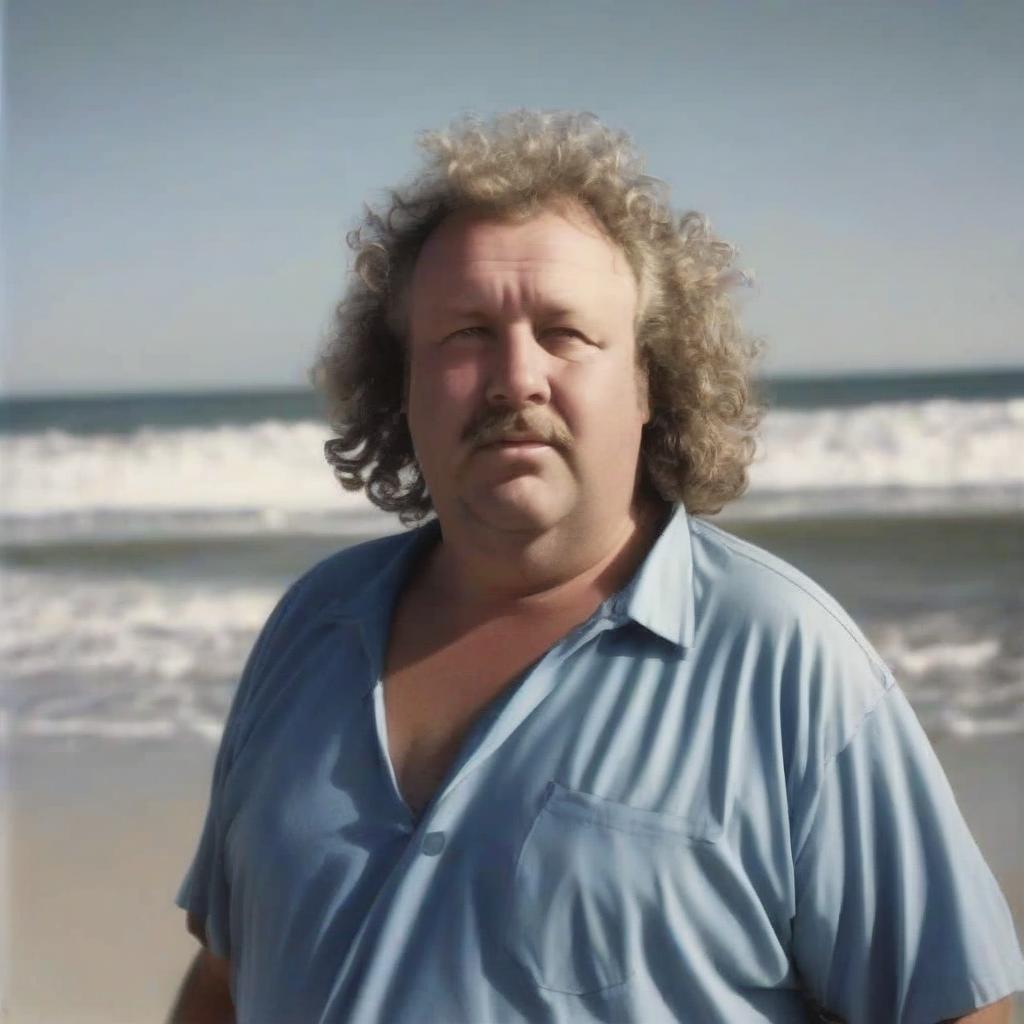}} &
        {\includegraphics[valign=c, width=\ww]{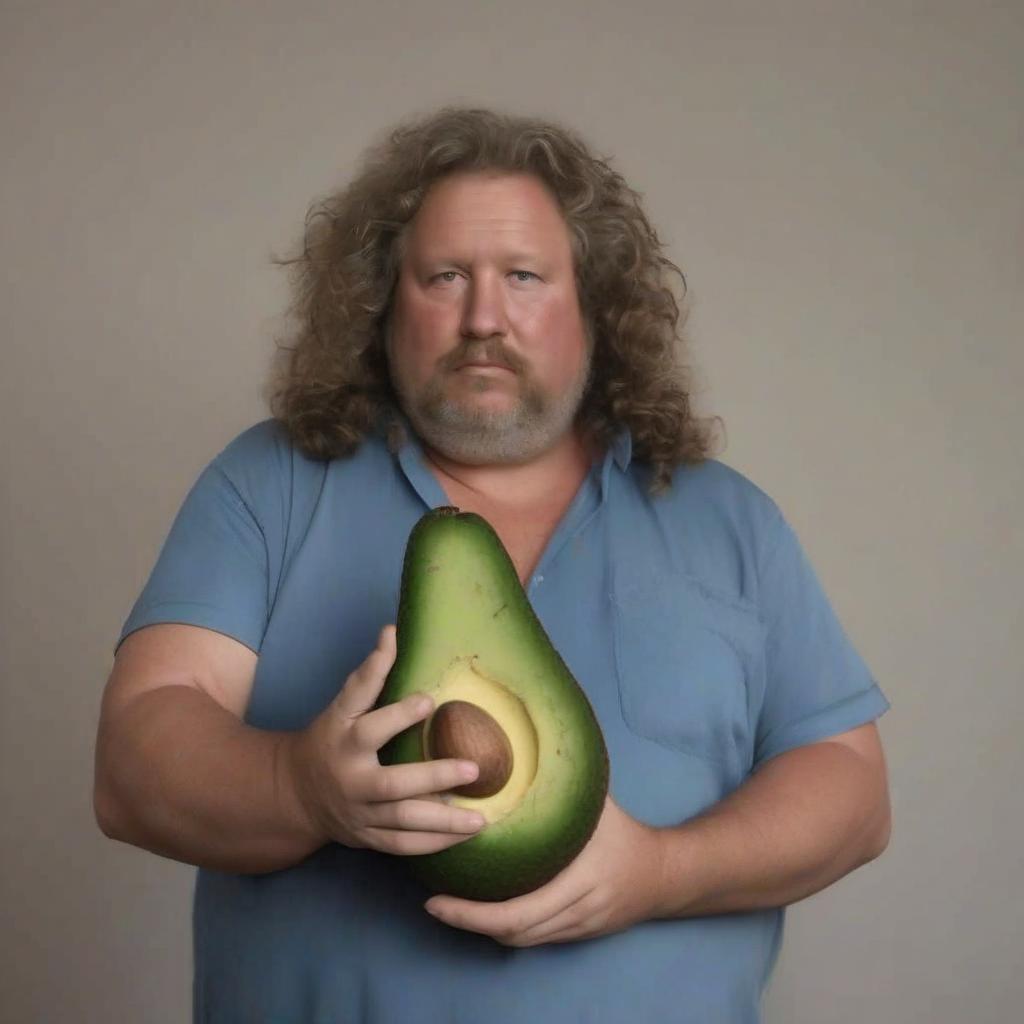}}

    \end{tabular}
    
    \caption{\textbf{Non-determinism.} By running our method multiple times, given the same prompt \textit{``a photo of a 50 years old man with curly hair''}, but using different initial seeds, we obtain different consistent characters corresponding to the text prompt.
    }
    \label{fig:nondeterminism_person}
\end{figure*}

%% file: figures/nondeterminism/fig_cat.tex
\begin{figure*}[t]
    \centering
    \setlength{\tabcolsep}{0.5pt}
    \renewcommand{\arraystretch}{1.0}
    \setlength{\ww}{0.4\columnwidth}
    \begin{tabular}{ccccc}
        &&&&
        \textit{``holding an}
        \\

        &
        \textit{``in the park''} &
        \textit{``reading a book''} &
        \textit{``at the beach''} &
        \textit{avocado''}
        \\

        {\includegraphics[valign=c, width=\ww]{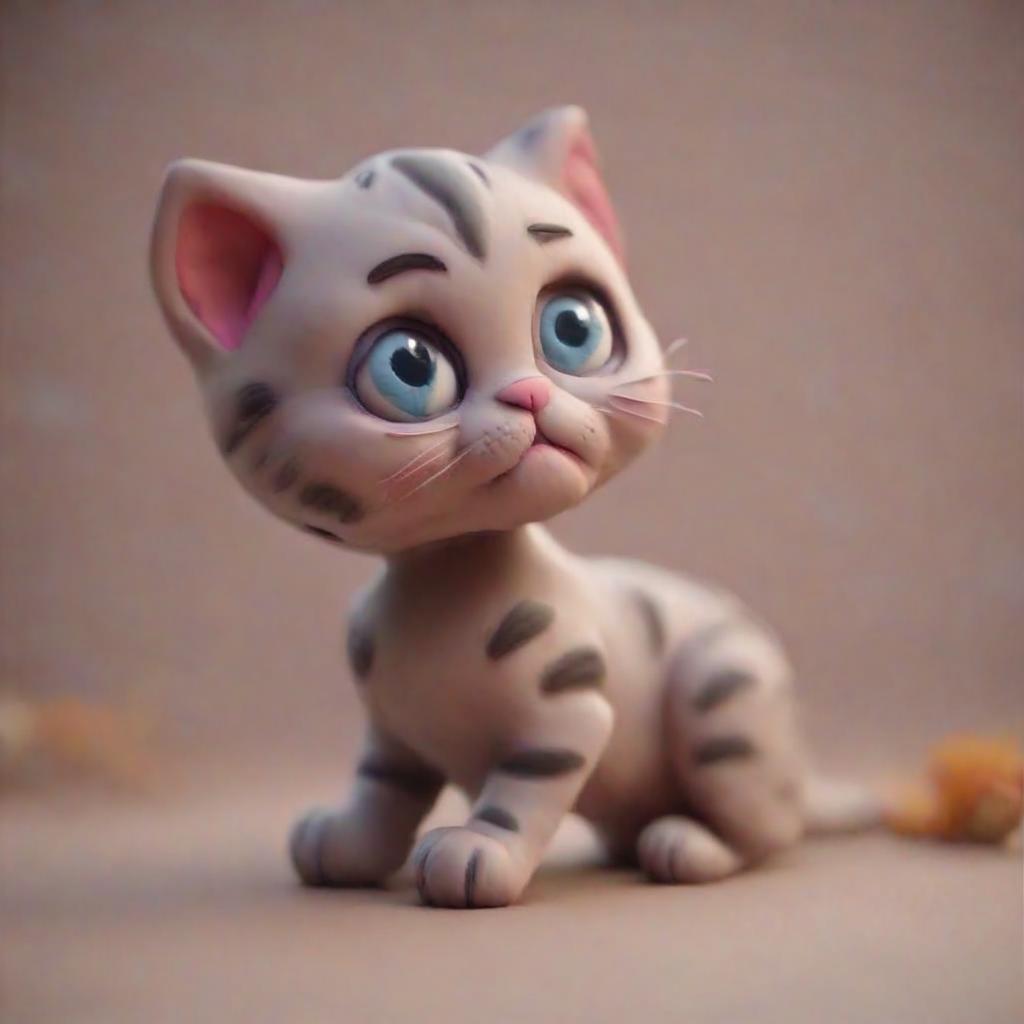}} &
        {\includegraphics[valign=c, width=\ww]{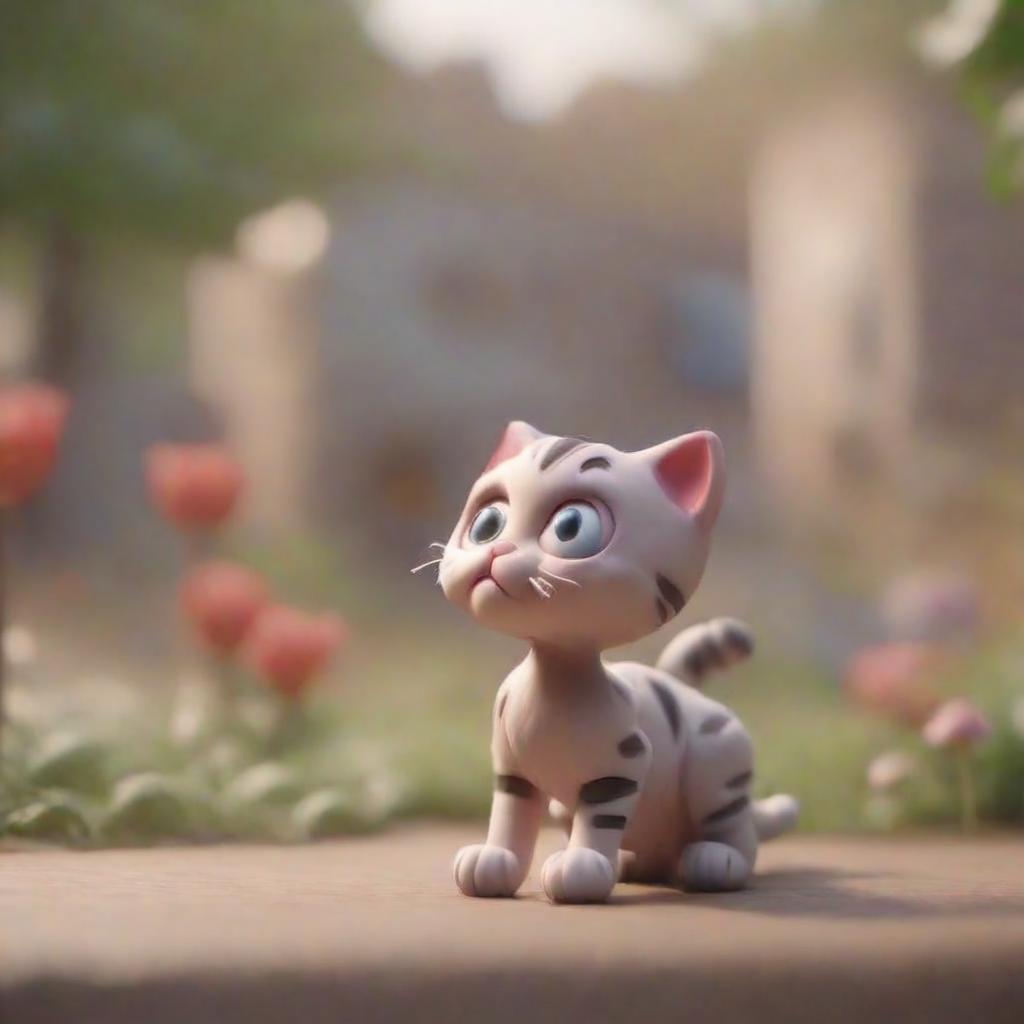}} &
        {\includegraphics[valign=c, width=\ww]{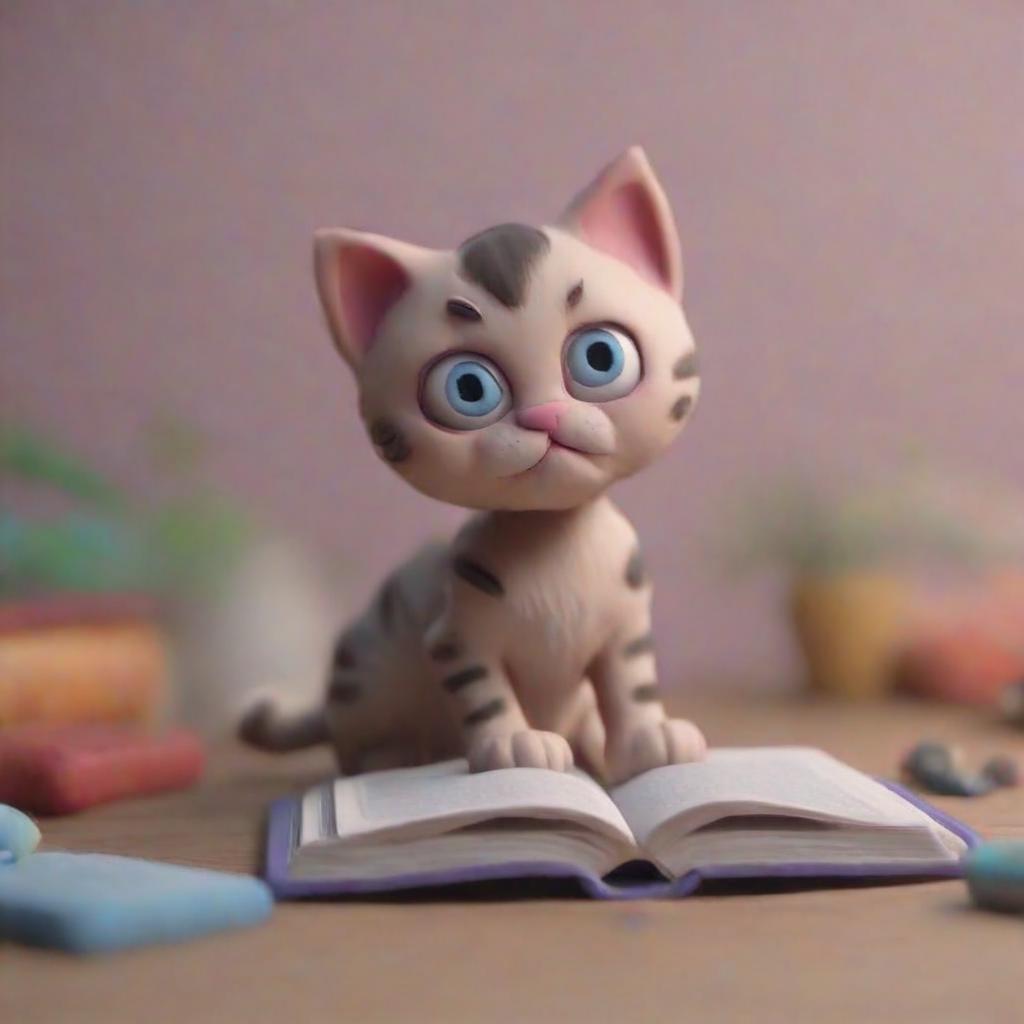}} &
        {\includegraphics[valign=c, width=\ww]{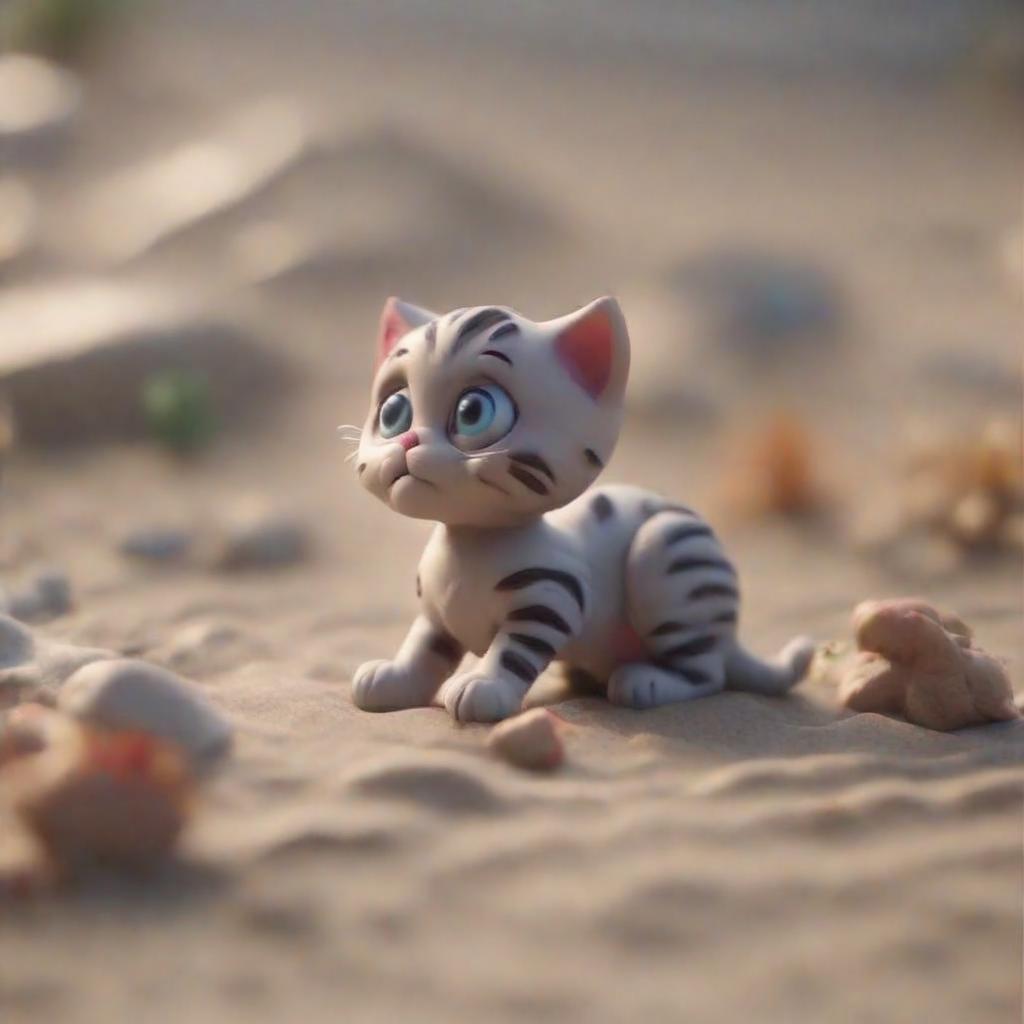}} &
        {\includegraphics[valign=c, width=\ww]{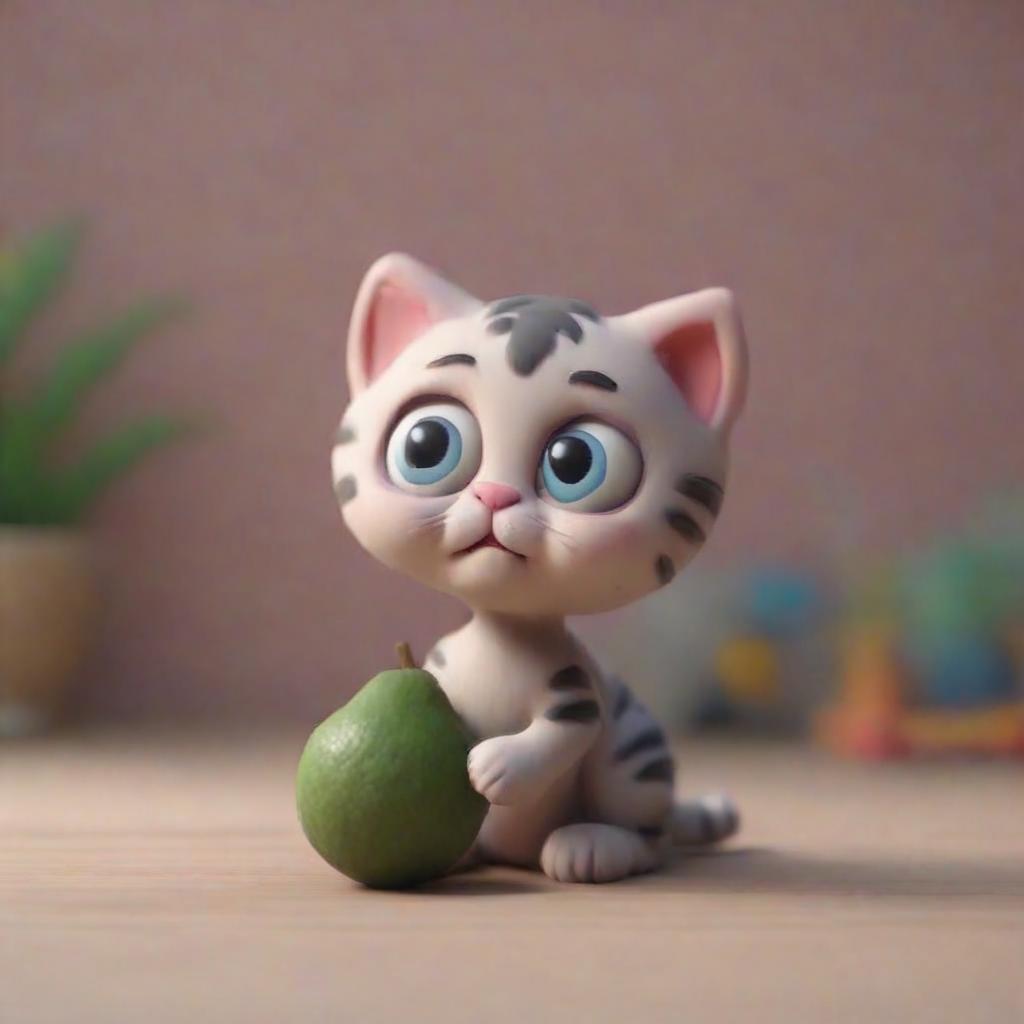}}
        \\
        \\

        {\includegraphics[valign=c, width=\ww]{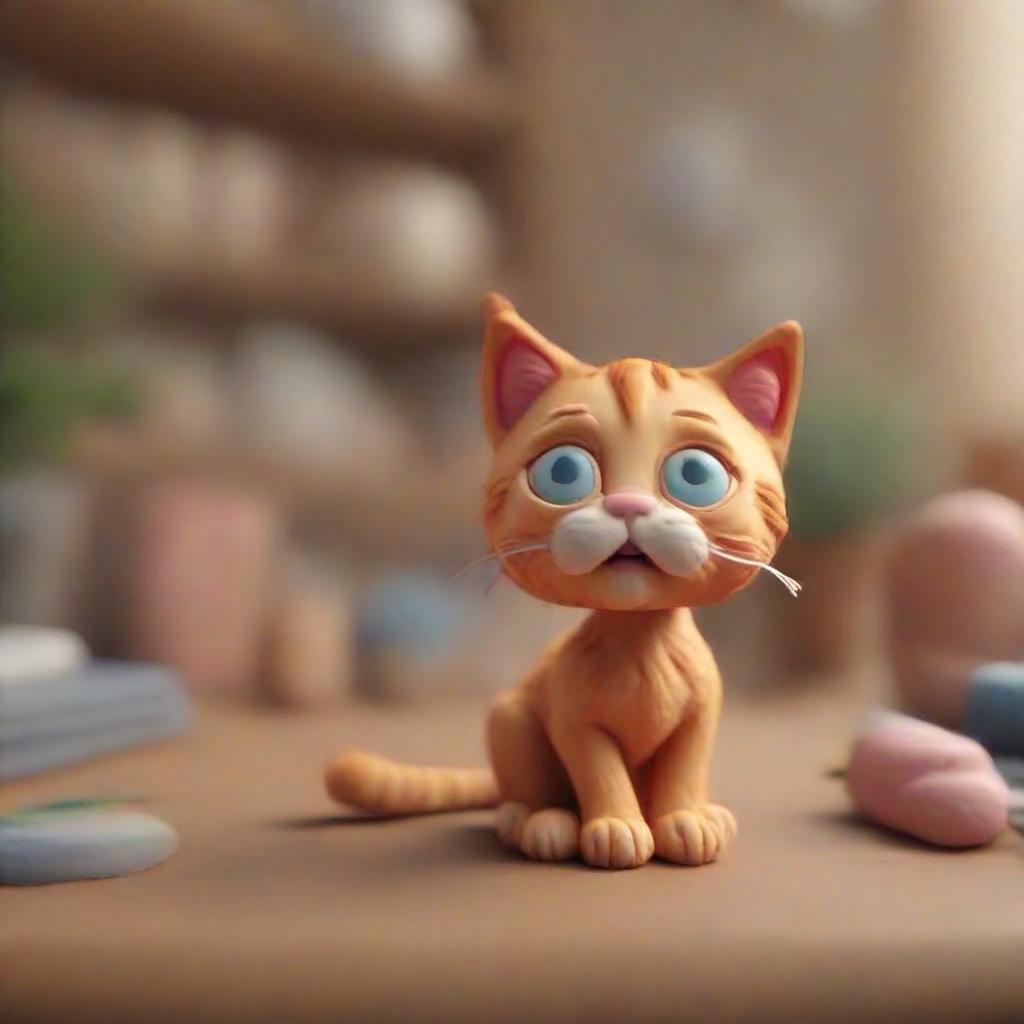}} &
        {\includegraphics[valign=c, width=\ww]{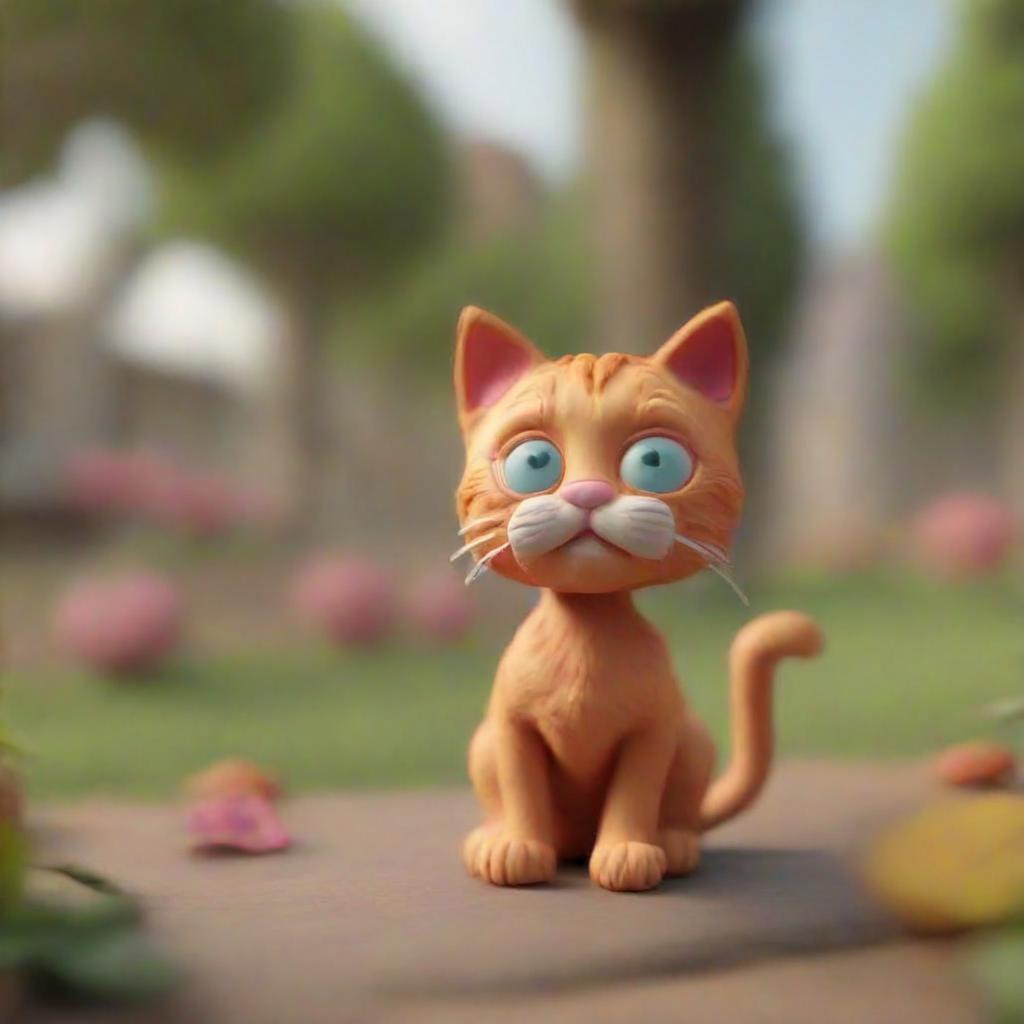}} &
        {\includegraphics[valign=c, width=\ww]{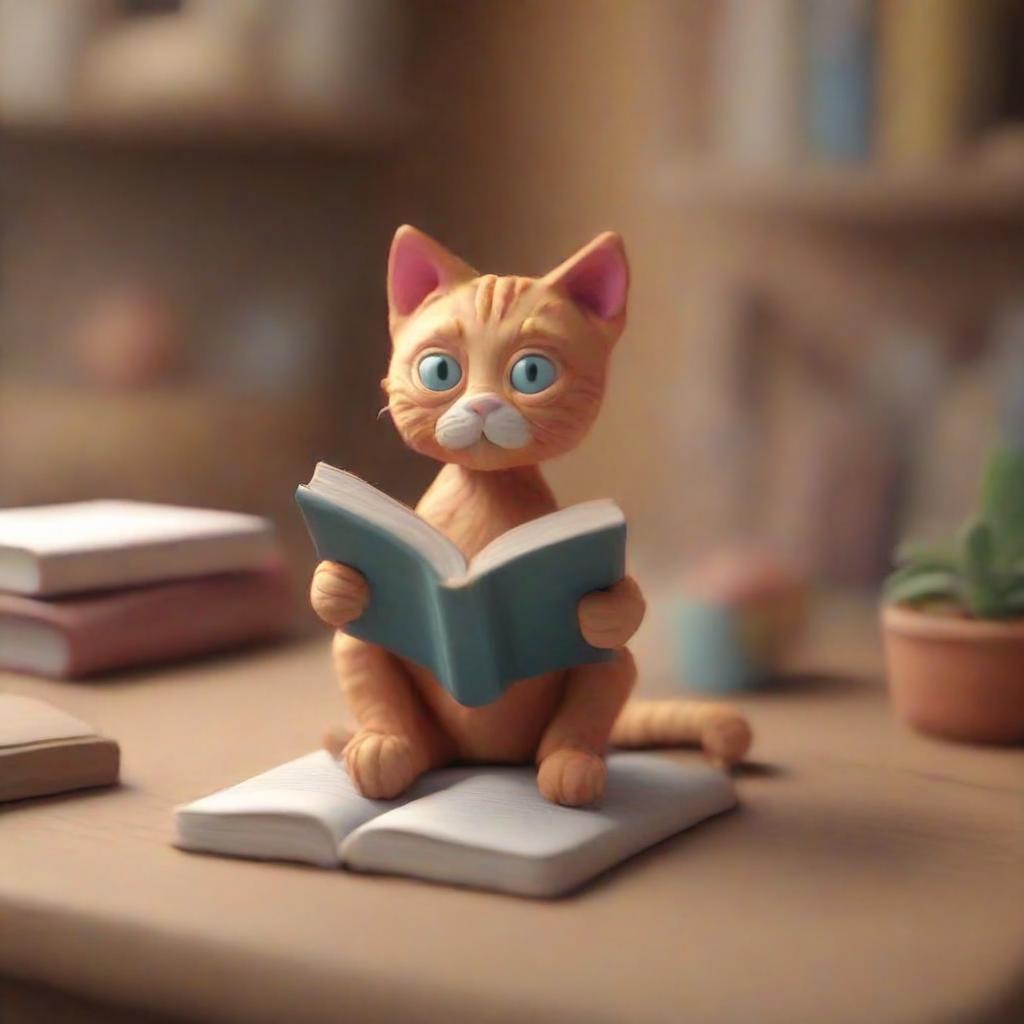}} &
        {\includegraphics[valign=c, width=\ww]{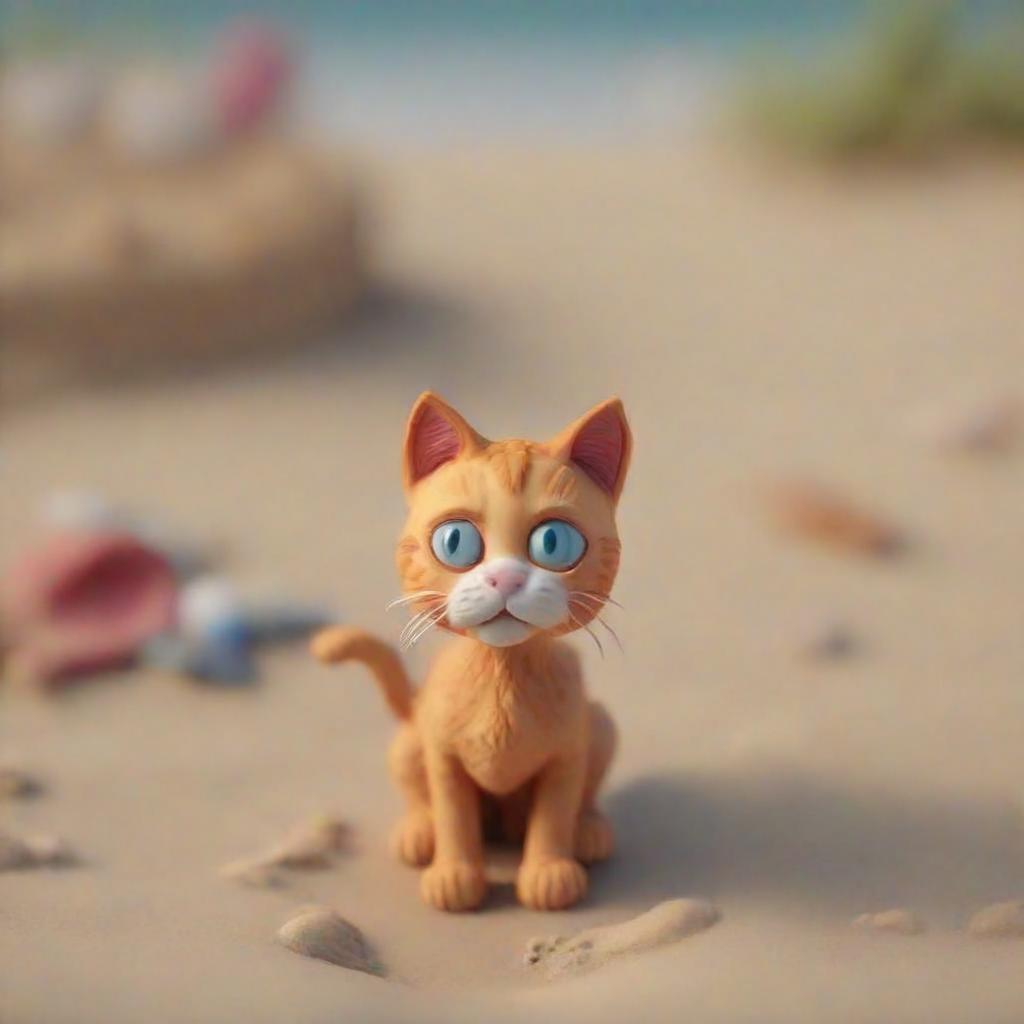}} &
        {\includegraphics[valign=c, width=\ww]{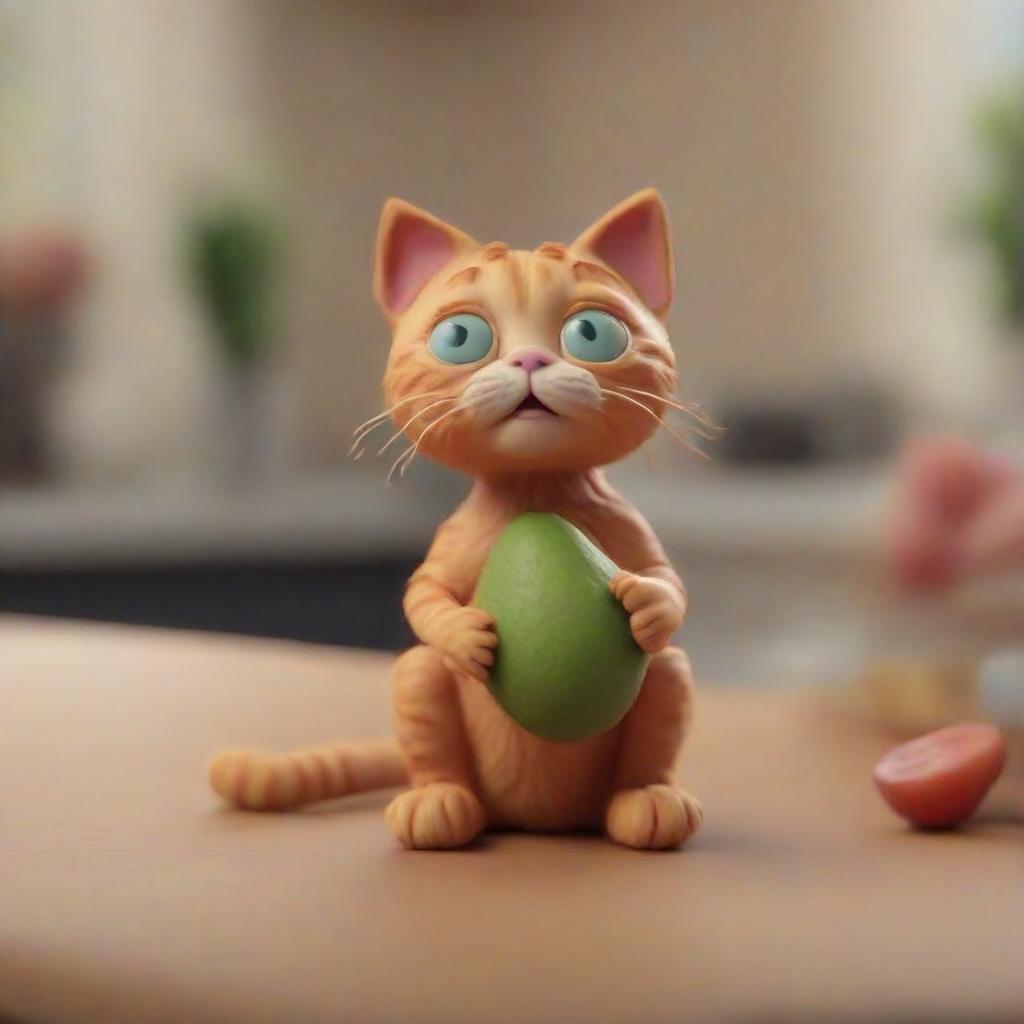}}
        \\
        \\

        {\includegraphics[valign=c, width=\ww]{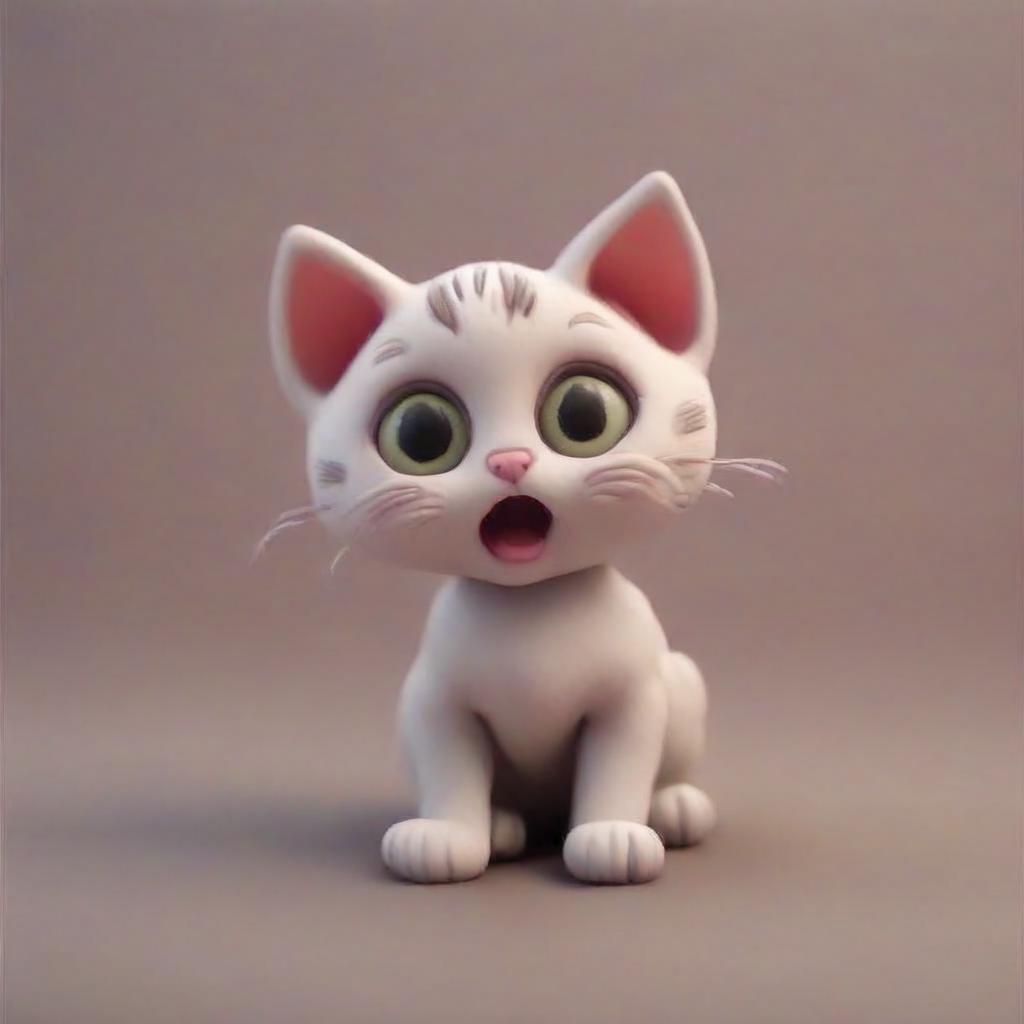}} &
        {\includegraphics[valign=c, width=\ww]{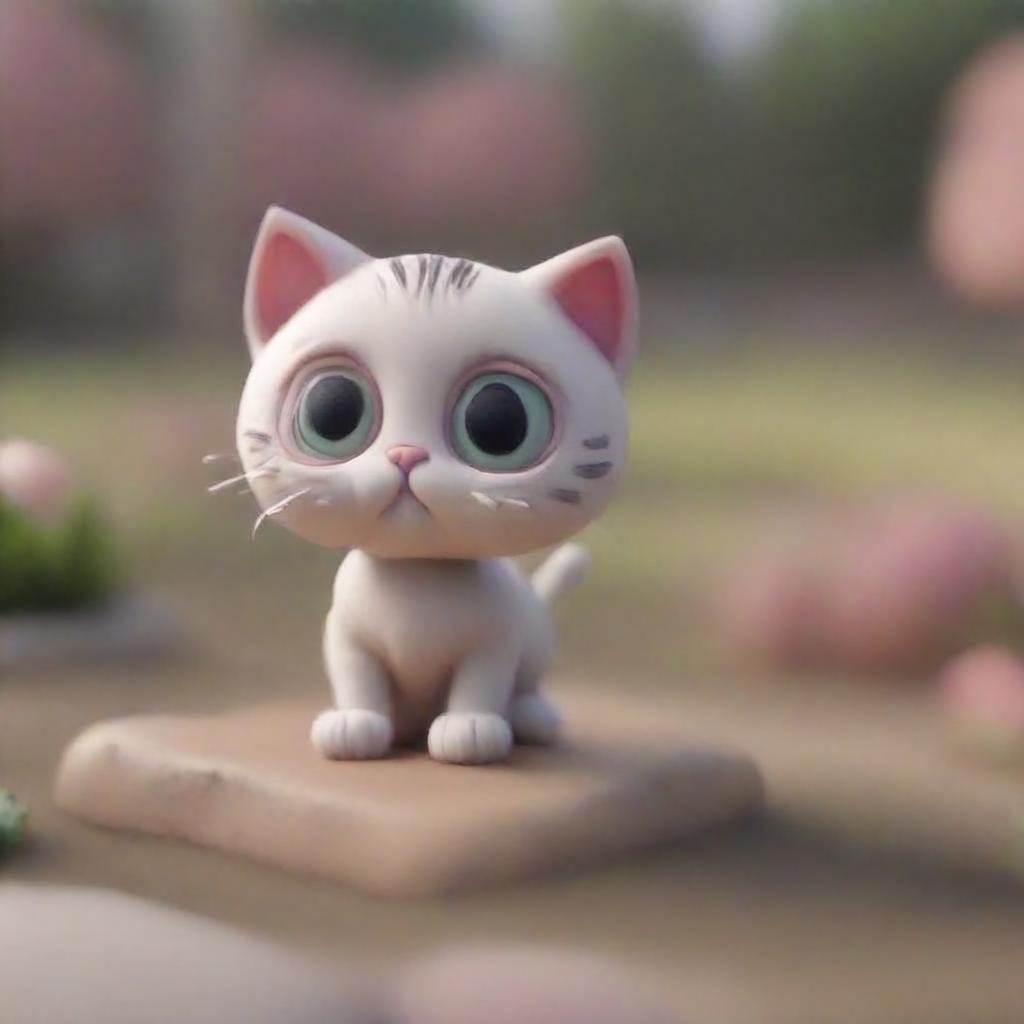}} &
        {\includegraphics[valign=c, width=\ww]{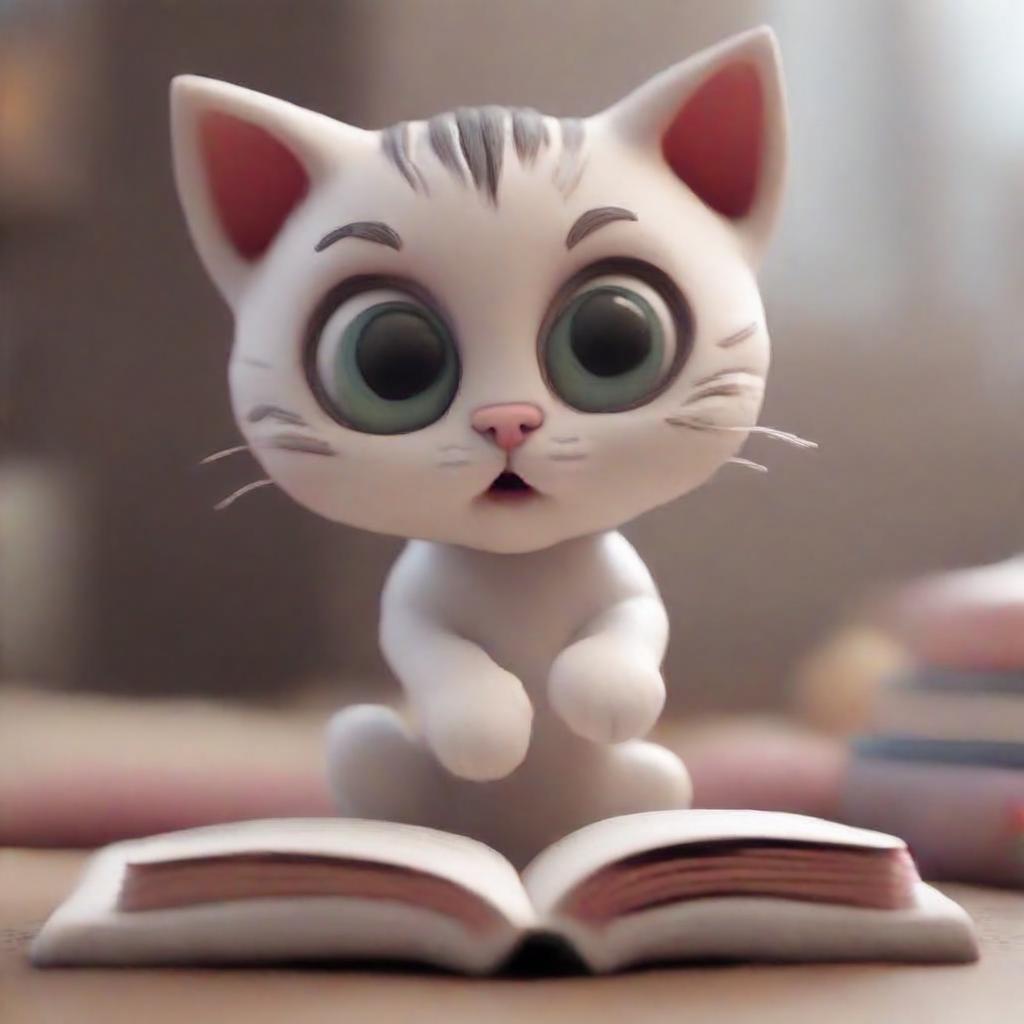}} &
        {\includegraphics[valign=c, width=\ww]{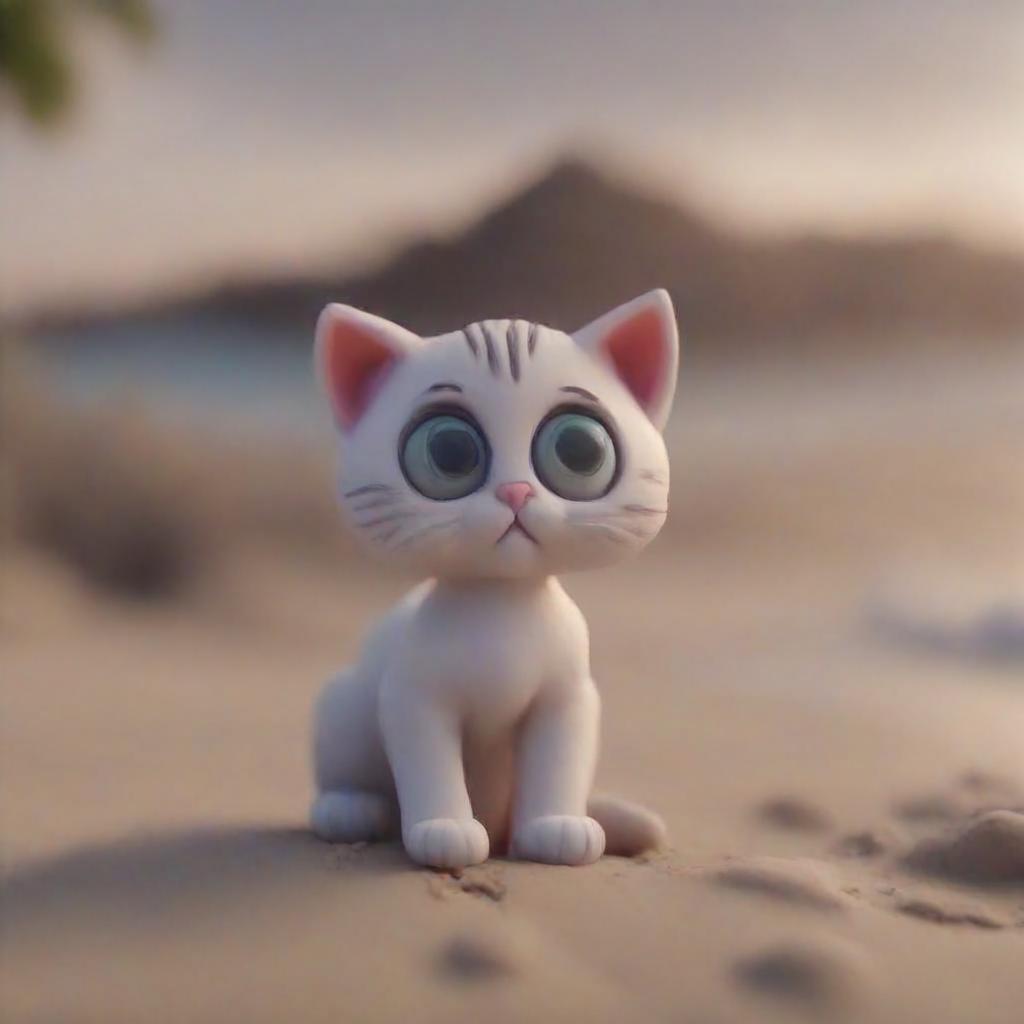}} &
        {\includegraphics[valign=c, width=\ww]{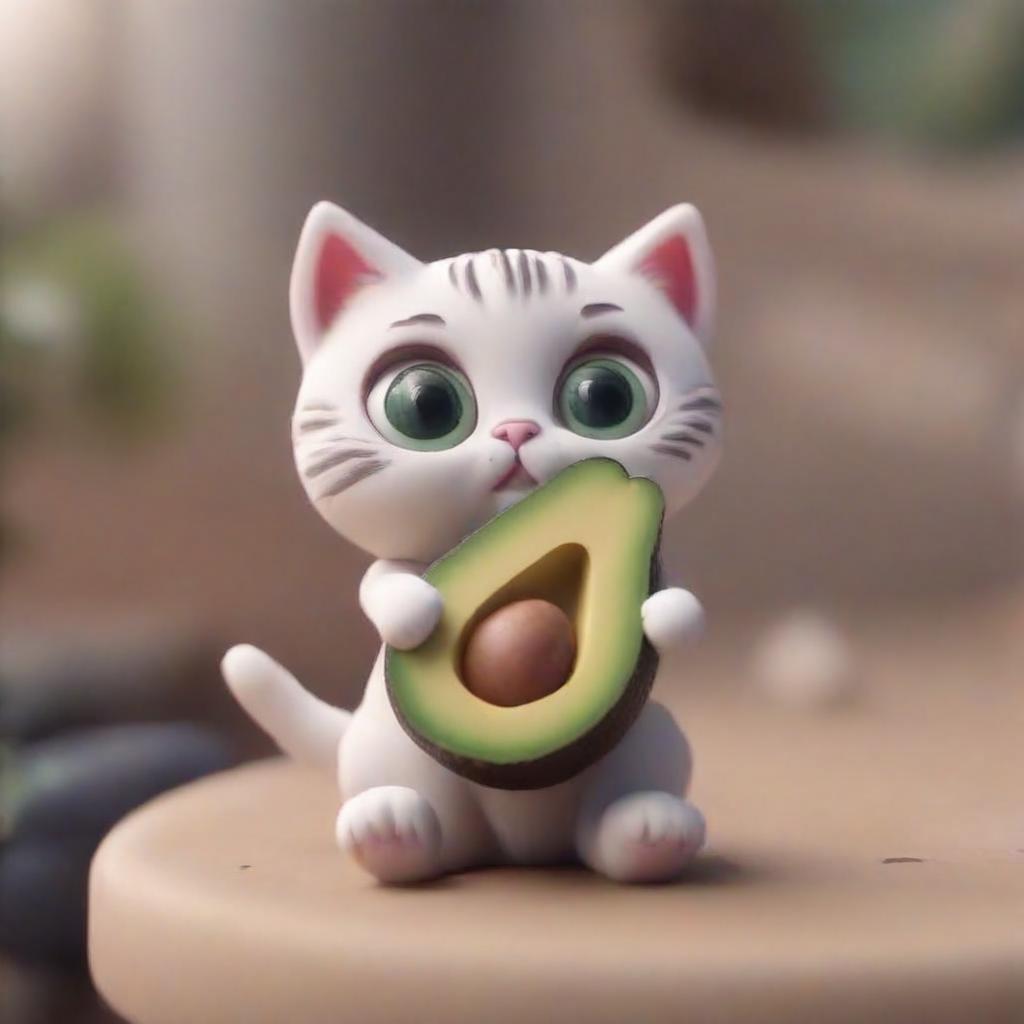}}
        \\
        \\

        {\includegraphics[valign=c, width=\ww]{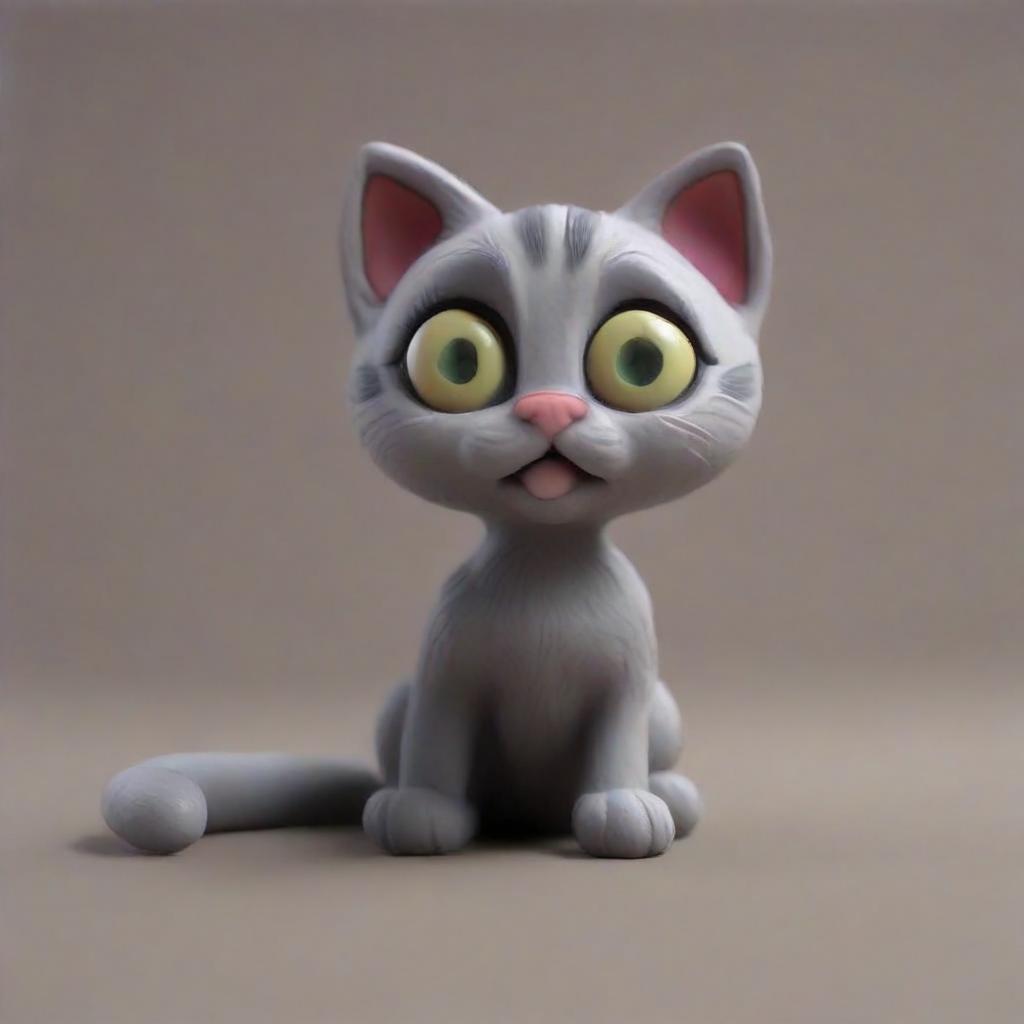}} &
        {\includegraphics[valign=c, width=\ww]{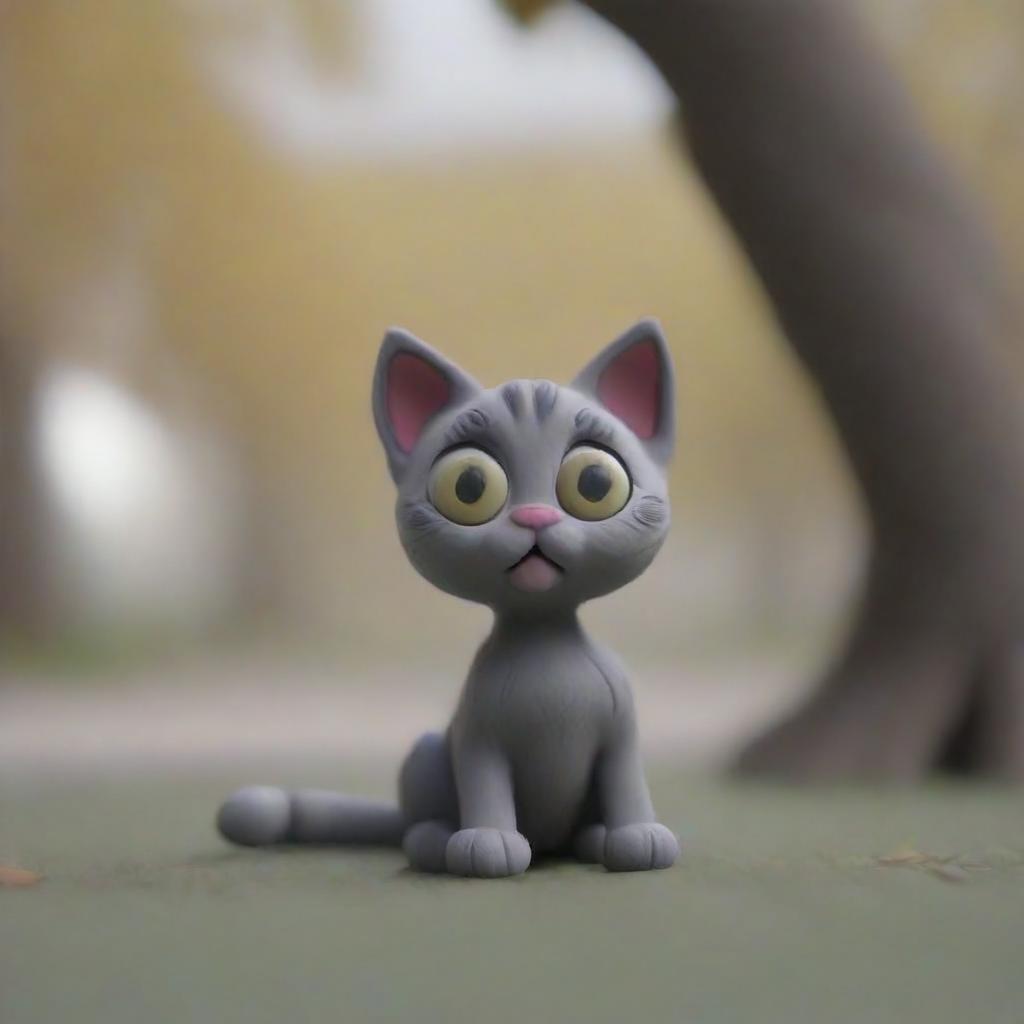}} &
        {\includegraphics[valign=c, width=\ww]{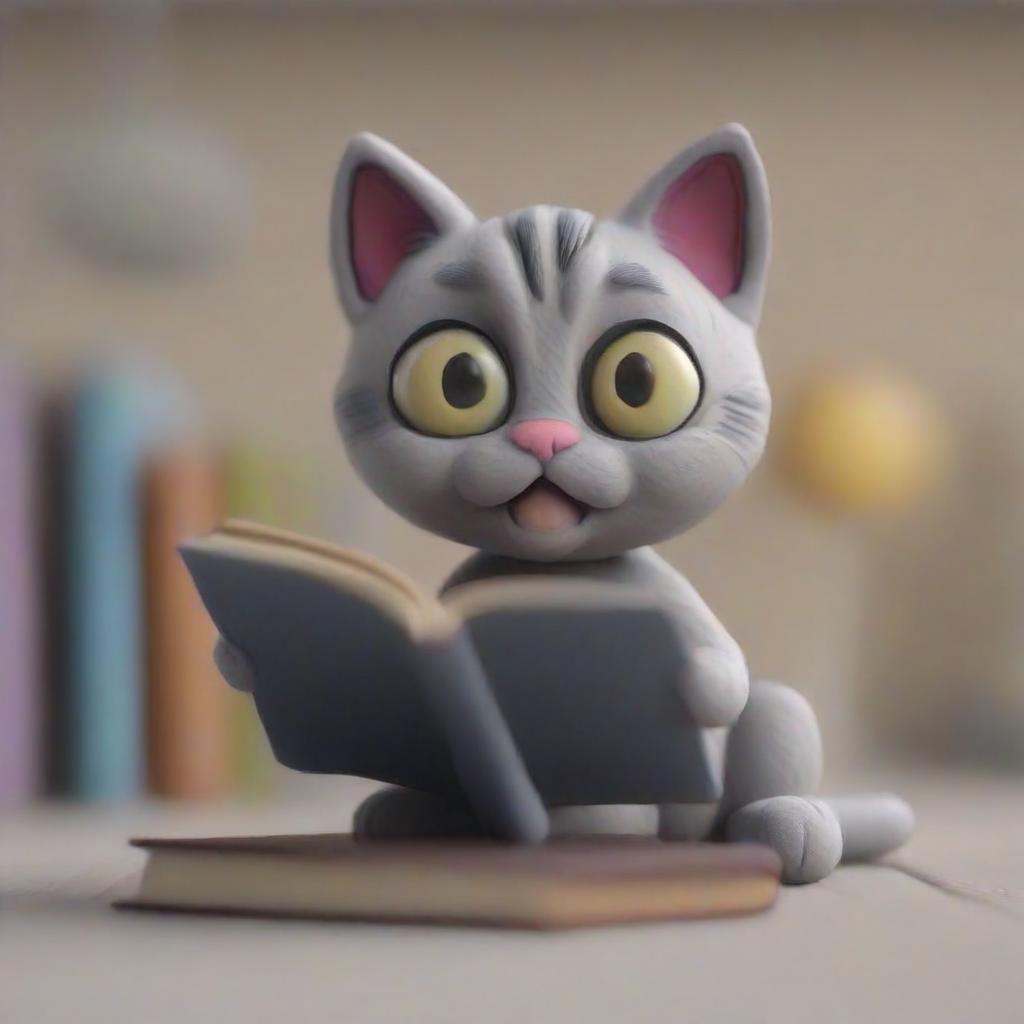}} &
        {\includegraphics[valign=c, width=\ww]{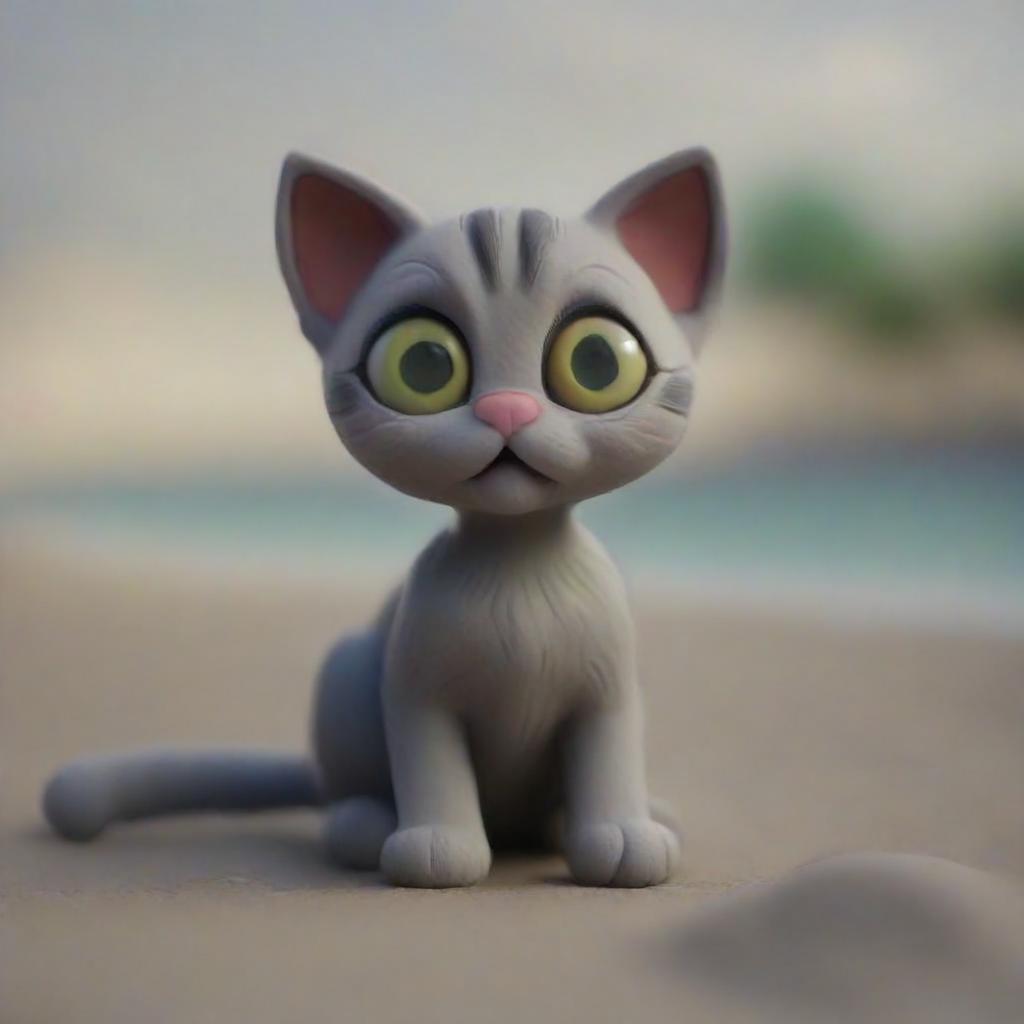}} &
        {\includegraphics[valign=c, width=\ww]{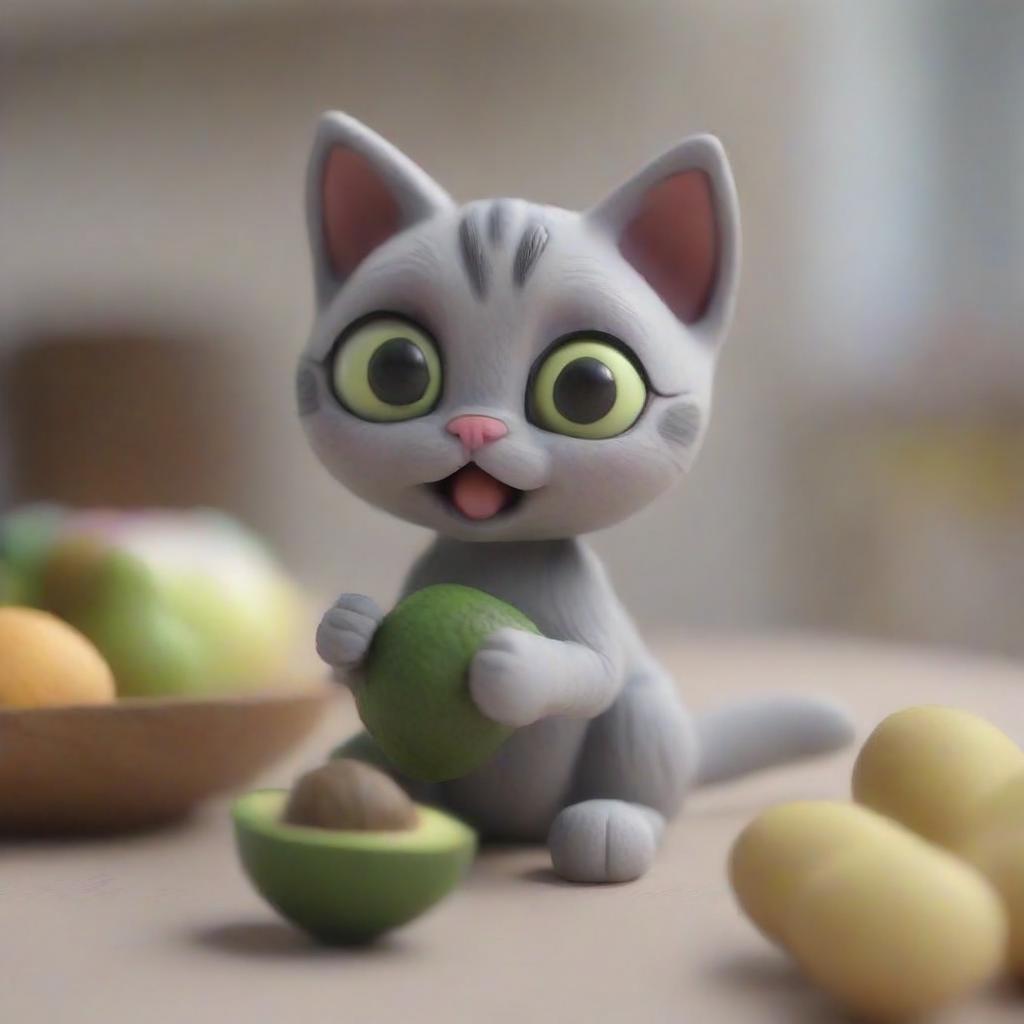}}

    \end{tabular}
    
    \caption{\textbf{Non-determinism.} By running our method multiple times, given the same prompt \textit{``a Plasticine of a cute baby cat with big eyes''}, but using different initial seeds, we obtain different consistent characters corresponding to the text prompt.}
    \label{fig:nondeterminism_cat}
\end{figure*}

%% file: figures/automatic_qualitative_comparison/fig_naive_baselines.tex
\begin{figure*}[t]
    \centering
    \setlength{\tabcolsep}{3.5pt}
    \renewcommand{\arraystretch}{0.4}
    \setlength{\ww}{0.34\columnwidth}
    \begin{tabular}{cccc}
        &
        \textbf{TI multi} &
        \textbf{LoRA DB multi} &
        \textbf{Ours}
        \\

        \rotatebox[origin=c]{90}{\phantom{a}}
        \rotatebox[origin=c]{90}{\textit{``drinking a beer''}} &
        {\includegraphics[valign=c, width=\ww]{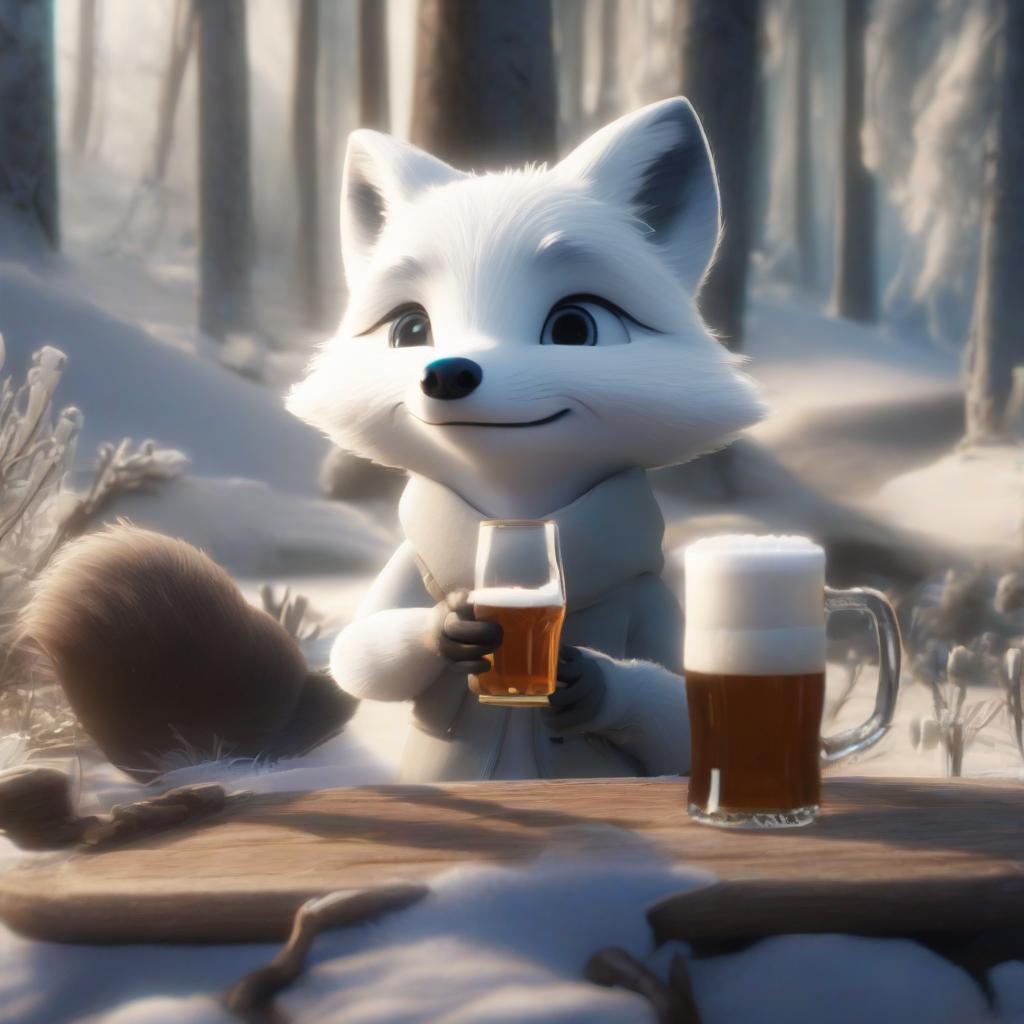}} &
        {\includegraphics[valign=c, width=\ww]{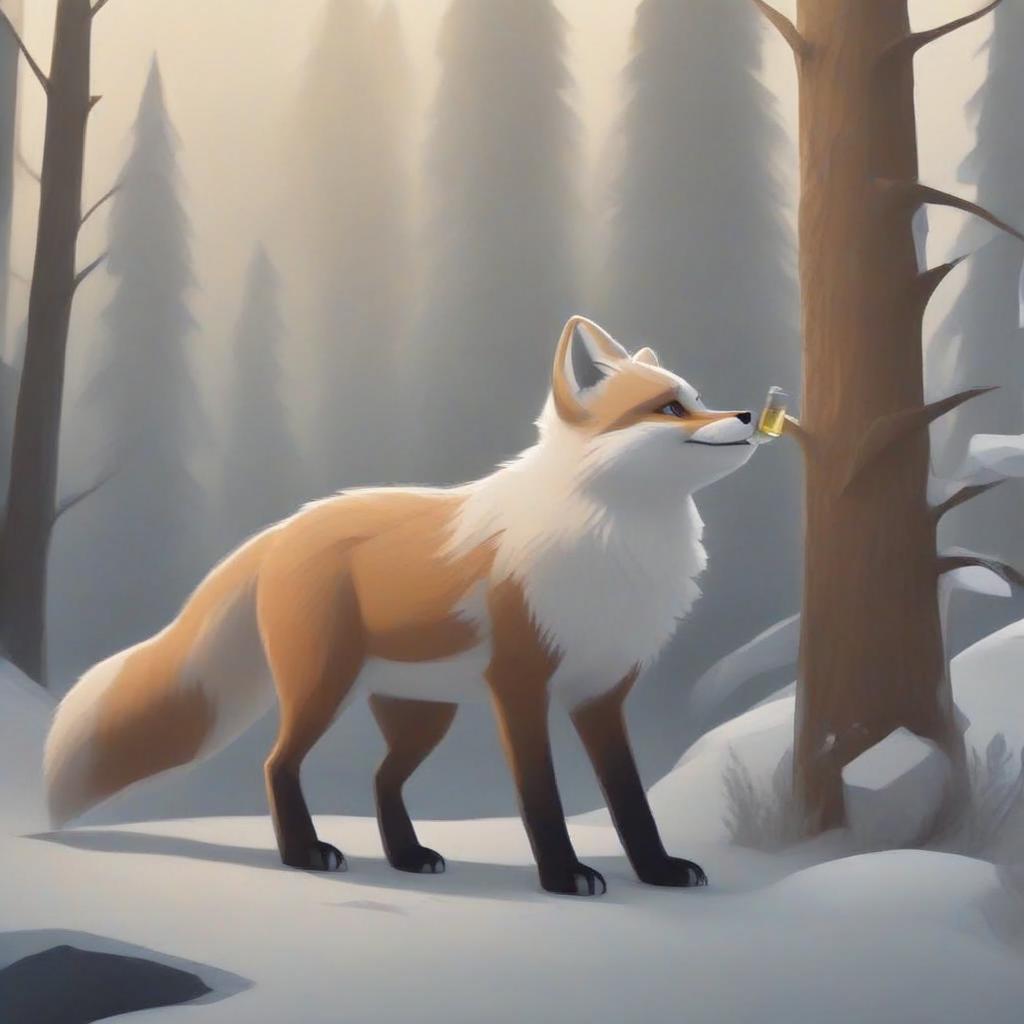}} &
        {\includegraphics[valign=c, width=\ww]{figures/automatic_qualitative_comparison/assets/fox/ours/beer_0.jpg}}
        \\
        \\

        \rotatebox[origin=c]{90}{\textit{``with a city in}}
        \rotatebox[origin=c]{90}{\textit{the background''}} &
        {\includegraphics[valign=c, width=\ww]{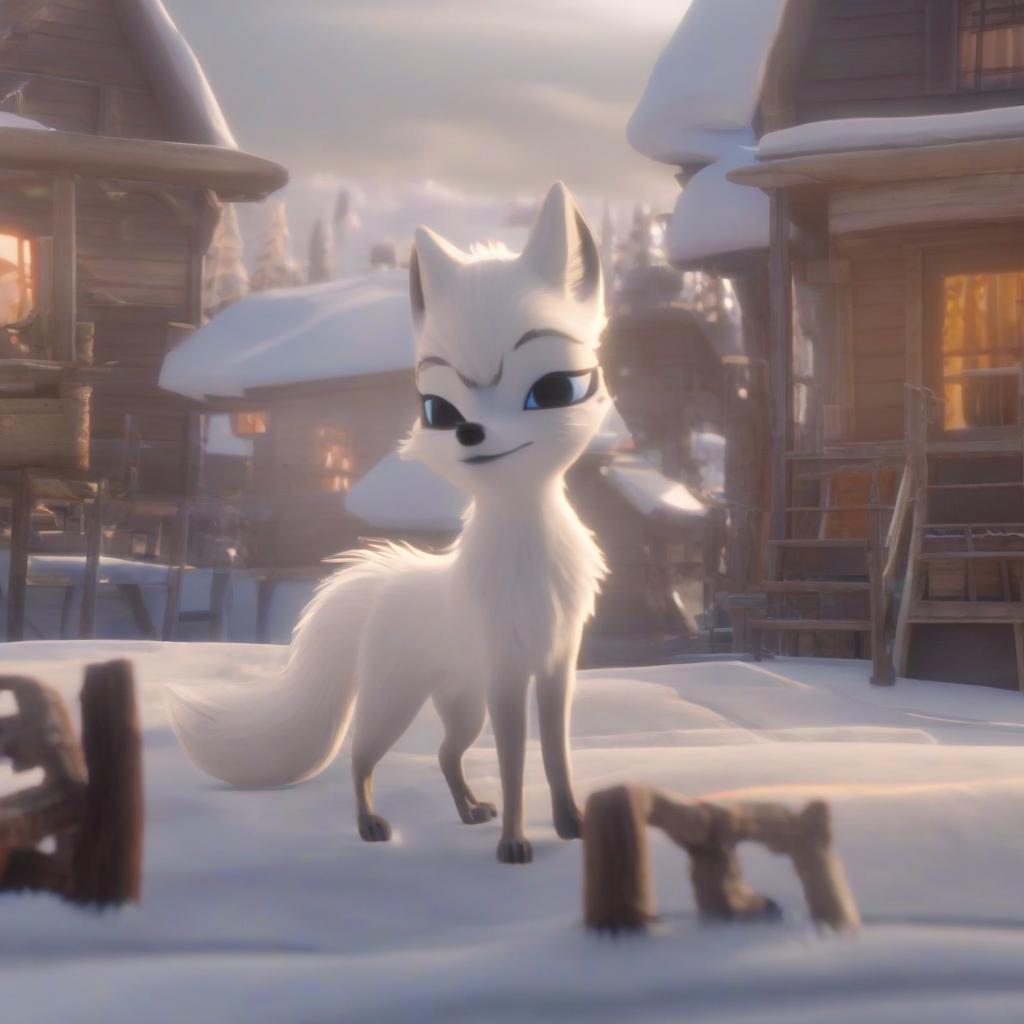}} &
        {\includegraphics[valign=c, width=\ww]{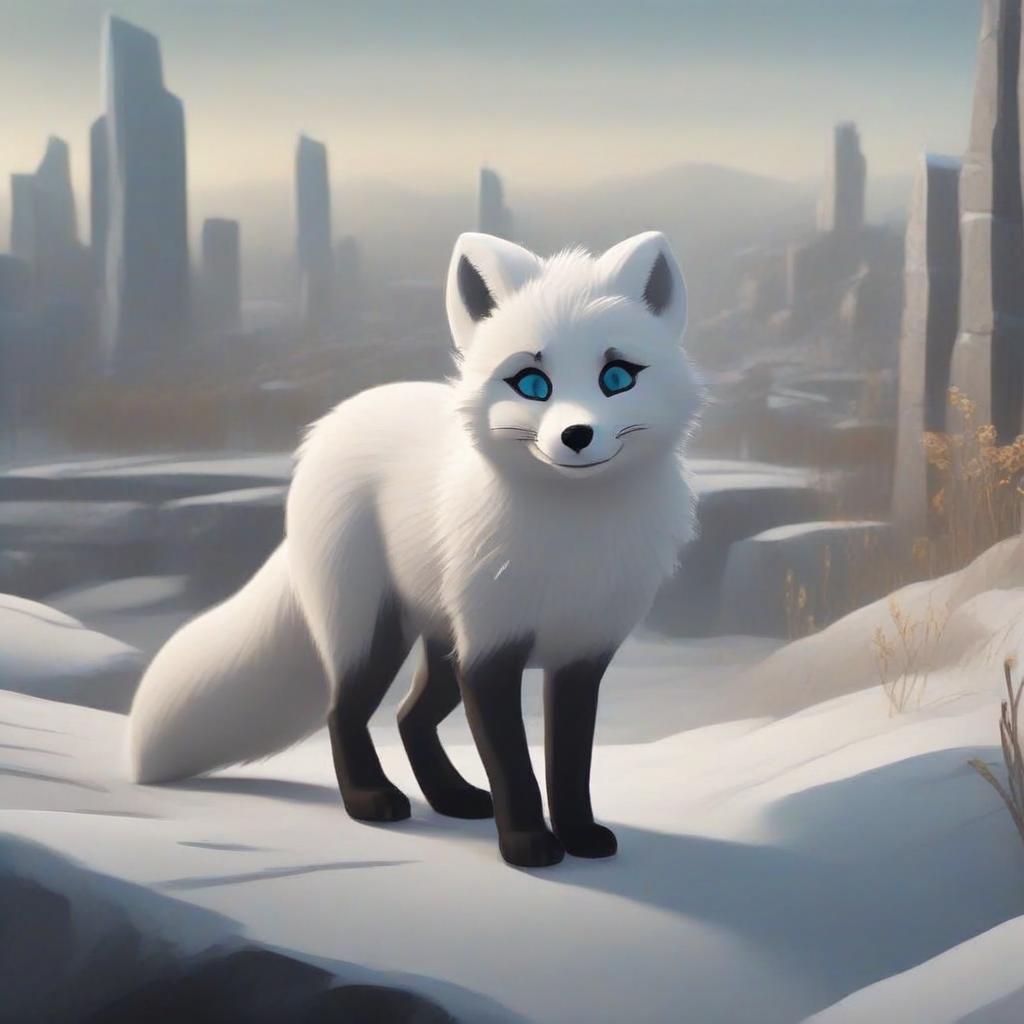}} &
        {\includegraphics[valign=c, width=\ww]{figures/automatic_qualitative_comparison/assets/fox/ours/city_0.jpg}}
        \\
        \\

        \multicolumn{4}{c}{\textit{``a 2D animation of captivating Arctic fox with fluffy fur, bright eyes}}
        \\
        \multicolumn{4}{c}{\textit{and nimble movements, bringing the magic of the icy wilderness}}
        \\
        \multicolumn{4}{c}{\textit{to animated life''}}
        \\
        \\
        \midrule

        \\
        \rotatebox[origin=c]{90}{\phantom{a}}
        \rotatebox[origin=c]{90}{\textit{``eating a burger''}} &
        {\includegraphics[valign=c, width=\ww]{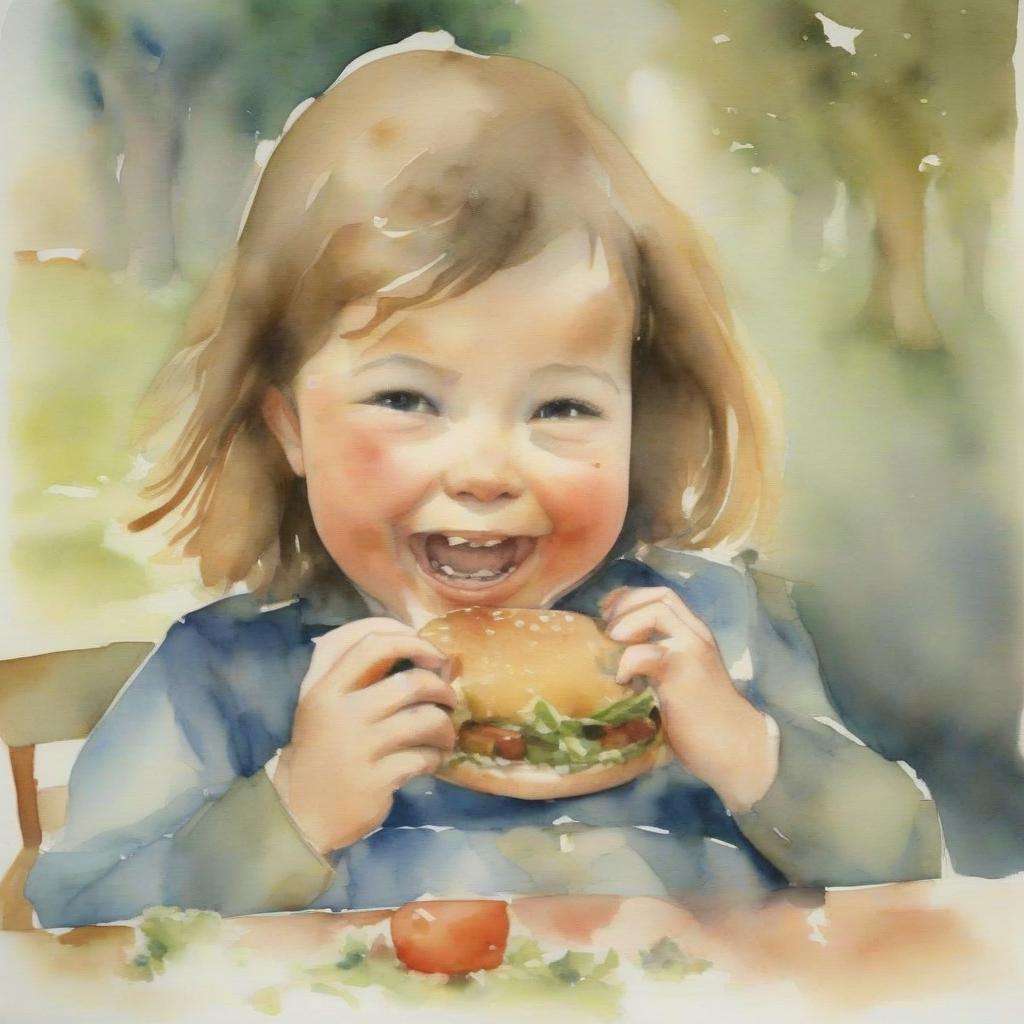}} &
        {\includegraphics[valign=c, width=\ww]{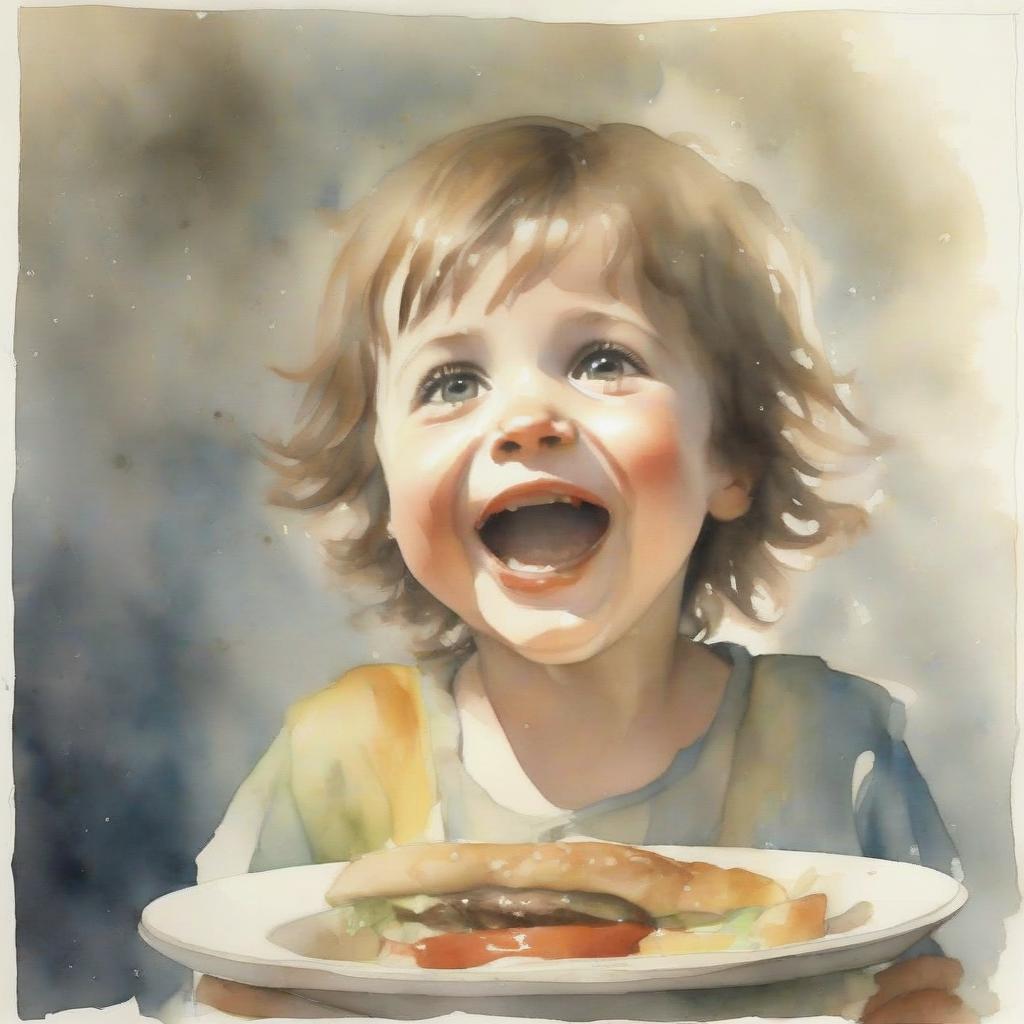}} &
        {\includegraphics[valign=c, width=\ww]{figures/automatic_qualitative_comparison/assets/child/ours/burger.jpg}}
        \\
        \\

        \rotatebox[origin=c]{90}{\textit{``wearing a}}
        \rotatebox[origin=c]{90}{\textit{blue hat''}} &
        {\includegraphics[valign=c, width=\ww]{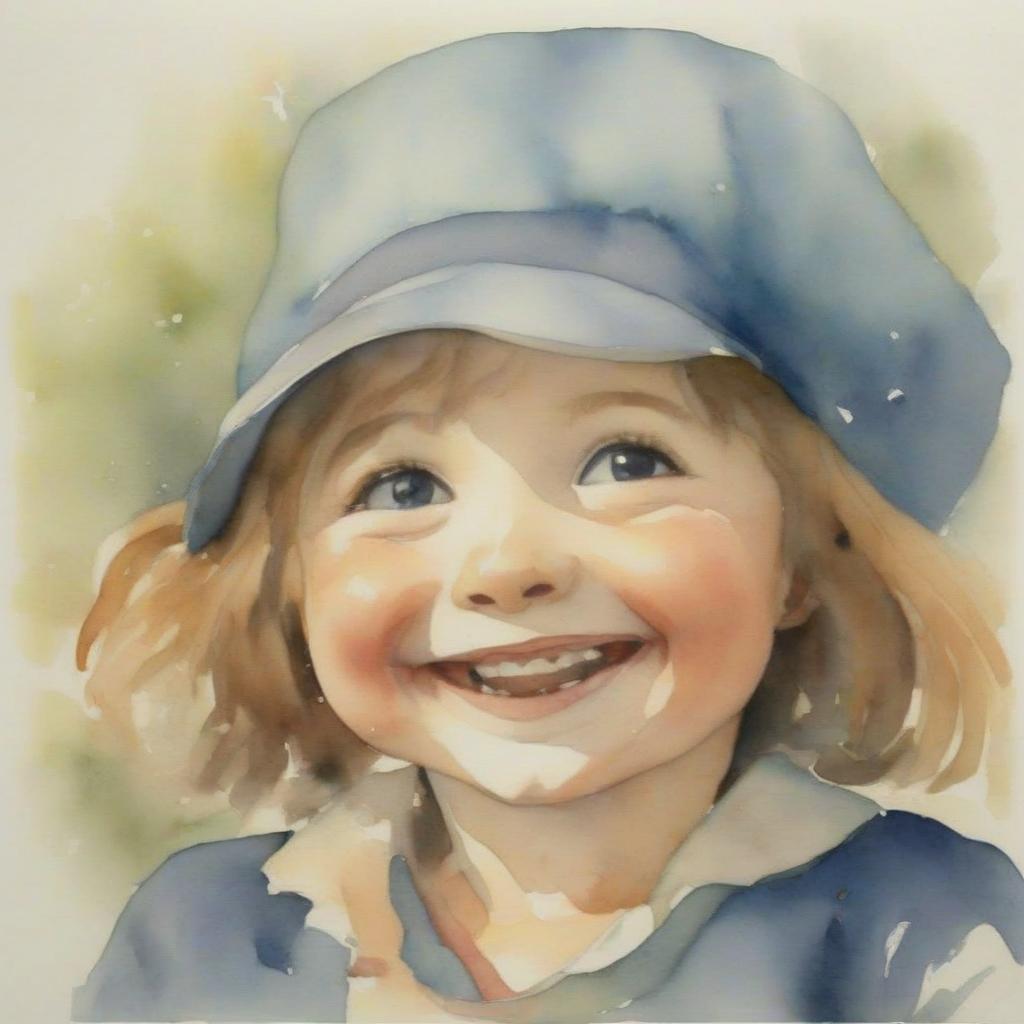}} &
        {\includegraphics[valign=c, width=\ww]{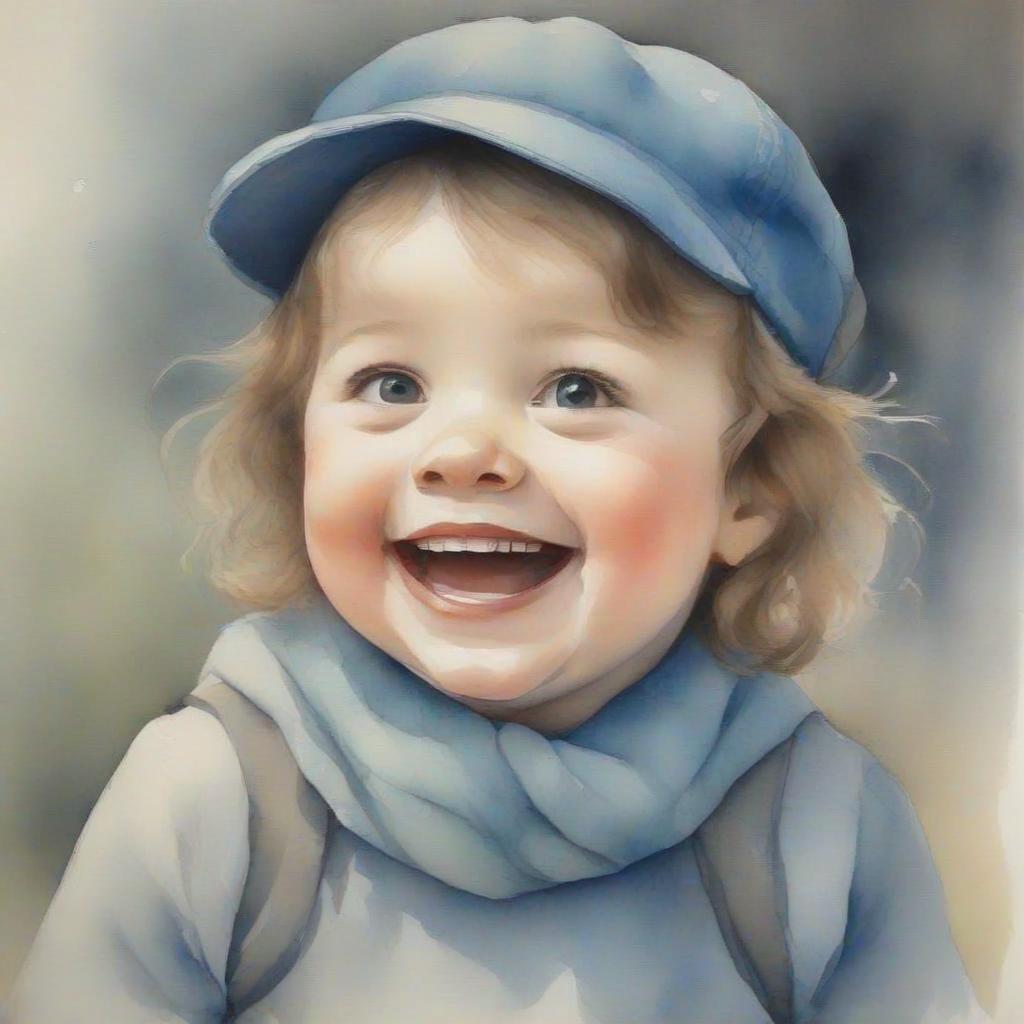}} &
        {\includegraphics[valign=c, width=\ww]{figures/automatic_qualitative_comparison/assets/child/ours/hat.jpg}}
        \\
        \\

        \multicolumn{4}{c}{\textit{``a watercolor portrayal of a joyful child, radiating innocence and}}
        \\
        \multicolumn{4}{c}{\textit{wonder with rosy cheeks and a genuine, wide-eyed smile''}}
        \\
        \\
        \midrule

        \\
        \rotatebox[origin=c]{90}{\textit{``near the Statue}}
        \rotatebox[origin=c]{90}{\textit{of Liberty''}} &
        {\includegraphics[valign=c, width=\ww]{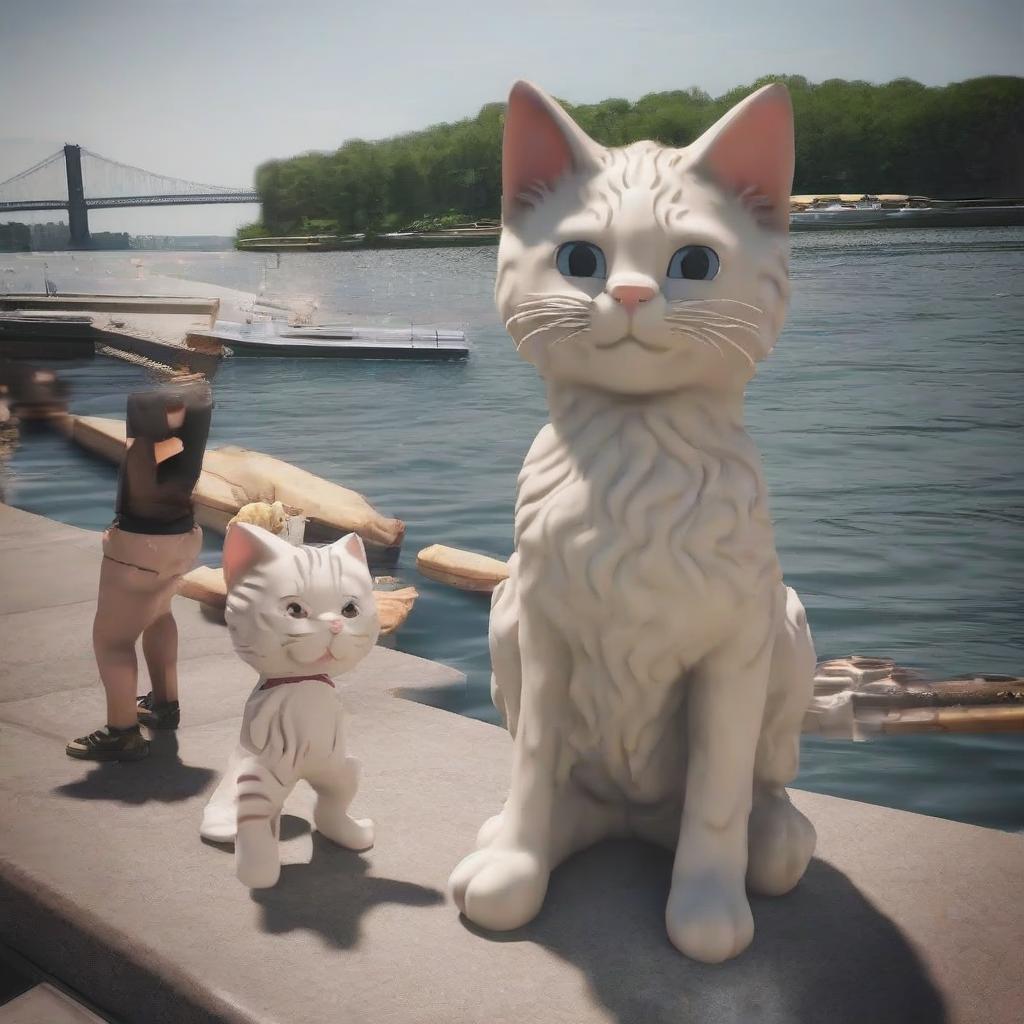}} &
        {\includegraphics[valign=c, width=\ww]{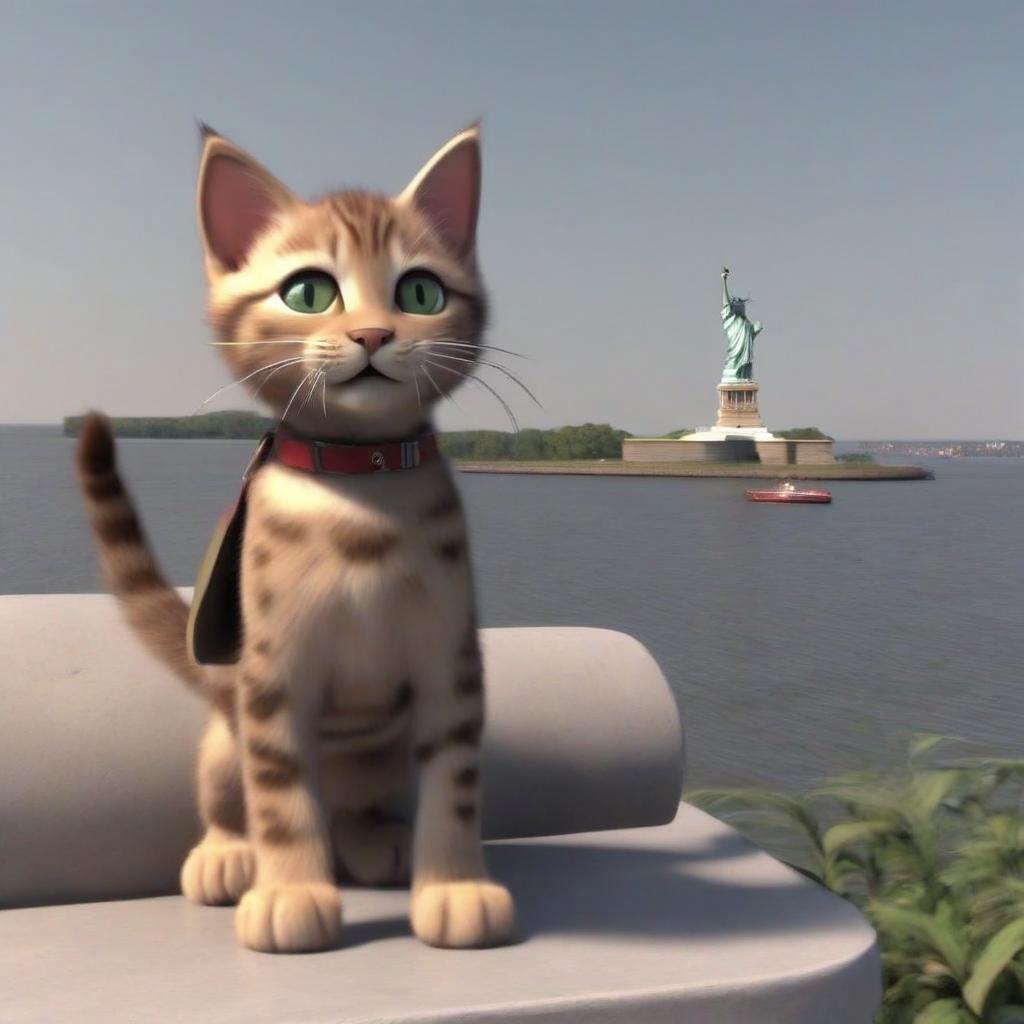}} &
        {\includegraphics[valign=c, width=\ww]{figures/automatic_qualitative_comparison/assets/cat/ours/liberty.jpg}}
        \\
        \\

        \rotatebox[origin=c]{90}{\textit{``as a police}}
        \rotatebox[origin=c]{90}{\textit{officer''}} &
        {\includegraphics[valign=c, width=\ww]{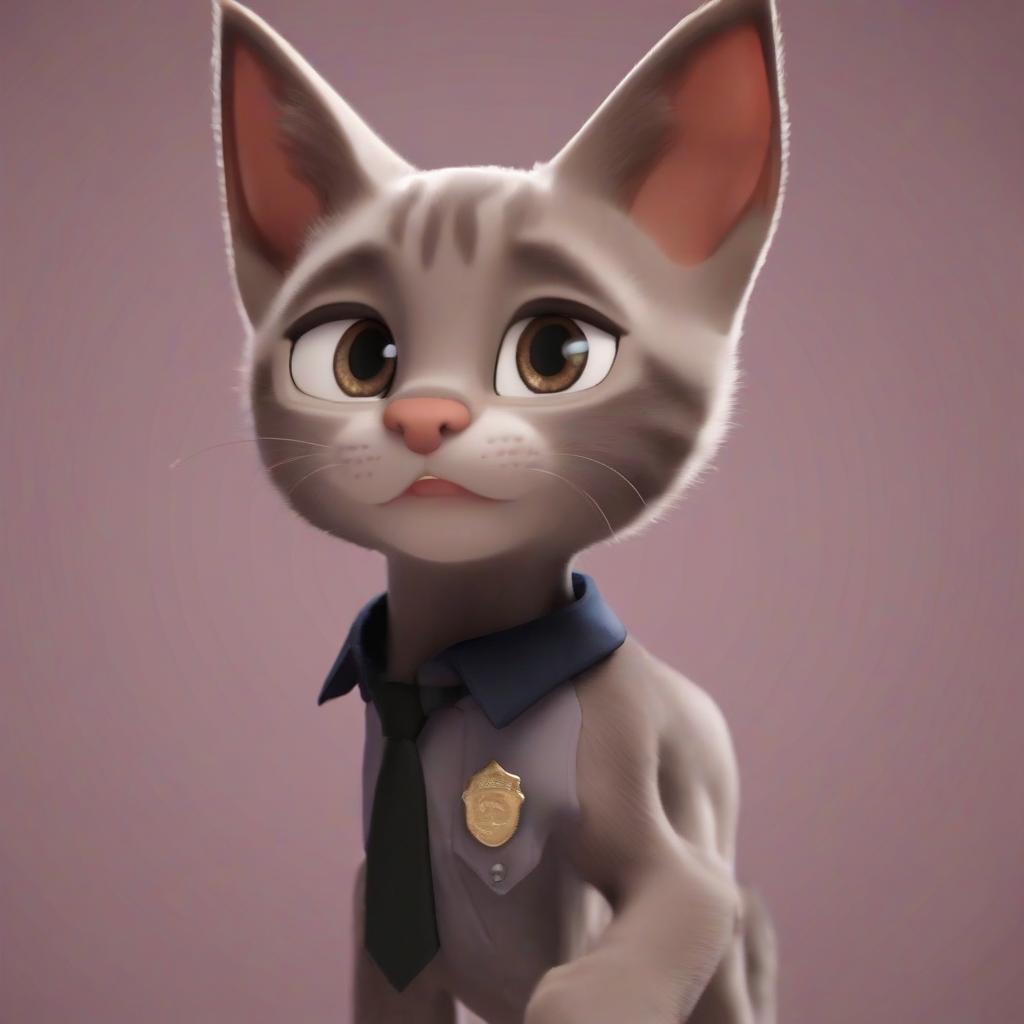}} &
        {\includegraphics[valign=c, width=\ww]{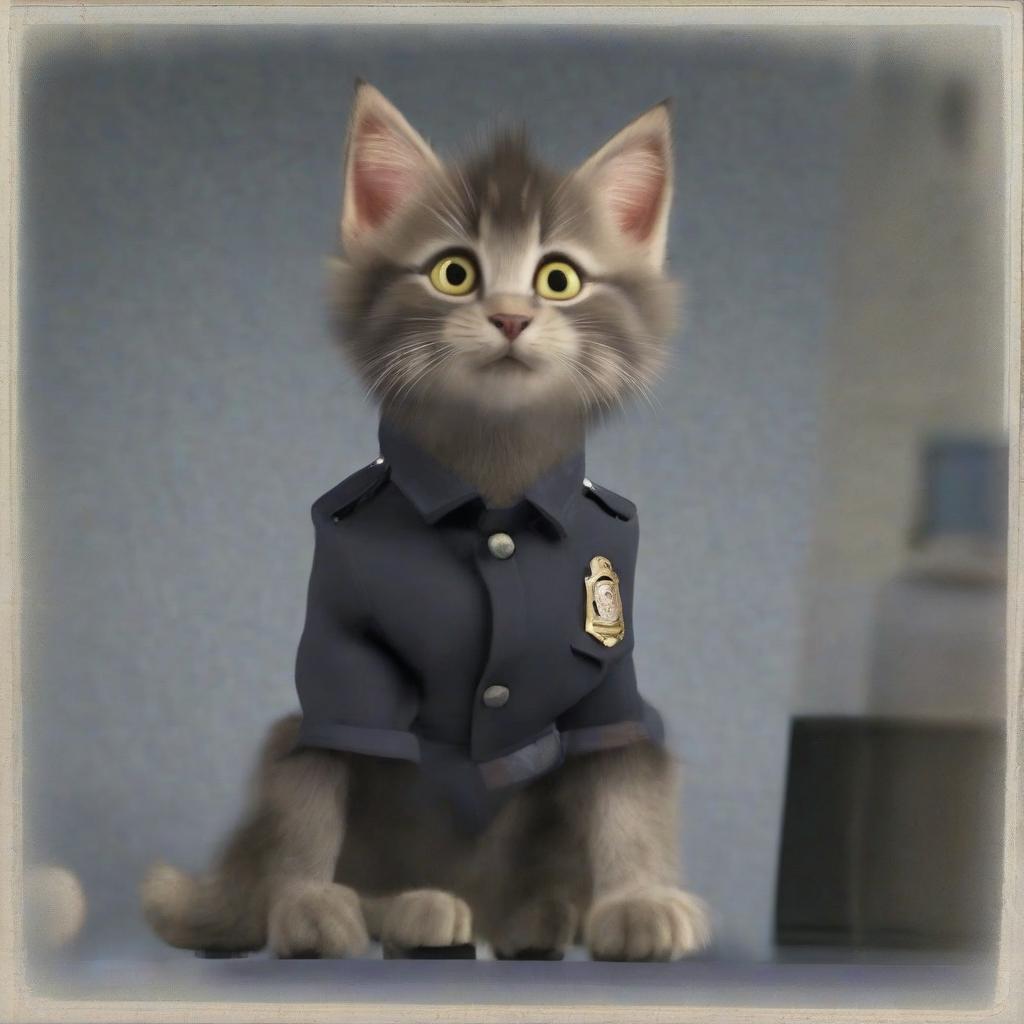}} &
        {\includegraphics[valign=c, width=\ww]{figures/automatic_qualitative_comparison/assets/cat/ours/police.jpg}}
        \\
        \\

        \multicolumn{4}{c}{\textit{``a 3D animation of a playful kitten, with bright eyes and}}
        \\
        \multicolumn{4}{c}{\textit{a mischievous expression, embodying youthful curiosity and joy''}}

    \end{tabular}
    
    \caption{\textbf{Qualitative comparison to \naive{} baselines.} We tested two additional \naive{} baselines against our method: TI~\cite{Gal2022AnII} and LoRA DB~\cite{lora_diffusion} that were trained on a small dataset of 5 images generated from the same prompt. The baselines are referred to as \emph{TI multi} (left column) and \emph{LoRA DB multi} (middle column). As can be seen, both of these baselines fail to extract a consistent identity.}
    \label{fig:qualitative_naive_baselines}
\end{figure*}

%% file: figures/quantitative_comparison/naive_baselines_comparison.tex
\begin{figure}[tp]
    \begin{tikzpicture} [thick,scale=0.9, every node/.style={scale=1}]

        \def\MarkSize{.75em}
        \protected\def\ToWest#1{
          \llap{#1\kern\MarkSize}\phantom{#1}
        }
        \protected\def\ToSouth#1{
          \sbox0{#1}
          \smash{
            \rlap{
              \kern-.5\dimexpr\wd0 + \MarkSize\relax
              \lower\dimexpr.375em+\ht0\relax\copy0
            }
          }
          \hphantom{#1}
        }

        \begin{axis}[
            xlabel={Automatic prompt similarity ($\rightarrow$)}, 
            ylabel={Automatic identity consistency ($\rightarrow$)},
            compat=newest,
            xmax=0.202,
        ]
            \addplot[
                scatter/classes={a={blue}, b={red}, c={green}, o={orange}},
                scatter,
                mark=*, 
                only marks, 
                scatter src=explicit symbolic,
                nodes near coords*={\Label},
                visualization depends on={value \thisrow{label} \as \Label}
            ] table [meta=class] {
                x y class label
                0.1869359639 0.7645850891 o \scriptsize{TI multi}
                0.1965330373 0.72537582 o \scriptsize{LoRA DB multi}
                0.1657731262 0.8447209117 c \scriptsize{Ours}
            };
        \end{axis}
    \end{tikzpicture}

    \caption{\textbf{Comparison to \naive{} baselines.} We tested two additional \naive{} baselines against our method: TI~\cite{Gal2022AnII} and LoRA DB~\cite{lora_diffusion} that were trained on a small dataset of 5 images generated from the same prompt. The baselines are referred to as \emph{TI multi} and \emph{LoRA DB multi}.
    Our automatic testing procedure, described in
    \Cref{sec:comparisons},
    measures identity consistency and prompt similarity. As can be seen, both of these baselines fail to achieve high identity consistency.}
    \label{fig:naive_baselines_comparison}
\end{figure}
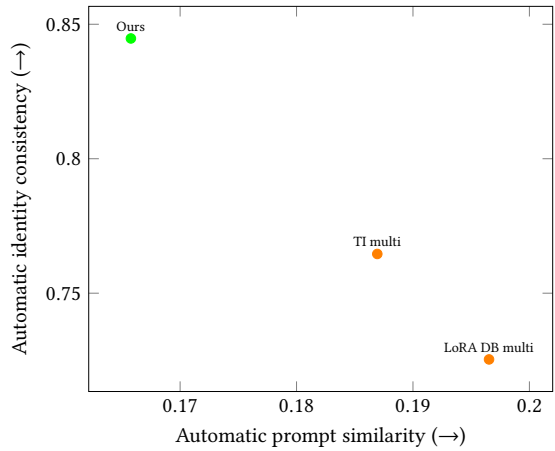

%% file: figures/dalle3/fig_dalle3.tex
\begin{figure}[tp]
    \vspace{30px}  %
    \centering
    \setlength{\tabcolsep}{0.8pt}
    \renewcommand{\arraystretch}{0.5}
    \setlength{\ww}{0.224\columnwidth}
    \begin{tabular}{ccccc}
        &&&&
        \scriptsize{\textit{``holding an}}
        \\

        &
        \scriptsize{\textit{``in the park''}} &
        \scriptsize{\textit{``reading a book''}} &
        \scriptsize{\textit{``at the beach''}} &
        \scriptsize{\textit{avocado''}}
        \\

        \rotatebox[origin=c]{90}{\scriptsize{\DALLE{} 3}} &
        {\includegraphics[valign=c, width=\ww]{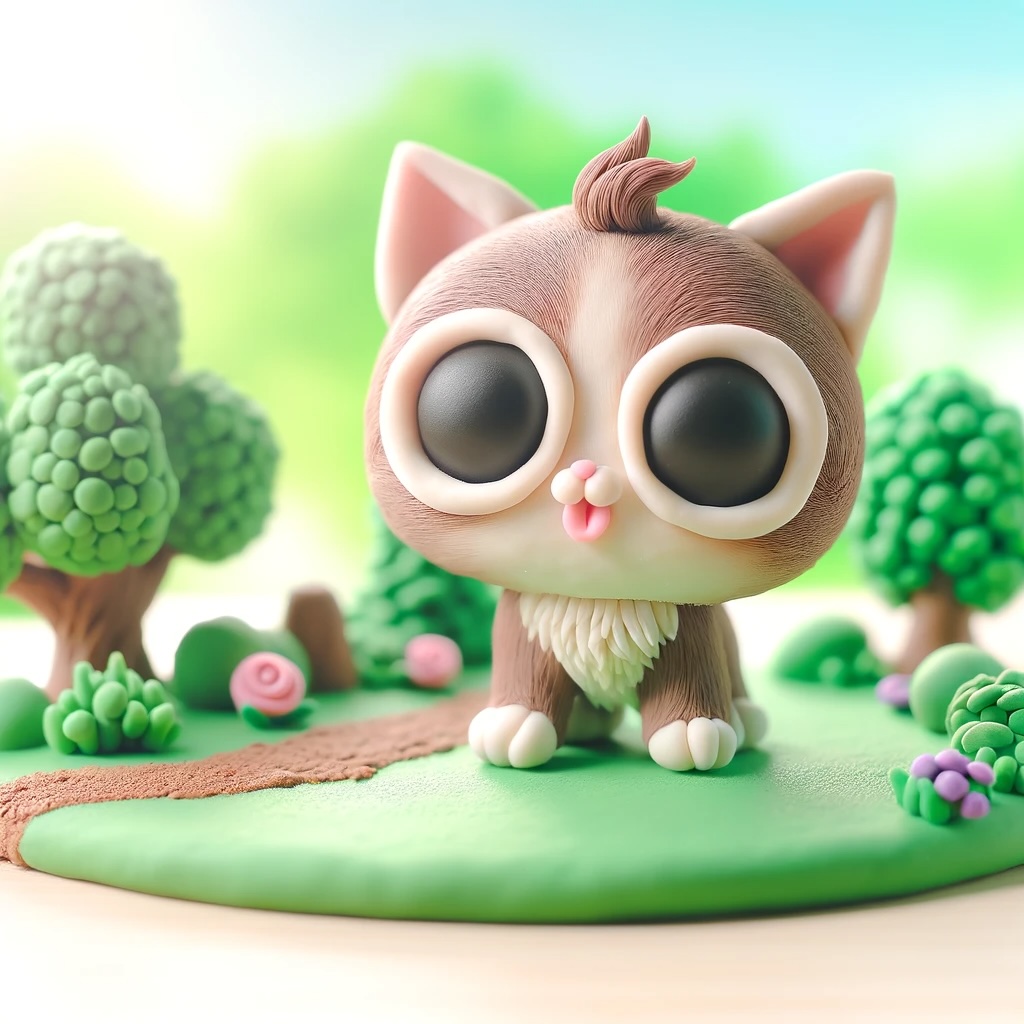}} &
        {\includegraphics[valign=c, width=\ww]{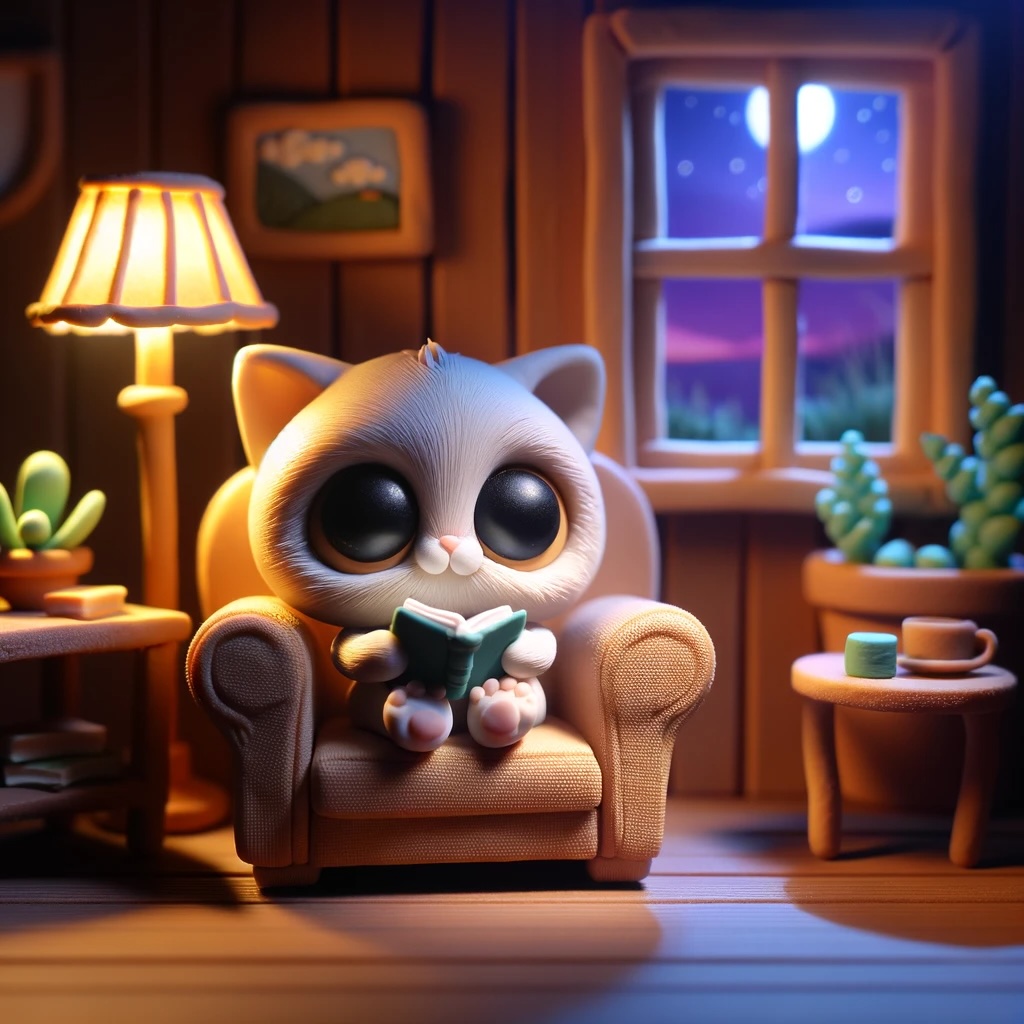}} &
        {\includegraphics[valign=c, width=\ww]{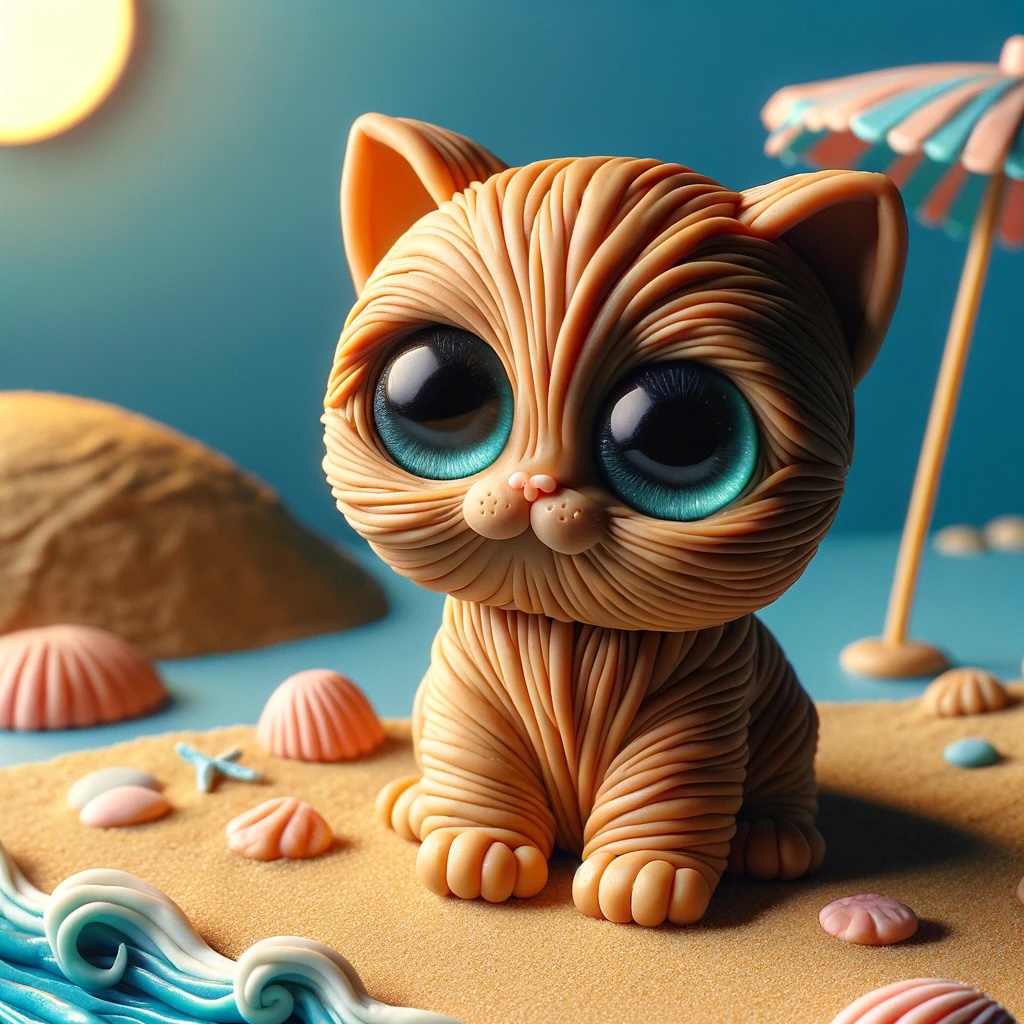}} &
        {\includegraphics[valign=c, width=\ww]{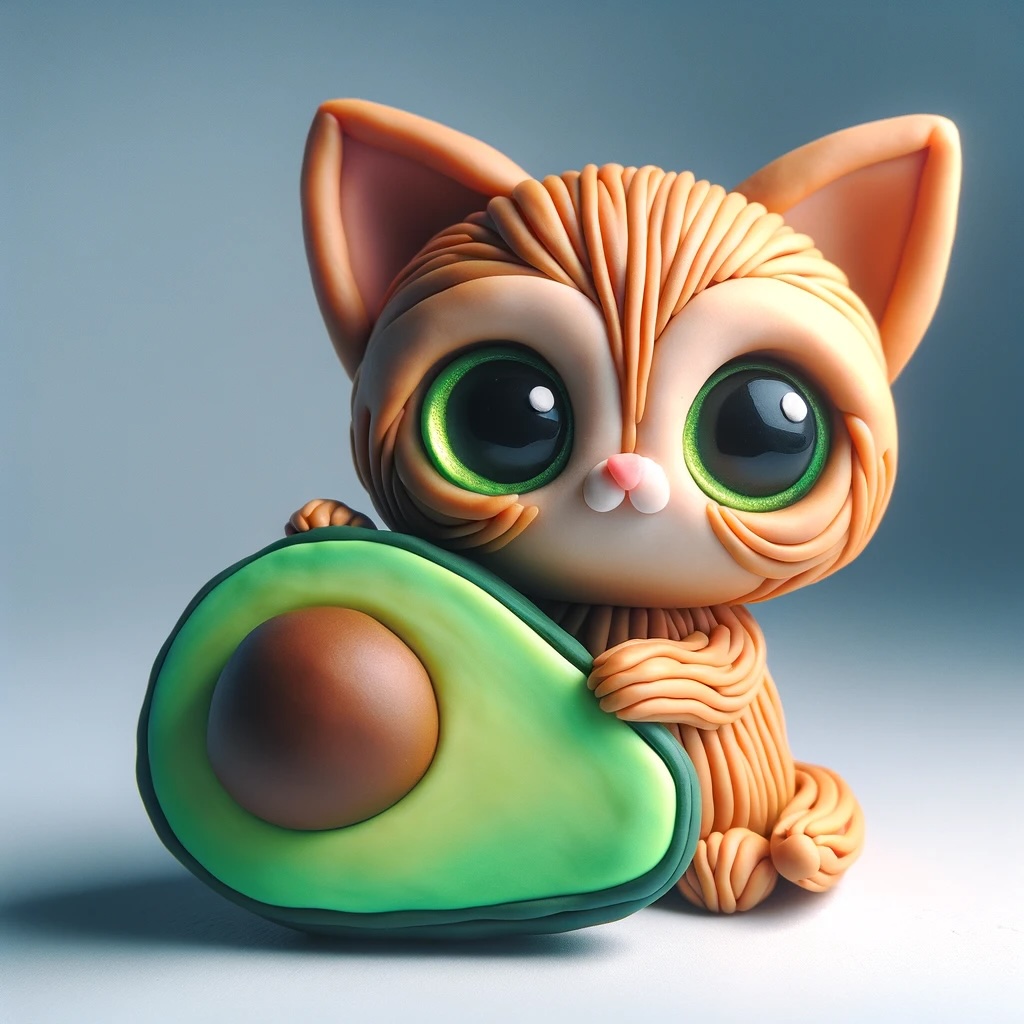}}
        \\
        \\

        \rotatebox[origin=c]{90}{\scriptsize{Ours}} &
        {\includegraphics[valign=c, width=\ww]{figures/nondeterminism/cat/instance1/in_the_park.jpg}} &
        {\includegraphics[valign=c, width=\ww]{figures/nondeterminism/cat/instance1/reading_a_book.jpg}} &
        {\includegraphics[valign=c, width=\ww]{figures/nondeterminism/cat/instance1/at_the_beach.jpg}} &
        {\includegraphics[valign=c, width=\ww]{figures/nondeterminism/cat/instance1/holding_an_avocado.jpg}}

    \end{tabular}
    
    \caption{\textbf{\DALLE{} 3 comparison.} We attempted to create a consistent character using the commercial ChatGPT Plus system, for the given prompt \textit{``a Plasticine of a cute baby cat with big eyes''}. As can be seen, the \DALLE{}~3 \cite{BetkerImprovingIG} generated characters share only some of the characteristics (\eg, big eyes) but not all of them (\eg, colors, textures and shapes).}
    \label{fig:dalle3_comparison}
\end{figure}

%% file: figures/quantitative_comparison/feature_extractors_comparison.tex
\begin{figure}[tp]
    \begin{tikzpicture} [thick,scale=0.9, every node/.style={scale=1}]

        \def\MarkSize{.75em}
        \protected\def\ToWest#1{
          \llap{#1\kern\MarkSize}\phantom{#1}
        }
        \protected\def\ToSouth#1{
          \sbox0{#1}
          \smash{
            \rlap{
              \kern-.5\dimexpr\wd0 + \MarkSize\relax
              \lower\dimexpr.375em+\ht0\relax\copy0
            }
          }
          \hphantom{#1}
        }

        \begin{axis}[
            xlabel={Automatic prompt similarity ($\rightarrow$)}, 
            ylabel={Automatic identity consistency ($\rightarrow$)},
            compat=newest,
            xmin=0.156,
            xmax=0.1755,
        ]
            \addplot[
                scatter/classes={a={blue}, b={red}, c={green}, o={orange}},
                scatter,
                mark=*, 
                only marks, 
                scatter src=explicit symbolic,
                nodes near coords*={\Label},
                visualization depends on={value \thisrow{label} \as \Label}
            ] table [meta=class] {
                x y class label
                0.1734933557 0.8116963196 o \scriptsize{Ours w CLIP}
                0.159085077 0.8658229405 o \scriptsize{Ours w DINOv1}
                0.1657731262 0.8447209117 c \scriptsize{Ours w DINOv2}
            };
        \end{axis}
    \end{tikzpicture}

    \caption{\textbf{Comparison of feature extractors.} We tested two additional feature extractors in our method: DINOv1 \cite{Caron2021EmergingPI} and CLIP \cite{Radford2021LearningTV}. Our automatic testing procedure, described in \Cref{sec:comparisons},
    measures identity consistency and prompt similarity. As can be seen, DINOv1 produces higher identity consistency by sacrificing prompt similarity, while CLIP results in higher prompt similarity at the expense of lower identity consistency. In practice, however, the DINOv1 results are similar to those obtained with DINOv2 features in terms of prompt adherence (see \Cref{fig:automatic_qualitative_feature_extractor_comparison}).}
    \label{fig:feature_extractors_comparison}
\end{figure}
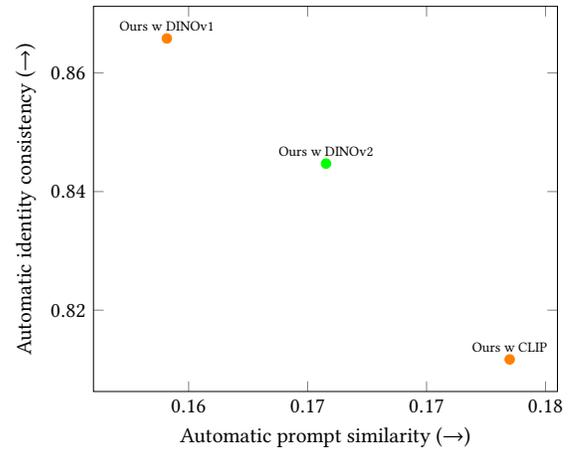

%% file: figures/automatic_qualitative_comparison/fig_feature_extractors.tex
\begin{figure*}[t]
    \centering
    \setlength{\tabcolsep}{3.5pt}
    \renewcommand{\arraystretch}{0.4}
    \setlength{\ww}{0.34\columnwidth}
    \begin{tabular}{cccc}
        &
        \textbf{Ours with CLIP} &
        \textbf{Ours with DINOv1} &
        \textbf{Ours}
        \\

        \rotatebox[origin=c]{90}{\phantom{a}}
        \rotatebox[origin=c]{90}{\textit{``drinking a beer''}} &
        {\includegraphics[valign=c, width=\ww]{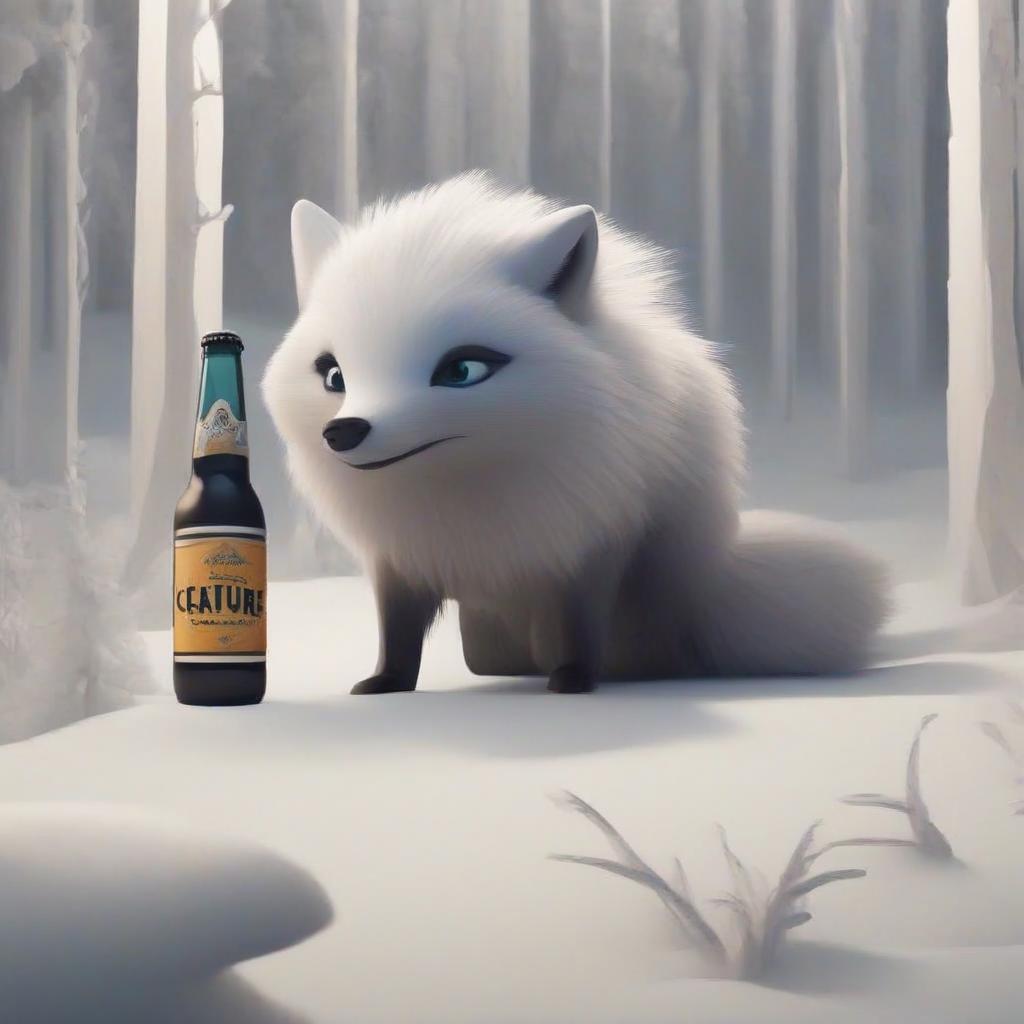}} &
        {\includegraphics[valign=c, width=\ww]{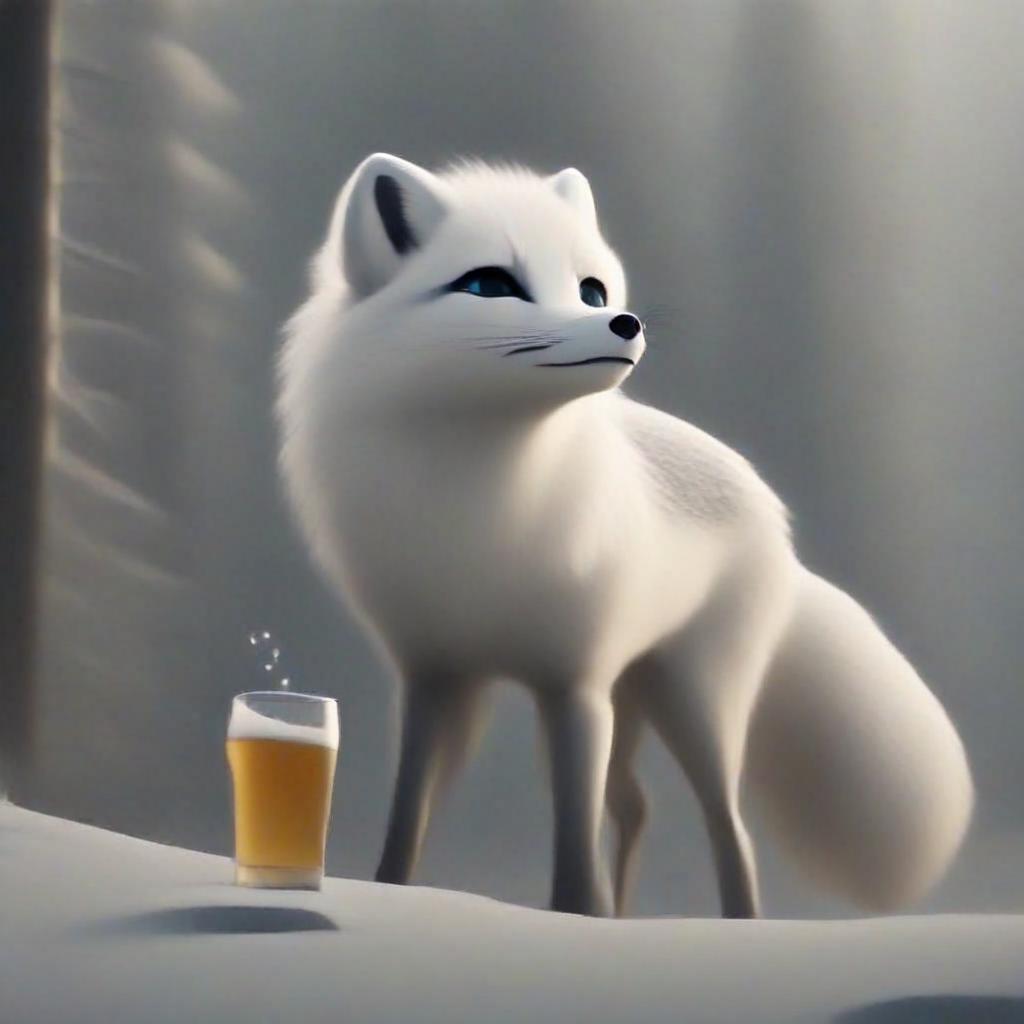}} &
        {\includegraphics[valign=c, width=\ww]{figures/automatic_qualitative_comparison/assets/fox/ours/beer_0.jpg}}
        \\
        \\

        \rotatebox[origin=c]{90}{\textit{``with a city in}}
        \rotatebox[origin=c]{90}{\textit{the background''}} &
        {\includegraphics[valign=c, width=\ww]{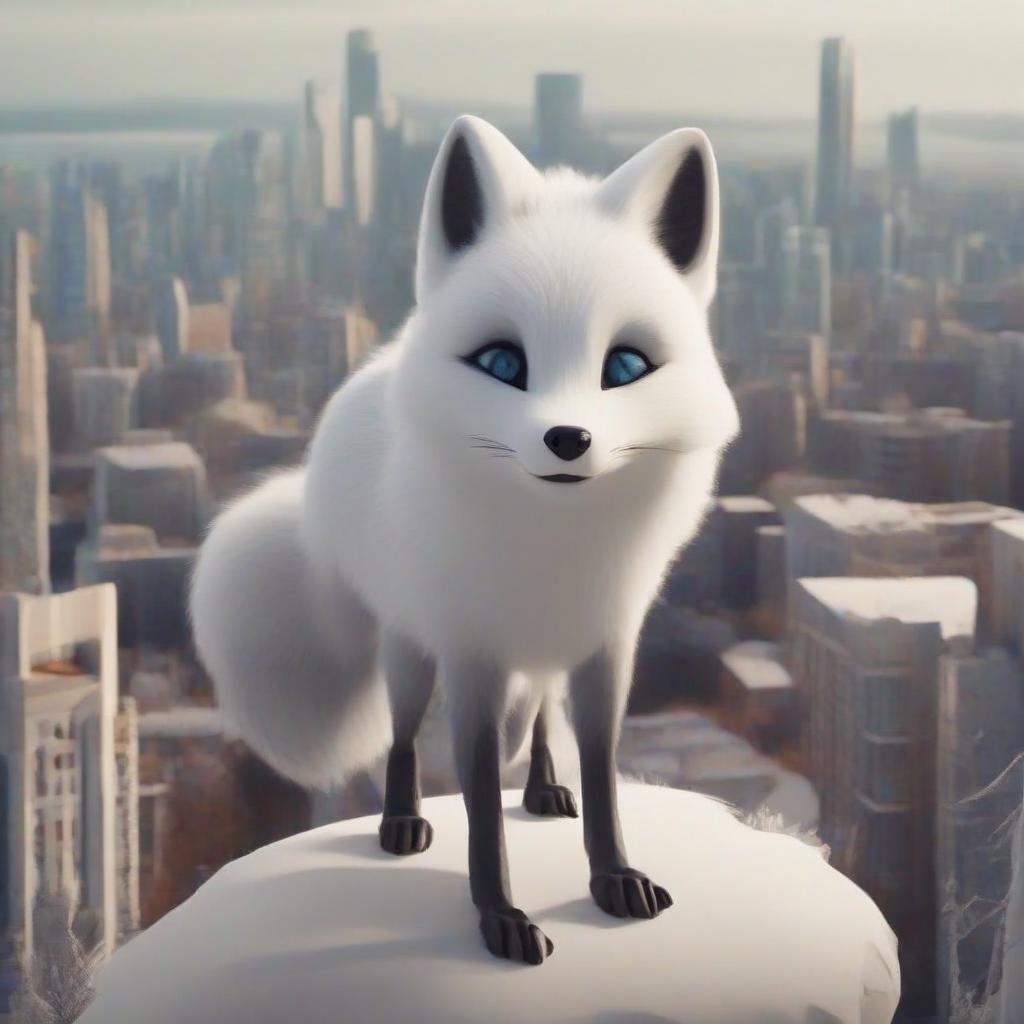}} &
        {\includegraphics[valign=c, width=\ww]{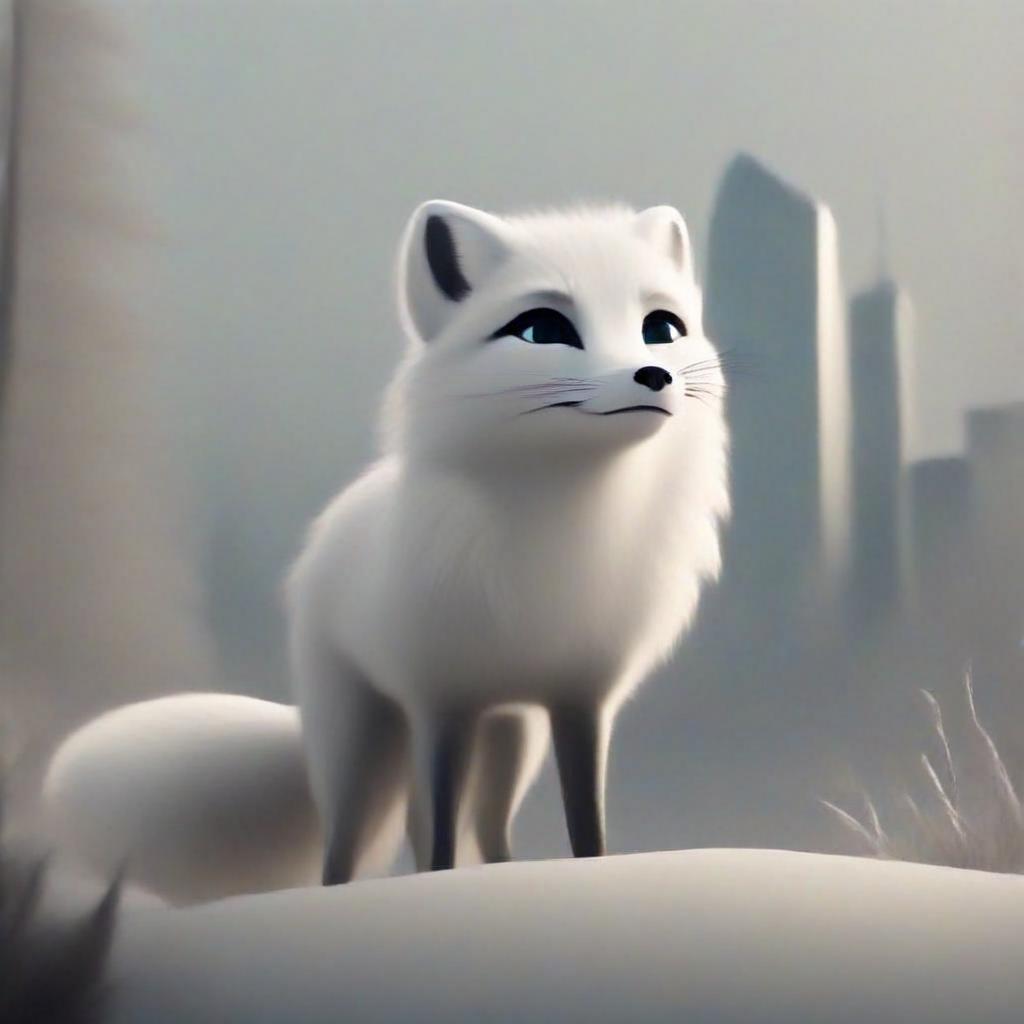}} &
        {\includegraphics[valign=c, width=\ww]{figures/automatic_qualitative_comparison/assets/fox/ours/city_0.jpg}}
        \\
        \\

        \multicolumn{4}{c}{\textit{``a 2D animation of captivating Arctic fox with fluffy fur, bright eyes}}
        \\
        \multicolumn{4}{c}{\textit{and nimble movements, bringing the magic of the icy wilderness}}
        \\
        \multicolumn{4}{c}{\textit{to animated life''}}
        \\
        \\
        \midrule

        \\
        \rotatebox[origin=c]{90}{\phantom{a}}
        \rotatebox[origin=c]{90}{\textit{``eating a burger''}} &
        {\includegraphics[valign=c, width=\ww]{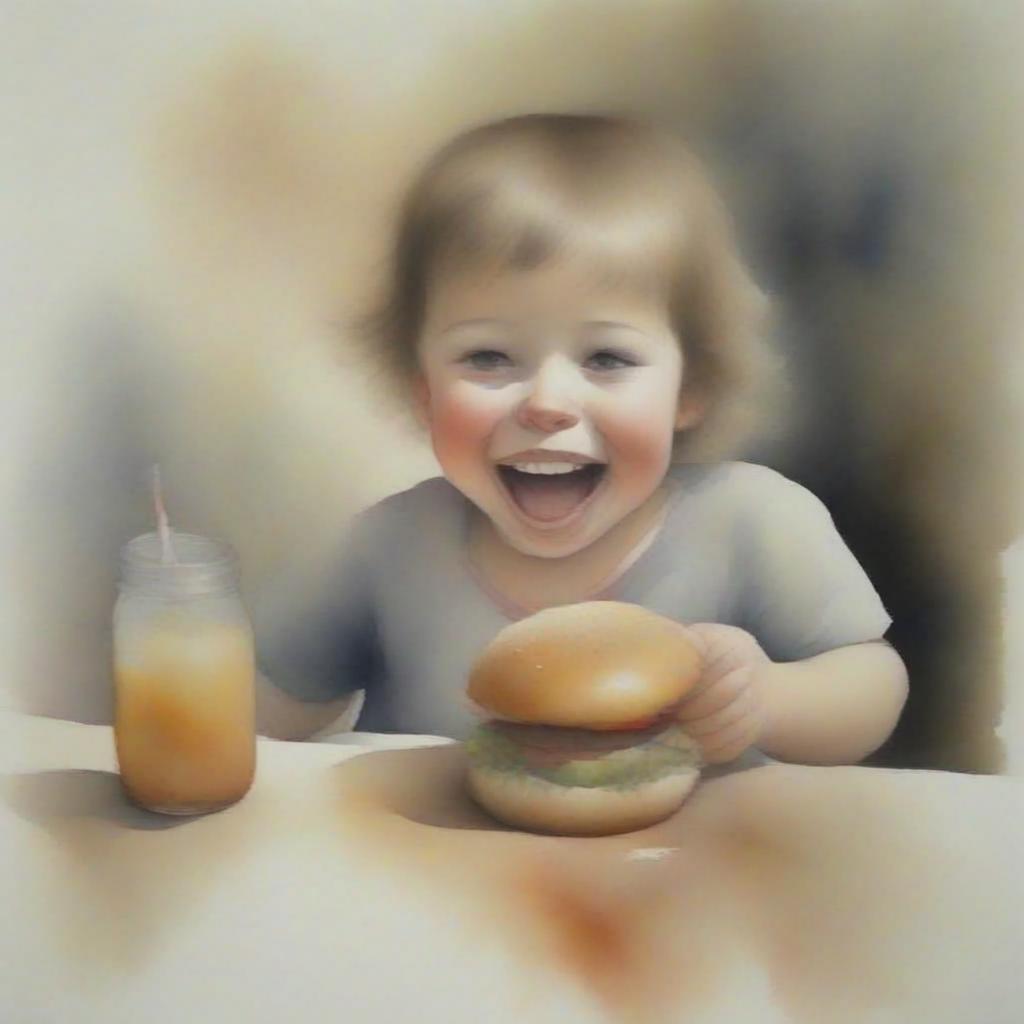}} &
        {\includegraphics[valign=c, width=\ww]{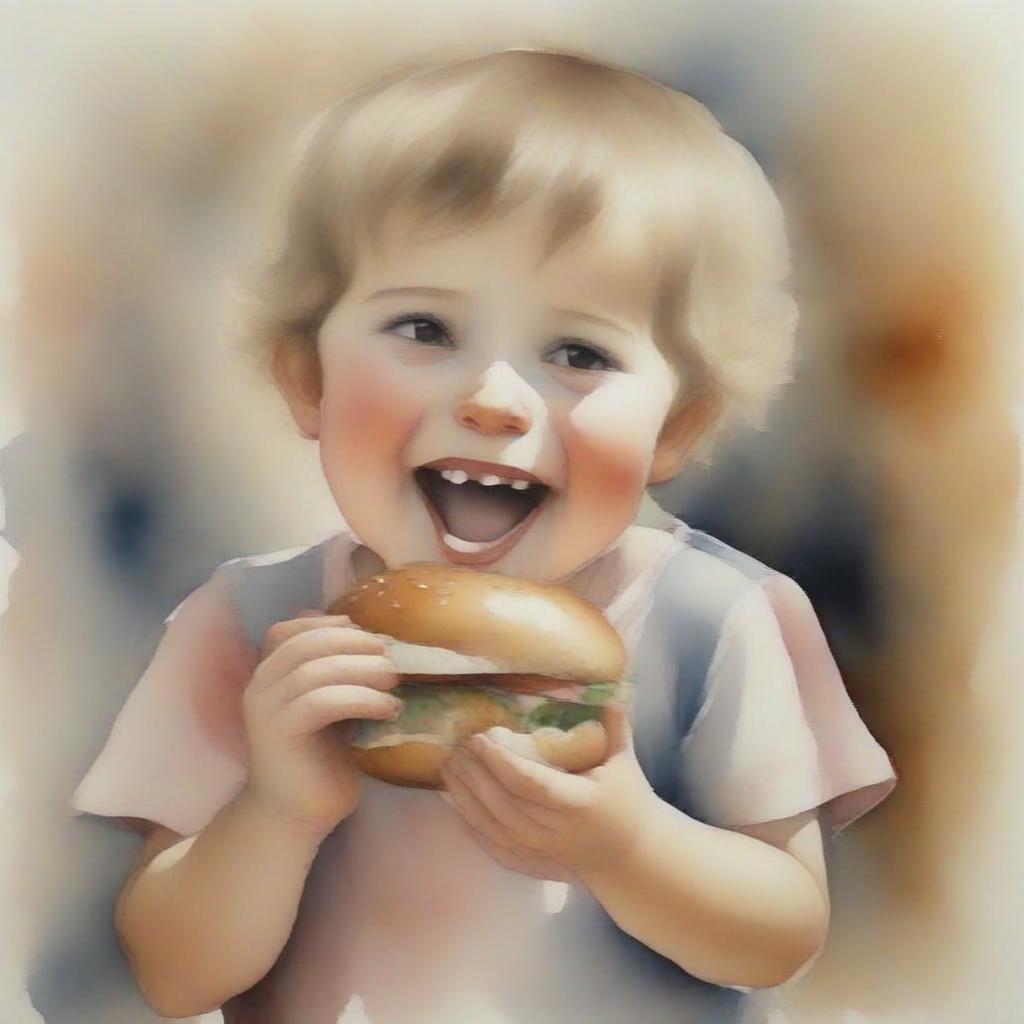}} &
        {\includegraphics[valign=c, width=\ww]{figures/automatic_qualitative_comparison/assets/child/ours/burger.jpg}}
        \\
        \\

        \rotatebox[origin=c]{90}{\textit{``wearing a}}
        \rotatebox[origin=c]{90}{\textit{blue hat''}} &
        {\includegraphics[valign=c, width=\ww]{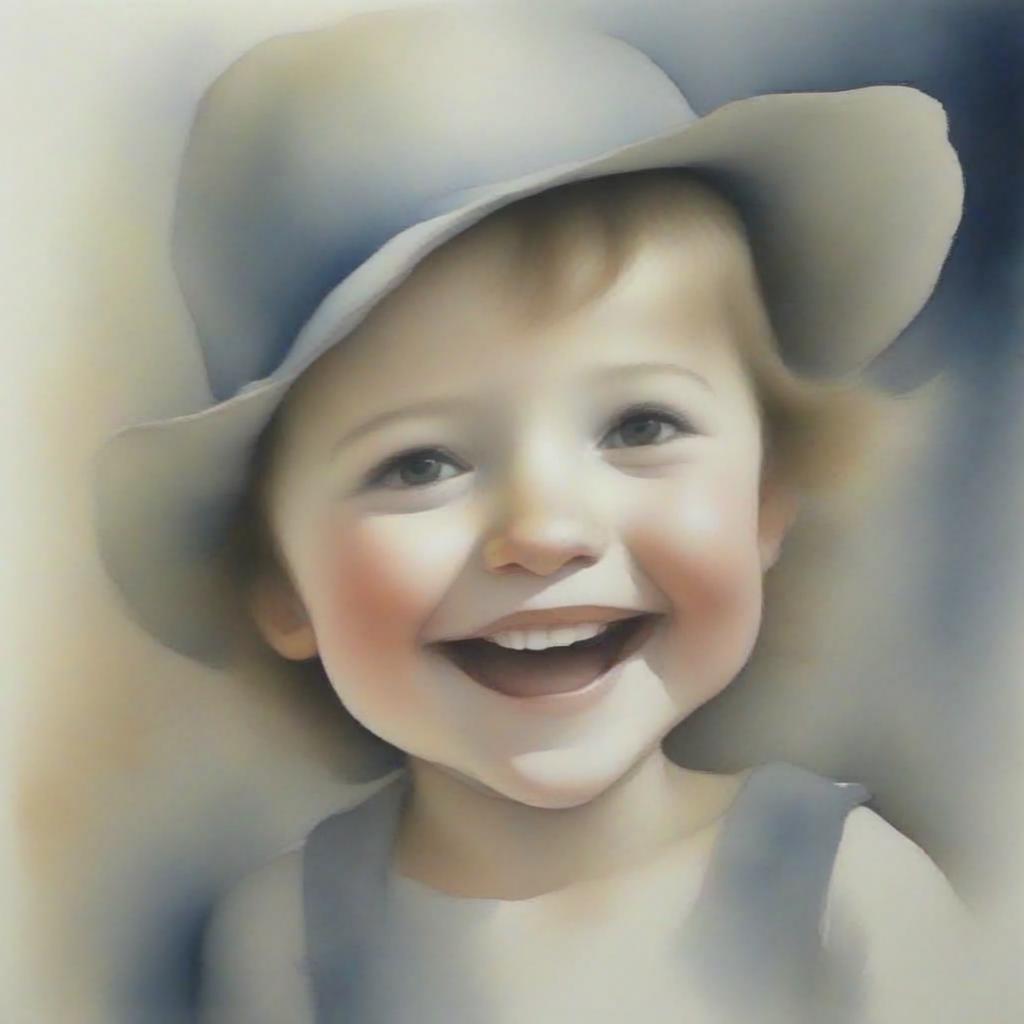}} &
        {\includegraphics[valign=c, width=\ww]{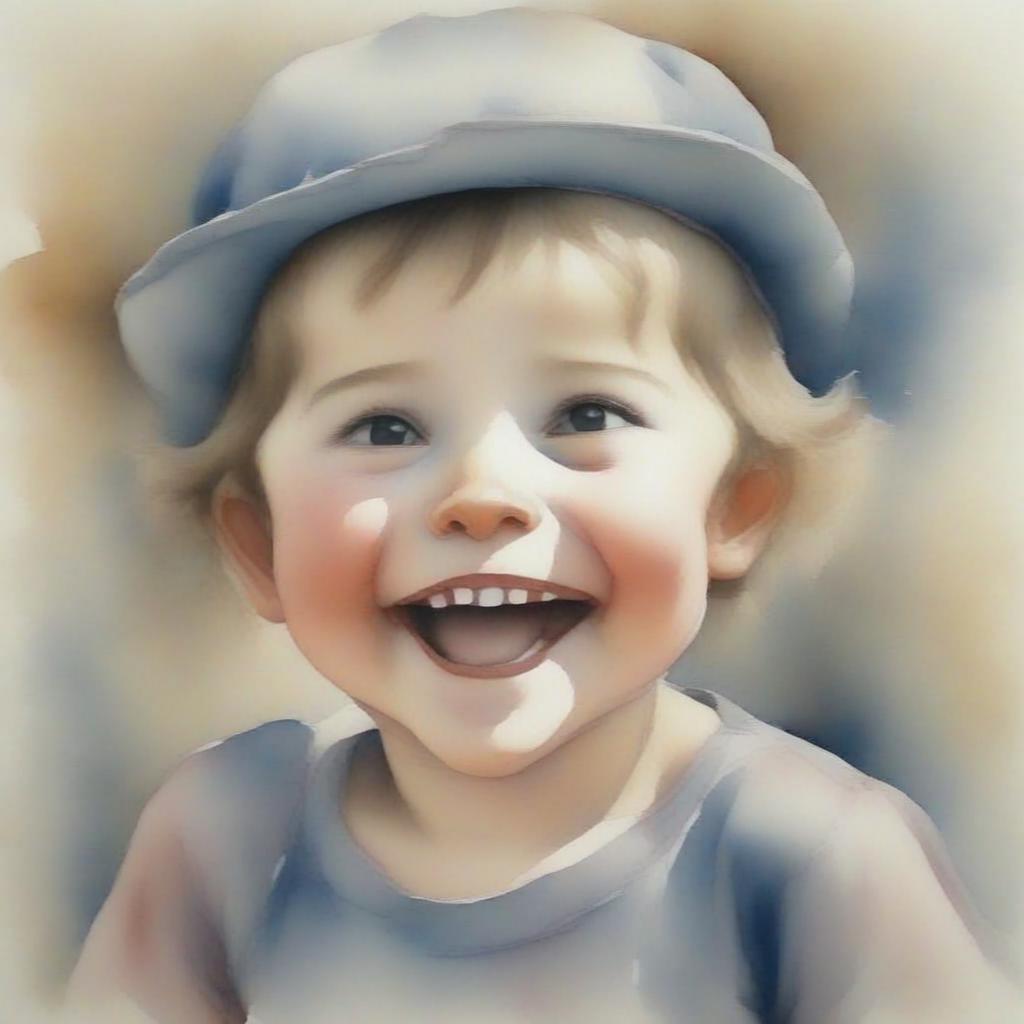}} &
        {\includegraphics[valign=c, width=\ww]{figures/automatic_qualitative_comparison/assets/child/ours/hat.jpg}}
        \\
        \\

        \multicolumn{4}{c}{\textit{``a watercolor portrayal of a joyful child, radiating innocence and}}
        \\
        \multicolumn{4}{c}{\textit{wonder with rosy cheeks and a genuine, wide-eyed smile''}}
        \\
        \\
        \midrule

        \\
        \rotatebox[origin=c]{90}{\textit{``near the Statue}}
        \rotatebox[origin=c]{90}{\textit{of Liberty''}} &
        {\includegraphics[valign=c, width=\ww]{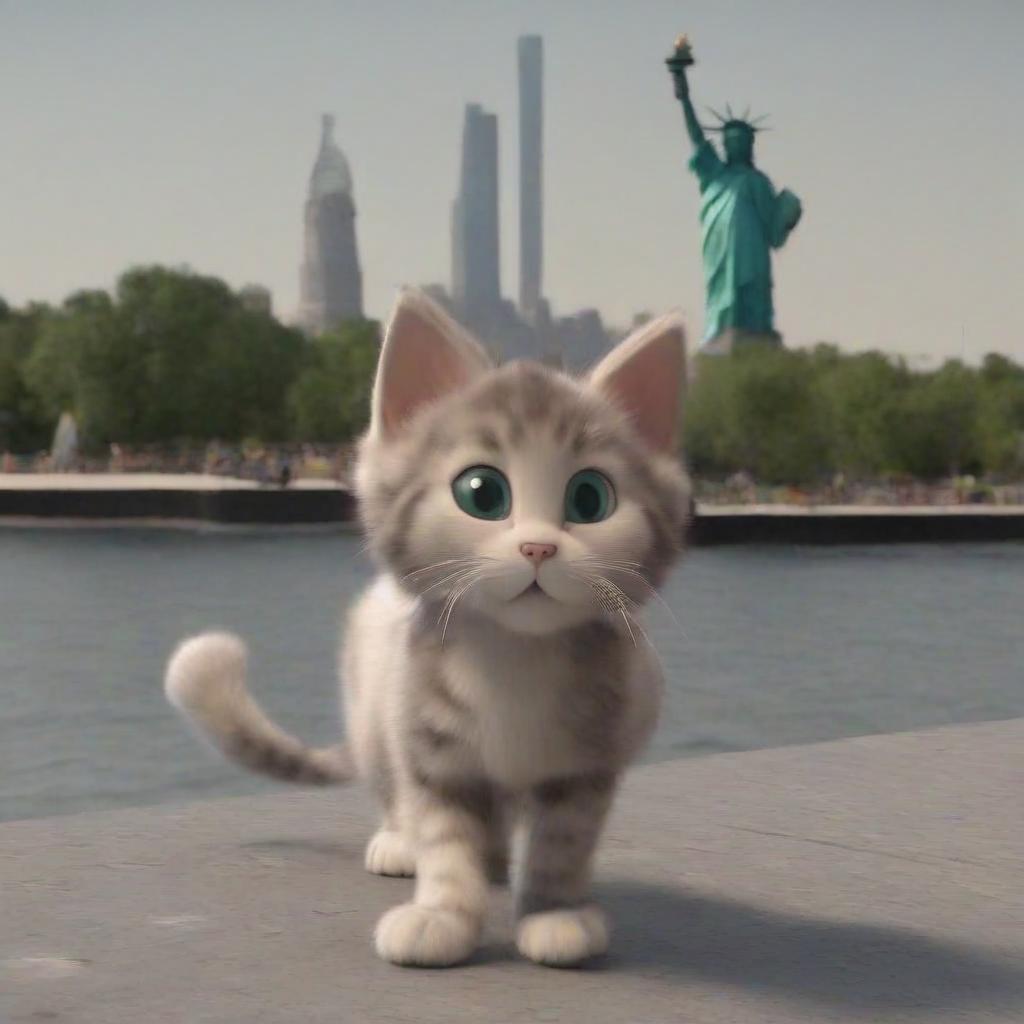}} &
        {\includegraphics[valign=c, width=\ww]{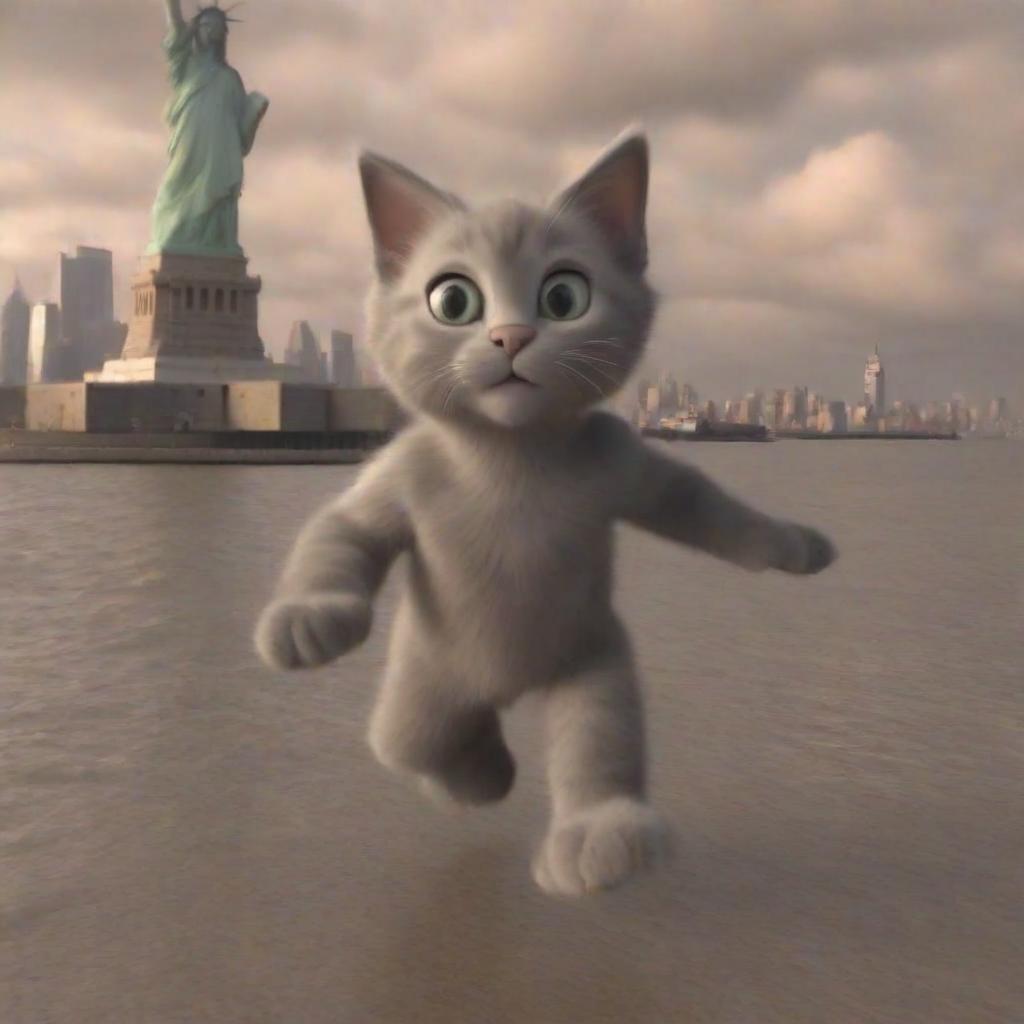}} &
        {\includegraphics[valign=c, width=\ww]{figures/automatic_qualitative_comparison/assets/cat/ours/liberty.jpg}}
        \\
        \\

        \rotatebox[origin=c]{90}{\textit{``as a police}}
        \rotatebox[origin=c]{90}{\textit{officer''}} &
        {\includegraphics[valign=c, width=\ww]{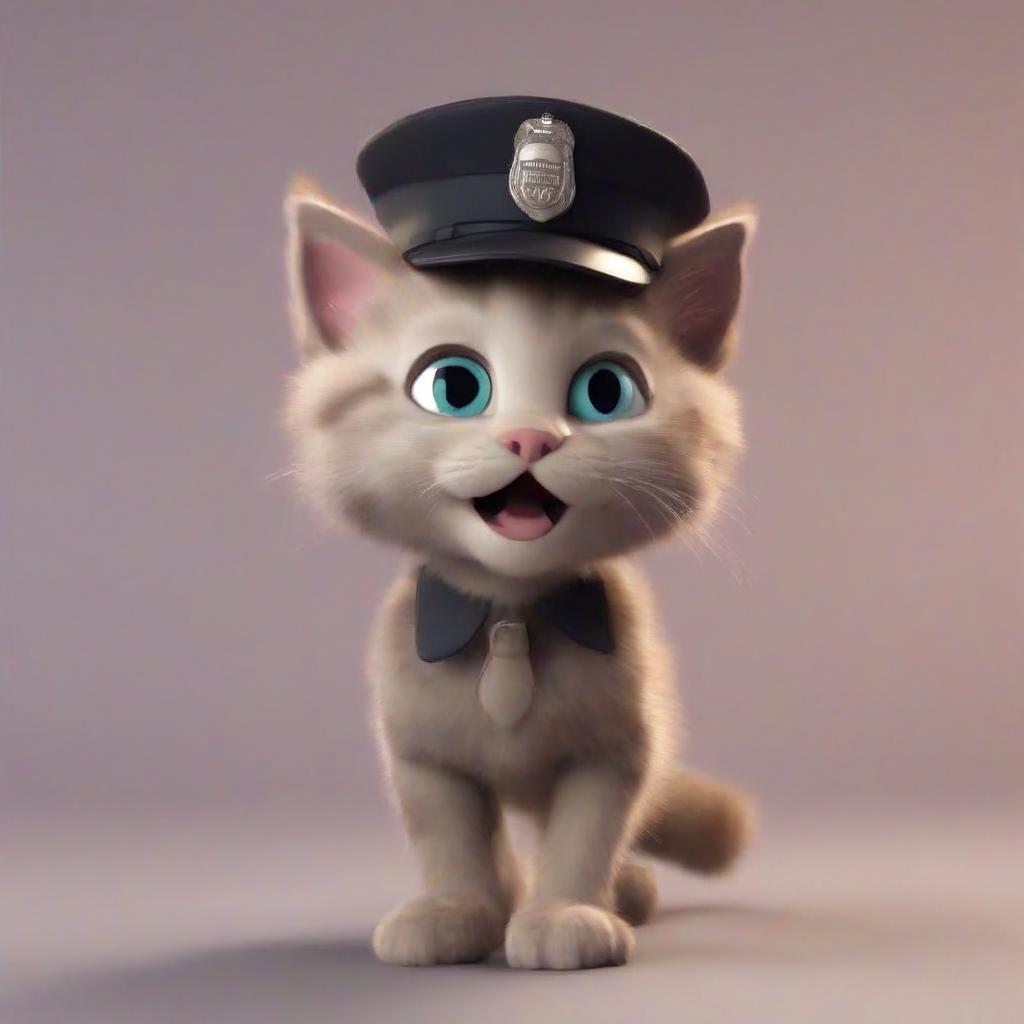}} &
        {\includegraphics[valign=c, width=\ww]{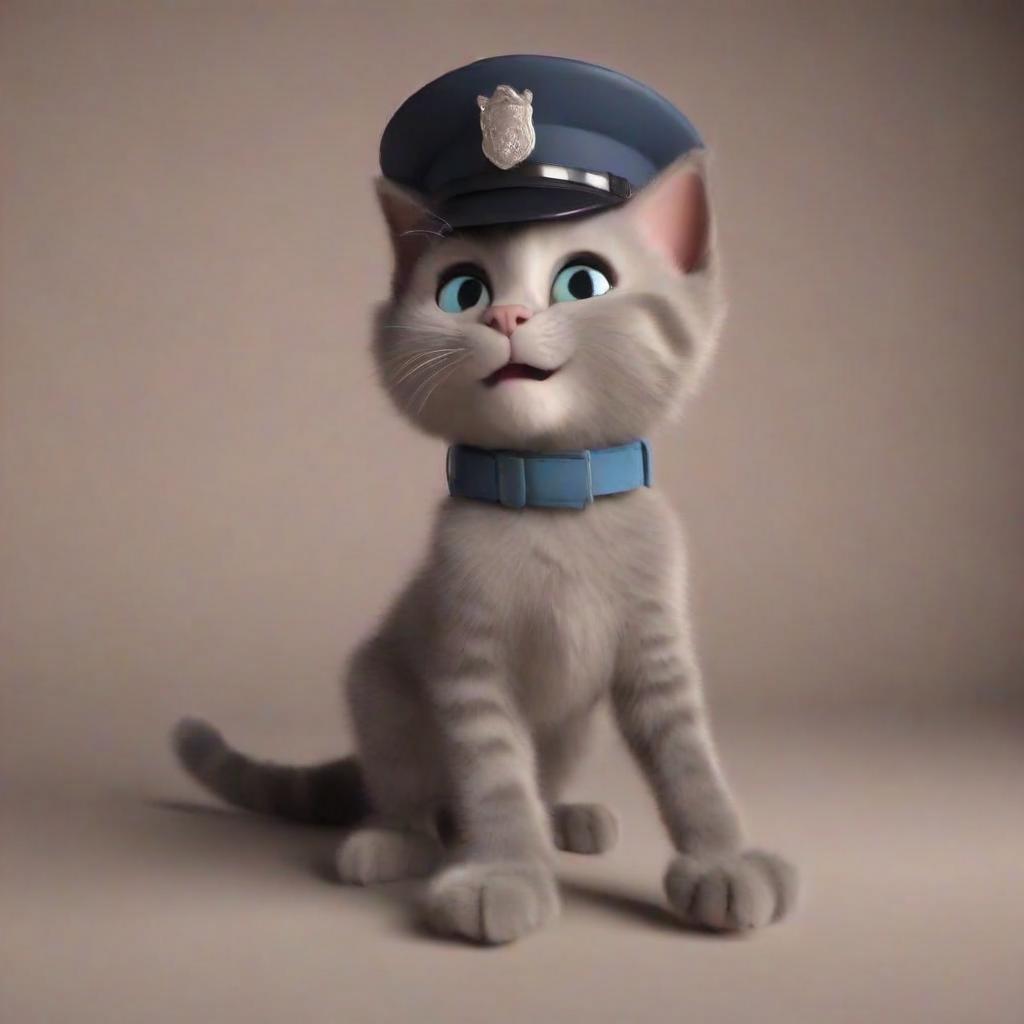}} &
        {\includegraphics[valign=c, width=\ww]{figures/automatic_qualitative_comparison/assets/cat/ours/police.jpg}}
        \\
        \\

        \multicolumn{4}{c}{\textit{``a 3D animation of a playful kitten, with bright eyes and a}}
        \\
        \multicolumn{4}{c}{\textit{mischievous expression, embodying youthful curiosity and joy''}}

    \end{tabular}
    
    \caption{\textbf{Comparison of feature extractors.} We experimented with two additional feature extractors in our method: DINOv1 \cite{Caron2021EmergingPI} and CLIP \cite{Radford2021LearningTV}. As can be seen, DINOv1 results are qualitatively similar to DINOv2, whereas CLIP produces results with a slightly lower identity consistency.}
    \label{fig:automatic_qualitative_feature_extractor_comparison}
\end{figure*}

%% file: figures/clustering_visualization/fig.tex
\begin{figure*}[t]
    \centering
    \setlength{\tabcolsep}{0.8pt}
    \renewcommand{\arraystretch}{0.5}
    \setlength{\ww}{0.38\columnwidth}
    \begin{tabular}{cccccc}
        \rotatebox[origin=c]{90}{Cluster 1} &
        {\includegraphics[valign=c, width=\ww]{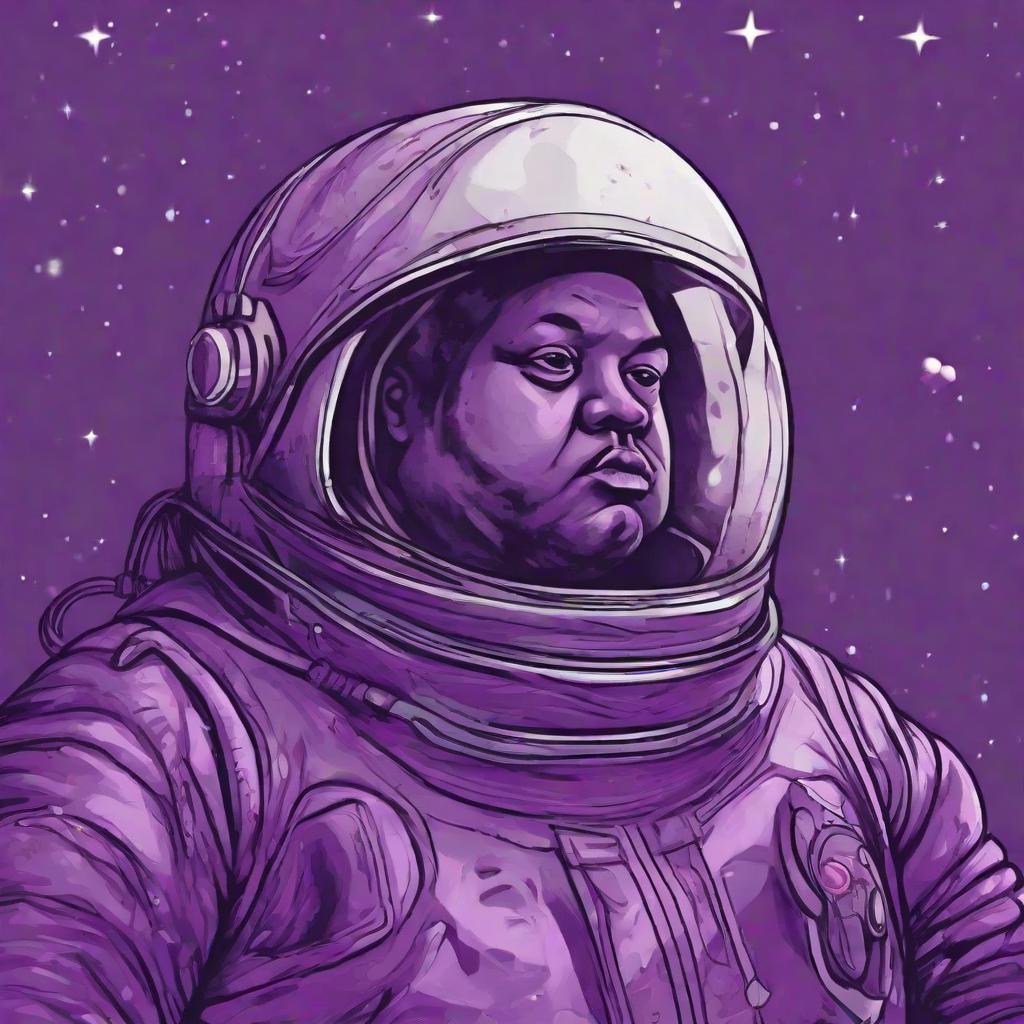}} &
        {\includegraphics[valign=c, width=\ww]{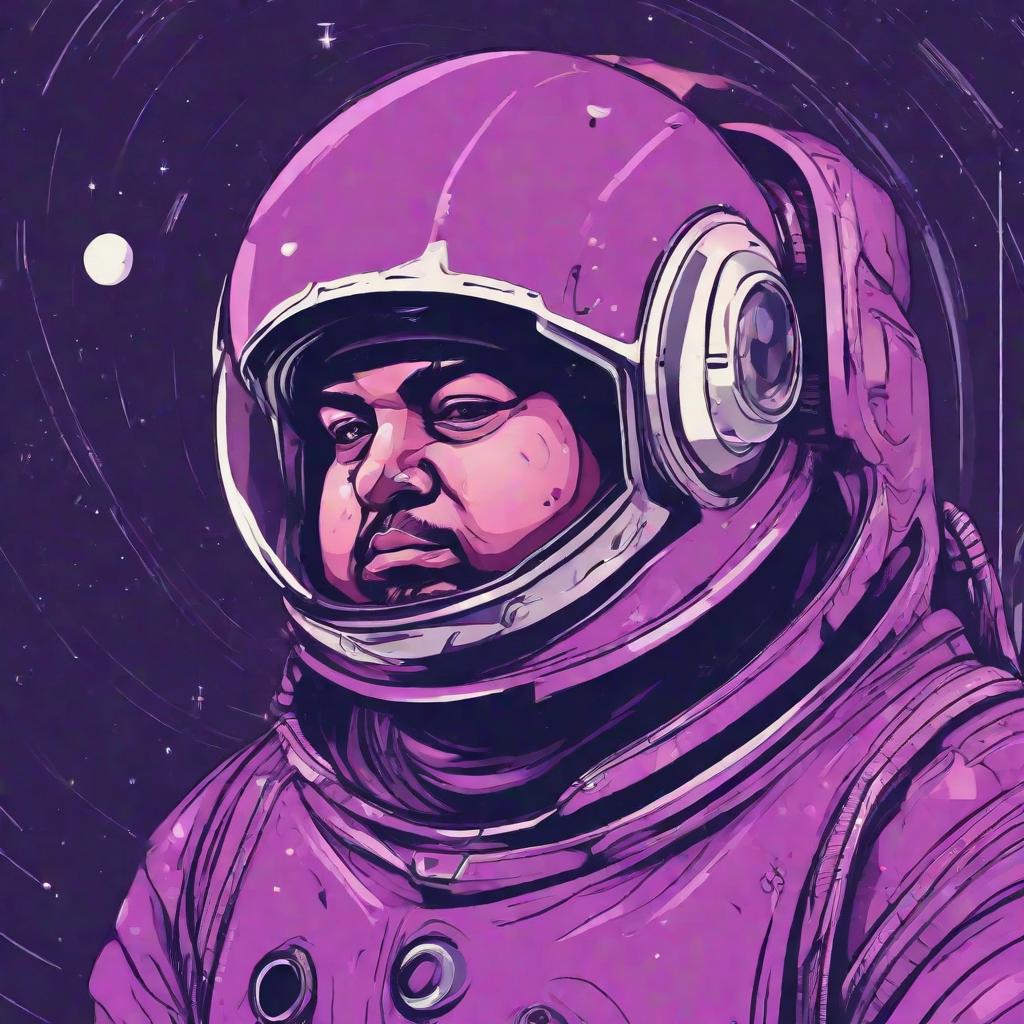}} &
        {\includegraphics[valign=c, width=\ww]{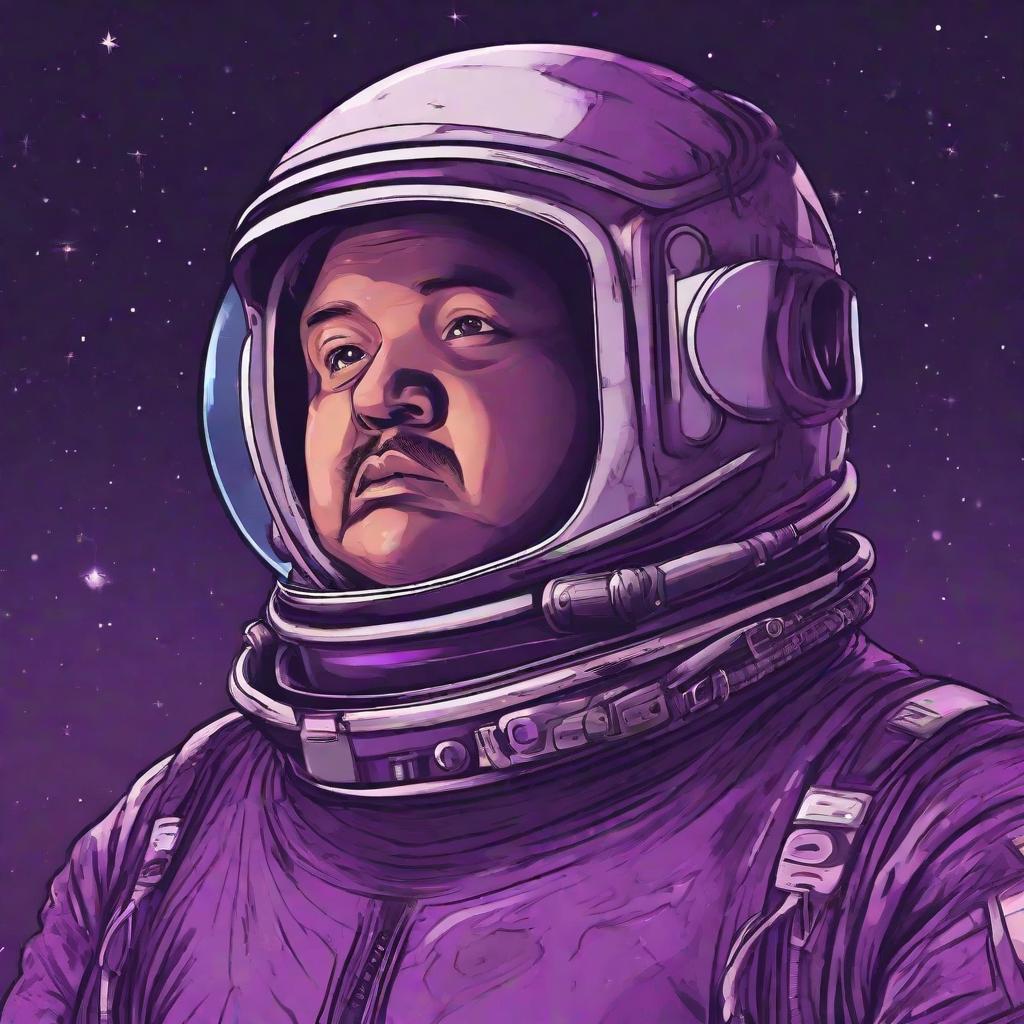}} &
        {\includegraphics[valign=c, width=\ww]{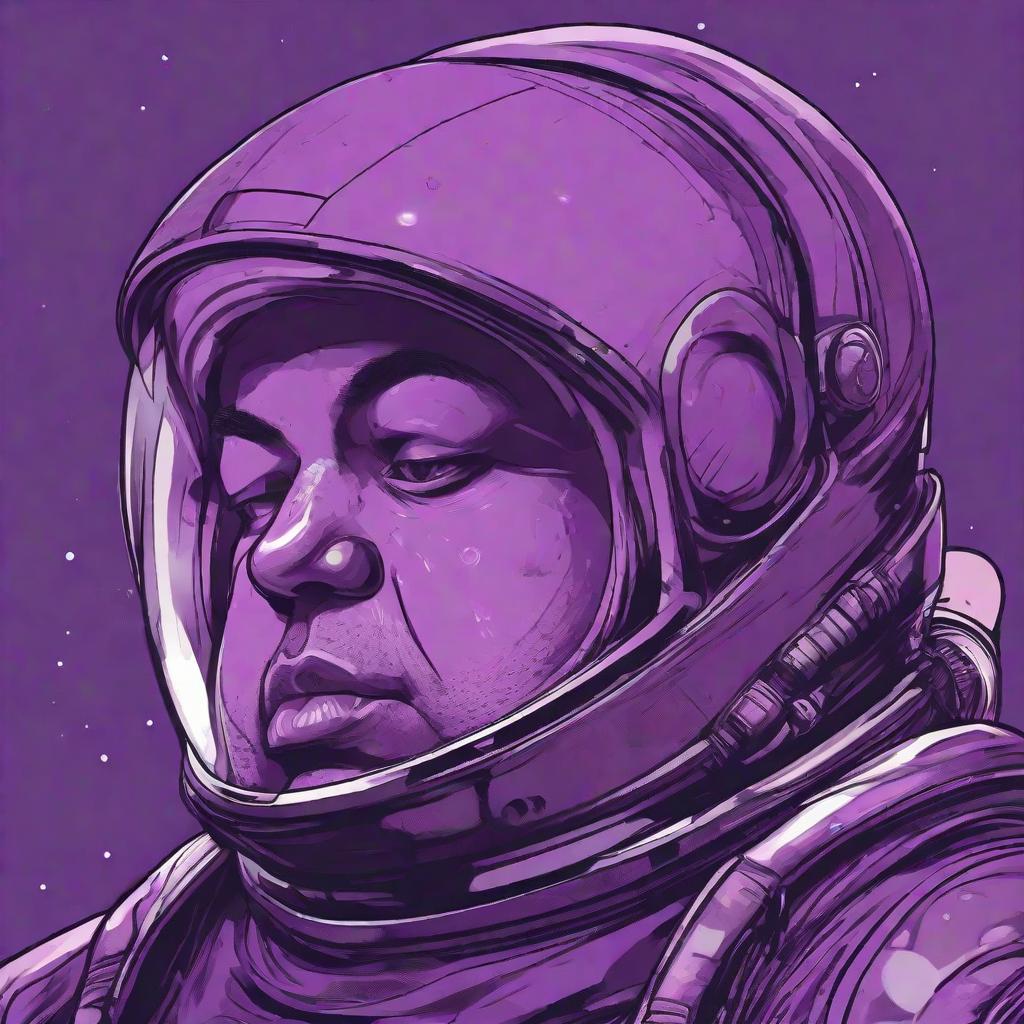}} &
        {\includegraphics[valign=c, width=\ww]{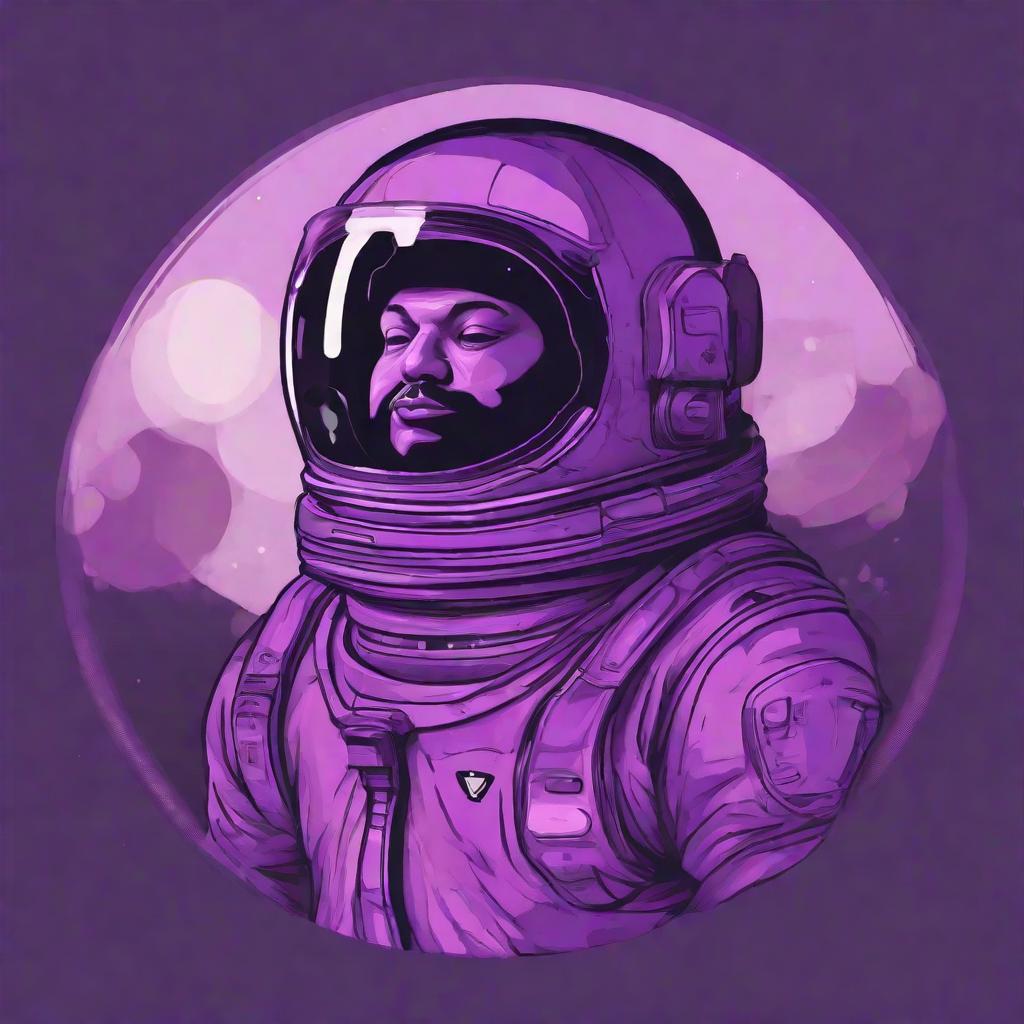}}
        \\
        \\

        \rotatebox[origin=c]{90}{Cluster 2} &
        {\includegraphics[valign=c, width=\ww]{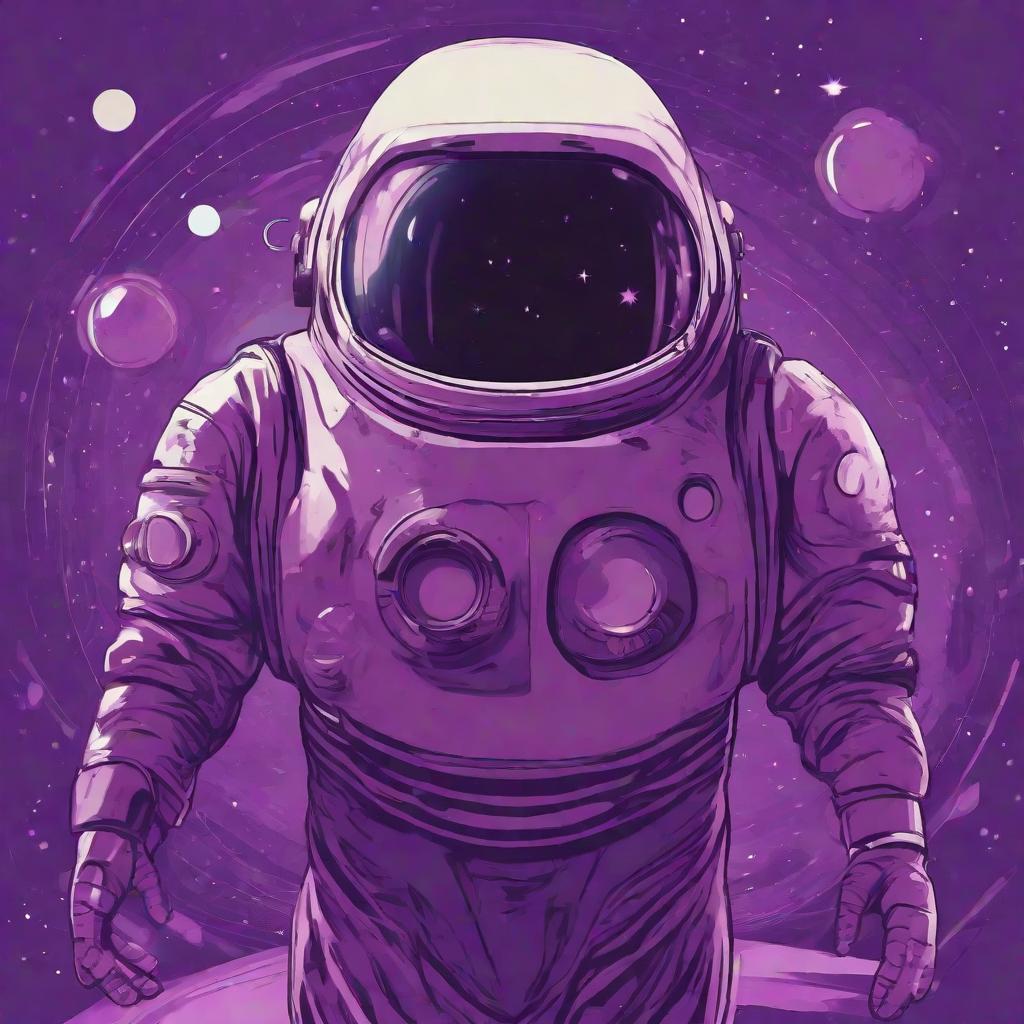}} &
        {\includegraphics[valign=c, width=\ww]{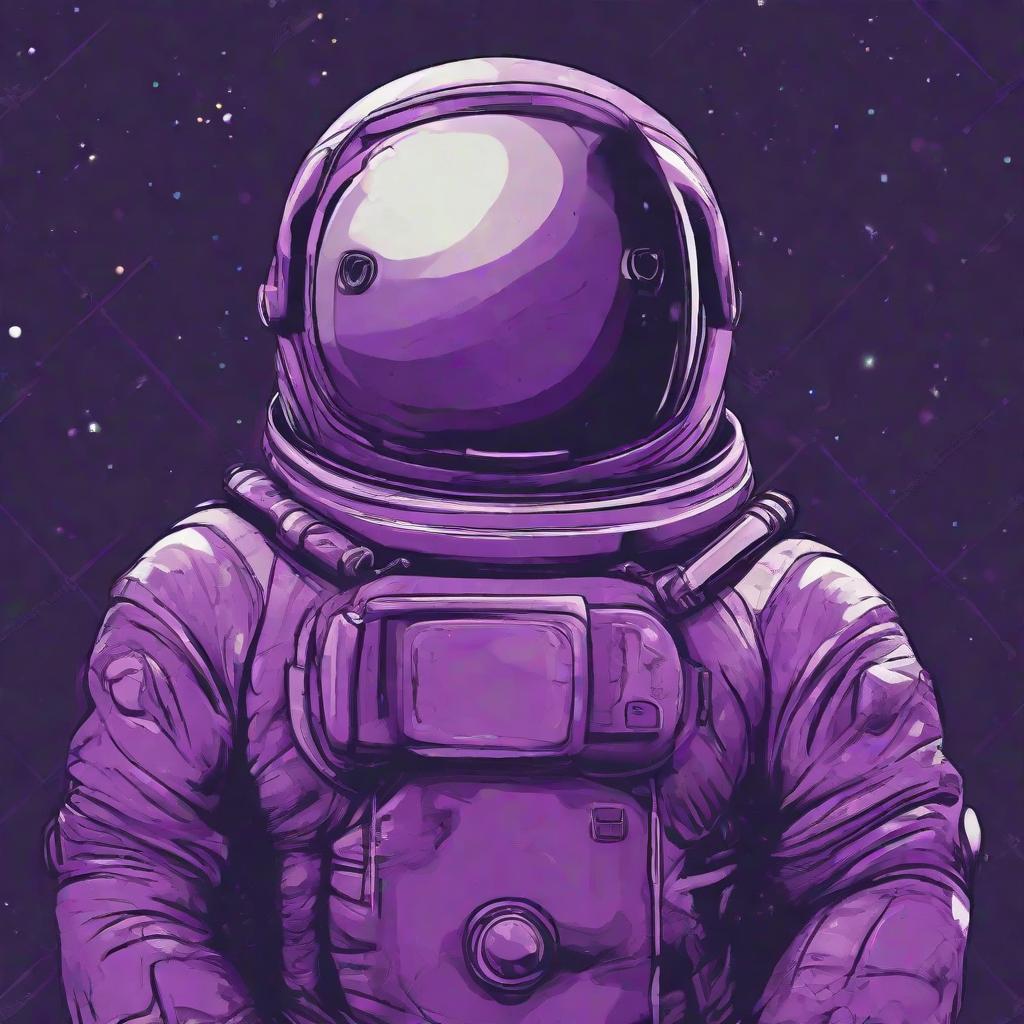}} &
        {\includegraphics[valign=c, width=\ww]{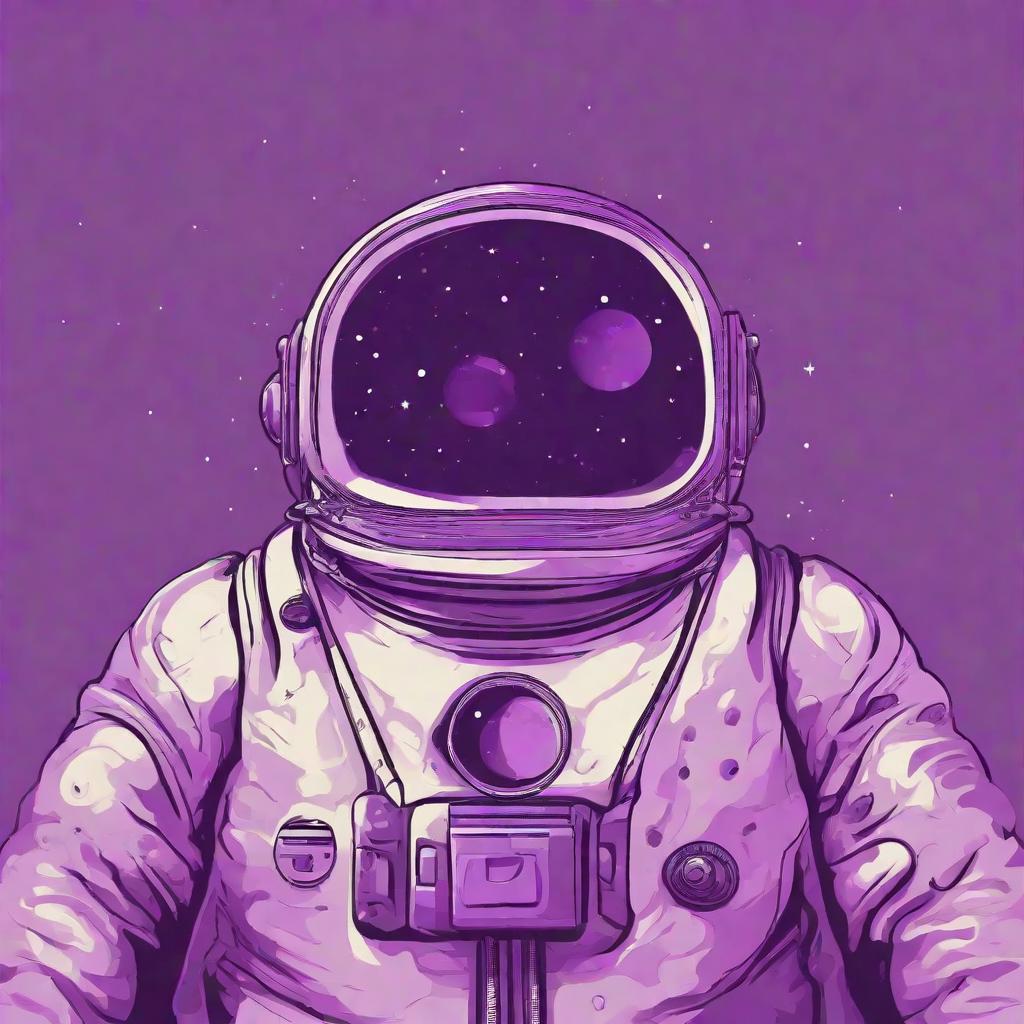}} &
        {\includegraphics[valign=c, width=\ww]{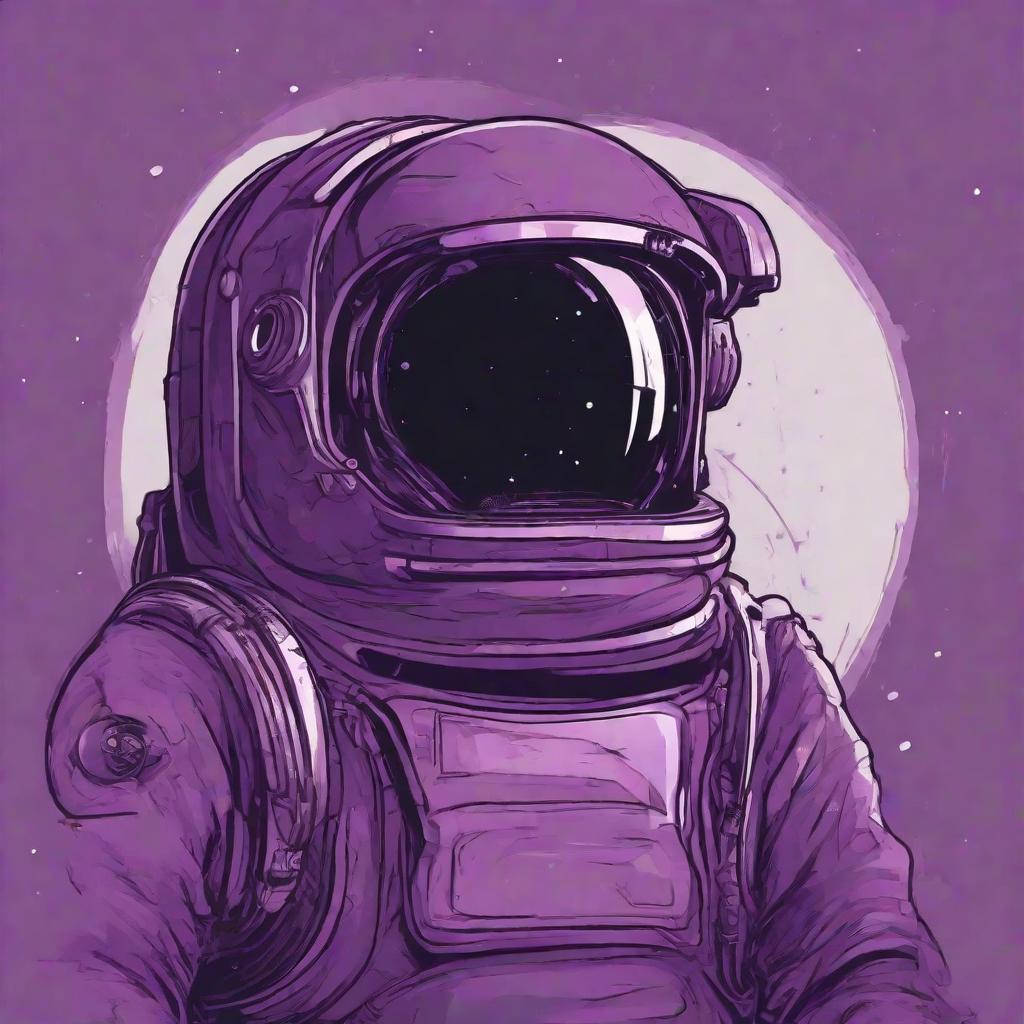}} &
        {\includegraphics[valign=c, width=\ww]{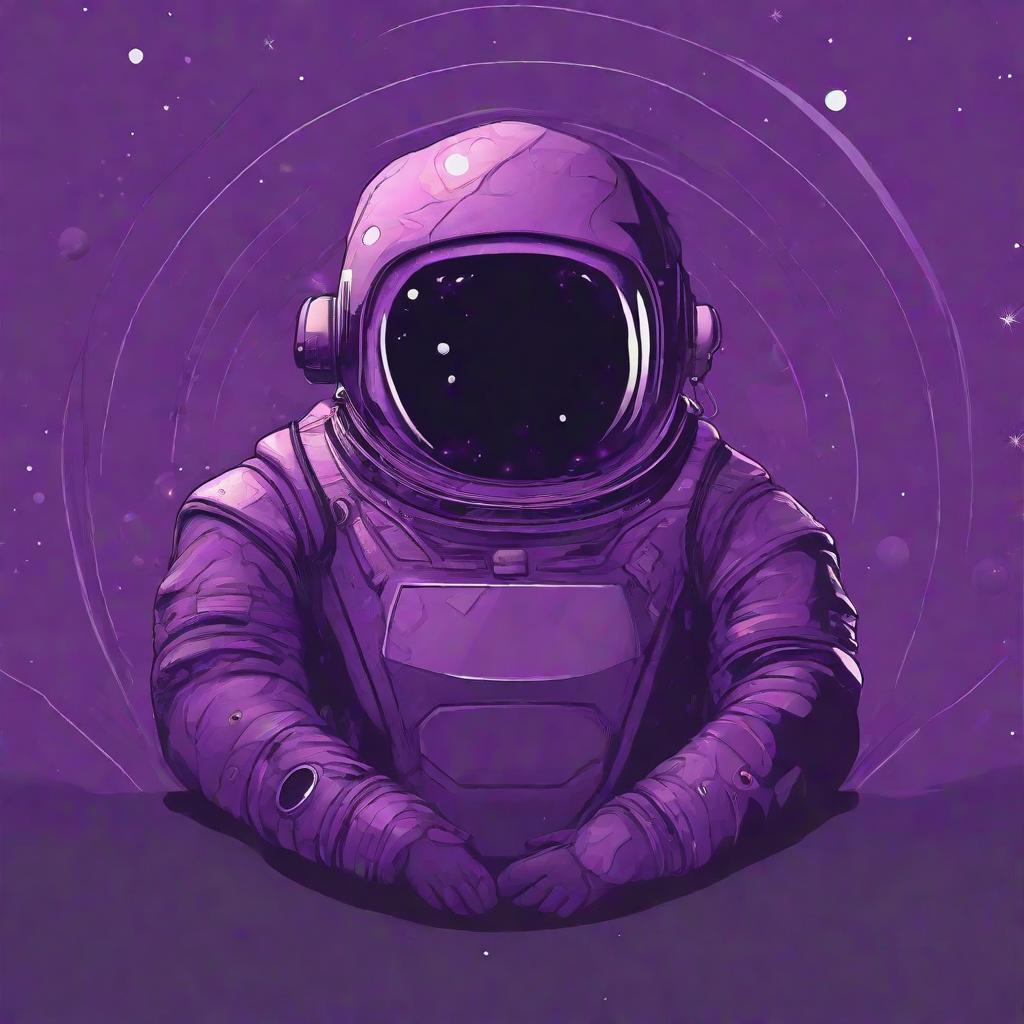}}
        \\
        \\

        \rotatebox[origin=c]{90}{Cluster 3} &
        {\includegraphics[valign=c, width=\ww]{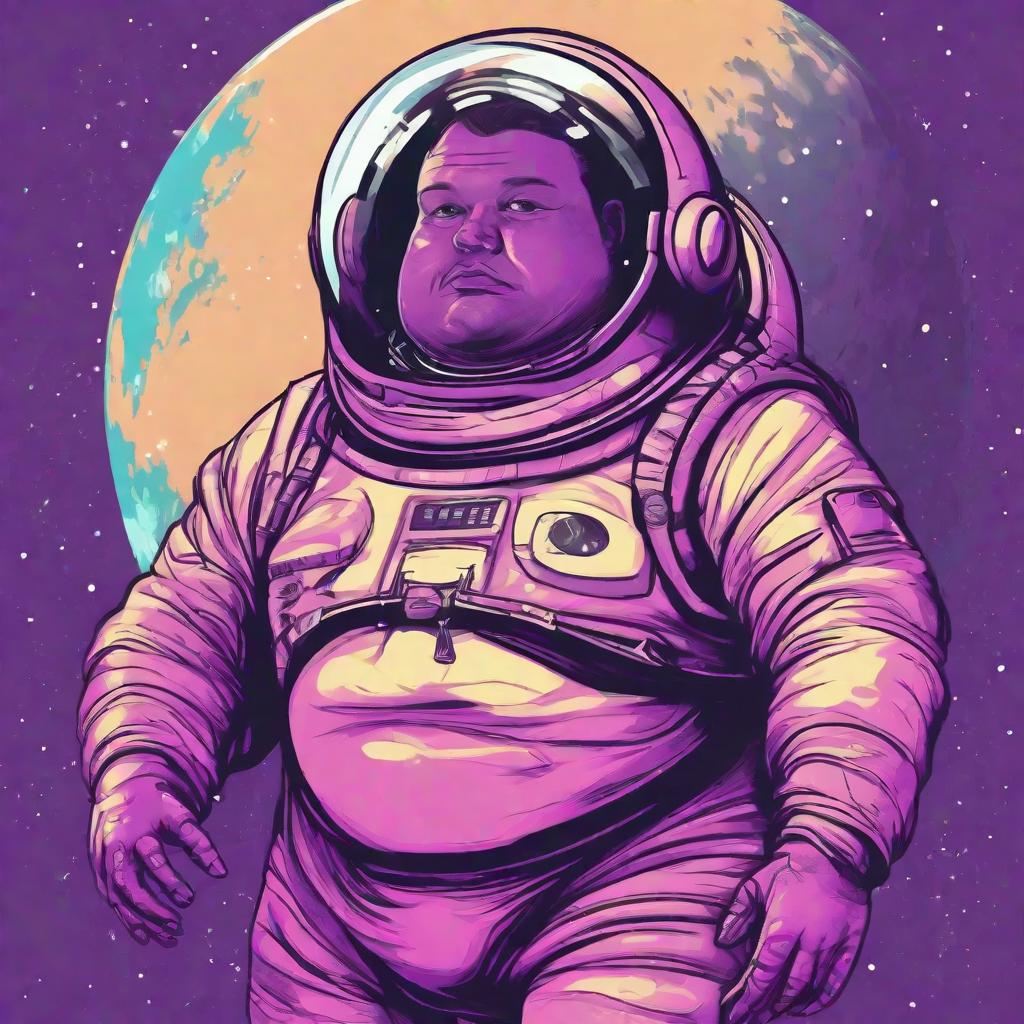}} &
        {\includegraphics[valign=c, width=\ww]{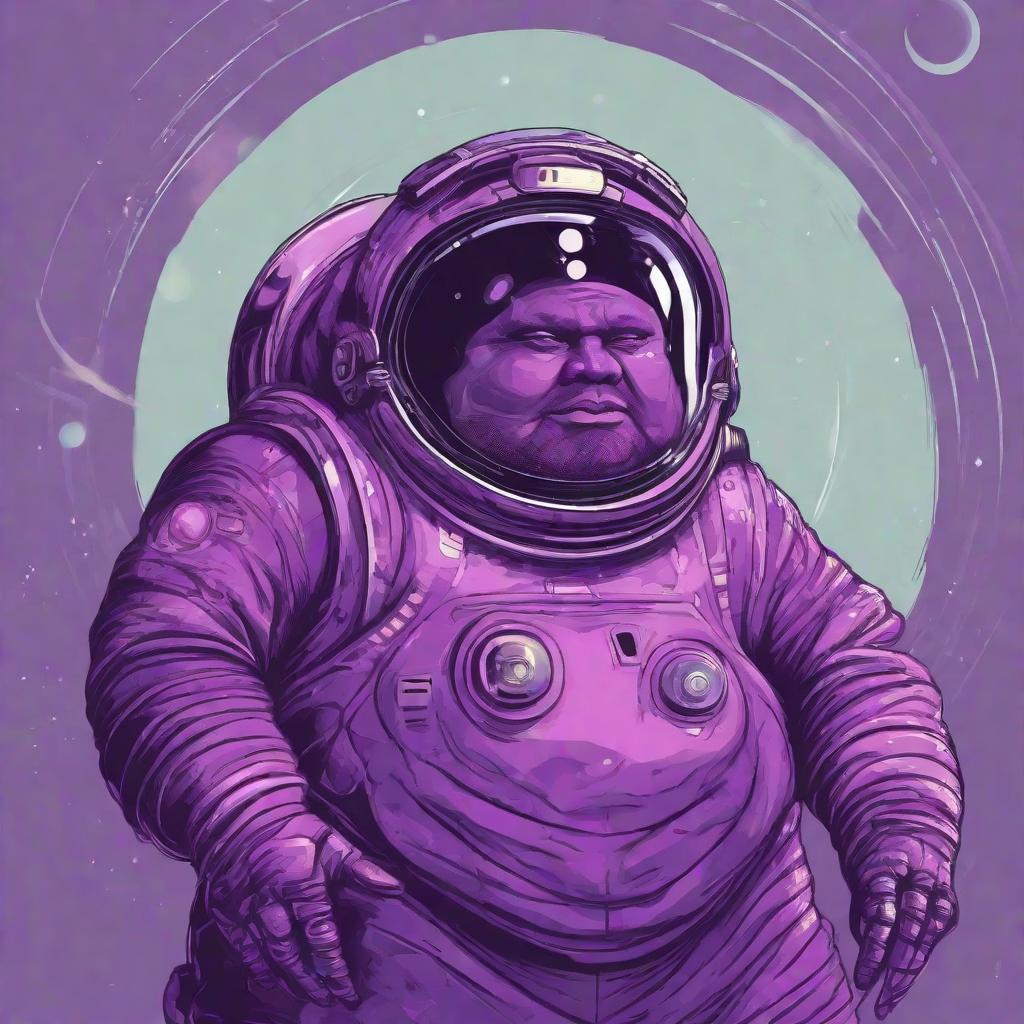}} &
        {\includegraphics[valign=c, width=\ww]{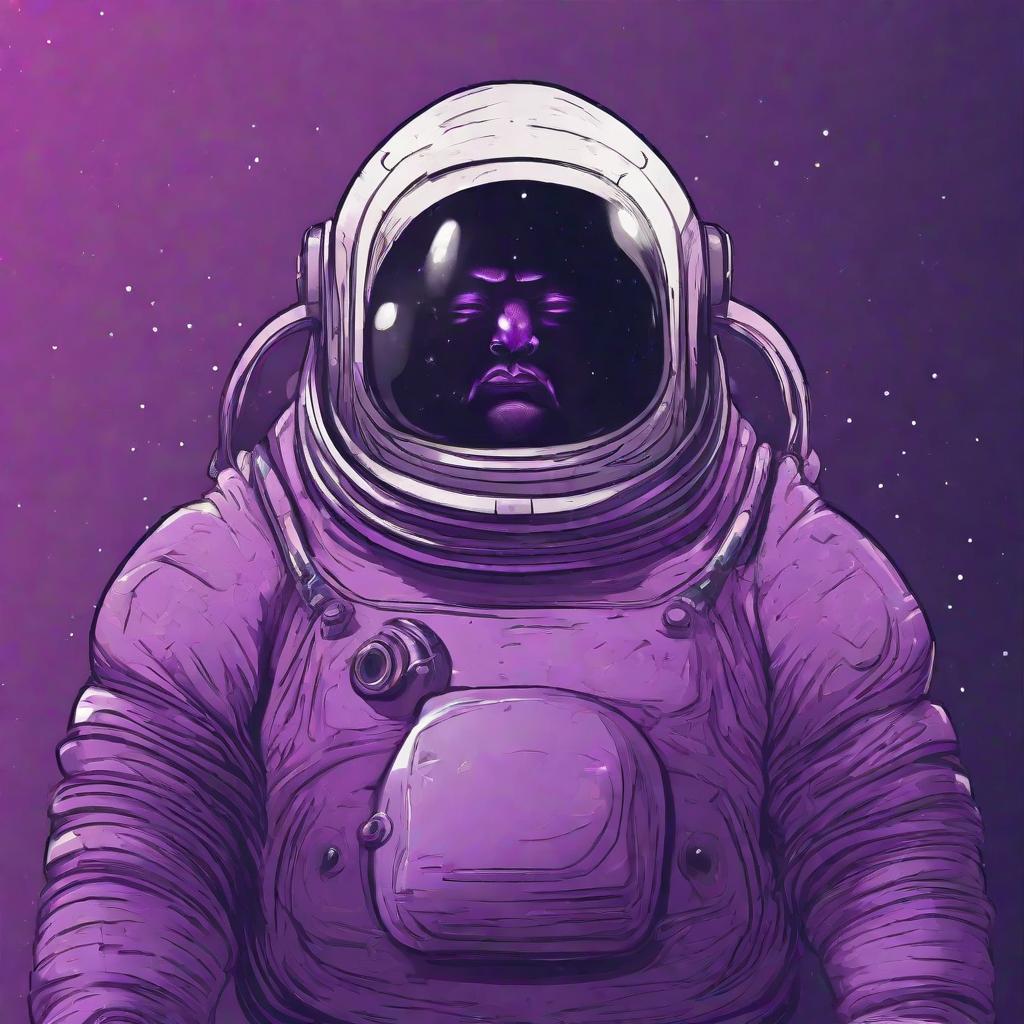}} &
        {\includegraphics[valign=c, width=\ww]{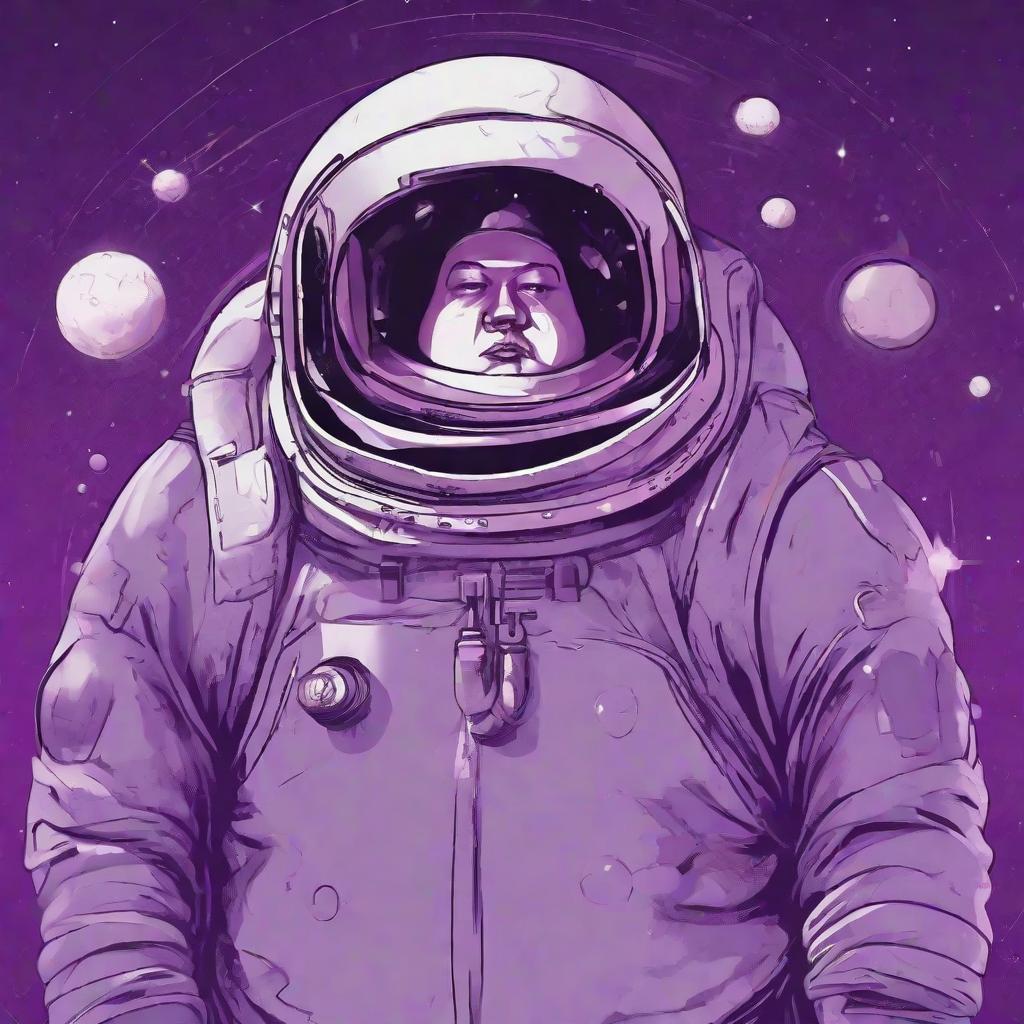}} &
        {\includegraphics[valign=c, width=\ww]{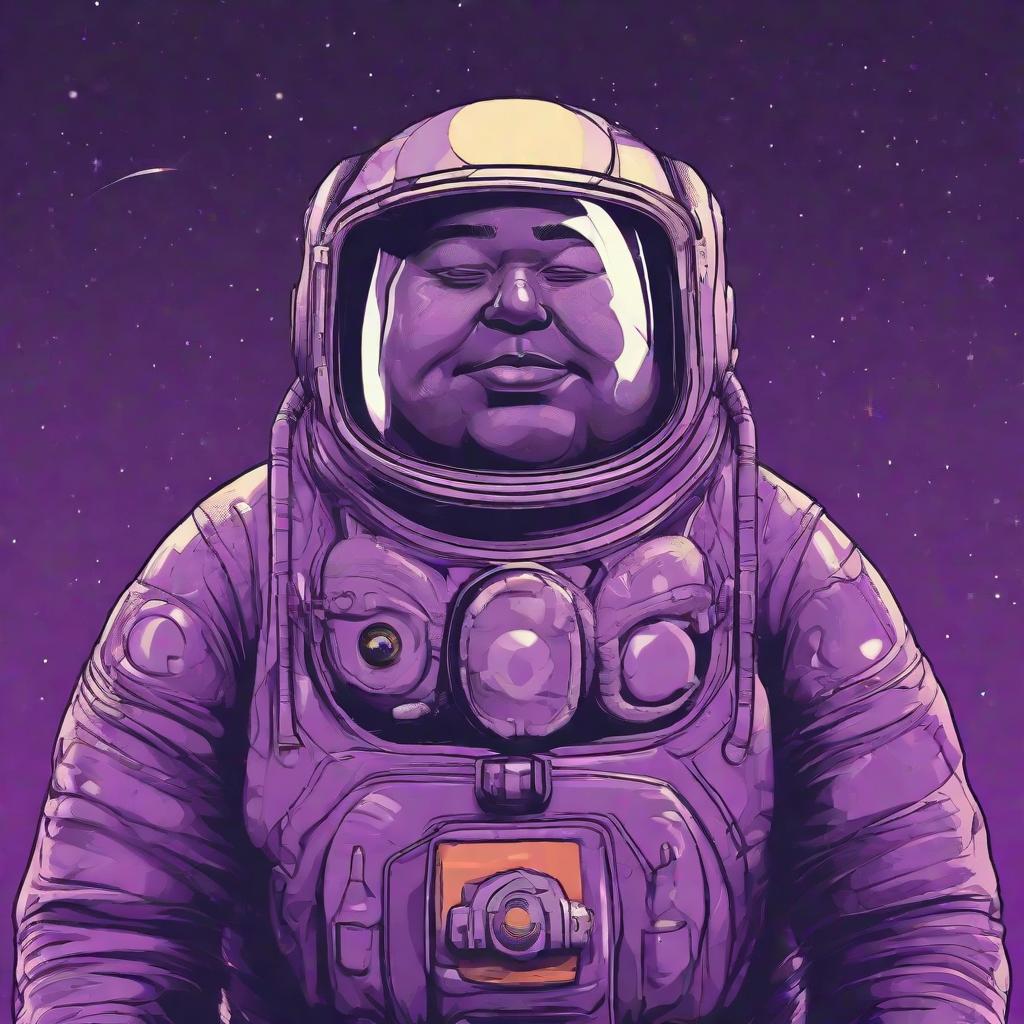}}
        \\
        \\

        \midrule
        \\

        \rotatebox[origin=c]{90}{Cluster 1} &
        {\includegraphics[valign=c, width=\ww]{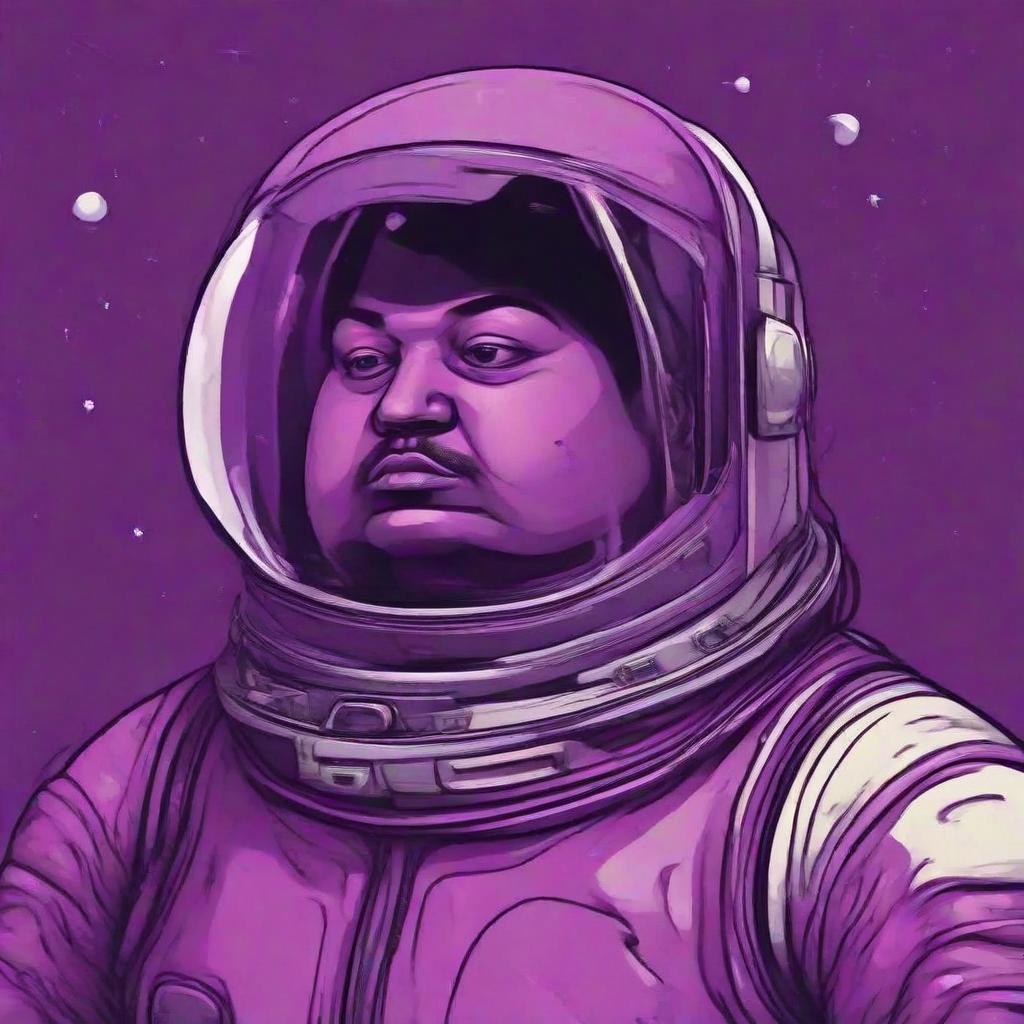}} &
        {\includegraphics[valign=c, width=\ww]{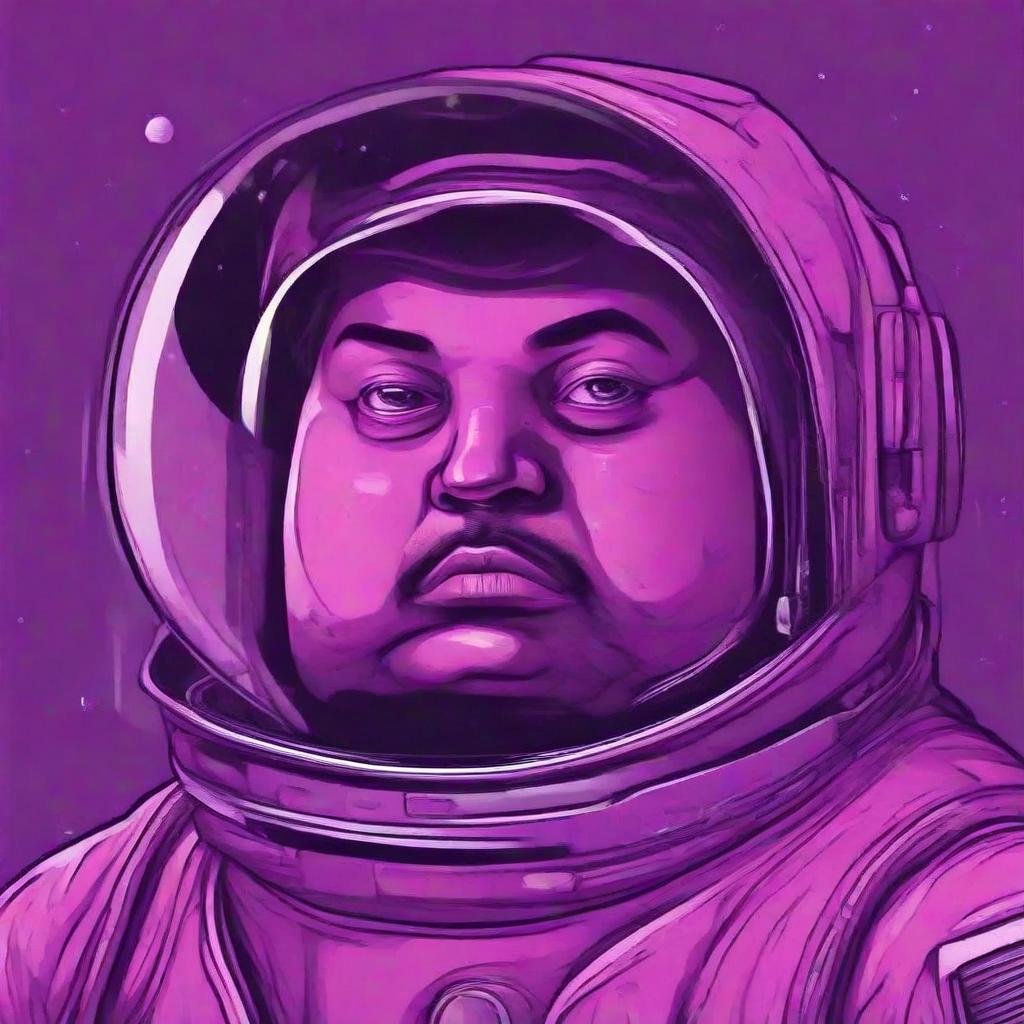}} &
        {\includegraphics[valign=c, width=\ww]{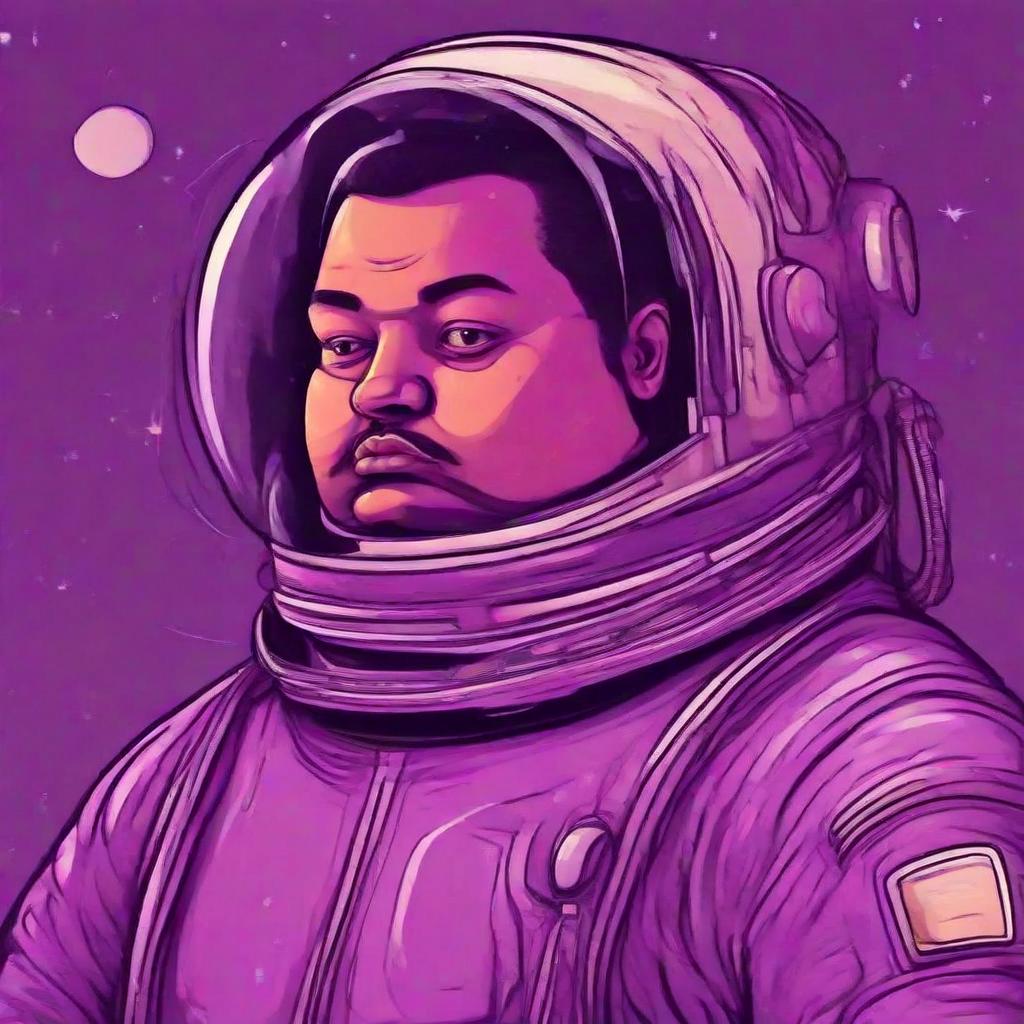}} &
        {\includegraphics[valign=c, width=\ww]{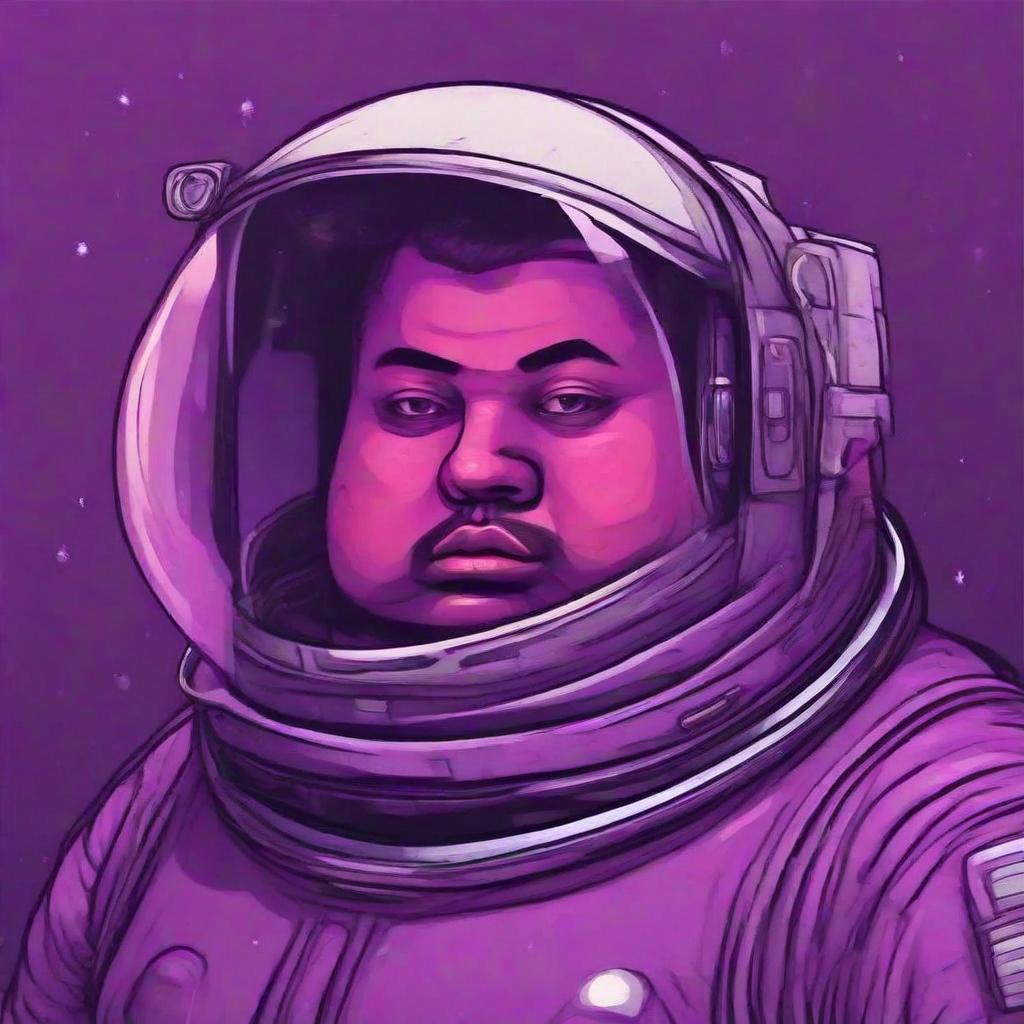}} &
        {\includegraphics[valign=c, width=\ww]{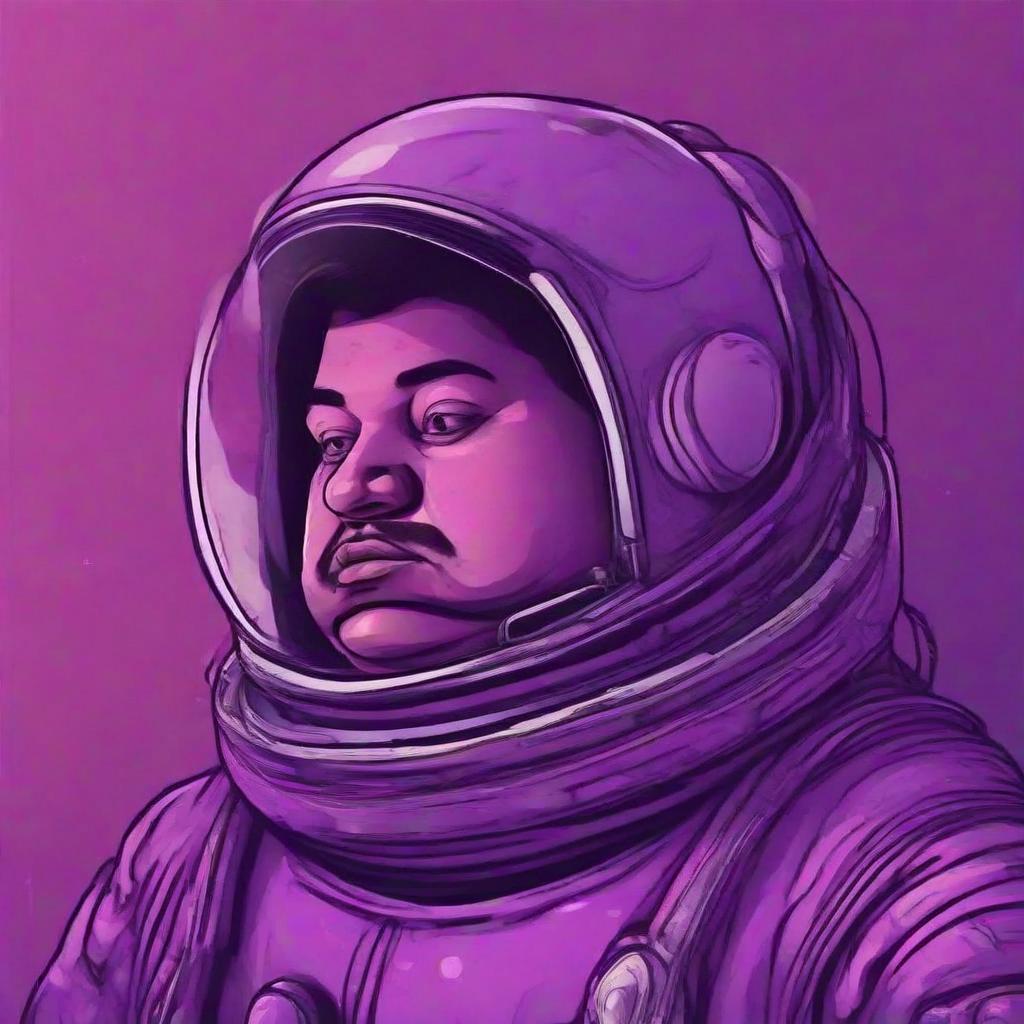}}
        \\
        \\

        \rotatebox[origin=c]{90}{Cluster 2} &
        {\includegraphics[valign=c, width=\ww]{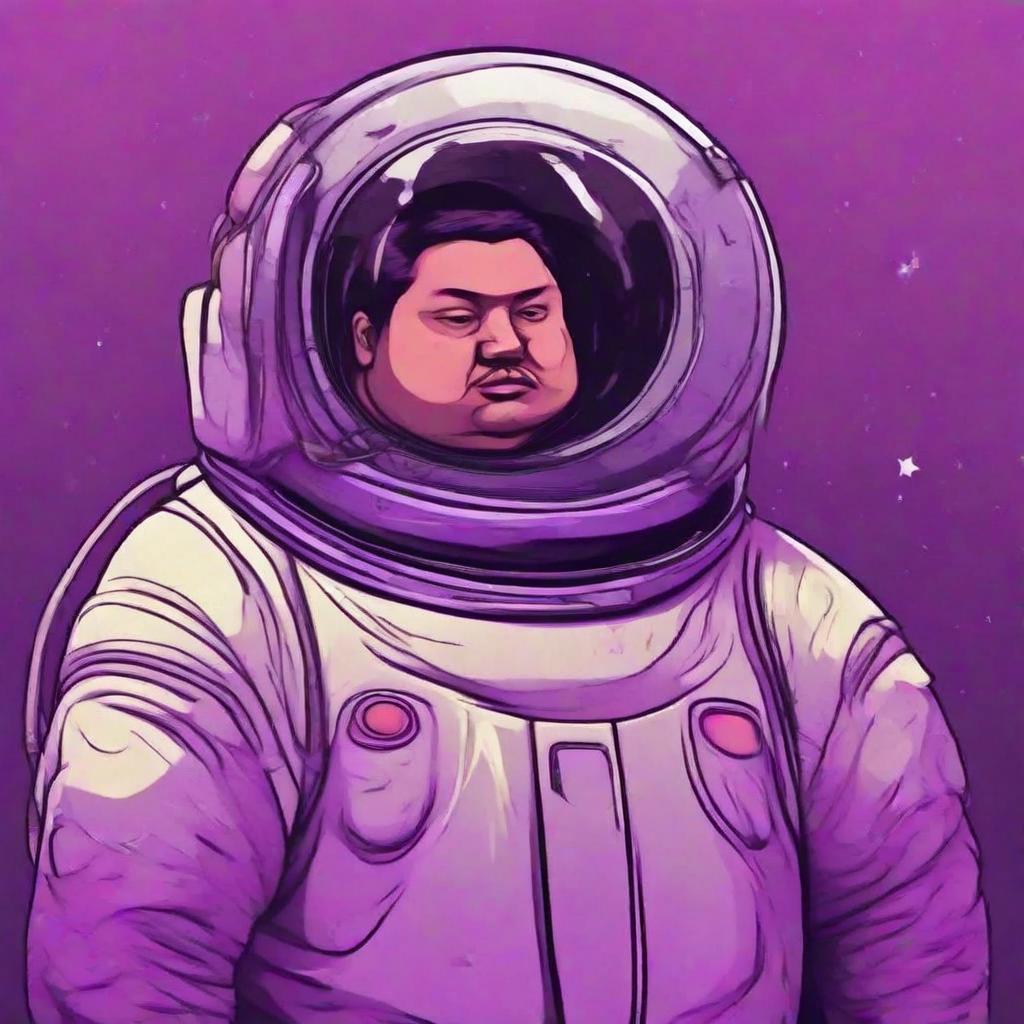}} &
        {\includegraphics[valign=c, width=\ww]{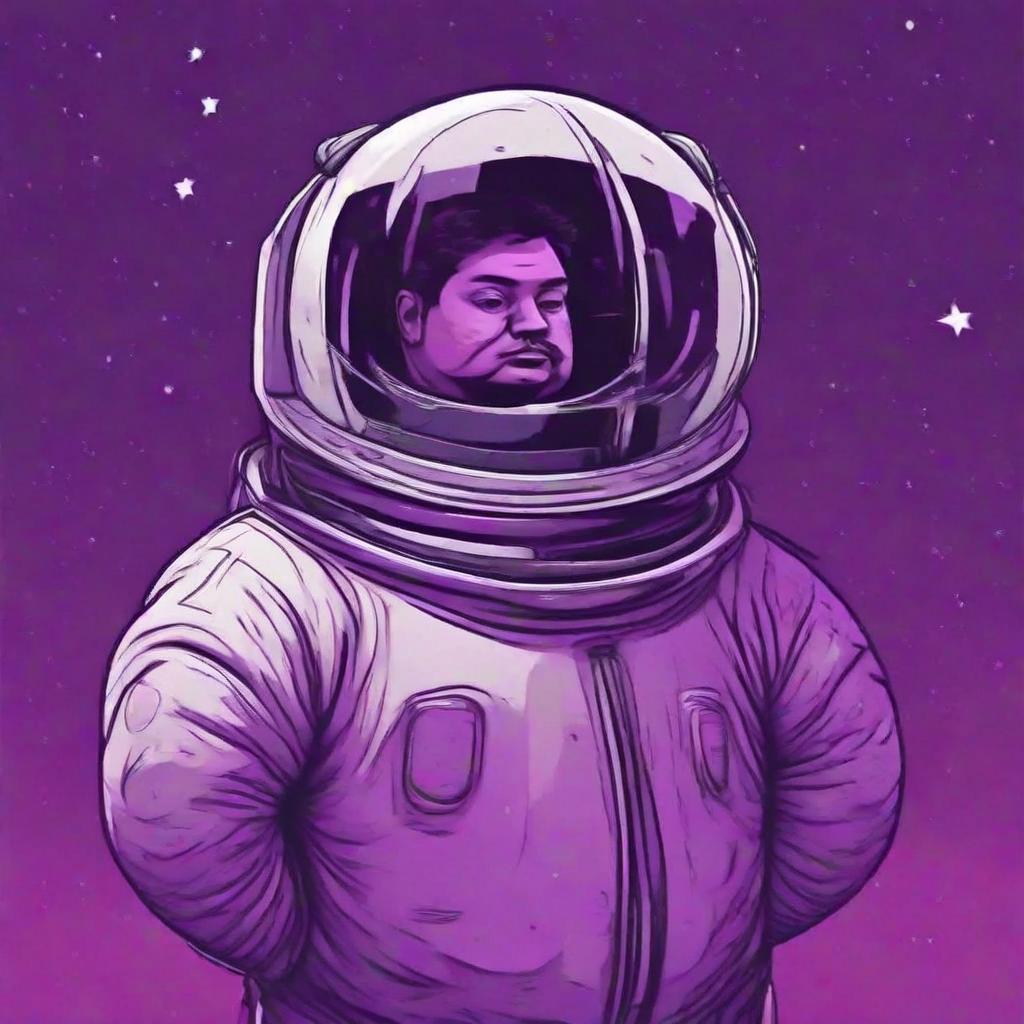}} &
        {\includegraphics[valign=c, width=\ww]{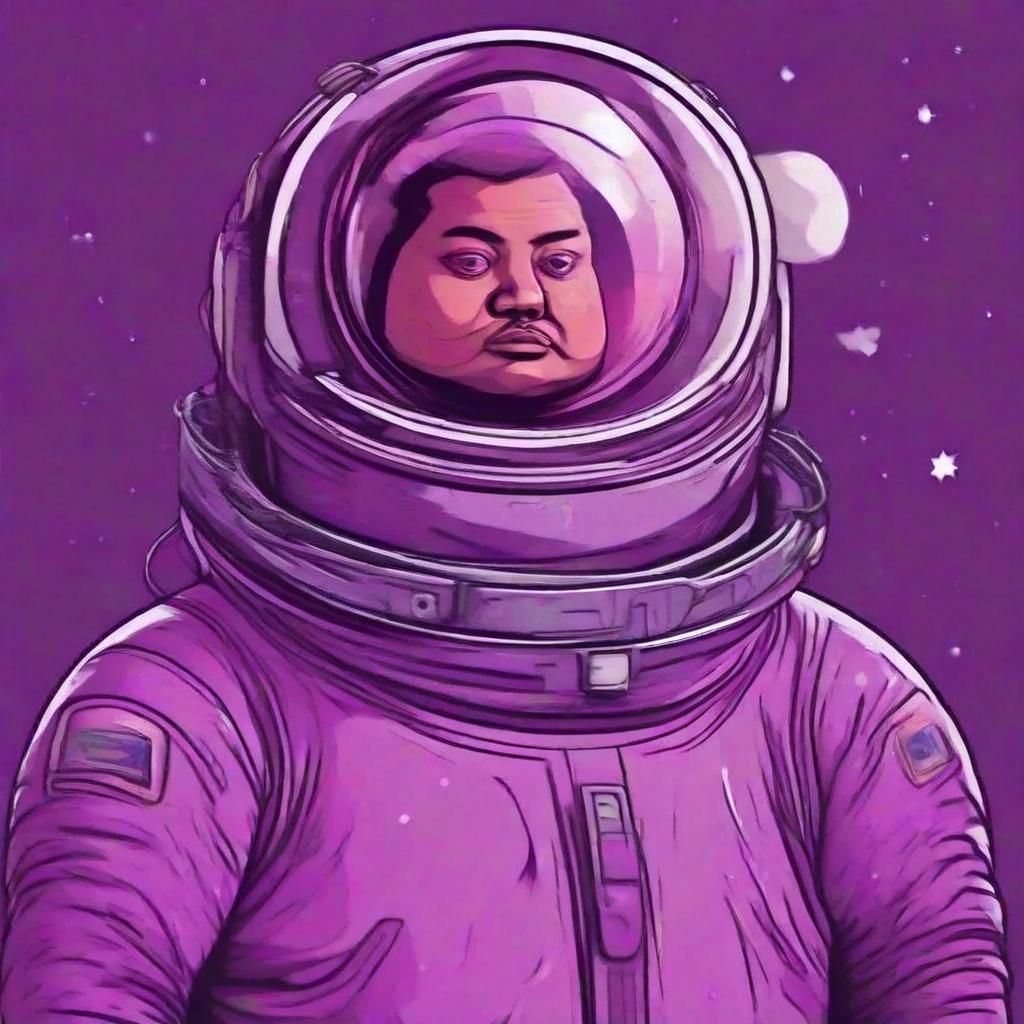}} &
        {\includegraphics[valign=c, width=\ww]{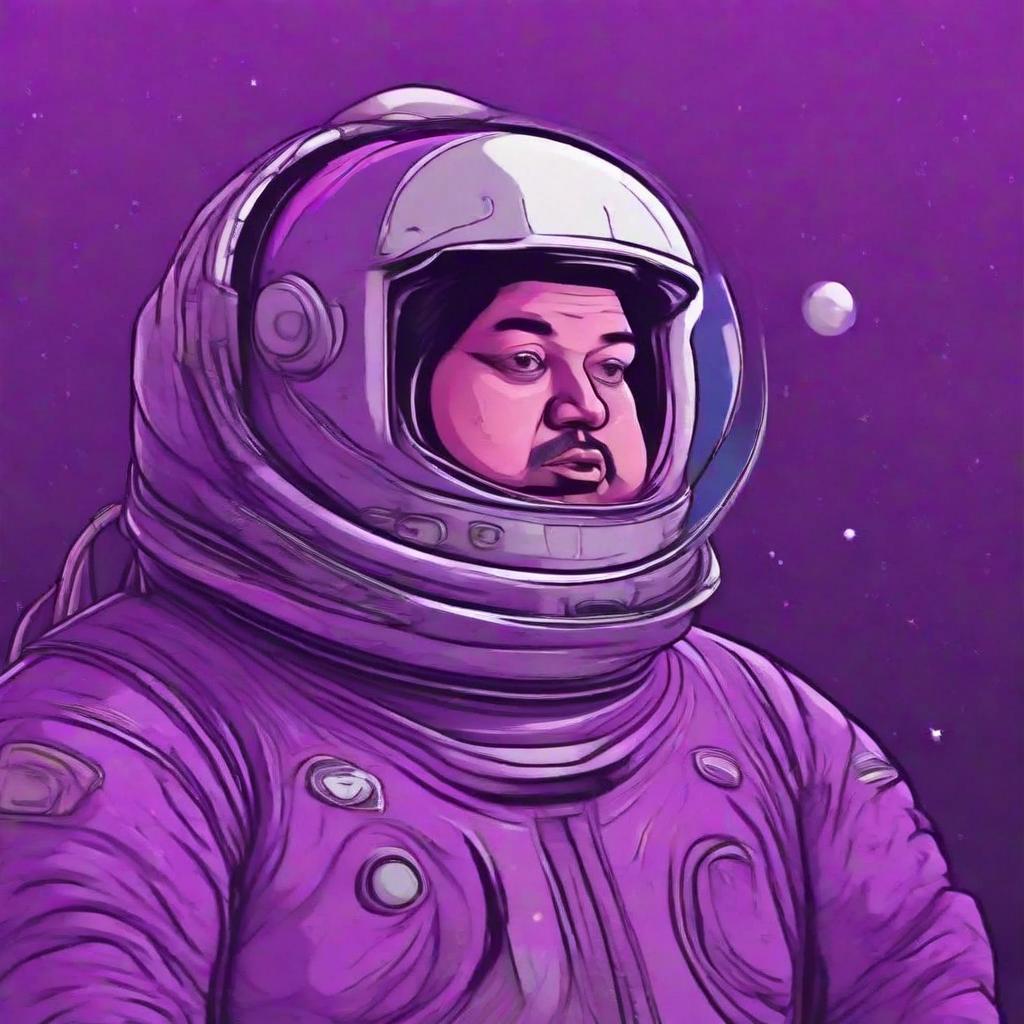}} &
        {\includegraphics[valign=c, width=\ww]{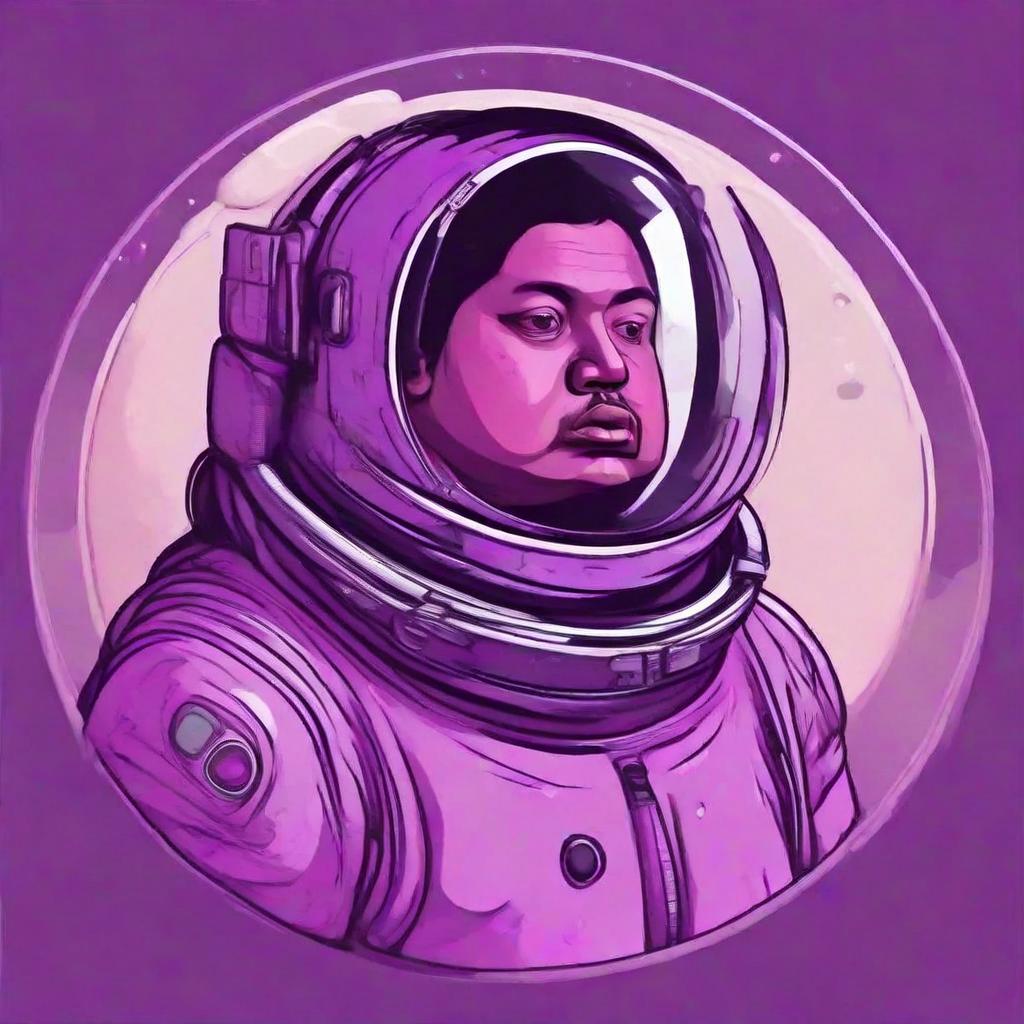}}
        \\
        \\

        \rotatebox[origin=c]{90}{Cluster 3} &
        {\includegraphics[valign=c, width=\ww]{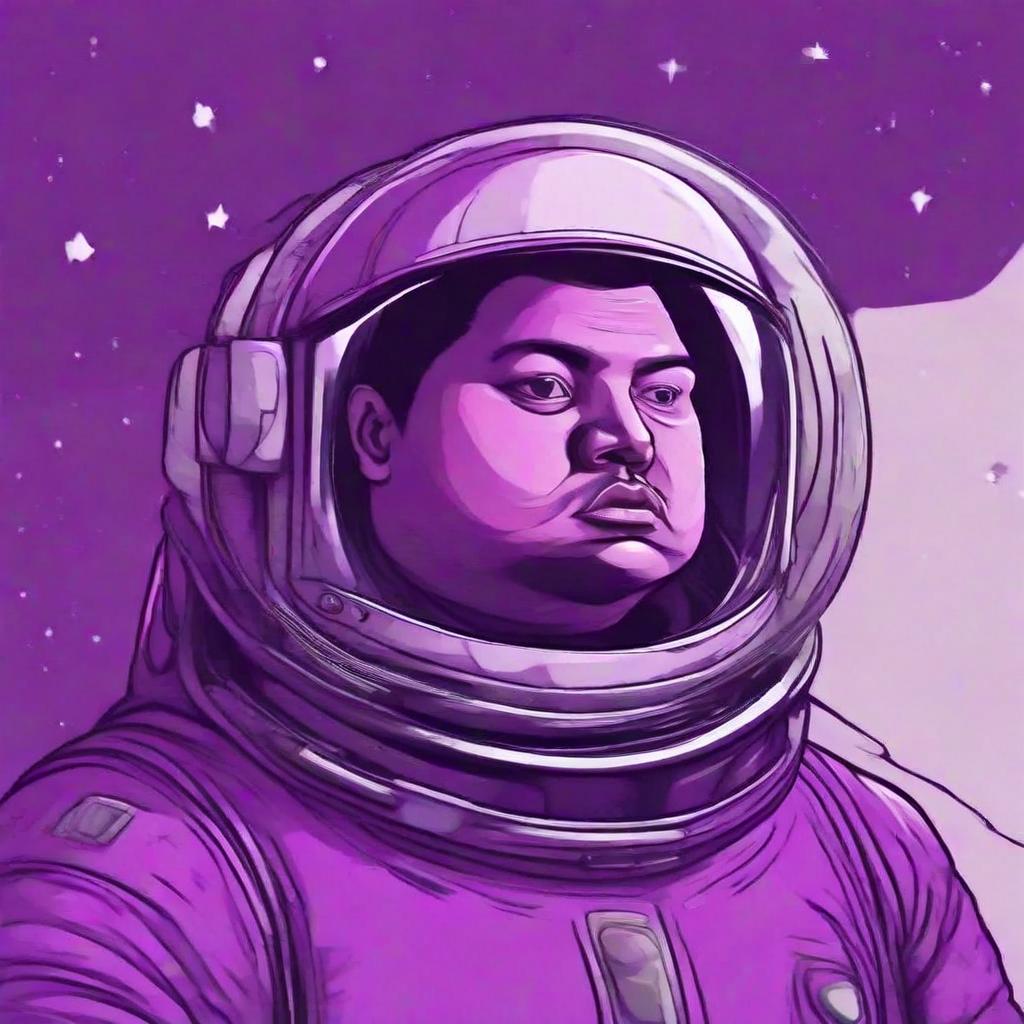}} &
        {\includegraphics[valign=c, width=\ww]{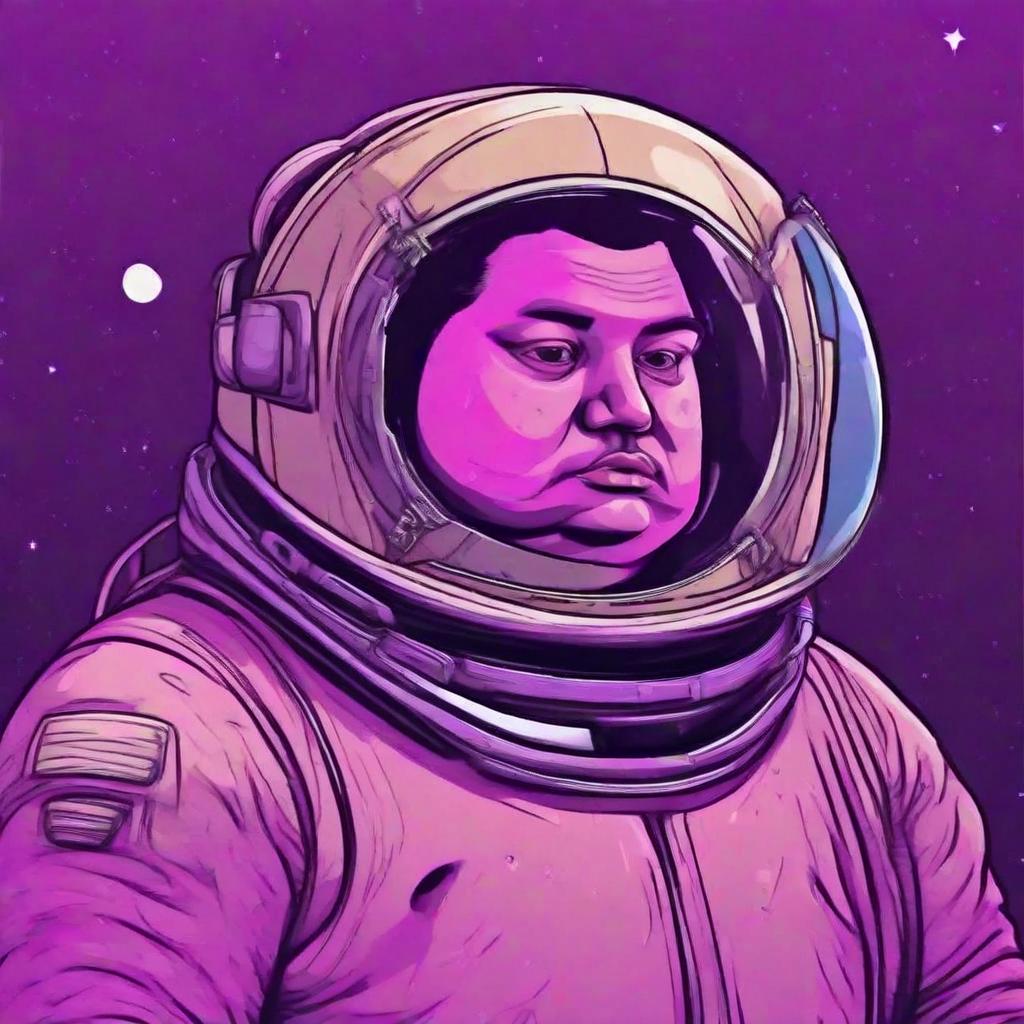}} &
        {\includegraphics[valign=c, width=\ww]{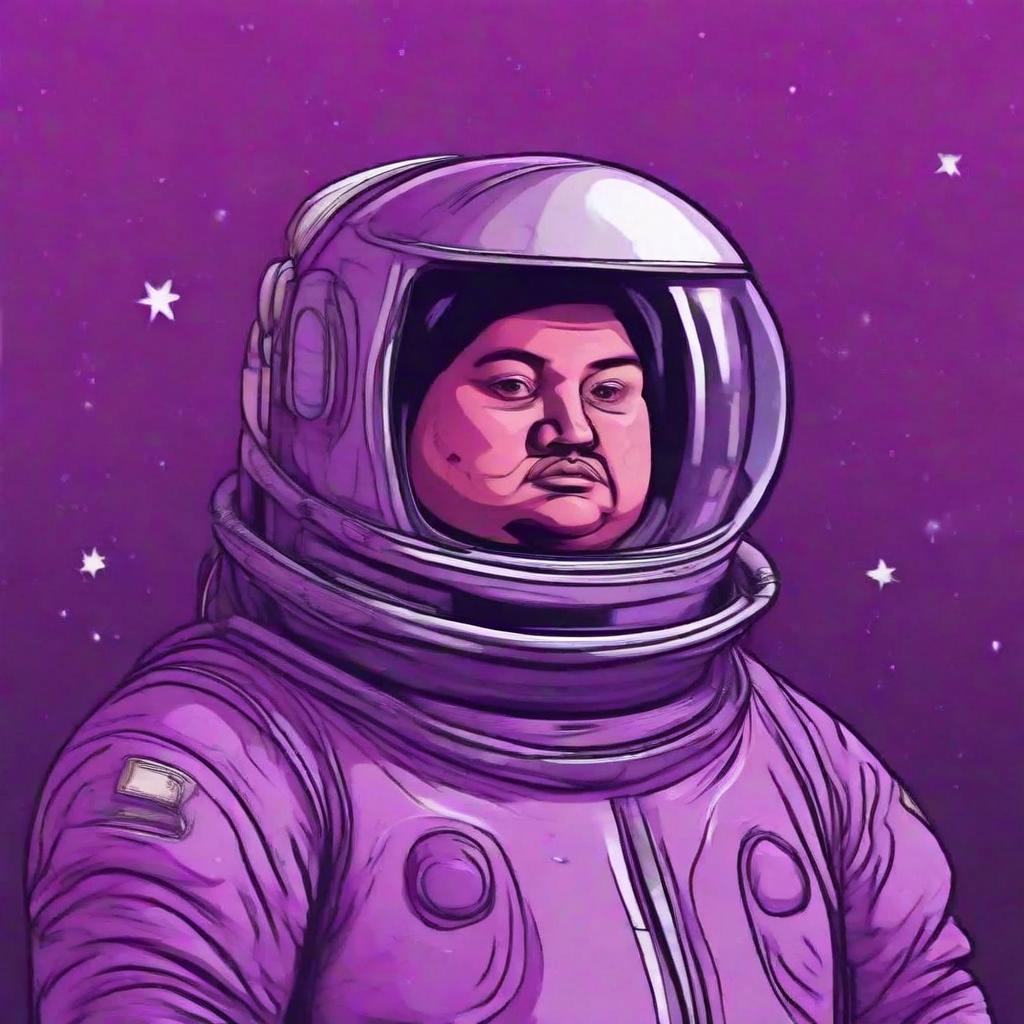}} &
        {\includegraphics[valign=c, width=\ww]{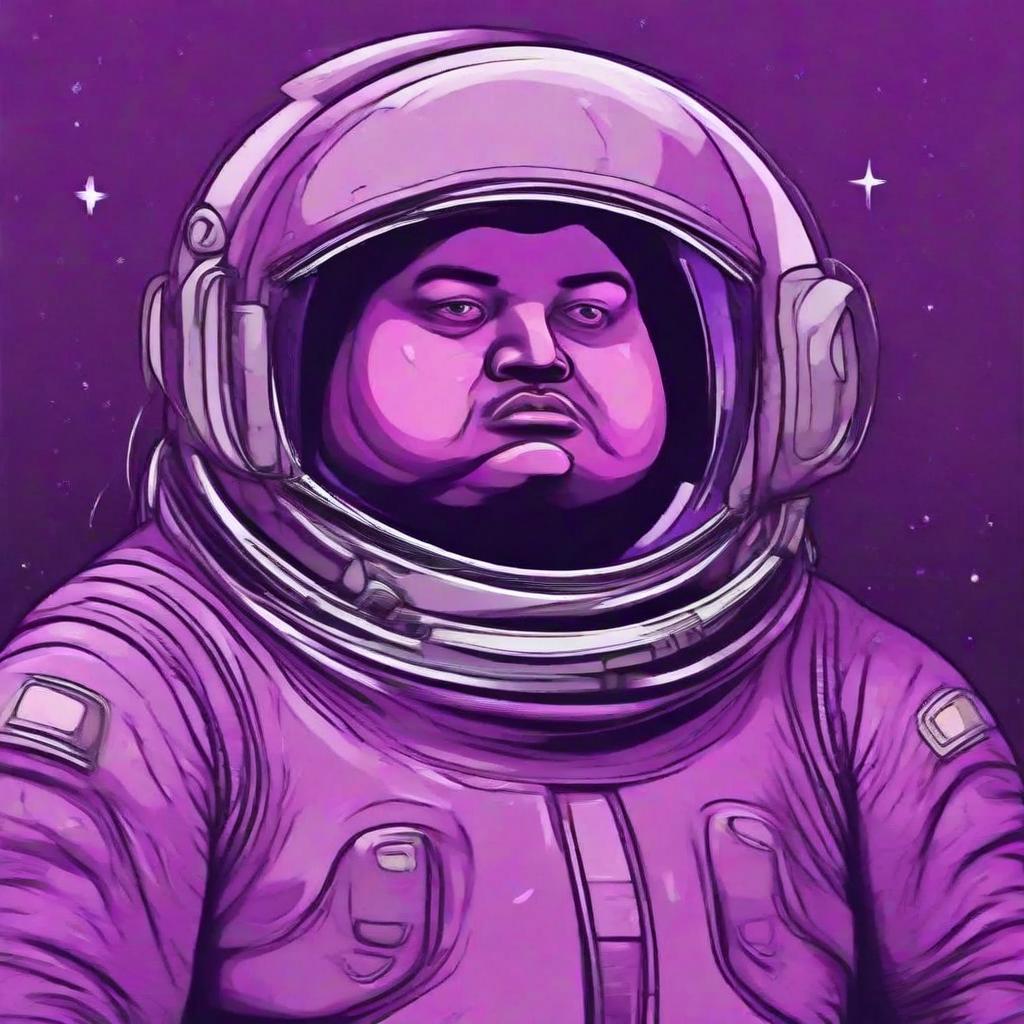}} &
        {\includegraphics[valign=c, width=\ww]{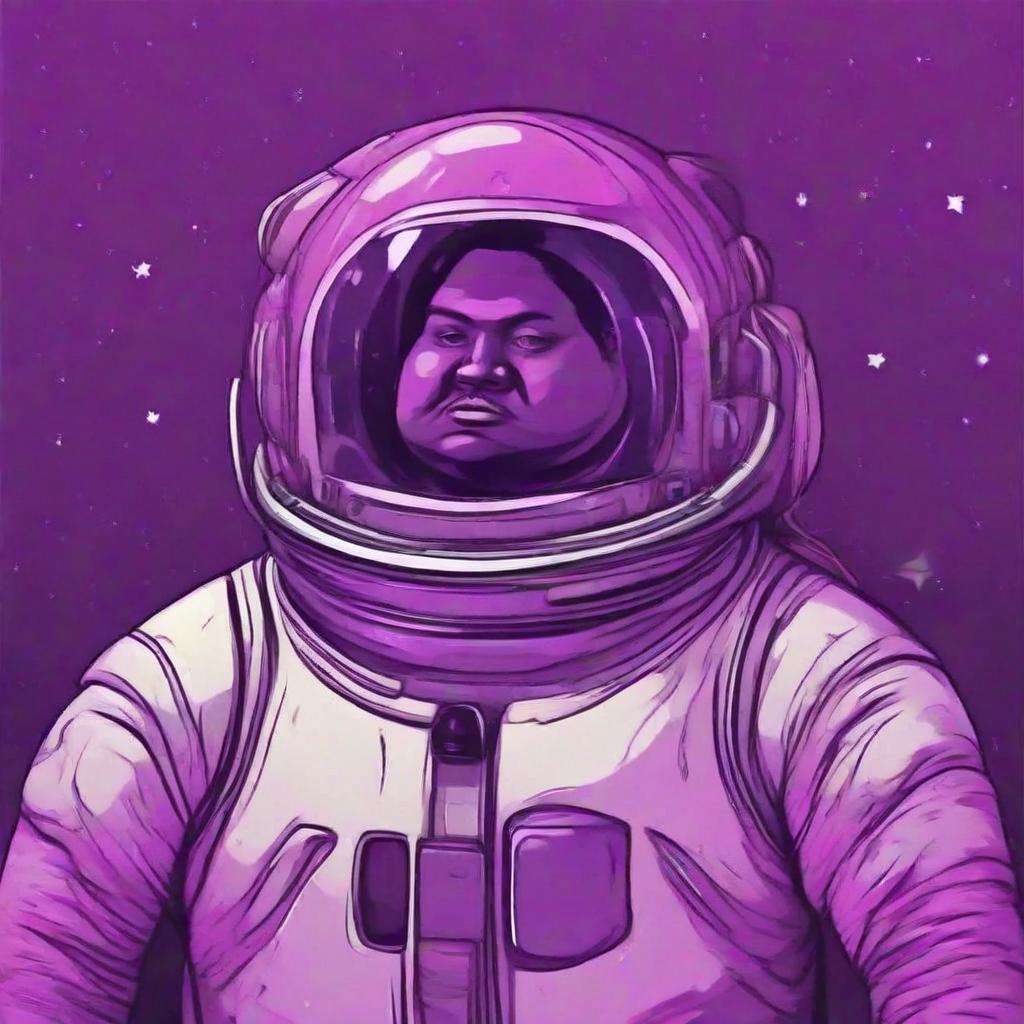}}

    \end{tabular}
    
    \caption{\textbf{Clustering visualization.} We visualize the clustering of images generated with the prompt \textit{``a purple astronaut, digital art, smooth, sharp focus, vector art''}. In the initial iteration (top three rows), our algorithm divides the generated images into three clusters: (1) emphasizing the astronaut's head, (2) an astronaut without a face, and (3) a full-body astronaut. Cluster 1 (top row) is the most cohesive cluster, and it is chosen for the identity extraction phase. In the subsequent iteration (bottom three rows), all images adopt the same extracted identity, and the clusters mainly differ from each other in the pose of the character.}
    \label{fig:clustering_visualization}
\end{figure*}

%% file: figures/dataset_non_memorization/fig.tex
\begin{figure*}[t]
    \centering
    \setlength{\tabcolsep}{0.5pt}
    \renewcommand{\arraystretch}{0.6}
    \setlength{\ww}{0.33\columnwidth}
    \begin{tabular}{c @{\hspace{20\tabcolsep}} ccccc}
        Generated &
        \\

        character &
        \multicolumn{5}{c}{Top 5 nearest neighbors}
        \\

        {\includegraphics[valign=c, width=\ww]{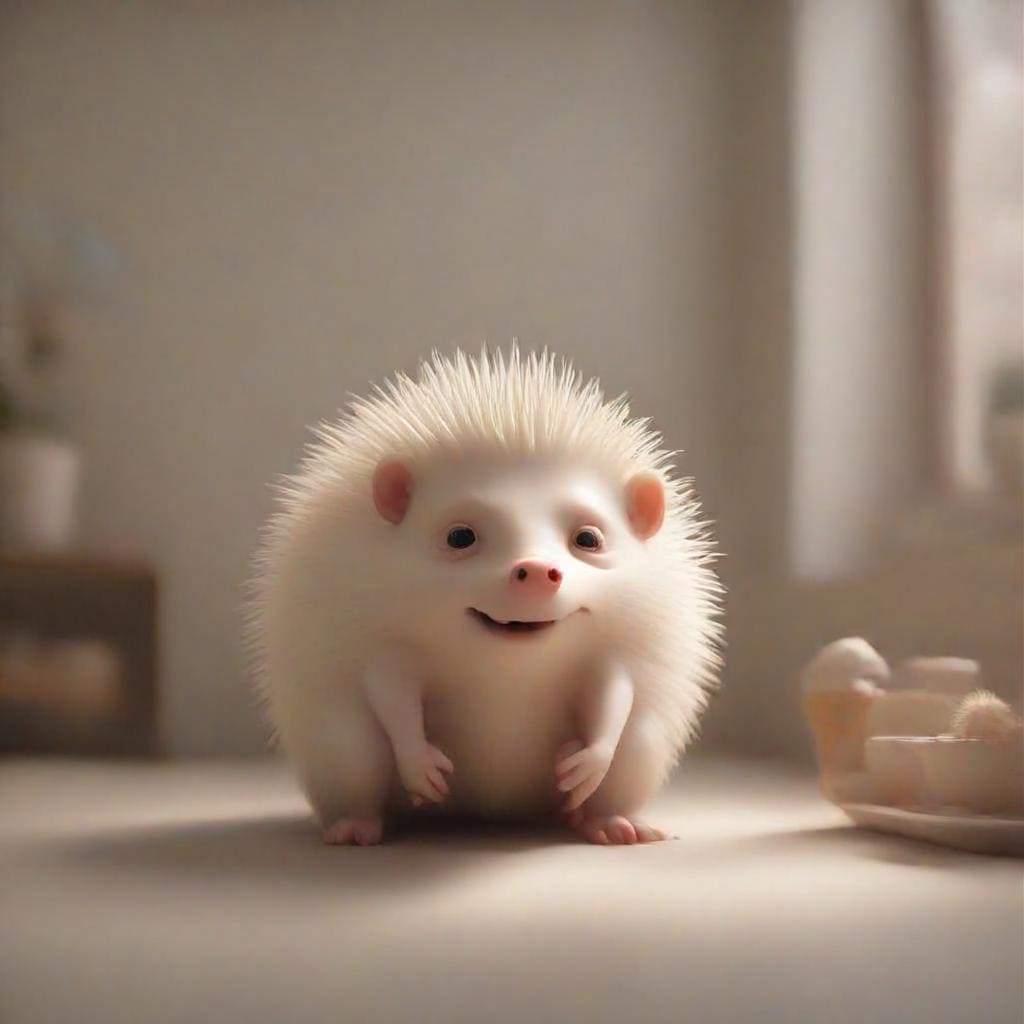}} &
        {\includegraphics[valign=c, width=\ww]{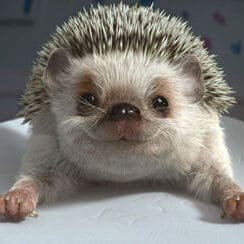}} &
        {\includegraphics[valign=c, width=\ww]{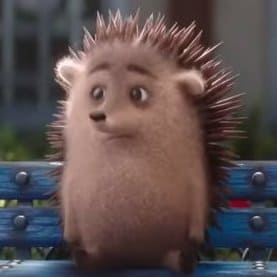}} &
        {\includegraphics[valign=c, width=\ww]{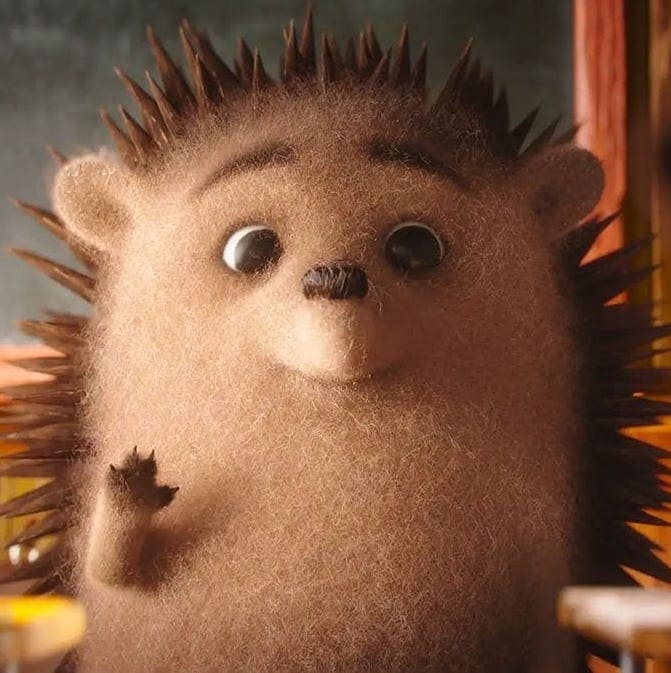}} &
        {\includegraphics[valign=c, width=\ww]{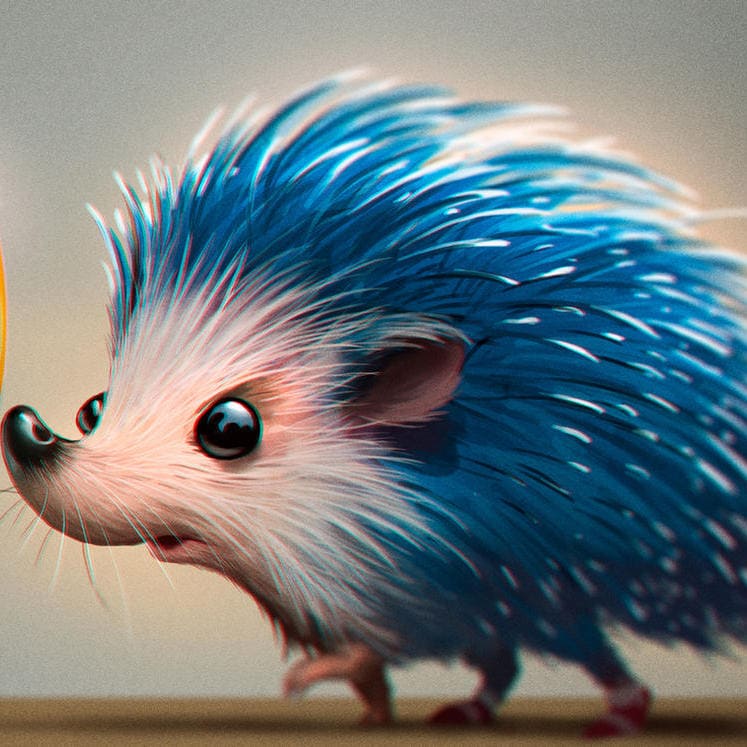}} &
        {\includegraphics[valign=c, width=\ww]{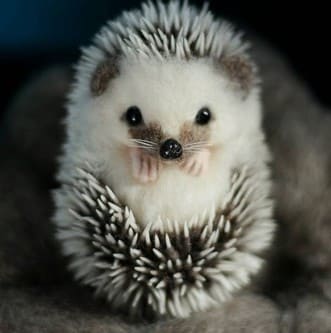}}
        \\
        \\

        {\includegraphics[valign=c, width=\ww]{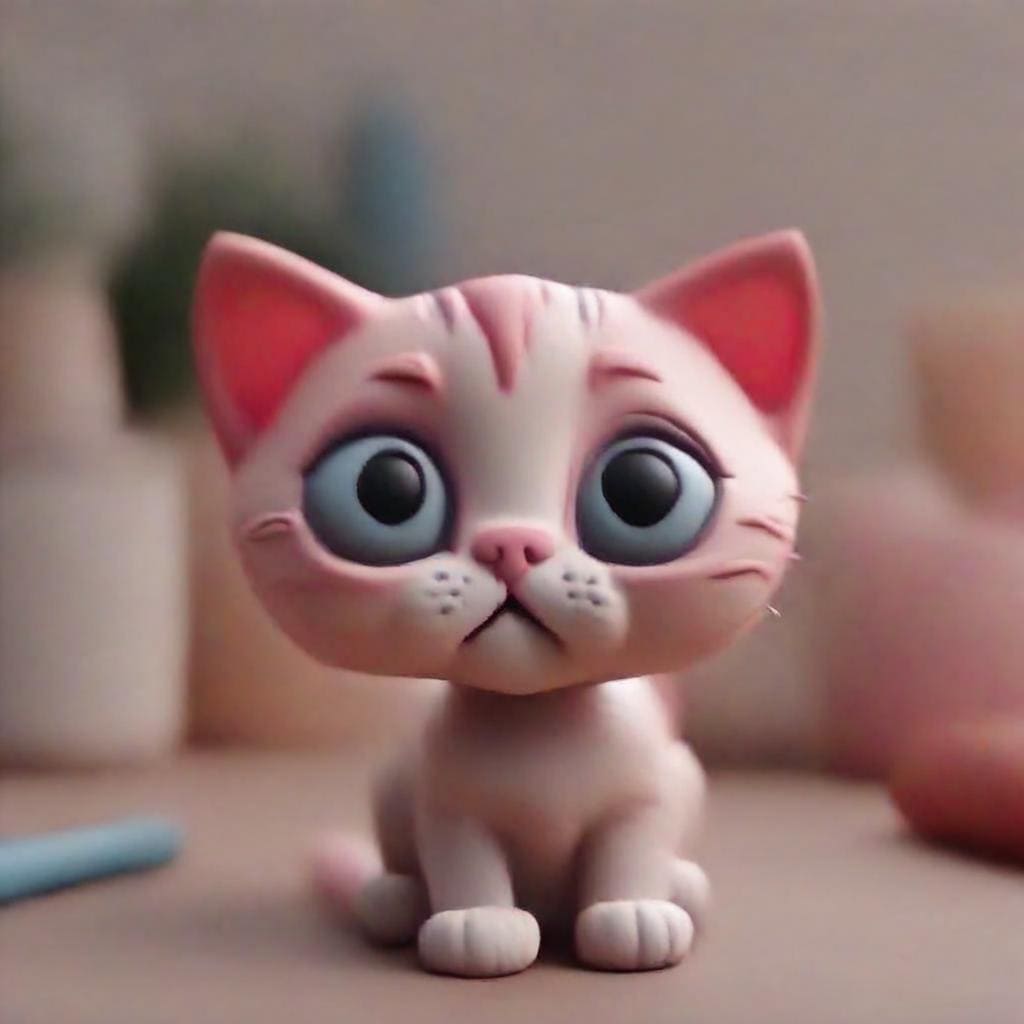}} &
        {\includegraphics[valign=c, width=\ww]{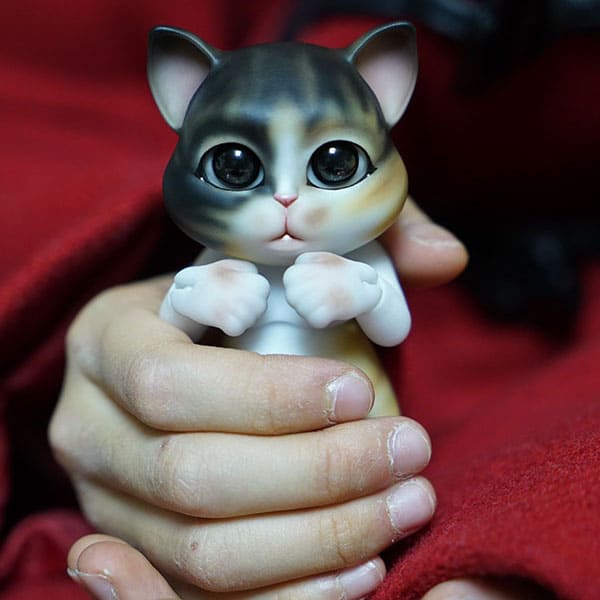}} &
        {\includegraphics[valign=c, width=\ww]{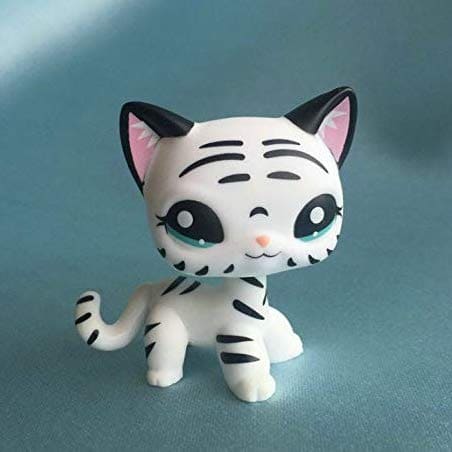}} &
        {\includegraphics[valign=c, width=\ww]{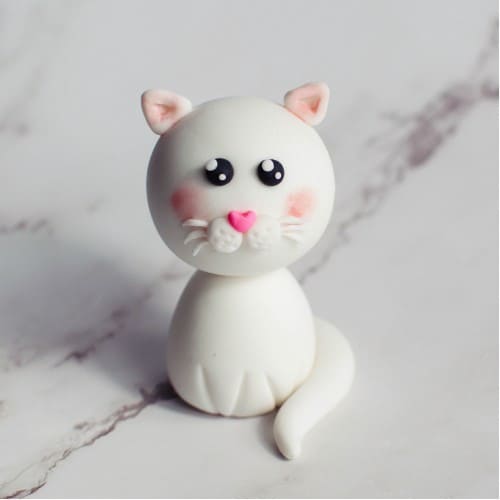}} &
        {\includegraphics[valign=c, width=\ww]{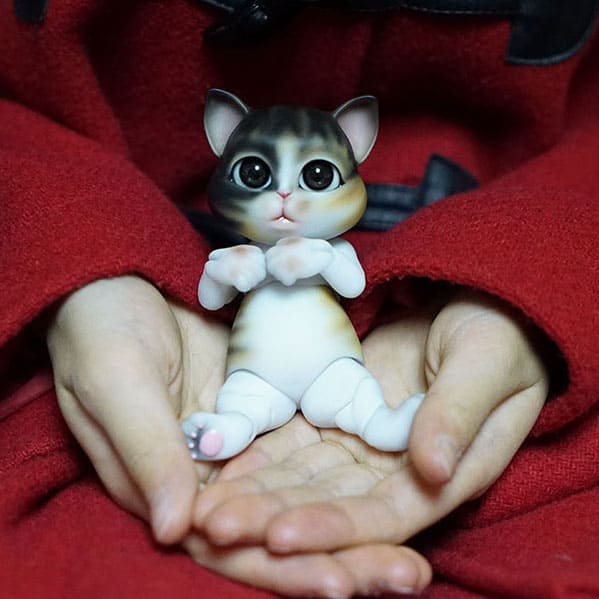}} &
        {\includegraphics[valign=c, width=\ww]{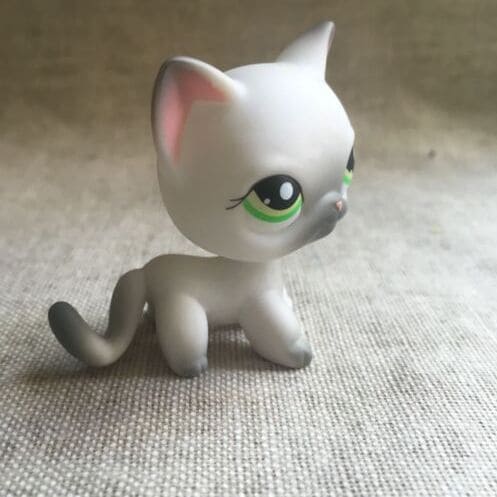}}
        \\
        \\

        {\includegraphics[valign=c, width=\ww]{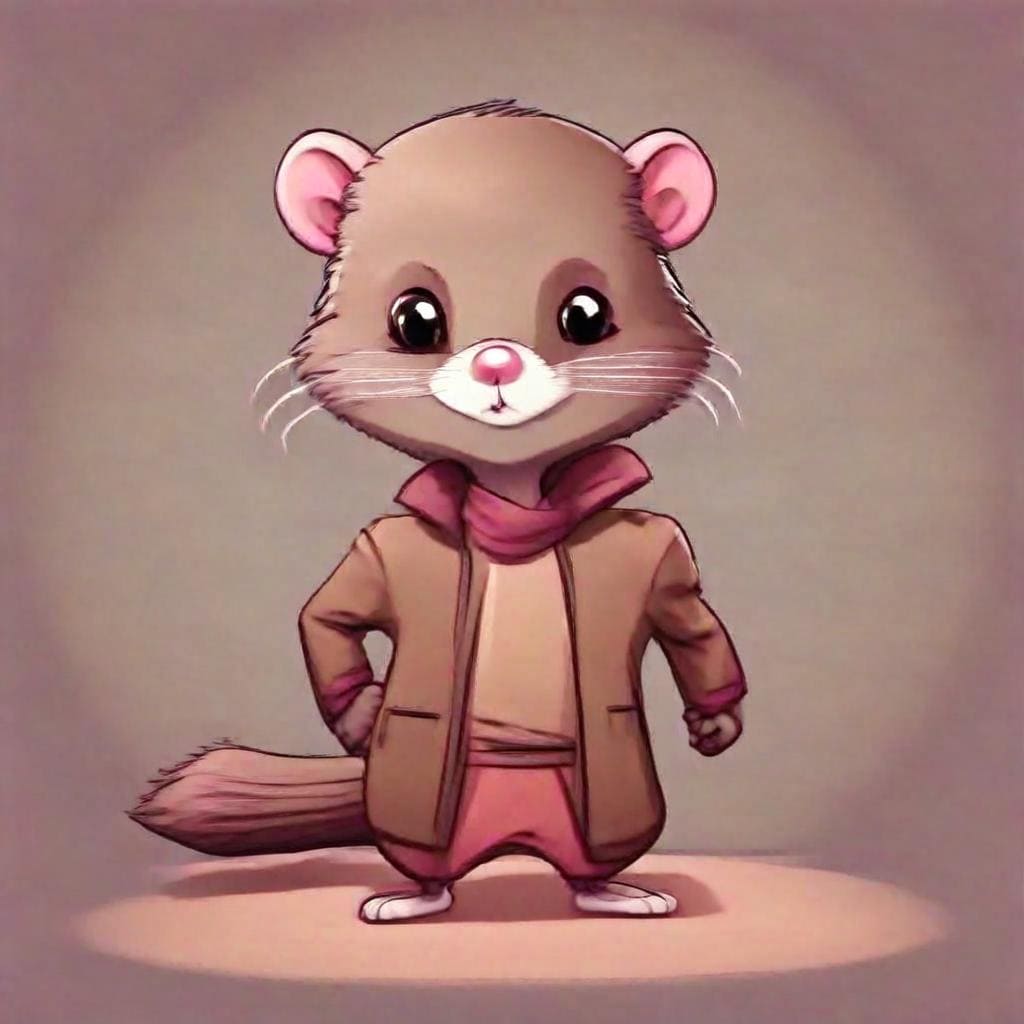}} &
        {\includegraphics[valign=c, width=\ww]{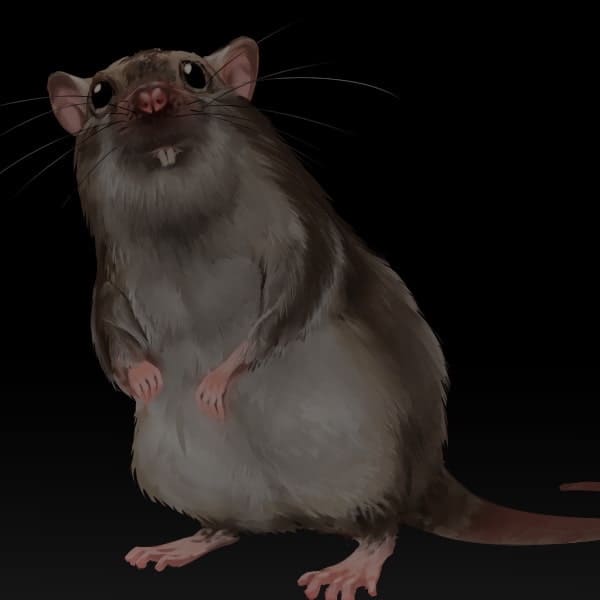}} &
        {\includegraphics[valign=c, width=\ww]{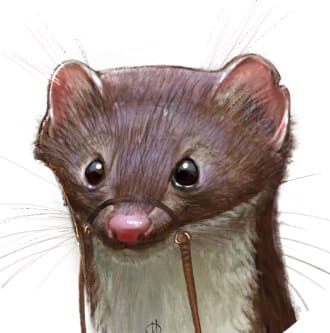}} &
        {\includegraphics[valign=c, width=\ww]{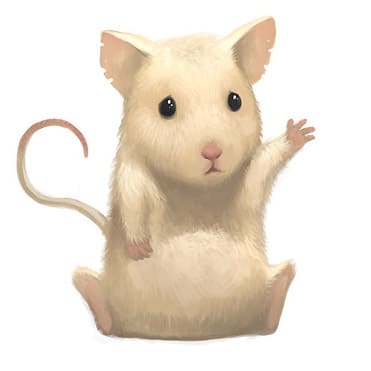}} &
        {\includegraphics[valign=c, width=\ww]{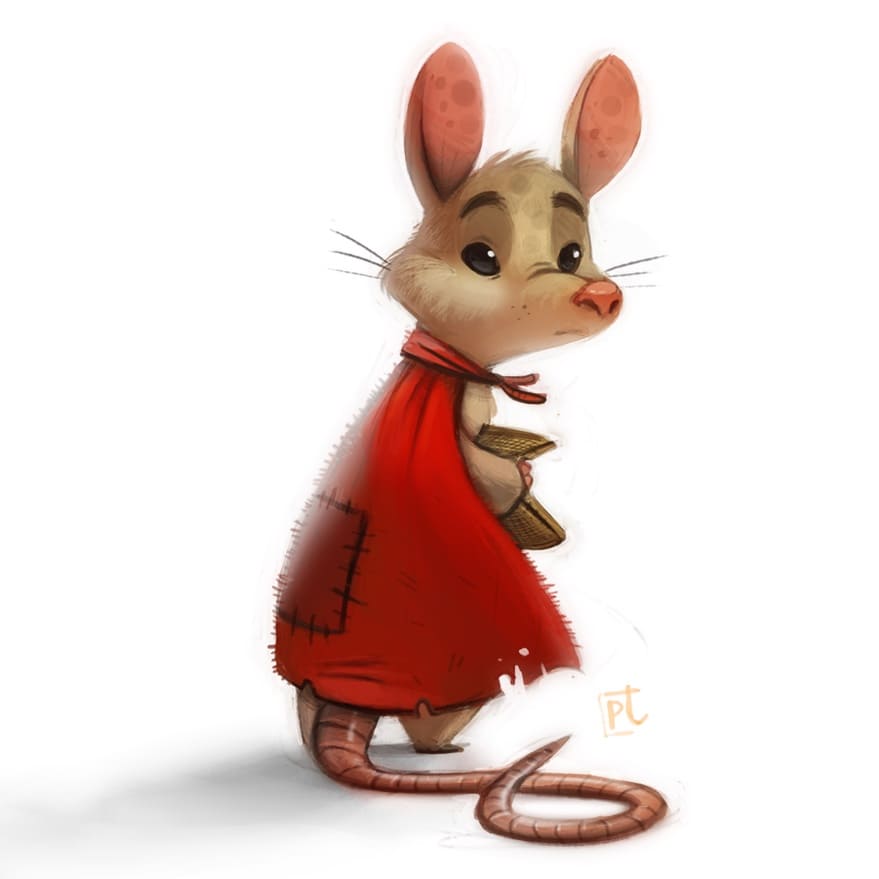}} &
        {\includegraphics[valign=c, width=\ww]{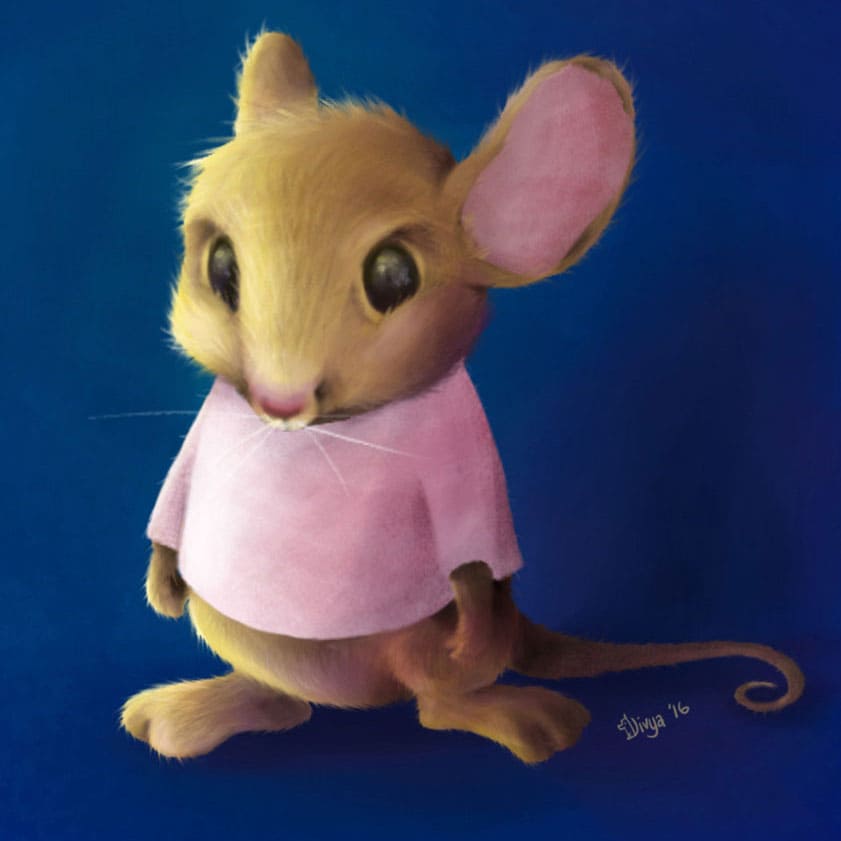}}
        \\
        \\

        {\includegraphics[valign=c, width=\ww]{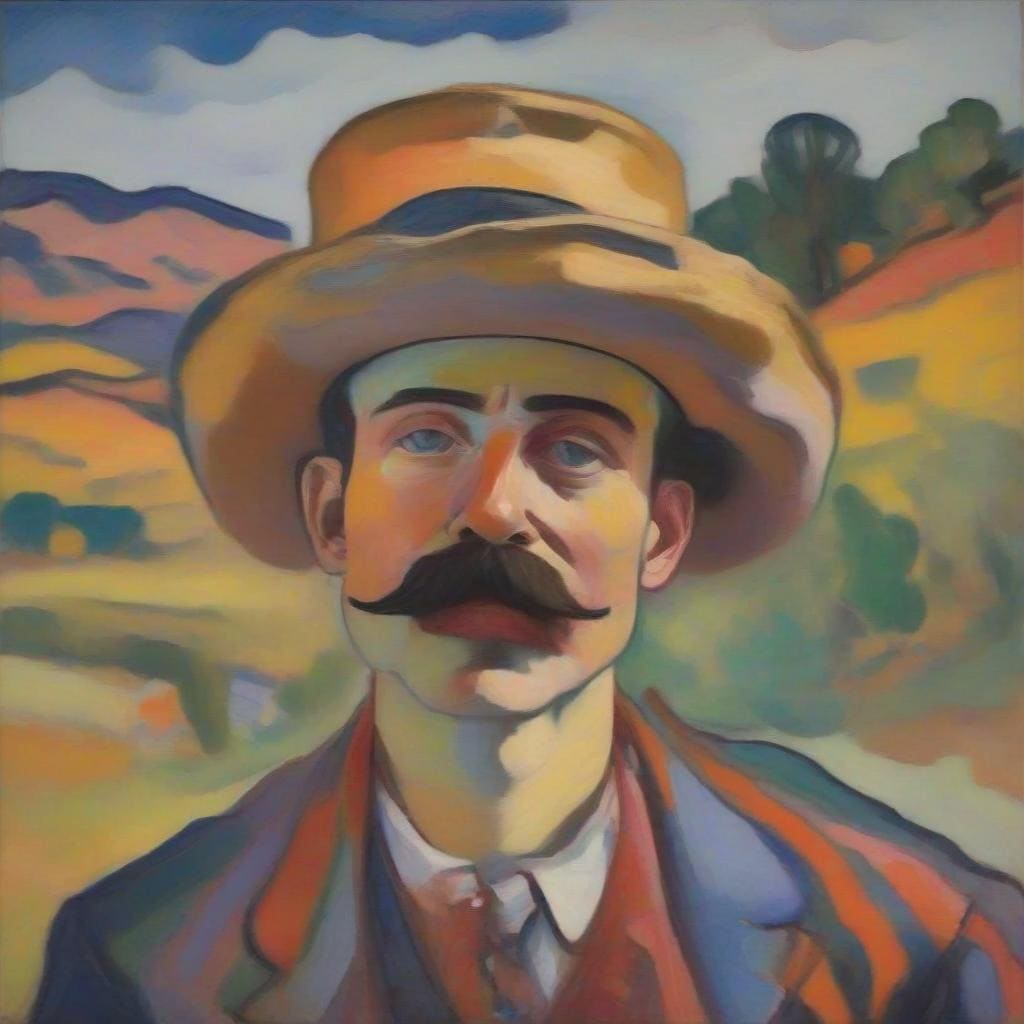}} &
        {\includegraphics[valign=c, width=\ww]{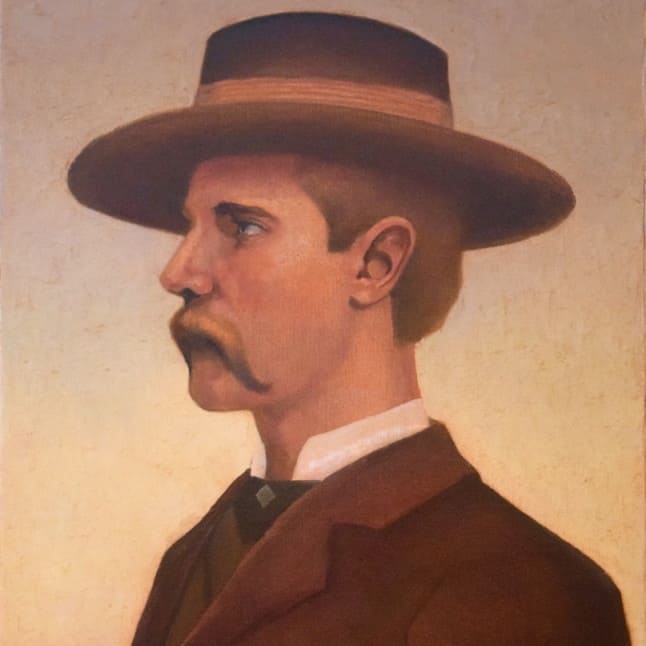}} &
        {\includegraphics[valign=c, width=\ww]{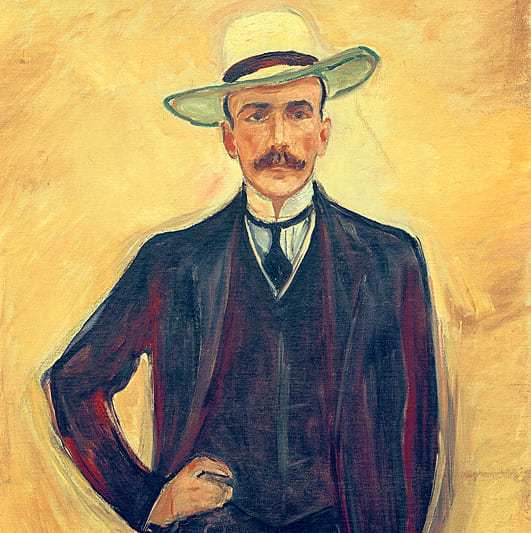}} &
        {\includegraphics[valign=c, width=\ww]{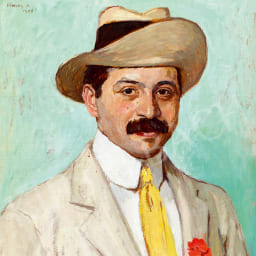}} &
        {\includegraphics[valign=c, width=\ww]{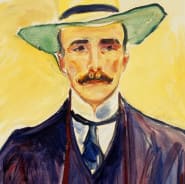}} &
        {\includegraphics[valign=c, width=\ww]{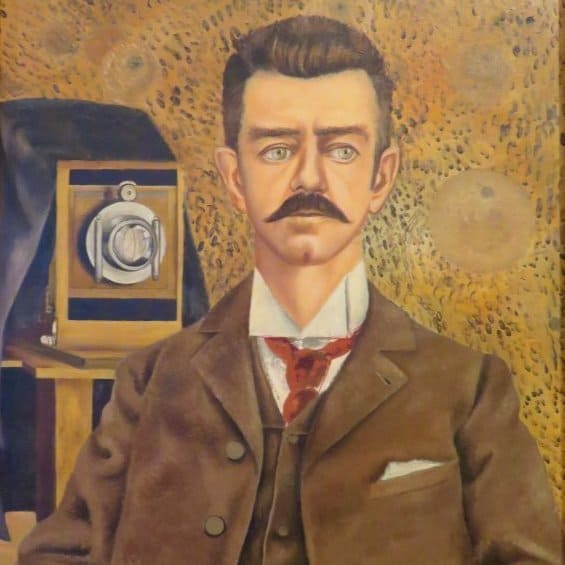}}
        \\
        \\

        {\includegraphics[valign=c, width=\ww]{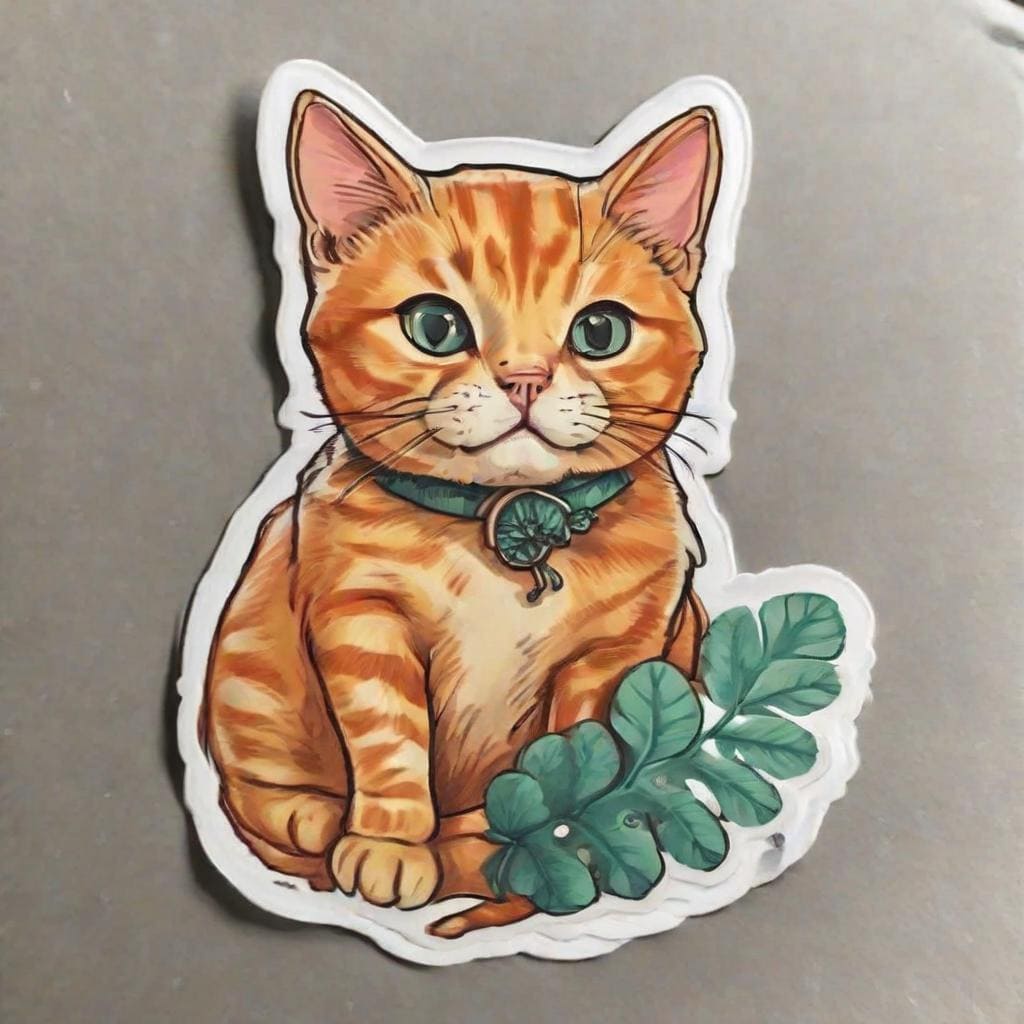}} &
        {\includegraphics[valign=c, width=\ww]{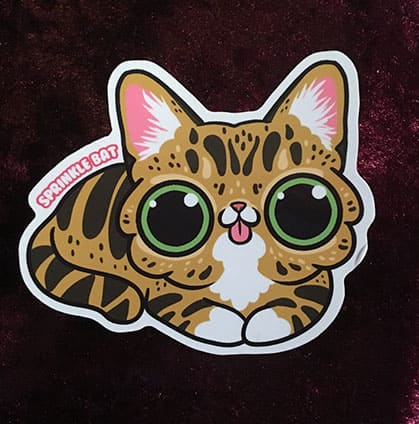}} &
        {\includegraphics[valign=c, width=\ww]{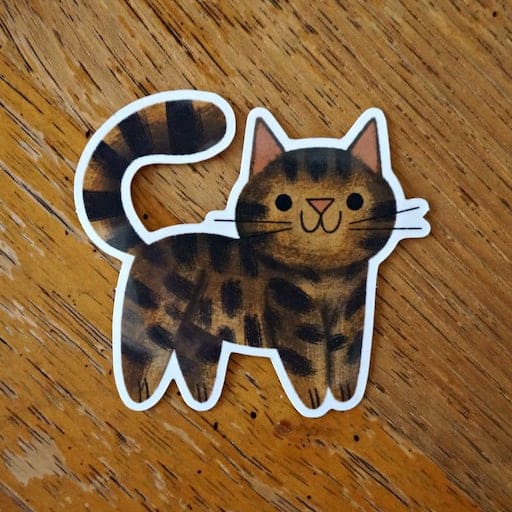}} &
        {\includegraphics[valign=c, width=\ww]{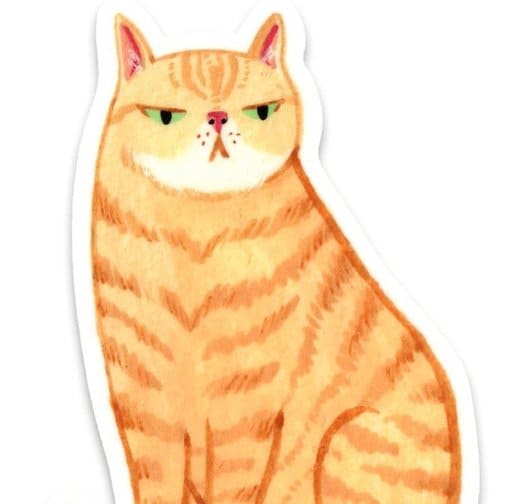}} &
        {\includegraphics[valign=c, width=\ww]{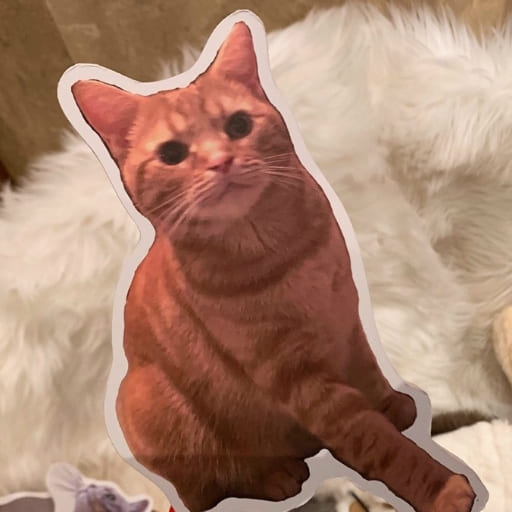}} &
        {\includegraphics[valign=c, width=\ww]{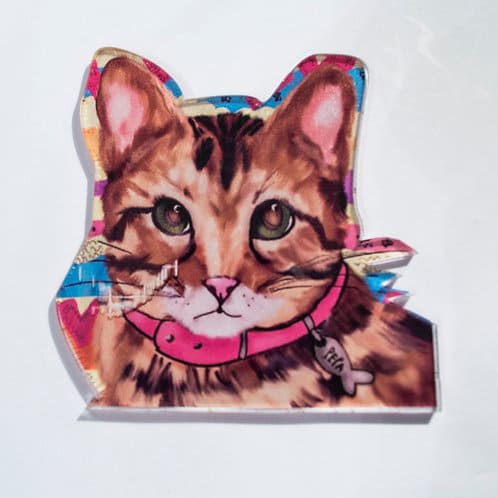}}
        \\
        \\

        {\includegraphics[valign=c, width=\ww]{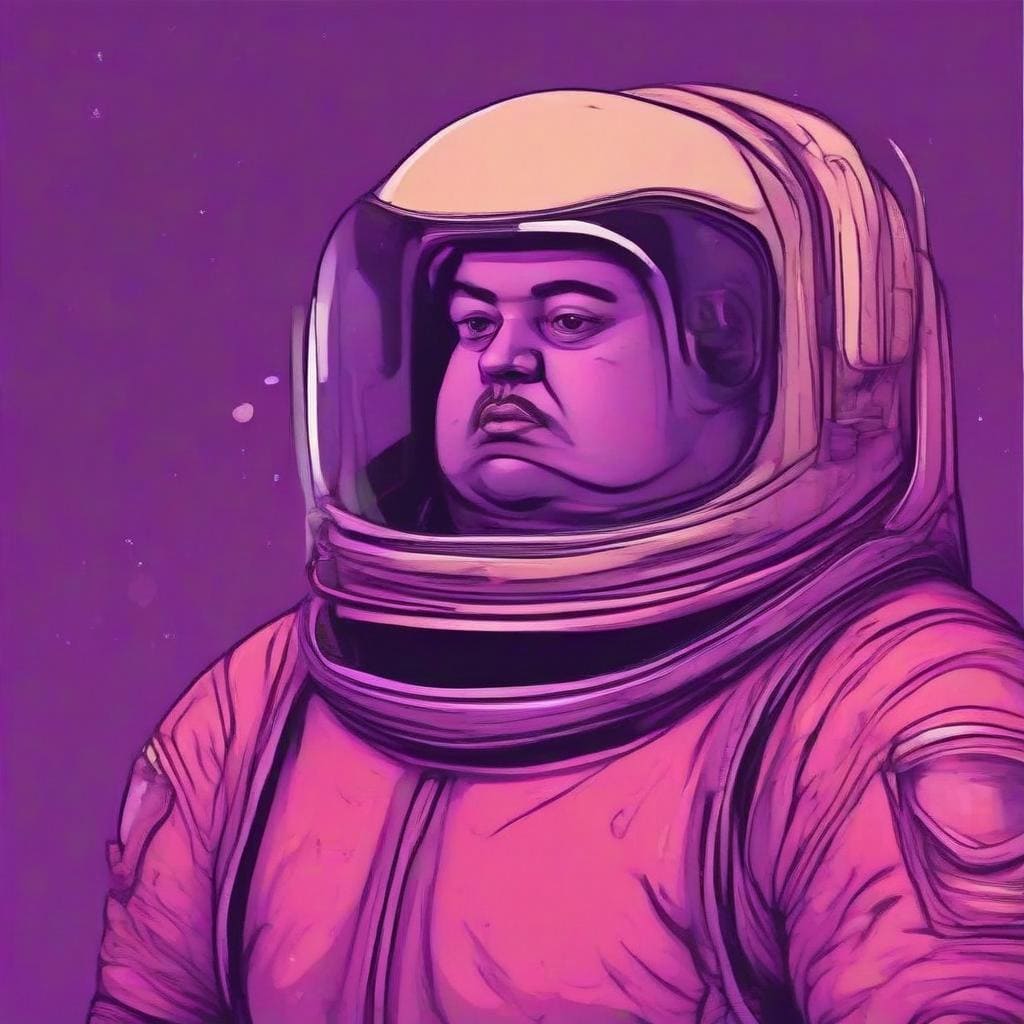}} &
        {\includegraphics[valign=c, width=\ww]{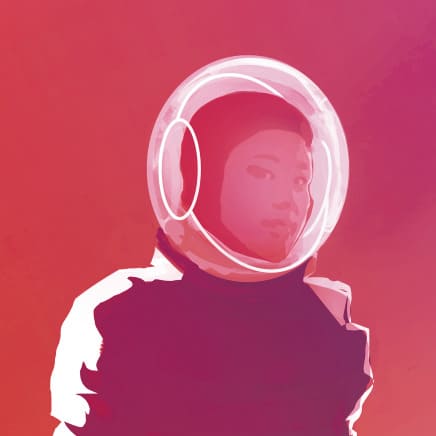}} &
        {\includegraphics[valign=c, width=\ww]{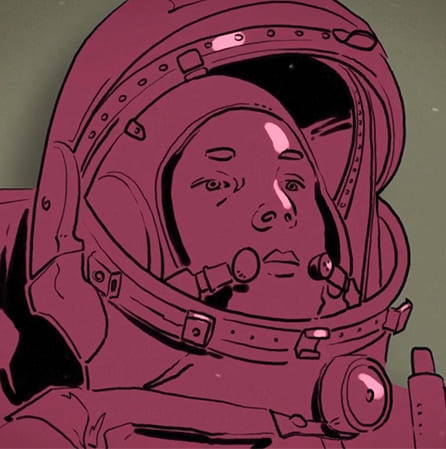}} &
        {\includegraphics[valign=c, width=\ww]{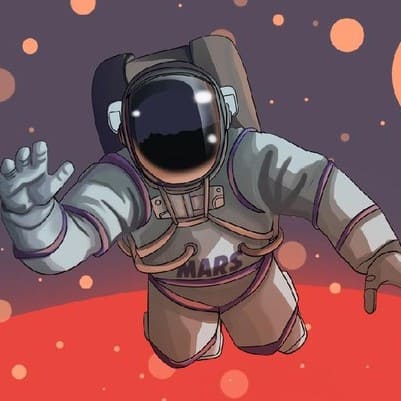}} &
        {\includegraphics[valign=c, width=\ww]{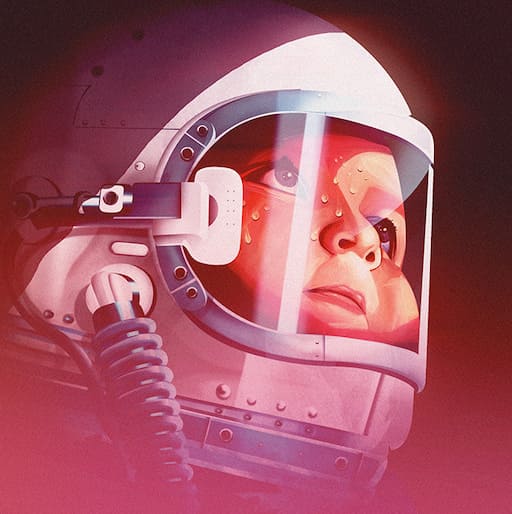}} &
        {\includegraphics[valign=c, width=\ww]{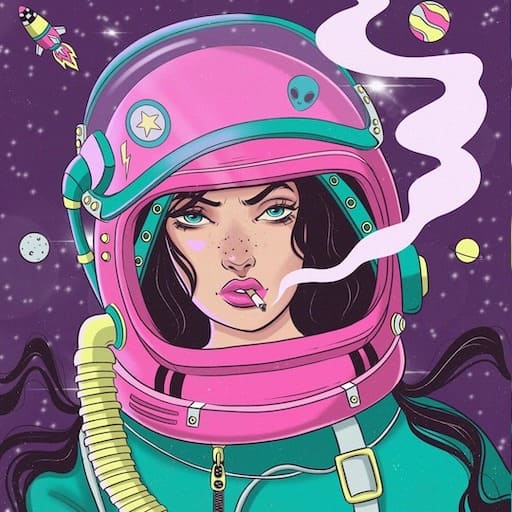}}
    \end{tabular}
    
    \caption{\textbf{Dataset non-memorization.} We found the top 5 nearest neighbors in the LAION-5B dataset~\cite{Schuhmann2022LAION5BAO}, in terms of CLIP~\cite{Radford2021LearningTV} image similarity, for a few representative characters from our paper, using an open-source solution \cite{clip_retrival}. As can be seen, our method does not simply memorize images from the LAION-5B dataset.}
    \label{fig:dataset_non_memorization}
\end{figure*}

%% file: figures/sd2_results/fig.tex
\begin{figure*}[t]
    \centering
    \setlength{\tabcolsep}{0.5pt}
    \renewcommand{\arraystretch}{1.0}
    \setlength{\ww}{0.4\columnwidth}
    \begin{tabular}{ccccc}
        &&&&
        \textit{``holding an}
        \\

        &
        \textit{``in the park''} &
        \textit{``reading a book''} &
        \textit{``at the beach''} &
        \textit{avocado''}
        \\

        {\includegraphics[valign=c, width=\ww]{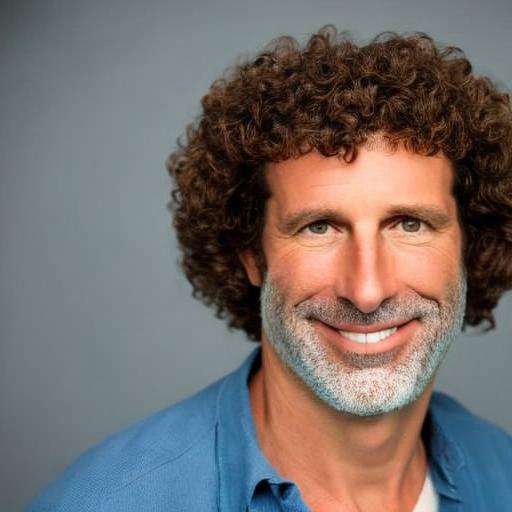}} &
        {\includegraphics[valign=c, width=\ww]{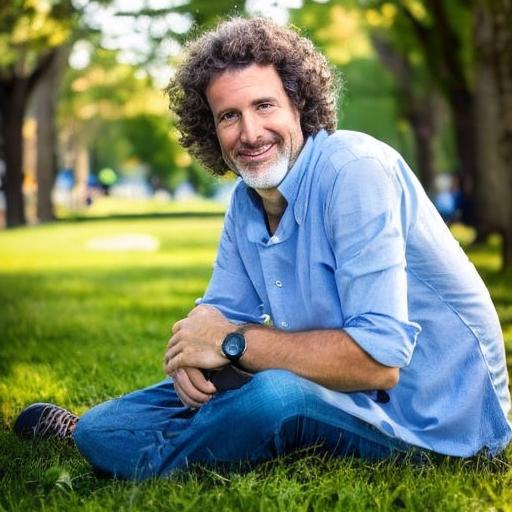}} &
        {\includegraphics[valign=c, width=\ww]{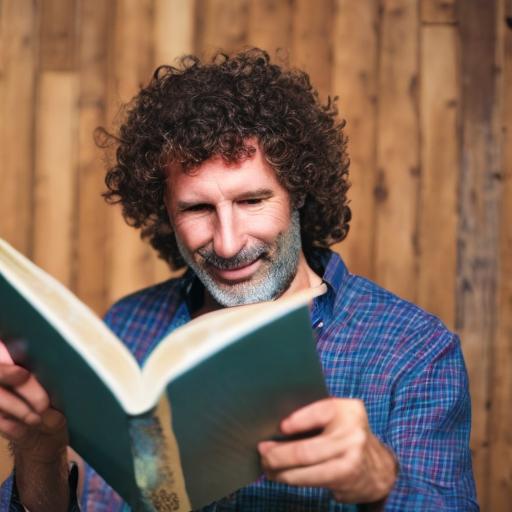}} &
        {\includegraphics[valign=c, width=\ww]{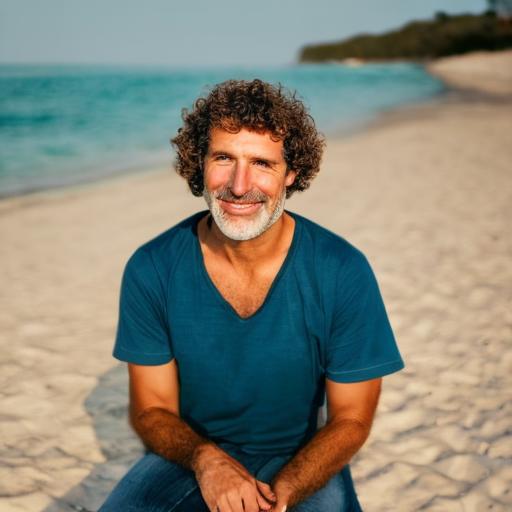}} &
        {\includegraphics[valign=c, width=\ww]{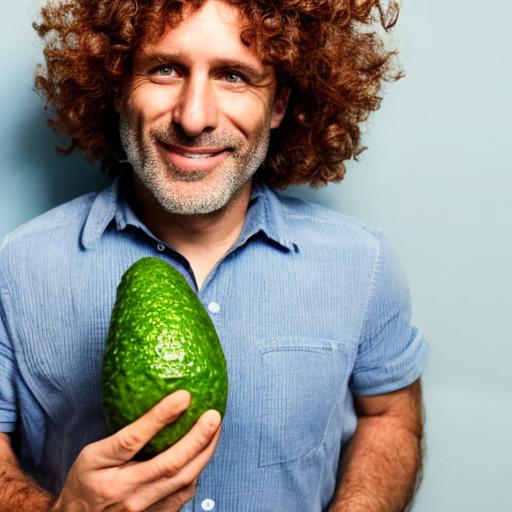}}
        \\

        \multicolumn{5}{c}{\textit{``a photo of a 50 years old man with curly hair''}}
        \\
        \\

        {\includegraphics[valign=c, width=\ww]{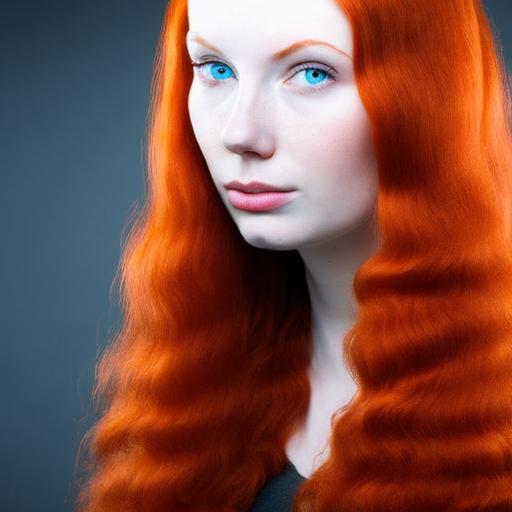}} &
        {\includegraphics[valign=c, width=\ww]{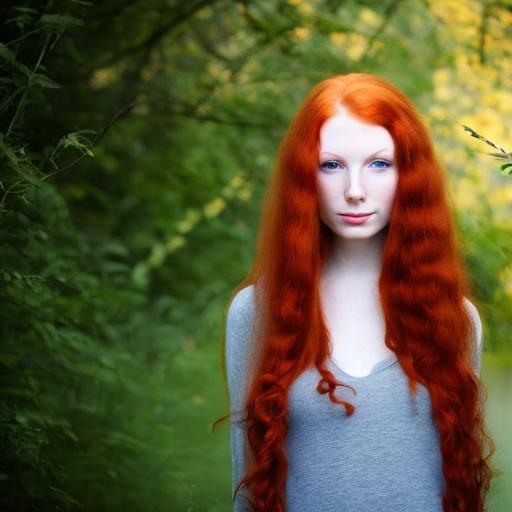}} &
        {\includegraphics[valign=c, width=\ww]{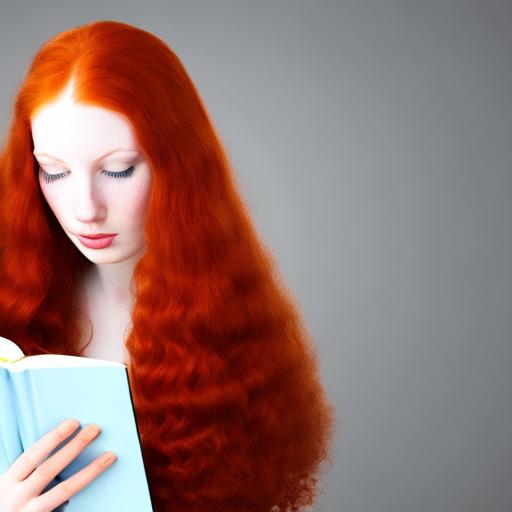}} &
        {\includegraphics[valign=c, width=\ww]{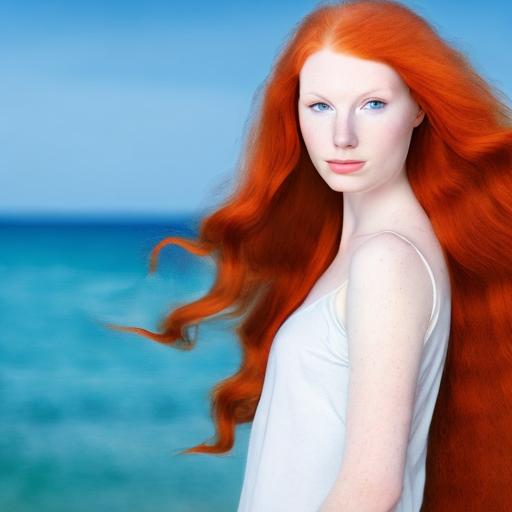}} &
        {\includegraphics[valign=c, width=\ww]{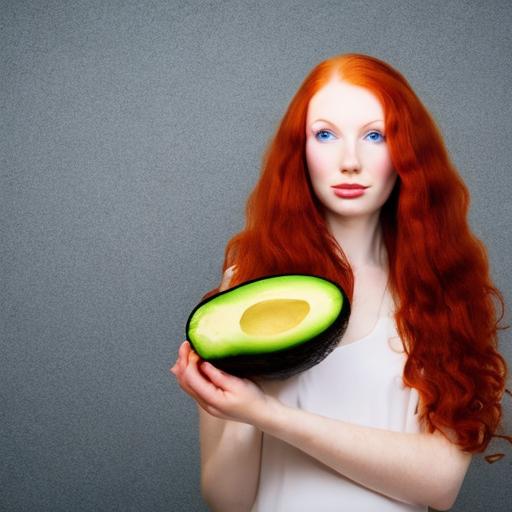}}
        \\

        \multicolumn{5}{c}{\textit{``a photo of a woman with long ginger hair''}}
        \\
        \\

        {\includegraphics[valign=c, width=\ww]{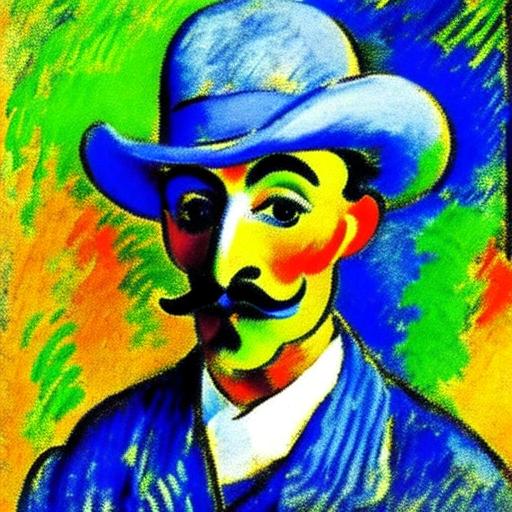}} &
        {\includegraphics[valign=c, width=\ww]{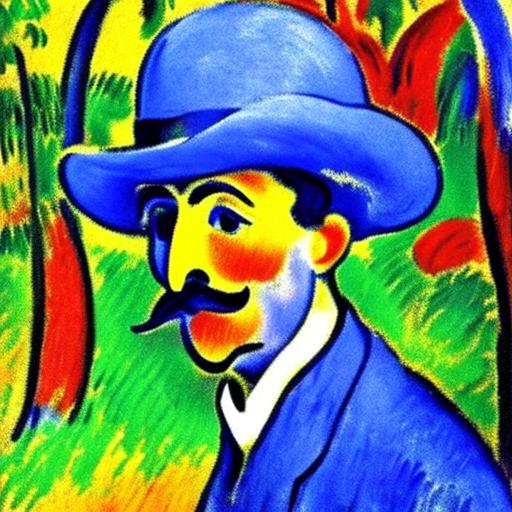}} &
        {\includegraphics[valign=c, width=\ww]{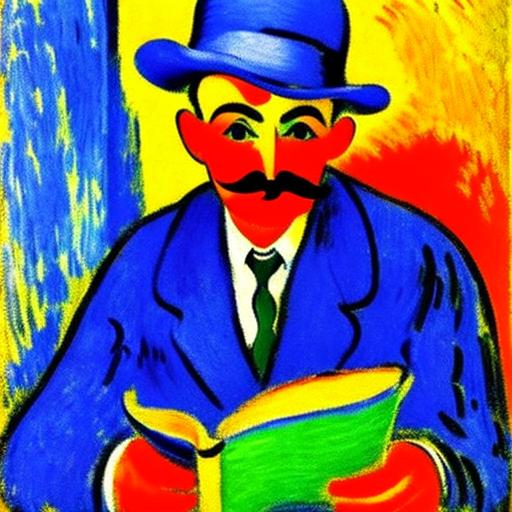}} &
        {\includegraphics[valign=c, width=\ww]{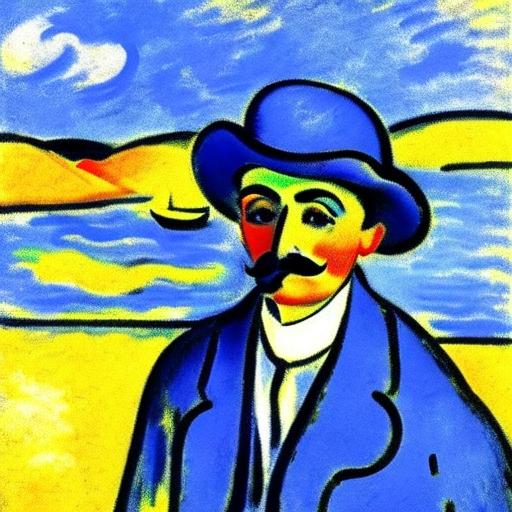}} &
        {\includegraphics[valign=c, width=\ww]{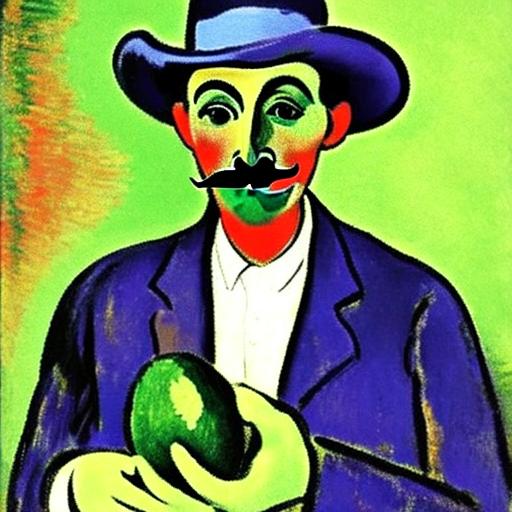}}
        \\

        \multicolumn{5}{c}{\textit{``a portrait of a
        man with a mustache and a hat, fauvism''}}
        \\
        \\

        {\includegraphics[valign=c, width=\ww]{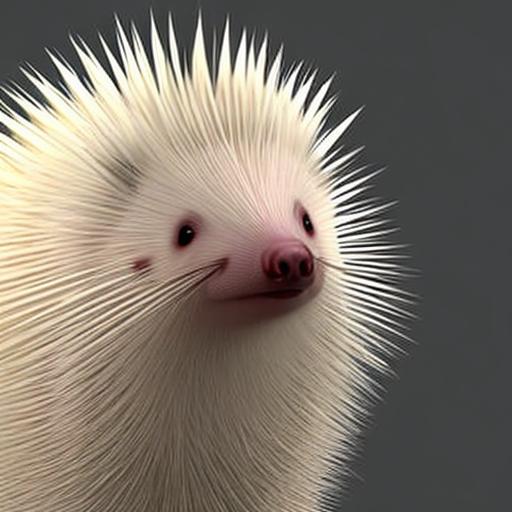}} &
        {\includegraphics[valign=c, width=\ww]{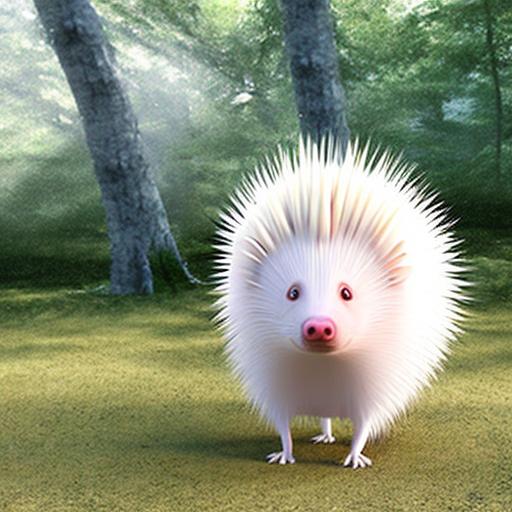}} &
        {\includegraphics[valign=c, width=\ww]{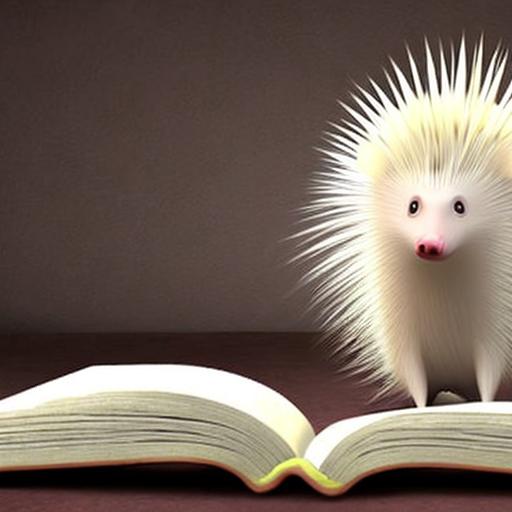}} &
        {\includegraphics[valign=c, width=\ww]{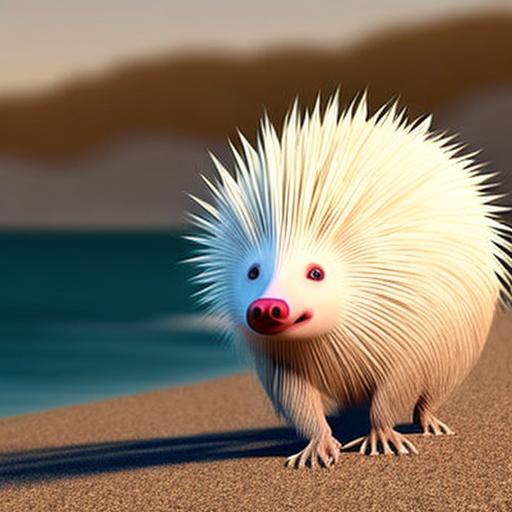}} &
        {\includegraphics[valign=c, width=\ww]{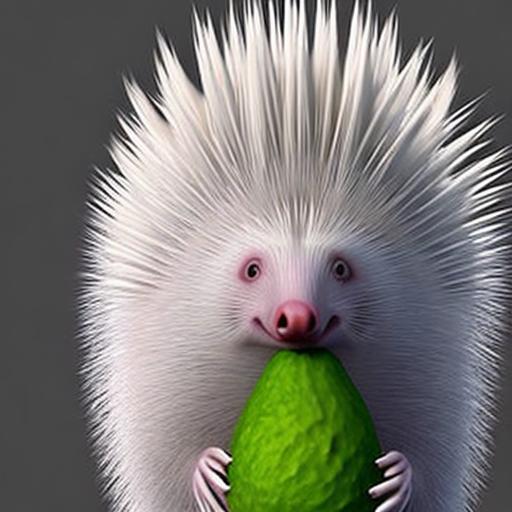}}
        \\

        \multicolumn{5}{c}{\textit{``a rendering of a cute albino porcupine, cozy indoor lighting''}}
        \\
        \\

    \end{tabular}
    
    \caption{\textbf{Our method using Stable Diffusion v2.1 backbone.} We experimented with a version of our method that uses the Stable Diffusion v2.1~\cite{Rombach2021HighResolutionIS} model. As can be seen, our method can extract a consistent character, however, as expected, the results are of a lower quality than when using the SDXL ~\cite{Podell2023SDXLIL} backbone that we use in the rest of this paper.}
    \label{fig:sd2_results}
\end{figure*}

%% file: sections/appendices/implmentation_details.tex
\section{Implementation Details}
\label{sec:implementation_details}

In this section, we provide the implementation details that were omitted from the main paper. In \Cref{sec:method_implementation_details} we provide the implementation details of our method and the baselines. Then, in \Cref{sec:automatic_metrics_implementation_details} we provide the implementation details of the automatic metrics that we used to evaluate our method against the baselines. In \Cref{sec:user_study_details} we provide the implementation details and the statistical analysis for the user study we conducted. Lastly, in \Cref{sec:applications_implementation_details} we provide the implementation details for the applications we presented.

\subsection{Method Implementation Details}
\label{sec:method_implementation_details}

We based our method, and all the baselines (except ELITE~\cite{Wei2023ELITEEV} and BLIP-diffusion~\cite{Li2023BLIPDiffusionPS}) on Stable Diffusion XL (SDXL)~\cite{Podell2023SDXLIL}, which is the state-of-the-art open source text-to-image model, at the writing of this paper. We used the official ELITE implementation, that uses Stable Diffusion V1.4, and the official implementation of BLIP-diffusion, that uses Stable Diffusion V1.5. We could not replace these two baselines to SDXL backbone, as the encoders were trained on these specific models. As for the rest of the baselines, we used the same SDXL architecture and weights.

For our method, we generated a set of $N = 128$ images at each iteration, which we found to be sufficient, empirically. We utilized the Adam optimizer \cite{Kingma2014AdamAM} with learning rate of 3e-5, $\beta_1 = 0.9$, $\beta_2 = 0.99$ and weight decay of 1e-2. In each identity extraction iteration of our method, we used 500 steps. We also found empirically that we can set the convergence criterion $\dconv$ adaptively to be 80\% of the average pairwise Euclidean distance between all $N$ initial image embeddings of the first iteration. In most cases, our method converges in 1--2 iterations, which takes about 13--26 minutes on A100 NVIDIA GPU when using bfloat16 mixed precision. In addition, we found that encouraging small clusters is beneficial by setting the minimum cluster size $\dminc$, and the target cluster size $\dtgtc$ to $\dminc = \dtgtc = 5$, which is the recommended image set size in the personalization setting \cite{Ruiz2022DreamBoothFT,Gal2022AnII}.

List of the third-party packages that we used:
\begin{itemize}
    \item Official SDXL~\cite{{Podell2023SDXLIL}} implementation by HuggingFace Diffusers~\cite{von-platen-etal-2022-diffusers} at \url{https://github.com/huggingface/diffusers}
    \item Official SDXL LoRA DB implementation by HuggingFace Diffusers~\cite{von-platen-etal-2022-diffusers} at \\ \url{https://github.com/huggingface/diffusers}.
    \item Official ELITE~\cite{Wei2023ELITEEV} implementation at \url{https://github.com/csyxwei/ELITE}
    \item Official BLIP-diffusion~\cite{Li2023BLIPDiffusionPS} implementation at \url{https://github.com/salesforce/LAVIS/tree/main/projects/blip-diffusion}
    \item Official IP-adapter~\cite{Ye2023IPAdapterTC} implementation at \url{https://github.com/tencent-ailab/IP-Adapter}
    \item DINOv2~\cite{Oquab2023DINOv2LR} ViT-g/14, DINOv1~\cite{Caron2021EmergingPI} ViT-B/16 and CLIP~\cite{Radford2021LearningTV} ViT-L/14 implementation by HuggingFace Transformers~\cite{wolf-etal-2020-transformers} at \url{https://github.com/huggingface/transformers}
\end{itemize}

\subsection{Automatic Metrics Implementation Details}
\label{sec:automatic_metrics_implementation_details}

In order to automatically evaluate our method and the baselines quantitatively, we instructed ChatGPT~\cite{chatgpt} to generate prompts for characters of different types (\eg, animals, creatures, objects, \etc) in different styles (\eg, stickers, animations, photorealistic images, \etc). These prompts were then used to generate a set of consistent characters by our method and by each of the baselines. Next, these prompts were used to generate these characters in a predefined collection of novel contexts from the following list:

\begin{itemize}
    \item ``a photo of [v] at the beach''
    \item ``a photo of [v] in the jungle''
    \item ``a photo of [v] in the snow''
    \item ``a photo of [v] in the street''
    \item ``a photo of [v] with a city in the background''
    \item ``a photo of [v] with a mountain in the background''
    \item ``a photo of [v] with the Eiffel Tower in the background''
    \item ``a photo of [v] near the Statue of Liberty''
    \item ``a photo of [v] near the Sydney Opera House''
    \item ``a photo of [v] floating on top of water''
    \item ``a photo of [v] eating a burger''
    \item ``a photo of [v] drinking a beer''
    \item ``a photo of [v] wearing a blue hat''
    \item ``a photo of [v] wearing sunglasses''
    \item ``a photo of [v] playing with a ball''
    \item ``a photo of [v] as a police officer''
\end{itemize}
where [v] is the newly-added token that represents the consistent character.

\subsection{User Study Details}
\label{sec:user_study_details}

\input{tables/user_study_variances.tex}

\input{tables/statistical_analysis.tex}

As explained in \Cref{sec:user_study},
we conducted a user study to evaluate our method, using the Amazon Mechanical Turk (AMT) platform \cite{amt}. We used the same generated prompts and samples that were used in \Cref{sec:comparisons},
and asked the evaluators to rate the prompt similarity and identity consistency of each result on a Likert scale of 1--5. For ranking the prompt similarity, the evaluators were instructed the following: ``For each of the following images, please rank on a scale of 1 to 5 its correspondence to this text description: \{PROMPT\}. The character in the image can be anything (e.g., a person, an animal, a toy etc.'' where \{PROMPT\} is the target text prompt (in which we replaced the special token with the word ``character''). All the baselines, as well as our method, were presented in the same page, and the evaluators were asked to rate each one of the results using a slider from 1 (``Do not match at all'') to 5 (``Match perfectly''). Next, to assess identity consistency, we took for each one of the characters two generated images that correspond to \emph{different} target text prompts, put them next to each other, and instructed the evaluators the following: ``For each of the following image pairs, please rank on a scale of 1 to 5 if they contain the same character (1 means that they contain totally different characters and 5 means that they contain exactly the same character). The images can have different backgrounds''. We put all the compared images on the same page, and the evaluators were asked to rate each one of the pairs using a slider from 1 (``Totally different characters'') to 5 (``Exactly the same character'').

We collected three ratings per question, resulting in 1104 ratings per task (prompt similarity and identity consistency). The time allotted per task was one hour, to allow the raters to properly evaluate the results without time pressure. The means and variances of the user study responses are reported in \Cref{tab:user_study_variances}.

In addition, we conducted a statistical analysis of our user study by validating that the difference between all the conditions is statistically significant using Kruskal-Wallis \cite{Kruskal1952UseOR} test ($p < \expnumber{1}{-28}$ for the text similarity test and $p < \expnumber{1}{-76}$ for the identity consistency text). Lastly, we used Tukey's honestly significant difference procedure \cite{Tukey1949ComparingIM} to show that the comparison of our method against all the baselines is statistically significant, as detailed in \Cref{tab:statistical_analysis}.

\subsection{Applications Implementation Details}
\label{sec:applications_implementation_details}
In \Cref{sec:applications},
we presented three downstream applications of our method.

\paragraph{Story illustration.} Given a long story, \eg, \textit{``This is a story about Jasper, a cute mink with a brown jacket and red pants. Jasper started his day by jogging on the beach, and afterwards, he enjoyed a coffee meetup with a friend in the heart of New York City. As the day drew to a close, he settled into his cozy apartment to review a paper''}, one can create a consistent character from the main character description (\textit{``a cute mink with a brown jacket and red pants''}), then they can generate the various scenes by simply rephrasing the sentence:
\begin{enumerate}
    \item \textit{``[v] jogging on the beach''}
    \item \textit{``[v] drinking coffee with his friend in the heart of New York City''}
    \item \textit{``[v] reviewing a paper in his cozy apartment''} 
\end{enumerate}

\paragraph{Local image editing.} Our method can be simply integrated with Blended Latent Diffusion~\cite{avrahami2023blendedlatent,blended_2022_CVPR} for editing images locally: given a text prompt, we start by running our method to extract a consistent identity, then, given an input image and mask, we can plant the character in the image within the mask boundaries. In addition, we can provide a local text description for the character.

\paragraph{Additional pose control.} Our method can be integrated with ControlNet~\cite{zhang2023controlnet}: given a text prompt, we first apply our method to extract a consistent identity $\crep = (\theta,\tau)$, where $\theta$ are the LoRA weights and $\tau$ is a set of custom text embeddings. Then, we can take an off-the-shelf pre-trained ControlNet model, plug-in our representation $\crep$, and use it to generate the consistent character in different poses given by the user.

%% file: tables/user_study_variances.tex
\begin{table}
    \centering
    \caption{\textbf{Users' rankings means and variances.} The means and variances of the rankings that are reported in the user study.}
    \begin{adjustbox}{width=1\columnwidth}
        \begin{tabular}{>{\columncolor[gray]{0.95}}lcc}
            \toprule
            
            \textbf{Method} & 
            Prompt similarity $(\uparrow)$ &
            Identity consistency $(\uparrow)$
            \\
            
            \midrule

            TI \cite{Gal2022AnII} &
            $3.31 \pm 1.43$ &
            $3.17 \pm 1.17$
            \\

            LoRA DB \cite{lora_diffusion} &
            $3.03 \pm 1.43$ &
            $3.67 \pm 1.20$
            \\

            ELITE \cite{Wei2023ELITEEV} &
            $2.87 \pm 1.46$ &
            $3.20 \pm 1.21$
            \\

            BLIP-Diffusion \cite{Li2023BLIPDiffusionPS} &
            $3.35 \pm 1.41$ &
            $2.76 \pm 1.31$
            \\

            IP-Adapter \cite{Ye2023IPAdapterTC} &
            $3.25 \pm 1.42$ &
            $2.99 \pm 1.28$
            \\

            Ours &
            $3.30 \pm 1.36$ &
            $3.48 \pm 1.20$
            \\
            
            \bottomrule
        \end{tabular}
    \end{adjustbox}
    \label{tab:user_study_variances}
\end{table}

%% file: tables/statistical_analysis.tex
\begin{table}
    \centering
    \caption{\textbf{Statistical analysis.} We use Tukey's  honestly significant difference procedure \cite{Tukey1949ComparingIM} to test whether the differences between mean scores in our user study are statistically significant.}
    \begin{adjustbox}{width=1\columnwidth}
        \begin{tabular}{>{\columncolor[gray]{0.95}}l>{\columncolor[gray]{0.95}}lcc}
            \toprule
            
            \textbf{Method 1} &
            \textbf{Method 2} &
            Prompt similarity &
            Identities similarity
            \\

            &
            &
            p-value &
            p-value
            \\
            \midrule

            TI \cite{Gal2022AnII} &
            Ours &
            $p < 0.001$ &
            $p < \expnumber{1}{-10}$
            \\

            LoRA DB \cite{lora_diffusion} &
            Ours &
            $p < \expnumber{1}{-13}$ &
            $\expnumber{1}{-4}$
            \\

            ELITE \cite{Wei2023ELITEEV} &
            Ours &
            $p < \expnumber{1}{-13}$ &
            $p < \expnumber{1}{-7}$
            \\

            BLIP-Diffusion \cite{Li2023BLIPDiffusionPS} &
            Ours &
            $p < 0.01$ &
            $p < \expnumber{1}{-13}$
            \\

            IP-Adapter \cite{Ye2023IPAdapterTC} &
            Ours &
            $p < \expnumber{1}{-5}$ &
            $p < \expnumber{1}{-13}$
            \\
            
            \bottomrule
        \end{tabular}
    \end{adjustbox}
    \label{tab:statistical_analysis}
\end{table}

%% file: sections/appendices/societal_impact.tex
\section{Societal Impact}
\label{sec:soceital_impact}

We believe that the emergence of technology that facilitates the effortless creation of consistent characters holds exciting promise in a variety of creative and practical applications. It can empower storytellers and content creators to bring their narratives to life with vivid and unique characters, enhancing the immersive quality of their work. In addition, it may offer accessibility to those who may not possess traditional artistic skills, democratizing character design in the creative industry. Furthermore, it can reduce the cost of advertising, and open up new opportunities for small and underprivileged entrepreneurs, enabling them to reach a wider audience and compete in the market more effectively.

On the other hand, as any other generative AI technology, it can be misused by creating false and misleading visual content for deceptive purposes. Creating fake characters or personas can be used for online scams, disinformation campaigns, \etc, making it challenging to discern genuine information from fabricated content. Such technologies underscore the vital importance of developing generated content detection systems, making it a compelling research direction to address. In addition, since our method uses a clustering algorithm, there exists a risk of automatically choosing a cluster with improper content, which may result in creating an improper consistent character.

%% file: main.bbl

\begin{thebibliography}{101}


\ifx \showCODEN    \undefined \def \showCODEN     #1{\unskip}     \fi
\ifx \showDOI      \undefined \def \showDOI       #1{#1}\fi
\ifx \showISBNx    \undefined \def \showISBNx     #1{\unskip}     \fi
\ifx \showISBNxiii \undefined \def \showISBNxiii  #1{\unskip}     \fi
\ifx \showISSN     \undefined \def \showISSN      #1{\unskip}     \fi
\ifx \showLCCN     \undefined \def \showLCCN      #1{\unskip}     \fi
\ifx \shownote     \undefined \def \shownote      #1{#1}          \fi
\ifx \showarticletitle \undefined \def \showarticletitle #1{#1}   \fi
\ifx \showURL      \undefined \def \showURL       {\relax}        \fi
\providecommand\bibfield[2]{#2}
\providecommand\bibinfo[2]{#2}
\providecommand\natexlab[1]{#1}
\providecommand\showeprint[2][]{arXiv:#2}

\bibitem[Ahn et~al\mbox{.}(2023)]%
        {Ahn2023DreamStylerPB}
\bibfield{author}{\bibinfo{person}{Namhyuk Ahn}, \bibinfo{person}{Junsoo Lee}, \bibinfo{person}{Chunggi Lee}, \bibinfo{person}{Kunhee Kim}, \bibinfo{person}{Daesik Kim}, \bibinfo{person}{Seung-Hun Nam}, {and} \bibinfo{person}{Kibeom Hong}.} \bibinfo{year}{2023}\natexlab{}.
\newblock \showarticletitle{DreamStyler: Paint by Style Inversion with Text-to-Image Diffusion Models}.
\newblock \bibinfo{journal}{\emph{ArXiv}}  \bibinfo{volume}{abs/2309.06933} (\bibinfo{year}{2023}).
\newblock
\urldef\tempurl%
\url{https://api.semanticscholar.org/CorpusID:261706081}
\showURL{%
\tempurl}


\bibitem[Alaluf et~al\mbox{.}(2023)]%
        {Alaluf2023ANS}
\bibfield{author}{\bibinfo{person}{Yuval Alaluf}, \bibinfo{person}{Elad Richardson}, \bibinfo{person}{Gal Metzer}, {and} \bibinfo{person}{Daniel Cohen-Or}.} \bibinfo{year}{2023}\natexlab{}.
\newblock \showarticletitle{A Neural Space-Time Representation for Text-to-Image Personalization}.
\newblock \bibinfo{journal}{\emph{ArXiv}}  \bibinfo{volume}{abs/2305.15391} (\bibinfo{year}{2023}).
\newblock
\urldef\tempurl%
\url{https://api.semanticscholar.org/CorpusID:258866047}
\showURL{%
\tempurl}


\bibitem[Amazon(2023)]%
        {amt}
\bibfield{author}{\bibinfo{person}{Amazon}.} \bibinfo{year}{2023}\natexlab{}.
\newblock \bibinfo{title}{Amazon Mechanical Turk}.
\newblock \bibinfo{howpublished}{\url{https://www.mturk.com/}}.
\newblock


\bibitem[Arar et~al\mbox{.}(2023)]%
        {arar2023domain}
\bibfield{author}{\bibinfo{person}{Moab Arar}, \bibinfo{person}{Rinon Gal}, \bibinfo{person}{Yuval Atzmon}, \bibinfo{person}{Gal Chechik}, \bibinfo{person}{Daniel Cohen-Or}, \bibinfo{person}{Ariel Shamir}, {and} \bibinfo{person}{Amit~H Bermano}.} \bibinfo{year}{2023}\natexlab{}.
\newblock \showarticletitle{Domain-agnostic tuning-encoder for fast personalization of text-to-image models}.
\newblock \bibinfo{journal}{\emph{arXiv preprint arXiv:2307.06925}} (\bibinfo{year}{2023}).
\newblock


\bibitem[Arthur and Vassilvitskii(2007)]%
        {Arthur2007kmeansTA}
\bibfield{author}{\bibinfo{person}{David Arthur} {and} \bibinfo{person}{Sergei Vassilvitskii}.} \bibinfo{year}{2007}\natexlab{}.
\newblock \showarticletitle{k-means++: the advantages of careful seeding}. In \bibinfo{booktitle}{\emph{ACM-SIAM Symposium on Discrete Algorithms}}.
\newblock
\urldef\tempurl%
\url{https://api.semanticscholar.org/CorpusID:1782131}
\showURL{%
\tempurl}


\bibitem[Avrahami et~al\mbox{.}(2023a)]%
        {Avrahami2023BreakASceneEM}
\bibfield{author}{\bibinfo{person}{Omri Avrahami}, \bibinfo{person}{Kfir Aberman}, \bibinfo{person}{Ohad Fried}, \bibinfo{person}{Daniel Cohen-Or}, {and} \bibinfo{person}{Dani Lischinski}.} \bibinfo{year}{2023}\natexlab{a}.
\newblock \showarticletitle{Break-A-Scene: Extracting Multiple Concepts from a Single Image}.
\newblock \bibinfo{journal}{\emph{ArXiv}}  \bibinfo{volume}{abs/2305.16311} (\bibinfo{year}{2023}).
\newblock
\urldef\tempurl%
\url{https://api.semanticscholar.org/CorpusID:258888228}
\showURL{%
\tempurl}


\bibitem[Avrahami et~al\mbox{.}(2023b)]%
        {avrahami2023blendedlatent}
\bibfield{author}{\bibinfo{person}{Omri Avrahami}, \bibinfo{person}{Ohad Fried}, {and} \bibinfo{person}{Dani Lischinski}.} \bibinfo{year}{2023}\natexlab{b}.
\newblock \showarticletitle{Blended Latent Diffusion}.
\newblock \bibinfo{journal}{\emph{ACM Trans. Graph.}} \bibinfo{volume}{42}, \bibinfo{number}{4}, Article \bibinfo{articleno}{149} (\bibinfo{date}{jul} \bibinfo{year}{2023}), \bibinfo{numpages}{11}~pages.
\newblock
\showISSN{0730-0301}
\urldef\tempurl%
\url{https://doi.org/10.1145/3592450}
\showDOI{\tempurl}


\bibitem[Avrahami et~al\mbox{.}(2023c)]%
        {spatext_2023_CVPR}
\bibfield{author}{\bibinfo{person}{Omri Avrahami}, \bibinfo{person}{Thomas Hayes}, \bibinfo{person}{Oran Gafni}, \bibinfo{person}{Sonal Gupta}, \bibinfo{person}{Yaniv Taigman}, \bibinfo{person}{Devi Parikh}, \bibinfo{person}{Dani Lischinski}, \bibinfo{person}{Ohad Fried}, {and} \bibinfo{person}{Xi Yin}.} \bibinfo{year}{2023}\natexlab{c}.
\newblock \showarticletitle{SpaText: Spatio-Textual Representation for Controllable Image Generation}. In \bibinfo{booktitle}{\emph{Proceedings of the IEEE/CVF Conference on Computer Vision and Pattern Recognition (CVPR)}}. \bibinfo{pages}{18370--18380}.
\newblock


\bibitem[Avrahami et~al\mbox{.}(2022)]%
        {blended_2022_CVPR}
\bibfield{author}{\bibinfo{person}{Omri Avrahami}, \bibinfo{person}{Dani Lischinski}, {and} \bibinfo{person}{Ohad Fried}.} \bibinfo{year}{2022}\natexlab{}.
\newblock \showarticletitle{Blended Diffusion for Text-Driven Editing of Natural Images}. In \bibinfo{booktitle}{\emph{Proceedings of the IEEE/CVF Conference on Computer Vision and Pattern Recognition (CVPR)}}. \bibinfo{pages}{18208--18218}.
\newblock


\bibitem[Balaji et~al\mbox{.}(2022)]%
        {Balaji2022eDiffITD}
\bibfield{author}{\bibinfo{person}{Yogesh Balaji}, \bibinfo{person}{Seungjun Nah}, \bibinfo{person}{Xun Huang}, \bibinfo{person}{Arash Vahdat}, \bibinfo{person}{Jiaming Song}, \bibinfo{person}{Qinsheng Zhang}, \bibinfo{person}{Karsten Kreis}, \bibinfo{person}{Miika Aittala}, \bibinfo{person}{Timo Aila}, \bibinfo{person}{Samuli Laine}, \bibinfo{person}{Bryan Catanzaro}, \bibinfo{person}{Tero Karras}, {and} \bibinfo{person}{Ming-Yu Liu}.} \bibinfo{year}{2022}\natexlab{}.
\newblock \showarticletitle{eDiff-I: Text-to-Image Diffusion Models with an Ensemble of Expert Denoisers}.
\newblock \bibinfo{journal}{\emph{ArXiv}}  \bibinfo{volume}{abs/2211.01324} (\bibinfo{year}{2022}).
\newblock
\urldef\tempurl%
\url{https://api.semanticscholar.org/CorpusID:253254800}
\showURL{%
\tempurl}


\bibitem[Bar-Tal et~al\mbox{.}(2022)]%
        {bar2022text2live}
\bibfield{author}{\bibinfo{person}{Omer Bar-Tal}, \bibinfo{person}{Dolev Ofri-Amar}, \bibinfo{person}{Rafail Fridman}, \bibinfo{person}{Yoni Kasten}, {and} \bibinfo{person}{Tali Dekel}.} \bibinfo{year}{2022}\natexlab{}.
\newblock \showarticletitle{Text2live: Text-driven layered image and video editing}. In \bibinfo{booktitle}{\emph{European conference on computer vision}}. Springer, \bibinfo{pages}{707--723}.
\newblock


\bibitem[Benaim et~al\mbox{.}(2022)]%
        {Benaim2022VolumetricDF}
\bibfield{author}{\bibinfo{person}{Sagie Benaim}, \bibinfo{person}{Frederik Warburg}, \bibinfo{person}{Peter~Ebert Christensen}, {and} \bibinfo{person}{Serge~J. Belongie}.} \bibinfo{year}{2022}\natexlab{}.
\newblock \showarticletitle{Volumetric Disentanglement for 3D Scene Manipulation}.
\newblock \bibinfo{journal}{\emph{ArXiv}}  \bibinfo{volume}{abs/2206.02776} (\bibinfo{year}{2022}).
\newblock
\urldef\tempurl%
\url{https://api.semanticscholar.org/CorpusID:249394623}
\showURL{%
\tempurl}


\bibitem[Betker et~al\mbox{.}(2023)]%
        {BetkerImprovingIG}
\bibfield{author}{\bibinfo{person}{James Betker}, \bibinfo{person}{Gabriel Goh}, \bibinfo{person}{Li Jing}, \bibinfo{person}{Tim Brooks}, \bibinfo{person}{Jianfeng Wang}, \bibinfo{person}{Linjie Li}, \bibinfo{person}{Long Ouyang}, \bibinfo{person}{Juntang Zhuang}, \bibinfo{person}{Joyce Lee}, \bibinfo{person}{Yufei Guo}, {et~al\mbox{.}}} \bibinfo{year}{2023}\natexlab{}.
\newblock \showarticletitle{Improving image generation with better captions}.
\newblock


\bibitem[Cao et~al\mbox{.}(2023)]%
        {cao2023masactrl}
\bibfield{author}{\bibinfo{person}{Mingdeng Cao}, \bibinfo{person}{Xintao Wang}, \bibinfo{person}{Zhongang Qi}, \bibinfo{person}{Ying Shan}, \bibinfo{person}{Xiaohu Qie}, {and} \bibinfo{person}{Yinqiang Zheng}.} \bibinfo{year}{2023}\natexlab{}.
\newblock \showarticletitle{{MasaCtrl:} Tuning-Free Mutual Self-Attention Control for Consistent Image Synthesis and Editing}. In \bibinfo{booktitle}{\emph{Proceedings of the IEEE/CVF International Conference on Computer Vision (ICCV)}}. \bibinfo{pages}{22560--22570}.
\newblock


\bibitem[Caron et~al\mbox{.}(2021)]%
        {Caron2021EmergingPI}
\bibfield{author}{\bibinfo{person}{Mathilde Caron}, \bibinfo{person}{Hugo Touvron}, \bibinfo{person}{Ishan Misra}, \bibinfo{person}{{Herv\'{e}} Jegou}, \bibinfo{person}{Julien Mairal}, \bibinfo{person}{Piotr Bojanowski}, {and} \bibinfo{person}{Armand Joulin}.} \bibinfo{year}{2021}\natexlab{}.
\newblock \showarticletitle{Emerging Properties in Self-Supervised Vision Transformers}. In \bibinfo{booktitle}{\emph{2021 IEEE/CVF International Conference on Computer Vision (ICCV)}}. \bibinfo{pages}{9630--9640}.
\newblock


\bibitem[Chefer et~al\mbox{.}(2023)]%
        {Chefer2023AttendandExciteAS}
\bibfield{author}{\bibinfo{person}{Hila Chefer}, \bibinfo{person}{Yuval Alaluf}, \bibinfo{person}{Yael Vinker}, \bibinfo{person}{Lior Wolf}, {and} \bibinfo{person}{Daniel Cohen-Or}.} \bibinfo{year}{2023}\natexlab{}.
\newblock \showarticletitle{Attend-and-Excite: Attention-Based Semantic Guidance for Text-to-Image Diffusion Models}.
\newblock \bibinfo{journal}{\emph{ACM Transactions on Graphics (TOG)}}  \bibinfo{volume}{42} (\bibinfo{year}{2023}), \bibinfo{pages}{1 -- 10}.
\newblock
\urldef\tempurl%
\url{https://api.semanticscholar.org/CorpusID:256416326}
\showURL{%
\tempurl}


\bibitem[Chen et~al\mbox{.}(2023a)]%
        {Chen2023SubjectdrivenTG}
\bibfield{author}{\bibinfo{person}{Wenhu Chen}, \bibinfo{person}{Hexiang Hu}, \bibinfo{person}{Yandong Li}, \bibinfo{person}{Nataniel Rui}, \bibinfo{person}{Xuhui Jia}, \bibinfo{person}{Ming-Wei Chang}, {and} \bibinfo{person}{William~W. Cohen}.} \bibinfo{year}{2023}\natexlab{a}.
\newblock \showarticletitle{Subject-driven Text-to-Image Generation via Apprenticeship Learning}.
\newblock \bibinfo{journal}{\emph{ArXiv}}  \bibinfo{volume}{abs/2304.00186} (\bibinfo{year}{2023}).
\newblock


\bibitem[Chen et~al\mbox{.}(2023b)]%
        {Chen2023AnyDoorZO}
\bibfield{author}{\bibinfo{person}{Xi Chen}, \bibinfo{person}{Lianghua Huang}, \bibinfo{person}{Yu Liu}, \bibinfo{person}{Yujun Shen}, \bibinfo{person}{Deli Zhao}, {and} \bibinfo{person}{Hengshuang Zhao}.} \bibinfo{year}{2023}\natexlab{b}.
\newblock \showarticletitle{AnyDoor: Zero-shot Object-level Image Customization}.
\newblock \bibinfo{journal}{\emph{ArXiv}}  \bibinfo{volume}{abs/2307.09481} (\bibinfo{year}{2023}).
\newblock
\urldef\tempurl%
\url{https://api.semanticscholar.org/CorpusID:259951373}
\showURL{%
\tempurl}


\bibitem[Couairon et~al\mbox{.}(2023)]%
        {Couairon2023ZeroshotSL}
\bibfield{author}{\bibinfo{person}{Guillaume Couairon}, \bibinfo{person}{Marlene Careil}, \bibinfo{person}{Matthieu Cord}, \bibinfo{person}{St{\'e}phane Lathuili{\`e}re}, {and} \bibinfo{person}{Jakob Verbeek}.} \bibinfo{year}{2023}\natexlab{}.
\newblock \showarticletitle{Zero-shot spatial layout conditioning for text-to-image diffusion models}.
\newblock \bibinfo{journal}{\emph{ArXiv}}  \bibinfo{volume}{abs/2306.13754} (\bibinfo{year}{2023}).
\newblock
\urldef\tempurl%
\url{https://api.semanticscholar.org/CorpusID:259252153}
\showURL{%
\tempurl}


\bibitem[Foundations(2023)]%
        {design_sheet_trick}
\bibfield{author}{\bibinfo{person}{AI Foundations}.} \bibinfo{year}{2023}\natexlab{}.
\newblock \bibinfo{title}{How to Create Consistent Characters in Midjourney}.
\newblock \bibinfo{howpublished}{\url{https://www.youtube.com/watch?v=Z7_ta3RHijQ}}.
\newblock


\bibitem[Fridman et~al\mbox{.}(2023)]%
        {Fridman2023SceneScapeTC}
\bibfield{author}{\bibinfo{person}{Rafail Fridman}, \bibinfo{person}{Amit Abecasis}, \bibinfo{person}{Yoni Kasten}, {and} \bibinfo{person}{Tali Dekel}.} \bibinfo{year}{2023}\natexlab{}.
\newblock \showarticletitle{SceneScape: Text-Driven Consistent Scene Generation}.
\newblock \bibinfo{journal}{\emph{ArXiv}}  \bibinfo{volume}{abs/2302.01133} (\bibinfo{year}{2023}).
\newblock
\urldef\tempurl%
\url{https://api.semanticscholar.org/CorpusID:256503775}
\showURL{%
\tempurl}


\bibitem[Gal et~al\mbox{.}(2022)]%
        {Gal2022AnII}
\bibfield{author}{\bibinfo{person}{Rinon Gal}, \bibinfo{person}{Yuval Alaluf}, \bibinfo{person}{Yuval Atzmon}, \bibinfo{person}{Or Patashnik}, \bibinfo{person}{Amit~Haim Bermano}, \bibinfo{person}{Gal Chechik}, {and} \bibinfo{person}{Daniel Cohen-or}.} \bibinfo{year}{2022}\natexlab{}.
\newblock \showarticletitle{An Image is Worth One Word: Personalizing Text-to-Image Generation using Textual Inversion}. In \bibinfo{booktitle}{\emph{The Eleventh International Conference on Learning Representations}}.
\newblock


\bibitem[Gal et~al\mbox{.}(2023)]%
        {gal2023encoder}
\bibfield{author}{\bibinfo{person}{Rinon Gal}, \bibinfo{person}{Moab Arar}, \bibinfo{person}{Yuval Atzmon}, \bibinfo{person}{Amit~H Bermano}, \bibinfo{person}{Gal Chechik}, {and} \bibinfo{person}{Daniel Cohen-Or}.} \bibinfo{year}{2023}\natexlab{}.
\newblock \showarticletitle{Encoder-based domain tuning for fast personalization of text-to-image models}.
\newblock \bibinfo{journal}{\emph{ACM Transactions on Graphics (TOG)}} \bibinfo{volume}{42}, \bibinfo{number}{4} (\bibinfo{year}{2023}), \bibinfo{pages}{1--13}.
\newblock


\bibitem[Ge et~al\mbox{.}(2023)]%
        {Ge2023ExpressiveTG}
\bibfield{author}{\bibinfo{person}{Songwei Ge}, \bibinfo{person}{Taesung Park}, \bibinfo{person}{Jun-Yan Zhu}, {and} \bibinfo{person}{Jia-Bin Huang}.} \bibinfo{year}{2023}\natexlab{}.
\newblock \showarticletitle{Expressive Text-to-Image Generation with Rich Text}.
\newblock \bibinfo{journal}{\emph{ArXiv}}  \bibinfo{volume}{abs/2304.06720} (\bibinfo{year}{2023}).
\newblock
\urldef\tempurl%
\url{https://api.semanticscholar.org/CorpusID:258108187}
\showURL{%
\tempurl}


\bibitem[Geyer et~al\mbox{.}(2023)]%
        {geyer2023tokenflow}
\bibfield{author}{\bibinfo{person}{Michal Geyer}, \bibinfo{person}{Omer Bar-Tal}, \bibinfo{person}{Shai Bagon}, {and} \bibinfo{person}{Tali Dekel}.} \bibinfo{year}{2023}\natexlab{}.
\newblock \showarticletitle{Tokenflow: Consistent diffusion features for consistent video editing}.
\newblock \bibinfo{journal}{\emph{arXiv preprint arXiv:2307.10373}} (\bibinfo{year}{2023}).
\newblock


\bibitem[Gong et~al\mbox{.}(2023)]%
        {Gong2023TaleCrafterIS}
\bibfield{author}{\bibinfo{person}{Yuan Gong}, \bibinfo{person}{Youxin Pang}, \bibinfo{person}{Xiaodong Cun}, \bibinfo{person}{Menghan Xia}, \bibinfo{person}{Haoxin Chen}, \bibinfo{person}{Longyue Wang}, \bibinfo{person}{Yong Zhang}, \bibinfo{person}{Xintao Wang}, \bibinfo{person}{Ying Shan}, {and} \bibinfo{person}{Yujiu Yang}.} \bibinfo{year}{2023}\natexlab{}.
\newblock \showarticletitle{{TaleCrafter:} Interactive Story Visualization with Multiple Characters}.
\newblock \bibinfo{journal}{\emph{ArXiv}}  \bibinfo{volume}{abs/2305.18247} (\bibinfo{year}{2023}).
\newblock
\urldef\tempurl%
\url{https://api.semanticscholar.org/CorpusID:258960665}
\showURL{%
\tempurl}


\bibitem[Gordon et~al\mbox{.}(2023)]%
        {Gordon2023BlendedNeRFZO}
\bibfield{author}{\bibinfo{person}{Ori Gordon}, \bibinfo{person}{Omri Avrahami}, {and} \bibinfo{person}{Dani Lischinski}.} \bibinfo{year}{2023}\natexlab{}.
\newblock \showarticletitle{Blended-NeRF: Zero-Shot Object Generation and Blending in Existing Neural Radiance Fields}.
\newblock \bibinfo{journal}{\emph{ArXiv}}  \bibinfo{volume}{abs/2306.12760} (\bibinfo{year}{2023}).
\newblock
\urldef\tempurl%
\url{https://api.semanticscholar.org/CorpusID:259224726}
\showURL{%
\tempurl}


\bibitem[Han et~al\mbox{.}(2023)]%
        {Han2023SVDiffCP}
\bibfield{author}{\bibinfo{person}{Ligong Han}, \bibinfo{person}{Yinxiao Li}, \bibinfo{person}{Han Zhang}, \bibinfo{person}{Peyman Milanfar}, \bibinfo{person}{Dimitris~N. Metaxas}, {and} \bibinfo{person}{Feng Yang}.} \bibinfo{year}{2023}\natexlab{}.
\newblock \showarticletitle{SVDiff: Compact Parameter Space for Diffusion Fine-Tuning}.
\newblock \bibinfo{journal}{\emph{ArXiv}}  \bibinfo{volume}{abs/2303.11305} (\bibinfo{year}{2023}).
\newblock


\bibitem[Hertz et~al\mbox{.}(2023)]%
        {hertz2023delta}
\bibfield{author}{\bibinfo{person}{Amir Hertz}, \bibinfo{person}{Kfir Aberman}, {and} \bibinfo{person}{Daniel Cohen-Or}.} \bibinfo{year}{2023}\natexlab{}.
\newblock \showarticletitle{Delta denoising score}. In \bibinfo{booktitle}{\emph{Proceedings of the IEEE/CVF International Conference on Computer Vision}}. \bibinfo{pages}{2328--2337}.
\newblock


\bibitem[Hertz et~al\mbox{.}(2022)]%
        {hertz2022prompt}
\bibfield{author}{\bibinfo{person}{Amir Hertz}, \bibinfo{person}{Ron Mokady}, \bibinfo{person}{Jay Tenenbaum}, \bibinfo{person}{Kfir Aberman}, \bibinfo{person}{Yael Pritch}, {and} \bibinfo{person}{Daniel Cohen-Or}.} \bibinfo{year}{2022}\natexlab{}.
\newblock \showarticletitle{Prompt-to-prompt image editing with cross attention control}.
\newblock \bibinfo{journal}{\emph{arXiv preprint arXiv:2208.01626}} (\bibinfo{year}{2022}).
\newblock


\bibitem[Hinton and Roweis(2002)]%
        {Hinton2002StochasticNE}
\bibfield{author}{\bibinfo{person}{Geoffrey~E. Hinton} {and} \bibinfo{person}{Sam~T. Roweis}.} \bibinfo{year}{2002}\natexlab{}.
\newblock \showarticletitle{Stochastic Neighbor Embedding}. In \bibinfo{booktitle}{\emph{NIPS}}.
\newblock
\urldef\tempurl%
\url{https://api.semanticscholar.org/CorpusID:20240}
\showURL{%
\tempurl}


\bibitem[Ho et~al\mbox{.}(2020)]%
        {ho2020denoising}
\bibfield{author}{\bibinfo{person}{Jonathan Ho}, \bibinfo{person}{Ajay Jain}, {and} \bibinfo{person}{Pieter Abbeel}.} \bibinfo{year}{2020}\natexlab{}.
\newblock \showarticletitle{Denoising Diffusion Probabilistic Models}. In \bibinfo{booktitle}{\emph{Proc.~NeurIPS}}.
\newblock


\bibitem[H{\"o}llein et~al\mbox{.}(2023)]%
        {Hllein2023Text2RoomET}
\bibfield{author}{\bibinfo{person}{Lukas H{\"o}llein}, \bibinfo{person}{Ang Cao}, \bibinfo{person}{Andrew Owens}, \bibinfo{person}{Justin Johnson}, {and} \bibinfo{person}{Matthias Nie{\ss}ner}.} \bibinfo{year}{2023}\natexlab{}.
\newblock \showarticletitle{Text2Room: Extracting Textured 3D Meshes from 2D Text-to-Image Models}.
\newblock \bibinfo{journal}{\emph{ArXiv}}  \bibinfo{volume}{abs/2303.11989} (\bibinfo{year}{2023}).
\newblock
\urldef\tempurl%
\url{https://api.semanticscholar.org/CorpusID:257636653}
\showURL{%
\tempurl}


\bibitem[Horwitz and Hoshen(2022)]%
        {Horwitz2022ConffusionCI}
\bibfield{author}{\bibinfo{person}{Eliahu Horwitz} {and} \bibinfo{person}{Yedid Hoshen}.} \bibinfo{year}{2022}\natexlab{}.
\newblock \showarticletitle{Conffusion: Confidence Intervals for Diffusion Models}.
\newblock \bibinfo{journal}{\emph{ArXiv}}  \bibinfo{volume}{abs/2211.09795} (\bibinfo{year}{2022}).
\newblock


\bibitem[Hu et~al\mbox{.}(2021)]%
        {lora}
\bibfield{author}{\bibinfo{person}{Edward~J Hu}, \bibinfo{person}{Phillip Wallis}, \bibinfo{person}{Zeyuan Allen-Zhu}, \bibinfo{person}{Yuanzhi Li}, \bibinfo{person}{Shean Wang}, \bibinfo{person}{Lu Wang}, \bibinfo{person}{Weizhu Chen}, {et~al\mbox{.}}} \bibinfo{year}{2021}\natexlab{}.
\newblock \showarticletitle{{LoRA}: Low-Rank Adaptation of Large Language Models}. In \bibinfo{booktitle}{\emph{International Conference on Learning Representations}}.
\newblock


\bibitem[Ilharco et~al\mbox{.}(2021)]%
        {OpenCLIP}
\bibfield{author}{\bibinfo{person}{Gabriel Ilharco}, \bibinfo{person}{Mitchell Wortsman}, \bibinfo{person}{Ross Wightman}, \bibinfo{person}{Cade Gordon}, \bibinfo{person}{Nicholas Carlini}, \bibinfo{person}{Rohan Taori}, \bibinfo{person}{Achal Dave}, \bibinfo{person}{Vaishaal Shankar}, \bibinfo{person}{Hongseok Namkoong}, \bibinfo{person}{John Miller}, \bibinfo{person}{Hannaneh Hajishirzi}, \bibinfo{person}{Ali Farhadi}, {and} \bibinfo{person}{Ludwig Schmidt}.} \bibinfo{year}{2021}\natexlab{}.
\newblock \bibinfo{title}{{OpenCLIP}}.
\newblock
\newblock
\urldef\tempurl%
\url{https://doi.org/10.5281/zenodo.5143773}
\showDOI{\tempurl}


\bibitem[Iluz et~al\mbox{.}(2023)]%
        {Iluz2023WordAsImageFS}
\bibfield{author}{\bibinfo{person}{Shira Iluz}, \bibinfo{person}{Yael Vinker}, \bibinfo{person}{Amir Hertz}, \bibinfo{person}{Daniel Berio}, \bibinfo{person}{Daniel Cohen-Or}, {and} \bibinfo{person}{Ariel Shamir}.} \bibinfo{year}{2023}\natexlab{}.
\newblock \showarticletitle{Word-As-Image for Semantic Typography}.
\newblock \bibinfo{journal}{\emph{ACM Transactions on Graphics (TOG)}}  \bibinfo{volume}{42} (\bibinfo{year}{2023}), \bibinfo{pages}{1 -- 11}.
\newblock
\urldef\tempurl%
\url{https://api.semanticscholar.org/CorpusID:257353586}
\showURL{%
\tempurl}


\bibitem[Jeong et~al\mbox{.}(2023)]%
        {Jeong2023ZeroshotGO}
\bibfield{author}{\bibinfo{person}{Hyeonho Jeong}, \bibinfo{person}{Gihyun Kwon}, {and} \bibinfo{person}{Jong-Chul Ye}.} \bibinfo{year}{2023}\natexlab{}.
\newblock \showarticletitle{Zero-shot Generation of Coherent Storybook from Plain Text Story using Diffusion Models}.
\newblock \bibinfo{journal}{\emph{ArXiv}}  \bibinfo{volume}{abs/2302.03900} (\bibinfo{year}{2023}).
\newblock
\urldef\tempurl%
\url{https://api.semanticscholar.org/CorpusID:256662241}
\showURL{%
\tempurl}


\bibitem[Jia et~al\mbox{.}(2023)]%
        {Jia2023TamingEF}
\bibfield{author}{\bibinfo{person}{Xuhui Jia}, \bibinfo{person}{Yang Zhao}, \bibinfo{person}{Kelvin C.~K. Chan}, \bibinfo{person}{Yandong Li}, \bibinfo{person}{Han-Ying Zhang}, \bibinfo{person}{Boqing Gong}, \bibinfo{person}{Tingbo Hou}, \bibinfo{person}{H. Wang}, {and} \bibinfo{person}{Yu-Chuan Su}.} \bibinfo{year}{2023}\natexlab{}.
\newblock \showarticletitle{Taming Encoder for Zero Fine-tuning Image Customization with Text-to-Image Diffusion Models}.
\newblock \bibinfo{journal}{\emph{ArXiv}}  \bibinfo{volume}{abs/2304.02642} (\bibinfo{year}{2023}).
\newblock


\bibitem[JoshGreat(2023)]%
        {consistent_generation_tricks}
\bibfield{author}{\bibinfo{person}{JoshGreat}.} \bibinfo{year}{2023}\natexlab{}.
\newblock \bibinfo{title}{8 ways to generate consistent characters (for comics, storyboards, books etc) : StableDiffusion}.
\newblock \bibinfo{howpublished}{\url{https://www.reddit.com/r/StableDiffusion/comments/10yxz3m/8_ways_to_generate_consistent_characters_for/}}.
\newblock


\bibitem[Kawar et~al\mbox{.}(2023)]%
        {Kawar2022ImagicTR}
\bibfield{author}{\bibinfo{person}{Bahjat Kawar}, \bibinfo{person}{Shiran Zada}, \bibinfo{person}{Oran Lang}, \bibinfo{person}{Omer Tov}, \bibinfo{person}{Huiwen Chang}, \bibinfo{person}{Tali Dekel}, \bibinfo{person}{Inbar Mosseri}, {and} \bibinfo{person}{Michal Irani}.} \bibinfo{year}{2023}\natexlab{}.
\newblock \showarticletitle{Imagic: Text-based real image editing with diffusion models}. In \bibinfo{booktitle}{\emph{Proceedings of the IEEE/CVF Conference on Computer Vision and Pattern Recognition}}. \bibinfo{pages}{6007--6017}.
\newblock


\bibitem[Kingma and Ba(2014)]%
        {Kingma2014AdamAM}
\bibfield{author}{\bibinfo{person}{Diederik~P. Kingma} {and} \bibinfo{person}{Jimmy Ba}.} \bibinfo{year}{2014}\natexlab{}.
\newblock \showarticletitle{Adam: A Method for Stochastic Optimization}.
\newblock \bibinfo{journal}{\emph{CoRR}}  \bibinfo{volume}{abs/1412.6980} (\bibinfo{year}{2014}).
\newblock


\bibitem[Kruskal and Wallis(1952)]%
        {Kruskal1952UseOR}
\bibfield{author}{\bibinfo{person}{William~H. Kruskal} {and} \bibinfo{person}{Wilson~Allen Wallis}.} \bibinfo{year}{1952}\natexlab{}.
\newblock \showarticletitle{Use of Ranks in One-Criterion Variance Analysis}.
\newblock \bibinfo{journal}{\emph{J. Amer. Statist. Assoc.}}  \bibinfo{volume}{47} (\bibinfo{year}{1952}), \bibinfo{pages}{583--621}.
\newblock


\bibitem[Kumari et~al\mbox{.}(2023)]%
        {Kumari2022MultiConceptCO}
\bibfield{author}{\bibinfo{person}{Nupur Kumari}, \bibinfo{person}{Bingliang Zhang}, \bibinfo{person}{Richard Zhang}, \bibinfo{person}{Eli Shechtman}, {and} \bibinfo{person}{Jun-Yan Zhu}.} \bibinfo{year}{2023}\natexlab{}.
\newblock \showarticletitle{Multi-concept customization of text-to-image diffusion}. In \bibinfo{booktitle}{\emph{Proceedings of the IEEE/CVF Conference on Computer Vision and Pattern Recognition}}. \bibinfo{pages}{1931--1941}.
\newblock


\bibitem[Li et~al\mbox{.}(2023)]%
        {Li2023BLIPDiffusionPS}
\bibfield{author}{\bibinfo{person}{Dongxu Li}, \bibinfo{person}{Junnan Li}, {and} \bibinfo{person}{Steven C.~H. Hoi}.} \bibinfo{year}{2023}\natexlab{}.
\newblock \showarticletitle{{BLIP-Diffusion}: Pre-trained Subject Representation for Controllable Text-to-Image Generation and Editing}.
\newblock \bibinfo{journal}{\emph{ArXiv}}  \bibinfo{volume}{abs/2305.14720} (\bibinfo{year}{2023}).
\newblock
\urldef\tempurl%
\url{https://api.semanticscholar.org/CorpusID:258865473}
\showURL{%
\tempurl}


\bibitem[Li et~al\mbox{.}(2019)]%
        {li2018storygan}
\bibfield{author}{\bibinfo{person}{Yitong Li}, \bibinfo{person}{Zhe Gan}, \bibinfo{person}{Yelong Shen}, \bibinfo{person}{Jingjing Liu}, \bibinfo{person}{Yu Cheng}, \bibinfo{person}{Yuexin Wu}, \bibinfo{person}{Lawrence Carin}, \bibinfo{person}{David Carlson}, {and} \bibinfo{person}{Jianfeng Gao}.} \bibinfo{year}{2019}\natexlab{}.
\newblock \showarticletitle{StoryGAN: A Sequential Conditional GAN for Story Visualization}.
\newblock \bibinfo{journal}{\emph{CVPR}} (\bibinfo{year}{2019}).
\newblock


\bibitem[Liu et~al\mbox{.}(2023a)]%
        {liu2023video}
\bibfield{author}{\bibinfo{person}{Shaoteng Liu}, \bibinfo{person}{Yuechen Zhang}, \bibinfo{person}{Wenbo Li}, \bibinfo{person}{Zhe Lin}, {and} \bibinfo{person}{Jiaya Jia}.} \bibinfo{year}{2023}\natexlab{a}.
\newblock \showarticletitle{Video-p2p: Video editing with cross-attention control}.
\newblock \bibinfo{journal}{\emph{arXiv preprint arXiv:2303.04761}} (\bibinfo{year}{2023}).
\newblock


\bibitem[Liu et~al\mbox{.}(2023b)]%
        {Liu2023VideoP2PVE}
\bibfield{author}{\bibinfo{person}{Shaoteng Liu}, \bibinfo{person}{Yuecheng Zhang}, \bibinfo{person}{Wenbo Li}, \bibinfo{person}{Zhe Lin}, {and} \bibinfo{person}{Jiaya Jia}.} \bibinfo{year}{2023}\natexlab{b}.
\newblock \showarticletitle{Video-P2P: Video Editing with Cross-attention Control}.
\newblock \bibinfo{journal}{\emph{ArXiv}}  \bibinfo{volume}{abs/2303.04761} (\bibinfo{year}{2023}).
\newblock
\urldef\tempurl%
\url{https://api.semanticscholar.org/CorpusID:257405406}
\showURL{%
\tempurl}


\bibitem[Maharana et~al\mbox{.}(2022)]%
        {maharana2022storydall}
\bibfield{author}{\bibinfo{person}{Adyasha Maharana}, \bibinfo{person}{Darryl Hannan}, {and} \bibinfo{person}{Mohit Bansal}.} \bibinfo{year}{2022}\natexlab{}.
\newblock \showarticletitle{Storydall-e: Adapting pretrained text-to-image transformers for story continuation}. In \bibinfo{booktitle}{\emph{European Conference on Computer Vision}}. Springer, \bibinfo{pages}{70--87}.
\newblock


\bibitem[Meng et~al\mbox{.}(2021)]%
        {meng2021sdedit}
\bibfield{author}{\bibinfo{person}{Chenlin Meng}, \bibinfo{person}{Yutong He}, \bibinfo{person}{Yang Song}, \bibinfo{person}{Jiaming Song}, \bibinfo{person}{Jiajun Wu}, \bibinfo{person}{Jun-Yan Zhu}, {and} \bibinfo{person}{Stefano Ermon}.} \bibinfo{year}{2021}\natexlab{}.
\newblock \showarticletitle{SDEdit: Guided Image Synthesis and Editing with Stochastic Differential Equations}. In \bibinfo{booktitle}{\emph{International Conference on Learning Representations}}.
\newblock


\bibitem[Metzer et~al\mbox{.}(2023)]%
        {metzer2023latent}
\bibfield{author}{\bibinfo{person}{Gal Metzer}, \bibinfo{person}{Elad Richardson}, \bibinfo{person}{Or Patashnik}, \bibinfo{person}{Raja Giryes}, {and} \bibinfo{person}{Daniel Cohen-Or}.} \bibinfo{year}{2023}\natexlab{}.
\newblock \showarticletitle{Latent-nerf for shape-guided generation of 3d shapes and textures}. In \bibinfo{booktitle}{\emph{Proceedings of the IEEE/CVF Conference on Computer Vision and Pattern Recognition}}. \bibinfo{pages}{12663--12673}.
\newblock


\bibitem[Mokady et~al\mbox{.}(2023)]%
        {mokady2022null}
\bibfield{author}{\bibinfo{person}{Ron Mokady}, \bibinfo{person}{Amir Hertz}, \bibinfo{person}{Kfir Aberman}, \bibinfo{person}{Yael Pritch}, {and} \bibinfo{person}{Daniel Cohen-Or}.} \bibinfo{year}{2023}\natexlab{}.
\newblock \showarticletitle{Null-text inversion for editing real images using guided diffusion models}. In \bibinfo{booktitle}{\emph{Proceedings of the IEEE/CVF Conference on Computer Vision and Pattern Recognition}}. \bibinfo{pages}{6038--6047}.
\newblock


\bibitem[Molad et~al\mbox{.}(2023)]%
        {Molad2023DreamixVD}
\bibfield{author}{\bibinfo{person}{Eyal Molad}, \bibinfo{person}{Eliahu Horwitz}, \bibinfo{person}{Dani Valevski}, \bibinfo{person}{Alex~Rav Acha}, \bibinfo{person}{Y. Matias}, \bibinfo{person}{Yael Pritch}, \bibinfo{person}{Yaniv Leviathan}, {and} \bibinfo{person}{Yedid Hoshen}.} \bibinfo{year}{2023}\natexlab{}.
\newblock \showarticletitle{Dreamix: Video Diffusion Models are General Video Editors}.
\newblock \bibinfo{journal}{\emph{ArXiv}}  \bibinfo{volume}{abs/2302.01329} (\bibinfo{year}{2023}).
\newblock


\bibitem[Mou et~al\mbox{.}(2023)]%
        {mou2023t2i}
\bibfield{author}{\bibinfo{person}{Chong Mou}, \bibinfo{person}{Xintao Wang}, \bibinfo{person}{Liangbin Xie}, \bibinfo{person}{Yanze Wu}, \bibinfo{person}{Jian Zhang}, \bibinfo{person}{Zhongang Qi}, \bibinfo{person}{Ying Shan}, {and} \bibinfo{person}{Xiaohu Qie}.} \bibinfo{year}{2023}\natexlab{}.
\newblock \showarticletitle{T2i-adapter: Learning adapters to dig out more controllable ability for text-to-image diffusion models}.
\newblock \bibinfo{journal}{\emph{arXiv preprint arXiv:2302.08453}} (\bibinfo{year}{2023}).
\newblock


\bibitem[Nichol et~al\mbox{.}(2021)]%
        {nichol2021glide}
\bibfield{author}{\bibinfo{person}{Alex Nichol}, \bibinfo{person}{Prafulla Dhariwal}, \bibinfo{person}{Aditya Ramesh}, \bibinfo{person}{Pranav Shyam}, \bibinfo{person}{Pamela Mishkin}, \bibinfo{person}{Bob McGrew}, \bibinfo{person}{Ilya Sutskever}, {and} \bibinfo{person}{Mark Chen}.} \bibinfo{year}{2021}\natexlab{}.
\newblock \showarticletitle{GLIDE: Towards Photorealistic Image Generation and Editing with Text-Guided Diffusion Models}. In \bibinfo{booktitle}{\emph{International Conference on Machine Learning}}.
\newblock
\urldef\tempurl%
\url{https://api.semanticscholar.org/CorpusID:245335086}
\showURL{%
\tempurl}


\bibitem[OpenAI(2022)]%
        {chatgpt}
\bibfield{author}{\bibinfo{person}{OpenAI}.} \bibinfo{year}{2022}\natexlab{}.
\newblock \bibinfo{title}{{ChatGPT}}.
\newblock \bibinfo{howpublished}{\url{https://chat.openai.com/}}.
\newblock
\newblock
\shownote{Accessed: 2023-10-15}.


\bibitem[Oquab et~al\mbox{.}(2023)]%
        {Oquab2023DINOv2LR}
\bibfield{author}{\bibinfo{person}{Maxime Oquab}, \bibinfo{person}{Timoth{\'e}e Darcet}, \bibinfo{person}{Th{\'e}o Moutakanni}, \bibinfo{person}{Huy~Q. Vo}, \bibinfo{person}{Marc Szafraniec}, \bibinfo{person}{Vasil Khalidov}, \bibinfo{person}{Pierre Fernandez}, \bibinfo{person}{Daniel Haziza}, \bibinfo{person}{Francisco Massa}, \bibinfo{person}{Alaaeldin El-Nouby}, \bibinfo{person}{Mahmoud Assran}, \bibinfo{person}{Nicolas Ballas}, \bibinfo{person}{Wojciech Galuba}, \bibinfo{person}{Russ Howes}, \bibinfo{person}{Po-Yao~(Bernie) Huang}, \bibinfo{person}{Shang-Wen Li}, \bibinfo{person}{Ishan Misra}, \bibinfo{person}{Michael~G. Rabbat}, \bibinfo{person}{Vasu Sharma}, \bibinfo{person}{Gabriel Synnaeve}, \bibinfo{person}{Huijiao Xu}, \bibinfo{person}{Herv{\'e} J{\'e}gou}, \bibinfo{person}{Julien Mairal}, \bibinfo{person}{Patrick Labatut}, \bibinfo{person}{Armand Joulin}, {and} \bibinfo{person}{Piotr Bojanowski}.} \bibinfo{year}{2023}\natexlab{}.
\newblock \showarticletitle{{DINOv2}: Learning Robust Visual Features without Supervision}.
\newblock \bibinfo{journal}{\emph{ArXiv}}  \bibinfo{volume}{abs/2304.07193} (\bibinfo{year}{2023}).
\newblock
\urldef\tempurl%
\url{https://api.semanticscholar.org/CorpusID:258170077}
\showURL{%
\tempurl}


\bibitem[Patashnik et~al\mbox{.}(2023)]%
        {Patashnik2023LocalizingOS}
\bibfield{author}{\bibinfo{person}{Or Patashnik}, \bibinfo{person}{Daniel Garibi}, \bibinfo{person}{Idan Azuri}, \bibinfo{person}{Hadar Averbuch-Elor}, {and} \bibinfo{person}{Daniel Cohen-Or}.} \bibinfo{year}{2023}\natexlab{}.
\newblock \showarticletitle{Localizing Object-level Shape Variations with Text-to-Image Diffusion Models}.
\newblock \bibinfo{journal}{\emph{ArXiv}}  \bibinfo{volume}{abs/2303.11306} (\bibinfo{year}{2023}).
\newblock


\bibitem[Po et~al\mbox{.}(2023)]%
        {Po2023StateOT}
\bibfield{author}{\bibinfo{person}{Ryan Po}, \bibinfo{person}{Wang Yifan}, \bibinfo{person}{Vladislav Golyanik}, \bibinfo{person}{Kfir Aberman}, \bibinfo{person}{Jonathan~T. Barron}, \bibinfo{person}{Amit~H. Bermano}, \bibinfo{person}{Eric~Ryan Chan}, \bibinfo{person}{Tali Dekel}, \bibinfo{person}{Aleksander Holynski}, \bibinfo{person}{Angjoo Kanazawa}, \bibinfo{person}{C.~Karen Liu}, \bibinfo{person}{Lingjie Liu}, \bibinfo{person}{Ben Mildenhall}, \bibinfo{person}{Matthias Nie{\ss}ner}, \bibinfo{person}{Bjorn Ommer}, \bibinfo{person}{Christian Theobalt}, \bibinfo{person}{Peter Wonka}, {and} \bibinfo{person}{Gordon Wetzstein}.} \bibinfo{year}{2023}\natexlab{}.
\newblock \showarticletitle{State of the Art on Diffusion Models for Visual Computing}.
\newblock \bibinfo{journal}{\emph{ArXiv}}  \bibinfo{volume}{abs/2310.07204} (\bibinfo{year}{2023}).
\newblock
\urldef\tempurl%
\url{https://api.semanticscholar.org/CorpusID:263835355}
\showURL{%
\tempurl}


\bibitem[Podell et~al\mbox{.}(2023)]%
        {Podell2023SDXLIL}
\bibfield{author}{\bibinfo{person}{Dustin Podell}, \bibinfo{person}{Zion English}, \bibinfo{person}{Kyle Lacey}, \bibinfo{person}{A. Blattmann}, \bibinfo{person}{Tim Dockhorn}, \bibinfo{person}{Jonas Muller}, \bibinfo{person}{Joe Penna}, {and} \bibinfo{person}{Robin Rombach}.} \bibinfo{year}{2023}\natexlab{}.
\newblock \showarticletitle{{SDXL}: Improving Latent Diffusion Models for High-Resolution Image Synthesis}.
\newblock \bibinfo{journal}{\emph{ArXiv}}  \bibinfo{volume}{abs/2307.01952} (\bibinfo{year}{2023}).
\newblock
\urldef\tempurl%
\url{https://api.semanticscholar.org/CorpusID:259341735}
\showURL{%
\tempurl}


\bibitem[Poole et~al\mbox{.}(2022)]%
        {poole2022dreamfusion}
\bibfield{author}{\bibinfo{person}{Ben Poole}, \bibinfo{person}{Ajay Jain}, \bibinfo{person}{Jonathan~T Barron}, {and} \bibinfo{person}{Ben Mildenhall}.} \bibinfo{year}{2022}\natexlab{}.
\newblock \showarticletitle{Dreamfusion: Text-to-3d using 2d diffusion}.
\newblock \bibinfo{journal}{\emph{arXiv preprint arXiv:2209.14988}} (\bibinfo{year}{2022}).
\newblock


\bibitem[Qi et~al\mbox{.}(2023)]%
        {qi2023fatezero}
\bibfield{author}{\bibinfo{person}{Chenyang Qi}, \bibinfo{person}{Xiaodong Cun}, \bibinfo{person}{Yong Zhang}, \bibinfo{person}{Chenyang Lei}, \bibinfo{person}{Xintao Wang}, \bibinfo{person}{Ying Shan}, {and} \bibinfo{person}{Qifeng Chen}.} \bibinfo{year}{2023}\natexlab{}.
\newblock \showarticletitle{Fatezero: Fusing attentions for zero-shot text-based video editing}.
\newblock \bibinfo{journal}{\emph{arXiv preprint arXiv:2303.09535}} (\bibinfo{year}{2023}).
\newblock


\bibitem[Raab et~al\mbox{.}(2023)]%
        {Raab2023SingleMD}
\bibfield{author}{\bibinfo{person}{Sigal Raab}, \bibinfo{person}{Inbal Leibovitch}, \bibinfo{person}{Guy Tevet}, \bibinfo{person}{Moab Arar}, \bibinfo{person}{Amit~H. Bermano}, {and} \bibinfo{person}{Daniel Cohen-Or}.} \bibinfo{year}{2023}\natexlab{}.
\newblock \showarticletitle{Single Motion Diffusion}.
\newblock \bibinfo{journal}{\emph{ArXiv}}  \bibinfo{volume}{abs/2302.05905} (\bibinfo{year}{2023}).
\newblock
\urldef\tempurl%
\url{https://api.semanticscholar.org/CorpusID:256827051}
\showURL{%
\tempurl}


\bibitem[Radford et~al\mbox{.}(2021)]%
        {Radford2021LearningTV}
\bibfield{author}{\bibinfo{person}{Alec Radford}, \bibinfo{person}{Jong~Wook Kim}, \bibinfo{person}{Chris Hallacy}, \bibinfo{person}{Aditya Ramesh}, \bibinfo{person}{Gabriel Goh}, \bibinfo{person}{Sandhini Agarwal}, \bibinfo{person}{Girish Sastry}, \bibinfo{person}{Amanda Askell}, \bibinfo{person}{Pamela Mishkin}, \bibinfo{person}{Jack Clark}, \bibinfo{person}{Gretchen Krueger}, {and} \bibinfo{person}{Ilya Sutskever}.} \bibinfo{year}{2021}\natexlab{}.
\newblock \showarticletitle{Learning Transferable Visual Models From Natural Language Supervision}. In \bibinfo{booktitle}{\emph{International Conference on Machine Learning}}.
\newblock


\bibitem[Rahman et~al\mbox{.}(2022)]%
        {Rahman2022MakeAStoryVM}
\bibfield{author}{\bibinfo{person}{Tanzila Rahman}, \bibinfo{person}{Hsin-Ying Lee}, \bibinfo{person}{Jian Ren}, \bibinfo{person}{S. Tulyakov}, \bibinfo{person}{Shweta Mahajan}, {and} \bibinfo{person}{Leonid Sigal}.} \bibinfo{year}{2022}\natexlab{}.
\newblock \showarticletitle{Make-A-Story: Visual Memory Conditioned Consistent Story Generation}.
\newblock \bibinfo{journal}{\emph{2023 IEEE/CVF Conference on Computer Vision and Pattern Recognition (CVPR)}} (\bibinfo{year}{2022}), \bibinfo{pages}{2493--2502}.
\newblock
\urldef\tempurl%
\url{https://api.semanticscholar.org/CorpusID:254017562}
\showURL{%
\tempurl}


\bibitem[Ramesh et~al\mbox{.}(2022)]%
        {ramesh2022hierarchical}
\bibfield{author}{\bibinfo{person}{Aditya Ramesh}, \bibinfo{person}{Prafulla Dhariwal}, \bibinfo{person}{Alex Nichol}, \bibinfo{person}{Casey Chu}, {and} \bibinfo{person}{Mark Chen}.} \bibinfo{year}{2022}\natexlab{}.
\newblock \showarticletitle{Hierarchical text-conditional image generation with {CLIP} latents}.
\newblock \bibinfo{journal}{\emph{arXiv preprint arXiv:2204.06125}} (\bibinfo{year}{2022}).
\newblock


\bibitem[Richardson et~al\mbox{.}(2023a)]%
        {richardson2023conceptlab}
\bibfield{author}{\bibinfo{person}{Elad Richardson}, \bibinfo{person}{Kfir Goldberg}, \bibinfo{person}{Yuval Alaluf}, {and} \bibinfo{person}{Daniel Cohen-Or}.} \bibinfo{year}{2023}\natexlab{a}.
\newblock \showarticletitle{ConceptLab: Creative Generation using Diffusion Prior Constraints}.
\newblock \bibinfo{journal}{\emph{arXiv preprint arXiv:2308.02669}} (\bibinfo{year}{2023}).
\newblock


\bibitem[Richardson et~al\mbox{.}(2023b)]%
        {Richardson2023TEXTureTT}
\bibfield{author}{\bibinfo{person}{Elad Richardson}, \bibinfo{person}{Gal Metzer}, \bibinfo{person}{Yuval Alaluf}, \bibinfo{person}{Raja Giryes}, {and} \bibinfo{person}{Daniel Cohen-Or}.} \bibinfo{year}{2023}\natexlab{b}.
\newblock \showarticletitle{TEXTure: Text-Guided Texturing of 3D Shapes}.
\newblock \bibinfo{journal}{\emph{ACM SIGGRAPH 2023 Conference Proceedings}} (\bibinfo{year}{2023}).
\newblock
\urldef\tempurl%
\url{https://api.semanticscholar.org/CorpusID:256597953}
\showURL{%
\tempurl}


\bibitem[Romain Beaumont(2023)]%
        {clip_retrival}
Romain Beaumont \bibinfo{year}{2023}\natexlab{}.
\newblock \bibinfo{title}{CLIP Retrival}.
\newblock \bibinfo{howpublished}{\url{https://github.com/rom1504/clip-retrieval}}.
\newblock


\bibitem[Rombach et~al\mbox{.}(2021)]%
        {Rombach2021HighResolutionIS}
\bibfield{author}{\bibinfo{person}{Robin Rombach}, \bibinfo{person}{A. Blattmann}, \bibinfo{person}{Dominik Lorenz}, \bibinfo{person}{Patrick Esser}, {and} \bibinfo{person}{Bj{\"o}rn Ommer}.} \bibinfo{year}{2021}\natexlab{}.
\newblock \showarticletitle{High-Resolution Image Synthesis with Latent Diffusion Models}.
\newblock \bibinfo{journal}{\emph{2022 IEEE/CVF Conference on Computer Vision and Pattern Recognition (CVPR)}} (\bibinfo{year}{2021}), \bibinfo{pages}{10674--10685}.
\newblock


\bibitem[Ruiz et~al\mbox{.}(2023)]%
        {Ruiz2022DreamBoothFT}
\bibfield{author}{\bibinfo{person}{Nataniel Ruiz}, \bibinfo{person}{Yuanzhen Li}, \bibinfo{person}{Varun Jampani}, \bibinfo{person}{Yael Pritch}, \bibinfo{person}{Michael Rubinstein}, {and} \bibinfo{person}{Kfir Aberman}.} \bibinfo{year}{2023}\natexlab{}.
\newblock \showarticletitle{{DreamBooth}: Fine tuning text-to-image diffusion models for subject-driven generation}. In \bibinfo{booktitle}{\emph{Proceedings of the IEEE/CVF Conference on Computer Vision and Pattern Recognition}}. \bibinfo{pages}{22500--22510}.
\newblock


\bibitem[Ryu(2022)]%
        {lora_diffusion}
\bibfield{author}{\bibinfo{person}{Simo Ryu}.} \bibinfo{year}{2022}\natexlab{}.
\newblock \bibinfo{title}{Low-rank Adaptation for Fast Text-to-Image Diffusion Fine-tuning}.
\newblock \bibinfo{howpublished}{\url{https://github.com/cloneofsimo/lora}}.
\newblock


\bibitem[Saharia et~al\mbox{.}(2022)]%
        {Saharia2022PhotorealisticTD}
\bibfield{author}{\bibinfo{person}{Chitwan Saharia}, \bibinfo{person}{William Chan}, \bibinfo{person}{Saurabh Saxena}, \bibinfo{person}{Lala Li}, \bibinfo{person}{Jay Whang}, \bibinfo{person}{Emily~L Denton}, \bibinfo{person}{Kamyar Ghasemipour}, \bibinfo{person}{Raphael Gontijo~Lopes}, \bibinfo{person}{Burcu Karagol~Ayan}, \bibinfo{person}{Tim Salimans}, {et~al\mbox{.}}} \bibinfo{year}{2022}\natexlab{}.
\newblock \showarticletitle{Photorealistic text-to-image diffusion models with deep language understanding}.
\newblock \bibinfo{journal}{\emph{Advances in Neural Information Processing Systems}}  \bibinfo{volume}{35} (\bibinfo{year}{2022}), \bibinfo{pages}{36479--36494}.
\newblock


\bibitem[Schuhmann et~al\mbox{.}(2022)]%
        {Schuhmann2022LAION5BAO}
\bibfield{author}{\bibinfo{person}{Christoph Schuhmann}, \bibinfo{person}{Romain Beaumont}, \bibinfo{person}{Richard Vencu}, \bibinfo{person}{Cade Gordon}, \bibinfo{person}{Ross Wightman}, \bibinfo{person}{Mehdi Cherti}, \bibinfo{person}{Theo Coombes}, \bibinfo{person}{Aarush Katta}, \bibinfo{person}{Clayton Mullis}, \bibinfo{person}{Mitchell Wortsman}, \bibinfo{person}{Patrick Schramowski}, \bibinfo{person}{Srivatsa Kundurthy}, \bibinfo{person}{Katherine Crowson}, \bibinfo{person}{Ludwig Schmidt}, \bibinfo{person}{Robert Kaczmarczyk}, {and} \bibinfo{person}{Jenia Jitsev}.} \bibinfo{year}{2022}\natexlab{}.
\newblock \showarticletitle{LAION-5B: An open large-scale dataset for training next generation image-text models}.
\newblock \bibinfo{journal}{\emph{ArXiv}}  \bibinfo{volume}{abs/2210.08402} (\bibinfo{year}{2022}).
\newblock
\urldef\tempurl%
\url{https://api.semanticscholar.org/CorpusID:252917726}
\showURL{%
\tempurl}


\bibitem[Sella et~al\mbox{.}(2023)]%
        {Sella2023VoxETV}
\bibfield{author}{\bibinfo{person}{Etai Sella}, \bibinfo{person}{Gal Fiebelman}, \bibinfo{person}{Peter Hedman}, {and} \bibinfo{person}{Hadar Averbuch-Elor}.} \bibinfo{year}{2023}\natexlab{}.
\newblock \showarticletitle{Vox-E: Text-guided Voxel Editing of 3D Objects}.
\newblock \bibinfo{journal}{\emph{ArXiv}}  \bibinfo{volume}{abs/2303.12048} (\bibinfo{year}{2023}).
\newblock
\urldef\tempurl%
\url{https://api.semanticscholar.org/CorpusID:257636627}
\showURL{%
\tempurl}


\bibitem[Sheynin et~al\mbox{.}(2022)]%
        {sheynin2022knn}
\bibfield{author}{\bibinfo{person}{Shelly Sheynin}, \bibinfo{person}{Oron Ashual}, \bibinfo{person}{Adam Polyak}, \bibinfo{person}{Uriel Singer}, \bibinfo{person}{Oran Gafni}, \bibinfo{person}{Eliya Nachmani}, {and} \bibinfo{person}{Yaniv Taigman}.} \bibinfo{year}{2022}\natexlab{}.
\newblock \showarticletitle{kNN-Diffusion: Image Generation via Large-Scale Retrieval}. In \bibinfo{booktitle}{\emph{The Eleventh International Conference on Learning Representations}}.
\newblock


\bibitem[Shi et~al\mbox{.}(2023)]%
        {Shi2023InstantBoothPT}
\bibfield{author}{\bibinfo{person}{Jing Shi}, \bibinfo{person}{Wei Xiong}, \bibinfo{person}{Zhe~L. Lin}, {and} \bibinfo{person}{Hyun~Joon Jung}.} \bibinfo{year}{2023}\natexlab{}.
\newblock \showarticletitle{InstantBooth: Personalized Text-to-Image Generation without Test-Time Finetuning}.
\newblock \bibinfo{journal}{\emph{ArXiv}}  \bibinfo{volume}{abs/2304.03411} (\bibinfo{year}{2023}).
\newblock


\bibitem[Sohl-Dickstein et~al\mbox{.}(2015)]%
        {sohl2015deep}
\bibfield{author}{\bibinfo{person}{Jascha Sohl-Dickstein}, \bibinfo{person}{Eric Weiss}, \bibinfo{person}{Niru Maheswaranathan}, {and} \bibinfo{person}{Surya Ganguli}.} \bibinfo{year}{2015}\natexlab{}.
\newblock \showarticletitle{Deep unsupervised learning using nonequilibrium thermodynamics}. In \bibinfo{booktitle}{\emph{International Conference on Machine Learning}}. PMLR, \bibinfo{pages}{2256--2265}.
\newblock


\bibitem[Sohn et~al\mbox{.}(2023)]%
        {Sohn2023StyleDropTG}
\bibfield{author}{\bibinfo{person}{Kihyuk Sohn}, \bibinfo{person}{Nataniel Ruiz}, \bibinfo{person}{Kimin Lee}, \bibinfo{person}{Daniel~Castro Chin}, \bibinfo{person}{Irina Blok}, \bibinfo{person}{Huiwen Chang}, \bibinfo{person}{Jarred Barber}, \bibinfo{person}{Lu Jiang}, \bibinfo{person}{Glenn Entis}, \bibinfo{person}{Yuanzhen Li}, \bibinfo{person}{Yuan Hao}, \bibinfo{person}{Irfan Essa}, \bibinfo{person}{Michael Rubinstein}, {and} \bibinfo{person}{Dilip Krishnan}.} \bibinfo{year}{2023}\natexlab{}.
\newblock \showarticletitle{StyleDrop: Text-to-Image Generation in Any Style}.
\newblock \bibinfo{journal}{\emph{ArXiv}}  \bibinfo{volume}{abs/2306.00983} (\bibinfo{year}{2023}).
\newblock
\urldef\tempurl%
\url{https://api.semanticscholar.org/CorpusID:258999204}
\showURL{%
\tempurl}


\bibitem[Song et~al\mbox{.}(2020)]%
        {song2020denoising}
\bibfield{author}{\bibinfo{person}{Jiaming Song}, \bibinfo{person}{Chenlin Meng}, {and} \bibinfo{person}{Stefano Ermon}.} \bibinfo{year}{2020}\natexlab{}.
\newblock \showarticletitle{Denoising Diffusion Implicit Models}. In \bibinfo{booktitle}{\emph{International Conference on Learning Representations}}.
\newblock


\bibitem[Song and Ermon(2019)]%
        {song2019generative}
\bibfield{author}{\bibinfo{person}{Yang Song} {and} \bibinfo{person}{Stefano Ermon}.} \bibinfo{year}{2019}\natexlab{}.
\newblock \showarticletitle{Generative modeling by estimating gradients of the data distribution}.
\newblock \bibinfo{journal}{\emph{Advances in Neural Information Processing Systems}}  \bibinfo{volume}{32} (\bibinfo{year}{2019}).
\newblock


\bibitem[stassius(2023)]%
        {celebrities_trick}
\bibfield{author}{\bibinfo{person}{stassius}.} \bibinfo{year}{2023}\natexlab{}.
\newblock \bibinfo{title}{How to create consistent character faces without training (info in the comments) : StableDiffusion}.
\newblock \bibinfo{howpublished}{\url{https://www.reddit.com/r/StableDiffusion/comments/12djxvz/how_to_create_consistent_character_faces_without/}}.
\newblock


\bibitem[Sz{\H{u}}cs and Al-Shouha(2022)]%
        {szHucs2022modular}
\bibfield{author}{\bibinfo{person}{G{\'a}bor Sz{\H{u}}cs} {and} \bibinfo{person}{Modafar Al-Shouha}.} \bibinfo{year}{2022}\natexlab{}.
\newblock \showarticletitle{Modular StoryGAN with background and theme awareness for story visualization}. In \bibinfo{booktitle}{\emph{International Conference on Pattern Recognition and Artificial Intelligence}}. Springer, \bibinfo{pages}{275--286}.
\newblock


\bibitem[Tevet et~al\mbox{.}(2022)]%
        {Tevet2022HumanMD}
\bibfield{author}{\bibinfo{person}{Guy Tevet}, \bibinfo{person}{Sigal Raab}, \bibinfo{person}{Brian Gordon}, \bibinfo{person}{Yonatan Shafir}, \bibinfo{person}{Daniel Cohen-Or}, {and} \bibinfo{person}{Amit~H. Bermano}.} \bibinfo{year}{2022}\natexlab{}.
\newblock \showarticletitle{Human Motion Diffusion Model}.
\newblock \bibinfo{journal}{\emph{ArXiv}}  \bibinfo{volume}{abs/2209.14916} (\bibinfo{year}{2022}).
\newblock
\urldef\tempurl%
\url{https://api.semanticscholar.org/CorpusID:252595883}
\showURL{%
\tempurl}


\bibitem[Tewel et~al\mbox{.}(2023)]%
        {Tewel2023KeyLockedRO}
\bibfield{author}{\bibinfo{person}{Yoad Tewel}, \bibinfo{person}{Rinon Gal}, \bibinfo{person}{Gal Chechik}, {and} \bibinfo{person}{Yuval Atzmon}.} \bibinfo{year}{2023}\natexlab{}.
\newblock \showarticletitle{Key-Locked Rank One Editing for Text-to-Image Personalization}.
\newblock \bibinfo{journal}{\emph{ACM SIGGRAPH 2023 Conference Proceedings}} (\bibinfo{year}{2023}).
\newblock
\urldef\tempurl%
\url{https://api.semanticscholar.org/CorpusID:258436985}
\showURL{%
\tempurl}


\bibitem[Tukey(1949)]%
        {Tukey1949ComparingIM}
\bibfield{author}{\bibinfo{person}{John~W. Tukey}.} \bibinfo{year}{1949}\natexlab{}.
\newblock \showarticletitle{Comparing individual means in the analysis of variance.}
\newblock \bibinfo{journal}{\emph{Biometrics}}  \bibinfo{volume}{5 2} (\bibinfo{year}{1949}), \bibinfo{pages}{99--114}.
\newblock


\bibitem[Tumanyan et~al\mbox{.}(2023)]%
        {pnpDiffusion2022}
\bibfield{author}{\bibinfo{person}{Narek Tumanyan}, \bibinfo{person}{Michal Geyer}, \bibinfo{person}{Shai Bagon}, {and} \bibinfo{person}{Tali Dekel}.} \bibinfo{year}{2023}\natexlab{}.
\newblock \showarticletitle{Plug-and-play diffusion features for text-driven image-to-image translation}. In \bibinfo{booktitle}{\emph{Proceedings of the IEEE/CVF Conference on Computer Vision and Pattern Recognition}}. \bibinfo{pages}{1921--1930}.
\newblock


\bibitem[Valevski et~al\mbox{.}(2023)]%
        {Valevski2023Face0IC}
\bibfield{author}{\bibinfo{person}{Dani Valevski}, \bibinfo{person}{Danny Lumen}, \bibinfo{person}{Yossi Matias}, {and} \bibinfo{person}{Yaniv Leviathan}.} \bibinfo{year}{2023}\natexlab{}.
\newblock \showarticletitle{Face0: Instantaneously Conditioning a Text-to-Image Model on a Face}.
\newblock \bibinfo{journal}{\emph{SIGGRAPH Asia 2023 Conference Papers}} (\bibinfo{year}{2023}).
\newblock
\urldef\tempurl%
\url{https://api.semanticscholar.org/CorpusID:259138505}
\showURL{%
\tempurl}


\bibitem[Vinker et~al\mbox{.}(2023)]%
        {Vinker2023ConceptDF}
\bibfield{author}{\bibinfo{person}{Yael Vinker}, \bibinfo{person}{Andrey Voynov}, \bibinfo{person}{Daniel Cohen-Or}, {and} \bibinfo{person}{Ariel Shamir}.} \bibinfo{year}{2023}\natexlab{}.
\newblock \showarticletitle{Concept Decomposition for Visual Exploration and Inspiration}.
\newblock \bibinfo{journal}{\emph{ArXiv}}  \bibinfo{volume}{abs/2305.18203} (\bibinfo{year}{2023}).
\newblock
\urldef\tempurl%
\url{https://api.semanticscholar.org/CorpusID:258959472}
\showURL{%
\tempurl}


\bibitem[von Platen et~al\mbox{.}(2022)]%
        {von-platen-etal-2022-diffusers}
\bibfield{author}{\bibinfo{person}{Patrick von Platen}, \bibinfo{person}{Suraj Patil}, \bibinfo{person}{Anton Lozhkov}, \bibinfo{person}{Pedro Cuenca}, \bibinfo{person}{Nathan Lambert}, \bibinfo{person}{Kashif Rasul}, \bibinfo{person}{Mishig Davaadorj}, {and} \bibinfo{person}{Thomas Wolf}.} \bibinfo{year}{2022}\natexlab{}.
\newblock \bibinfo{title}{Diffusers: State-of-the-art diffusion models}.
\newblock \bibinfo{howpublished}{\url{https://github.com/huggingface/diffusers}}.
\newblock


\bibitem[Voynov et~al\mbox{.}(2022)]%
        {voynov2022sketch}
\bibfield{author}{\bibinfo{person}{Andrey Voynov}, \bibinfo{person}{Kfir Aberman}, {and} \bibinfo{person}{Daniel Cohen-Or}.} \bibinfo{year}{2022}\natexlab{}.
\newblock \showarticletitle{Sketch-Guided Text-to-Image Diffusion Models}.
\newblock \bibinfo{journal}{\emph{arXiv preprint arXiv:2211.13752}} (\bibinfo{year}{2022}).
\newblock


\bibitem[Voynov et~al\mbox{.}(2023)]%
        {Voynov2023PET}
\bibfield{author}{\bibinfo{person}{Andrey Voynov}, \bibinfo{person}{Q. Chu}, \bibinfo{person}{Daniel Cohen-Or}, {and} \bibinfo{person}{Kfir Aberman}.} \bibinfo{year}{2023}\natexlab{}.
\newblock \showarticletitle{P+: Extended Textual Conditioning in Text-to-Image Generation}.
\newblock \bibinfo{journal}{\emph{ArXiv}}  \bibinfo{volume}{abs/2303.09522} (\bibinfo{year}{2023}).
\newblock


\bibitem[Wei(2023)]%
        {ELITE-implmentation}
\bibfield{author}{\bibinfo{person}{Yuxiang Wei}.} \bibinfo{year}{2023}\natexlab{}.
\newblock \bibinfo{title}{Official Implementation of {ELITE}}.
\newblock \bibinfo{howpublished}{\url{https://github.com/csyxwei/ELITE}}.
\newblock
\newblock
\shownote{Accessed: 2023-05-01}.


\bibitem[Wei et~al\mbox{.}(2023)]%
        {Wei2023ELITEEV}
\bibfield{author}{\bibinfo{person}{Yuxiang Wei}, \bibinfo{person}{Yabo Zhang}, \bibinfo{person}{Zhilong Ji}, \bibinfo{person}{Jinfeng Bai}, \bibinfo{person}{Lei Zhang}, {and} \bibinfo{person}{Wangmeng Zuo}.} \bibinfo{year}{2023}\natexlab{}.
\newblock \showarticletitle{{ELITE}: Encoding Visual Concepts into Textual Embeddings for Customized Text-to-Image Generation}.
\newblock \bibinfo{journal}{\emph{ArXiv}}  \bibinfo{volume}{abs/2302.13848} (\bibinfo{year}{2023}).
\newblock


\bibitem[Wolf et~al\mbox{.}(2020)]%
        {wolf-etal-2020-transformers}
\bibfield{author}{\bibinfo{person}{Thomas Wolf}, \bibinfo{person}{Lysandre Debut}, \bibinfo{person}{Victor Sanh}, \bibinfo{person}{Julien Chaumond}, \bibinfo{person}{Clement Delangue}, \bibinfo{person}{Anthony Moi}, \bibinfo{person}{Pierric Cistac}, \bibinfo{person}{Tim Rault}, \bibinfo{person}{Rémi Louf}, \bibinfo{person}{Morgan Funtowicz}, \bibinfo{person}{Joe Davison}, \bibinfo{person}{Sam Shleifer}, \bibinfo{person}{Patrick von Platen}, \bibinfo{person}{Clara Ma}, \bibinfo{person}{Yacine Jernite}, \bibinfo{person}{Julien Plu}, \bibinfo{person}{Canwen Xu}, \bibinfo{person}{Teven~Le Scao}, \bibinfo{person}{Sylvain Gugger}, \bibinfo{person}{Mariama Drame}, \bibinfo{person}{Quentin Lhoest}, {and} \bibinfo{person}{Alexander~M. Rush}.} \bibinfo{year}{2020}\natexlab{}.
\newblock \showarticletitle{Transformers: State-of-the-Art Natural Language Processing}. In \bibinfo{booktitle}{\emph{Proceedings of the 2020 Conference on Empirical Methods in Natural Language Processing: System Demonstrations}}. \bibinfo{publisher}{Association for Computational Linguistics}, \bibinfo{address}{Online}, \bibinfo{pages}{38--45}.
\newblock
\urldef\tempurl%
\url{https://www.aclweb.org/anthology/2020.emnlp-demos.6}
\showURL{%
\tempurl}


\bibitem[Yang et~al\mbox{.}(2023)]%
        {Yang2023RerenderAV}
\bibfield{author}{\bibinfo{person}{Shuai Yang}, \bibinfo{person}{Yifan Zhou}, \bibinfo{person}{Ziwei Liu}, {and} \bibinfo{person}{Chen~Change Loy}.} \bibinfo{year}{2023}\natexlab{}.
\newblock \showarticletitle{Rerender A Video: Zero-Shot Text-Guided Video-to-Video Translation}.
\newblock \bibinfo{journal}{\emph{ArXiv}}  \bibinfo{volume}{abs/2306.07954} (\bibinfo{year}{2023}).
\newblock
\urldef\tempurl%
\url{https://api.semanticscholar.org/CorpusID:259144797}
\showURL{%
\tempurl}


\bibitem[Ye et~al\mbox{.}(2023)]%
        {Ye2023IPAdapterTC}
\bibfield{author}{\bibinfo{person}{Hu Ye}, \bibinfo{person}{Jun Zhang}, \bibinfo{person}{Siyi Liu}, \bibinfo{person}{Xiao Han}, {and} \bibinfo{person}{Wei Yang}.} \bibinfo{year}{2023}\natexlab{}.
\newblock \showarticletitle{{IP-Adapter}: Text Compatible Image Prompt Adapter for Text-to-Image Diffusion Models}.
\newblock \bibinfo{journal}{\emph{ArXiv}}  \bibinfo{volume}{abs/2308.06721} (\bibinfo{year}{2023}).
\newblock
\urldef\tempurl%
\url{https://api.semanticscholar.org/CorpusID:260886966}
\showURL{%
\tempurl}


\bibitem[Yu et~al\mbox{.}(2022)]%
        {yu2022scaling}
\bibfield{author}{\bibinfo{person}{Jiahui Yu}, \bibinfo{person}{Yuanzhong Xu}, \bibinfo{person}{Jing~Yu Koh}, \bibinfo{person}{Thang Luong}, \bibinfo{person}{Gunjan Baid}, \bibinfo{person}{Zirui Wang}, \bibinfo{person}{Vijay Vasudevan}, \bibinfo{person}{Alexander Ku}, \bibinfo{person}{Yinfei Yang}, \bibinfo{person}{Burcu~Karagol Ayan}, {et~al\mbox{.}}} \bibinfo{year}{2022}\natexlab{}.
\newblock \showarticletitle{Scaling Autoregressive Models for Content-Rich Text-to-Image Generation}.
\newblock \bibinfo{journal}{\emph{arXiv preprint arXiv:2206.10789}} (\bibinfo{year}{2022}).
\newblock


\bibitem[Zhang et~al\mbox{.}(2023b)]%
        {Zhang2023TexttoimageDM}
\bibfield{author}{\bibinfo{person}{Chenshuang Zhang}, \bibinfo{person}{Chaoning Zhang}, \bibinfo{person}{Mengchun Zhang}, {and} \bibinfo{person}{In-So Kweon}.} \bibinfo{year}{2023}\natexlab{b}.
\newblock \showarticletitle{Text-to-image Diffusion Models in Generative AI: A Survey}.
\newblock \bibinfo{journal}{\emph{ArXiv}}  \bibinfo{volume}{abs/2303.07909} (\bibinfo{year}{2023}).
\newblock
\urldef\tempurl%
\url{https://api.semanticscholar.org/CorpusID:257505012}
\showURL{%
\tempurl}


\bibitem[Zhang et~al\mbox{.}(2023a)]%
        {zhang2023controlnet}
\bibfield{author}{\bibinfo{person}{Lvmin Zhang}, \bibinfo{person}{Anyi Rao}, {and} \bibinfo{person}{Maneesh Agrawala}.} \bibinfo{year}{2023}\natexlab{a}.
\newblock \showarticletitle{Adding Conditional Control to Text-to-Image Diffusion Models}. In \bibinfo{booktitle}{\emph{Proceedings of the IEEE/CVF International Conference on Computer Vision (ICCV)}}. \bibinfo{pages}{3836--3847}.
\newblock


\bibitem[Zhuang et~al\mbox{.}(2023)]%
        {Zhuang2023DreamEditorT3}
\bibfield{author}{\bibinfo{person}{Jingyu Zhuang}, \bibinfo{person}{Chen Wang}, \bibinfo{person}{Lingjie Liu}, \bibinfo{person}{Liang Lin}, {and} \bibinfo{person}{Guanbin Li}.} \bibinfo{year}{2023}\natexlab{}.
\newblock \showarticletitle{DreamEditor: Text-Driven 3D Scene Editing with Neural Fields}.
\newblock \bibinfo{journal}{\emph{ArXiv}}  \bibinfo{volume}{abs/2306.13455} (\bibinfo{year}{2023}).
\newblock
\urldef\tempurl%
\url{https://api.semanticscholar.org/CorpusID:259243782}
\showURL{%
\tempurl}


\end{thebibliography}
